\documentclass[10pt,twocolumn,letterpaper]{article}
\usepackage[pagenumbers]{cvpr}
\usepackage{verbatim}
\usepackage{cuted}
\usepackage{lipsum}
\usepackage[accsupp]{axessibility}

\usepackage[dvipsnames]{xcolor}
\usepackage[normalem]{ulem} 

\newcommand{\point}{\mathbf{P}}
\newcommand{\mesh}{\mathbf{M}}
\newcommand{\texture}{\mathbf{T}}
\newcommand{\textureStageOne}{\texture_\text{basic}} 
\newcommand{\textureStageTwo}{\texture_\text{cur}}
\newcommand{\textureStageFinal}{\texture_\text{final}}

\newcommand{\network}{D}

\newcommand{\imageSate}{I_{\text{sat}}}
\newcommand{\image}{I}
\newcommand{\renderFunc}{\mathcal{R}}
\newcommand{\camera}{\mathbf{C}}

\usepackage{setspace}
\usepackage{array}
\usepackage{multirow}
\usepackage{graphicx}
\usepackage{booktabs}
\usepackage{pgffor}
\usepackage{adjustbox}
\usepackage{xifthen}
\usepackage{tikz}
\usepackage{pgfplots}
\usetikzlibrary{arrows.meta}
\pgfplotsset{compat=1.18}

\usepackage{adjustbox}
\newcommand{\imagecell}[2][0.12]{%
    \adjustbox{width=#1\columnwidth, valign=c}{%
        \includegraphics[width=\columnwidth]{#2}%
    }%
}

\newcommand{\CroppedImageWithStarJAX}[4]{%
    \pgfmathsetlengthmacro{\targetW}{#2}%
    \pgfmathsetlengthmacro{\targetH}{0.75*#2}%
    \begin{tikzpicture}[baseline=(current bounding box.center)]
        \clip (0, 0) rectangle (\targetW, \targetH);
        \node[anchor=center, inner sep=0] at (0.5*\targetW, 0.5*\targetH) {%
            \includegraphics[width=\targetW]{#1}%
        };
        \node[text=red, inner sep=0pt, anchor=center] (starNode) at (#3*\targetW, #4*\targetH) {%
            \scalebox{0.5}{$\star$}%
        };
        \draw[-{Stealth[scale=0.6]}, red, very thin] (starNode.center) -- ++(0:0.15*\targetW);
    \end{tikzpicture}%
}

\newcommand{\CroppedImageWithStarNYC}[4]{%
    \pgfmathsetlengthmacro{\targetW}{#2}%
    \pgfmathsetlengthmacro{\targetH}{0.75*#2}%
    \begin{tikzpicture}[baseline=(current bounding box.center)]
        \clip (0, 0) rectangle (\targetW, \targetH);
        \node[anchor=center, inner sep=0] at (0.5*\targetW, 0.5*\targetH) {%
            \includegraphics[width=\targetW]{#1}%
        };
        \node[text=red, inner sep=0pt, anchor=center] (starNode) at (#3*\targetW, #4*\targetH) {%
            \scalebox{0.5}{$\star$}%
        };
        \draw[-{Stealth[scale=0.6]}, red, very thin, >=stealth] (starNode.center) -- ++(94:0.15*\targetW);
    \end{tikzpicture}%
}

\newcommand{\CroppedImageWithStarMC}[4]{%
    \pgfmathsetlengthmacro{\targetW}{#2}%
    \pgfmathsetlengthmacro{\targetH}{0.75*#2}%
    \begin{tikzpicture}[baseline=(current bounding box.center)]
        \clip (0, 0) rectangle (\targetW, \targetH);
        \node[anchor=center, inner sep=0] at (0.5*\targetW, 0.5*\targetH) {%
            \includegraphics[height=\targetH]{#1}%
        };
        \node[text=red, inner sep=0pt, anchor=center] (starNode) at (#3*\targetW, #4*\targetH) {%
            \scalebox{0.5}{$\star$}%
        };
        \draw[-{Stealth[scale=0.6]}, red, very thin] (starNode.center) -- ++(53:0.15*\targetW);
    \end{tikzpicture}%
}

\newcommand{\CroppedImageWithStarWJ}[4]{%
    \pgfmathsetlengthmacro{\targetW}{#2}%
    \pgfmathsetlengthmacro{\targetH}{0.75*#2}%
    \begin{tikzpicture}[baseline=(current bounding box.center)]
        \clip (0, 0) rectangle (\targetW, \targetH);
        \node[anchor=center, inner sep=0] at (0.5*\targetW, 0.5*\targetH) {%
            \includegraphics[height=\targetH]{#1}%
        };
        \node[text=red, inner sep=0pt, anchor=center] (starNode) at (#3*\targetW, #4*\targetH) {%
            \scalebox{0.5}{$\star$}%
        };
        \draw[-{Stealth[scale=0.6]}, red, very thin] (starNode.center) -- ++(115:0.15*\targetW);
    \end{tikzpicture}%
}

\definecolor{cvprblue}{rgb}{0.21,0.49,0.74}
\usepackage[pagebackref,breaklinks,colorlinks,allcolors=cvprblue]{hyperref}

\usepackage{colortbl}
\definecolor{best1}{RGB}{255,152,152}
\definecolor{best2}{RGB}{255,203,152}
\definecolor{best3}{RGB}{255,247,173}
\definecolor{orbit}{RGB}{68,147,244}
\definecolor{ground}{RGB}{239,191,70}

\def\paperID{16823} 
\def\confName{CVPR}
\def\confYear{2026}

\title{From \includegraphics[height=0.6cm]{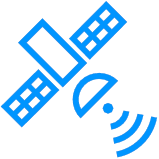}\ \textcolor{orbit}{Orbit} to \textcolor{ground}{Ground}: Generative City Photogrammetry  \\ from Extreme Off-Nadir Satellite Images}

\author{
    Fei Yu\textsuperscript{1,2,4*}, 
    Yu Liu\textsuperscript{2*}, 
    Luyang Tang\textsuperscript{2}, 
    Mingchao Sun\textsuperscript{2}, 
    Zengye Ge\textsuperscript{2}, 
    Rui Bu\textsuperscript{3}, 
    Yuchao Jin\textsuperscript{1}, 
    \\
    Haisen Zhao\textsuperscript{4}, 
    He Sun\textsuperscript{1}, 
    Yangyan Li\textsuperscript{3}, 
    Mu Xu\textsuperscript{2\dag}, 
    Wenzheng Chen\textsuperscript{1,5\dag}, 
    Baoquan Chen\textsuperscript{1\dag}
    \\
    \textsuperscript{1}Peking University \quad
    \textsuperscript{2}AMAP \quad
    \textsuperscript{3}Ant Group
    \\
    \textsuperscript{4}Shandong University \quad
    \textsuperscript{5}Beijing Academy of Artificial Intelligence
}

\begin{document}

\twocolumn[{%
    \renewcommand\twocolumn[1][]{#1}%
    \maketitle  
    \begin{center}
        \vspace{-15pt}
        \centering
        \includegraphics[width=\textwidth]
        {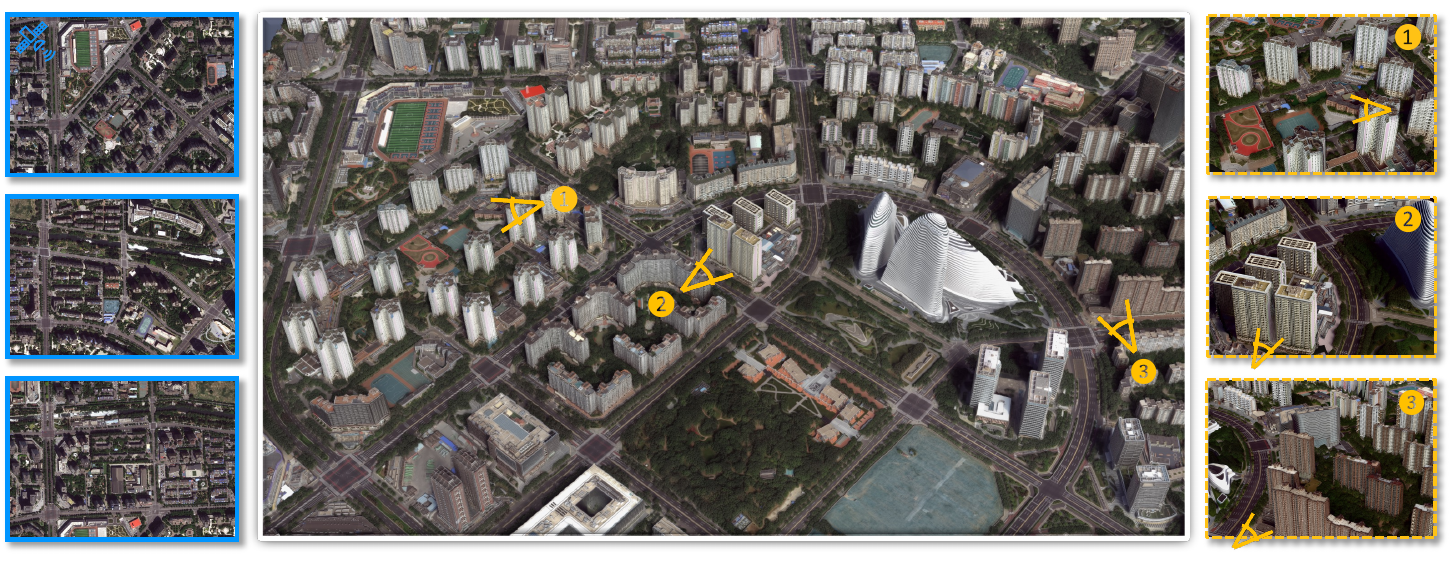}
        \vspace*{-0.5cm}
        \captionof{figure}{\textbf{City-Scale 3D Reconstruction from Satellite Imagery.}
        We reconstruct a $4\,\mathrm{km}^2$ real-world urban region from 11 sparse-view satellite images \emph{captured from orbit} that contain extremely limited parallax. The resulting 3D model, featuring crisp geometry and photorealistic appearance, enables \textbf{extreme viewpoint extrapolation}, supporting high-fidelity, close-range rendering from ground-level viewpoints. Please zoom in for details. 
       }
    \label{fig:teaser}
    \end{center}
 }]

\renewcommand{\thefootnote}{*}\footnotetext{Equal contribution.}
\renewcommand{\thefootnote}{\dag}\footnotetext{Corresponding authors.}

\begin{abstract}

\vspace*{-10pt}

City-scale 3D reconstruction from satellite imagery presents the challenge of \textbf{extreme viewpoint extrapolation}, where our goal is to synthesize ground-level novel views from sparse orbital images with minimal parallax. This requires inferring nearly $90^\circ$ viewpoint gaps from image sources with severely foreshortened facades and flawed textures, causing state-of-the-art reconstruction engines such as NeRF and 3DGS to fail.

To address this problem, we propose two design choices tailored for city structures and satellite inputs.
First, we model city geometry as a 2.5D height map, implemented as a Z-Monotonic signed distance field (SDF) that matches urban building layouts from top-down viewpoints.
This stabilizes geometry optimization under sparse, off-nadir satellite views and yields a watertight mesh with crisp roofs and clean, vertically extruded facades.
Second, we paint the mesh appearance from satellite images via differentiable rendering techniques. 
While the satellite inputs may contain long-range, blurry captures, we further train a generative texture restoration network to enhance the appearance, recovering high-frequency, plausible texture details from degraded inputs.

Our method's scalability and robustness are demonstrated through extensive experiments on large-scale urban reconstruction.
For example in \cref{fig:teaser}, we reconstruct a $4\,\mathrm{km}^2$ real-world region from only a few satellite images, achieving state-of-the-art performance in synthesizing photorealistic ground views. 
The resulting models are not only visually compelling but also serve as high-fidelity, application-ready assets for downstream tasks like urban planning and simulation.
Project page can be found at \url{https://pku-vcl-geometry.github.io/Orbit2Ground/}.

\vspace{-2pt}

\end{abstract}
    
\section{Introduction}
\label{sec:intro}

The past decade has witnessed remarkable advances in 3D reconstruction. 
Neural representations, particularly Neural Radiance Fields (NeRFs) and 3D Gaussian Splatting (3DGS)~\cite{mildenhall2021nerf,kerbl20233d}, have brought photorealistic novel view synthesis (NVS) closer to reality than ever before. 
While these methods have fundamentally revolutionized object- and street-level modeling~\cite{lin2024vastgaussian,liu2024citygaussian,liu2024citygaussianv2,gao2025citygs}, scaling this fidelity to expansive urban environments remains a critical challenge. 
Reconstructing a large city faces a fundamental data acquisition problem: ground-level capture—whether crowdsourced photographs or cameras on vehicles or drones—is logistically complex and prohibitively expensive for continuous, citywide coverage. 
In contrast, satellite imagery provides massive, readily available city coverage, positioning it as an economical and scalable data source for city-scale 3D modeling.

\begin{figure}
    \centering
    \includegraphics[width=0.95\linewidth]{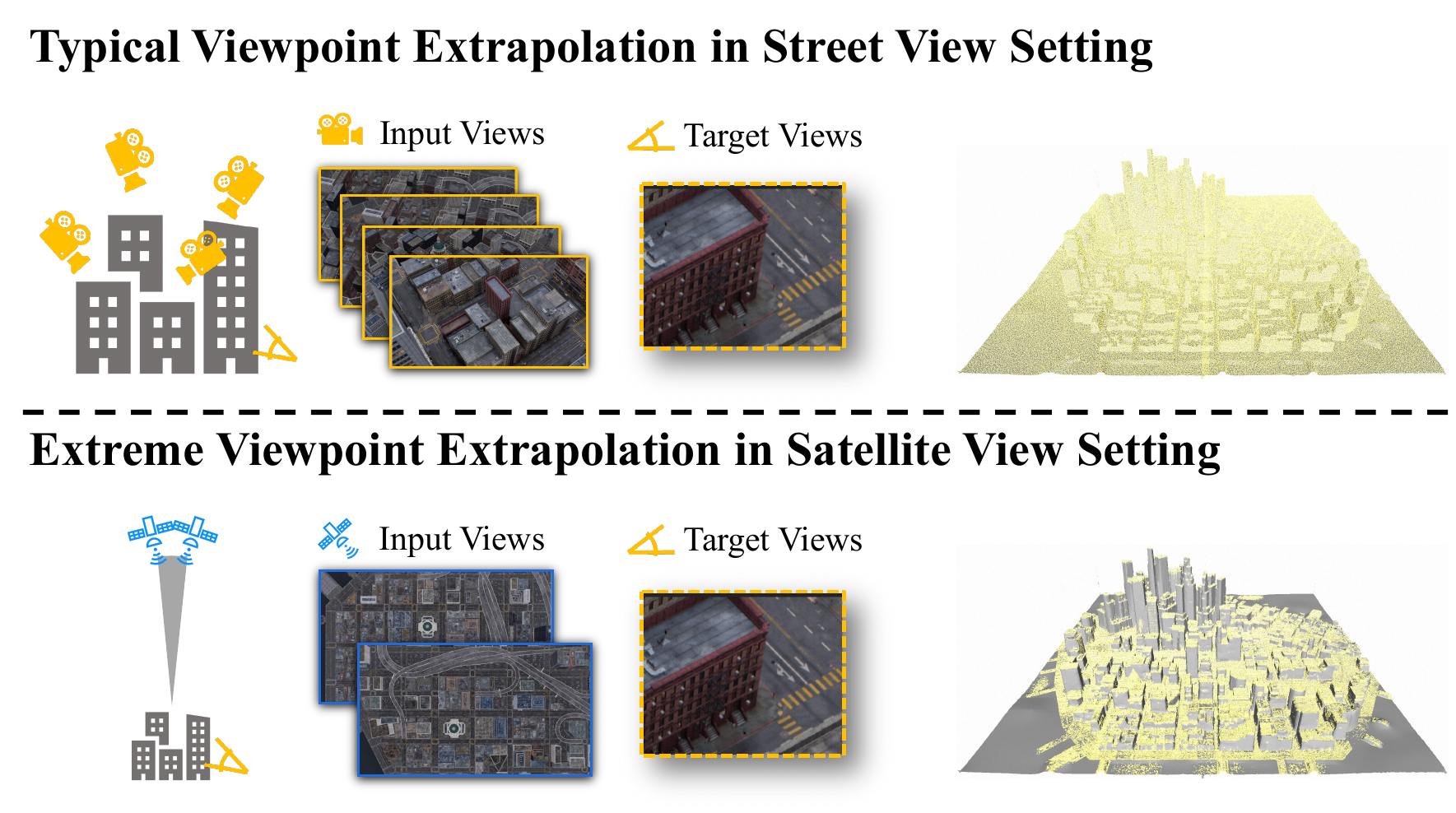}
    \caption{Unlike dense street views, satellite images are sparse and captured with extreme off-nadir angles. This leads to a severe deficiency in parallax for vertical structures. 
    Yellow points represent 3D locations determined by MVS, whereas satellite images only recover ground and roof surfaces.}
    \label{fig:intro-why-sat-hard}
    \vspace*{-10pt}
\end{figure}

However, reconstructing a high-fidelity city model from satellite imagery presents the \textbf{extreme viewpoint extrapolation} challenge. The satellite imagery is captured from top-down views with extreme off-nadir angles and minimal parallax,  while our goal is to synthesize ground-level novel views, resulting in an almost $90^\circ$ viewpoint gap between source and target. In satellite capture, as illustrated in \cref{fig:intro-why-sat-hard}, building facades suffer severe foreshortening, and vertical structures lack the viewpoint diversity needed to resolve geometry. Long-range atmospheric distortion and sensor limits further degrade textures. These combined conditions—sparse, minimal-parallax, and blurred inputs—fundamentally violate the dense-parallax and sharp-photometry assumptions required by NeRF and 3DGS, causing these methods to collapse.

To overcome this challenge, we adopt a two-stage design tailored for city photogrammetry from satellite inputs. In the first stage, our focus is on \textbf{geometry fidelity}: we leverage the inherent 2.5D structure of urban layouts from top-down views and propose a Z-Monotonic signed distance field (SDF) representation. %
This strong structural prior enables 2.5D mesh extraction via differentiable iso-surfacing~\cite{shen2023flexible} and yields coherent, watertight geometry with precise roofs and vertically extruded facades. Optimizing this SDF against off-nadir inputs preserves as much satellite-derived geometric information as possible, ensuring that even under extreme viewpoint shifts, the reconstructed structure remains faithful.

On this robust geometric foundation, the second stage addresses \textbf{appearance fidelity}. Naive back-projection texturing from satellite views produces blurred and distorted facades (\cref{fig:appearance-comparison}). To overcome these artifacts, we adapt the powerful generative prior of the FLUX foundation model~\cite{flux2024} into a high-fidelity restoration network. Novel close-range renders—degraded by projection gaps—are fed into this network, which synthesizes photorealistic appearances that drive the final texture optimization.

Our key contribution is a holistic approach that decouples this ill-posed problem into two complementary stages: robust geometry regularization and generative appearance refinement. By first establishing a stable geometric scaffold and then ``painting'' it with plausible, high-frequency details, our method effectively bridges the satellite-to-ground viewpoint gap where prior methods fail. We demonstrate through extensive experiments that this strategy yields state-of-the-art (SOTA) results, producing scalable and high-fidelity city photogrammetry from sparse satellite imagery.

\section{Related Work}
\label{sec:related-work}

\subsection{Large-Scale 3D Scene Reconstruction}

Large-scale scene reconstruction, such as entire cities, has been a key topic in computer vision and graphics.
Pioneering works \cite{schonberger2016structure, barnes2009patchmatch, furukawa2009accurate, schonberger2016pixelwise} based on Structure-from-Motion (SfM) and Multi-View Stereo (MVS) enabled city-scale reconstruction from massive, unordered photo collections. 
Recently, neural scene representations, exemplified by NeRF \cite{mildenhall2021nerf} and 3DGS \cite{kerbl20233d}, have been extended to large-scale scenes, achieving photorealistic city-scale reconstruction \cite{liu2024citygaussianv2, gao2025citygs}.
Block-NeRF \cite{tancik2022block}, Mega-NeRF \cite{turki2022mega} and Switch-NeRF \cite{zhenxing2022switch} divide the scene into blocks, each modeled by a separate NeRF.
Methods \cite{rematas2022urban, turki2023suds} incorporate multi-modal data such as LiDAR and 2D optical flow to improve the reconstruction quality.
Subsequent works \cite{xu2023grid, zhang2025efficient, song2024city} combine explicit feature grids with small MLPs to improve efficiency.
VastGaussian \cite{lin2024vastgaussian} first demonstrated the viability of 3DGS, tackling challenges like appearance variation across large areas. 
Subsequent efforts \cite{kerbl2024hierarchical, ren2024octree, liu2024citygaussian} introduce Level-of-Detail (LoD) techniques for efficient multi-scale visualization. 
Other works \cite{chen2024dogs, zhao2024scaling, feng2025flashgs} have developed distributed optimization frameworks to accelerate the training process and manage heavy memory overhead.

The aforementioned methods adopt the ``divide-and-conquer'' strategy, where each block requires sufficient parallax information from massive image collections to achieve high-quality reconstruction.
While effective, their dependency on dense, low-altitude aerial images makes data acquisition costly and complex, impeding scalable and economical city-scale 3D modeling.
In contrast, satellite imagery, with its inherent advantage of vast coverage, presents an ideal data source for this task. 
However, such images provide little parallax on vertical structures like building facades, which makes the direct application of existing parallax-dependent methods challenging, thus requiring more suitable representations.

\subsection{3D Reconstruction from Satellite Images}
Reconstructing urban scenes from satellite imagery is a long-standing goal in remote sensing and photogrammetry. 
Approaches in the traditional remote sensing field, while proficient with satellite data, pursue a distinct objective: generating metrically accurate 2.5D Digital Surface Models (DSMs) \cite{zhao2023review}, i.e., height maps.
Classic photogrammetric pipelines \cite{qin2016rpc, de2014automatic, zhang2019leveraging} and modern MVS systems, including recent deep learning-based methods \cite{gao2021rational, gao2023general, chen2024surface, yang2025learning}, are all optimized for this purpose. 
These works are fundamentally geared toward geospatial analysis, treating the output as a 2.5D grid of elevation values rather than a true 3D mesh, thus not directly suited for applications requiring visual realism or fine-grained geometric fidelity.
Recently, many works have started to improve the visual quality of satellite-based 3D reconstruction by adopting neural rendering techniques. Early efforts \cite{derksen2021shadow, mari2022sat, mari2023multi, zhang2023sparsesat, zhang2024satensorf} adapted NeRF to the satellite domain, while recent approaches \cite{aira2025gaussian, bai2025satgs, huang2025skysplat, lee2025skyfall} leverage 3DGS for its efficiency and real-time rendering.
However, as these methods supervise geometry solely through a photometric loss, they face challenges in resolving ambiguities from sparse, top-down satellite images. This leads to geometric inaccuracies that ultimately cap the achievable visual quality.

Unlike the 2.5D DSMs common in remote sensing, we introduce a fully differentiable 2.5D representation based on a Z-Monotonic SDF, which allows for fine-grained mesh extraction.
This robust geometric representation serves as a high-fidelity foundation for subsequent appearance optimization, enabling more accurate and visually realistic 3D reconstructions from satellite images.

\subsection{Generative Priors for 3D Reconstruction}
Generative priors \cite{saharia2022photorealistic, rombach2022high, peebles2023scalable, esser2024scaling}, learned from vast natural images, have unlocked a new capability for 3D reconstruction, especially in ill-posed settings like sparse-view or single-view tasks.
Seminal works such as DreamFusion \cite{poole2022dreamfusion} and Magic3D \cite{lin2023magic3d} pioneered this direction by introducing Score Distillation Sampling (SDS), which ``distills'' the knowledge from a 2D generative model into a 3D neural field. 
Subsequent works demonstrate that a powerful image-conditioned generative prior can be leveraged to infer a complete 3D object \cite{liu2023zero, long2024wonder3d, melas2023realfusion}, or even an entire scene \cite{yu2025wonderworld, team2025hunyuanworld, hua2025sat2city}. 
However, these powerful generative priors inherently favor plausible extrapolation over strict fidelity to the input, making them a double-edged sword for high-fidelity reconstruction tasks.

An alternative paradigm, which our work subscribes to, shifts the focus from generation to the refinement of existing 3D appearance \cite{wu2025difix3d+, liu20243dgs, fischer2025flowr, yin2025gsfixer}. 
A key challenge in this approach is the inherent stochasticity of generative models, which may produce conflicting details for the same scene. To address this, recent work Skyfall-GS \cite{lee2025skyfall} adopts a ``generate-and-average'' strategy by optimizing over an ensemble of candidates, which is effective but computationally expensive. 
In contrast, our work introduces a deterministic image restoration process. 
By fine-tuning a diffusion model to learn a direct mapping from degraded renders to sharp targets, we generate high-quality, holistically-consistent supervision in a single pass.
\section{Method}
\label{sec:method}

\begin{figure*}
	\centering
	\includegraphics[trim={0 25 0 0}, clip, width=1\linewidth]{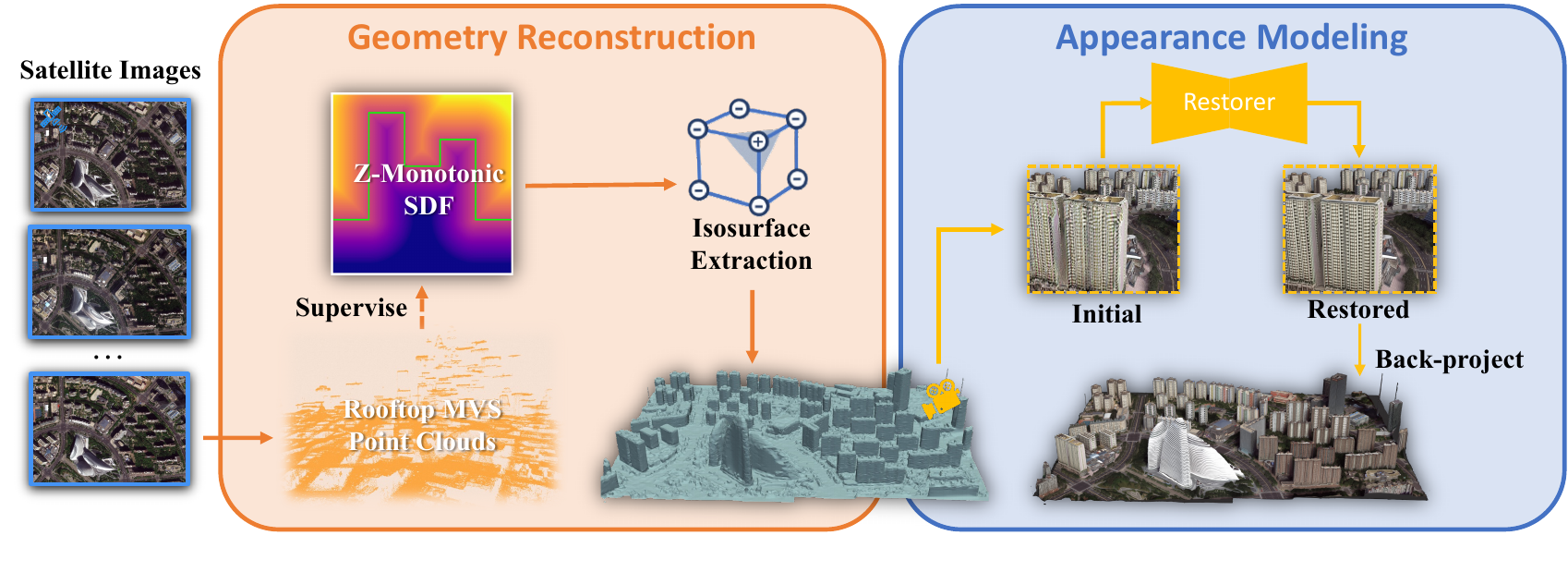}
	\caption{
		\label{fig:pipeline}
        \textbf{The framework of our method. Our pipeline first reconstructs city geometry, then refines its appearance.}
		\textbf{Stage 1 (Geometry):} We optimize a Z-Monotonic SDF against sparse MVS points to extract a high-fidelity, watertight mesh with clean vertical facades.
		\textbf{Stage 2 (Appearance):} Starting with an initial texture  
        (back-projected from source images), we use a restoration network to enhance close-range novel-view renderings, which further serve as sharp, high-fidelity supervision for final texture optimization. 
	}
	\vspace*{-10pt}
\end{figure*}

We now describe our method. As illustrated in \cref{fig:pipeline}, our method reconstructs a city in two main stages, each designed to address a core challenge of satellite-based reconstruction.
First, to overcome the extreme viewpoint gap inherent to satellite imagery, we introduce a strong geometric prior by representing the city as a Z-Monotonic SDF. This stabilizes optimization from sparse, top-down views and yields a high-fidelity, watertight mesh (\cref{sec:geometry}).
Second, to resolve the blur and visual artifacts, we leverage a generative prior by training a large-scale diffusion model to map degraded inputs to sharp, plausible textures (\cref{sec:appearance}).
Lastly, we provide implementation details in \cref{sub-sec:method:impl}.

\subsection{2.5D Geometry Representation}
\label{sec:geometry}

The primary challenge of satellite imagery for 3D reconstruction lies in its extreme viewpoint constraints. As shown in \cref{fig:intro-why-sat-hard}, standard MVS algorithms struggle to reconstruct vertical facades, yielding point clouds that are dense on rooftops and ground but virtually empty elsewhere. This massive data void on building facades creates a severely ill-posed condition for unconstrained 3D representations like NeRF or 3DGS. As a result, the optimizer is free to produce noisy, collapsed, or ``shrink-wrapped'' geometry that completely fails to capture the true city structure.

Our key insight is to regularize this ill-posed problem by introducing a strong, urban-specific structural prior: modeling the city as a 2.5D height map. This representation aligns perfectly with both the top-down nature of satellite data and the predominantly vertical geometry of urban architecture. While this design strategically sacrifices the ability to model non-monotonic structures (e.g., bridges), it provides a decisive gain in robustness against geometric ambiguity. For most urban scenes, this proves to be a highly effective trade-off, enabling our method to reconstruct high-fidelity geometry where unconstrained approaches inevitably fail.

\begin{figure}
    \centering
    \begin{spacing}{1}
    \setlength\tabcolsep{0pt}
    \begin{tabular}{ccc}
    \imagecell[0.32]{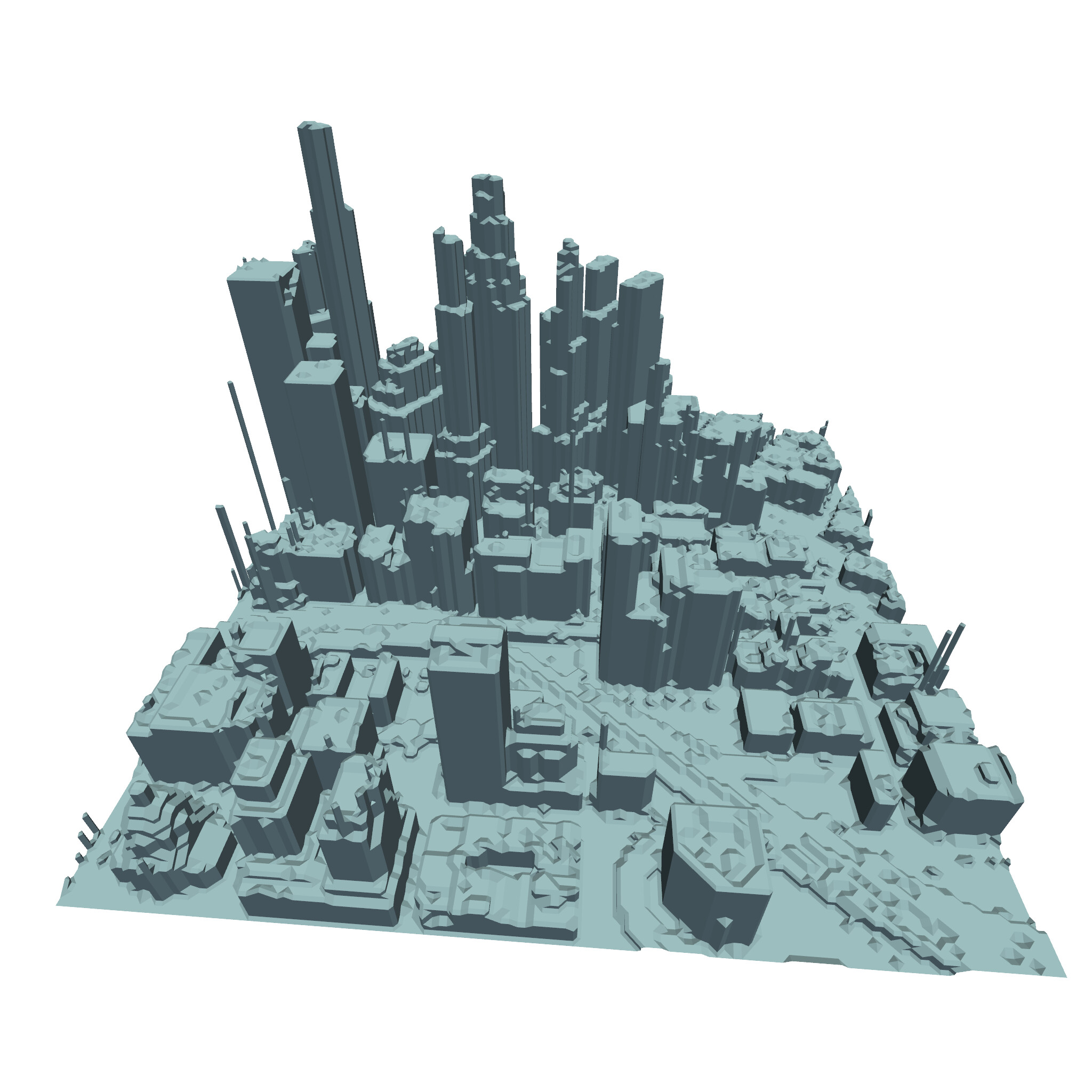} & 
    \imagecell[0.32]{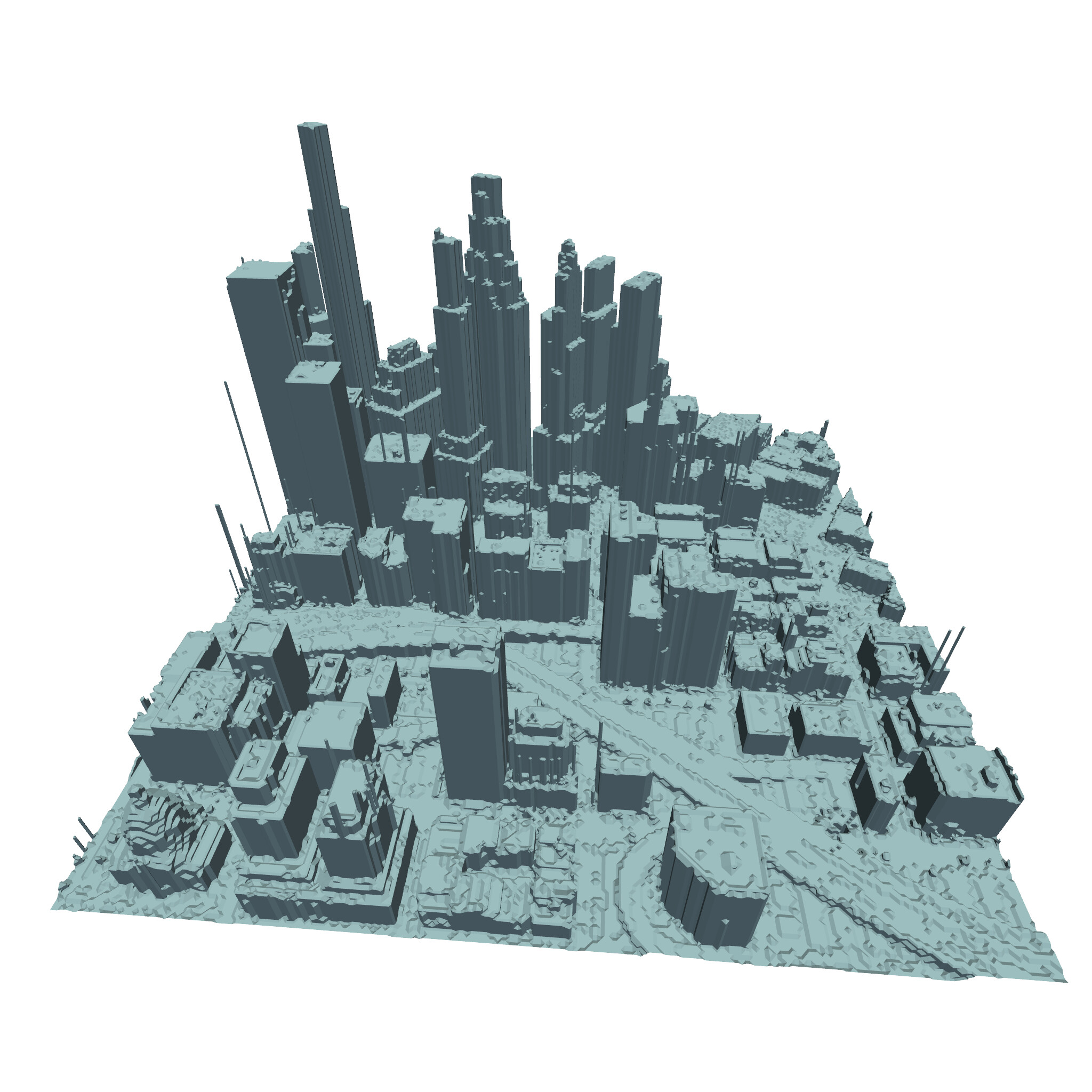} & 
    \imagecell[0.32]{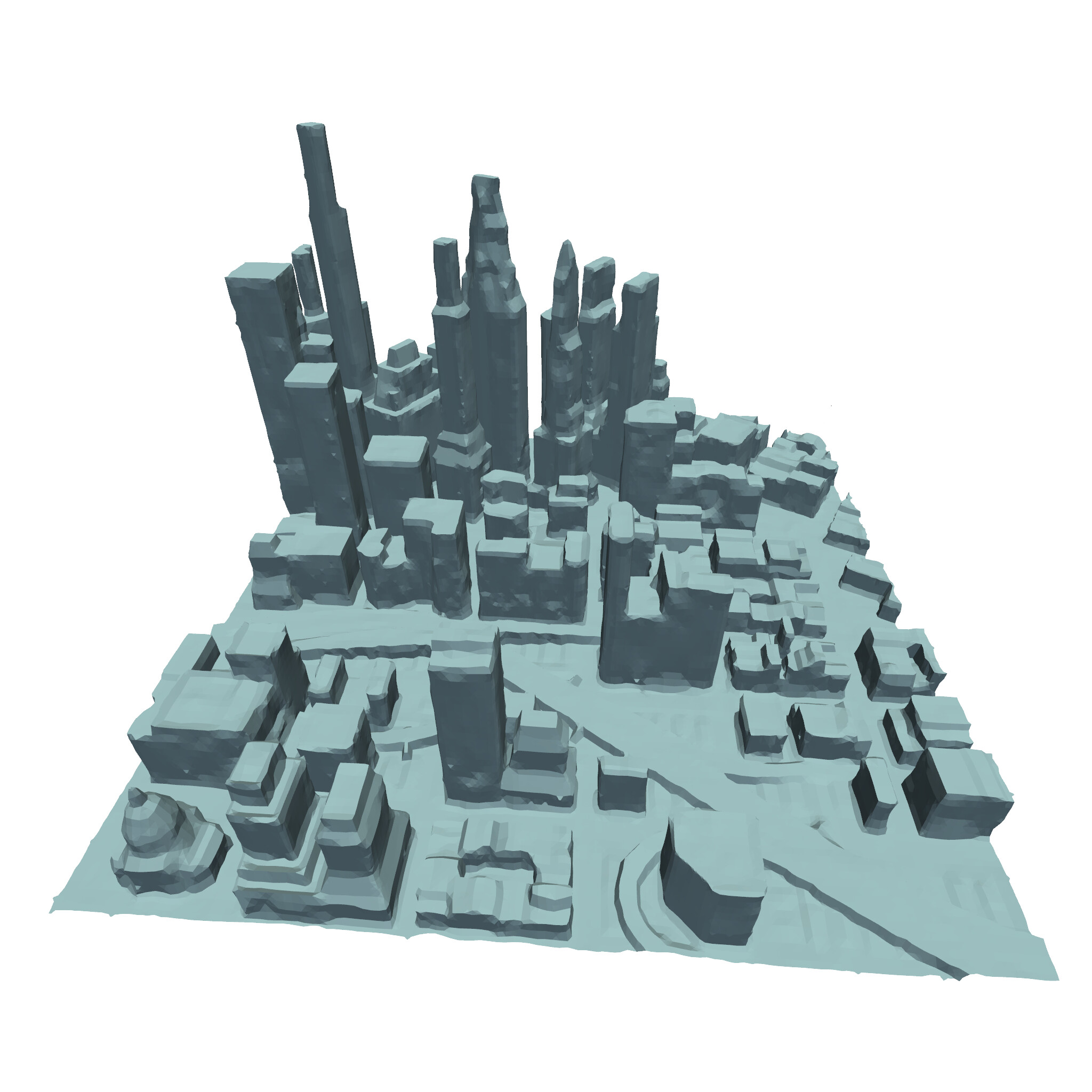} \\
    \vspace{-9pt} \\
    (a) { $\begin{matrix}
        \text{\small Marching Cubes} \\
        \text{\small Low Res. (128)}
    \end{matrix}$} & (b) $\begin{matrix}
        \text{\small Marching Cubes} \\
        \text{\small High Res. (256)}
    \end{matrix}$ & (c) $\begin{matrix}
        \text{\small Z-Monotonic} \\
        \text{\small SDF}
    \end{matrix}$
    \end{tabular}
    \end{spacing}
    \vspace{-7pt}
    \caption{\textbf{Z-Monotonic SDF vs. Naive Conversion.} 
        (a, b) A naive 2.5D mesh, generated by directly converting sparse MVS points into a voxel grid, suffers from severe ``stair-step'' artifacts and topological holes.
        (c) Our Z-Monotonic SDF representation optimizes a continuous field, resulting in a clean, watertight mesh with precise roofs and sharp vertical facades.
    }
    \label{fig:method-roof-morethan-facade}
    \vspace{-10pt}
\end{figure}

\vspace{-10pt}
\paragraph{Z-Monotonic SDF.}
\label{sec:z-sdf}
A 2.5D height map (also called DSM in remote sensing literature) defines the 3D surface as a single-valued height function $z = f(x, y)$ over the 2D ground plane.
A naive way to generate this 2.5D height map would be preparing a set of city point cloud $\point$ from the satellite images by applying MVS algorithms \cite{furukawa2009accurate}.
This point cloud $\point$ can be then directly converted  into a 2.5D height map, followed by filling a 3D voxel grid and extracting a surface via Marching Cubes \cite{lorensen1998marching}.
However, this naive conversion introduces significant aliasing or ``stair-step'' artifacts, primarily due to the noisy and sparse point cloud input, as illustrated in \cref{fig:method-roof-morethan-facade}.

To overcome this, we design a novel 2.5D representation using a \textbf{Z-Monotonic SDF}, avoiding aliasing and topological holes at building edges.
Specifically, we optimize the SDF field $s(\mathbf{x})$ with a special Z-constraint, {enforcing} its values to be non-decreasing along the vertical Z-axis:
\begin{equation}
    \frac{\partial s(x, y, z)}{\partial z} \ge 0 \quad \forall \mathbf{x} = (x, y, z).
    \label{eq:Z-Monotonic}
\end{equation}
With this constraint, our geometric representation supports continuous surface extraction for both \textbf{continuous surfaces} (e.g., roofs and ground) and \textbf{discontinuous surfaces} (e.g., facades).
At a continuous surface, the equation $s(x, y, z)=0$ has a {unique solution} $z$, defining the height.
Alternatively, at a building edge, the function becomes a vertical plateau where $s(x, y, z)=0$ for all $z \in [z_{\text{ground}}, z_{\text{roof}}]$, procedurally defining a perfect vertical facade.
Thus, the 0-level-set $\{ \mathbf{p} \mid s(\mathbf{p}) = 0 \}$ implicitly defines this 2.5D surface,
transforming the ill-posed facade reconstruction problem into a well-constrained optimization.

We implement the Z-Monotonic SDF via learning monotonic curves on each 2D X-Y plane grid. The SDF value $s(x, y, z)$ can be naturally interpolated. During optimization, we extract a mesh $M$ from the SDF using a differentiable iso-surfacing technique \cite{shen2023flexible}. We then optimize the SDF (parametrized by monotonic curves) by minimizing the Z-axis distance from the MVS points $P$ to the extracted mesh $M$, together with a Laplacian regularization term and a normal consistency term:
\begin{equation}
    \mathcal{L}_{\text{geo}} = \sum_{\mathbf{p} \in P} \min_{\mathbf{m} \in M} \|\mathbf{p}_z - \mathbf{m}^*(\mathbf{p})_z\|_1 + \lambda_{\text{Lap}} \mathcal{L}_{\text{Lap}} + \lambda_{\text{Nrm}} \mathcal{L}_{\text{Nrm}},
    \label{eq:chamfer_loss}
\end{equation}
where $\mathbf{m}^*(\mathbf{p}) := \underset{\mathbf{m} \in M}{\operatorname{argmin}} \|\mathbf{p}_{xy} - \mathbf{m}_{xy}\|_2$.
This optimization process ensures a clean, watertight mesh with accurate vertical facades, as shown in \cref{fig:method-roof-morethan-facade} (c). More details are provided in Appendices~\ref{sub-sec:supp:pc} and \ref{sub-sec:supp:zsdf}.

\subsection{High-Fidelity Appearance Modeling}
\label{sec:appearance}

\begin{figure}
    \centering
    \begin{spacing}{1}
    \setlength\tabcolsep{2pt}
    \begin{tabular}{cc}
    \imagecell[0.48]{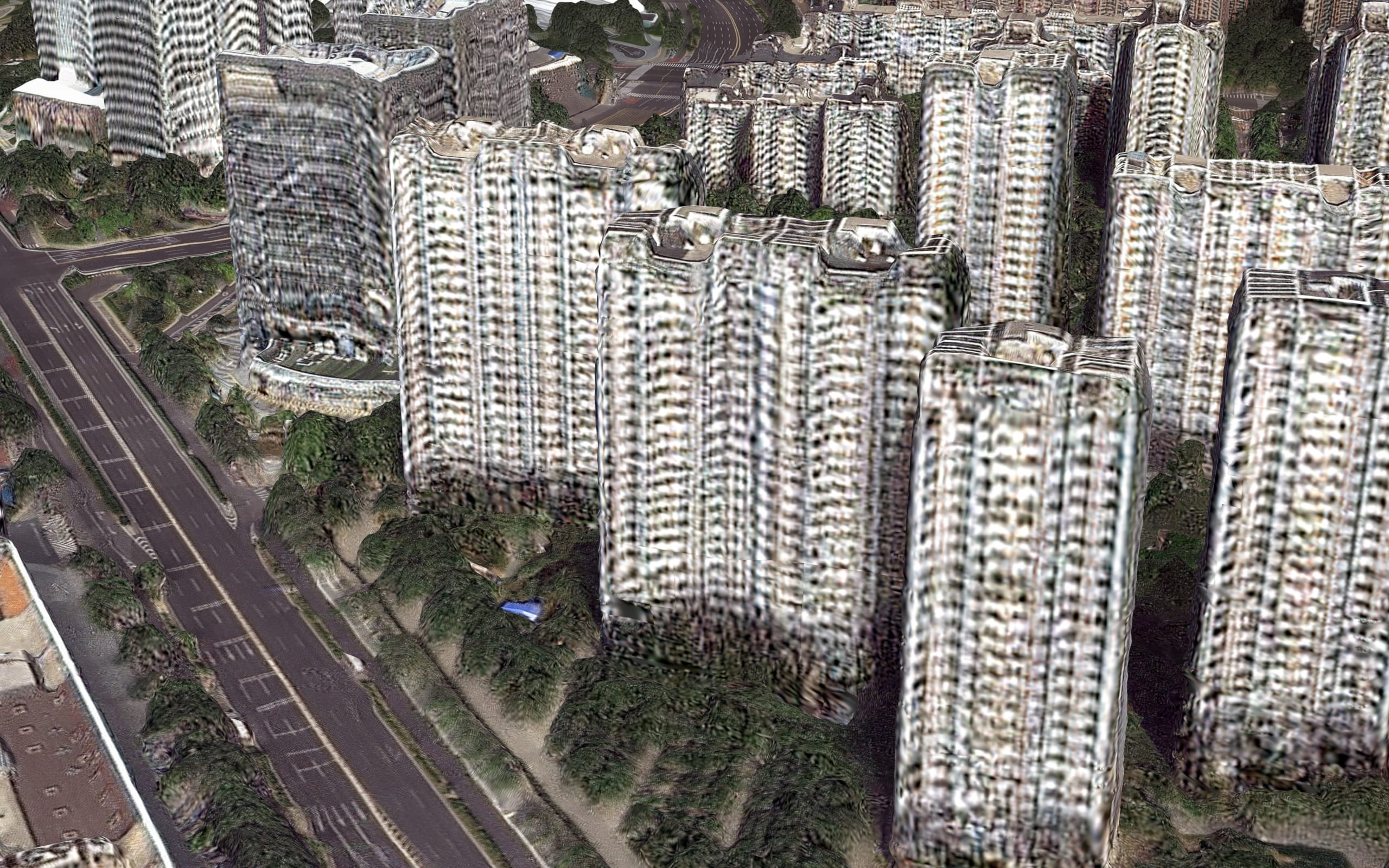} & 
    \imagecell[0.48]{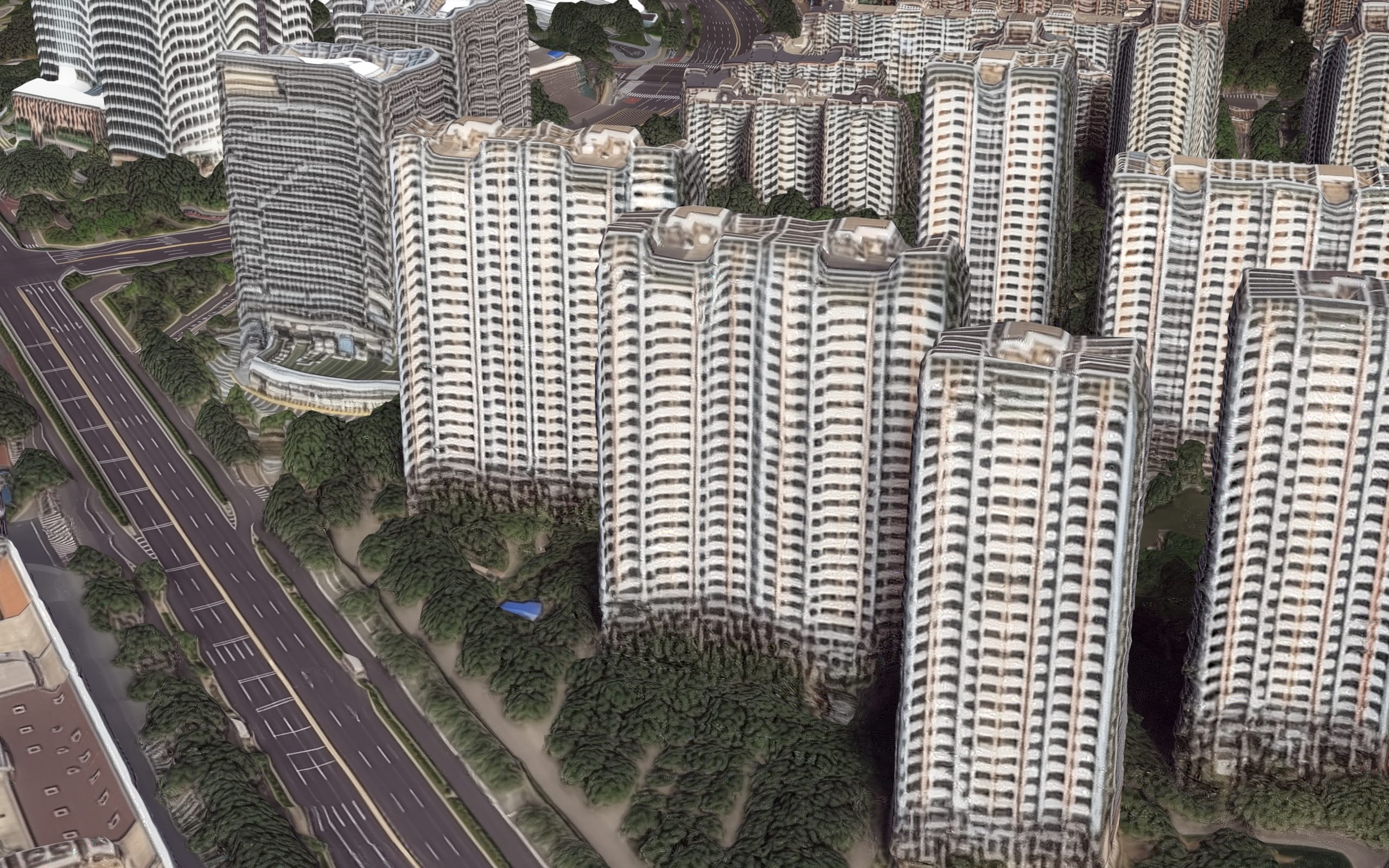} \\
    \vspace{-9pt} \\
    (a) Before Refinement & (b) After Refinement
    \end{tabular}
    \end{spacing}
    \vspace{-5pt}
    \caption{
        \textbf{Appearance Refinement.}
        (a) The basic texture $\textureStageOne$, created by naively back-projecting the blurry source satellite images, suffers from low fidelity and ``baked-in'' artifacts.
        (b) Our final texture $\textureStageFinal$, optimized using supervision from the restoration network, recovers sharp, photorealistic, and globally consistent details.
    }
    \label{fig:appearance-comparison}
    \vspace{-15pt}
\end{figure}

With the robust city-scale mesh $\mesh$ established, we proceed to its appearance modeling. A naive approach of optimizing a texture $\texture$ by back-projecting the source satellite images $\imageSate$ is fundamentally limited by the input quality. As demonstrated in \cref{fig:appearance-comparison}, this process ``bakes'' blurriness, aliasing, and projection artifacts directly into the texture, rendering any synthesized ground-level views unconvincing. To circumvent this, we propose an iterative refinement process that leverages a powerful generative prior to synthesize sharp, plausible textures from these degraded inputs.

\vspace{-10pt}
\paragraph{Basic Texture Creation.}
We first compute a preliminary texture $\textureStageOne$ to provide a coarse but geometrically-aligned starting point. This is achieved by optimizing a texture atlas to match the source satellite images $\imageSate$ via a differentiable renderer $\renderFunc$~\cite{pidhorskyi2024rasterized}, minimizing a combination of MSE and SSIM losses:
\begin{equation}
    \textureStageOne = \arg\min_{\texture} \sum_{\image_i \in \imageSate} \lambda_{\text{MSE}} \mathcal{L}_{\text{MSE}} + \lambda_{\text{SSIM}} \mathcal{L}_{\text{SSIM}} ,
    \label{eq:prelim-loss}
\end{equation}
where $\image_i$ is the $i$-th satellite image, $\camera_i$ is the corresponding camera parameter, $\mathcal{L}_{\text{MSE}} = \left\| \renderFunc(\mesh, \texture, \camera_i) - \image_i \right\|_2^2$ and $\mathcal{L}_{\text{SSIM}} = \left(1 - \text{SSIM}(\renderFunc(\mesh, \texture, \camera_i), \image_i) \right)$.
While visually degraded, this texture $\textureStageOne$ serves as a crucial foundation for the subsequent generative refinement.

\vspace{-10pt}
\paragraph{Image \textbf{Restoration} Network.}
\label{sec:restoration}

The core of our appearance enhancement lies in a powerful generative prior, which we implement by fine-tuning a pre-trained diffusion model. The key is to train this network, denoted $\network$, as a deterministic restorer. It learns to directly map degraded renders ($I_{\text{low}}$) to high-quality targets ($I_{\text{high}}$) from a paired large dataset of diverse 3D urban scenes. This harnesses the rich data distribution of urban appearance learned by the diffusion model while ensuring a stable, repeatable output. The network is optimized by minimizing a combination of a perceptual loss ($\mathcal{L}_{\text{LPIPS}}$) and a fidelity loss (Charbonnier loss $\mathcal{L}_{\text{CHAR}}$ \cite{charbonnier1994two}):
\begin{equation}
    \mathcal{L}_{\text{restorer}} = \mathcal{L}_{\text{LPIPS}}(\hat{I}, I_{\text{high}}) + \lambda_{\text{CHAR}} \mathcal{L}_{\text{CHAR}}(\hat{I}, I_{\text{high}}).
    \label{eq:restorer-loss}
\end{equation}
This process yields a robust restorer $\network$ that serves as an expert on photorealistic urban appearance, ready to guide the subsequent texture refinement.

\vspace{-10pt}
\paragraph{Iterative Texture Refinement.}
\label{sec:tex-opt}

The final texture atlas, $\textureStageFinal$, is achieved through an iterative refinement process that distills knowledge from our image restoration network $\network$. A single iteration unfolds in two steps: First, in the generation step, we render novel close-range views $I_{\text{low}, j} = \renderFunc(\mesh, \textureStageTwo, \camera_{\text{novel}, j})$ from the current texture atlas (initially $\textureStageOne$), using simulated UAV paths. These degraded views are then enhanced by $\network$ into sharp, high-fidelity pseudo-ground truth targets $\hat{I}_{\text{target,j}} = D(I_{\text{low,j}})$. Second, in the optimization step, the texture atlas is updated by minimizing a reconstruction loss against these targets, following the formulation of \cref{eq:prelim-loss}.

This creates a powerful feedback loop: the refined texture from the current iteration serves as a higher-quality input for the next generation step, allowing the model to progressively ``bootstrap'' its way to photorealism \cite{chung2023luciddreamer, yu2025wonderworld}.

\subsection{Implementation Details}
\label{sub-sec:method:impl}

\paragraph{Geometry.}

We implement the Z-Monotonic SDF by parameterizing a field of monotonic curves on a 2D $(x,y)$ grid of resolution $256 \times 256$. Each cell in this grid holds a learnable scalar parameter, $h$, which dictates the vertical offset of a local basis curve.
For any query point $\point = (x, y, z)$, the SDF value is not determined by a single curve, but is synthesized by smoothly interpolating the outputs of multiple basis curves from the local neighborhood on the grid. Specifically, for a set of $n$ neighboring grid locations whose xy-coordinates are $\{ (x_j, y_j) \}_{j=1}^n$ around the query's projection $(x,y)$, the curve $f(z; x, y)$ is represented by $n$ activation function defined as:
\begin{equation}
    f(z; x, y) = \sum_{j=1}^{n} w_j \cdot \tanh(k \cdot (z - h_j)),
\end{equation}
where $h_j$ are learnable parameters from the 2D grid, $k$ is a hyperparameter, and $w_j$ is an interpolation weight that depends on the proximity of the query location $(x,y)$ to the neighbor's location $(x_j, y_j)$.
The grid is optimized using the Adam \cite{kingma2014adam} optimizer with a learning rate of $0.01$. 
The optimization is supervised by the $\mathcal{L}_{\text{geo}}$ loss (\cref{eq:chamfer_loss}), which balances the Z-axis distance to MVS points and a Laplacian loss $\mathcal{L}_{\text{Lap}}$ and a normal consistency loss $\mathcal{L}_{\text{Nrm}}$ (weighted by $\lambda_{\text{Lap}}$ and $\lambda_{\text{Nrm}}$).

\vspace{-15pt}
\paragraph{Appearance.}

For stability, geometry and texture are optimized separately. The texture is parameterized as a UV atlas, with all rendering performed by EdgeGrad~\cite{pidhorskyi2024rasterized}. The preliminary texture $\textureStageOne$ is optimized for 100 epochs, with $\lambda_{\text{MSE}} = 0.8, \lambda_{\text{SSIM}} = 0.2$. Our restoration network $\network$ fine-tuned from the FLUX-Schnell ~\cite{flux2024} on 100,000 aerial image pairs, is trained for 10,000 iterations (batch size 96) with  $\lambda_{\text{CHAR}} = 1$ in $\mathcal{L}_{\text{restorer}}$ (\cref{eq:restorer-loss}). Finally, the texture is refined over 2 iterations for efficiency. For each iteration, we generate target images of resolution $2048 \times 2048$ by sampling novel views from a uniform $150\,\mathrm{m}$ grid over the mesh bounding box extended by $100\,\mathrm{m}$ (altitude $450\,\mathrm{m}$, pitch $45^{\circ}$, four cardinal orientations) and optimize $\textureStageFinal$ for $20$ epochs.
 
\vspace{-15pt}
\paragraph{Platform.}   
Our method runs on a single NVIDIA A6000 GPU and requires approximately 1.5 hours to process a $1 \, \mathrm{km}^2$ urban area. This time comprises 0.5 hours for geometry optimization and 1 hour for appearance refinement. 
Further details on hyperparameters and network architectures are provided in Appendix~\ref{sec:impl}.

\section{Experiments}
\label{sec:experiments}

\begin{table*}
\vspace{-3mm}
\begin{center}
\renewcommand{\arraystretch}{1.2}
\resizebox{\textwidth}{!}{ 
\begin{tabular}{c | ccccc | ccc | ccc}
\toprule
\multirow{2}{*}{\textbf{Method}} & \multicolumn{5}{c|}{\textbf{MatrixCity}} & \multicolumn{3}{c|}{\textbf{DFC 2019}} & \multicolumn{3}{c}{\textbf{Google Earth}} \\
 & \textbf{P} $\uparrow$ & \textbf{R} $\uparrow$ & \textbf{F1} $\uparrow$ & \textbf{CD} $\downarrow$ & \textbf{PSNR} $\uparrow$ & \textbf{PSNR} $\uparrow$ & \textbf{SSIM} $\uparrow$ & \textbf{LPIPS} $\downarrow$ & \textbf{PSNR} $\uparrow$ & \textbf{SSIM} $\uparrow$ & \textbf{LPIPS} $\downarrow$ \\ \midrule
Mip-Splatting & 0.282 & \cellcolor{best2}0.569 & \cellcolor{best3}0.377 & \cellcolor{best3}0.161 & \cellcolor{best2}15.992 & \cellcolor{best3}10.289 & \cellcolor{best2}0.346 & \cellcolor{best3}0.816 & 12.214 & \cellcolor{best2}0.245 & \cellcolor{best3}0.551 \\
2DGS & \cellcolor{best1}0.693 & \cellcolor{best3}0.464 & \cellcolor{best2}0.556 & \cellcolor{best2}0.073 & 13.469 & 7.366 & 0.304 & 0.848 & 11.022 & 0.196 & 0.622 \\
CityGS-X & 0.278 & 0.143 & 0.189 & 0.227 & 13.542 & FAIL & FAIL & FAIL & \cellcolor{best2}12.674 & 0.223 & 0.591 \\
Skyfall-GS & \cellcolor{best3}0.352 & 0.256 & 0.296 & 0.359 & \cellcolor{best3}15.949 & \cellcolor{best2}12.460 & \cellcolor{best3}0.330 & \cellcolor{best2}0.740 & \cellcolor{best3}12.456 & \cellcolor{best3}0.233 & \cellcolor{best1}0.521 \\
Ours & \cellcolor{best2}0.673 & \cellcolor{best1}0.615 & \cellcolor{best1}0.643 & \cellcolor{best1}0.036 & \cellcolor{best1}17.153 & \cellcolor{best1}13.059 & \cellcolor{best1}0.358 & \cellcolor{best1}0.556 & \cellcolor{best1}12.770 & \cellcolor{best1}0.253 & \cellcolor{best2}0.546 \\ 
\bottomrule
\end{tabular}
}
\end{center}
\vspace{-5mm}
\caption{
    \label{table:exp:all} 
    \textbf{Quantitative evaluation of our method with SOTA reconstruction works.}
    Our method outperforms existing approaches in both geometric accuracy and visual quality for satellite reconstruction.
    ``FAIL'' denotes the method fails to converge in experiment, manifested as program crashes.
}
\vspace{-11pt}
\end{table*}

\begin{figure*}[tb]
	\centering
    \begin{spacing}{1} 
    \setlength\tabcolsep{1pt}
    \begin{tabular}{ccccccc}
    
    {\rotatebox[origin=c]{90}{\textbf{MatrixCity}} \hspace{1pt}}  & 
    \imagecell[0.325]{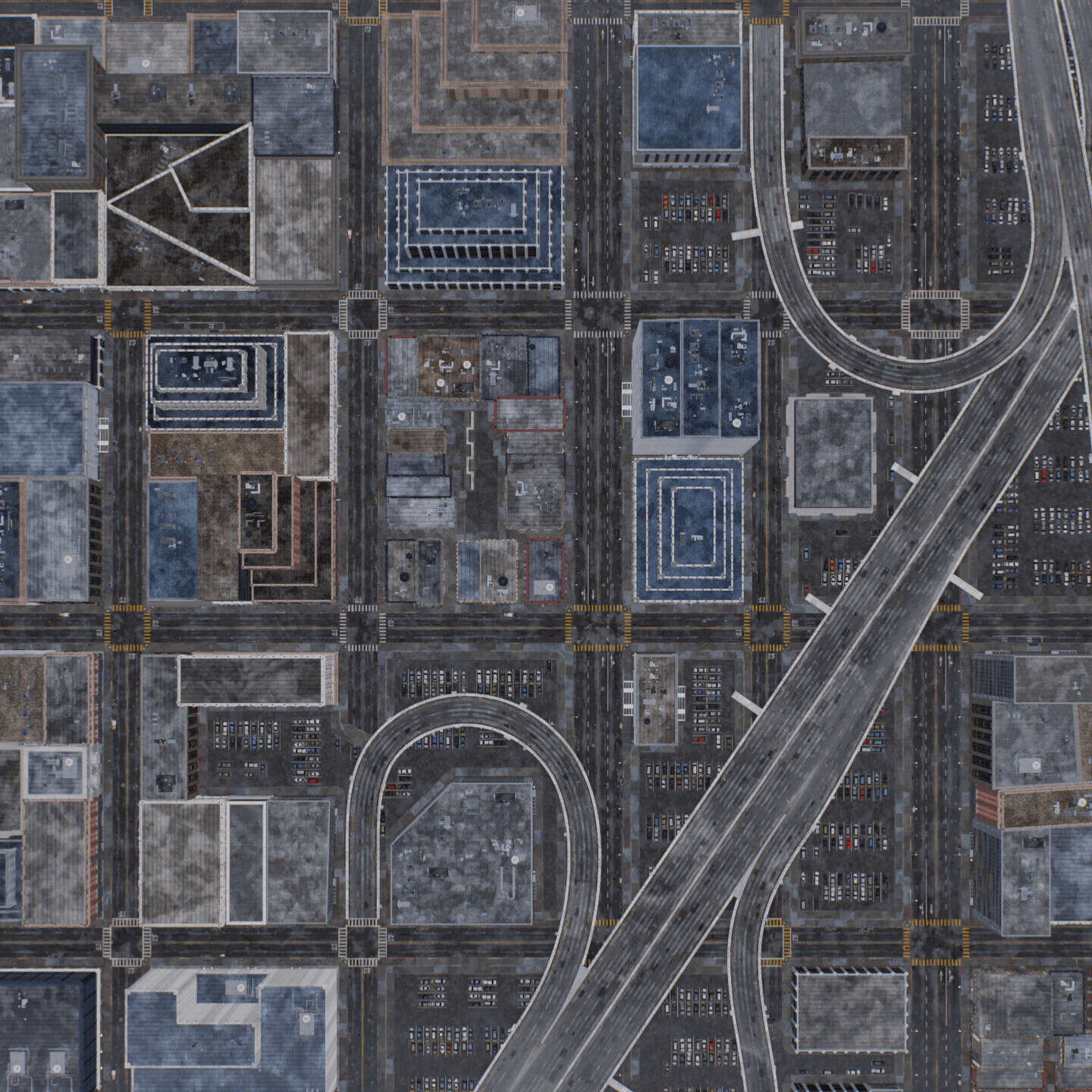} & 
    \imagecell[0.325]{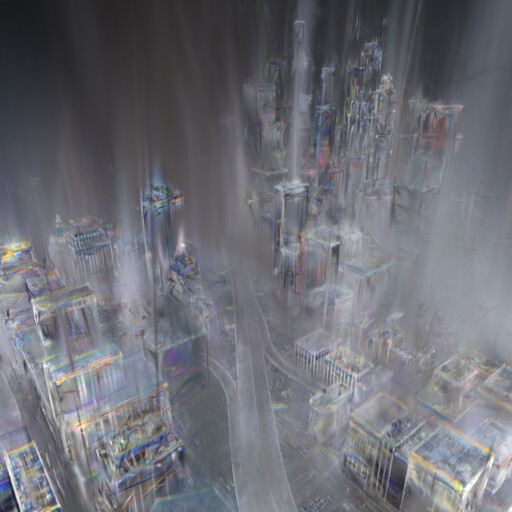} & 
    \imagecell[0.325]{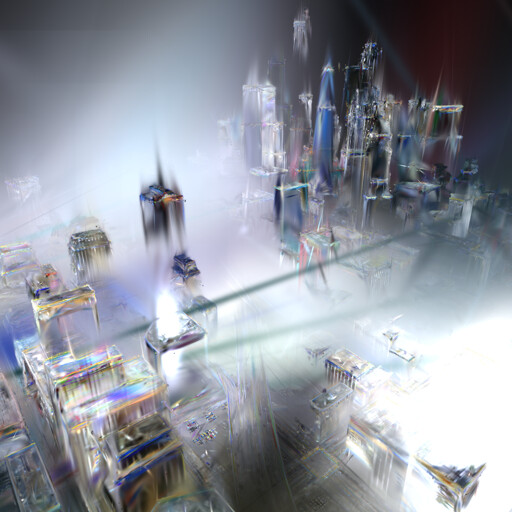} & 
    \imagecell[0.325]{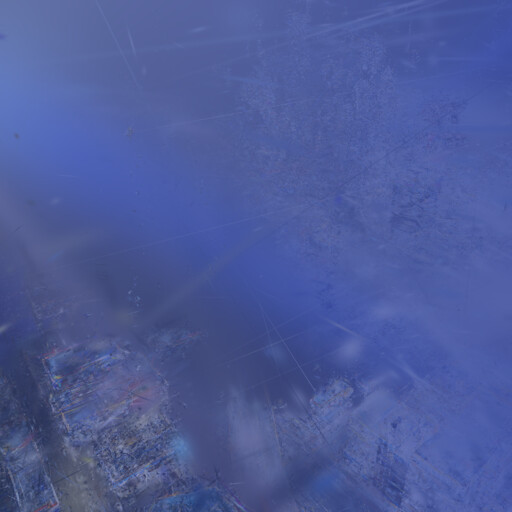} & 
    \imagecell[0.325]{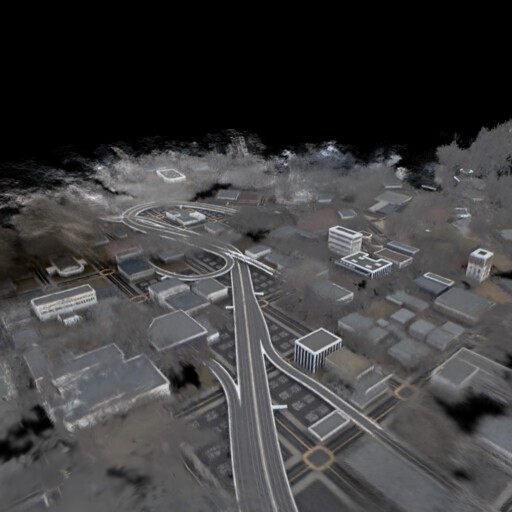} & 
    \imagecell[0.325]{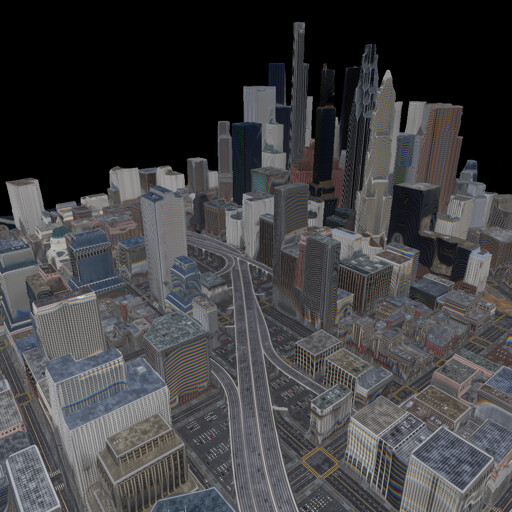} \\
    \vspace*{-10pt} \\

    {\rotatebox[origin=c]{90}{\textbf{DFC 2019}} \hspace{1pt}} & 
    \imagecell[0.325]{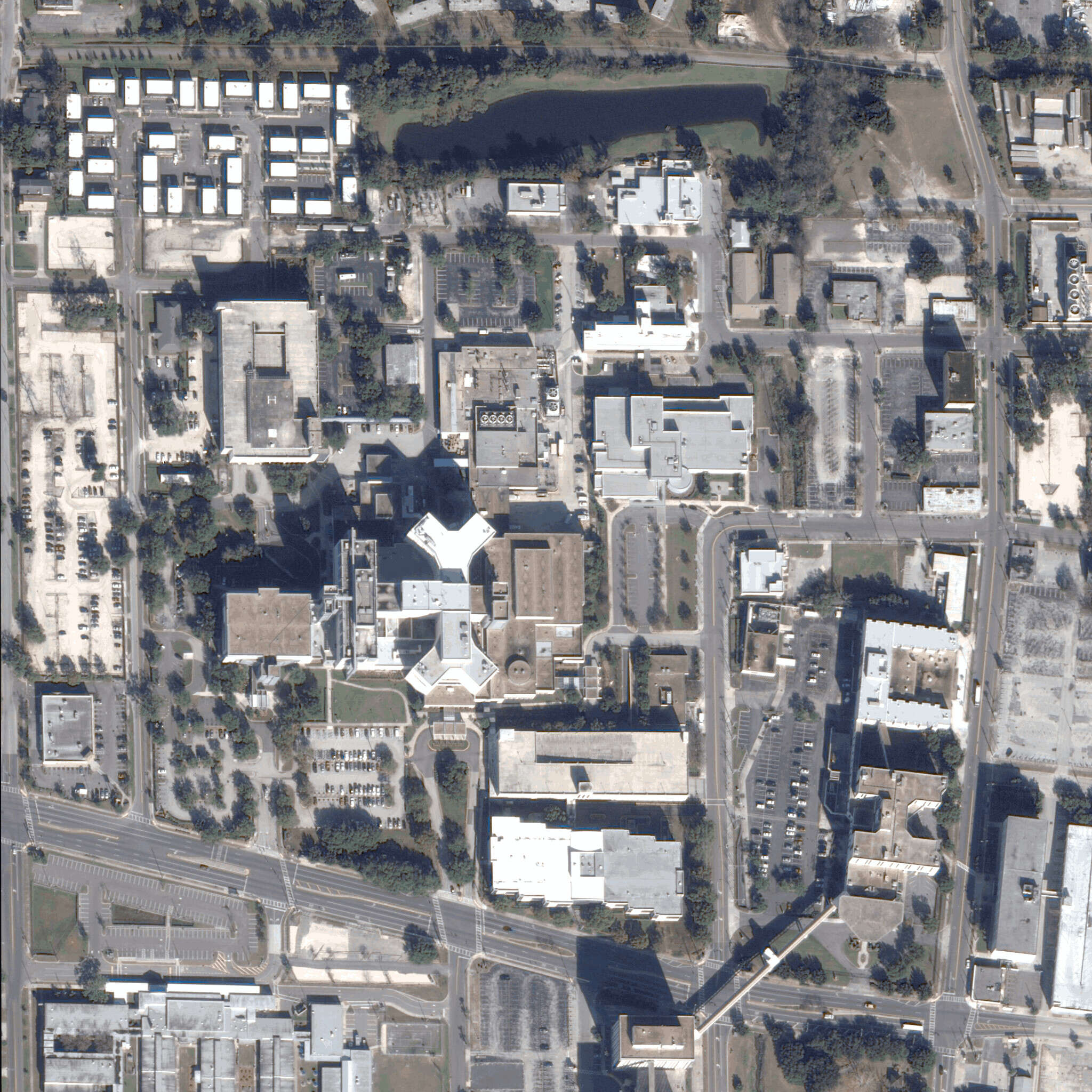} & 
    \imagecell[0.325]{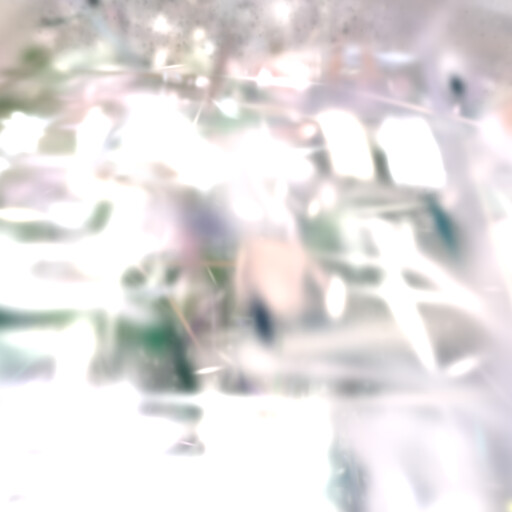} & 
    \imagecell[0.325]{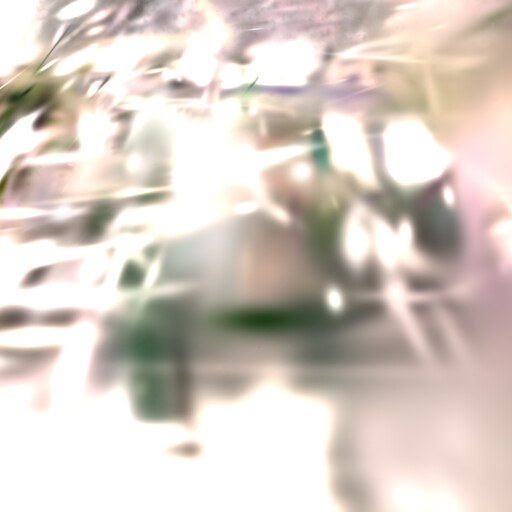} & 
    \imagecell[0.325]{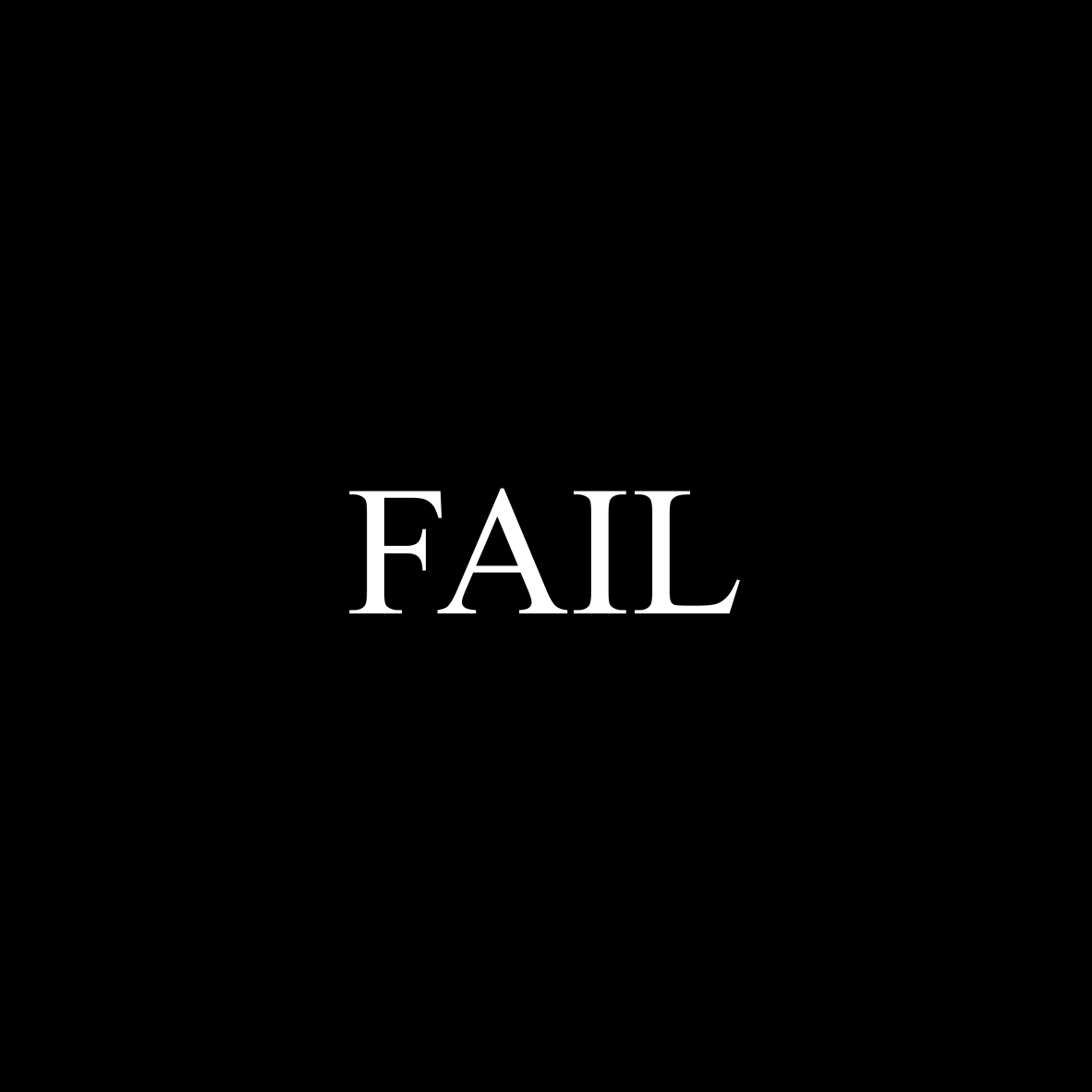} & 
    \imagecell[0.325]{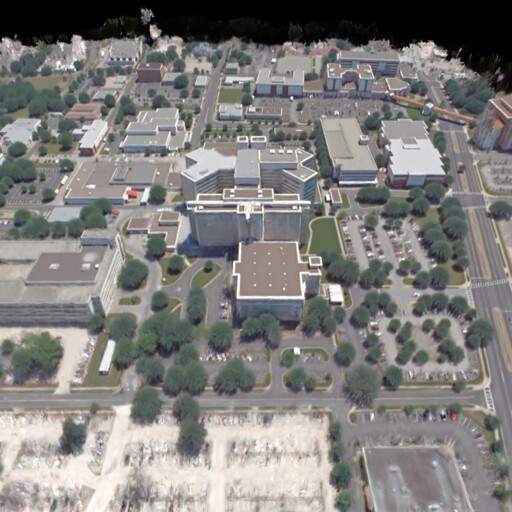} & 
    \imagecell[0.325]{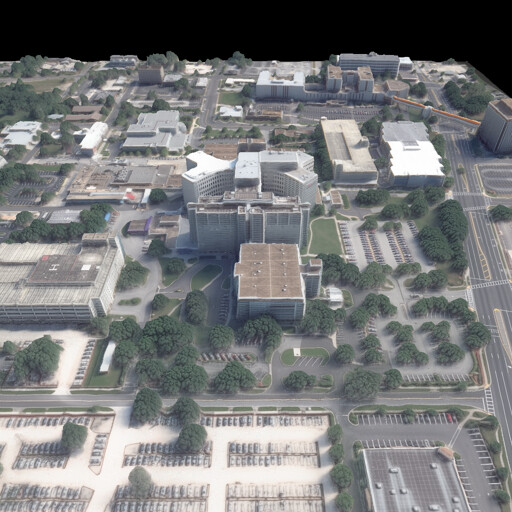} \\ 
    \vspace*{-10pt} \\

    {\rotatebox[origin=c]{90}{\textbf{GoogleEarth}} \hspace{1pt}} & 
    \imagecell[0.325]{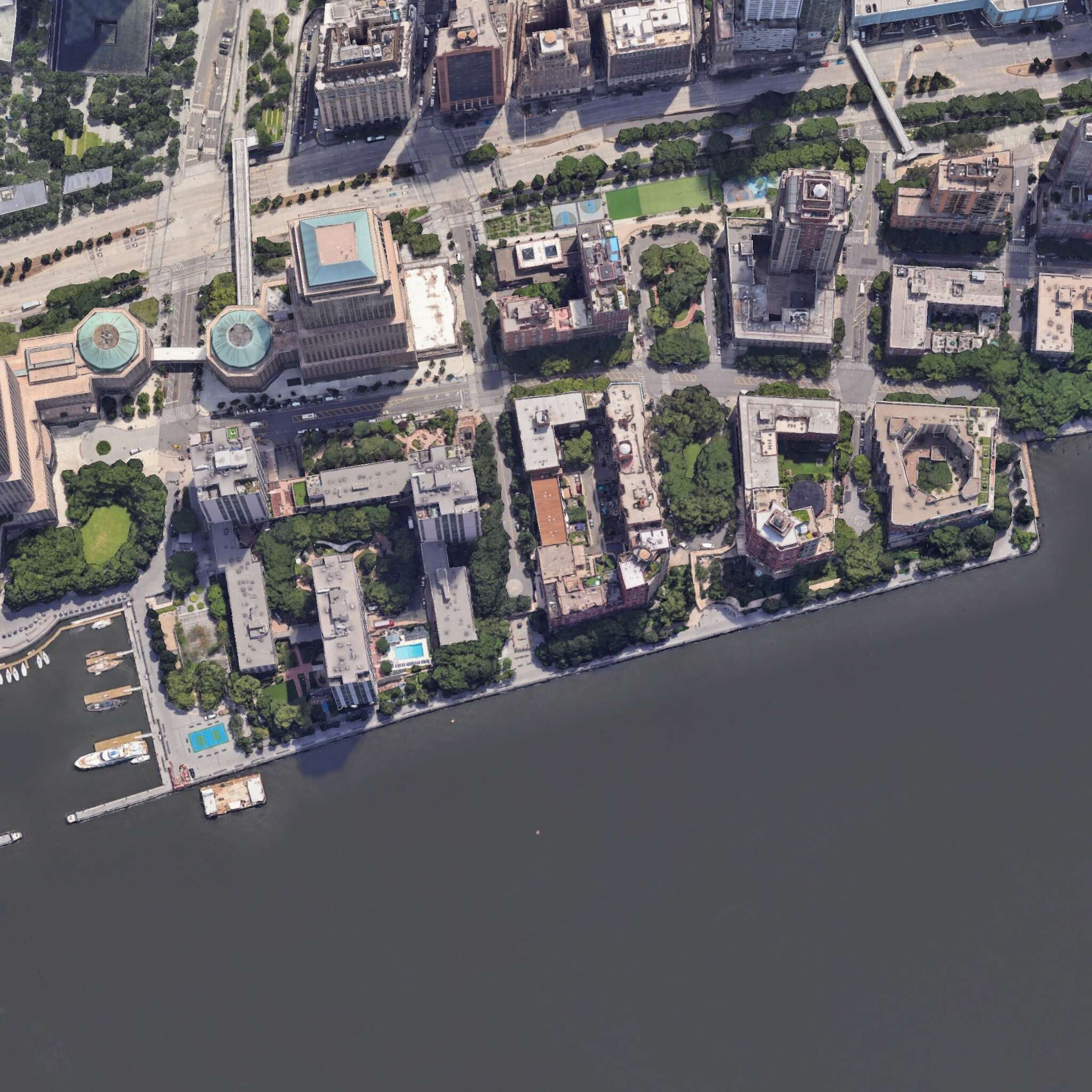} & 
    \imagecell[0.325]{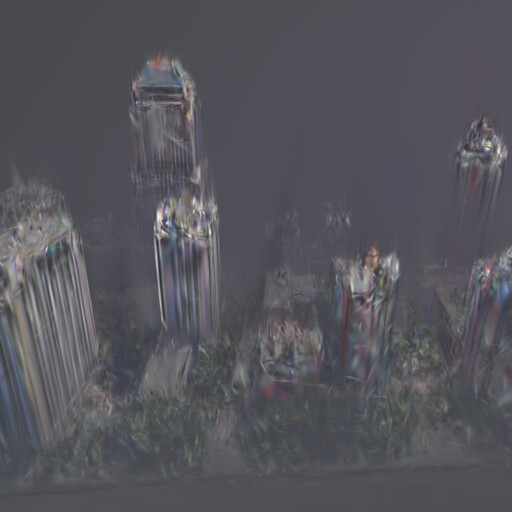} & 
    \imagecell[0.325]{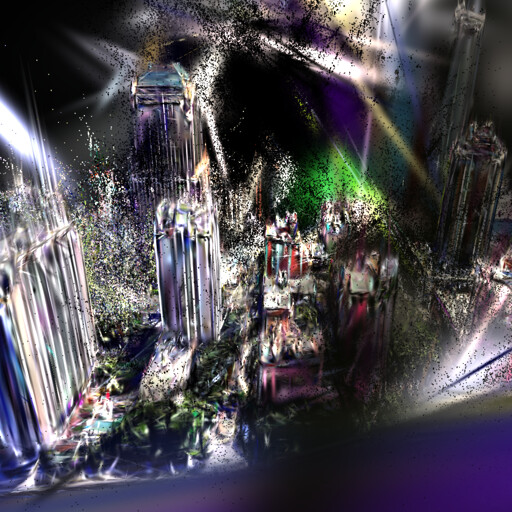} & 
    \imagecell[0.325]{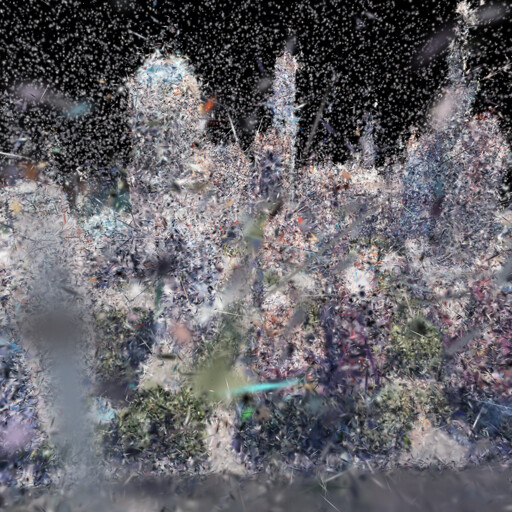} & 
    \imagecell[0.325]{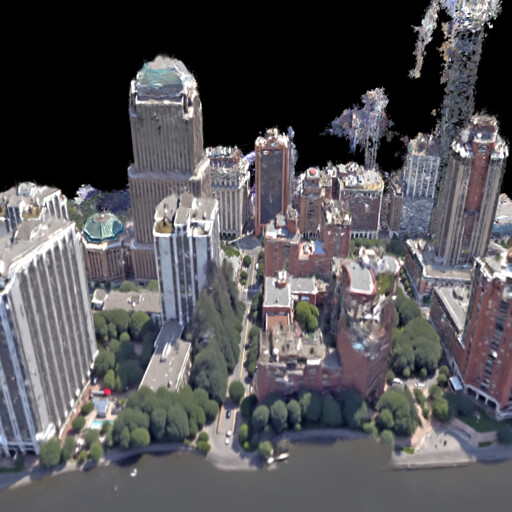} & 
    \imagecell[0.325]{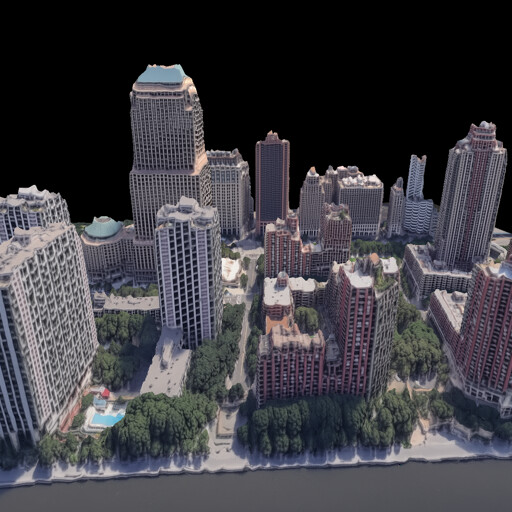} \\
    \vspace*{-10pt} \\

    {\rotatebox[origin=c]{90}{\textbf{Urban Scene}} \hspace{1pt}} & 
    \imagecell[0.325]{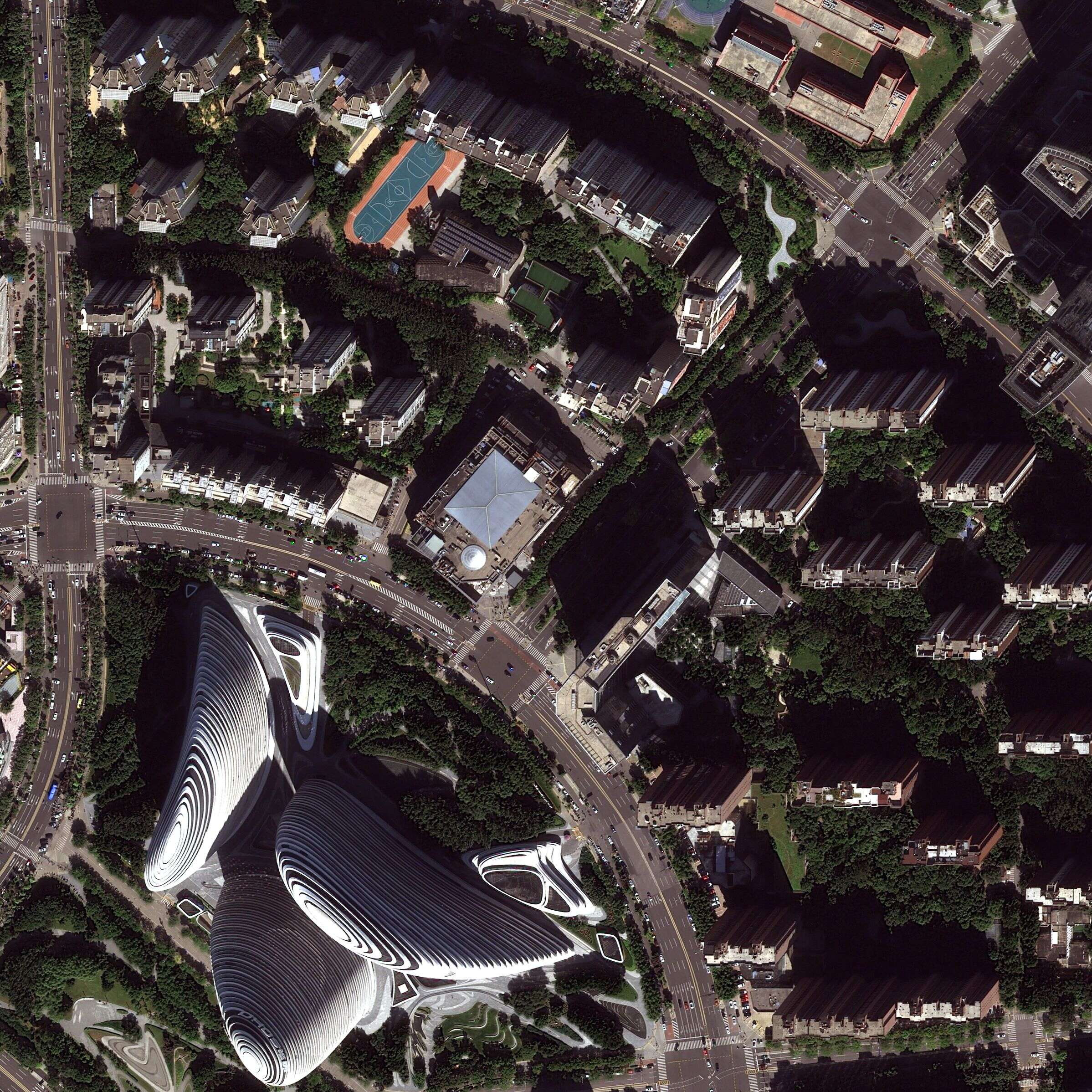} & 
    \imagecell[0.325]{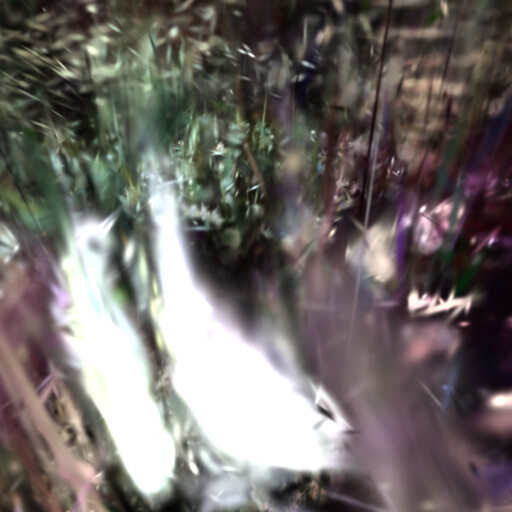} & 
    \imagecell[0.325]{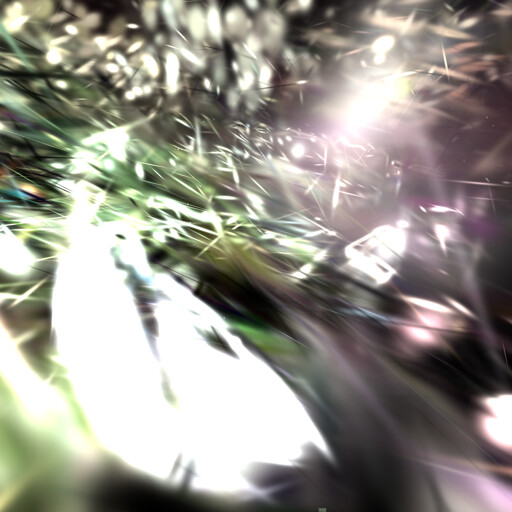} & 
    \imagecell[0.325]{figures/qualitative/fail.pdf} & 
    \imagecell[0.325]{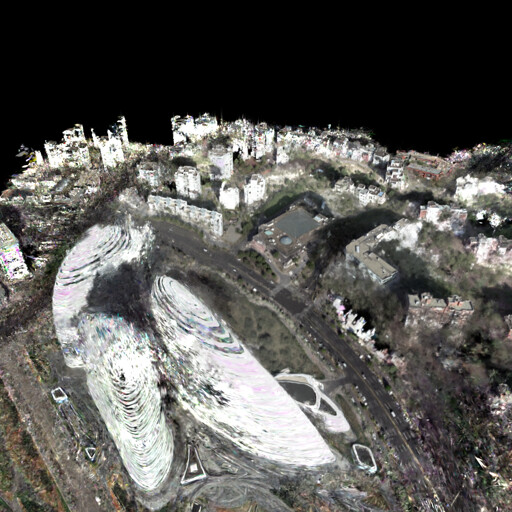} & 
    \imagecell[0.325]{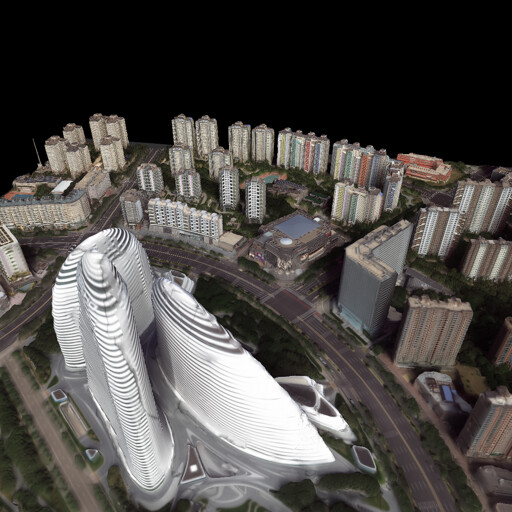} \\ 
    \vspace*{-10pt} \\
    
    \\
    \vspace*{-20pt}
    \\
    &
    Input Example & 
    Mip-Splatting & 
    2DGS & 
    CityGS-X & 
    Skyfall-GS & 
    Ours \\
    
    \end{tabular}
    \end{spacing}
	\caption{ 
    \textbf{Visualization of reconstruction results}. 
    Compared to baselines, our method successfully achieves high-quality city reconstruction from satellite imagery. ``FAIL'' denotes the method fails to converge in experiment, manifested as program crashes.
    The ground-view locations corresponding to these examples are provided in Appendix~\ref{sub-sec:supp:viewpoints}.
    }
    \label{fig:qualitative}
    \vspace*{-0.3cm}
\end{figure*}

\subsection{Experimental Setup}

\paragraph{Datasets.} 
To comprehensively validate the geometric accuracy, visual quality and real-world usage of our method, we evaluate on both synthetic and real-world datasets.
Specifically, we select four datasets:

\textit{MatrixCity-Satellite}. Currently there is no open benchmark for 3D reconstruction from satellite imagery that contains accurate ground-truth point clouds. To evaluate geometry quality, we utilize the synthetic 3D scene in MatrixCity dataset and UE5 engine \cite{li2023matrixcity} to synthesize satellite images with ground-truth point clouds, enabling quantitative evaluation of both novel view synthesis quality and geometry accuracy.
In particular, we synthesize 50 training images covering an area of $1 \, \mathrm{km}^2$ to maximally simulate satellite imagery.
For consistent and fair evaluation, we follow the \textit{MatrixCity-Aerial} test protocol, and evaluate geometric accuracy and novel view synthesis on the central $800 \, \mathrm{m} \times 800 \, \mathrm{m}$ area, thereby avoiding evaluation on regions with degraded reconstruction boundaries.

\textit{2019 IEEE GRSS Data Fusion Contest (DFC 2019)} dataset is a representative real-world satellite imagery dataset that features in high-quality WorldView-3 satellite images 
and is widely used in the remote sensing area \cite{zhang2024satensorf,mari2022sat,aira2025gaussian}. Following their protocol, we evaluate on four standard Areas of Interest (AOIs): JAX\_004, JAX\_068, JAX\_214, and JAX\_260.

\textit{GoogleEarth} dataset features an appearance-consistent unification of training satellite views and testing ground-level views of reconstructed New York City by Google Earth.
We follow \cite{lee2025skyfall} to evaluate visual quality on four AOIs (004, 010, 219, 336).

\textit{Urban Scene}. For a more rigorous evaluation of scalability and robustness, we curated a challenging real-world test case from satellite imagery of a modern metropolis. This scene is characterized by a large-scale, dense urban environment with numerous high-rise buildings and complex geometric layouts.

More details are provided in Appendix~\ref{sub-sec:datasets}.

\vspace{-10pt}
\paragraph{Baselines.}

To comprehensively evaluate our method for city-scale reconstruction, we benchmark it against representative SOTA methods from three key domains.
For \textit{geometric accuracy}, we compare against 2DGS \cite{huang20242d}, which aims to reconstruct a geometrically accurate radiance field.
For \textit{visual quality}, we benchmark against Mip-Splatting \cite{Yu2024MipSplatting}, a leading method for high-quality rendering.
For \textit{large-scale urban modeling}, we compare against CityGS-X \cite{gao2025citygs} as a SOTA method for city-scale scene reconstruction.
Additionally, we include a comparison with Skyfall-GS \cite{lee2025skyfall}, as it is a satellite-based reconstruction method that enables low-altitude novel view synthesis. 
These comparisons allow for a comprehensive evaluation of our method's capabilities across different aspects of this reconstruction task.

\vspace{-10pt}
\paragraph{Metrics.}
We evaluate our method from two perspectives: \textit{geometric accuracy} and \textit{visual quality}.
In terms of \textit{geometric accuracy}, we follow \cite{liu2024citygaussianv2,gao2025citygs} to evaluate three metrics Precision (P), Recall (R), F1-Score between sampled and GT point clouds. 
We also follow \cite{huang20242d} to evaluate Chamfer distance (CD) as a supplement. 
In terms of \textit{visual quality}, we adhere to standard practices by measuring PSNR, SSIM and LPIPS to evaluate the rendering quality of novel views.
Details about point cloud extraction, test view synthesis, \etc, are provided in Appendix~\ref{sub-sec:baselines} and \ref{sub-sec:metrics}.

\subsection{Experiment Results}

\paragraph{Quantitative Results.}

Quantitative results are provided in \cref{table:exp:all}.
Across all datasets, our method consistently achieves superior geometric accuracy and visual quality over existing approaches. 

For \textit{geometric accuracy}, it surpasses all baselines in Recall, F1-Score and Chamfer distance, improving F1-Score by $0.09$ and reducing Chamfer distance by $50\%$ relative to the best competitor. 
These gains stem from our 2.5D Z-Monotonic SDF, which enforces coherent roof–facade geometry under satellite views. 
Although 2DGS~\cite{huang20242d} attains slightly higher Precision by extracting detailed rooftops, its lack of meaningful facades produces ``shrink-wrapped'' structures (\cref{fig:qualitative}), validating our choice of a 2.5D prior.

For \textit{visual quality}, our method establishes a new SOTA, proving particularly dominant on challenging, low-altitude datasets like DFC 2019. While still highly competitive, our lead is less pronounced on the Google Earth benchmark, which we attribute to its higher-altitude test views that limit the visibility of facade texture details.

\vspace*{-10pt}
\paragraph{Qualitative Results.}
\begin{table}[tbp]
\vspace{-1mm}
\begin{center}
  \resizebox{1.0\linewidth}{!}{
  \begin{tabular}{l|ccc}
  \toprule
  \multirow{2}{*}{\textbf{Description}} & \multicolumn{2}{c}{\textbf{Geometry}} & \textbf{NVS} \\
  & \textbf{F1} $\uparrow$ & \textbf{CD} $\downarrow$ & \textbf{PSNR} $\uparrow$ \\
  \midrule
  \multicolumn{1}{l|}{\textbf{Geometry}} \\
  Naive Marching Cubes 128 & 0.279 & 0.0749 & 16.800 \\
  Naive Marching Cubes 256 & 0.412 & 0.0757 & 17.002 \\
  w/o Regularization Loss & 0.637 & 0.0364 & 17.115 \\
  \textbf{Full Modeling}  & \textbf{0.643} & \textbf{0.0357} & \textbf{17.153} \\
  \midrule
  \multicolumn{1}{l|}{\textbf{Appearance}} \\
  w/o Image Restoration Network & - & - & 17.038 \\
  \textbf{Full Modeling} & - & - & \textbf{17.153} \\
  \bottomrule
\end{tabular}
  }
  \end{center}
\vspace{-5mm}
\caption{
\label{table:exp:ablation}
\textbf{Ablation Studies.} We conduct extensive quantitative comparisons to validate the effectiveness of each module.
}
\vspace{-20pt}
\end{table}

\begin{figure}[tb]
    \centering
    \begin{spacing}{1}
    \setlength\tabcolsep{0pt}
    \begin{tabular}{cccc}
    \imagecell[0.23]{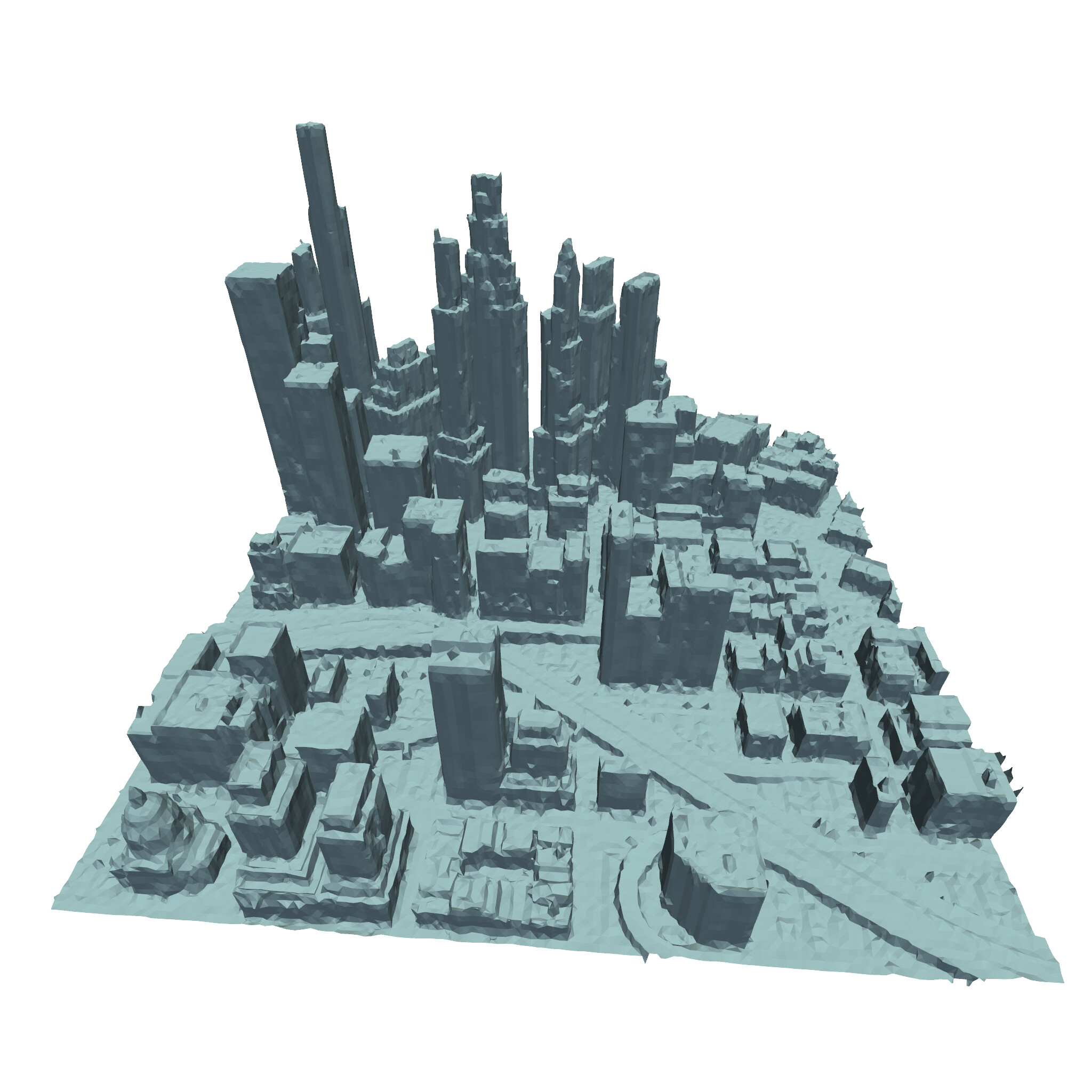} & 
    \imagecell[0.23]{figures/zsdf-vs-mc/zsdf.jpg} &
    \imagecell[0.23]{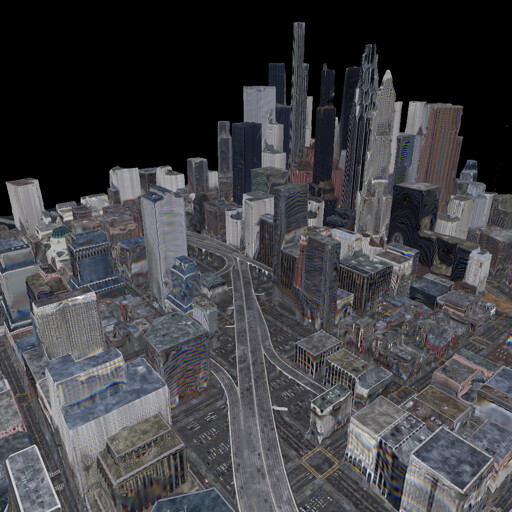} & 
    \imagecell[0.23]{figures/qualitative/ours-stage2_matrixcity.jpg}
    \\
    \vspace{-9pt} \\
    {\small (a) w/o Reg. Loss} & 
    {\small (b) Full Model} & 
    {\small (c) w/o Img. Res.} &
    {\small (d) Full Model}
    \end{tabular}
    \end{spacing}
    \vspace{-5pt}
    \caption{\textbf{Qualitative results on ablation study.} Please refer to \cref{fig:method-roof-morethan-facade} for Marching Cubes results. The absence of Z-Monotonic SDF, regularization loss and image restoration network consistently results in significant artifacts in geometry and texture.}
    \label{fig:ablation}
    \vspace{-15pt}
\end{figure}

Qualitative results are illustrated in \cref{fig:qualitative}.
Our qualitative results visually underscore the superiority of our decoupled geometry and appearance modeling. 
As shown, our method consistently yields clean geometry and sharp, photorealistic textures. 
In contrast, baseline approaches such as Mip-Splatting, 2DGS, and CityGS-X produce fragmented, ``floating'' geometry because their point-based representations lack a strong structural prior to regularize the ill-posed reconstruction from sparse views. 
While Skyfall-GS shows some robustness, its results suffer from blurry textures and cross-view inconsistencies, a common pitfall of generative approaches that fail to properly enforce global coherence.

We further demonstrate our method's scalability and robustness on our challenging ``Urban Scene'' test case. 
As seen in \cref{fig:qualitative}, our approach demonstrates remarkable resilience in this demanding scenario. 
The strong geometric prior proves essential, maintaining structural integrity across the vast and complex scene where other methods might fail. 
On this stable geometric scaffold, our generative refinement then synthesizes sharp, view-consistent details, preserving distinctive facades and vegetation across the entire district. 
This validates that our two-stage design is not only effective but also highly scalable and robust to the complexities of large-scale urban modeling. 

More experiment results are provided in Appendix~\ref{sec:more-results}.

\subsection{Ablation Study}

We conduct ablation studies on the MatrixCity dataset to validate our key design choices. The results, summarized in \cref{table:exp:ablation}, highlight the contribution of each component to both geometric and appearance quality.
More results are provided in Appendix~\ref{sub-sec:ablation-matrix}.

\vspace{-10pt}
\paragraph{Geometry}
Our Z-Monotonic SDF representation effectively avoids aliasing and other artifacts in surface extraction, ensuring the quality and precision of the reconstructed mesh.
Replacing it (optimized with \cite{shen2023flexible}) with a naive baseline that directly extracts a mesh from a voxel grid via Marching Cubes~\cite{lorensen1998marching} results in Precision and PSNR drops, regardless of the resolution setting.
Our regularization losses $\mathcal{L}_{\text{Lap}}$ and $\mathcal{L}_{\text{Nrm}}$ are designed to encourage smoother and more plausible surfaces. 
Removing these losses results in a degradation of geometric accuracy, adversely affecting visual quality.
Moreover, the geometry becomes much noisier when the regularization loss is removed (\cref{fig:ablation} (a, b)).

\vspace{-10pt}
\paragraph{Appearance}
Disabling our image restoration network also results in a PSNR drop.
Moreover, removing the restoration stage results in degraded textures with pronounced artifacts (\cref{fig:ablation} (c, d)), highlighting its critical role in bridging the satellite-to-ground viewpoint gap.

\begin{figure}
    \centering
        \includegraphics[width=0.8\linewidth]{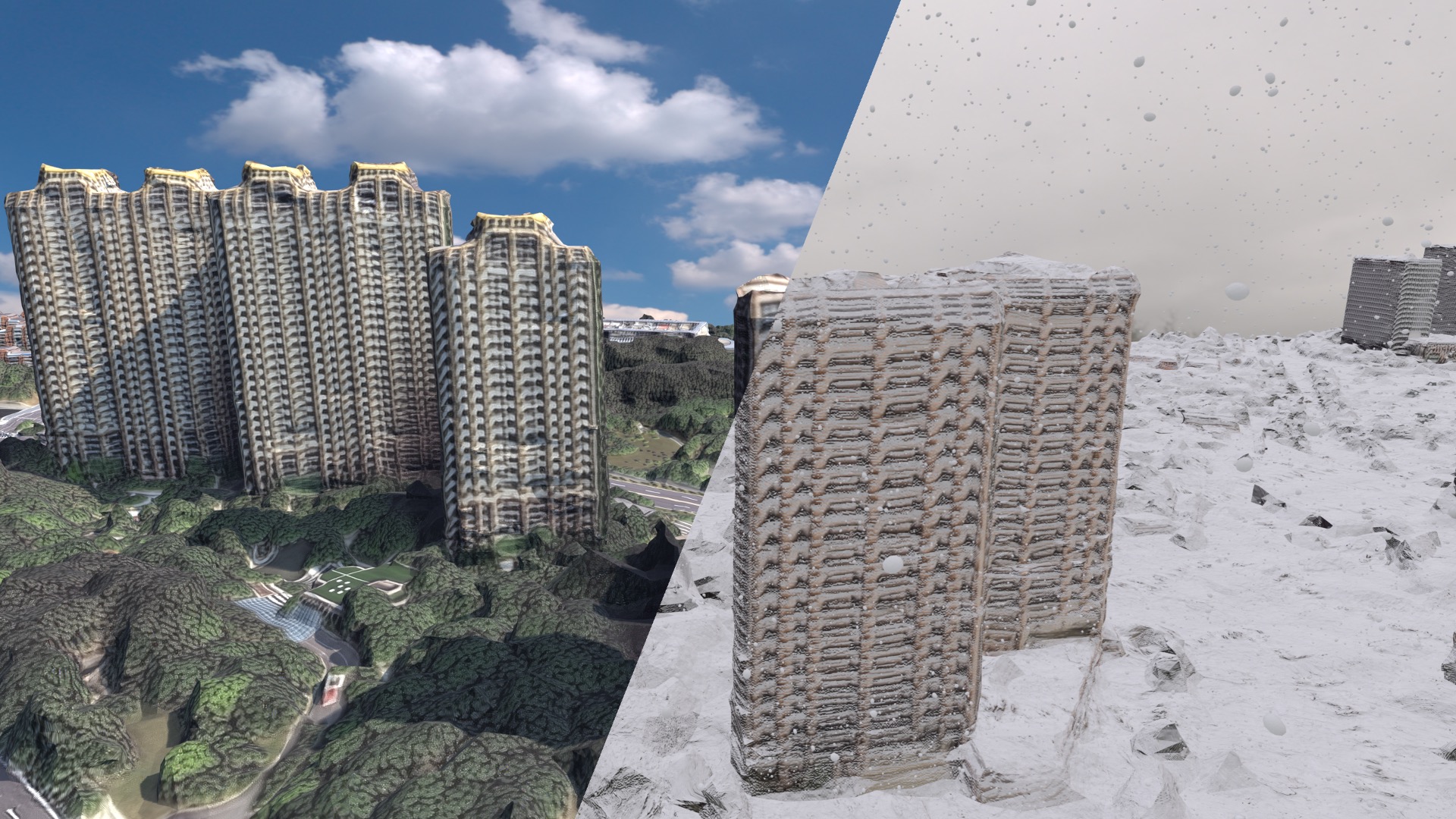}
    \caption{\textbf{Visualization of snow simulation in \textit{Urban Scene}.} Our mesh output enables wide downstream applications.}
    \label{fig:app-sim}
    \vspace{-15pt}
\end{figure}

\subsection{Applications}

A key application of our method is the creation and maintenance of city-scale Digital Twins from satellite imagery. 
Our high-fidelity geometry and elevation reconstructions provide the foundation for large-scale urban simulations such as snow modeling, as illustrated in \cref{fig:app-sim}.
Furthermore, our method's versatility extends to aerial view reconstruction, as demonstrated in Appendix~\ref{sub-sec:aerial}.

\vspace{-5pt}
\section{Discussion}
\label{sec:conclusion}

\paragraph{Conclusions} 

In this paper, we propose a novel framework for high-quality, city-scale 3D reconstruction from challenging satellite imagery. 
Our approach successfully addresses the critical limitations of traditional methods, which often fail under the narrow parallax and poor image quality inherent in satellite data. 
The core of our solution lies in two tailored design choices: (1) a Z-Monotonic SDF representation that robustly models urban geometry as a 2.5D height map, producing clean, watertight meshes with crisp roofs and vertically extruded facades; 
and (2) a large-scale texture restoration network that effectively recovers high-fidelity appearance details from blurry inputs. Through extensive experiments, we have demonstrated that our method achieves state-of-the-art performance in both geometric accuracy and texture fidelity.

\vspace{-10pt}
\paragraph{Limitations}

Our method's core strengths also define its primary limitations, suggesting two main avenues for future work.
First, our reliance on a strong 2.5D geometric prior, while crucial for stabilizing reconstruction from sparse views, imposes fundamental constraints. It inherently precludes modeling of non-monotonic structures (\eg, bridges) and makes the final quality contingent on a topologically sound initial mesh, as our pipeline is designed for refinement, not holistic correction. A failure case is illustrated in Appendix~\ref{sub-sec:supp:failure-cases}. Future work could explore hybrid representations that integrate local, full 3D models for complex areas.
Second, our generative appearance refinement prioritizes visual plausibility over strict factual accuracy. This can lead to the replacement of unique, real-world features with generic textures. Similarly, it synthesizes a single, canonical appearance, rather than explicitly modeling temporal variations present in multi-date satellite imagery. A promising direction is to enhance factual fidelity by incorporating multi-modal data, such as street-view imagery, or by developing time-conditioned appearance models.

\section{Acknowledgement}

This work is supported by the projects of Beijing Science and Technology Program (Z251100008125028).

{
    \small
    \bibliographystyle{ieeenat_fullname}
    \bibliography{main}
}

\clearpage
\setcounter{page}{1}
\maketitlesupplementaryonecol

\appendix
\setcounter{table}{0}
\setcounter{figure}{0}
\setcounter{equation}{0}
\renewcommand{\thetable}{A\arabic{table}}
\renewcommand{\thefigure}{A\arabic{figure}}
\renewcommand{\theequation}{A\arabic{equation}}

\section{Appendix Overview}

In this appendix, we provide additional details and results to complement the main paper.
Specifically, \cref{sec:impl} presents extended implementation details of our geometry and appearance pipelines, including the acquisition of satellite-based point clouds, the optimization of our Z-Monotonic SDF, and details for appearance modeling.
\cref{sec:exp-settings} further describes the experimental settings, including dataset configurations, baseline setups, and evaluation metrics.
Finally, \cref{sec:more-results} reports more extensive quantitative and qualitative results, ablation studies, and an additional application example that validate the scalability and robustness of our method.
We additionally provide an extra comparison with mesh-recovery baselines under sparse satellite observations, a real-scene geometric evaluation using partial LiDAR, a visualization of the ground-view locations used in the main-paper qualitative results, and further analyses on UV atlas parameterization and input sparsity.

\section{Implementation Details}
\label{sec:impl}

In this section, we introduce additional implementation details in our framework.

\subsection{Point Clouds from Multi-View Stereo}
\label{sub-sec:supp:pc}

In our framework, we first extract rooftop point clouds from MVS.
In implementation, we utilize MVSFormer++ and Colmap to extract point clouds.
We now describe the details.

To adapt MVSFormer++ \cite{cao2024mvsformer++} for satellite-based MVS, we constructed a new large-scale dataset comprising 2,600 unique satellite scenes with corresponding ground truth depth maps. 
This dataset was used to train the network specifically for satellite imagery. 
To enhance point cloud quality, we upgraded the feature extraction backbone in MVSFormer++ by replacing the pre-trained DINOv2 \cite{oquab2023dinov2} with the more powerful DINOv3 \cite{simeoni2025dinov3}. 
During inference on real-world multi-view satellite images with their corresponding RPC camera models, we first employ SatelliteSfM \cite{zhang2019leveraging} to obtain approximate pinhole camera projections, which then serve as the input for our trained MVSFormer++.

\subsection{Z-Monotonic SDF Optimization}
\label{sub-sec:supp:zsdf}
Our geometry stage reconstructs a high-fidelity, watertight mesh from an MVS point cloud $P$. We detail the optimization process below, which is applied to 4 spatial divisions of the full scene. The final city model is produced by merging the resulting meshes from each part.

\paragraph{Z-Monotonic SDF Representation}

For each of the 4 scene divisions, we implement the Z-Monotonic SDF by parameterizing a field of monotonic curves on a 2D $(x,y)$ grid. 
This is implemented as a learnable 2D parameter grid of resolution $256 \times 256$.
When querying the SDF value $s(\mathbf{x})$ at a 3D point $\mathbf{x} = (x, y, z)$, we compute the function from Eq.~5. This is achieved by:
\begin{itemize}
    \item Defining a 2D neighborhood of size $3 \times 3$ centered at the query's $(x,y)$ location.
    \item Retrieving the height offset parameters $\{h_j\}$ for this neighborhood by bilinearly sampling from the learnable parameter grid.
    \item Calculating the weights $\{w_j\}$ using a softmax of the inverse $xy$-plane distance to these neighbors.
    \item Computing the final weighted sum $f(z; x, y) = \sum_{j=1}^{n} w_j \cdot \tanh(k \cdot (z - h_j))$, where $k$ is a fixed hyperparameter controlling the curve sharpness, and we set it to $80$.
\end{itemize}
This formulation ensures the SDF is continuous, differentiable, and inherently Z-monotonic.

\paragraph{Differentiable Optimization}
We optimize the learnable parameters of the 2D grid for each scene part. 
We use the Adam optimizer with a learning rate of 0.01 and optimize the grid for 2000 steps. 
At each optimization step, we extract a mesh $M$ from the current SDF using FlexiCubes \cite{shen2023flexible}. 
The FlexiCubes module operates on a grid of resolution $128^3$, enabling gradients to flow from the mesh-based loss back to the 2D grid parameters.
The optimization is supervised by the $\mathcal{L}_{\text{geo}}$ loss function, as defined in Eq.~2:
$$
\mathcal{L}_{\text{geo}} = \sum_{\mathbf{p} \in P} \min_{\mathbf{m} \in M} \|\mathbf{p}_z - \mathbf{m}^*(\mathbf{p})_z\|_1 + \lambda_{\text{Lap}} \mathcal{L}_{\text{Lap}} + \lambda_{\text{Nrm}} \mathcal{L}_{\text{Nrm}}.
$$
We detail the definition for each loss term below.

\subparagraph{Height Supervision ($\sum_{\mathbf{p} \in P} \min_{\mathbf{m} \in M} \|\mathbf{p}_z - \mathbf{m}^*(\mathbf{p})_z\|_1$)}

We supervise the Z-Monotonic SDF optimization using height-based supervision.
To ensure both efficiency and numerical stability, we avoid computing per-point nearest neighbors on the mesh at each iteration.
Instead, we project the MVS point cloud onto a fixed 2D grid and compare it with a rasterized height map rendered from the current mesh.
Given the normalized MVS point cloud $P = \{\mathbf{p}_i\}_{i=1}^{N_P}$ in $[-1,1]^3$, we first convert it into a 2D height map on a regular grid of resolution $R \times R$.
For each point $\mathbf{p}_i = (x_i, y_i, z_i)$, we compute its grid indices
\begin{equation}
    u_i = \left\lfloor \frac{x_i + 1}{2} \, R \right\rfloor, 
    \quad
    v_i = \left\lfloor \frac{y_i + 1}{2} \, R \right\rfloor,
\end{equation}
and accumulate the maximum height per grid cell:
\begin{equation}
    H_{\text{target}}(u,v) = 
    \max \bigl\{\, z_i \;\big|\; (u_i, v_i) = (u, v) \,\bigr\},
\end{equation}
for all $(u,v)$ that receive at least one point.
Cells that do not receive any point are marked as invalid and excluded from the loss via a binary mask.

In parallel, at each optimization step we render a height map $H_{\text{pred}}$ from the current mesh $M$ by rasterizing it onto the same $R \times R$ grid.
We first transform the mesh vertices from normalized coordinates $[-1,1]^3$ to image coordinates:
\begin{equation}
    \tilde{\mathbf{v}} = \bigl( \tfrac{x+1}{2} R,\ \tfrac{y+1}{2} R,\ z \bigr),
\end{equation}
and then use a standard differentiable rasterizer to project the mesh triangles onto the $(u,v)$ plane.
For each pixel $(u,v)$, we compute the interpolated height $z$ via barycentric interpolation of the triangle vertices that cover that pixel.
This yields a dense height map $H_{\text{pred}}(u,v)$ that is fully differentiable with respect to the mesh vertices, and thus with respect to the underlying SDF parameters.

We then define the height-map loss as an $L_1$ difference between the two height maps, computed only on valid pixels where MVS observations exist:
\begin{equation}
    \mathcal{L}_{\text{H}} \;=\;
    \frac{1}{|\Omega|} \sum_{(u,v) \in \Omega}
    \bigl|\, H_{\text{pred}}(u,v) - H_{\text{target}}(u,v) \,\bigr|,
\end{equation}
where $\Omega$ denotes the set of pixels with valid MVS-derived heights.
In implementation, we set $R = 1024$.

\subparagraph{Laplacian Regularization ($\mathcal{L}_{\text{Lap}}$)}
We employ the standard laplacian loss. This loss encourages a smooth mesh surface by penalizing the $L_2$ distance between each vertex $\mathbf{v}_i$ and the uniform average of its 1-ring neighboring vertices $\mathcal{N}(i)$. The loss is defined as:
\begin{equation}
\mathcal{L}_{\text{Lap}} = \frac{1}{|V|} \sum_{\mathbf{v}_i \in V} \left\| \mathbf{v}_i - \frac{1}{|\mathcal{N}(i)|} \sum_{j \in \mathcal{N}(i)} \mathbf{v}_j \right\|_2^2,
\end{equation}
where $V$ is the set of all vertices in the extracted mesh $M$. This helps to reduce high-frequency ``bumpy'' artifacts from the MVS point cloud.

In implementation, we set $\lambda_{\text{Lap}} = 0.5$.

\subparagraph{Normal Consistency ($\mathcal{L}_{\text{Nrm}}$)}
The normal consistency term $\mathcal{L}_{\text{Nrm}}$ is implemented as a Total Variation (TV) loss on the \emph{rendered normal map} $\mathbf{N}$. This encourages smooth changes in surface orientation, which is particularly important for flat facades and ground planes.

At each optimization step, we render a normal map $\mathbf{N}$ from the current mesh $M$. The loss is defined as the sum of the mean $L_2$ norms of the finite differences of the normalized normal vectors $\hat{\mathbf{n}}$ between adjacent pixels, computed in the horizontal ($x$) and vertical ($y$) directions:
\begin{equation}
\mathcal{L}_{\text{Nrm}} = \mathbb{E}_{u,v} \left[ \|\hat{\mathbf{n}}_{u,v} - \hat{\mathbf{n}}_{u+1,v}\|_2 \right] + \mathbb{E}_{u,v} \left[ \|\hat{\mathbf{n}}_{u,v} - \hat{\mathbf{n}}_{u,v+1}\|_2 \right],
\end{equation}
where $\hat{\mathbf{n}}_{u,v}$ is the 3D unit normal vector at pixel $(u,v)$ in the rendered map $\mathbf{N}$. This loss effectively penalizes sharp, local changes in surface normals, promoting piecewise-planar structures.

In implementation, we set $\lambda_{\text{Nrm}} = 0.01$.

\paragraph{FlexiCubes \cite{shen2023flexible} Regularization}

Following \cite{shen2023flexible}, we also add a  regularization term $\lambda_{\text{reg}} \mathcal{L}_{\text{reg}}$ in the geometric optimization, where $\lambda_{\text{reg}} = 0.1$, and $\mathcal{L}_{\text{reg}}$ follows the same definition in \cite{shen2023flexible}.

\paragraph{Final Mesh Generation}
After the optimization converges for all 4 divisions, we extract the final mesh for each part. The vertices are un-normalized to their original world coordinates. These 4 meshes are then merged to produce the single city model. 
Finally, to prepare the mesh for the appearance stage, we compute a single UV atlas for the entire geometry with a resolution of $8192 \times 8192$.

\subsection{Appearance Modeling}
The core of our appearance enhancement is the image restoration network $D$,  a generative model fine-tuned to deterministically map degraded, low-quality renders to sharp, photorealistic targets.

\paragraph{Network Architecture} 
We construct our restorer $D$ by adapting the pre-trained FLUX-Schnell~\cite{flux2024}, a state-of-the-art diffusion transformer based on Rectified Flow (RF). While standard RF formulations model image synthesis as an Ordinary Differential Equation (ODE) initiated from Gaussian noise, this stochastic initialization introduces variance that is detrimental to our texture optimization. Our iterative framework strictly requires a stable, one-to-one mapping between the degraded render and the enhanced output to ensure convergence.

To enforce determinism, we reformulate the generation task as a direct, single-step restoration. Instead of sampling a random noise vector, we encode the degraded render 
$I_{\text{low}}$ into a latent code $z_{\text{low}} = E(I_{\text{low}})$ using the pre-trained VAE encoder. This latent code serves as the deterministic boundary condition for the ODE trajectory. Through fine-tuning, the network learns to map the corrupted latent $z_{\text{low}}$ directly to its clean counterpart $z_{\text{high}}$ in a single function evaluation. This establishes a deterministic mapping $D: z_{\text{low}} \rightarrow z_{\text{high}}$, providing the consistent supervisory signals essential for the stability of our optimization.

\paragraph{Fine-tuning Dataset Construction} To train $D$ effectively, we constructed a specialized dataset of 100{,}000 paired images. 
This dataset was generated from our extensive internal collection of high-quality 3D urban assets, which is completely disjoint from any of our test scenes to prevent data leakage. 
From each 3D asset, we rendered multiple sets of paired images $(I_{\text{low}}, I_{\text{high}})$ from low-altitude (e.g., $400$--$600\,\text{m}$ height) and oblique-angle (e.g., $45^\circ$--$60^\circ$ pitch). 
Each pair consists of a high-resolution, photorealistic rendering $I_{\text{high}}$ serving as the ground-truth target, and a corresponding degraded version $I_{\text{low}}$ rendered from the identical viewpoint but using a reduced Level-of-Detail (LoD). 
This degradation strategy naturally suppresses fine-grained textures and high-frequency details, thereby accurately simulating the coarse inputs encountered during the iterative refinement stage.

\section{Experimental Settings}
\label{sec:exp-settings}

\subsection{Datasets}
\label{sub-sec:datasets}

In the main paper, we evaluate our method on both synthetic and real-world datasets.
We now provide the detailed configurations for each benchmark.

\paragraph{MatrixCity-Satellite}

We build a synthetic satellite benchmark on top of the MatrixCity-Satellite \cite{li2023matrixcity} dataset using the UE5 engine, following the overall protocol described in the main paper.
We now detail our capture settings.

For training views, we adopt a clean and stable environment configuration with:
no fog, no dynamic weather, and no traffic or moving objects.
We place virtual cameras at a fixed altitude of $2000\,\text{m}$ above the ground (the maximal value, to the best of knowledge, that yields stable rendering in UE5), with a near-nadir viewing direction.
Concretely, we set the yaw angle to $89^\circ$ relative to the ground plane, a resolution of $2560 \times 1440$, and a field-of-view (FOV) of $22.42^\circ$ such that the resulting ground sampling distance (GSD) is $0.31\,\text{m/px}$, matching WorldView-3 satellite imagery.

We derive the field-of-view based on the desired ground sampling distance and image resolution.
Let $H$ denote the camera altitude above the ground, $W$ the image width in pixels, and $\theta$ the horizontal FOV.
The ground footprint width $L$ covered by the image is related to the FOV by
\begin{equation}
    L \;=\; 2 H \tan\!\left(\frac{\theta}{2}\right).
\end{equation}
For a target ground sampling distance $\mathrm{GSD}$, the footprint width is also given by
\begin{equation}
    L \;=\; W \cdot \mathrm{GSD}.
\end{equation}
Equating the two expressions for $L$ yields
\begin{equation}
    2 H \tan\!\left(\frac{\theta}{2}\right) \;=\; W \cdot \mathrm{GSD},
\end{equation}
from which the FOV can be solved as
\begin{equation}
    \theta 
    \;=\; 2 \arctan\!\left(\frac{W \cdot \mathrm{GSD}}{2 H}\right).
\end{equation}
Substituting our settings $H = 2000\,\text{m}$, $W = 2560$, and $\mathrm{GSD} = 0.31\,\text{m/px}$, we obtain
\begin{equation}
    \theta 
    \;=\; 2 \arctan\!\left(\frac{2560 \times 0.31}{2 \times 2000}\right)
    \;\approx\; 22.42^\circ,
\end{equation}
which is the FOV used in all our synthetic satellite captures.

To ensure sufficient overlap between neighboring satellite views, following practical experience, we enforce approximately $60\%$ horizontal overlap between adjacent captures.
Given the horizontal footprint width $L \approx 792.7\,\text{m}$ (derived from the FOV and altitude described above), this corresponds to a target stride of
\begin{equation}
    s \;=\; (1 - 0.6)\,L \;\approx\; 0.4 \times 792.7 \;\approx\; 317.1\,\text{m}.
\end{equation}
In practice, we set the sampling distance to $s = 317.44\,\text{m}$, which yields a horizontal overlap very close to $60\%$ between neighboring views.

We then translate the camera centers on a regular grid with this fixed stride along the $x$- and $y$-axes, covering the central $[-500, 500]\,\text{m} \times [-500, 500]\,\text{m}$ area of the scene.
At each grid location, we place two virtual cameras with different off-nadir viewing directions to simulate multi-directional satellite passes.
Concretely, we use two sets of rotations:
\begin{equation}
    ( \mathrm{pitch}, \mathrm{yaw}, \mathrm{roll} ) 
    \;=\; (0^\circ, 89^\circ, 0^\circ)
    \quad\text{and}\quad
    (0^\circ, 89^\circ, 90^\circ),
\end{equation}
corresponding to south-looking and west-looking near-nadir views, respectively.
This configuration results in a sparse multi-view satellite observation pattern that closely approximates real-world acquisition geometries.

For ground-truth point clouds, we use the point clouds provided by MatrixCity-Satellite and select the central $[-400, 400]\,\text{m}^2$ region for evaluation.
For test views, we follow the default \textit{MatrixCity-Aerial} evaluation protocol and extend it to two altitudes to assess multi-scale performance.
We place cameras at heights of $200\,\text{m}$ and $500\,\text{m}$, with a resolution of $1920\times1080$, FOV of $45^\circ$, and a pitch angle of $45^\circ$ toward the ground.
Camera centers are sampled on a uniform grid over the central $[-200, 200]\,\text{m}^2$ region with a fixed sampling interval ($45.01 \, \text{m}$) along both axes.
This yields $72$ test views that cover the evaluation area with sufficient angular diversity while avoiding boundary regions where the reconstruction quality naturally degrades.
These views are used for novel view synthesis evaluation in the main paper.

\paragraph{DFC 2019}

For the DFC 2019 benchmark, we follow the overall evaluation protocol of Skyfall-GS~\cite{lee2025skyfall} in terms of AOI selection and input modality.
We obtain approximate pinhole camera intrinsics and extrinsics from the provided RPC parameters using SatelliteSfM~\cite{zhang2019leveraging}, and use these cameras both for all competing methods and for rendering our results.

To obtain ground-level reference views, we follow Skyfall-GS~\cite{lee2025skyfall} and use video sequences from Google Earth Studio (GES) along a low-altitude orbital trajectory around each AOI as test views.
In practice, directly using the camera trajectories of~\cite{lee2025skyfall} may lead to invalid pixels near the image boundaries (\eg, black borders).
To ensure a fair and robust evaluation, we additionally apply a simple binary mask that removes boundary pixels before computing the image-based metrics.

\paragraph{GoogleEarth}

For the GoogleEarth benchmark, we also follow the evaluation protocol of Skyfall-GS~\cite{lee2025skyfall} in terms of AOI selection, input data, and camera trajectories.
For each AOI, we use the training and test views from Google Earth Studio, matching the satellite-like input configuration of~\cite{lee2025skyfall}.
Similar to the DFC 2019 case, these renders may contain invalid pixels near the image borders.
We therefore apply the same binary boundary mask as in the DFC 2019 evaluation.

In addition, we empirically observe a systematic misalignment in the NYC\_004 scene. 
The camera parameters provided in \cite{lee2025skyfall} do not accurately reproduce the released GES images, and this discrepancy cannot be resolved by simple refinement.
To avoid introducing bias in the quantitative comparison, we exclude NYC\_004 from our metric reporting and only include NYC\_010, NYC\_219, and NYC\_336 in the main paper.

\subsection{Baselines}
\label{sub-sec:baselines}

We compare our method against four representative baselines, as described in the main paper: Mip-Splatting~\cite{Yu2024MipSplatting}, 2DGS~\cite{huang20242d}, CityGS-X~\cite{gao2025citygs}, and Skyfall-GS~\cite{lee2025skyfall}.

For \cite{lee2025skyfall}, we use the official implementation and default hyperparameters recommended by the authors for satellite-based reconstruction.
Specifically, we evaluate the MatrixCity-Satellite dataset with the recommended DFC 2019 hyperparameters, and to match the scale of DFC 2019 scenes, we uniformly scale up the MatrixCity-Satellite scene by a factor of $50 \times$ before running Skyfall-GS.

For the other baselines, directly adopting their default hyperparameters (which are typically designed for dense aerial or street-view imagery) leads to numerical instability or divergence when applied to our extreme off-nadir satellite setting.
To ensure a fair and stable comparison, we carefully re-tune a small subset of hyperparameters while keeping the overall model architectures unchanged. 
In particular, for \cite{Yu2024MipSplatting,huang20242d,gao2025citygs}, we reduce the initial position learning rate from $1.6\times 10^{-4}$ to $1.6\times 10^{-5}$, decrease the densification ratio from $0.01$ to $0.001$, slightly increase the densification gradient threshold from $2.0\times 10^{-4}$ to $4.0\times 10^{-4}$, and disable frequent opacity resets. 
In addition, we extend the far plane from $100.0$ to $1.0\times 10^{8}$ and forward the principal point $(c_x, c_y)$ estimated by COLMAP into the projection matrix, so that these baselines can handle our kilometer-scale, satellite-like scenes without modifying their core models.

\subsection{Metrics}
\label{sub-sec:metrics}

For PSNR and SSIM, we follow standard image quality evaluation protocols.
For LPIPS, we adopt the official implementation with the AlexNet backbone, as commonly done in prior work.

For geometric evaluation, we report the Chamfer distance (CD) and the Precision/Recall/F1 metrics between the reconstructed and ground-truth point clouds, following \cite{gao2025citygs,liu2024citygaussianv2,huang20242d}.
The Precision/Recall/F1 metrics are adopted in \cite{gao2025citygs,liu2024citygaussianv2}. 
For the distance threshold $d_\tau$ for aerial reconstructions, it is typically set to $0.006$ in normalized scene units, which corresponds to an effective ground sampling distance (GSD) of about $5\,\text{cm/px}$ in aerial settings. 
In our satellite setting, however, the GSD is approximately $31\,\text{cm/px}$, \ie, about six times coarser than in the aerial case. 
To maintain a comparable geometric tolerance in physical units, we therefore scale the threshold proportionally and set $d_\tau = 0.036$ which yields a consistent relative matching criterion under our satellite imaging configuration.

For point cloud extraction, we follow the native pipelines of each method whenever possible: for \cite{huang20242d,gao2025citygs} we directly use their official mesh export routines, while for \cite{Yu2024MipSplatting,lee2025skyfall} we convert the optimized 3D Gaussian representations into meshes using the SOTA method \cite{A_G_Stuart_2025_ICCV}. 
We then uniformly sample points on all resulting meshes following the protocol of \cite{gao2025citygs} to obtain the predicted point clouds used in our geometric evaluation.

\section{More Results}
\label{sec:more-results}

\subsection{Quantitative Comparison}

\begin{table}[p]
\centering
\setlength{\tabcolsep}{18pt}
\begin{spacing}{1.0} 
\begin{tabular}{ccccc}
\toprule
\textbf{Scene} & \textbf{Method} & \textbf{PSNR $\uparrow$} & \textbf{SSIM $\uparrow$} & \textbf{LPIPS $\downarrow$} \\
\midrule
\multirow{6}{*}{\textbf{JAX\_004}} & Mip-Splatting & 13.124 & 0.262 & 0.868 \\
 & 2DGS & 6.958 & 0.209 & 0.906 \\
 & CityGS-X & FAIL & FAIL & FAIL \\
 & Skyfall-GS & 12.955 & 0.248 & 0.790 \\
 & Ours w/o Image Restoration   Network & 13.524 & 0.267 & 0.664 \\
 & Ours & 13.857 & 0.294 & 0.606 \\
\midrule
\multirow{6}{*}{\textbf{JAX\_068}} & Mip-Splatting & 7.950 & 0.314 & 0.834 \\
 & 2DGS & 8.079 & 0.300 & 0.839 \\
 & CityGS-X & FAIL & FAIL & FAIL \\
 & Skyfall-GS & 11.855 & 0.300 & 0.762 \\
 & Ours w/o Image Restoration   Network & 12.824 & 0.348 & 0.576 \\
 & Ours & 12.935 & 0.341 & 0.554 \\
\midrule
\multirow{6}{*}{\textbf{JAX\_214}} & Mip-Splatting & 9.003 & 0.404 & 0.766 \\
 & 2DGS & 6.794 & 0.357 & 0.813 \\
 & CityGS-X & FAIL & FAIL & FAIL \\
 & Skyfall-GS & 12.318 & 0.398 & 0.696 \\
 & Ours w/o Image Restoration   Network & 12.566 & 0.386 & 0.582 \\
 & Ours & 12.467 & 0.397 & 0.539 \\
\midrule
\multirow{6}{*}{\textbf{JAX\_260}} & Mip-Splatting & 11.078 & 0.402 & 0.795 \\
 & 2DGS & 7.633 & 0.349 & 0.832 \\
 & CityGS-X & FAIL & FAIL & FAIL \\
 & Skyfall-GS & 12.713 & 0.371 & 0.715 \\
 & Ours w/o Image Restoration   Network & 12.793 & 0.373 & 0.576 \\
 & Ours & 12.977 & 0.400 & 0.526 \\
\bottomrule
\end{tabular}
\end{spacing}
\caption{Quantitative evaluation of our method compared to prior works on every scene of \textbf{DFC 2019} datasets.}
\label{tab:supp:dfc2019}
\end{table}

\begin{table}[p]
\centering
\setlength{\tabcolsep}{18pt}
\begin{spacing}{1.0}
\begin{tabular}{ccccc}
\toprule
\textbf{Scene} & \textbf{Method} & \textbf{PSNR $\uparrow$} & \textbf{SSIM $\uparrow$} & \textbf{LPIPS $\downarrow$} \\
\midrule
\multirow{6}{*}{\textbf{NYC\_336}} & Mip-Splatting & 13.389 & 0.453 & 0.597 \\
 & 2DGS & 11.014 & 0.311 & 0.681 \\
 & CityGS-X & 14.143 & 0.393 & 0.661 \\
 & Skyfall-GS & 13.497 & 0.420 & 0.507 \\
 & Ours w/o Image Restoration   Network & 13.512 & 0.434 & 0.491 \\
 & Ours & 13.894 & 0.436 & 0.496 \\
\midrule
\multirow{6}{*}{\textbf{NYC\_010}} & Mip-Splatting & 12.022 & 0.148 & 0.533 \\
 & 2DGS & 10.899 & 0.151 & 0.642 \\
 & CityGS-X & 12.151 & 0.140 & 0.544 \\
 & Skyfall-GS & 12.184 & 0.142 & 0.531 \\
 & Ours w/o Image Restoration   Network & 12.007 & 0.164 & 0.538 \\
 & Ours & 12.513 & 0.162 & 0.577 \\
\midrule
\multirow{6}{*}{\textbf{NYC\_219}} & Mip-Splatting & 11.230 & 0.135 & 0.524 \\
 & 2DGS & 11.151 & 0.129 & 0.542 \\
 & CityGS-X & 11.727 & 0.135 & 0.569 \\
 & Skyfall-GS & 11.686 & 0.135 & 0.526 \\
 & Ours w/o Image Restoration   Network & 11.474 & 0.161 & 0.534 \\
 & Ours & 11.901 & 0.160 & 0.566 \\
\bottomrule
\end{tabular}
\end{spacing}
\caption{Quantitative evaluation of our method compared to prior works on every scene of \textbf{GoogleEarth} datasets.}
\label{tab:supp:ges}
\end{table}

\cref{tab:supp:dfc2019,tab:supp:ges} present per-scene quantitative comparisons on the DFC 2019 and GoogleEarth datasets, respectively.
Across all AOIs, our method consistently matches or surpasses state-of-the-art baselines in PSNR and SSIM, while achieving competitive or better LPIPS.
Notably, the ``Ours w/o Image Restoration Network'' variant already improves over most baselines, and further gains are obtained by our full appearance modeling pipeline, highlighting the contribution of our deterministic restoration network.

\subsection{Qualitative Comparison with Ground-Truth Views}

\begin{figure*}[p]
	\centering
    \begin{spacing}{1} 
    \setlength\tabcolsep{1pt}
    \begin{tabular}{cccccc}
    
    \imagecell[0.16]{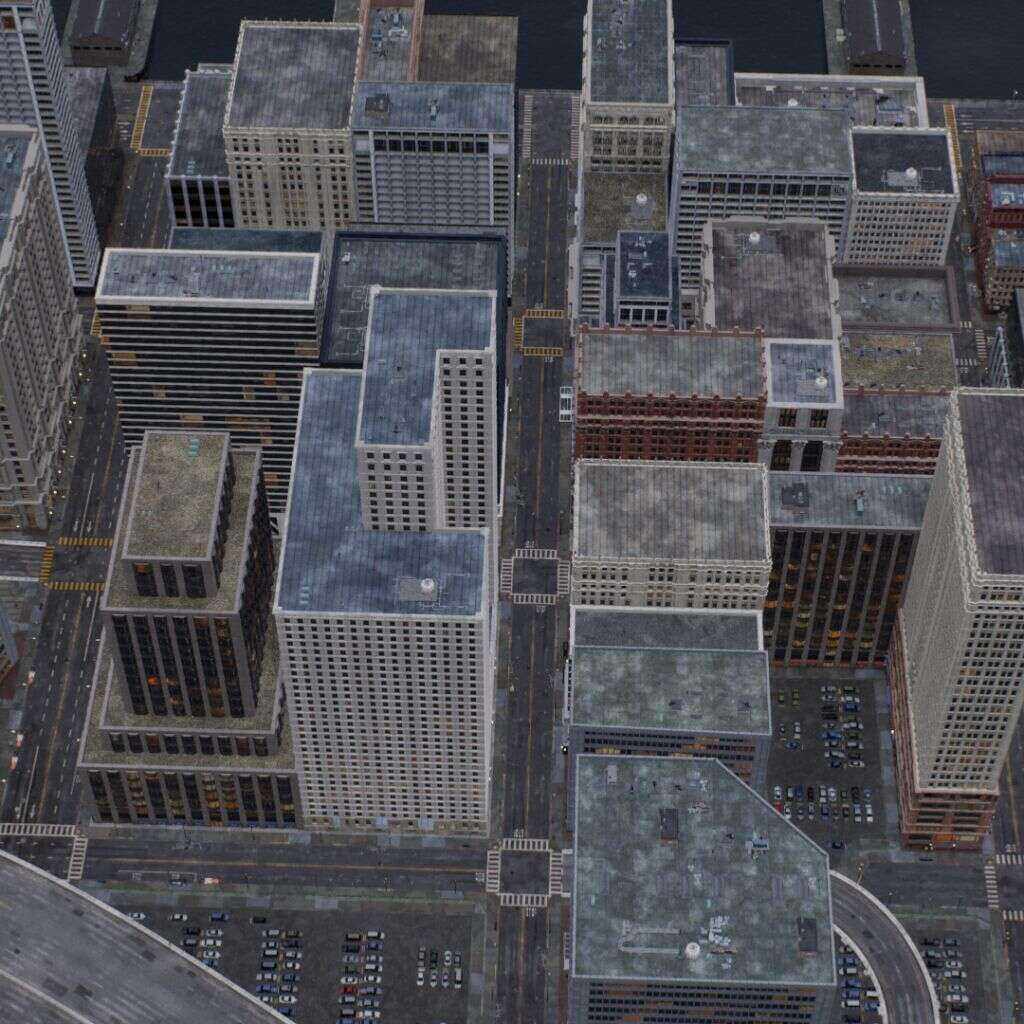} & 
    \imagecell[0.16]{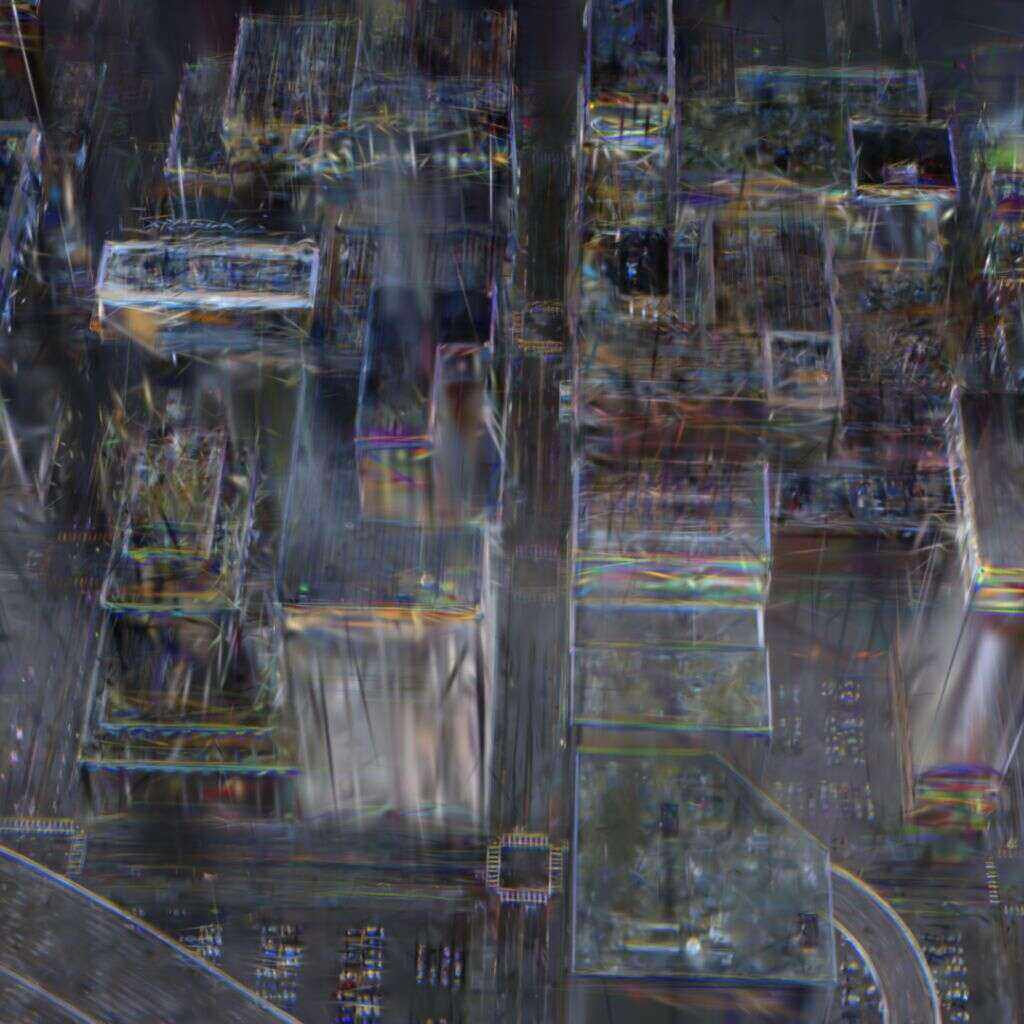} & 
    \imagecell[0.16]{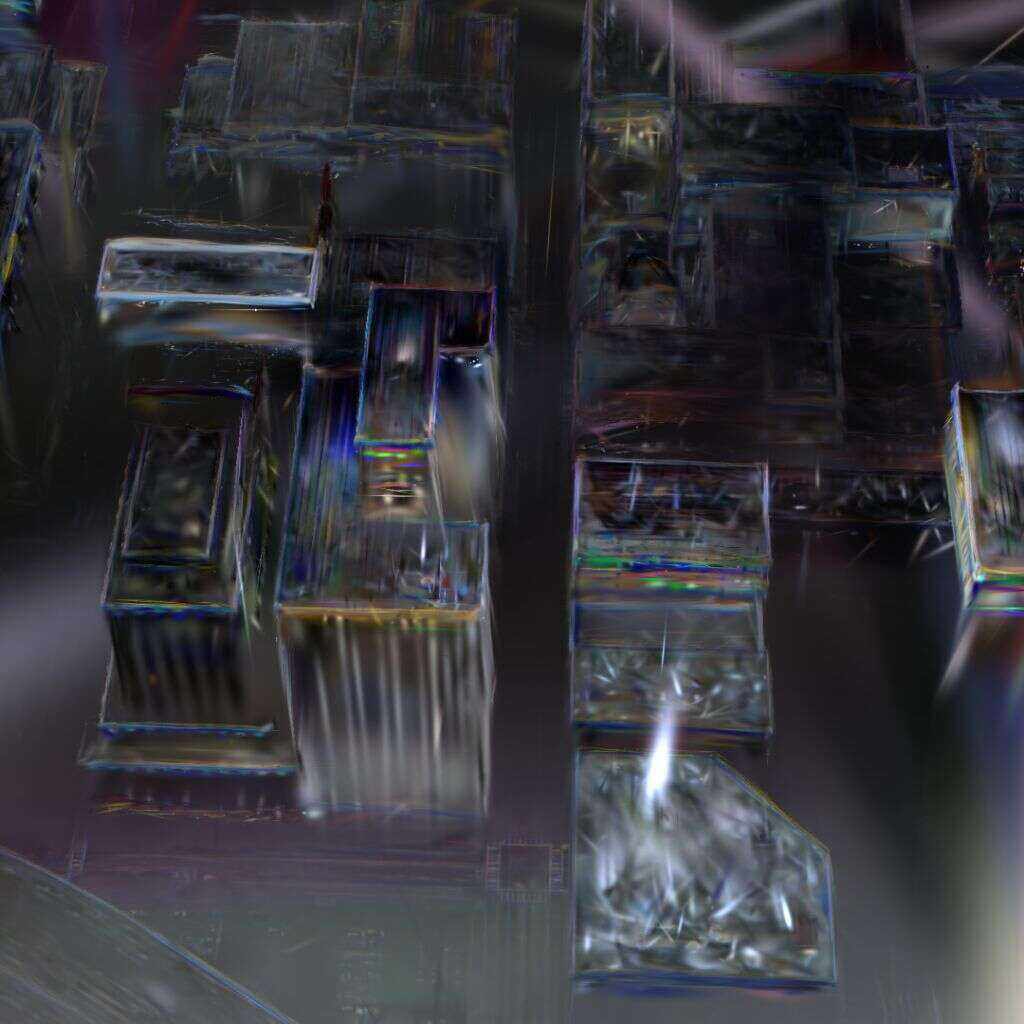} & 
    \imagecell[0.16]{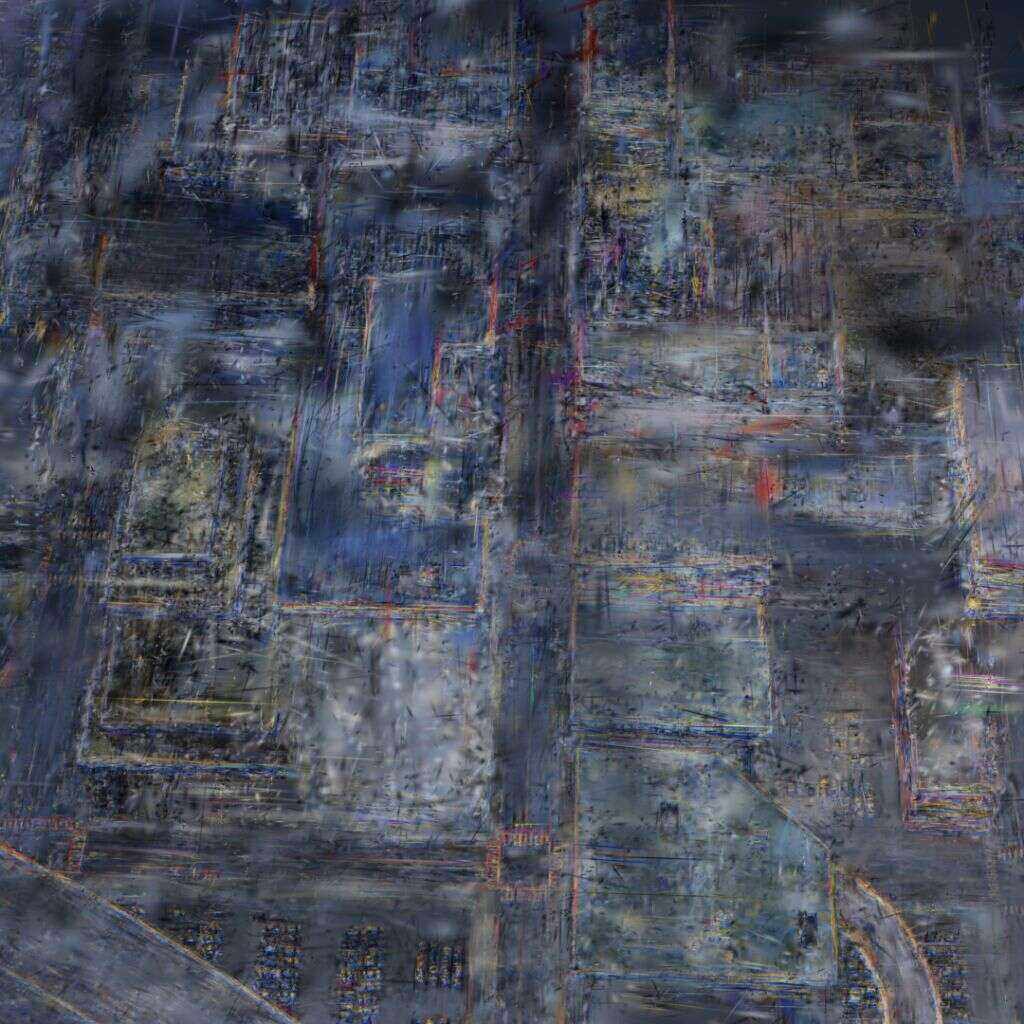} & 
    \imagecell[0.16]{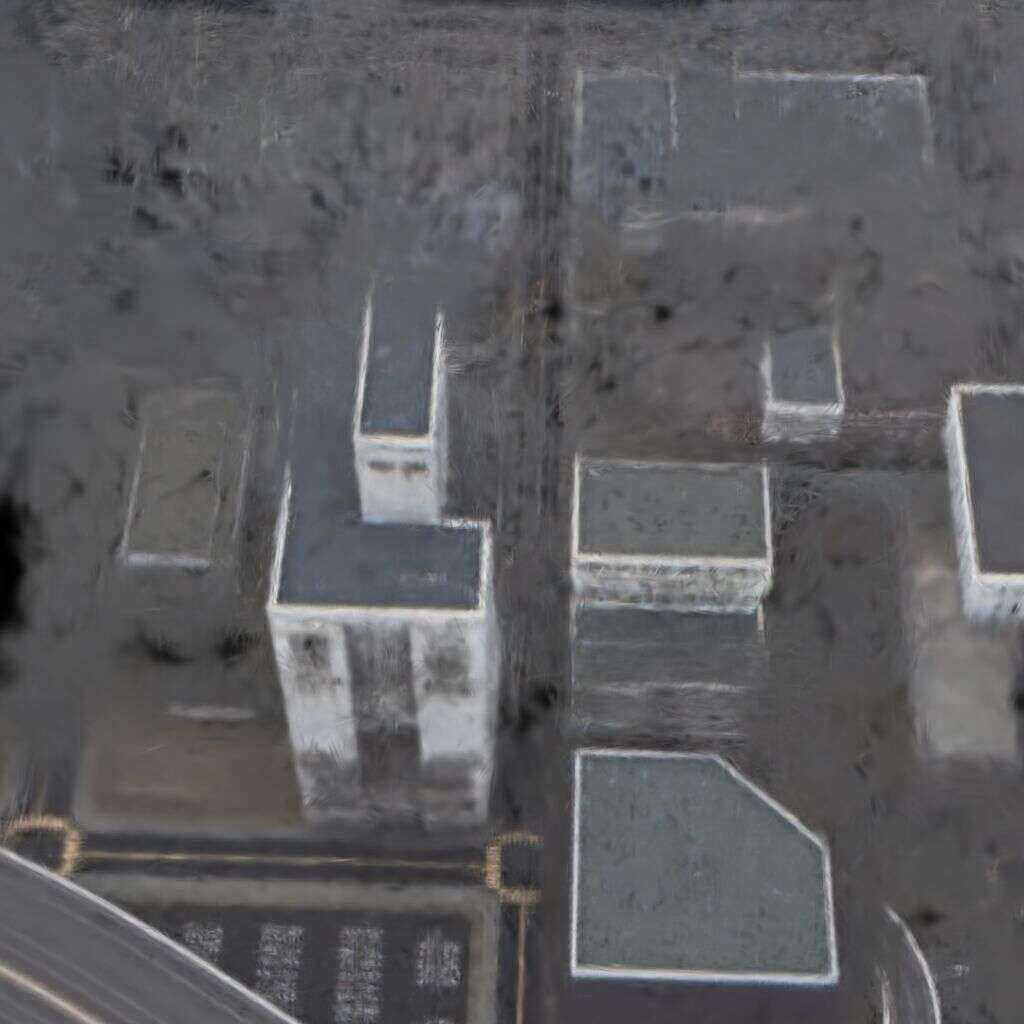} & 
    \imagecell[0.16]{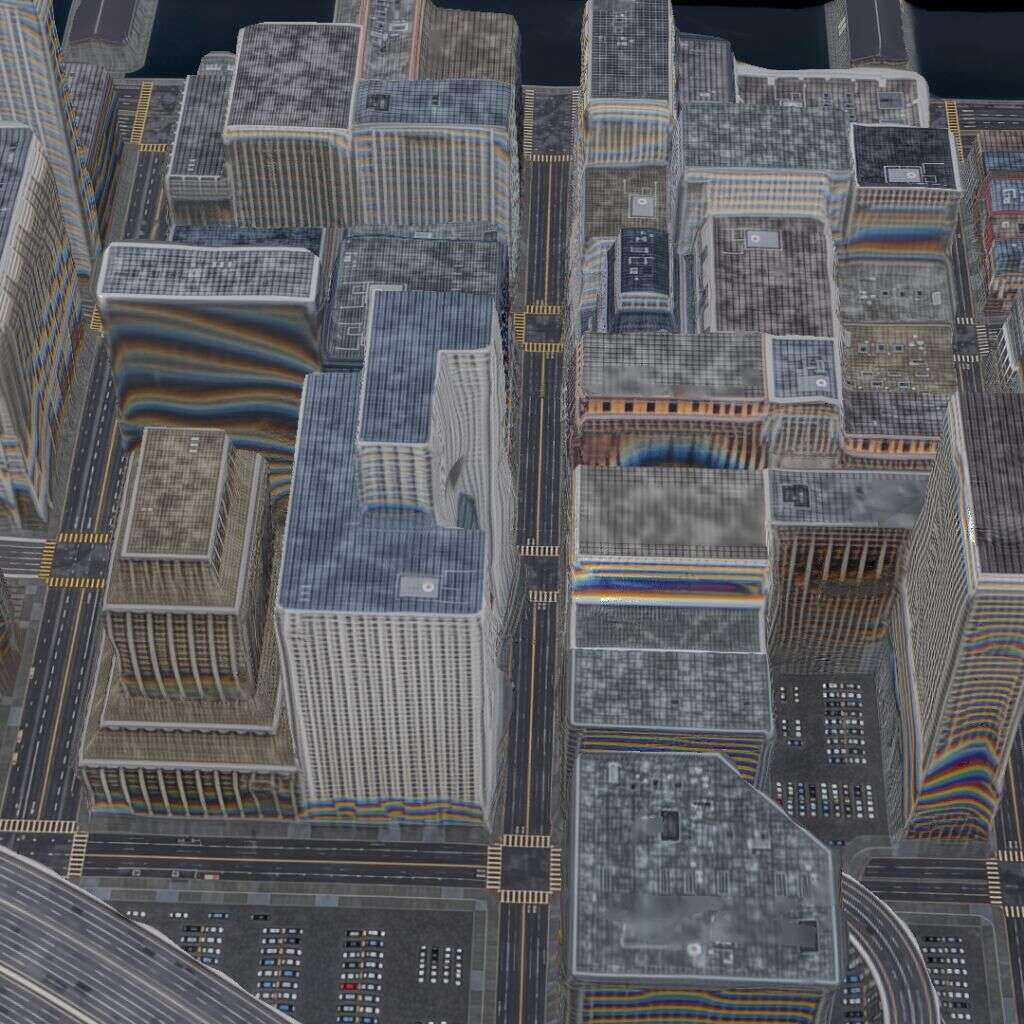} \\
    \vspace*{-10pt} \\   

    \imagecell[0.16]{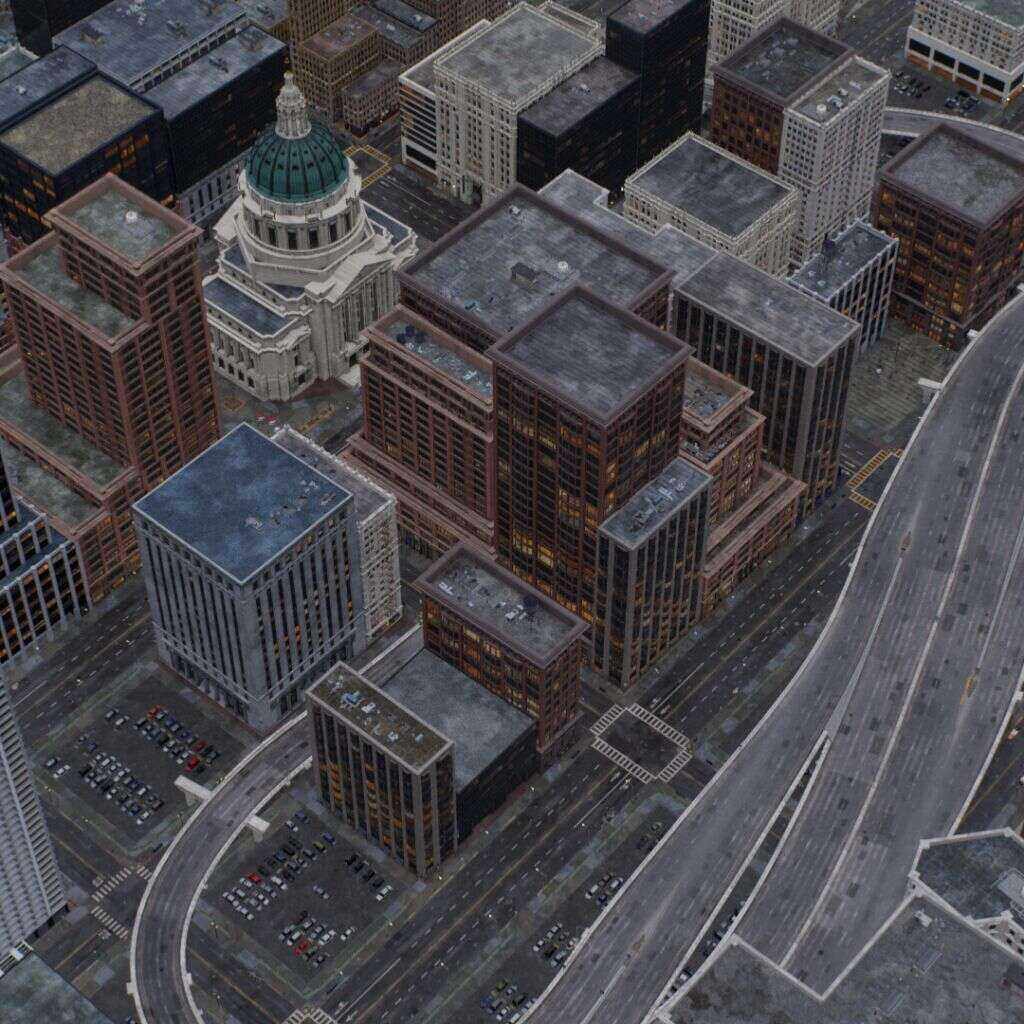} & 
    \imagecell[0.16]{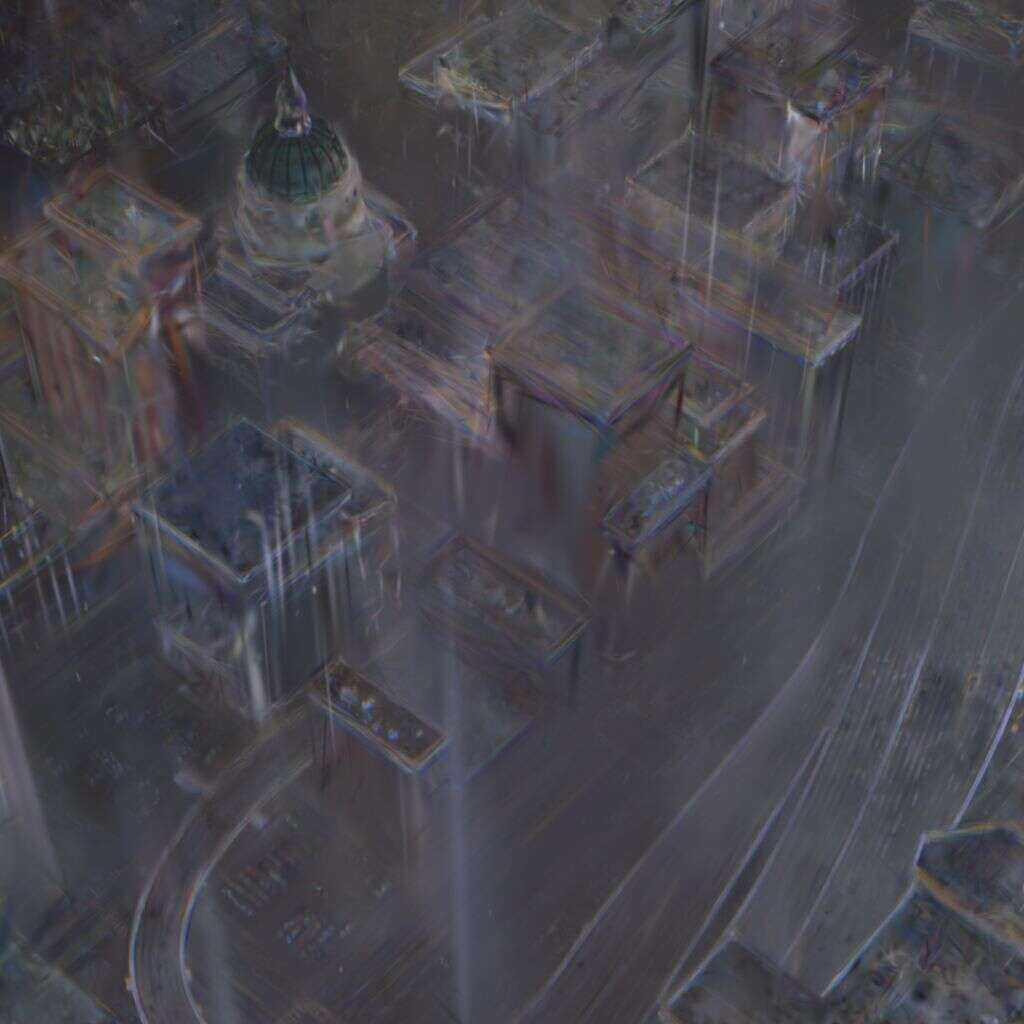} & 
    \imagecell[0.16]{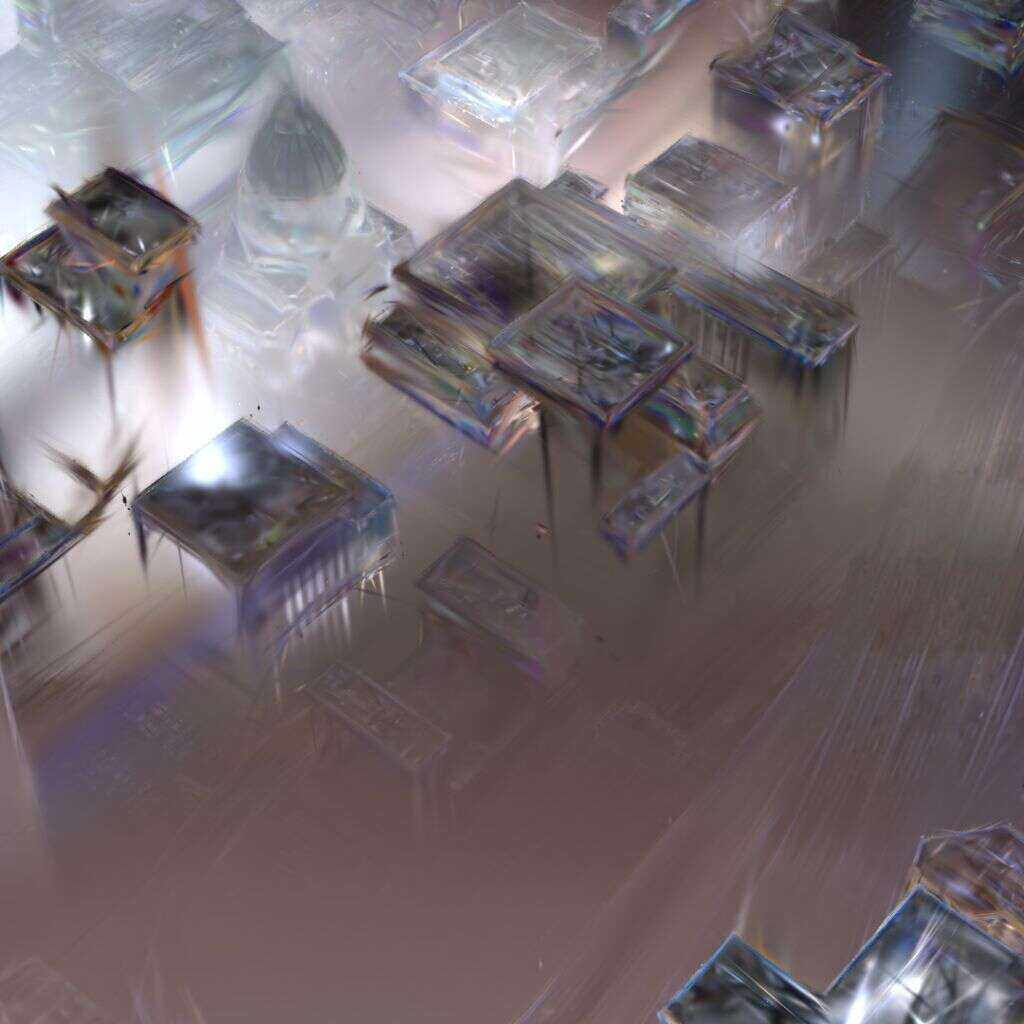} & 
    \imagecell[0.16]{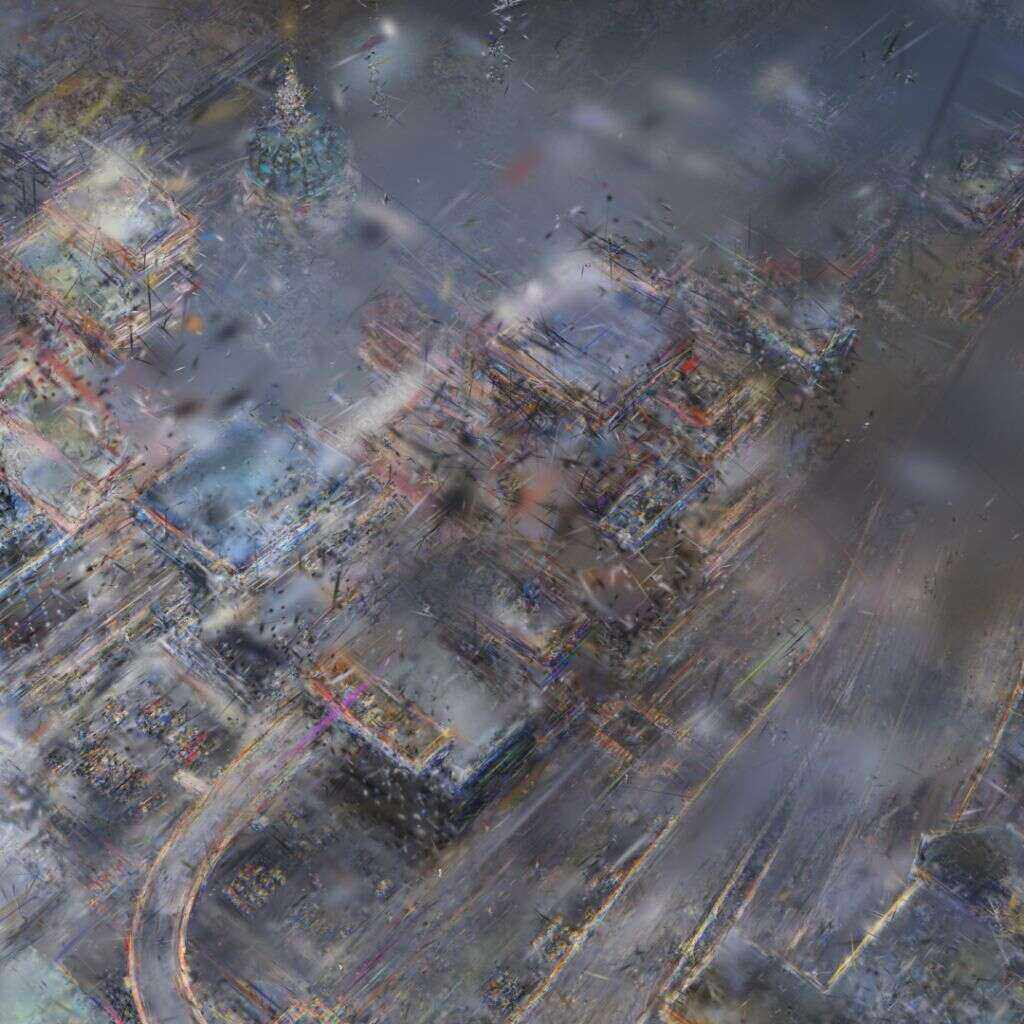} & 
    \imagecell[0.16]{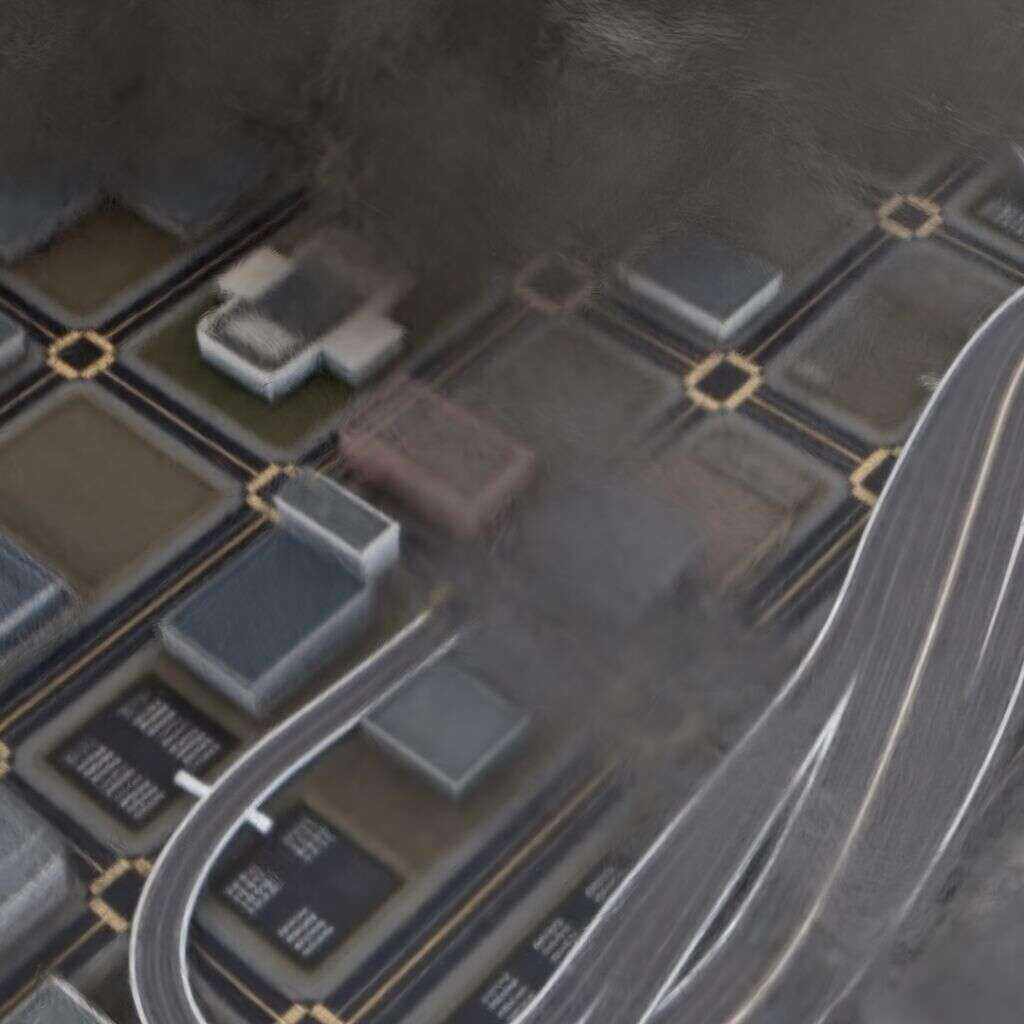} & 
    \imagecell[0.16]{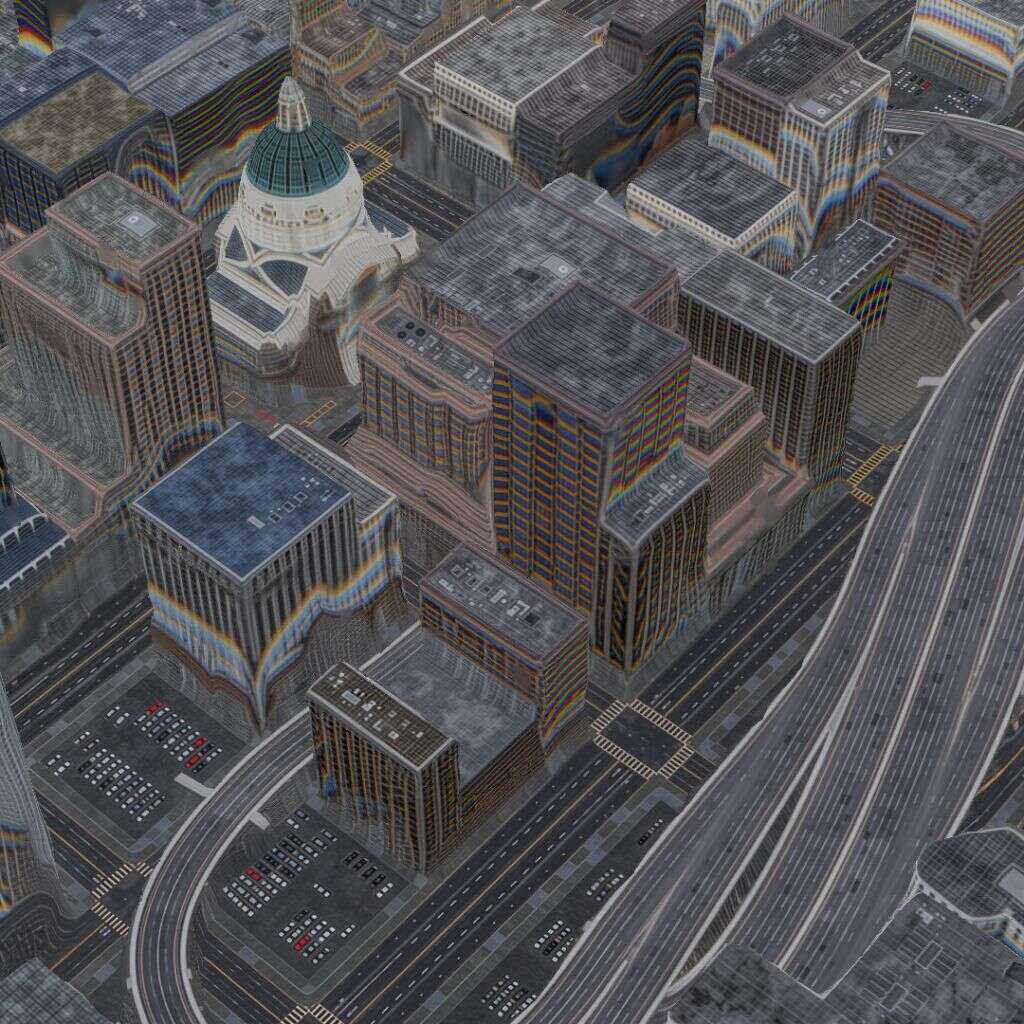} \\
    \vspace*{-10pt} \\

    \imagecell[0.16]{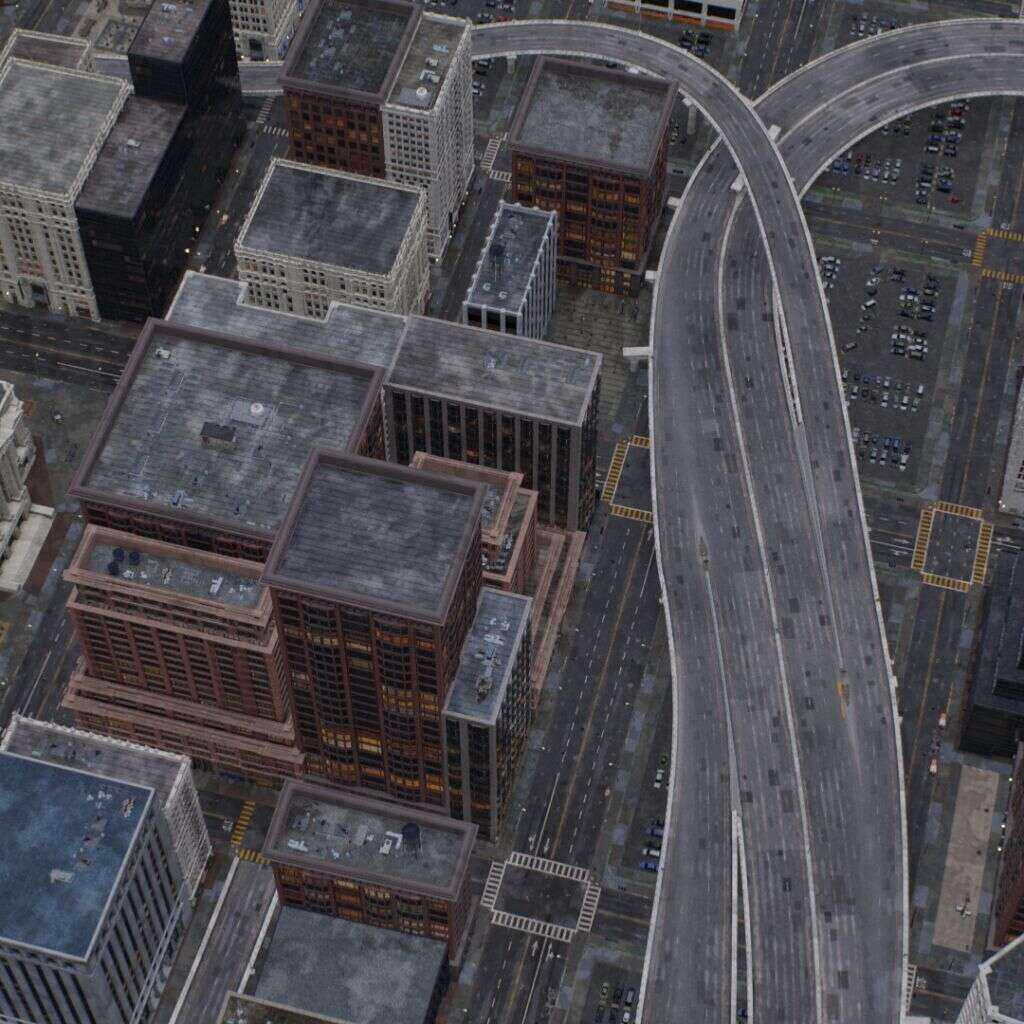} & 
    \imagecell[0.16]{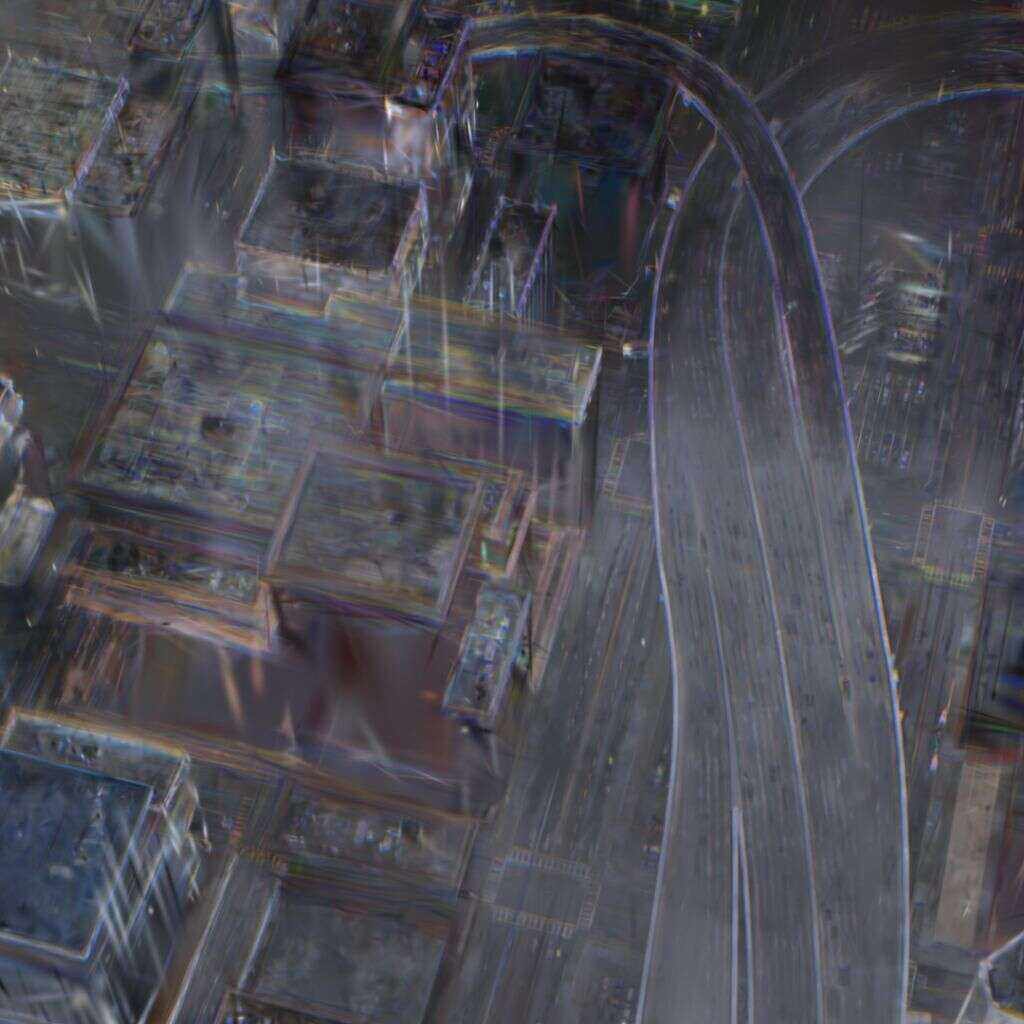} & 
    \imagecell[0.16]{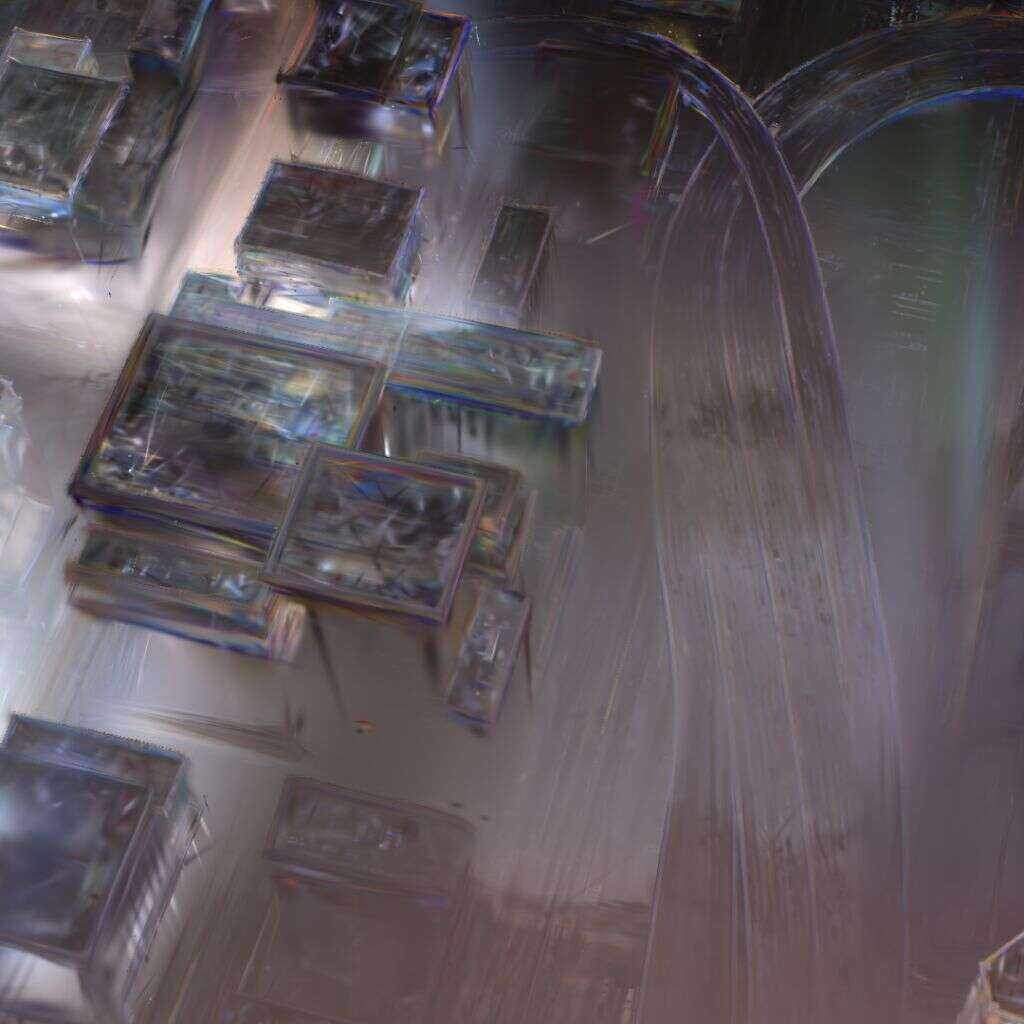} & 
    \imagecell[0.16]{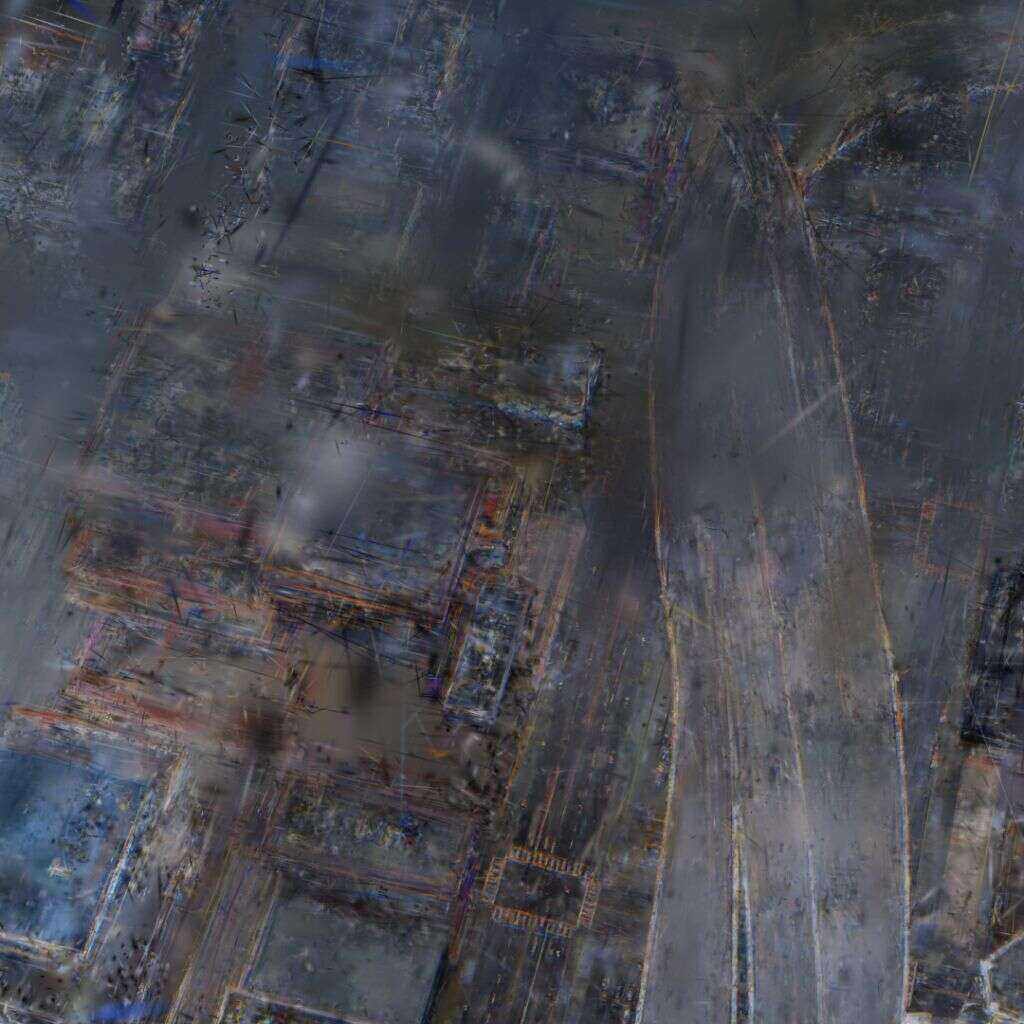} & 
    \imagecell[0.16]{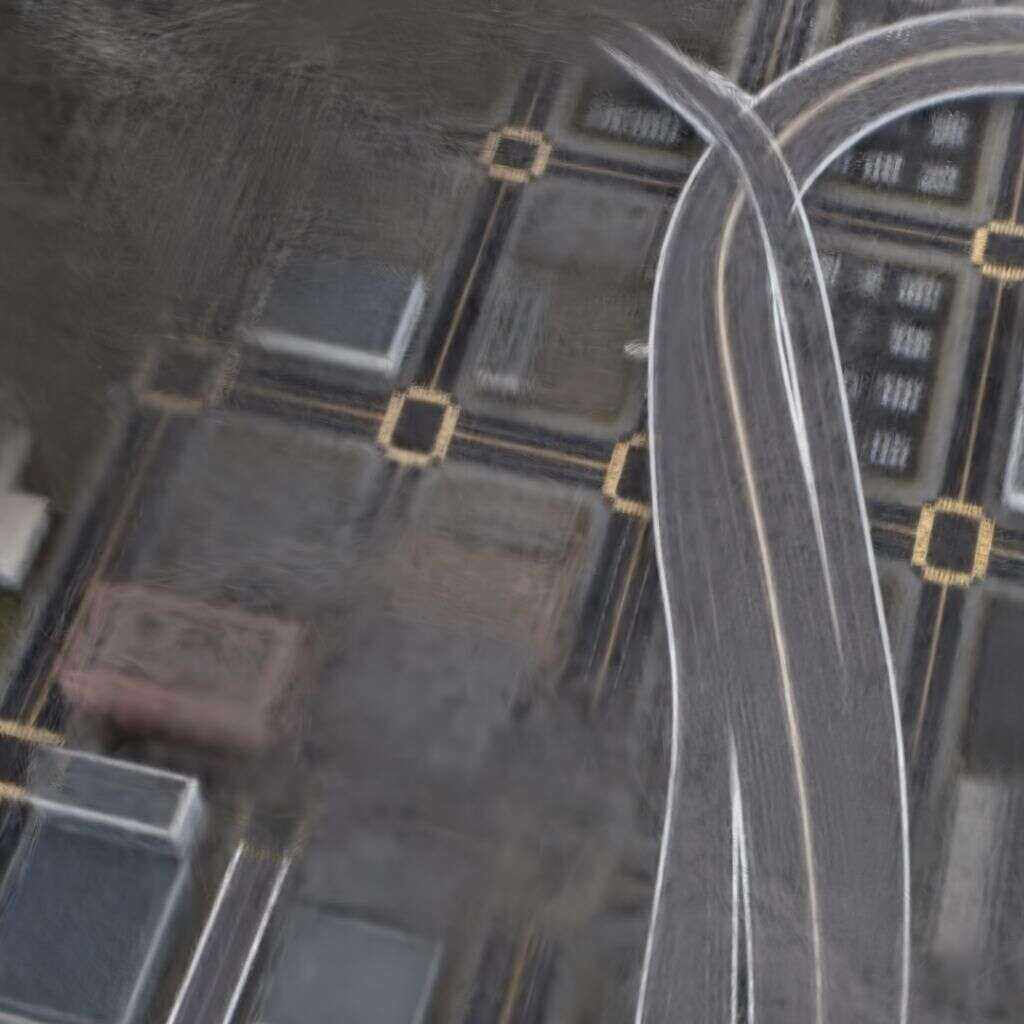} & 
    \imagecell[0.16]{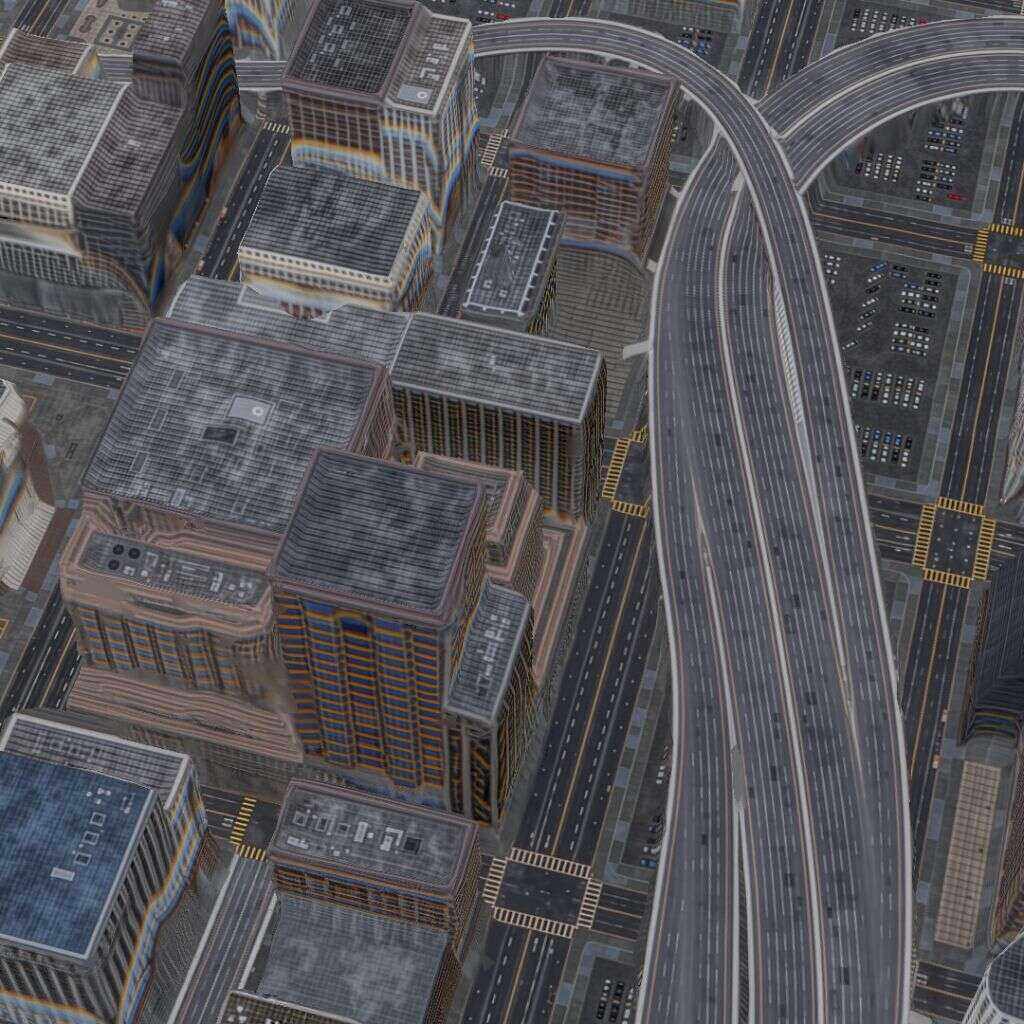} \\
    \vspace*{-10pt} \\
    
    \\
    \vspace*{-20pt}
    \\
    G.T. & 
    Mip-Splatting & 
    2DGS & 
    CityGS-X &
    Skyfall-GS & 
    Ours \\
    
    \end{tabular}
    \end{spacing}
	\caption{ 
    \textbf{Results of the MatrixCity-Satellite scene}. 
    Compared to baselines, our method successfully achieves high-quality city reconstruction from satellite imagery. 
    }
    \label{fig:supp:qual-mc-far}
    \vspace*{-0.3cm}
\end{figure*}

\begin{figure*}[p]
	\centering
    \begin{spacing}{1} 
    \setlength\tabcolsep{1pt}
    \begin{tabular}{ccccc}
    
    \imagecell[0.18]{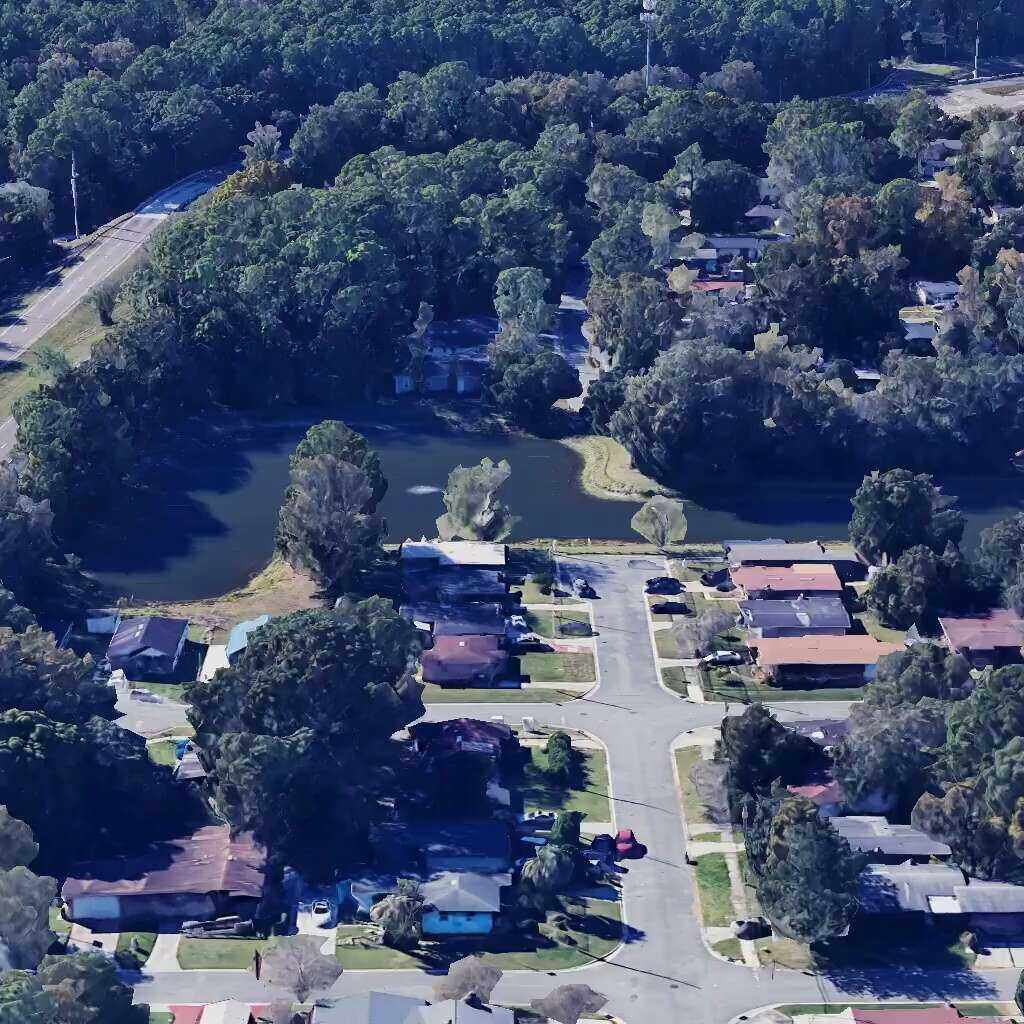} & 
    \imagecell[0.18]{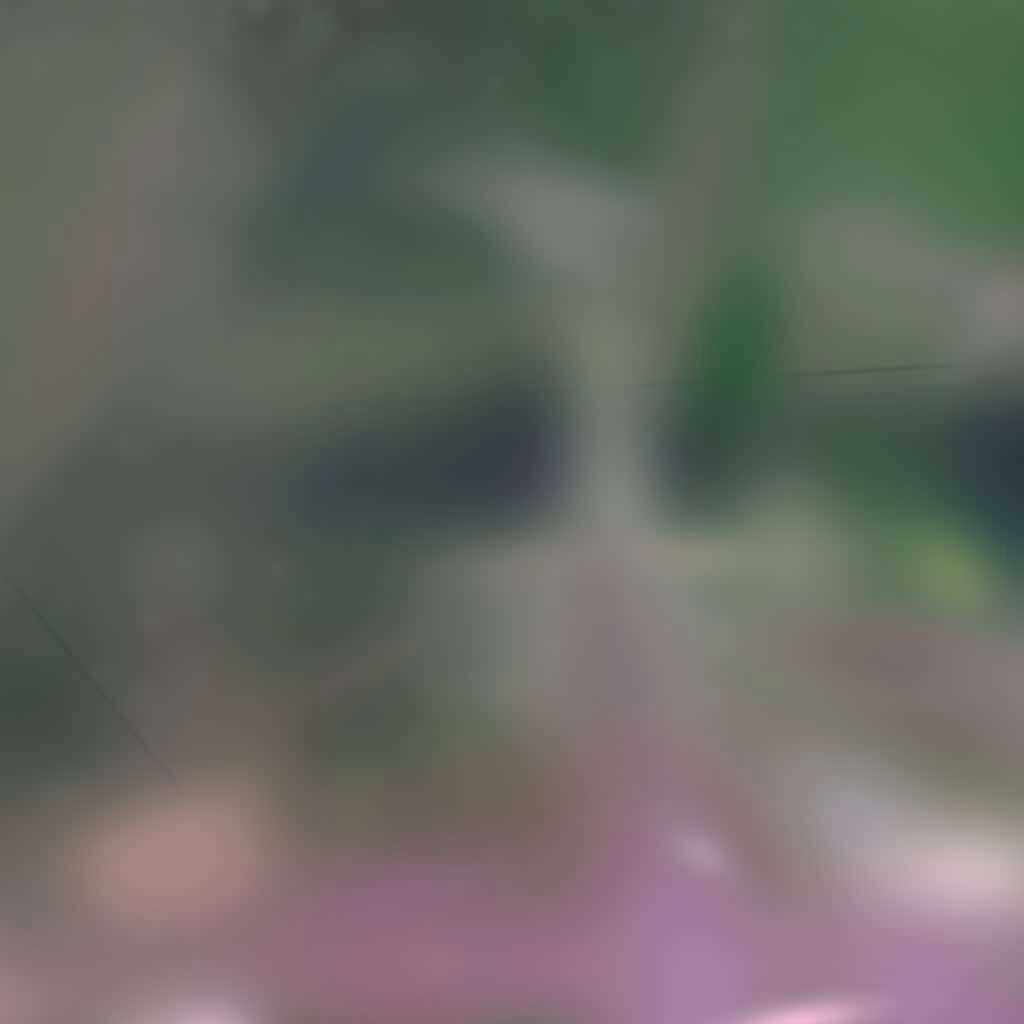} & 
    \imagecell[0.18]{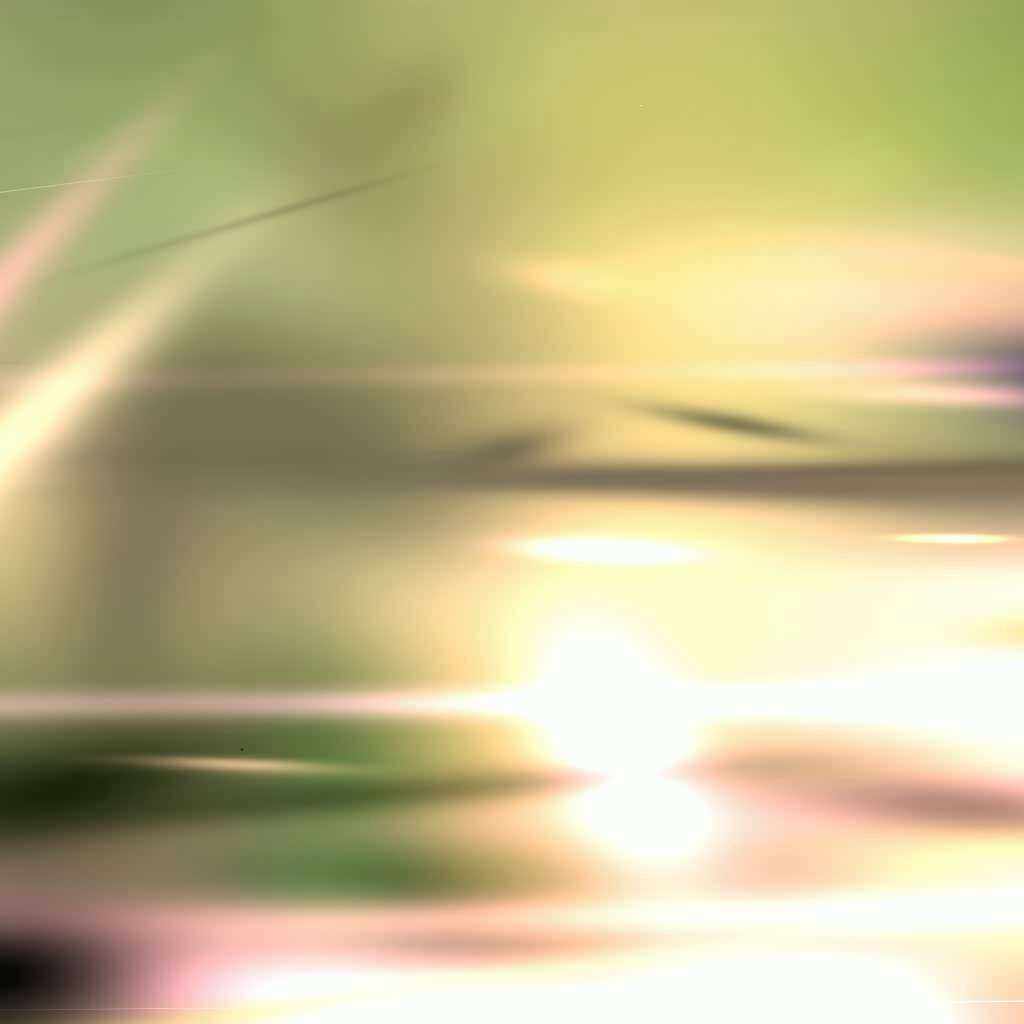} & 
    \imagecell[0.18]{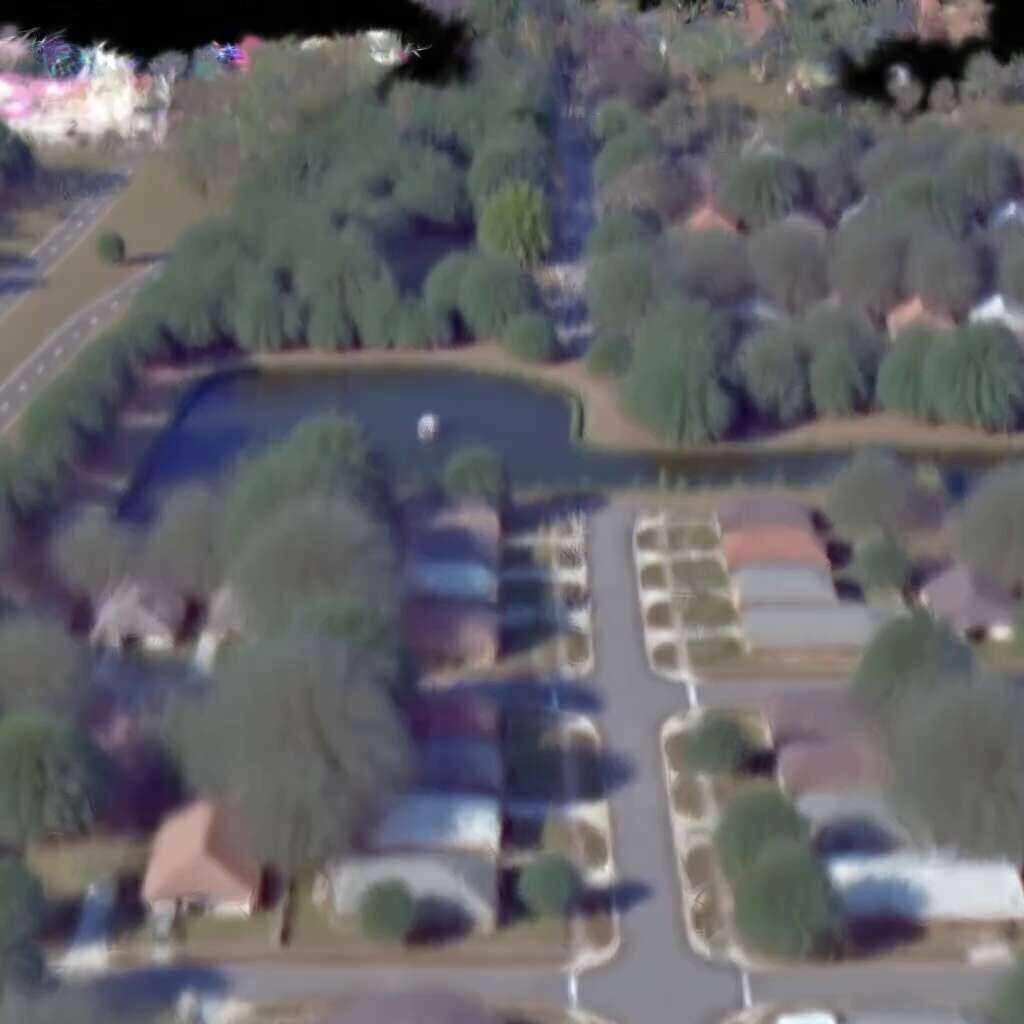} & 
    \imagecell[0.18]{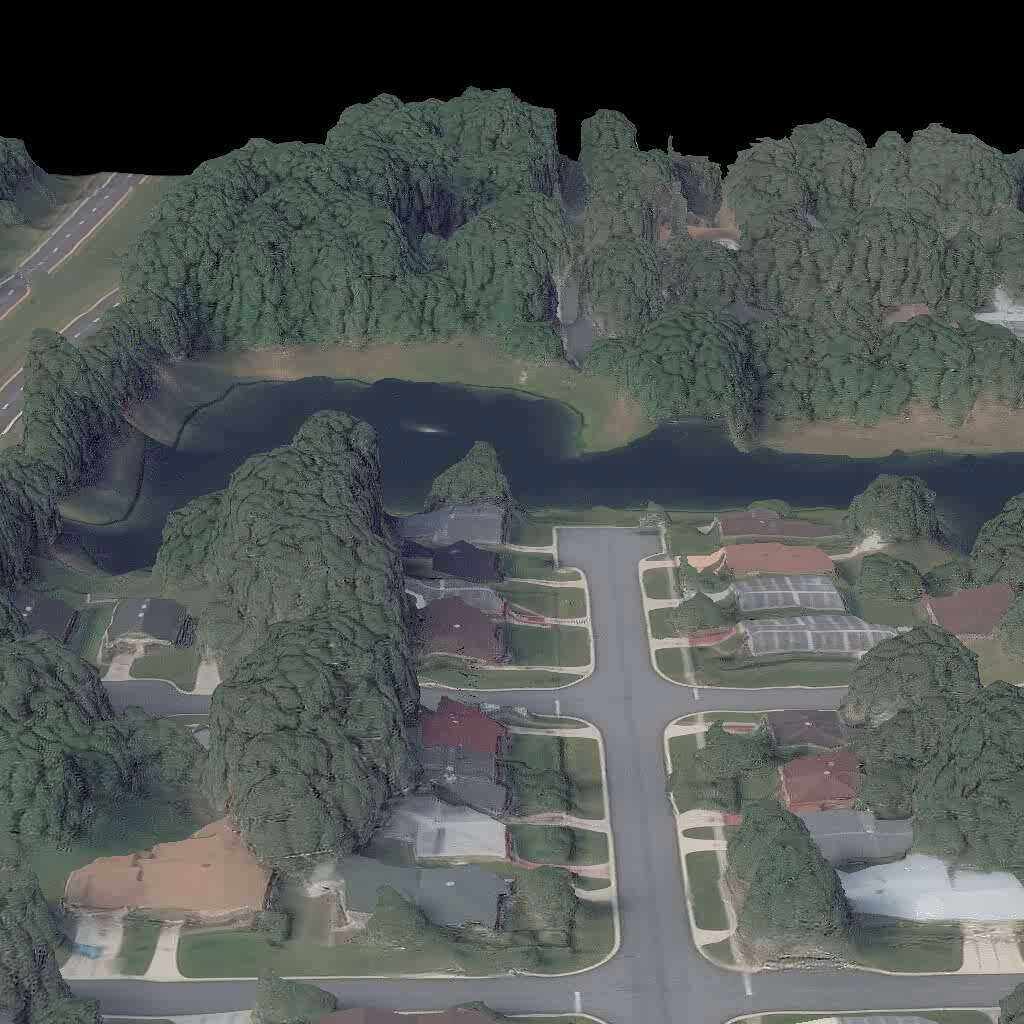} \\
    \vspace*{-10pt} \\  

    \imagecell[0.18]{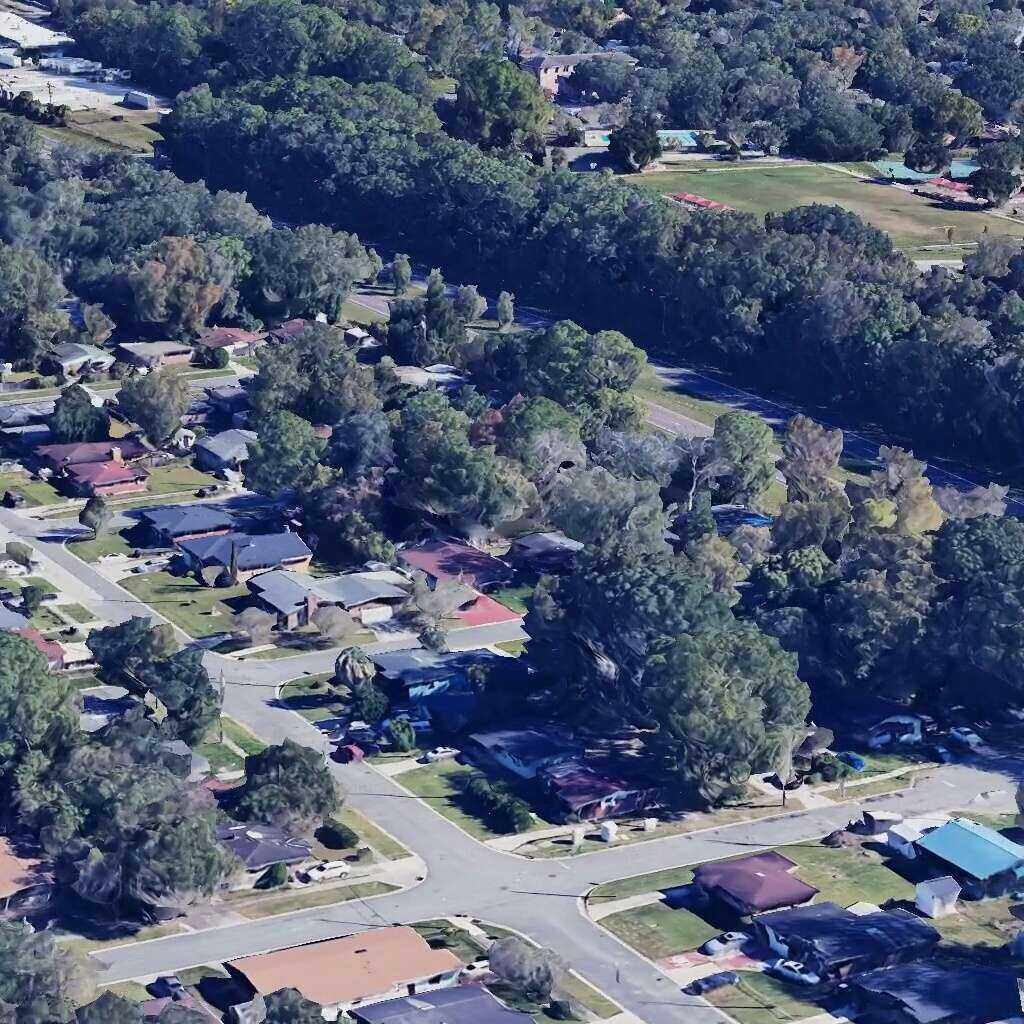} & 
    \imagecell[0.18]{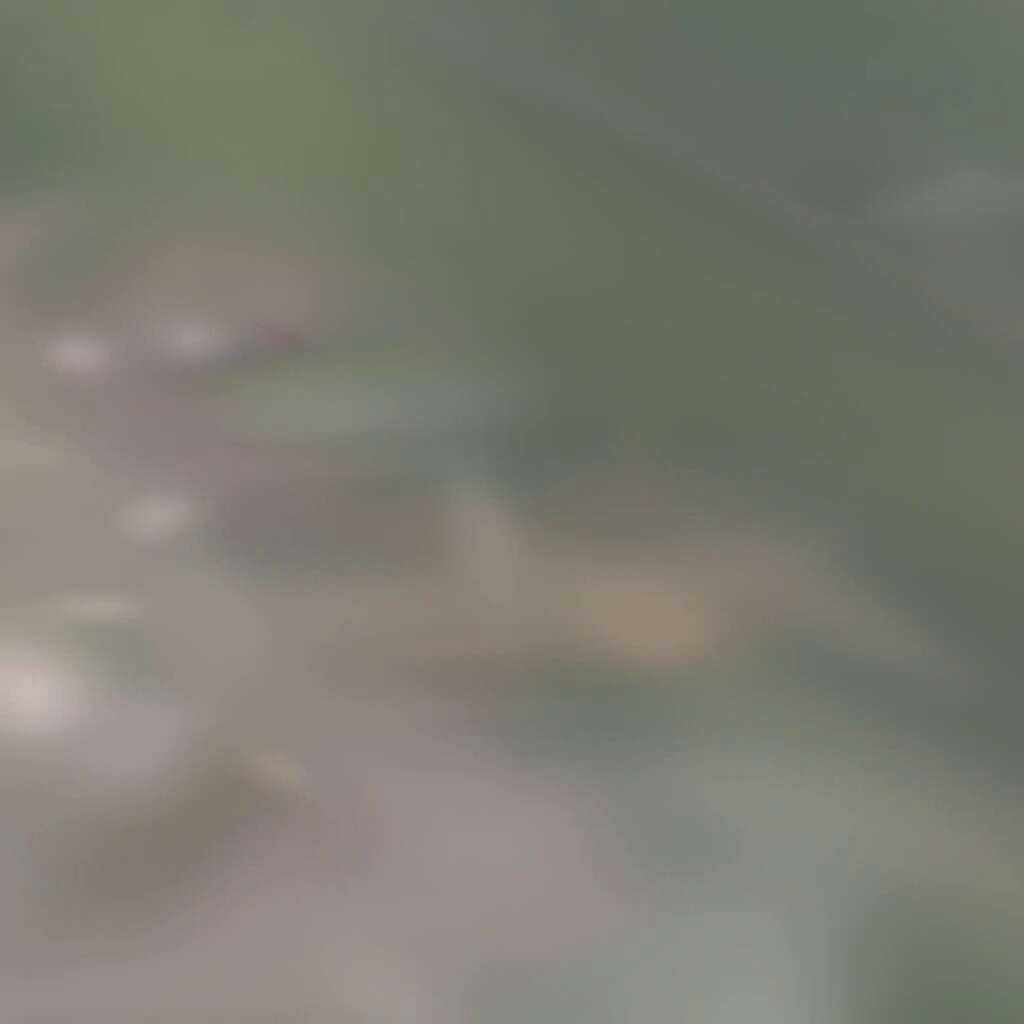} & 
    \imagecell[0.18]{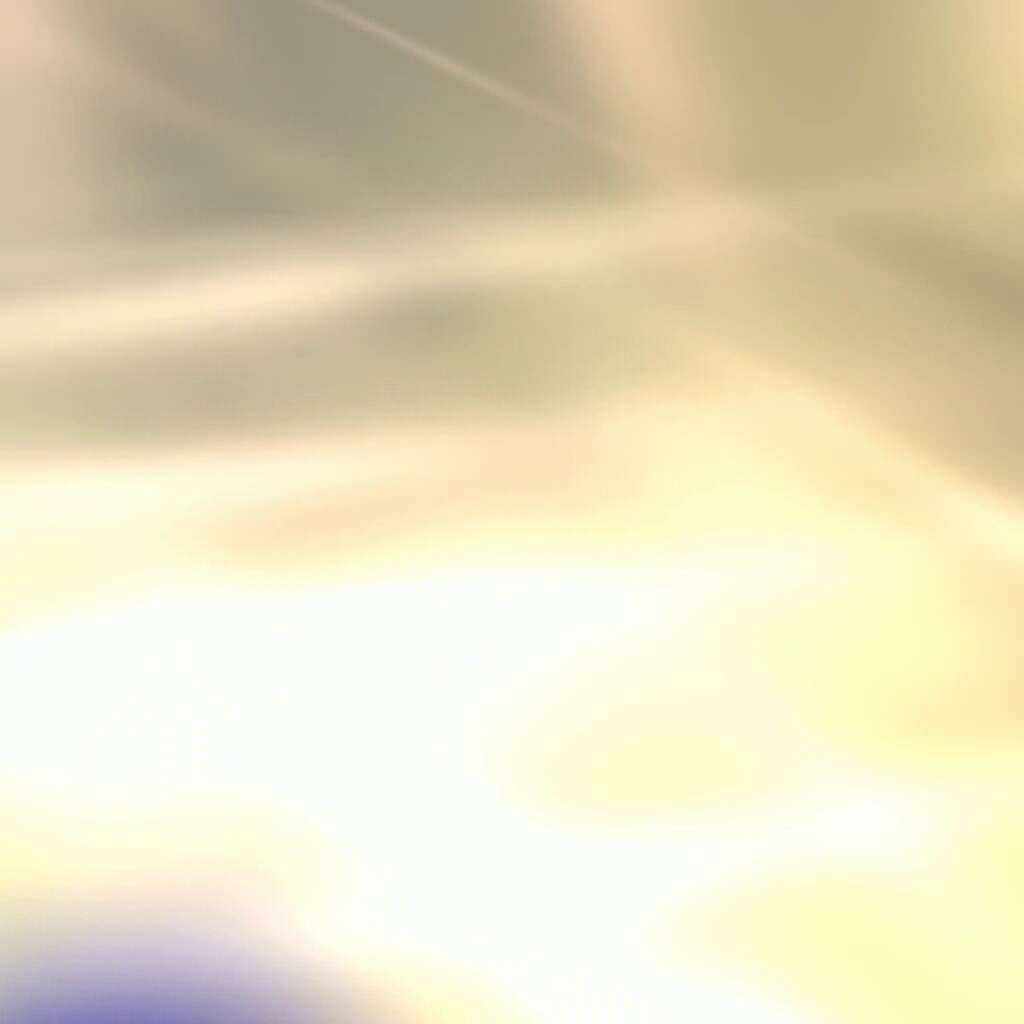} & 
    \imagecell[0.18]{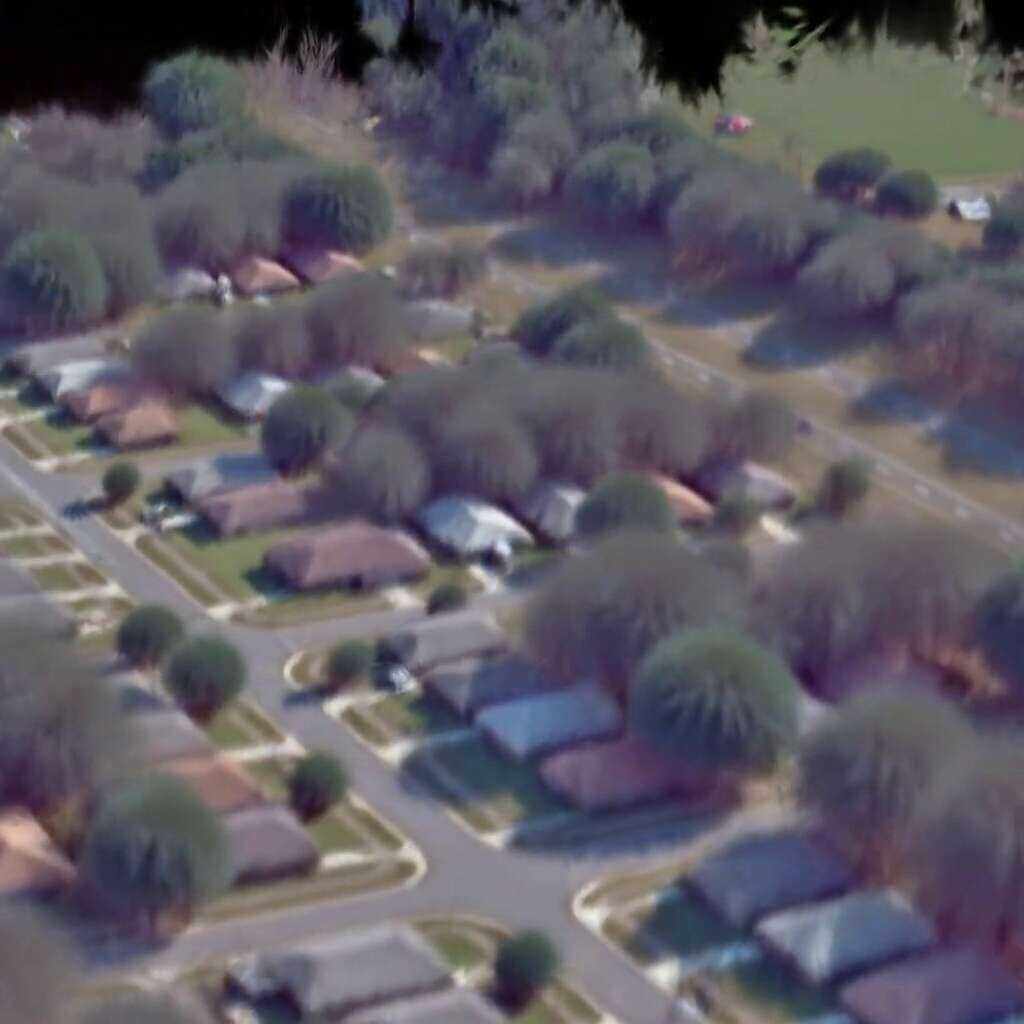} & 
    \imagecell[0.18]{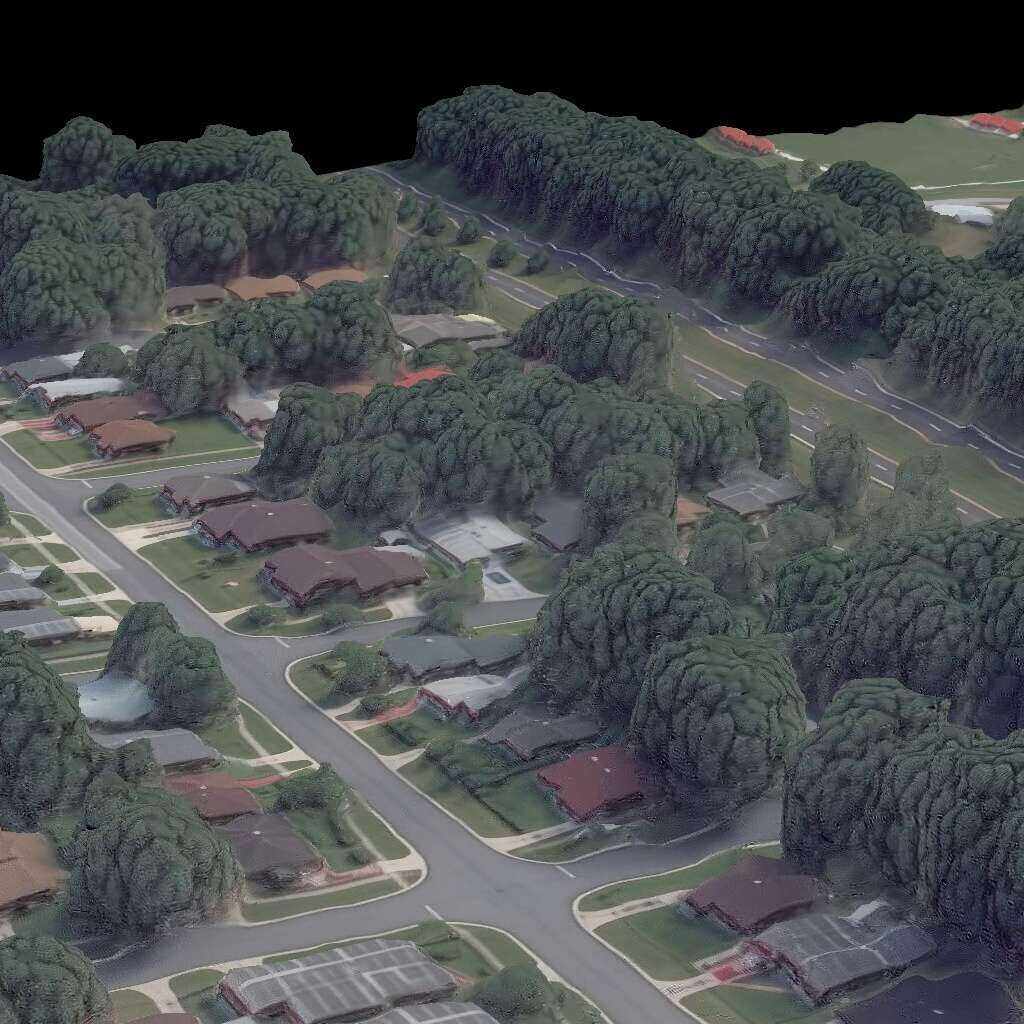} \\
    \vspace*{-10pt} \\

    \imagecell[0.18]{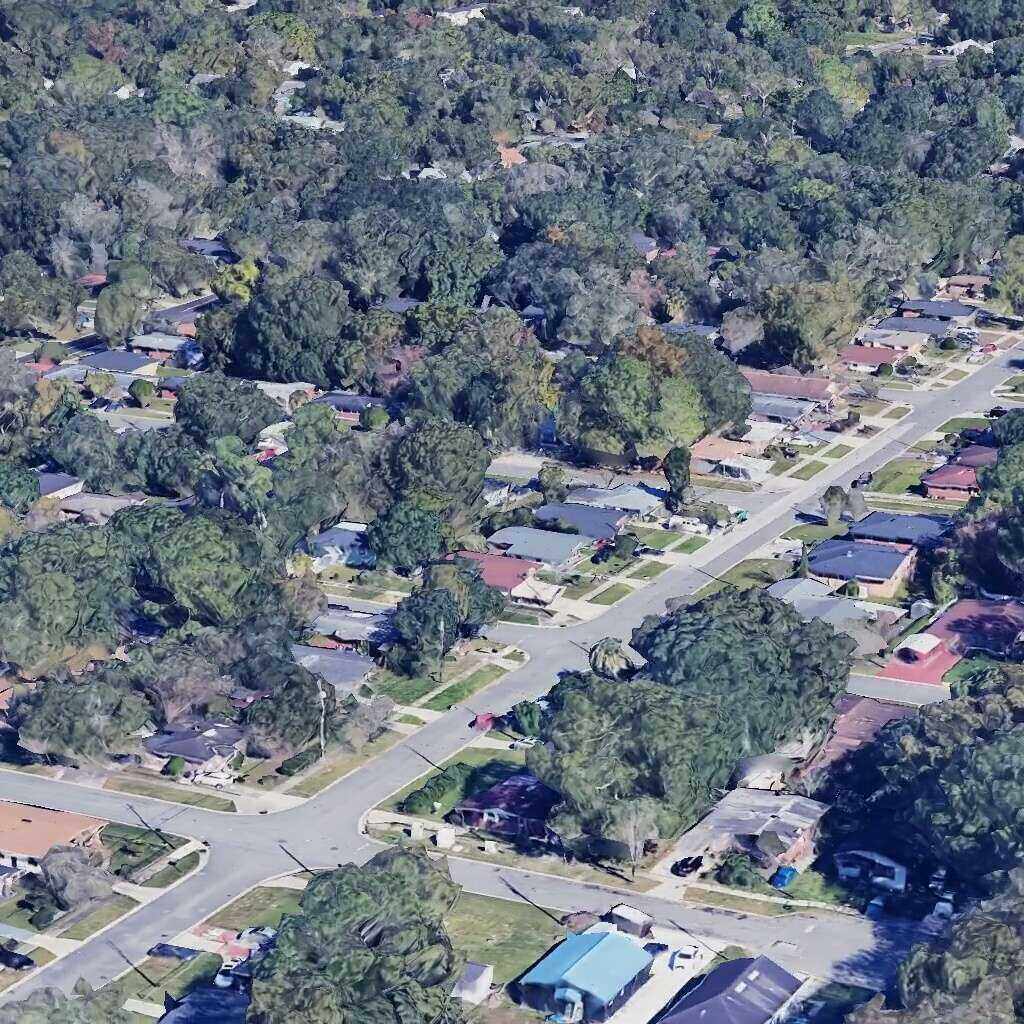} & 
    \imagecell[0.18]{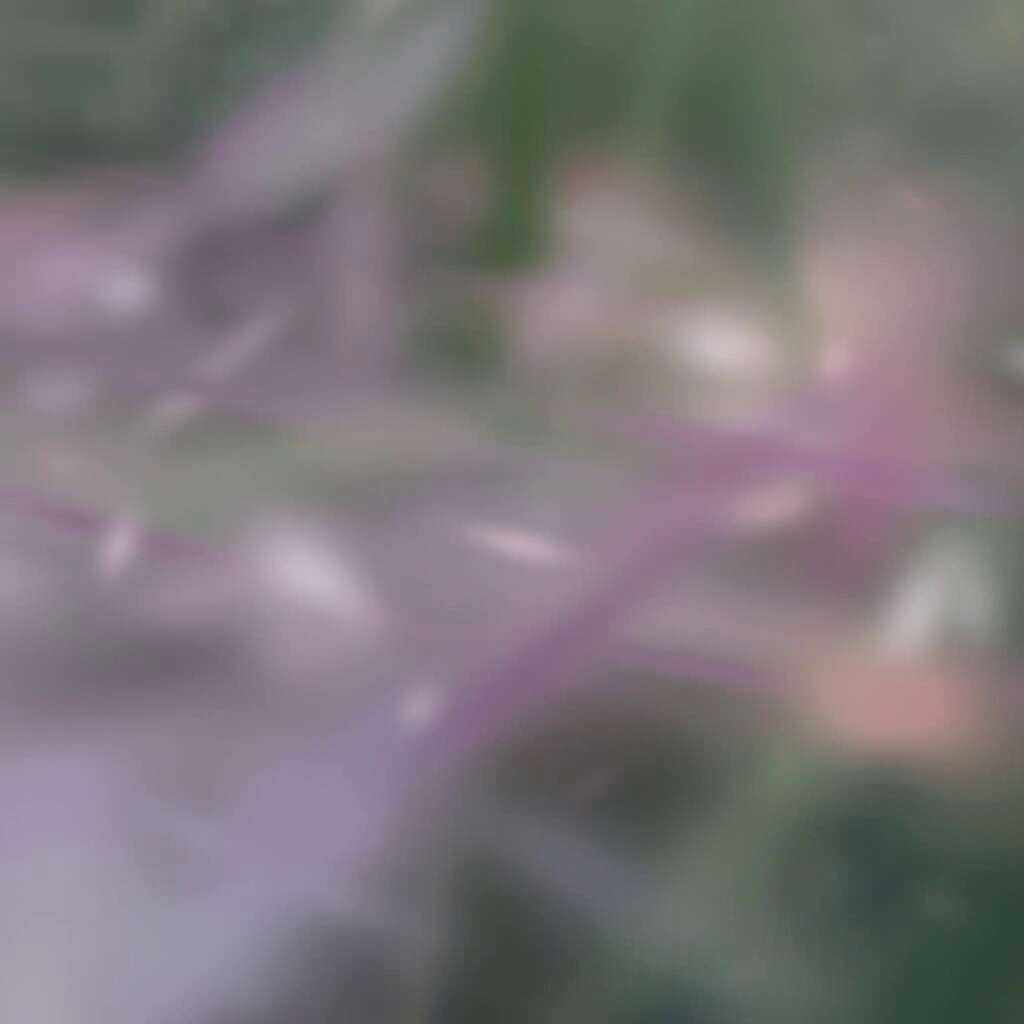} & 
    \imagecell[0.18]{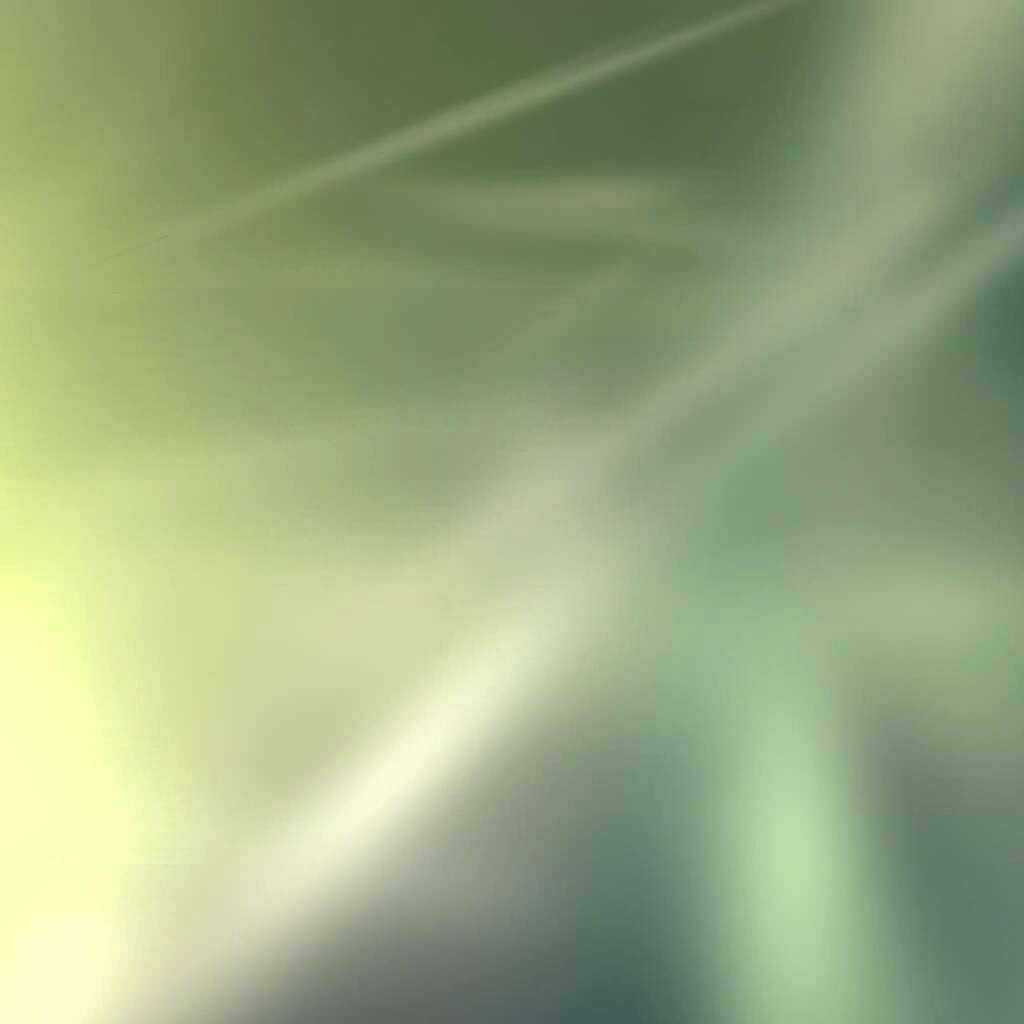} & 
    \imagecell[0.18]{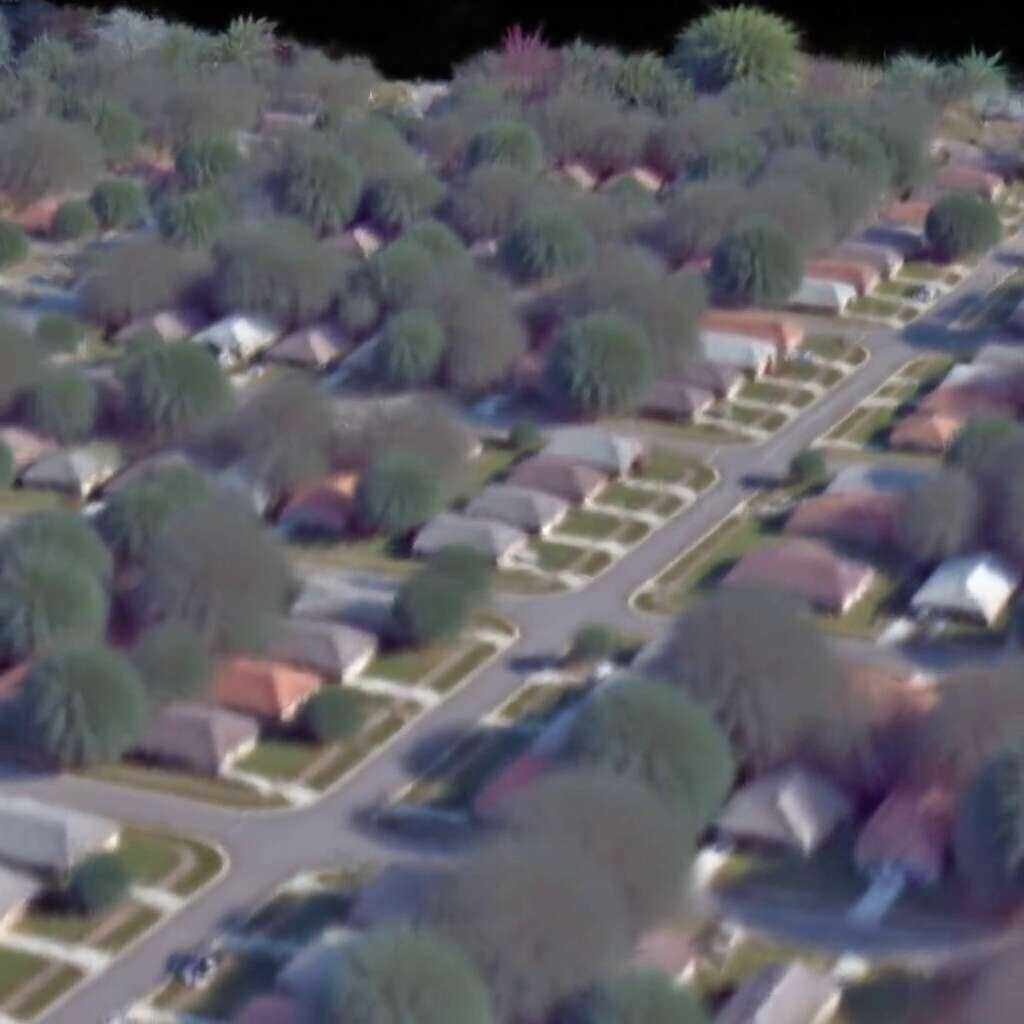} & 
    \imagecell[0.18]{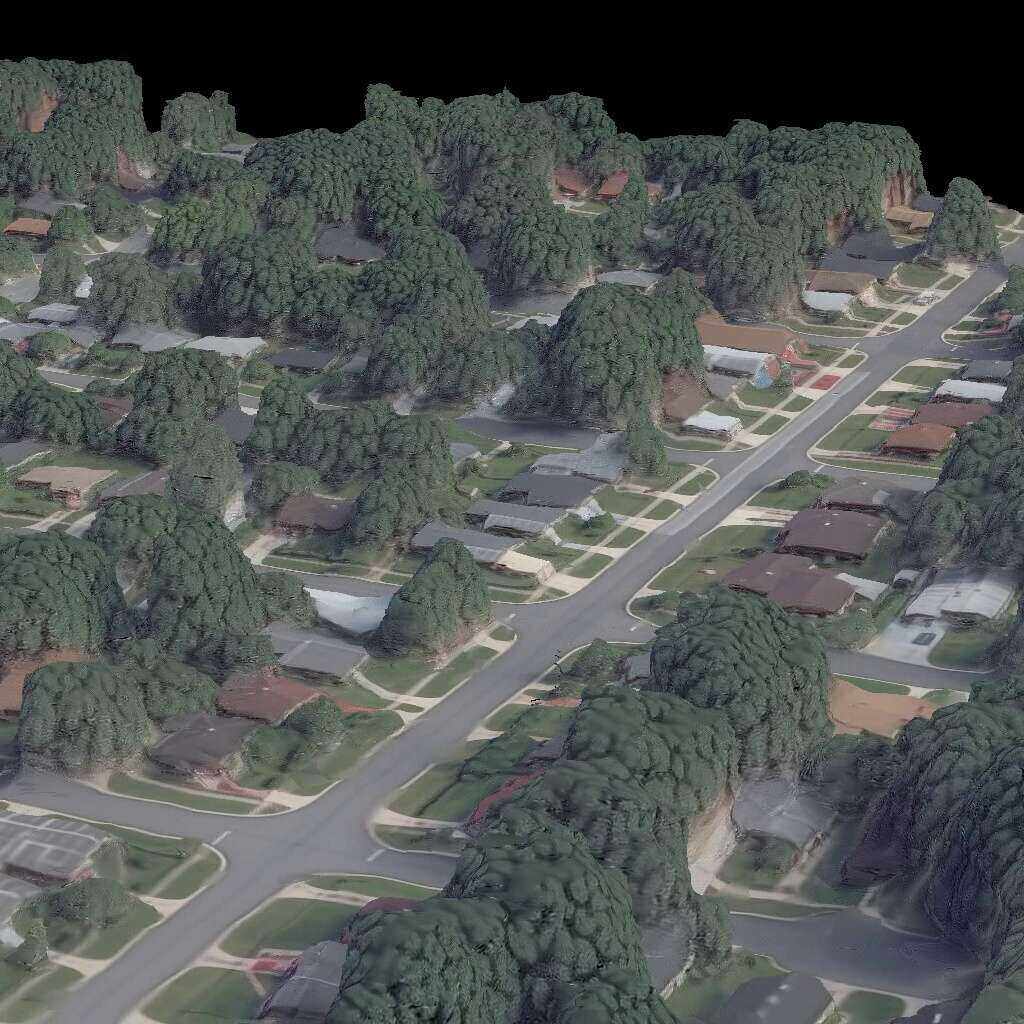} \\
    \vspace*{-10pt} \\

    \\
    \vspace*{-20pt}
    \\
    G.T. & 
    Mip-Splatting & 
    2DGS & 
    Skyfall-GS & 
    Ours \\
    
    \end{tabular}
    \end{spacing}
	\caption{ 
    \textbf{Results of the JAX\_004 scene in the DFC 2019 dataset}. 
    Compared to baselines, our method successfully achieves high-quality city reconstruction from satellite imagery. 
    Results of CityGS-X are removed since the method crashes while recovering this scene.
    }
    \label{fig:supp:qual-jax-004}
    \vspace*{-0.3cm}
\end{figure*}

\begin{figure*}[p]
	\centering
    \begin{spacing}{1} 
    \setlength\tabcolsep{1pt}
    \begin{tabular}{ccccc}  

    \imagecell[0.18]{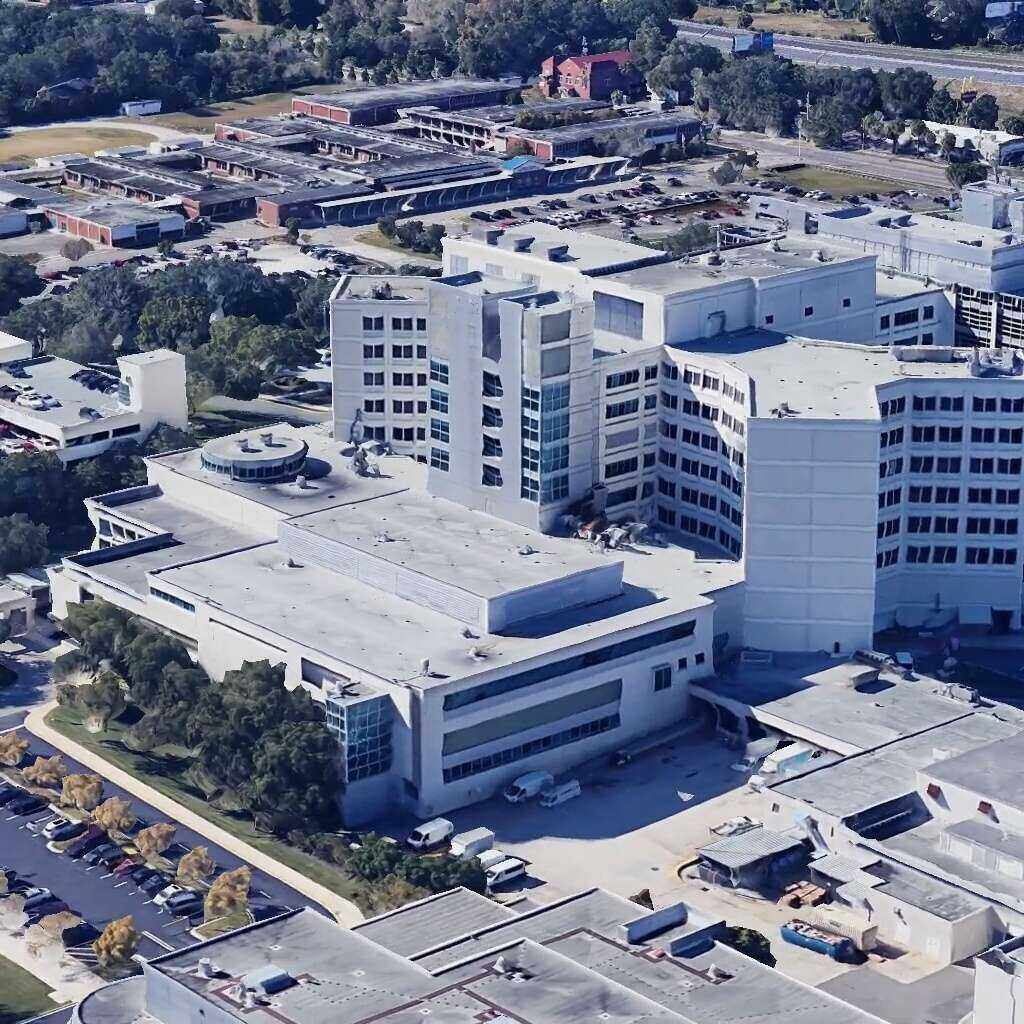} & 
    \imagecell[0.18]{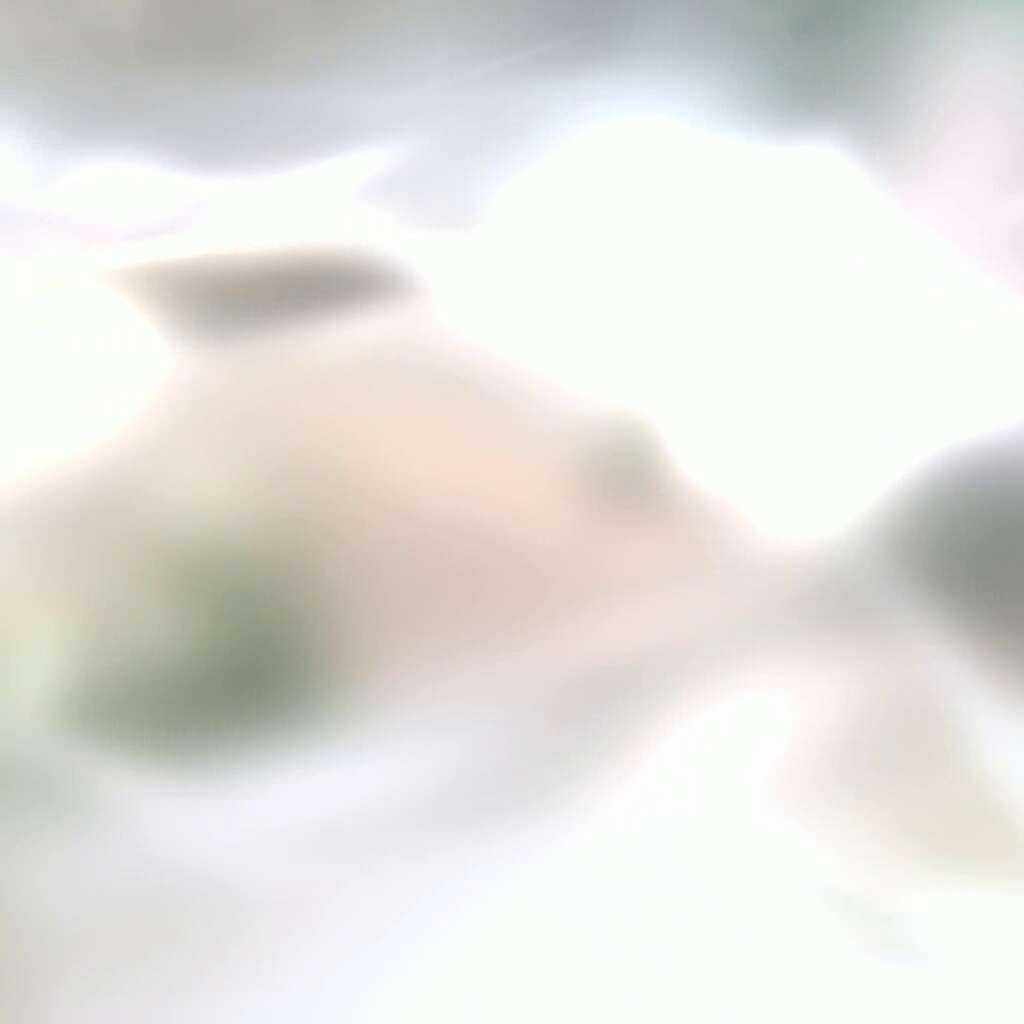} & 
    \imagecell[0.18]{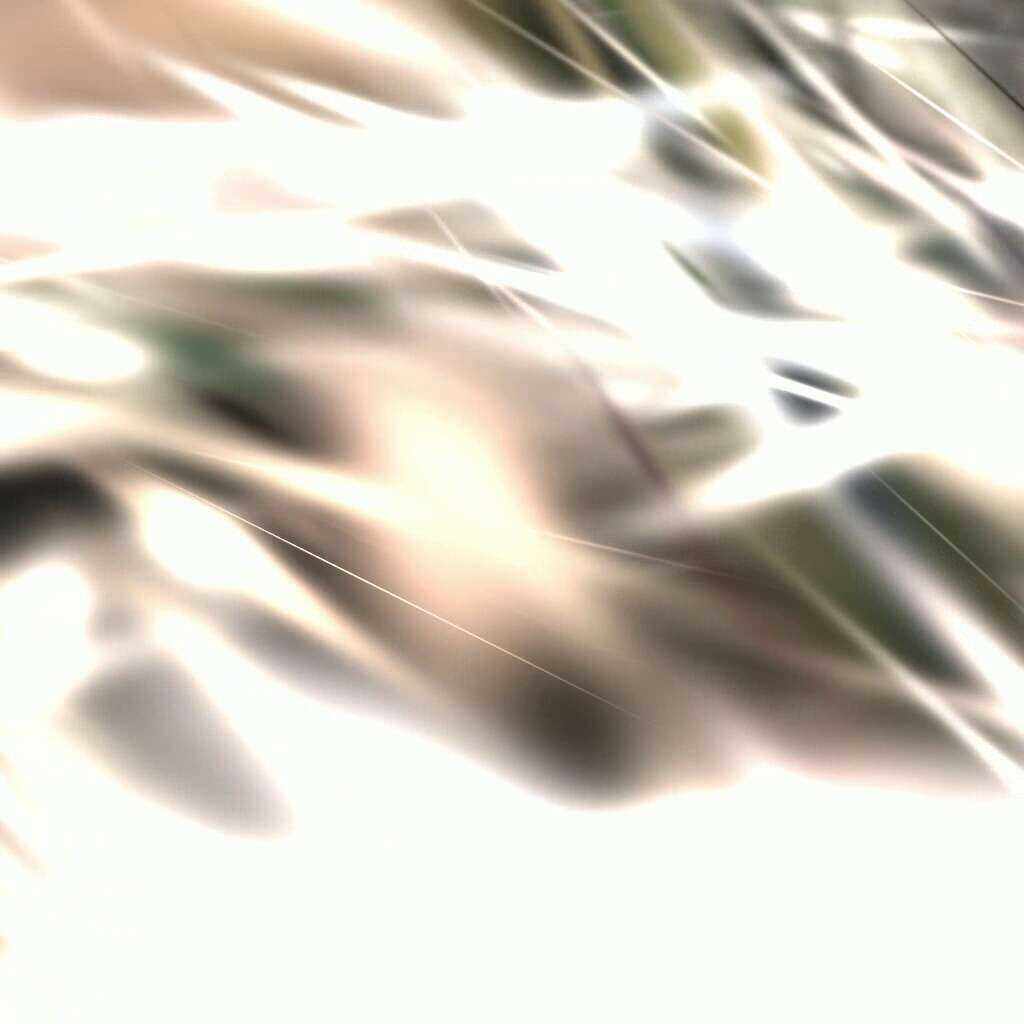} & 
    \imagecell[0.18]{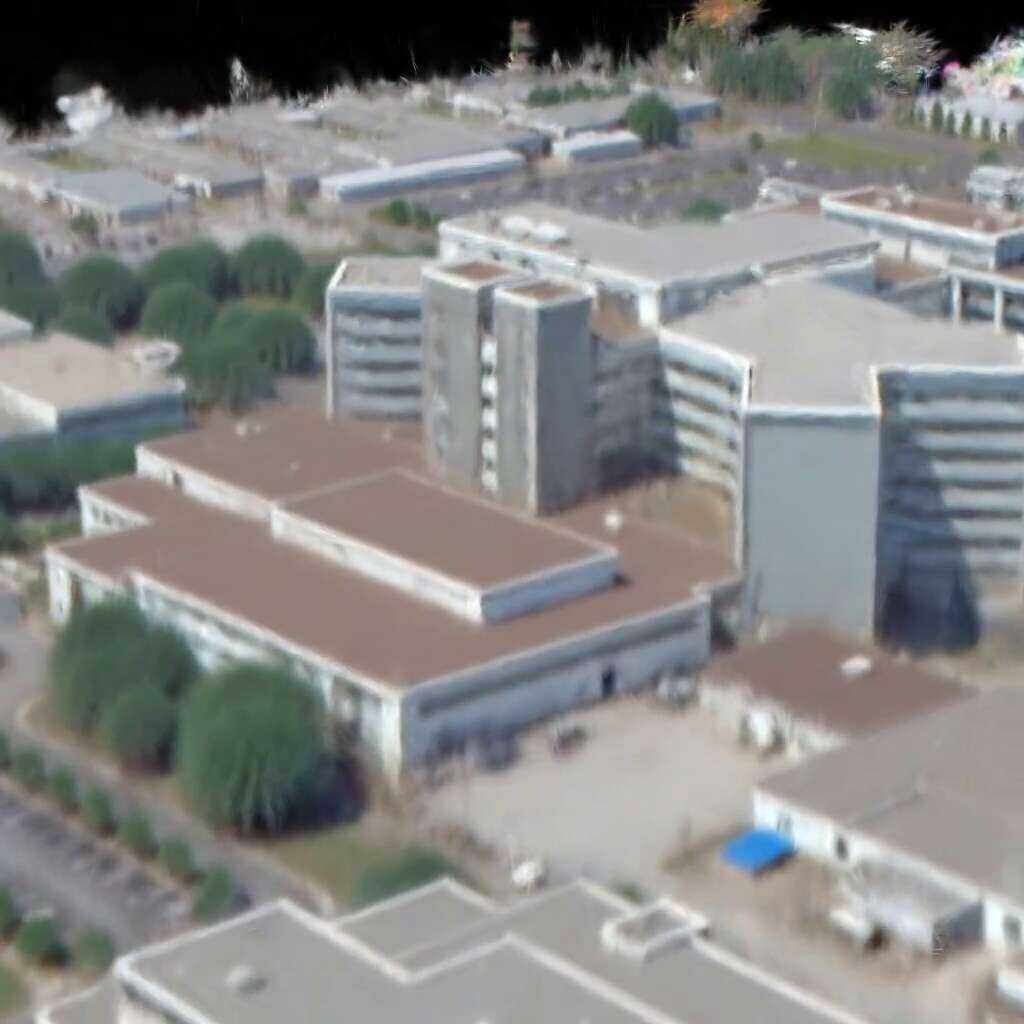} & 
    \imagecell[0.18]{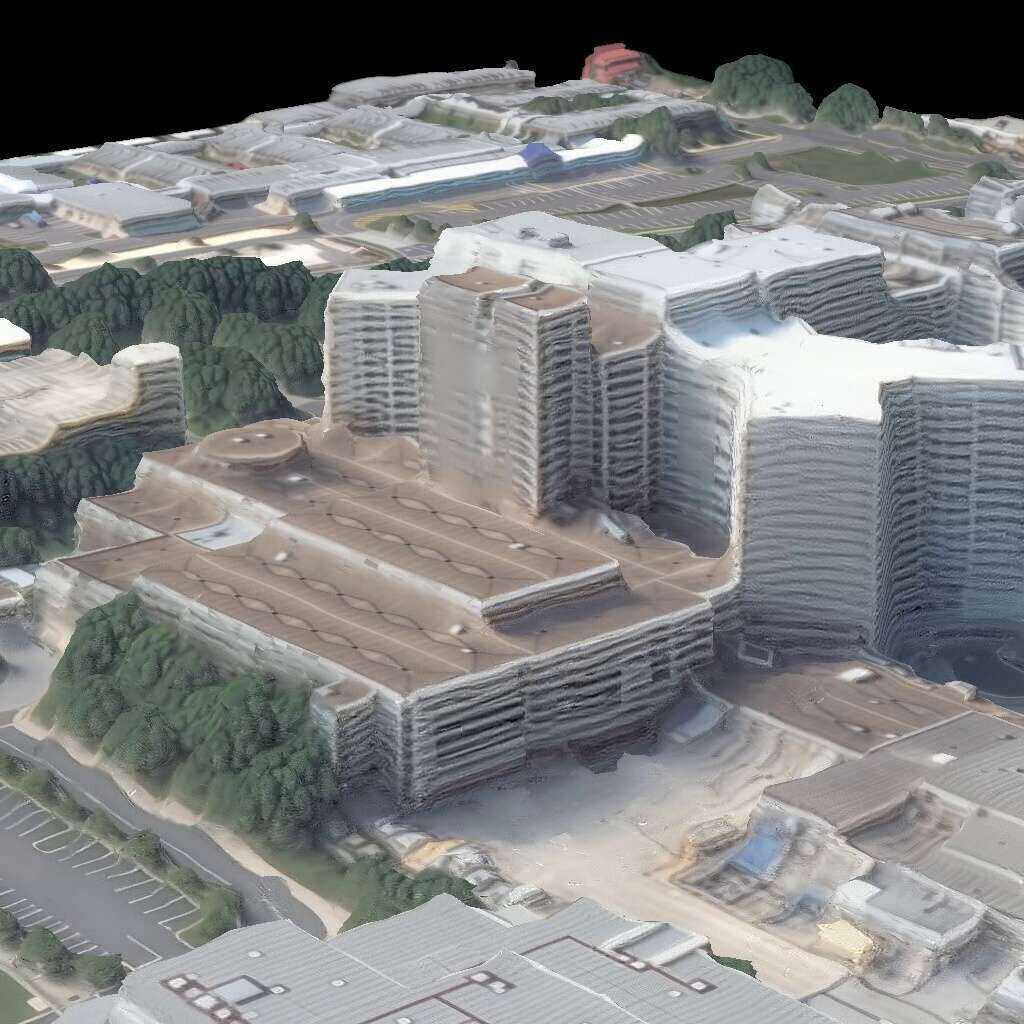} \\
    \vspace*{-10pt} \\    

    \imagecell[0.18]{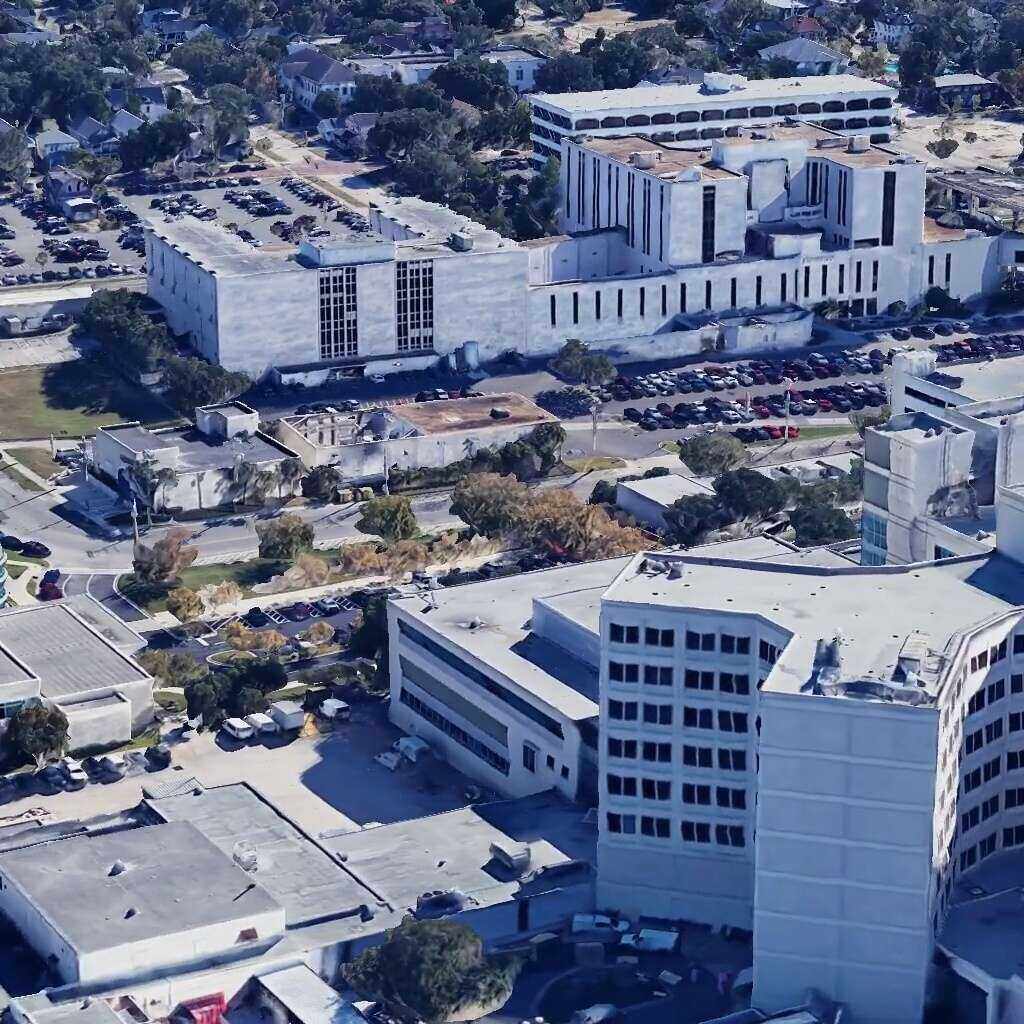} & 
    \imagecell[0.18]{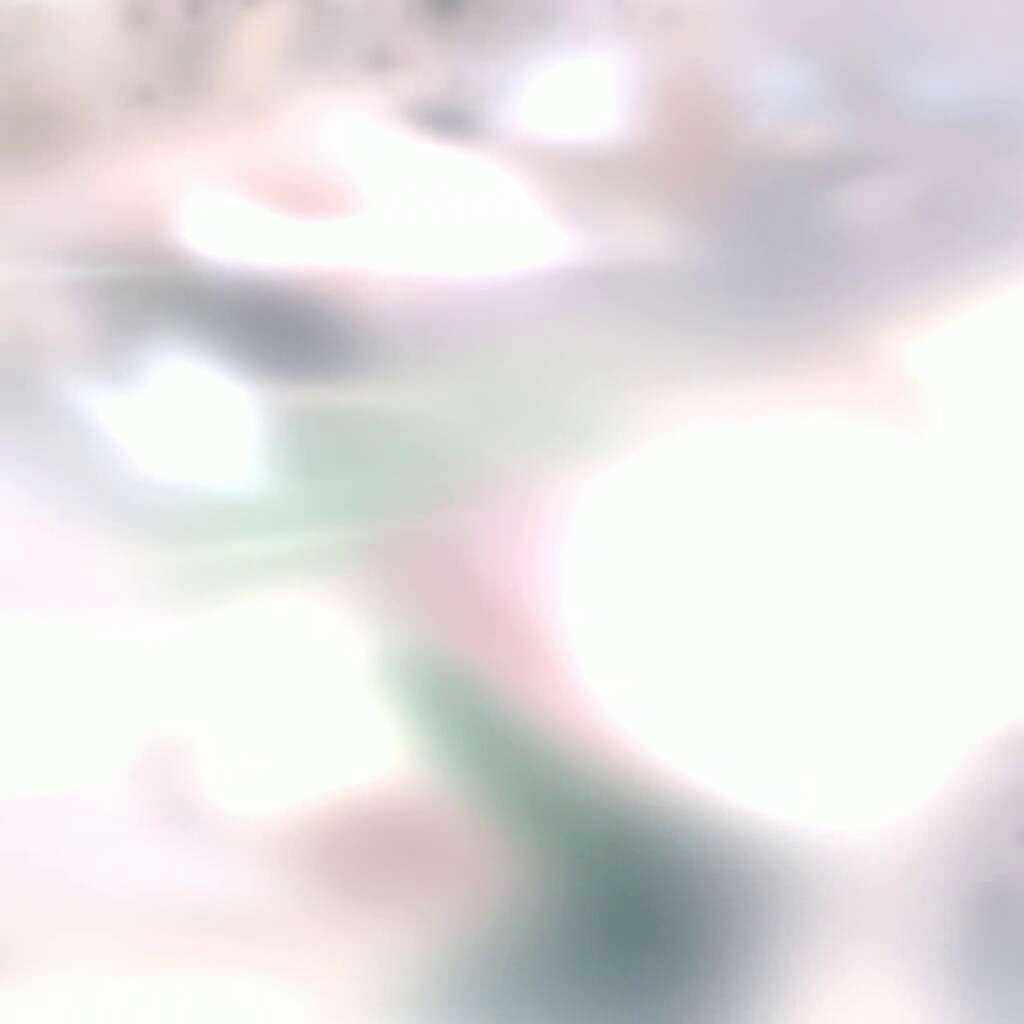} & 
    \imagecell[0.18]{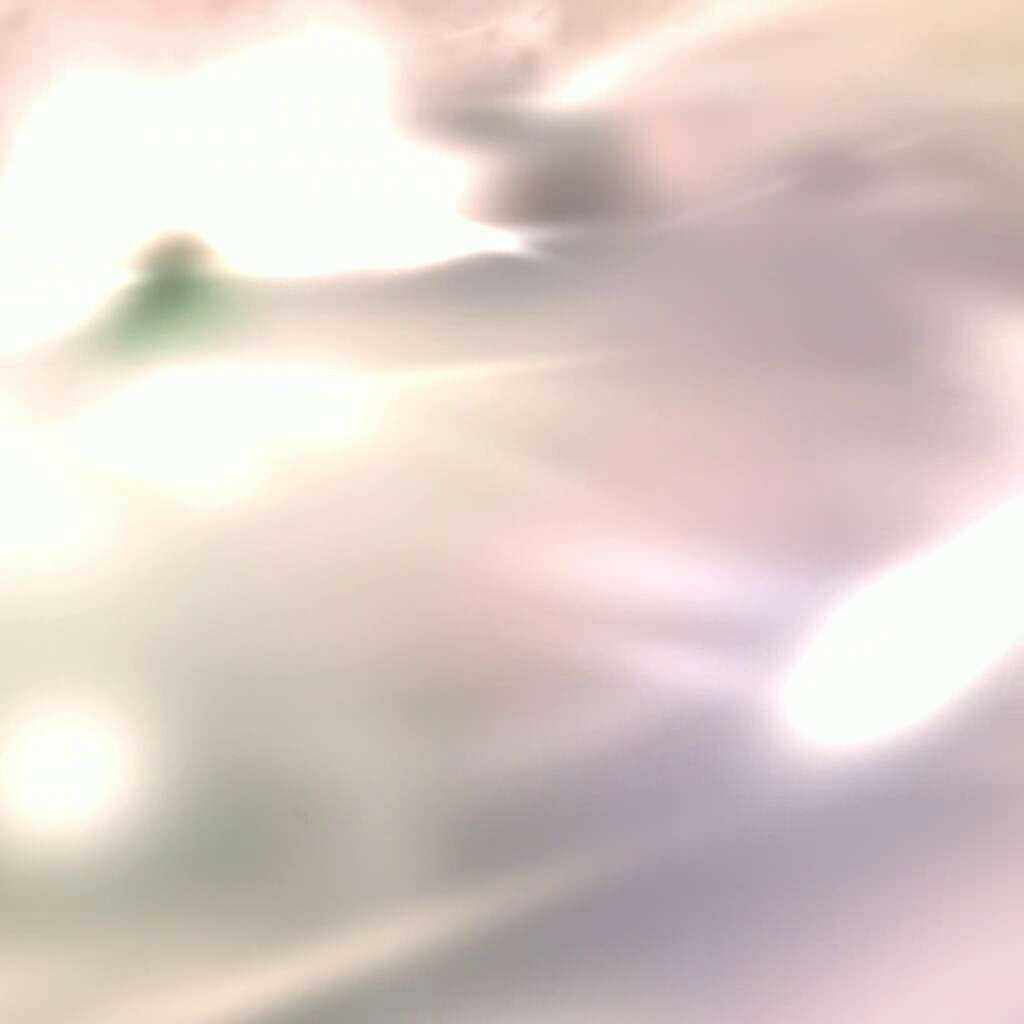} & 
    \imagecell[0.18]{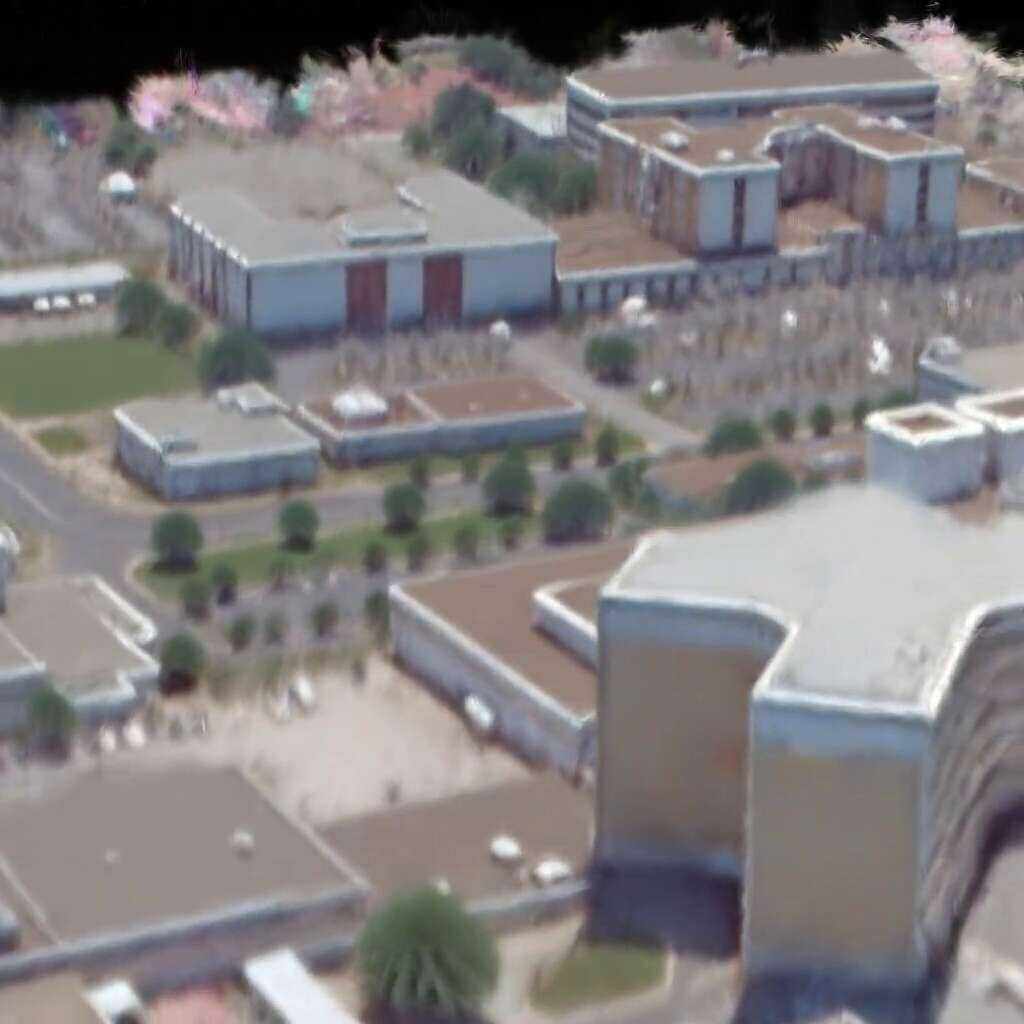} & 
    \imagecell[0.18]{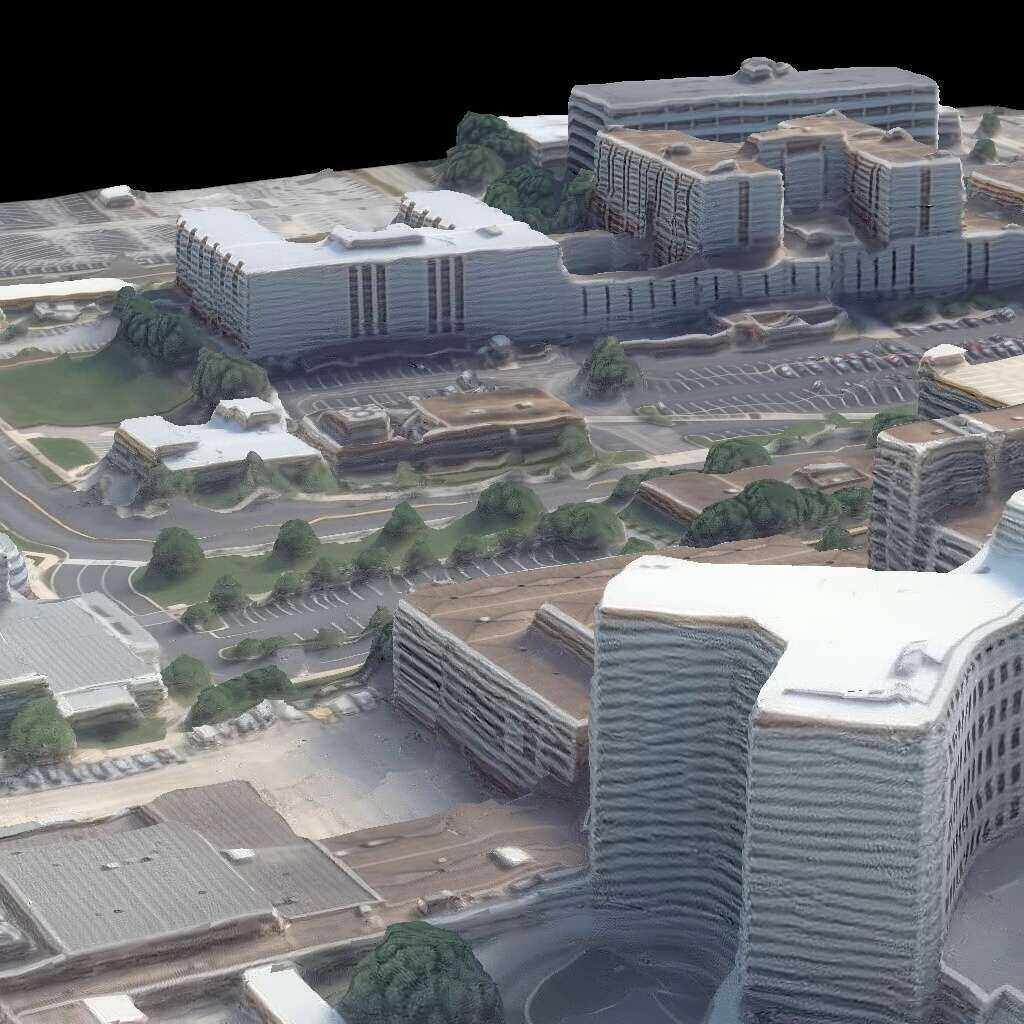} \\
    \vspace*{-10pt} \\
    
    \imagecell[0.18]{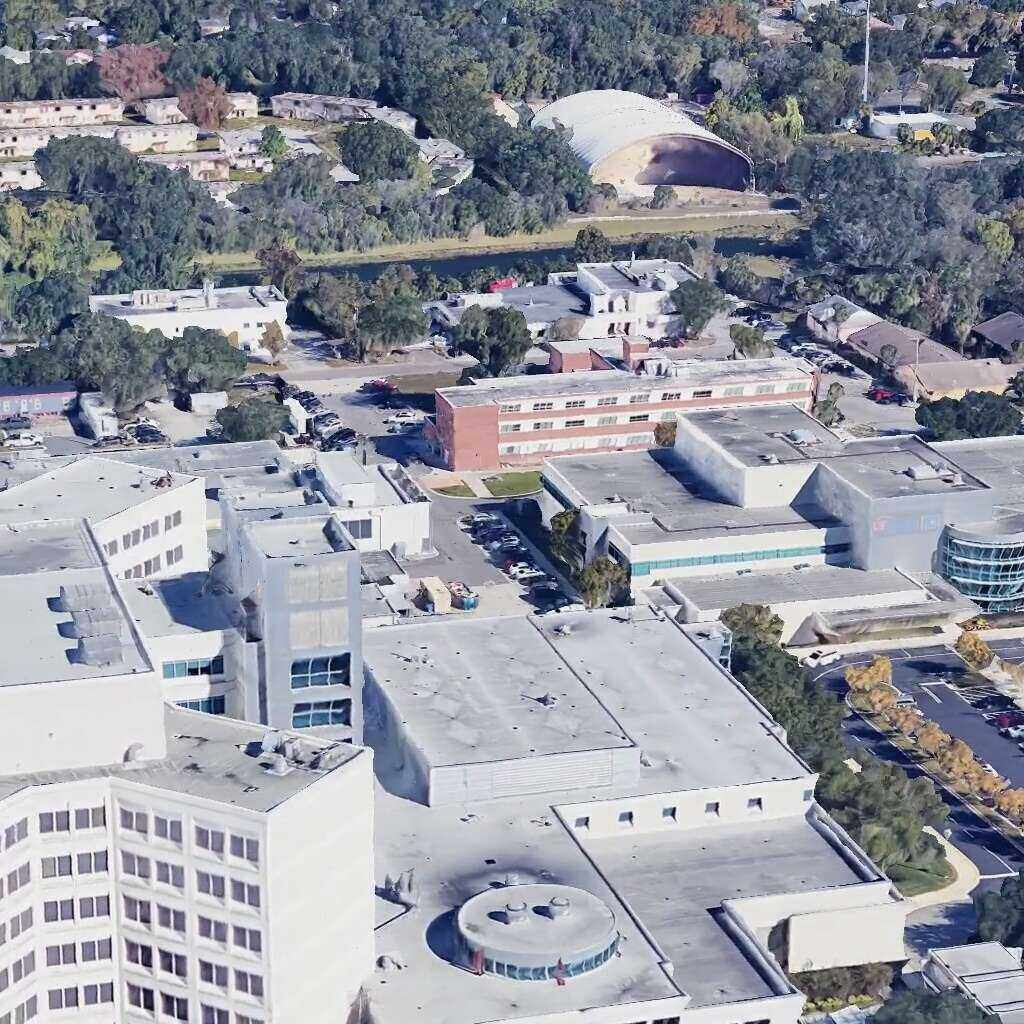} & 
    \imagecell[0.18]{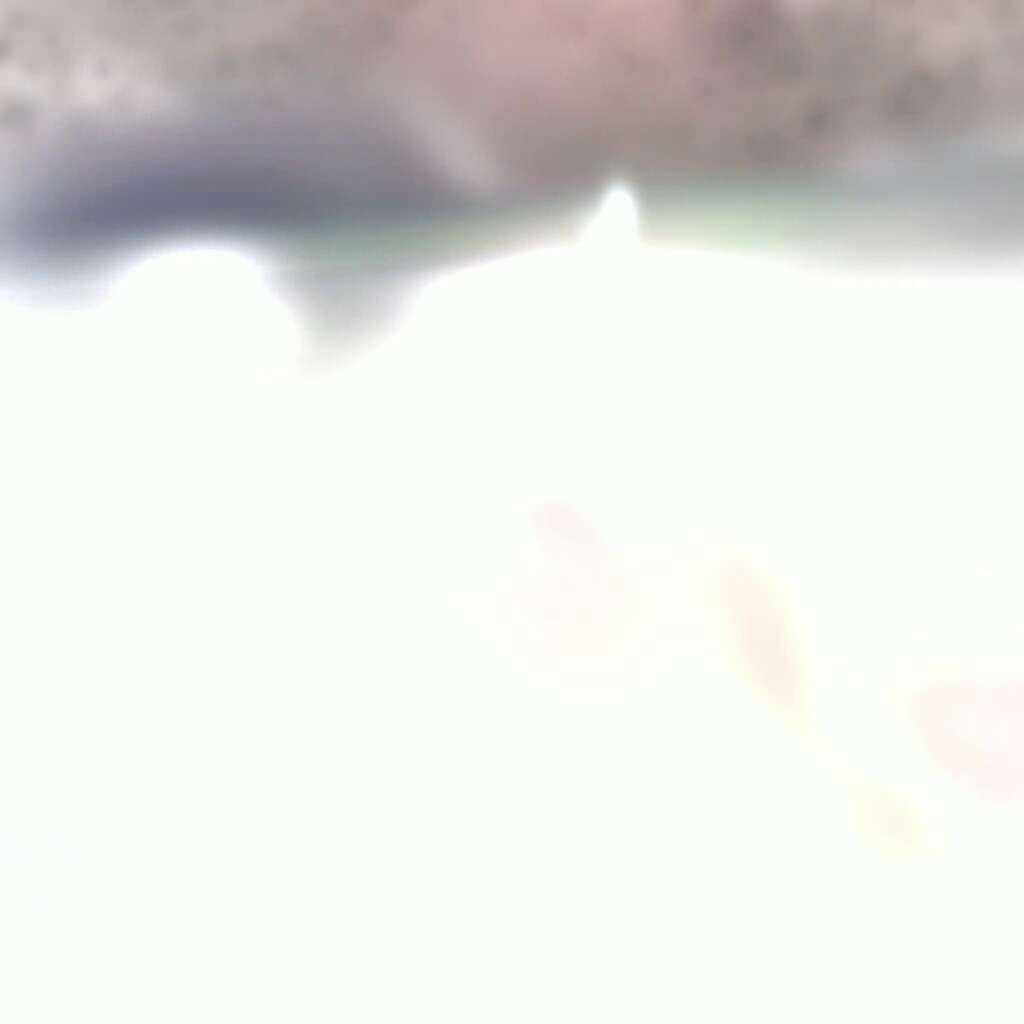} & 
    \imagecell[0.18]{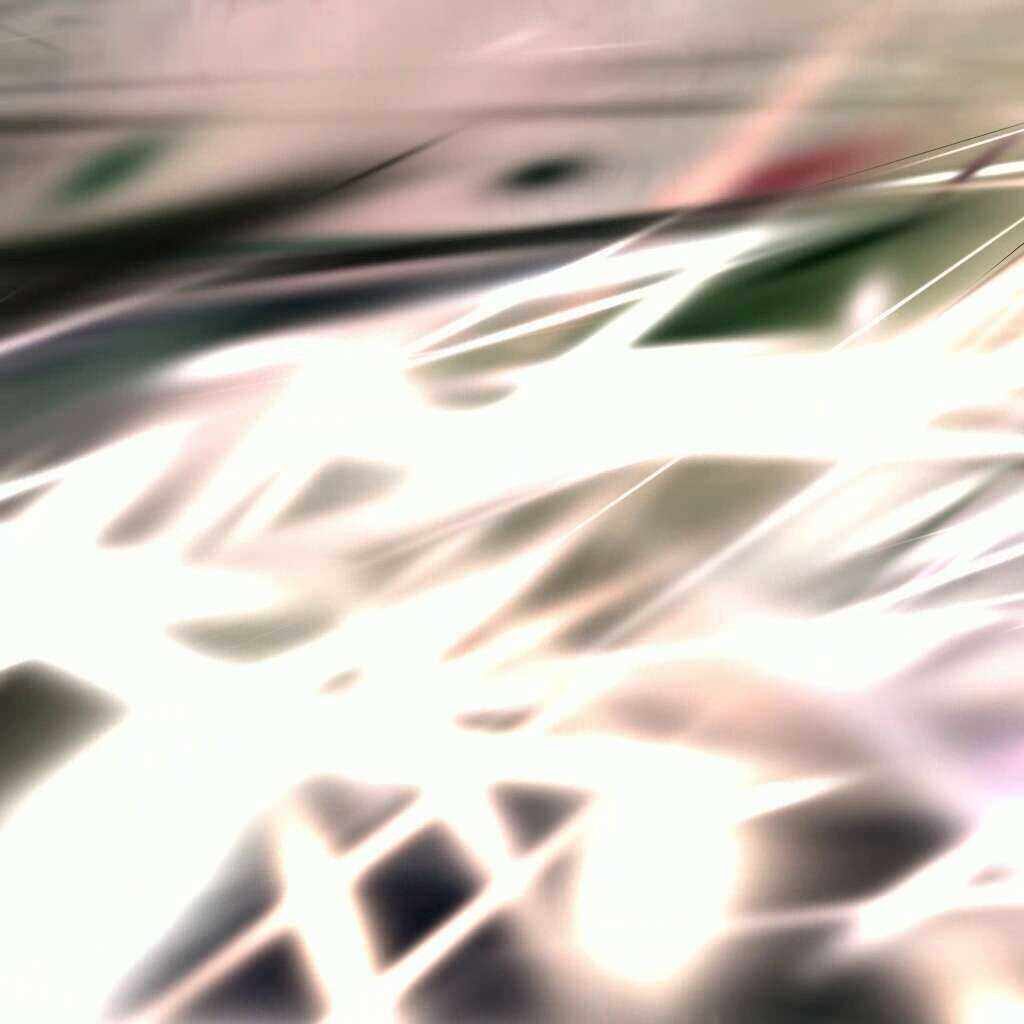} & 
    \imagecell[0.18]{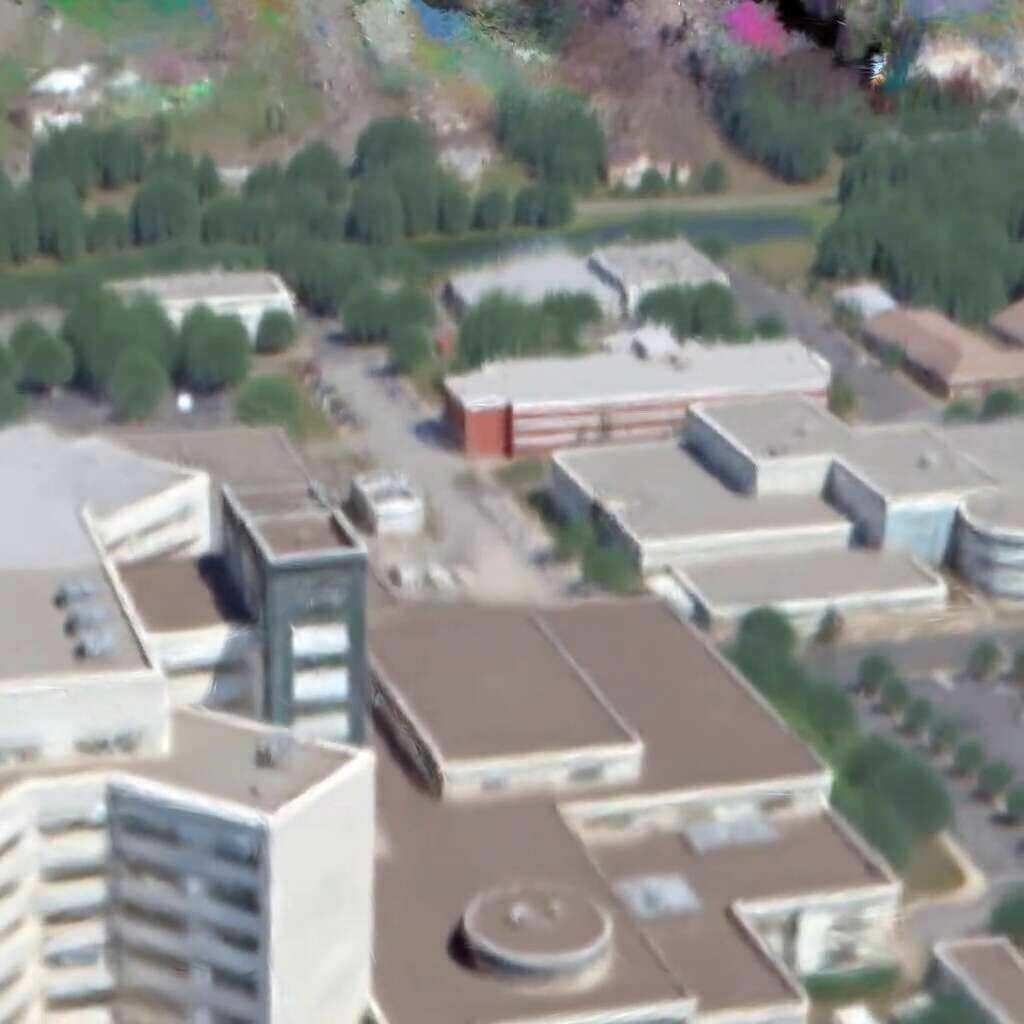} & 
    \imagecell[0.18]{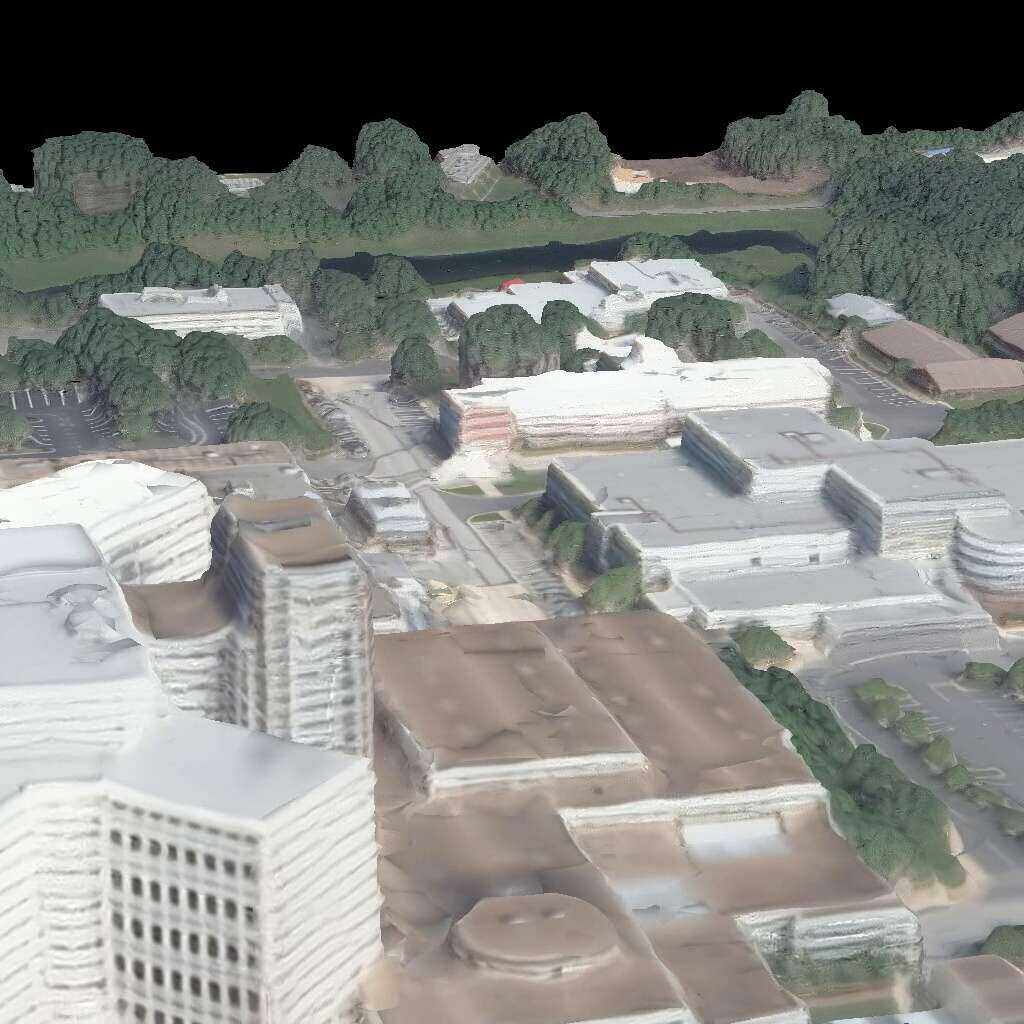} \\
    \vspace*{-10pt} \\

    \\
    \vspace*{-20pt}
    \\
    G.T. & 
    Mip-Splatting & 
    2DGS & 
    Skyfall-GS & 
    Ours \\
    
    \end{tabular}
    \end{spacing}
	\caption{ 
    \textbf{Results of the JAX\_068 scene in the DFC 2019 dataset}. 
    Compared to baselines, our method successfully achieves high-quality city reconstruction from satellite imagery. 
    Results of CityGS-X are removed since the method crashes while recovering this scene.
    }
    \label{fig:supp:qual-jax-068}
    \vspace*{-0.3cm}
\end{figure*}

\begin{figure*}[p]
	\centering
    \begin{spacing}{1} 
    \setlength\tabcolsep{1pt}
    \begin{tabular}{ccccc}

    \imagecell[0.18]{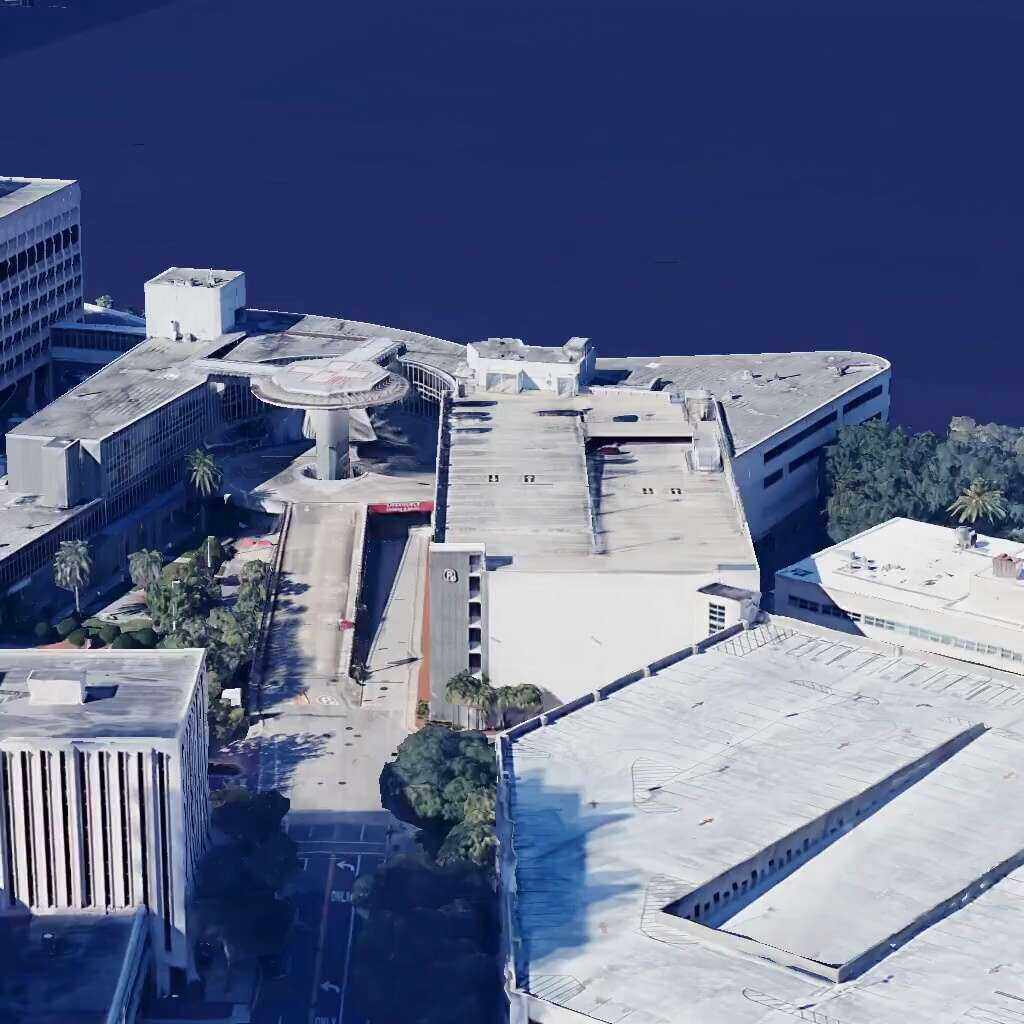} & 
    \imagecell[0.18]{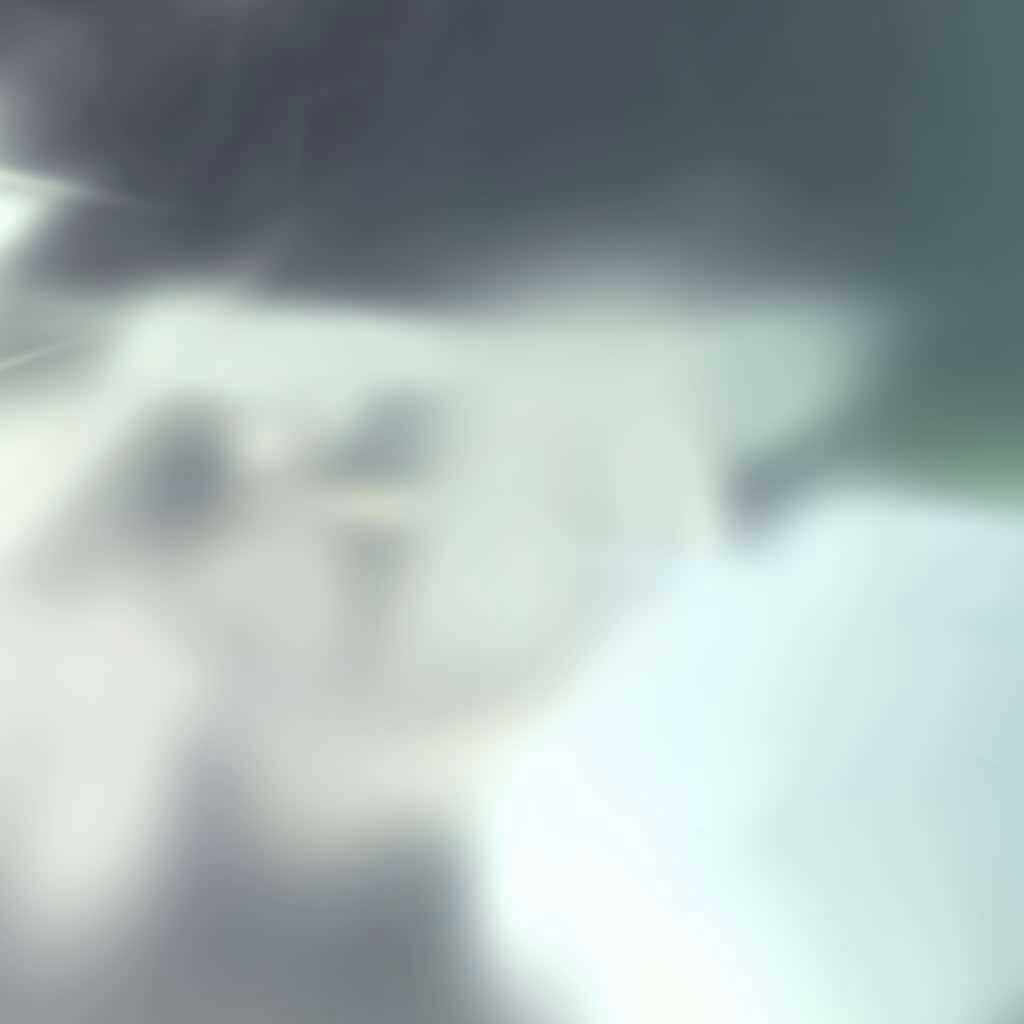} & 
    \imagecell[0.18]{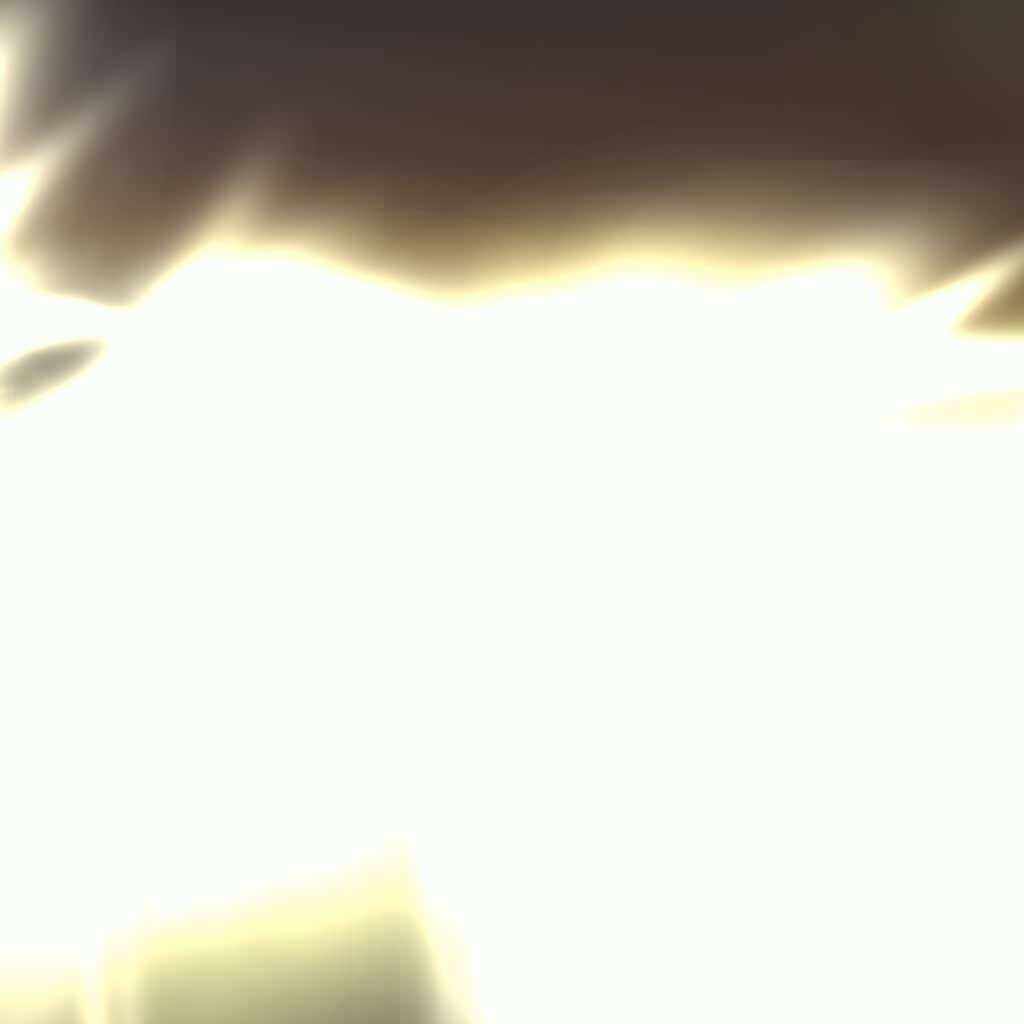} & 
    \imagecell[0.18]{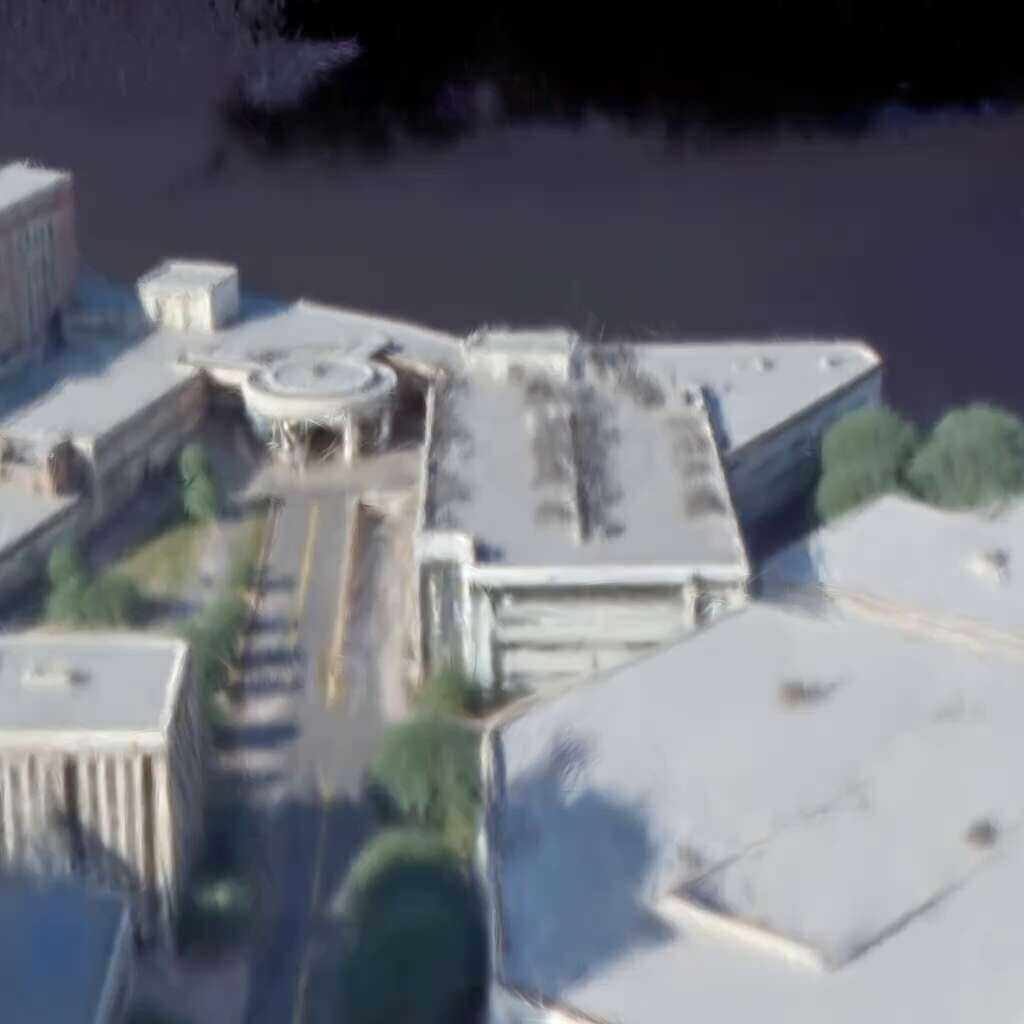} & 
    \imagecell[0.18]{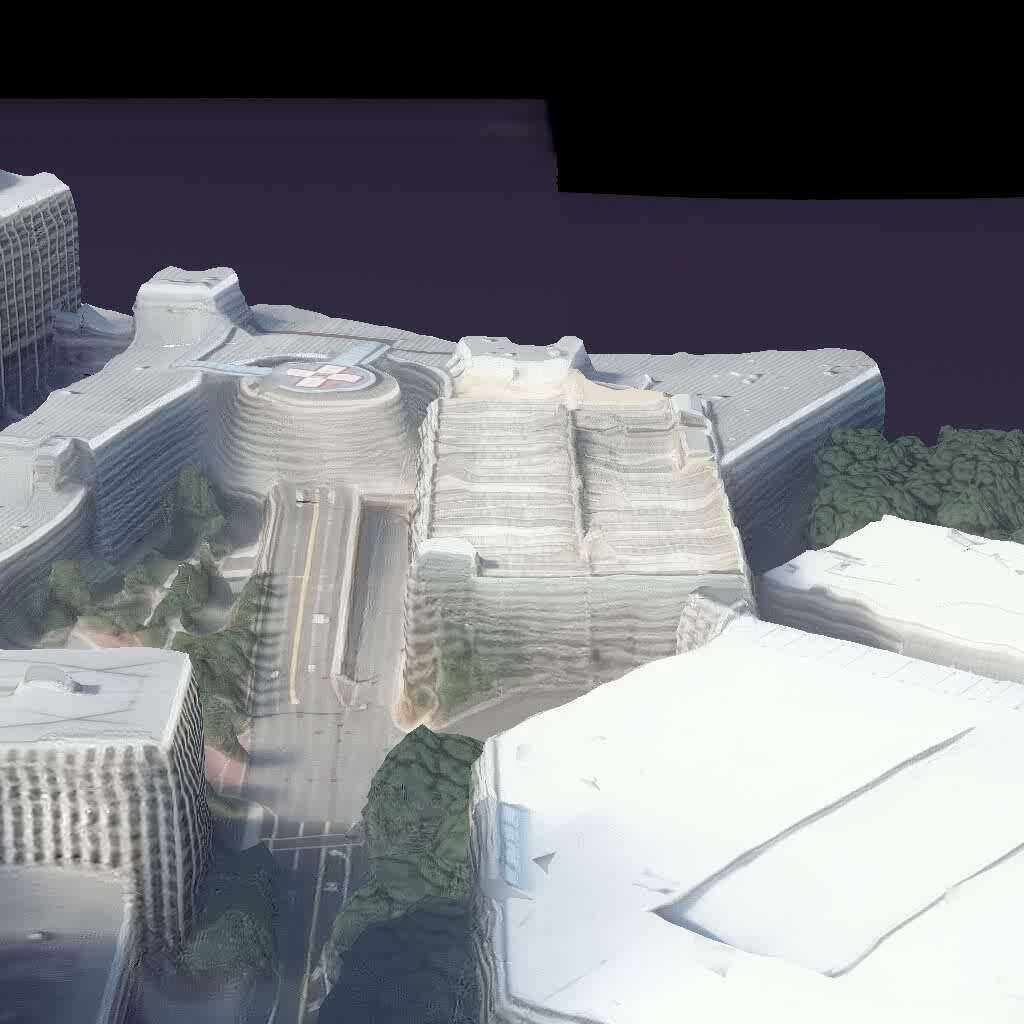} \\
    \vspace*{-10pt} \\

    \imagecell[0.18]{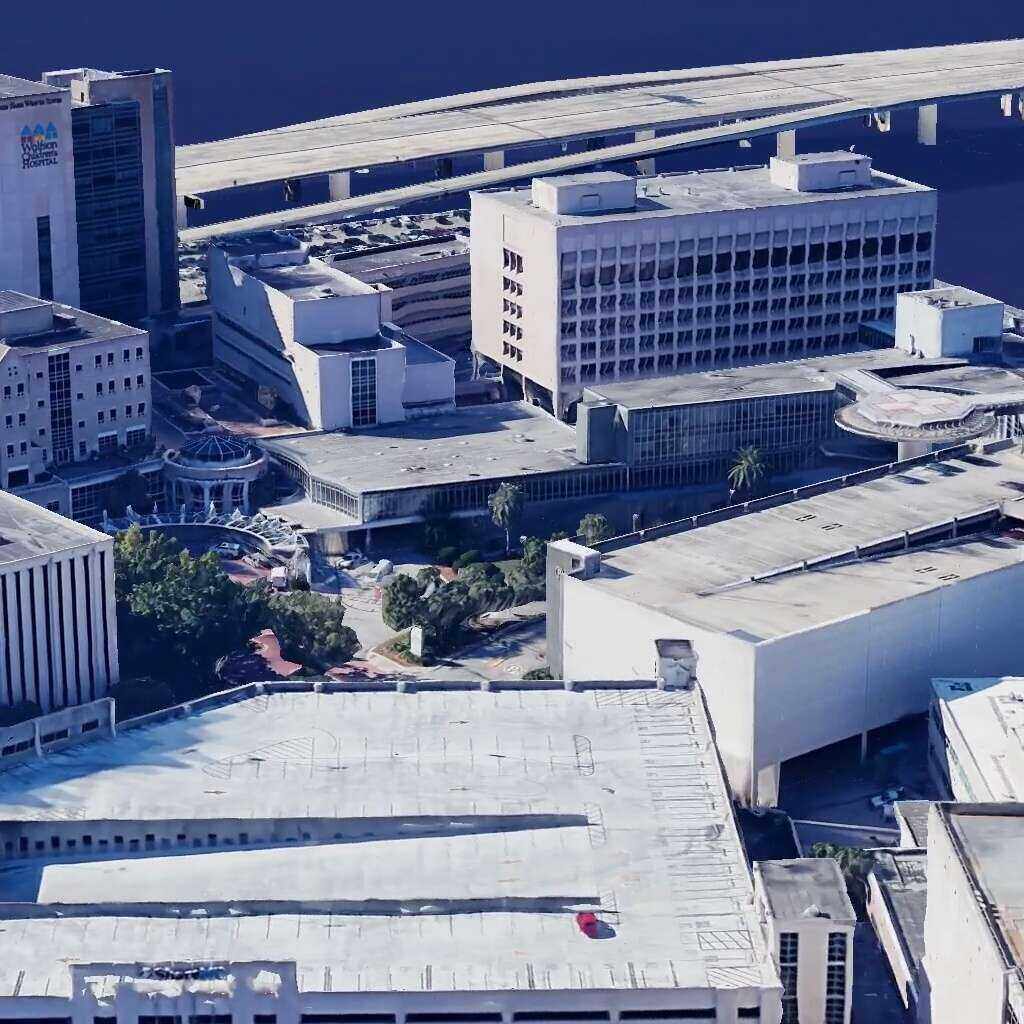} & 
    \imagecell[0.18]{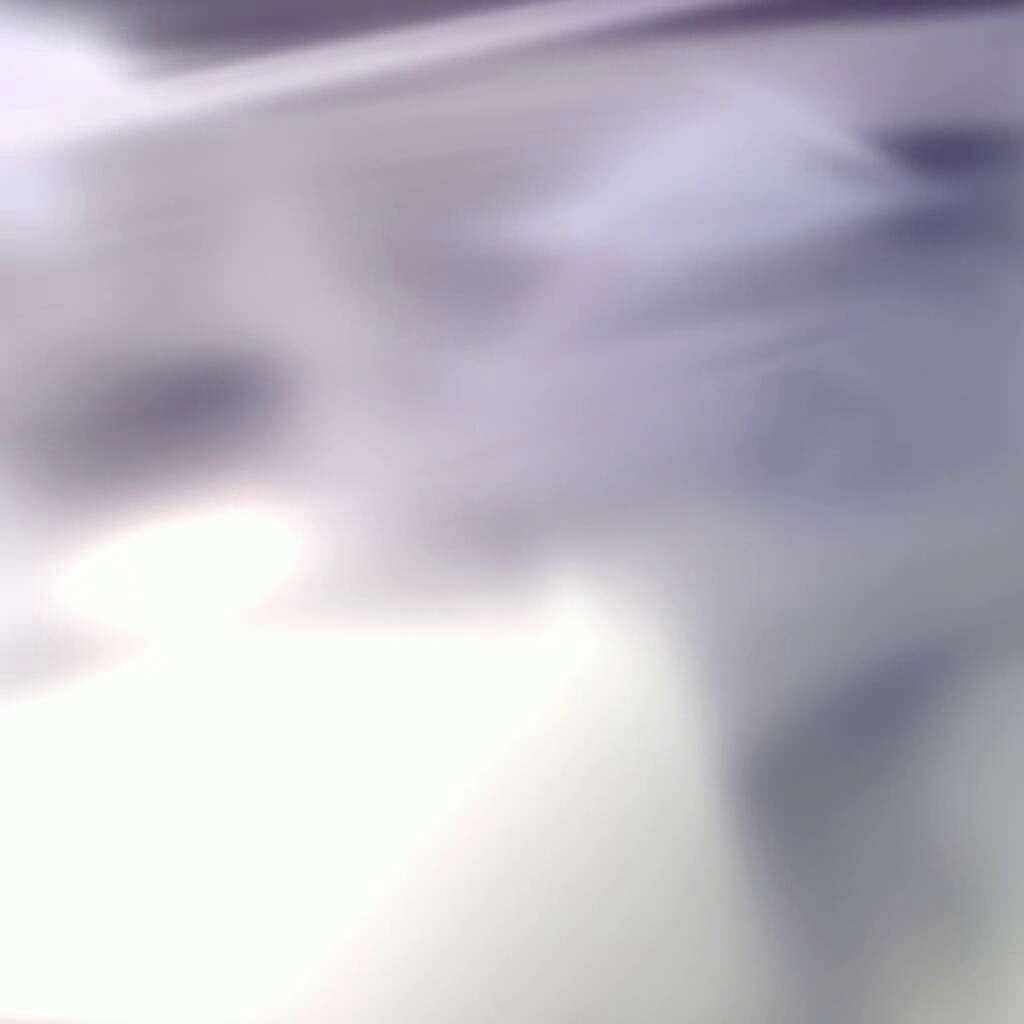} & 
    \imagecell[0.18]{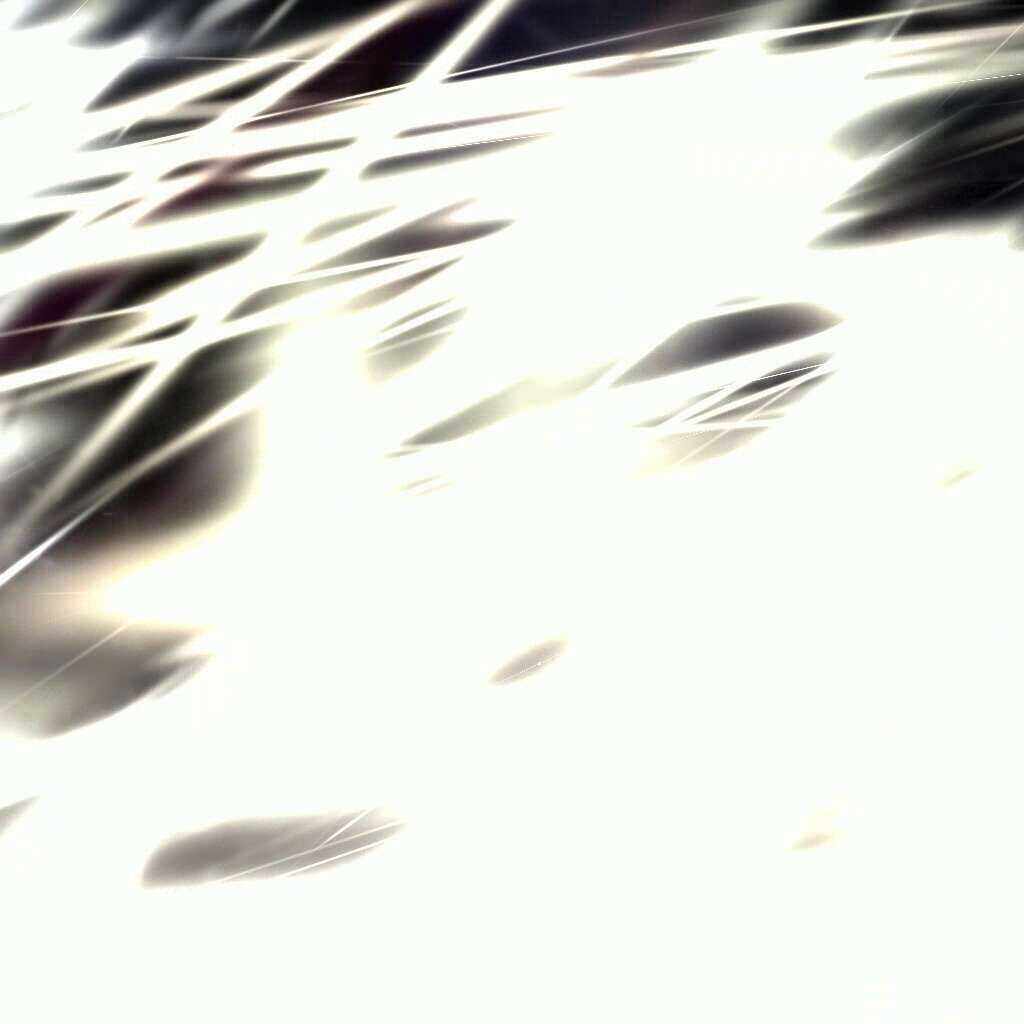} & 
    \imagecell[0.18]{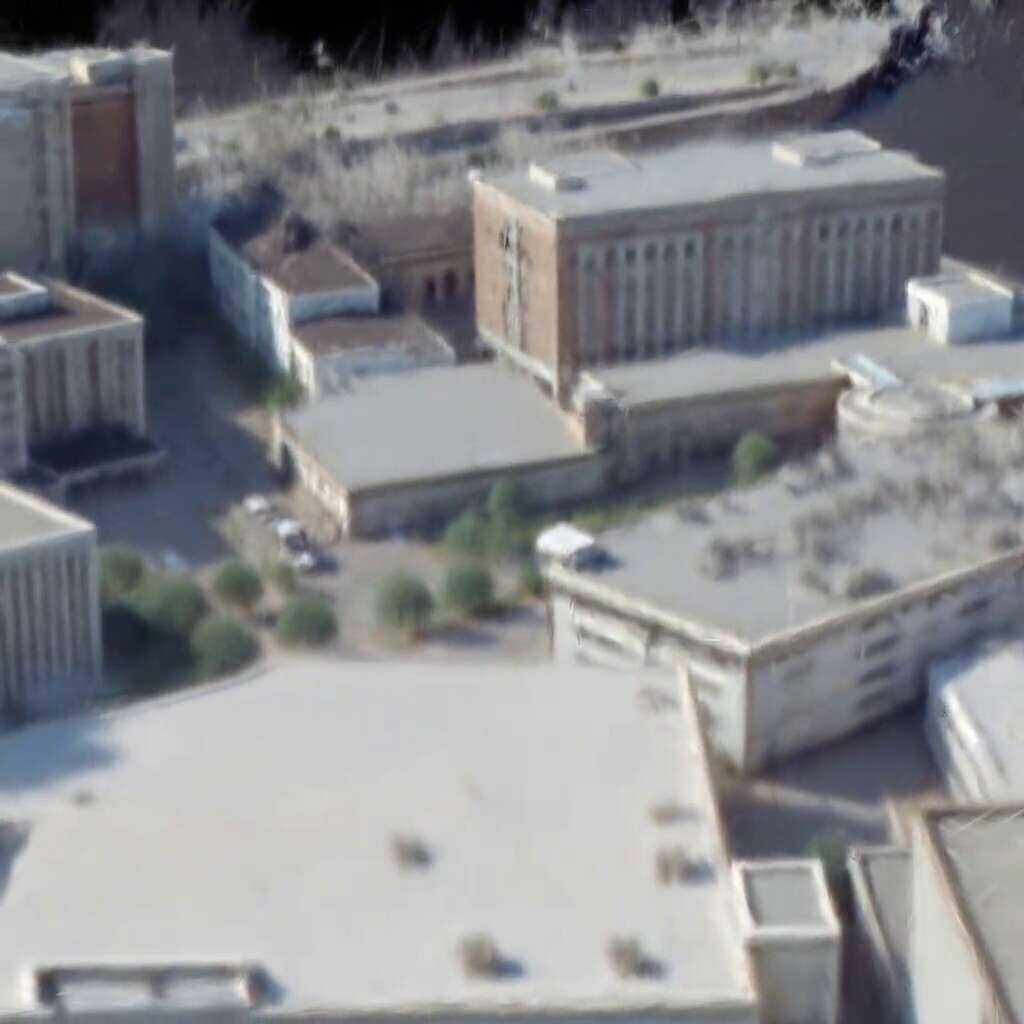} & 
    \imagecell[0.18]{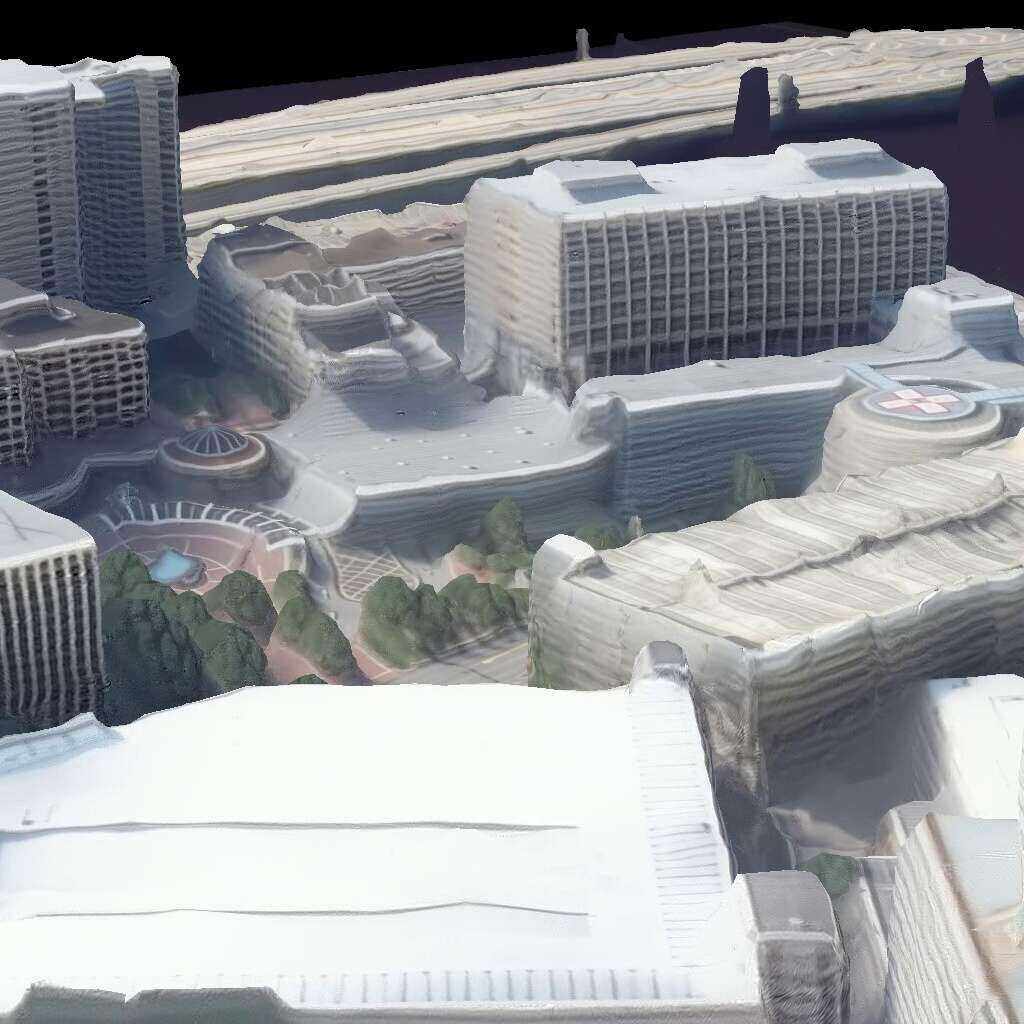} \\
    \vspace*{-10pt} \\

    \imagecell[0.18]{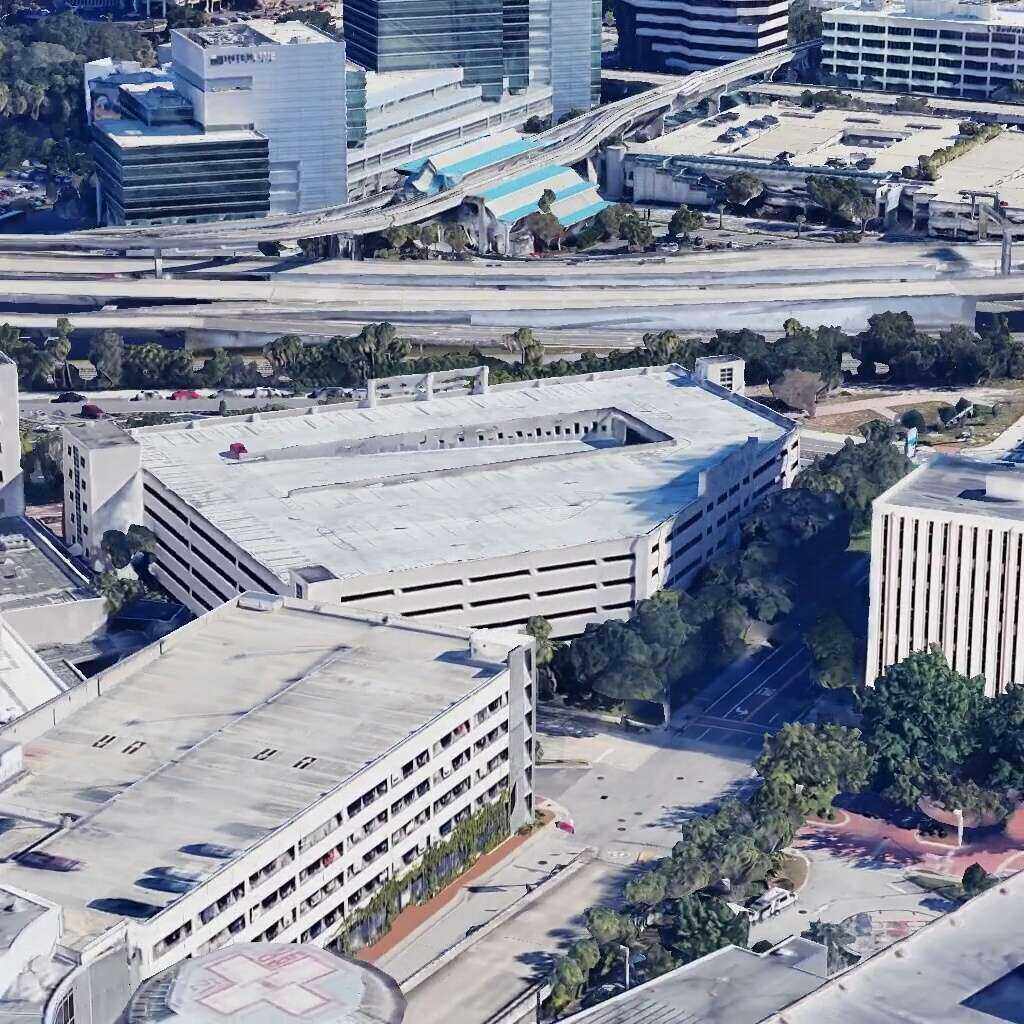} & 
    \imagecell[0.18]{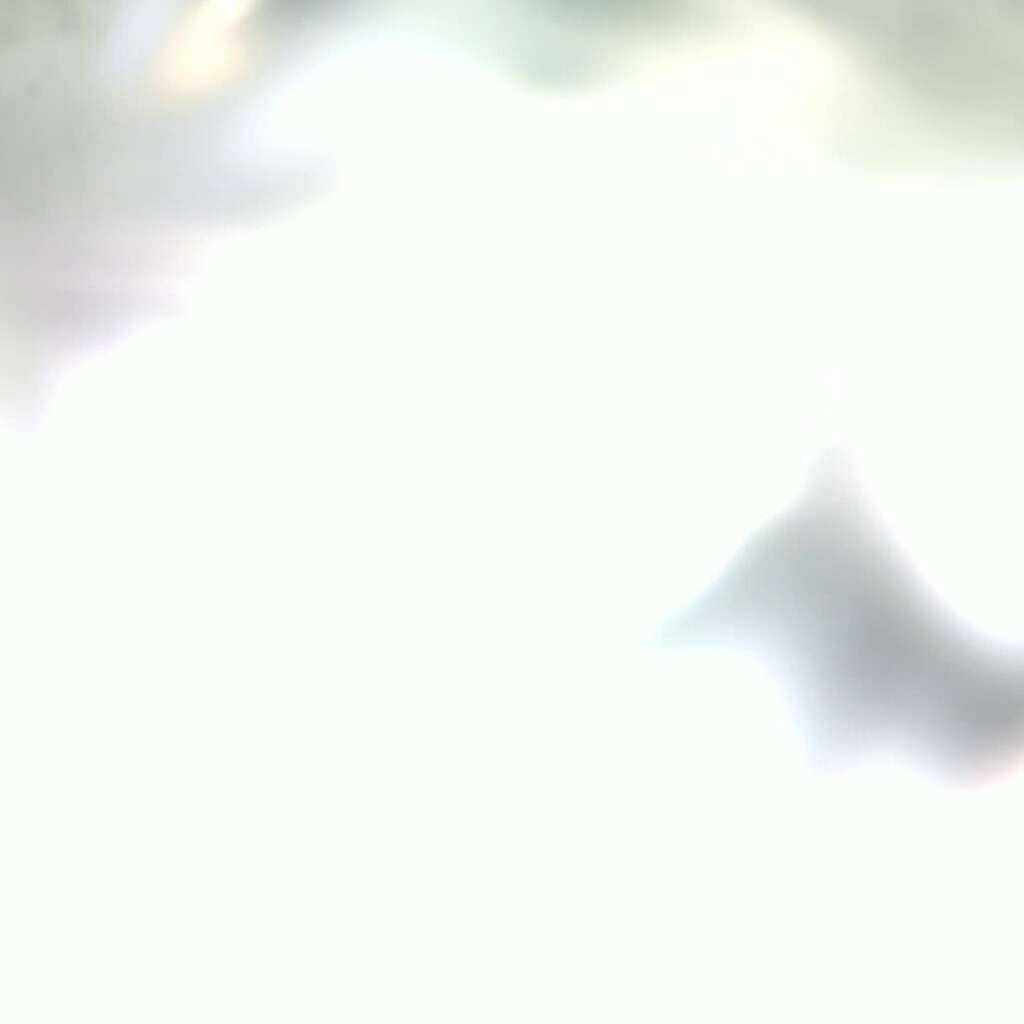} & 
    \imagecell[0.18]{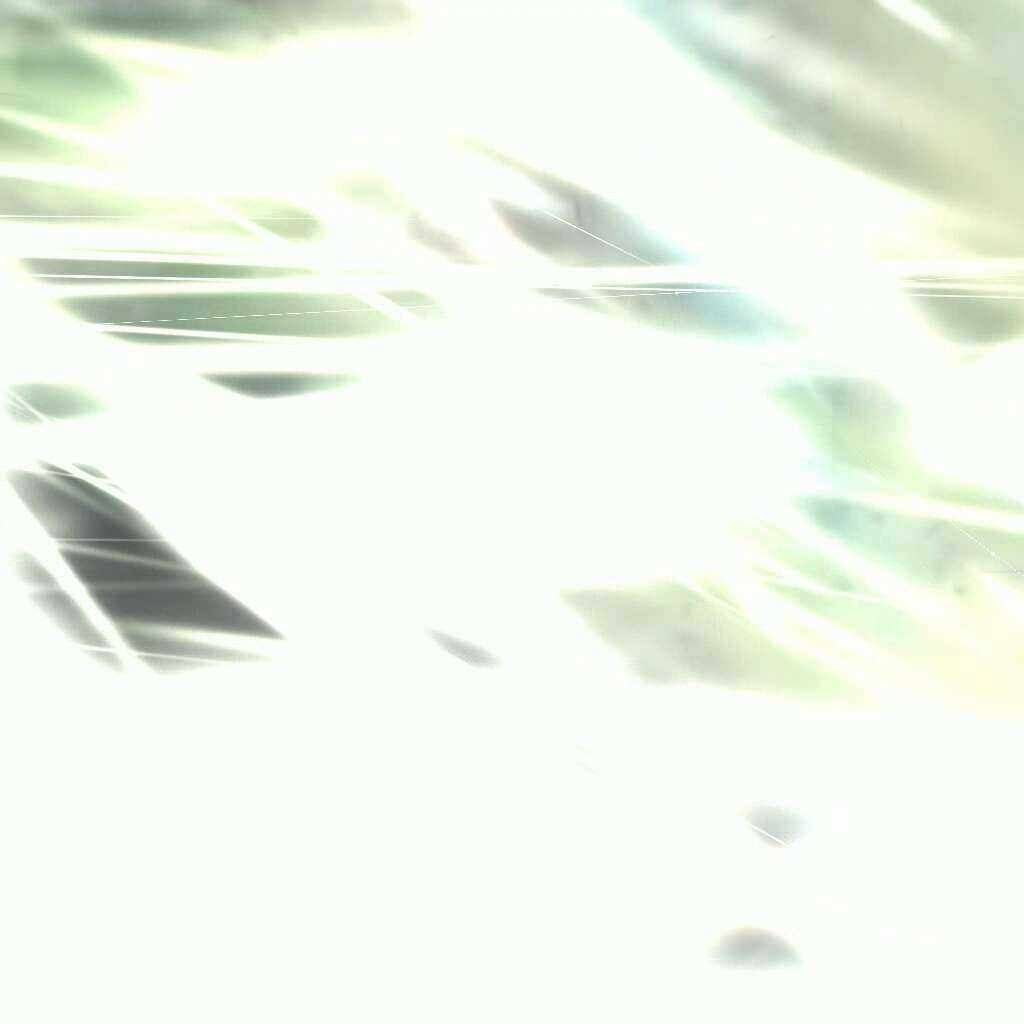} & 
    \imagecell[0.18]{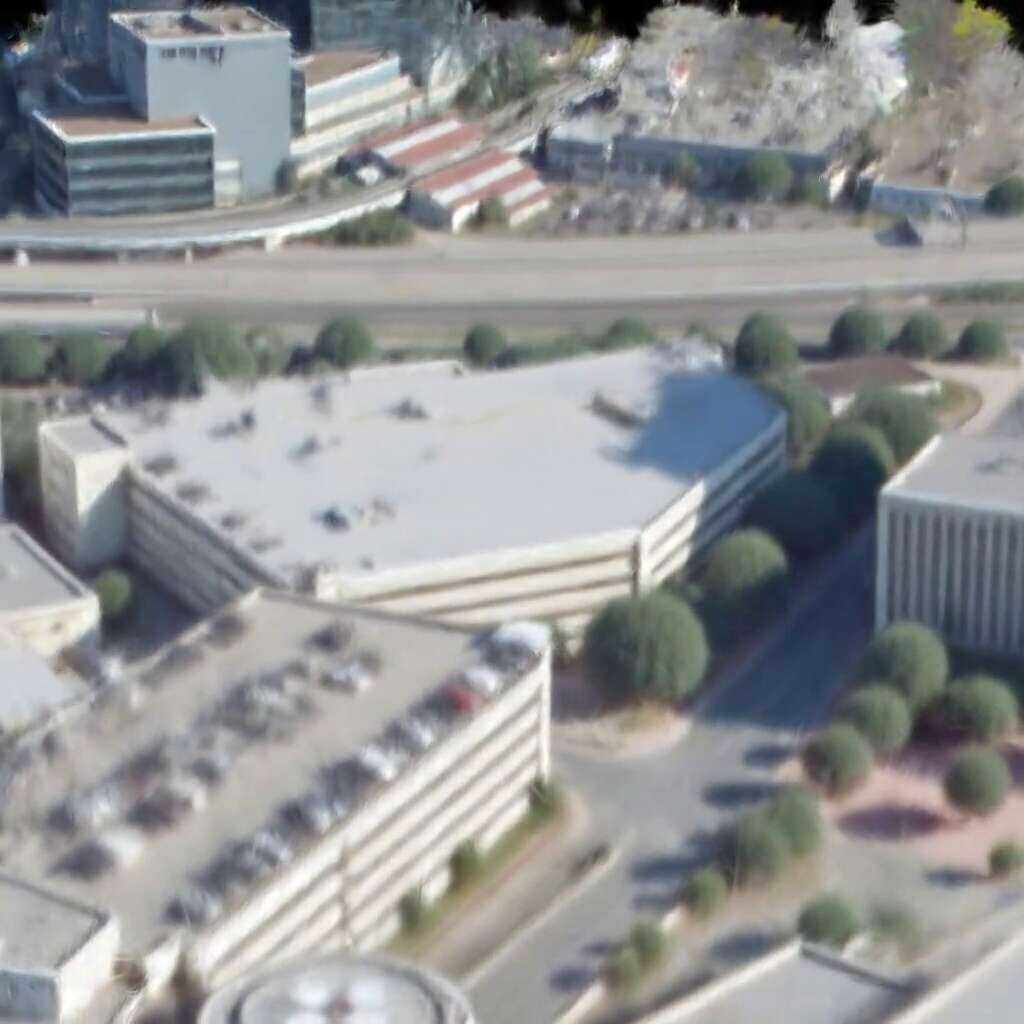} & 
    \imagecell[0.18]{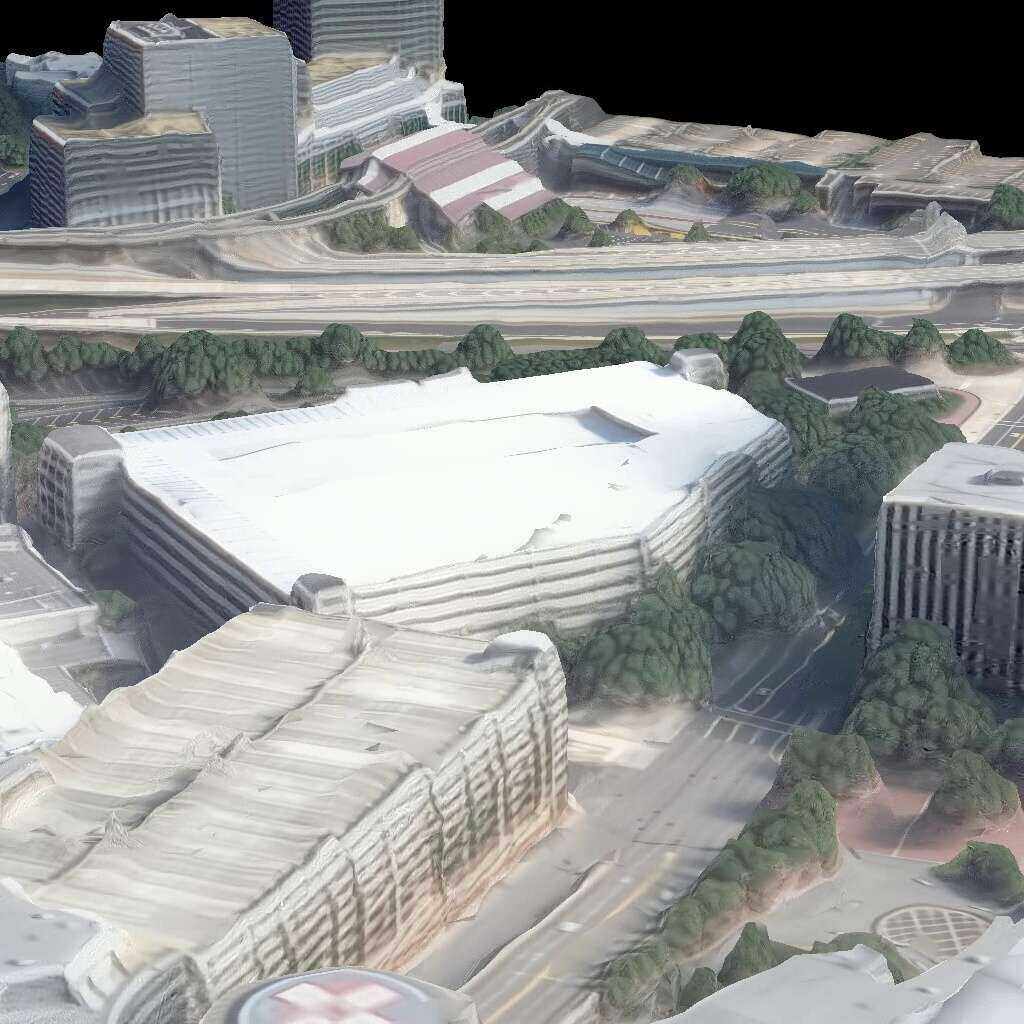} \\
    \vspace*{-10pt} \\
    
    \\
    \vspace*{-20pt}
    \\
    G.T. & 
    Mip-Splatting & 
    2DGS & 
    Skyfall-GS & 
    Ours \\
    
    \end{tabular}
    \end{spacing}
	\caption{ 
    \textbf{Results of the JAX\_214 scene in the DFC 2019 dataset}. 
    Compared to baselines, our method successfully achieves high-quality city reconstruction from satellite imagery. 
    Results of CityGS-X are removed since the method crashes while recovering this scene.
    }
    \label{fig:supp:qual-jax-214}
    \vspace*{-0.3cm}
\end{figure*}

\begin{figure*}[p]
	\centering
    \begin{spacing}{1} 
    \setlength\tabcolsep{1pt}
    \begin{tabular}{ccccc}

    \imagecell[0.18]{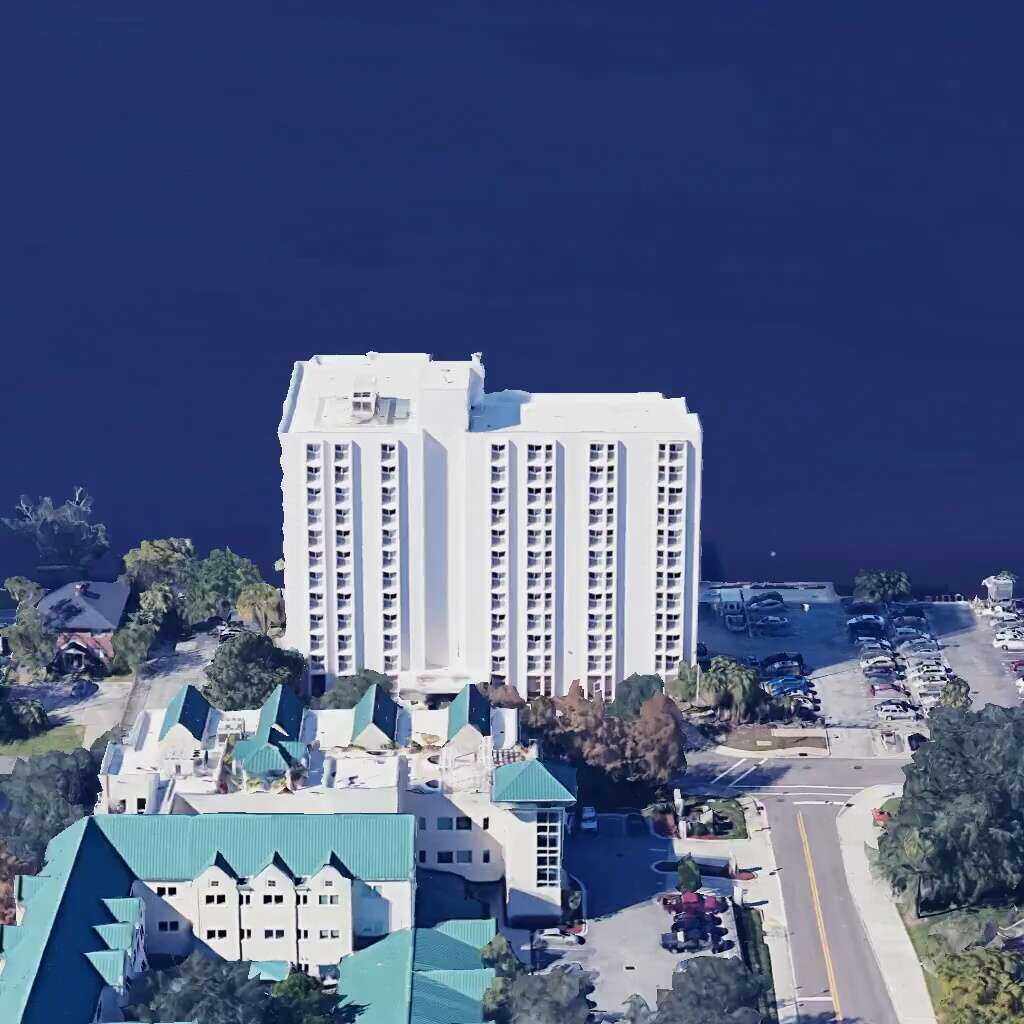} & 
    \imagecell[0.18]{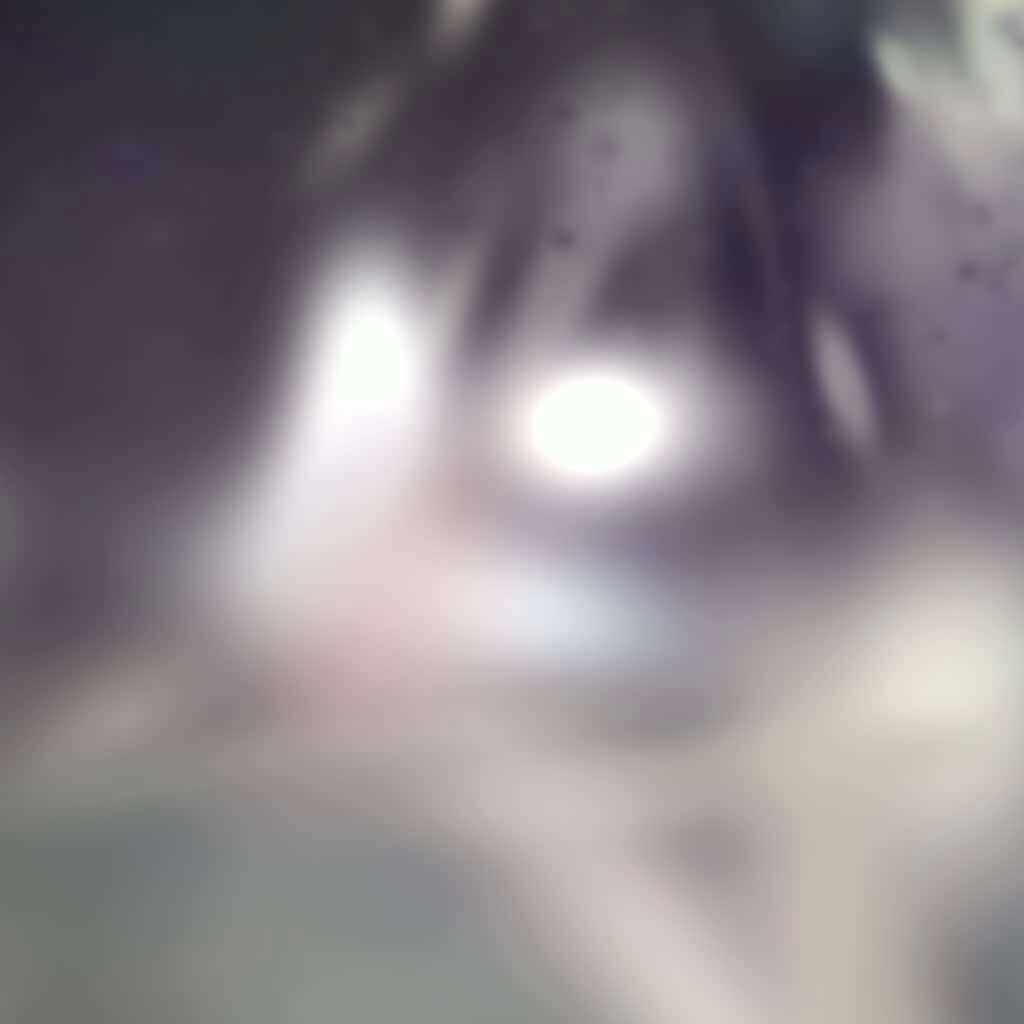} & 
    \imagecell[0.18]{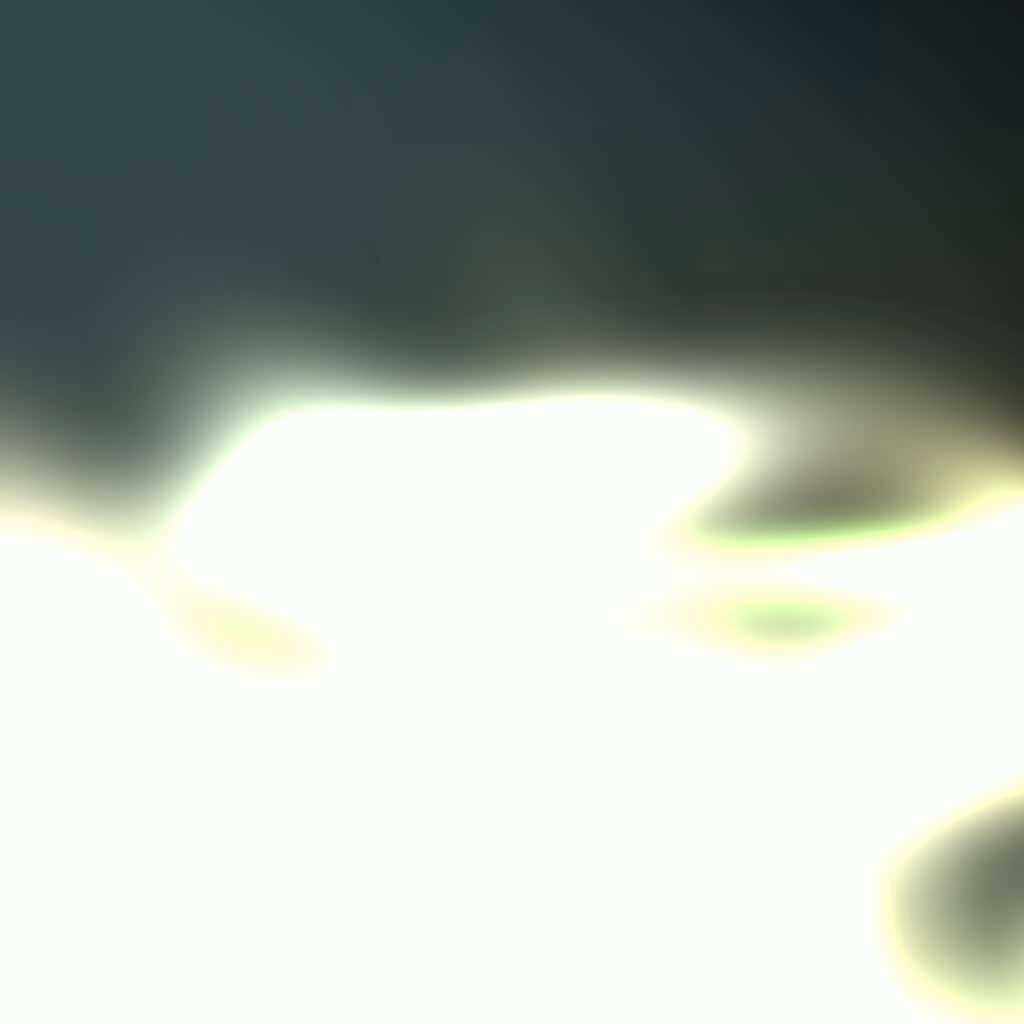} & 
    \imagecell[0.18]{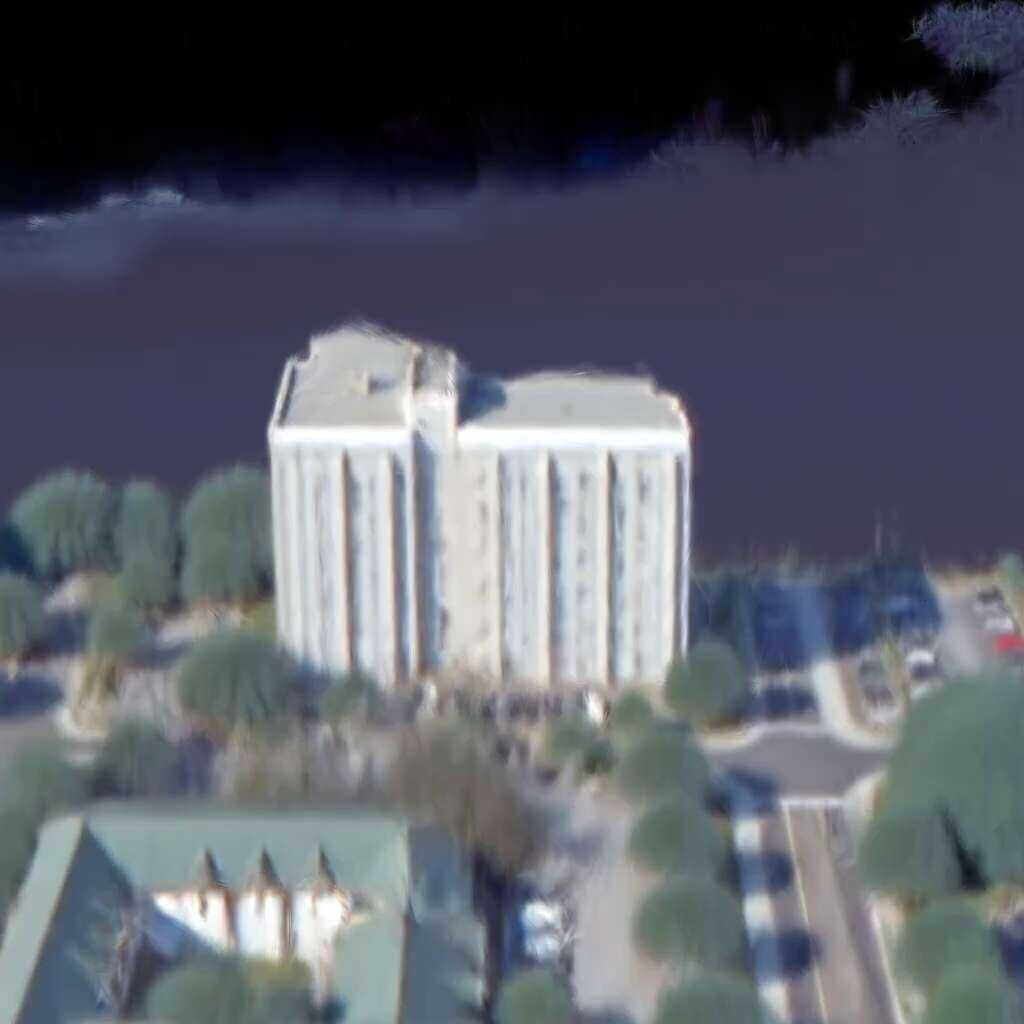} & 
    \imagecell[0.18]{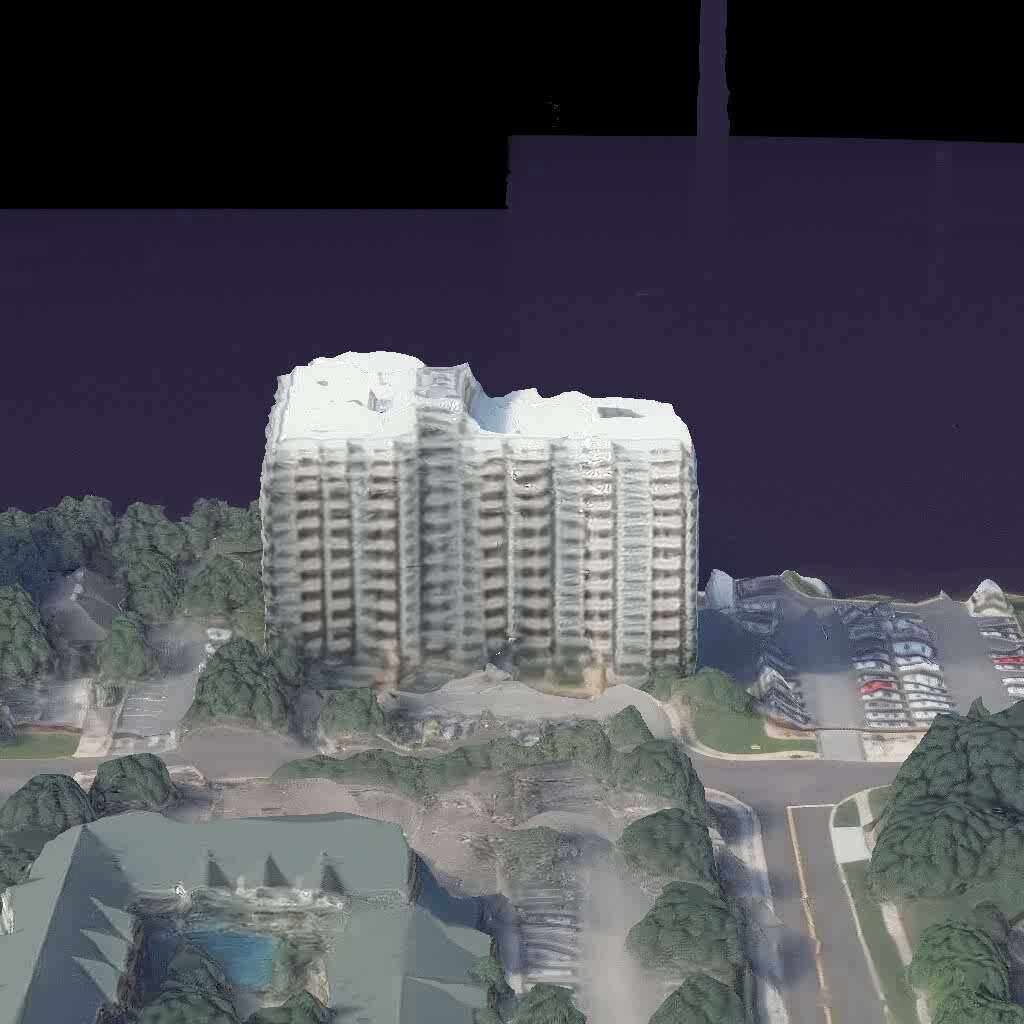} \\
    \vspace*{-10pt} \\

    \imagecell[0.18]{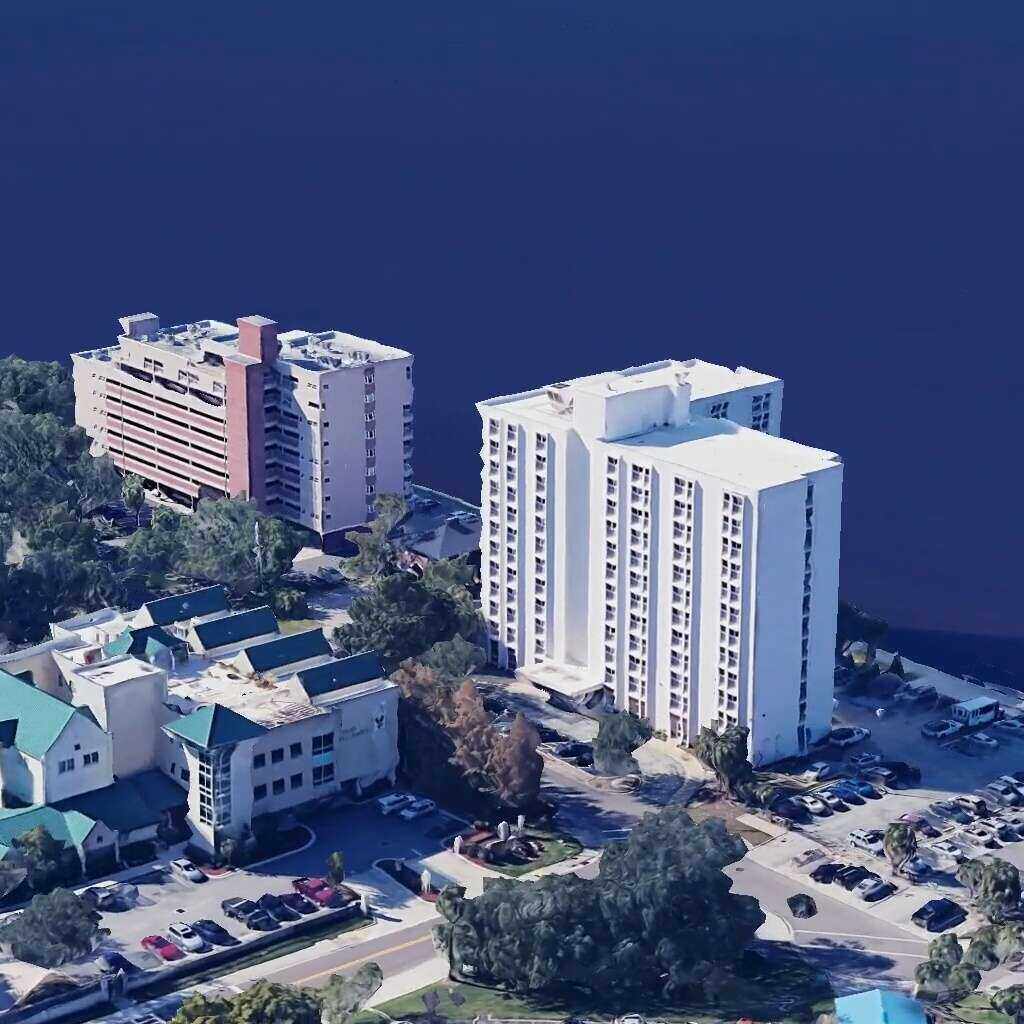} & 
    \imagecell[0.18]{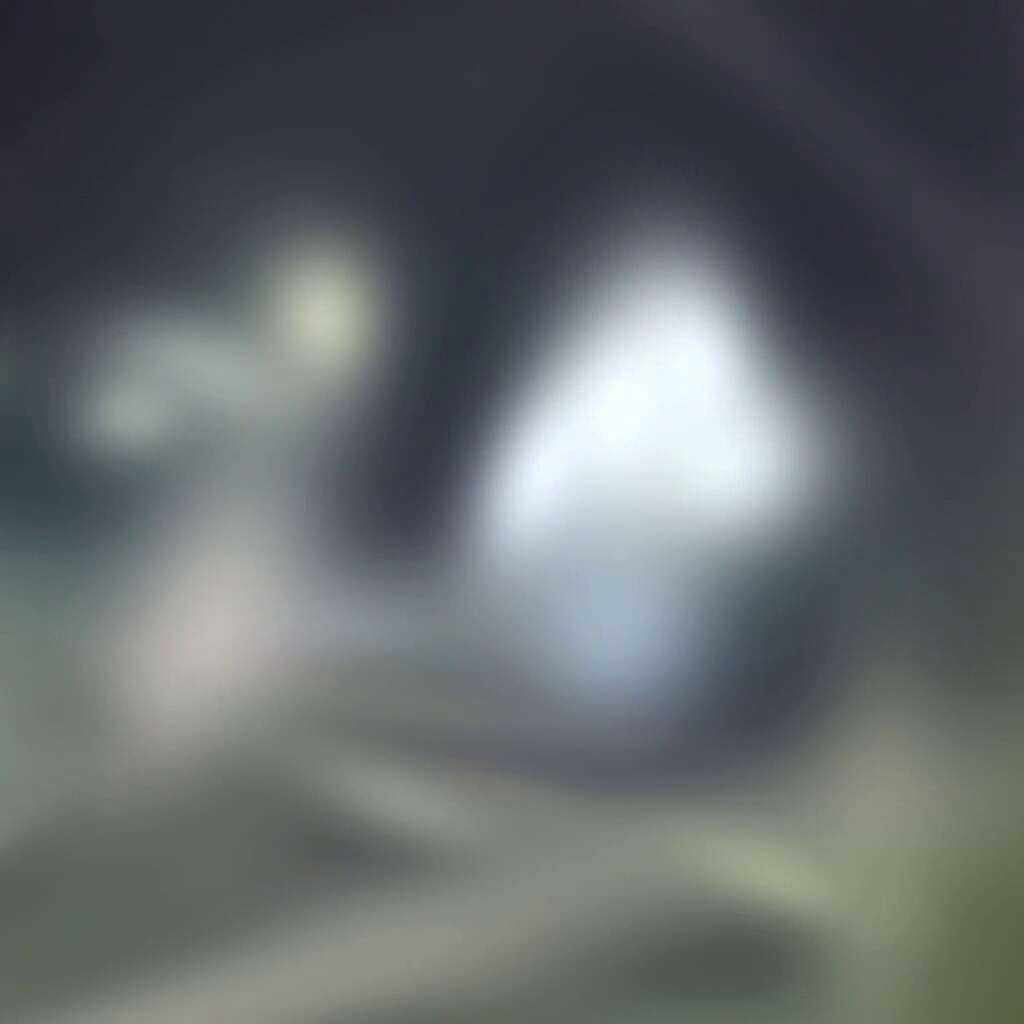} & 
    \imagecell[0.18]{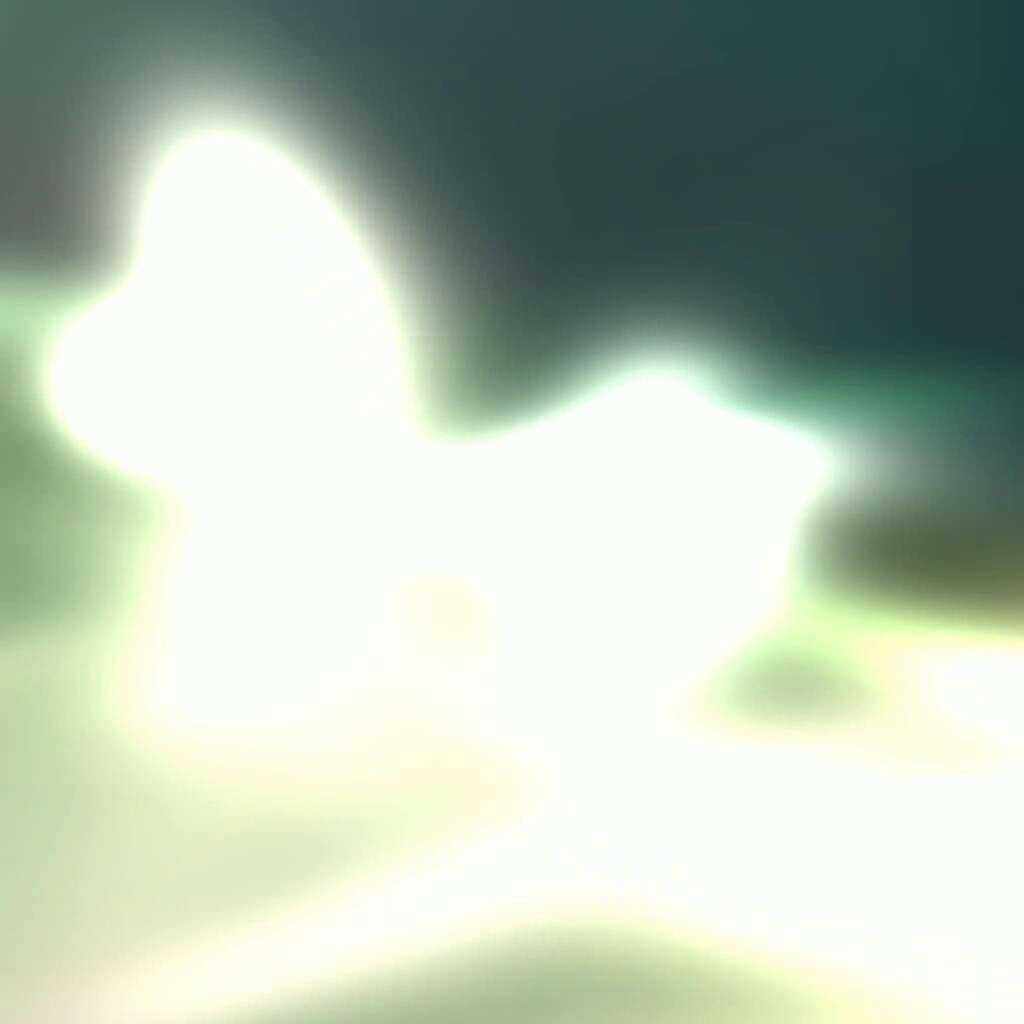} & 
    \imagecell[0.18]{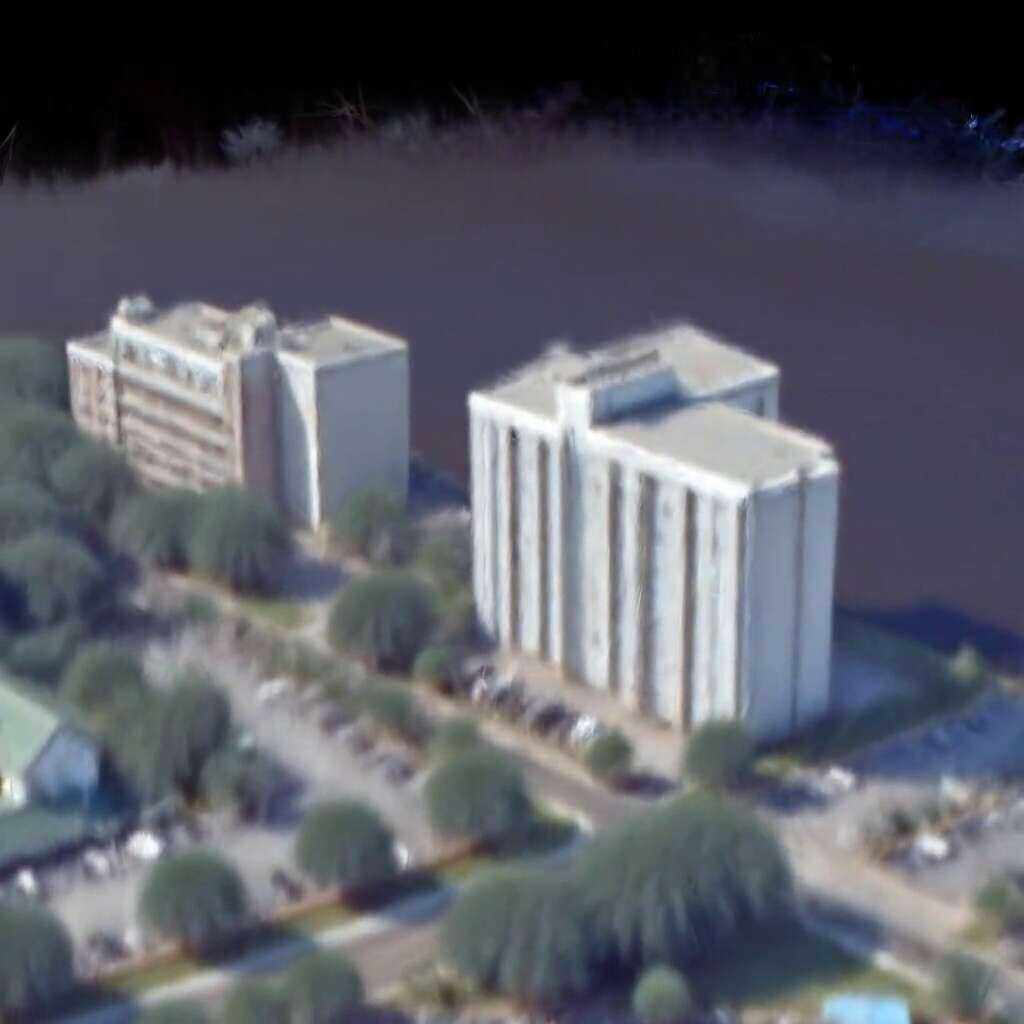} & 
    \imagecell[0.18]{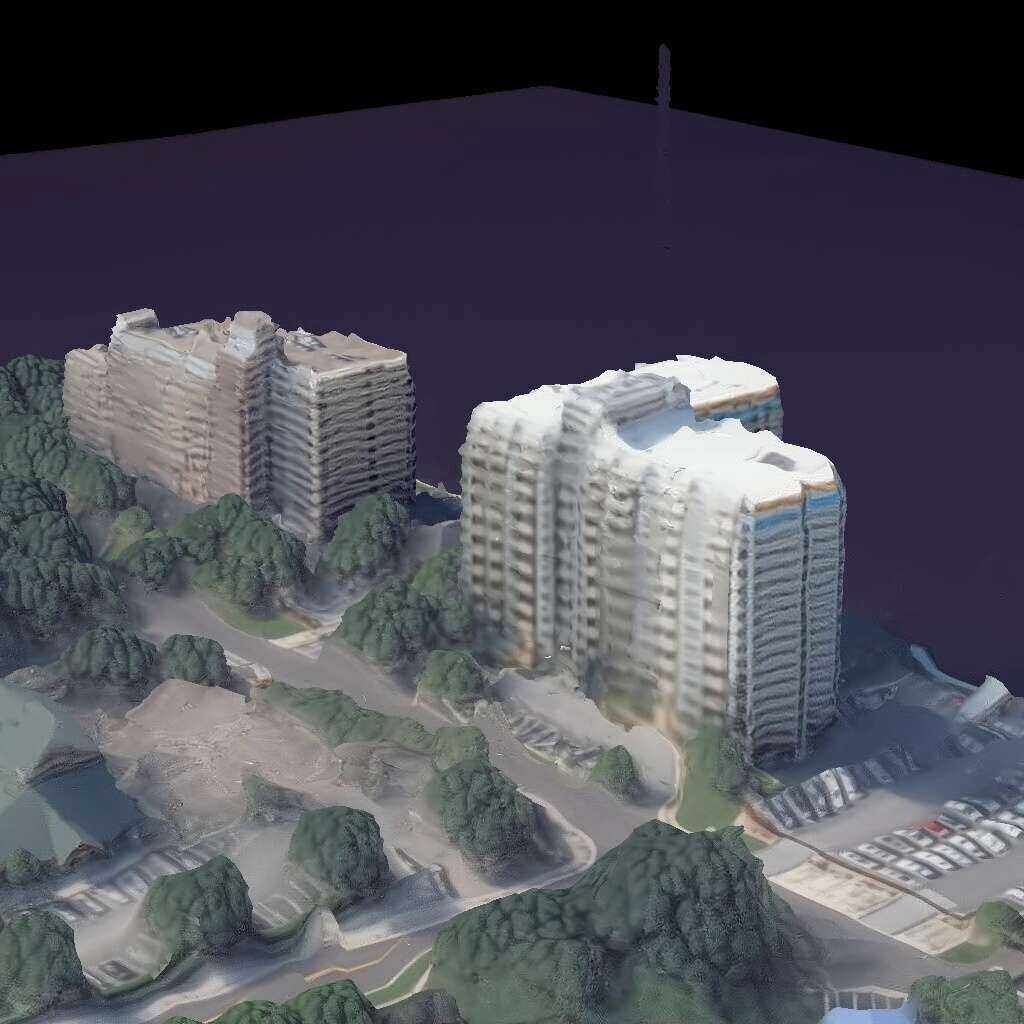} \\
    \vspace*{-10pt} \\
    
    \imagecell[0.18]{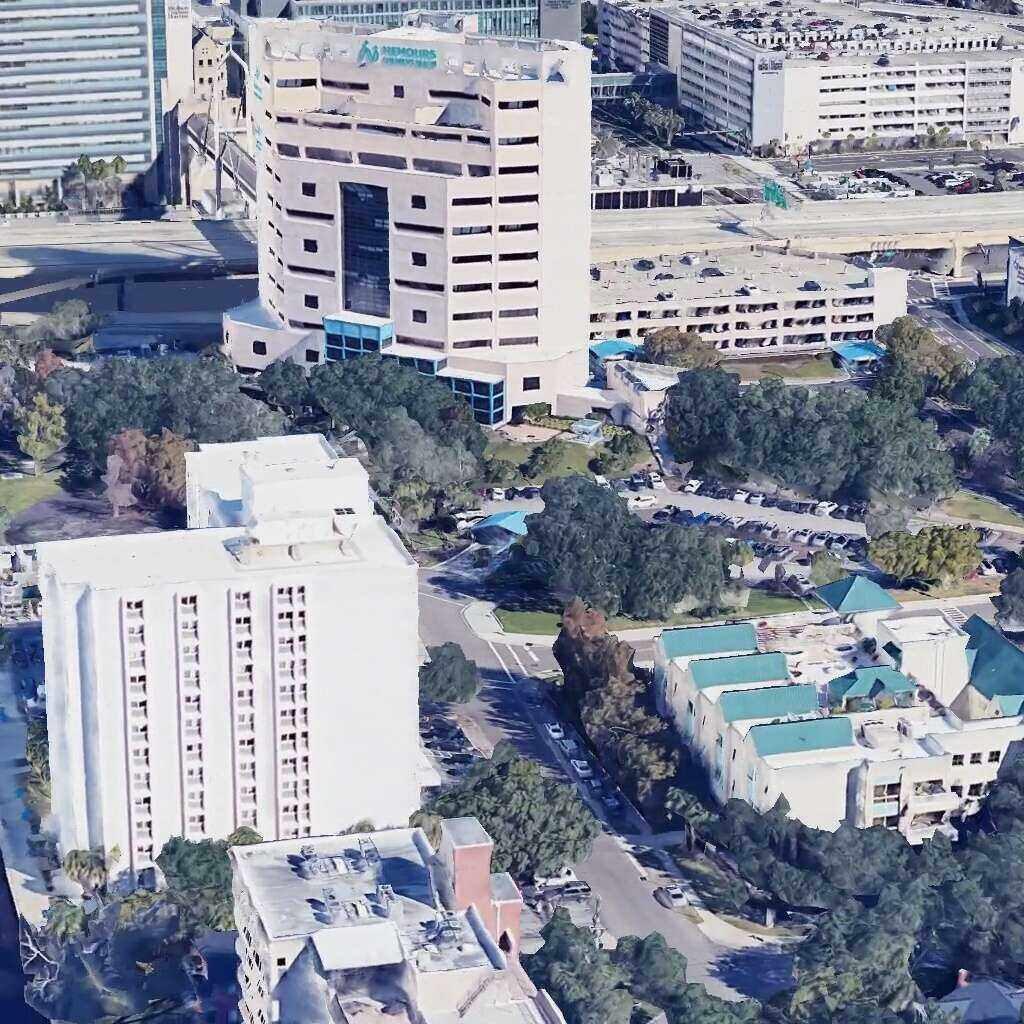} & 
    \imagecell[0.18]{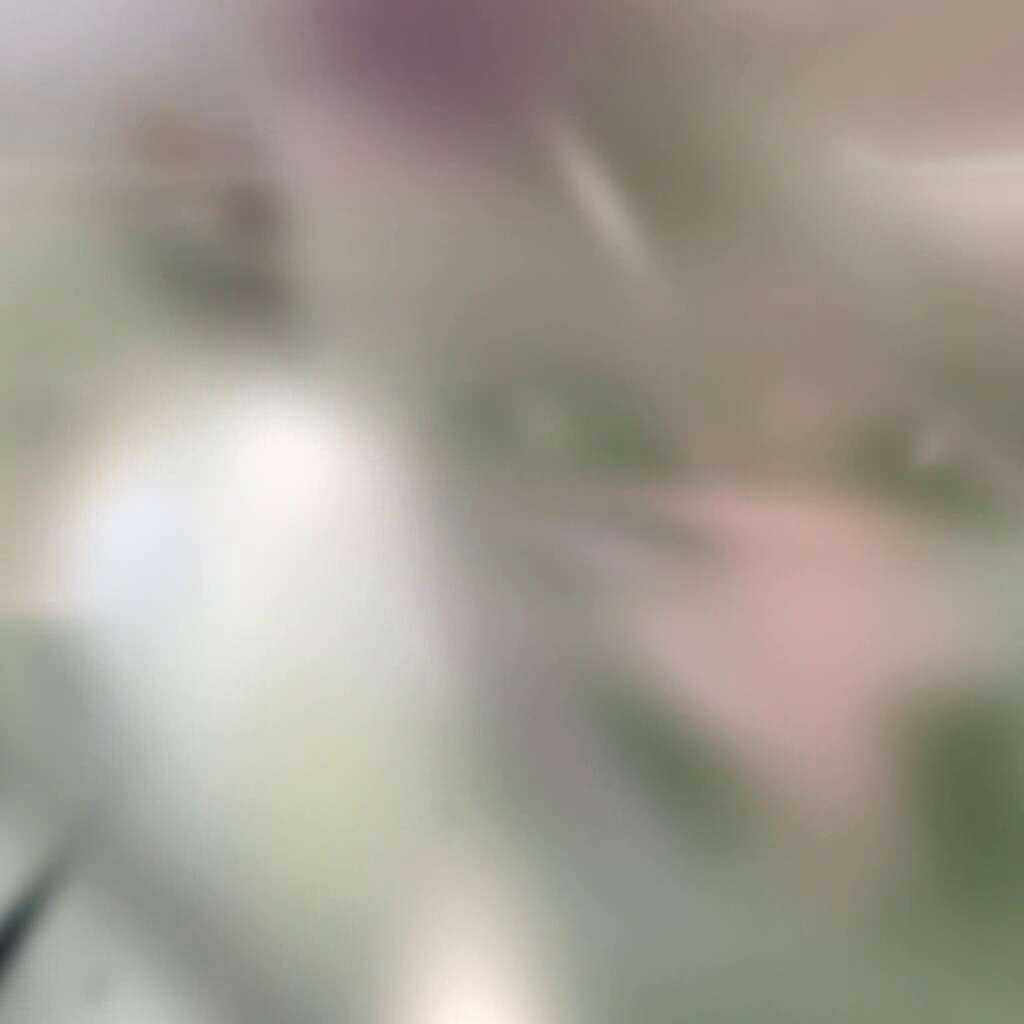} & 
    \imagecell[0.18]{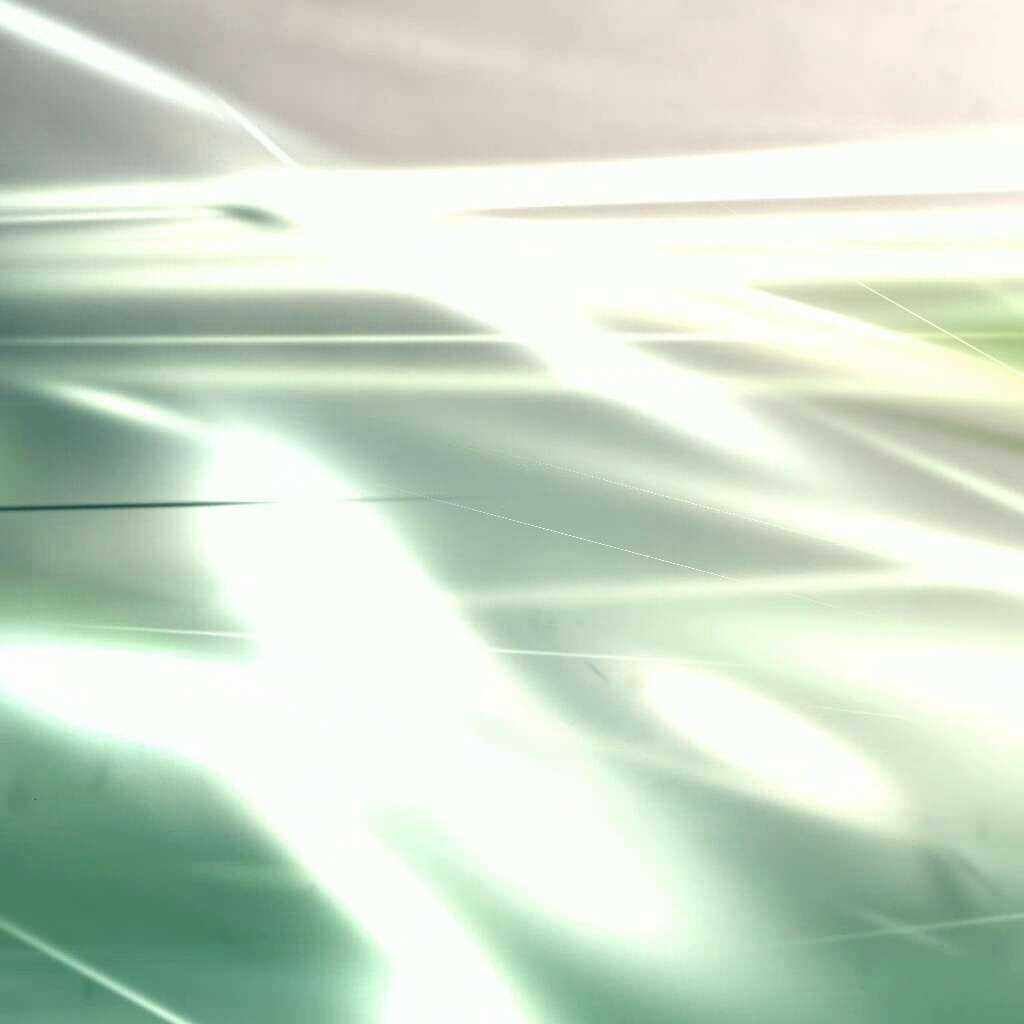} & 
    \imagecell[0.18]{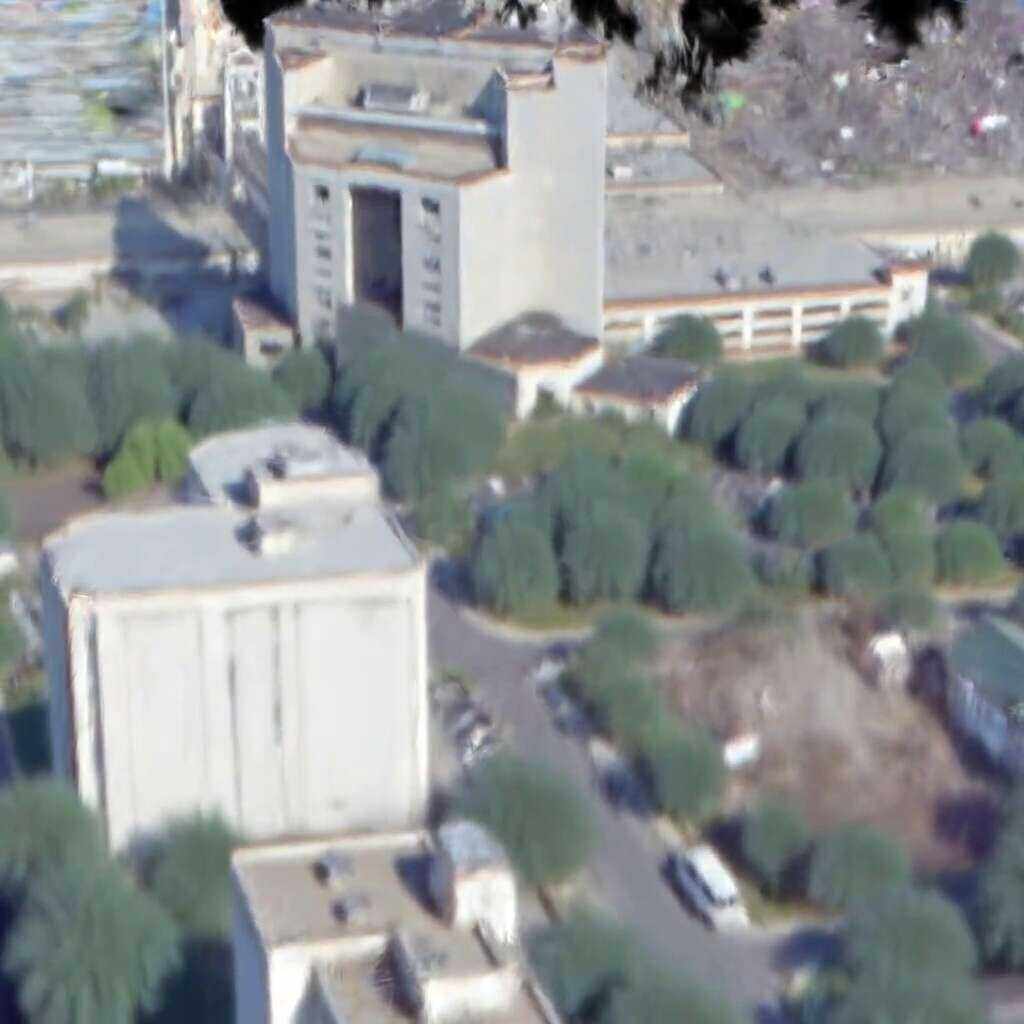} & 
    \imagecell[0.18]{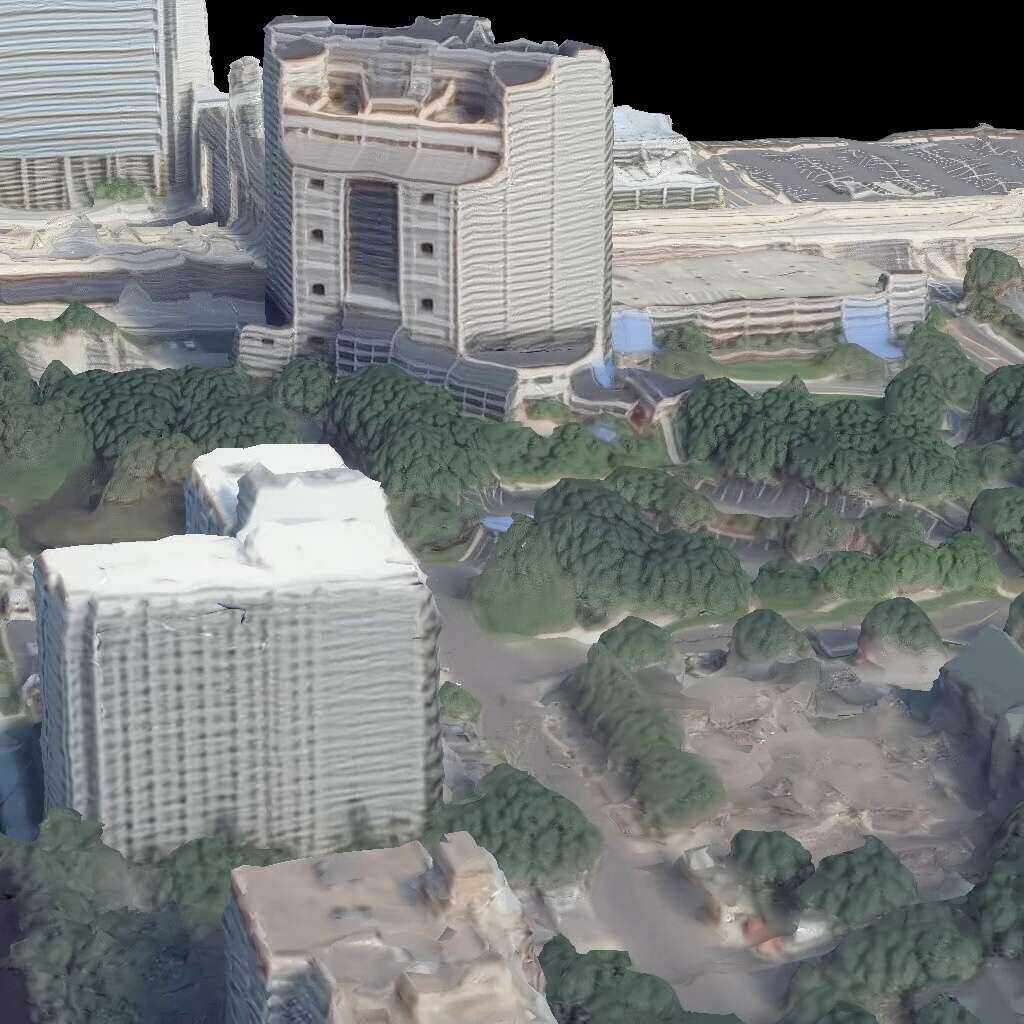} \\
    \vspace*{-10pt} \\

    \\
    \vspace*{-20pt}
    \\
    G.T. & 
    Mip-Splatting & 
    2DGS & 
    Skyfall-GS & 
    Ours \\
    
    \end{tabular}
    \end{spacing}
	\caption{ 
    \textbf{Results of the JAX\_260 scene in the DFC 2019 dataset}. 
    Compared to baselines, our method successfully achieves high-quality city reconstruction from satellite imagery. 
    Results of CityGS-X are removed since the method crashes while recovering this scene.
    }
    \label{fig:supp:qual-jax-260}
    \vspace*{-0.3cm}
\end{figure*}

\begin{figure*}[p]
	\centering
    \begin{spacing}{1} 
    \setlength\tabcolsep{1pt}
    \begin{tabular}{cccccc}

    \imagecell[0.16]{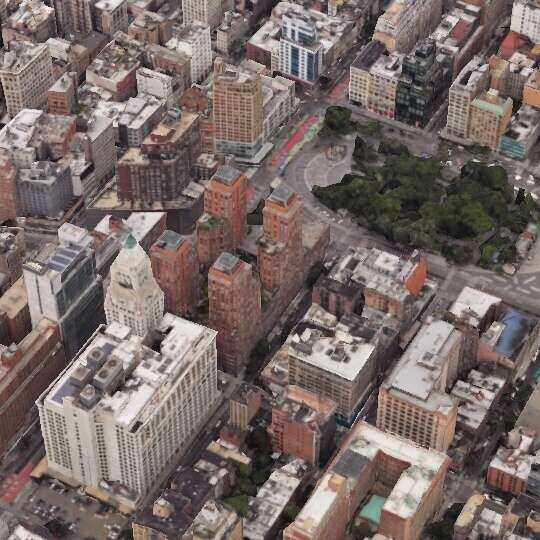} & 
    \imagecell[0.16]{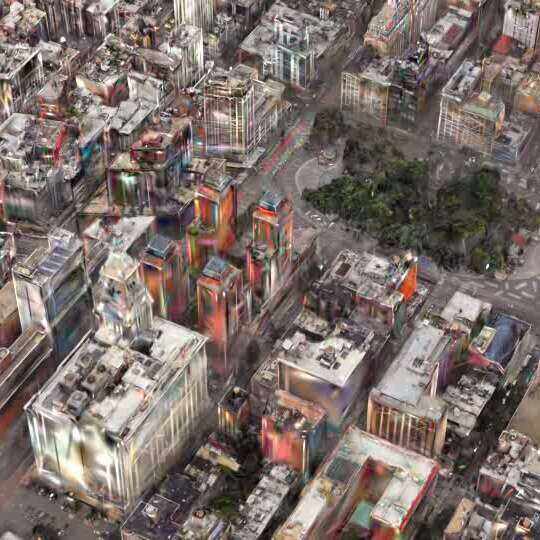} & 
    \imagecell[0.16]{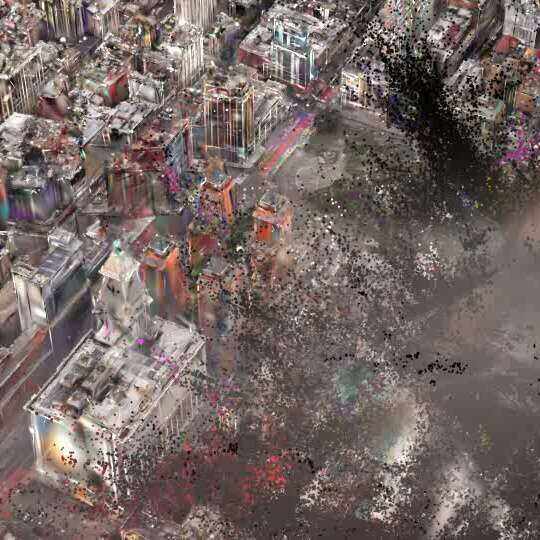} & 
    \imagecell[0.16]{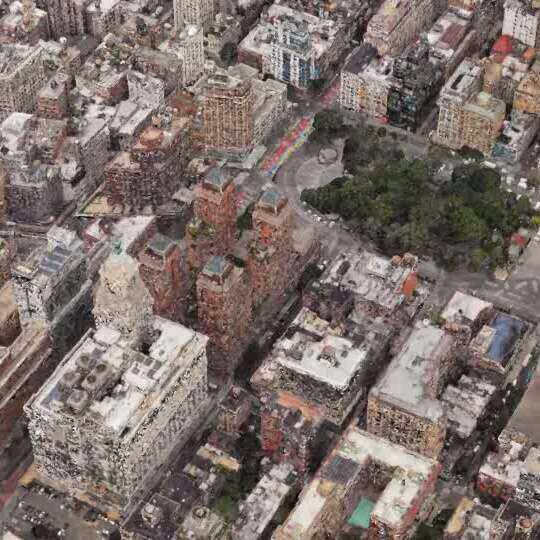} & 
    \imagecell[0.16]{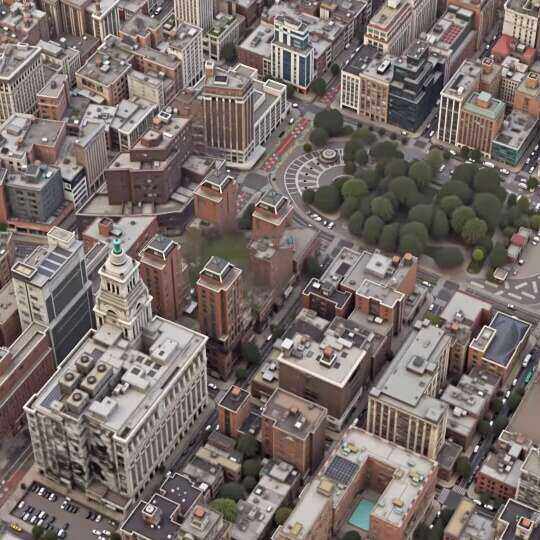} & 
    \imagecell[0.16]{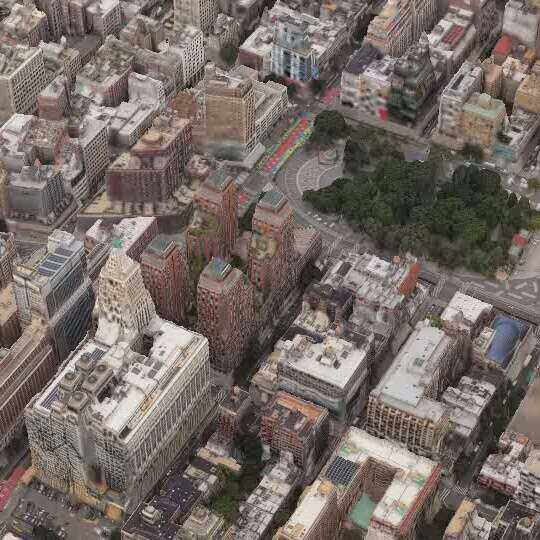} \\
    \vspace*{-10pt} \\   

    \imagecell[0.16]{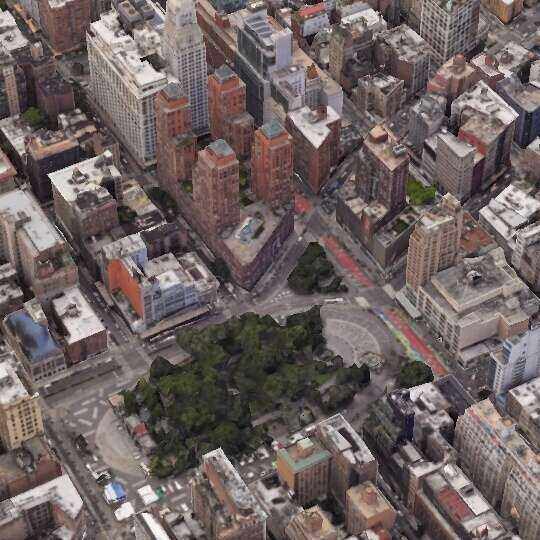} & 
    \imagecell[0.16]{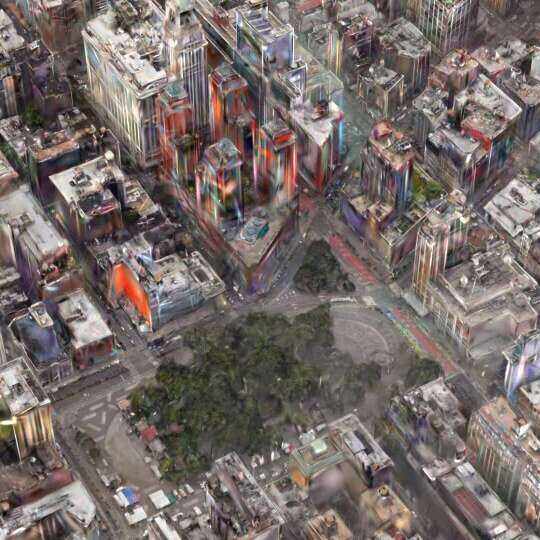} & 
    \imagecell[0.16]{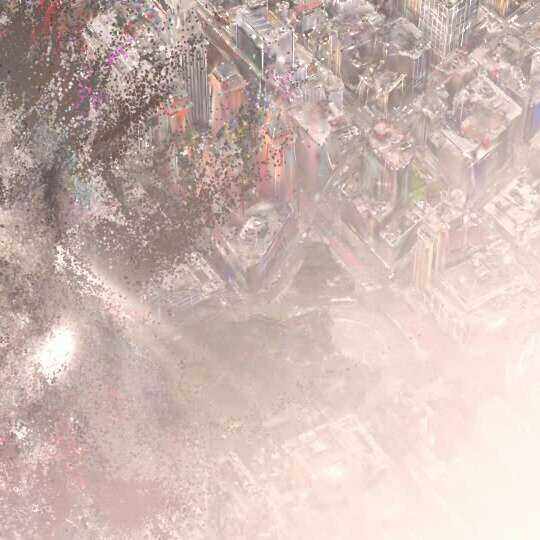} & 
    \imagecell[0.16]{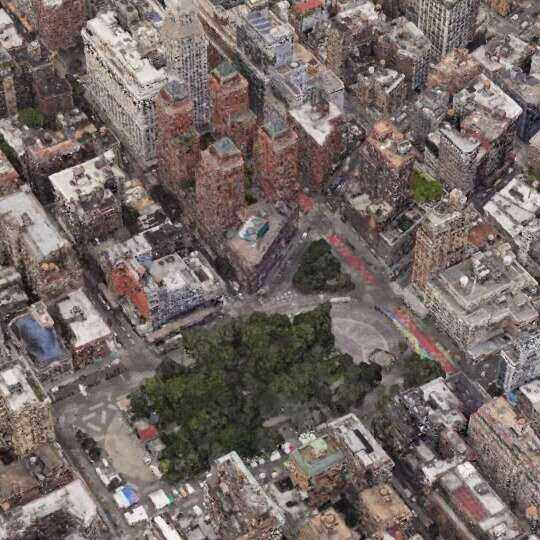} & 
    \imagecell[0.16]{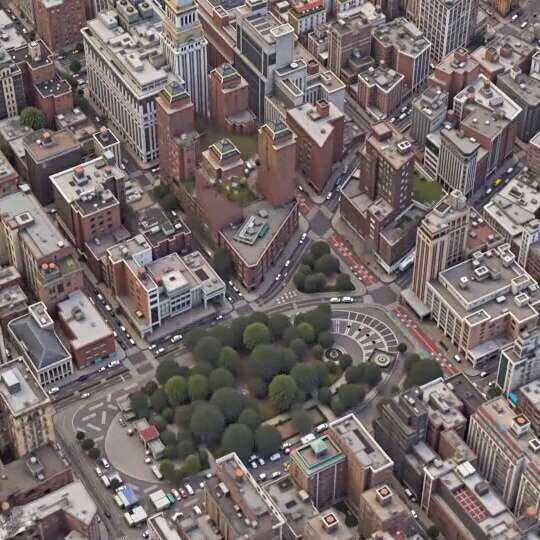} & 
    \imagecell[0.16]{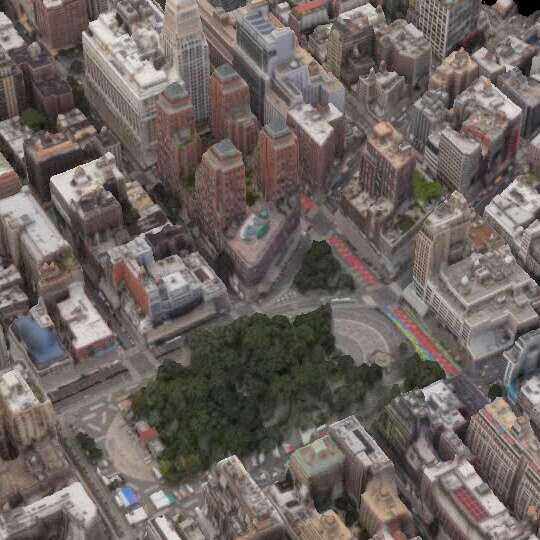} \\
    \vspace*{-10pt} \\

    \imagecell[0.16]{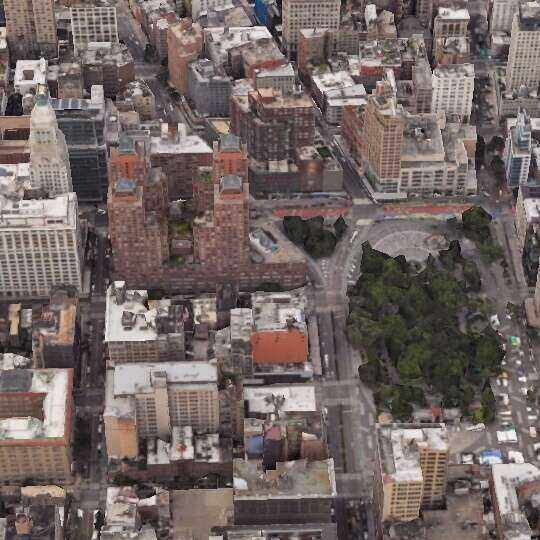} & 
    \imagecell[0.16]{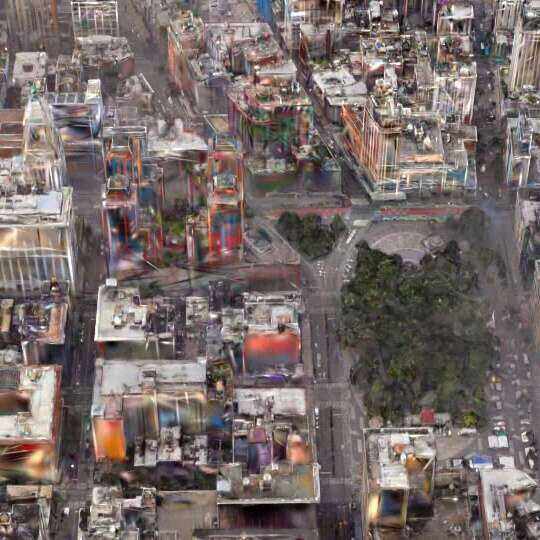} & 
    \imagecell[0.16]{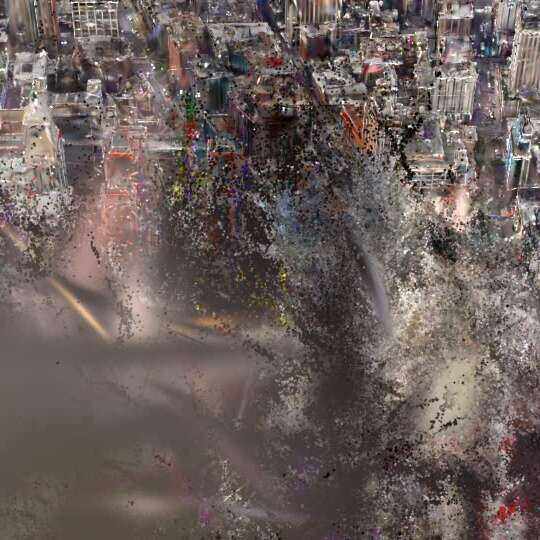} & 
    \imagecell[0.16]{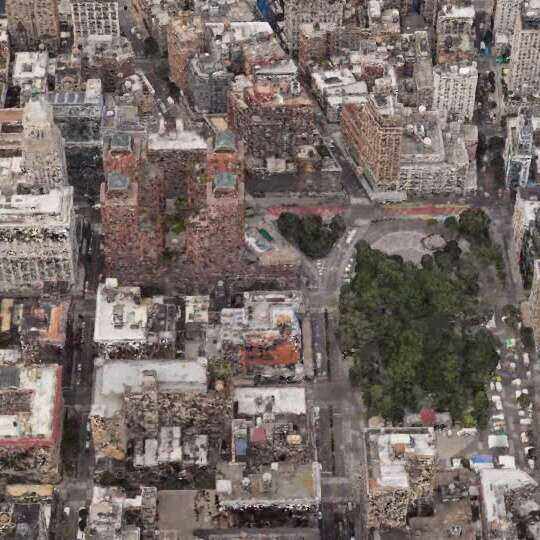} & 
    \imagecell[0.16]{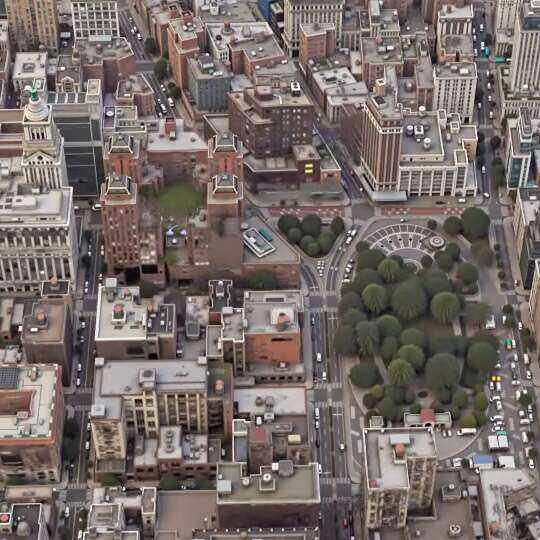} & 
    \imagecell[0.16]{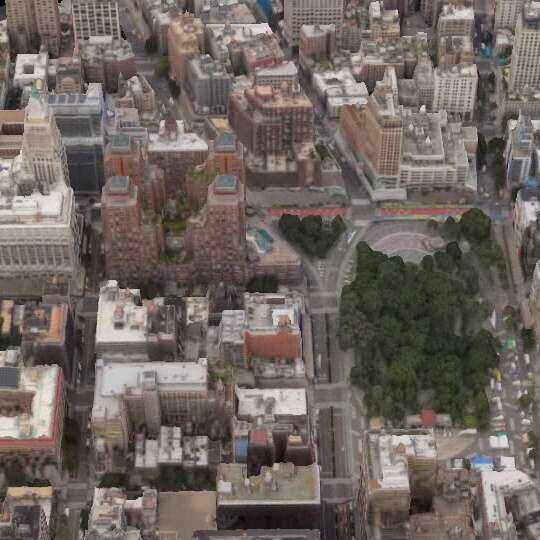} \\
    \vspace*{-10pt} \\
    
    \\
    \vspace*{-20pt}
    \\
    G.T. & 
    Mip-Splatting & 
    2DGS & 
    CityGS-X &
    Skyfall-GS & 
    Ours \\
    
    \end{tabular}
    \end{spacing}
	\caption{ 
    \textbf{Results of the NYC\_010 scene in the GoogleEarth dataset}. 
    Compared to baselines, our method successfully achieves high-quality city reconstruction from satellite imagery. 
    }
    \label{fig:supp:qual-nyc-010}
    \vspace*{-0.3cm}
\end{figure*}

\begin{figure*}[p]
	\centering
    \begin{spacing}{1} 
    \setlength\tabcolsep{1pt}
    \begin{tabular}{cccccc}

    \imagecell[0.16]{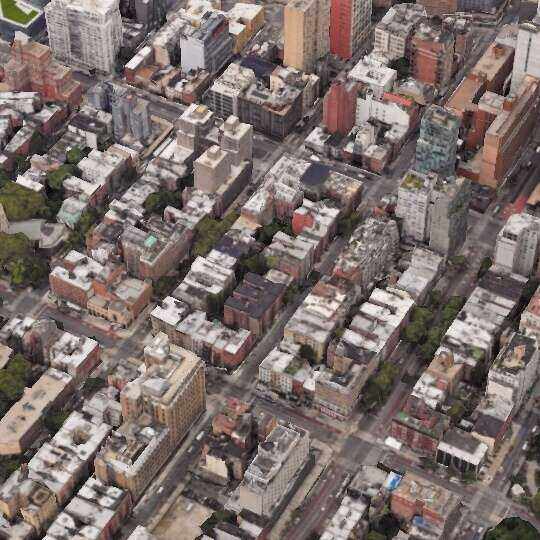} & 
    \imagecell[0.16]{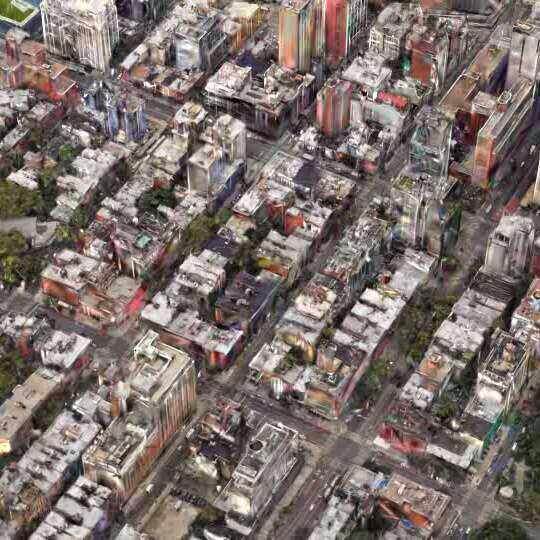} & 
    \imagecell[0.16]{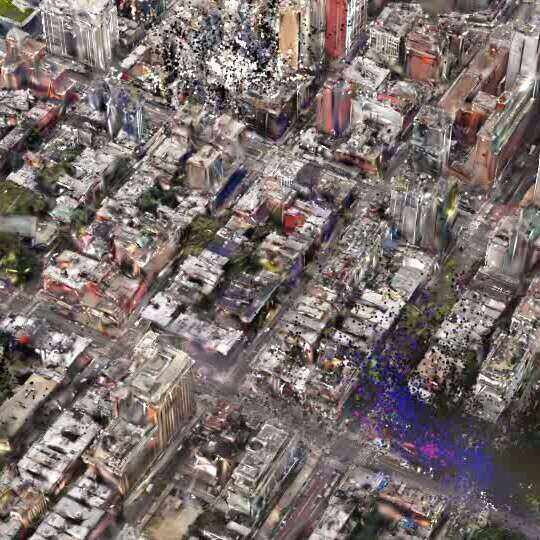} & 
    \imagecell[0.16]{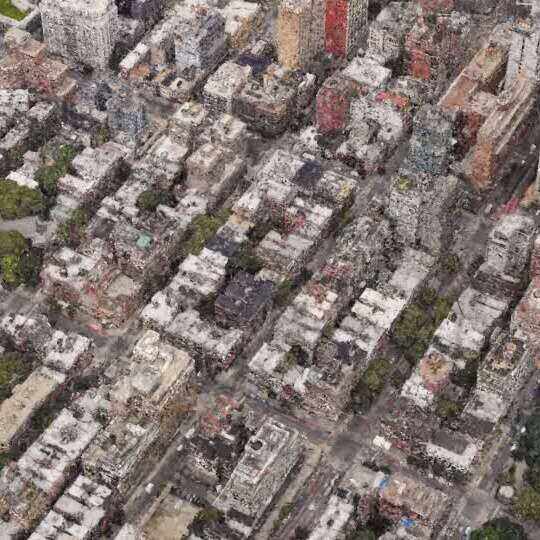} & 
    \imagecell[0.16]{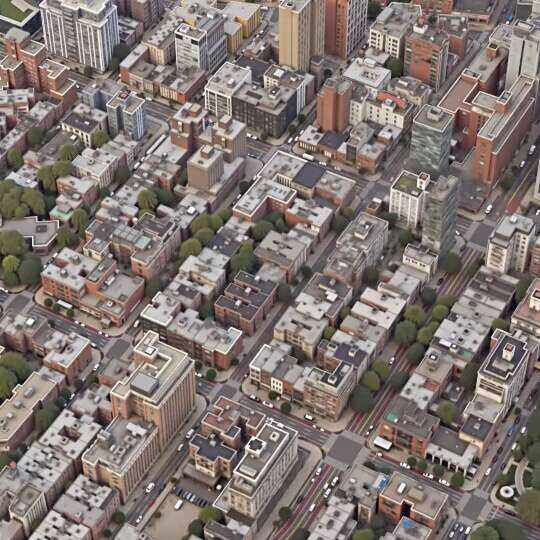} & 
    \imagecell[0.16]{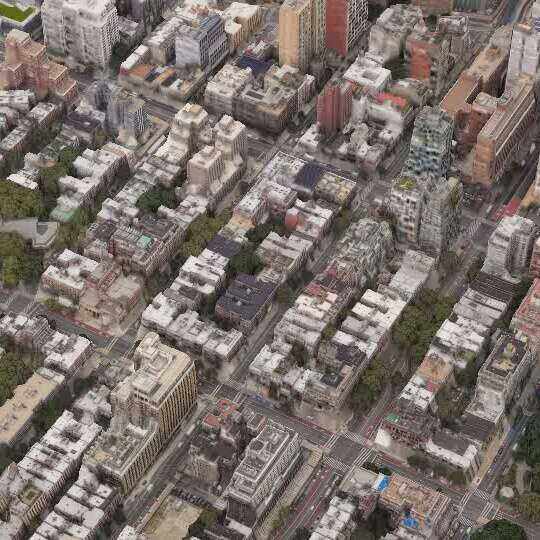} \\
    \vspace*{-10pt} \\

    \imagecell[0.16]{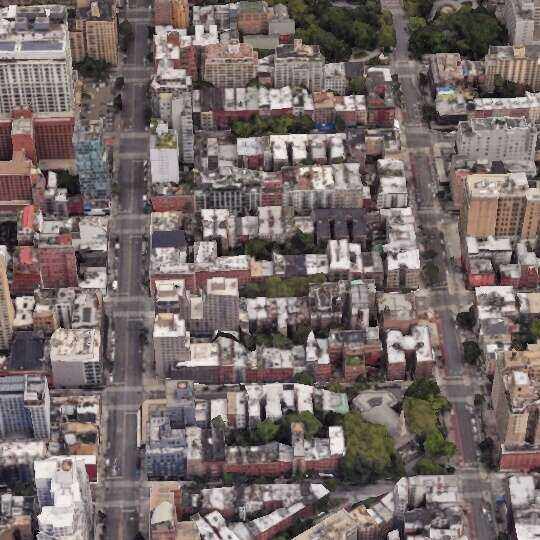} & 
    \imagecell[0.16]{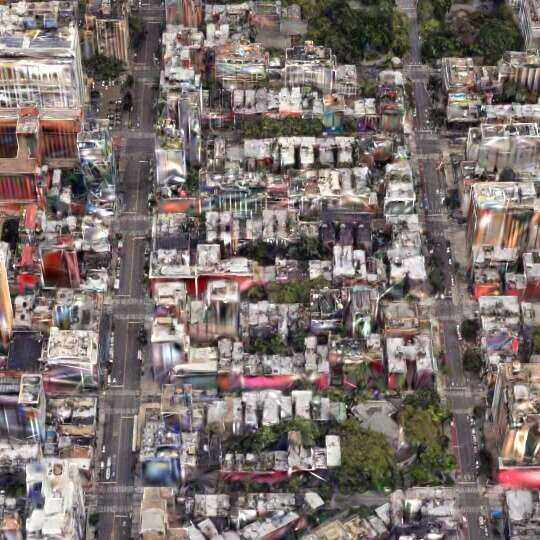} & 
    \imagecell[0.16]{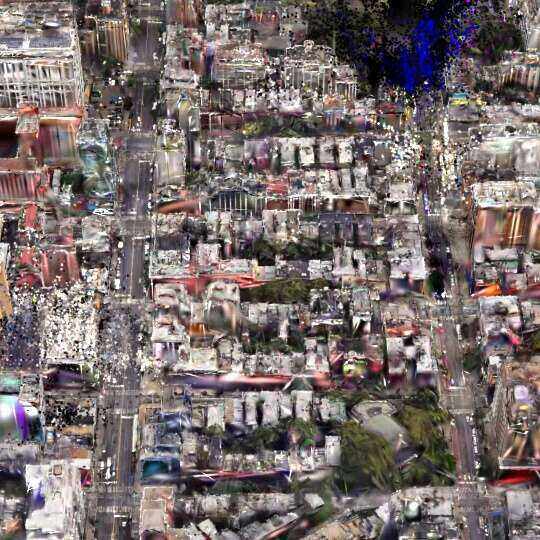} & 
    \imagecell[0.16]{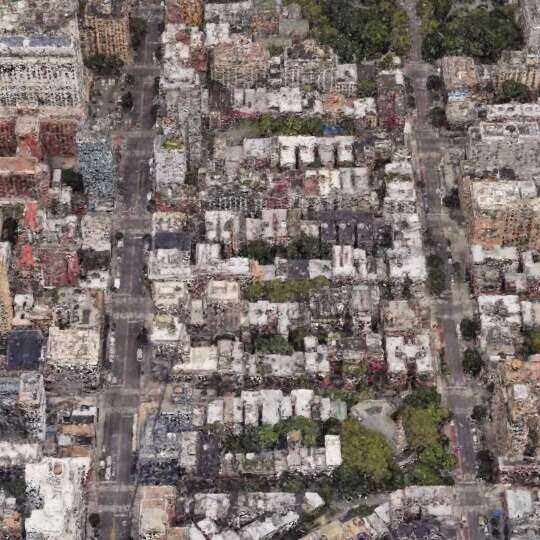} & 
    \imagecell[0.16]{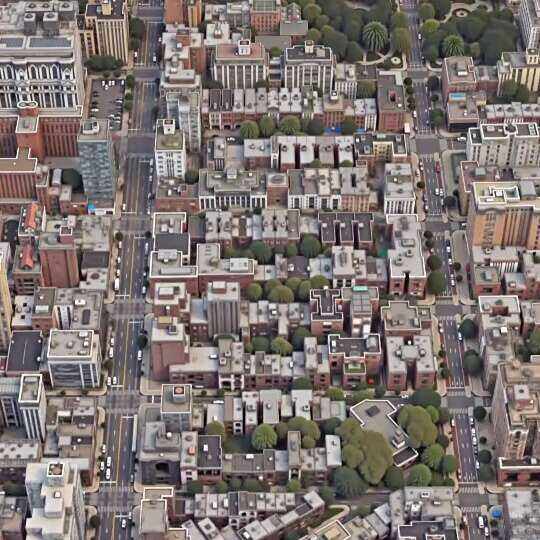} & 
    \imagecell[0.16]{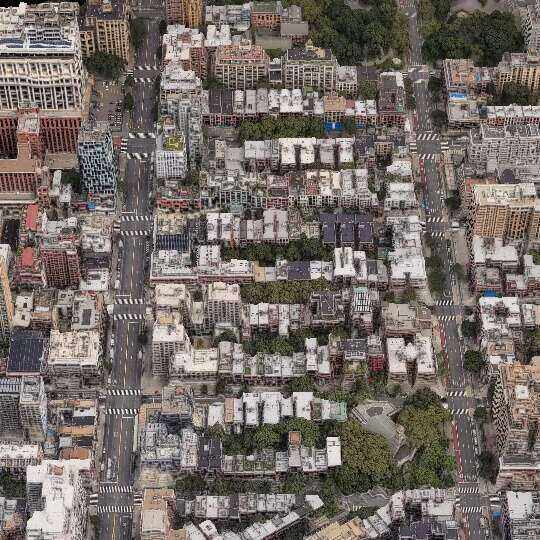} \\
    \vspace*{-10pt} \\

    \imagecell[0.16]{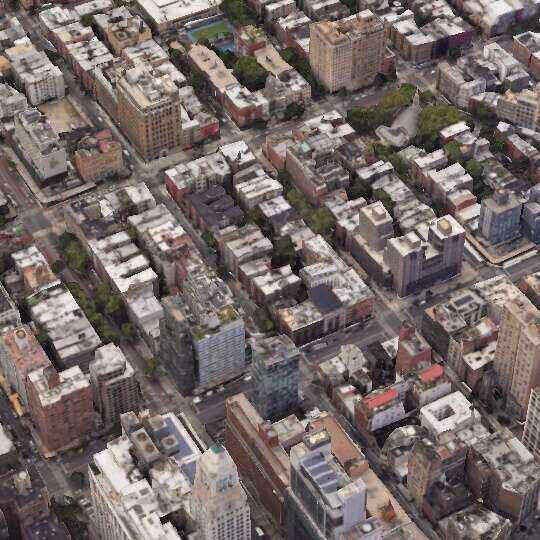} & 
    \imagecell[0.16]{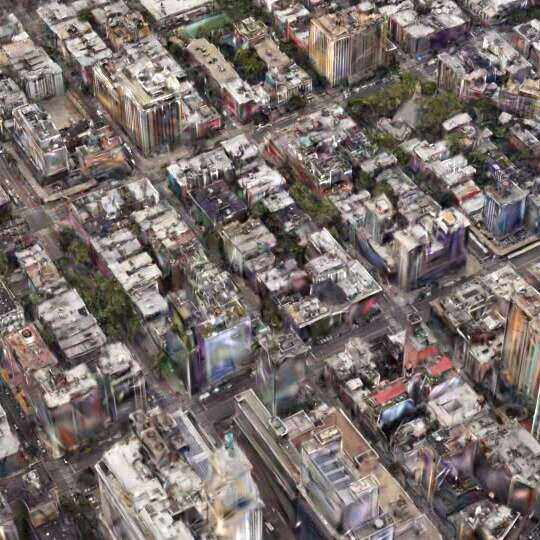} & 
    \imagecell[0.16]{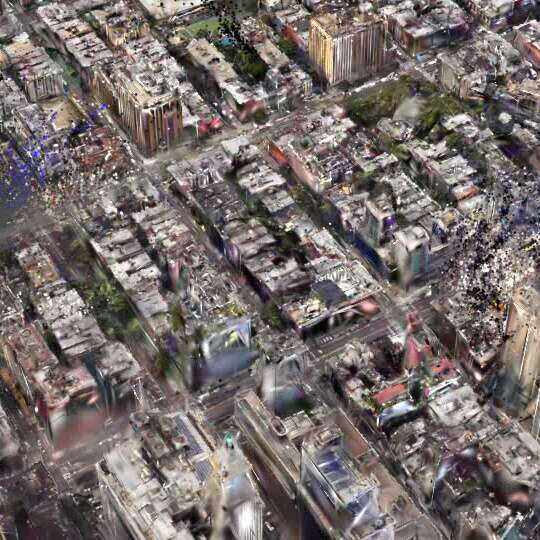} & 
    \imagecell[0.16]{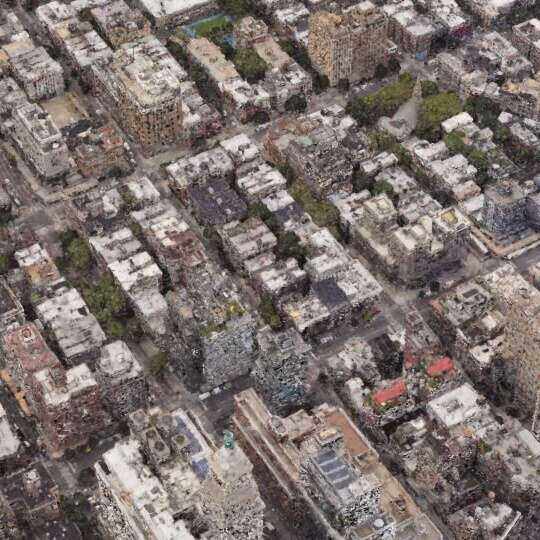} & 
    \imagecell[0.16]{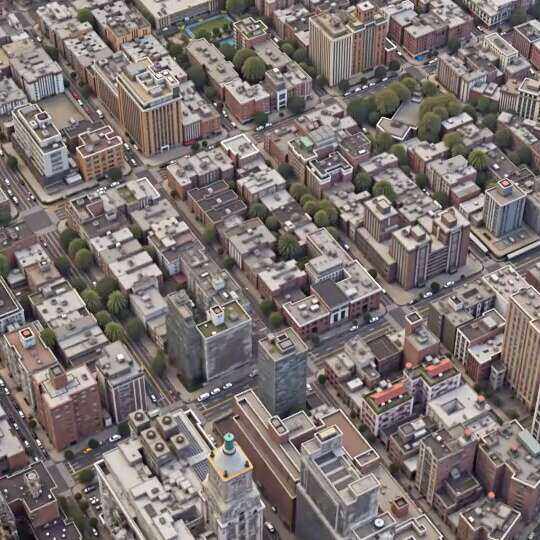} & 
    \imagecell[0.16]{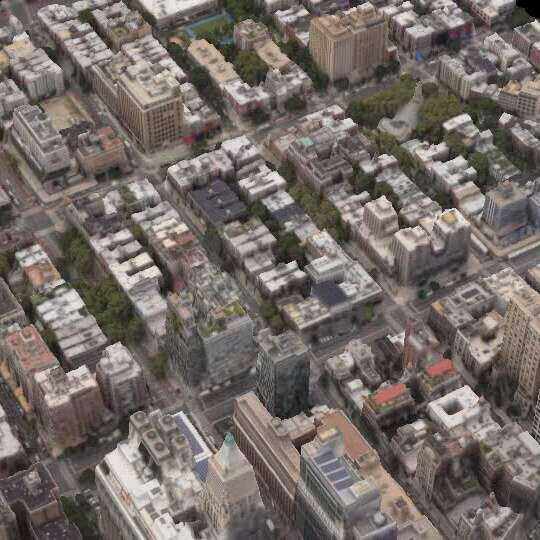} \\
    \vspace*{-10pt} \\
    
    \\
    \vspace*{-20pt}
    \\
    G.T. & 
    Mip-Splatting & 
    2DGS & 
    CityGS-X &
    Skyfall-GS & 
    Ours \\
    
    \end{tabular}
    \end{spacing}
	\caption{ 
    \textbf{Results of the NYC\_219 scene in the GoogleEarth dataset}. 
    Compared to baselines, our method successfully achieves high-quality city reconstruction from satellite imagery. 
    }
    \label{fig:supp:qual-nyc-219}
    \vspace*{-0.3cm}
\end{figure*}

\begin{figure*}[p]
	\centering
    \begin{spacing}{1} 
    \setlength\tabcolsep{1pt}
    \begin{tabular}{cccccc}

    \imagecell[0.16]{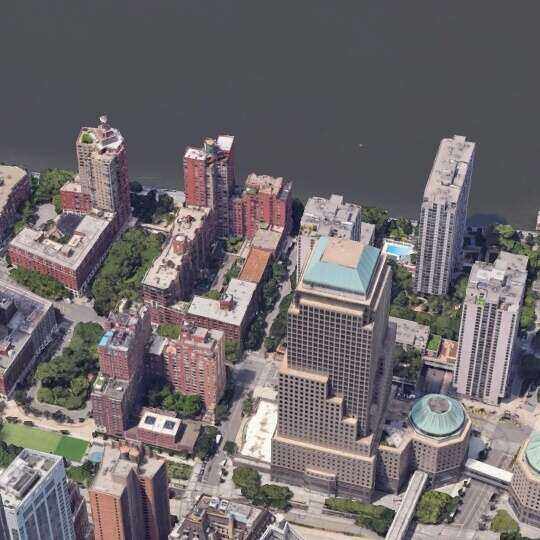} & 
    \imagecell[0.16]{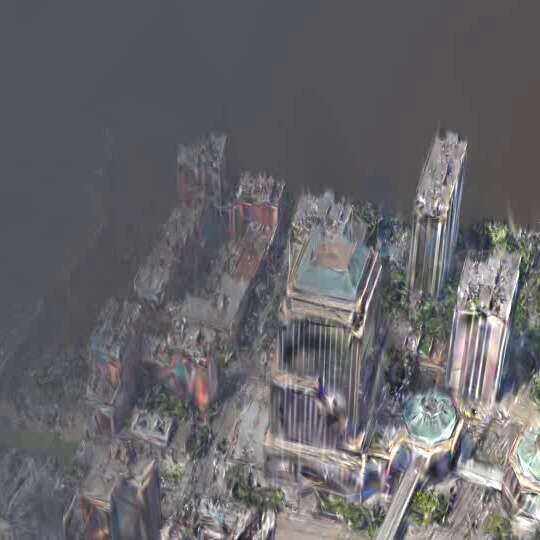} & 
    \imagecell[0.16]{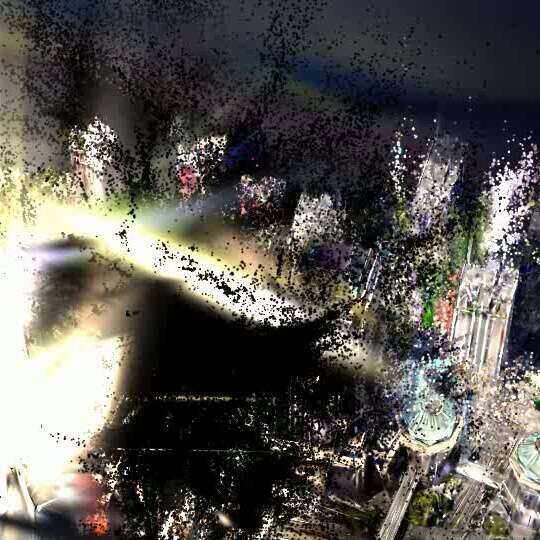} & 
    \imagecell[0.16]{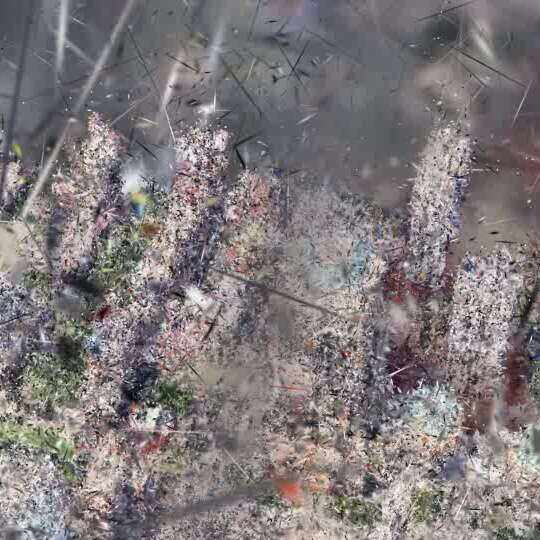} & 
    \imagecell[0.16]{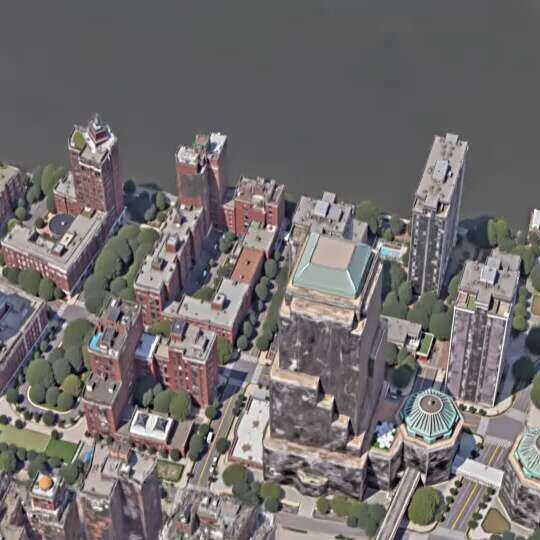} & 
    \imagecell[0.16]{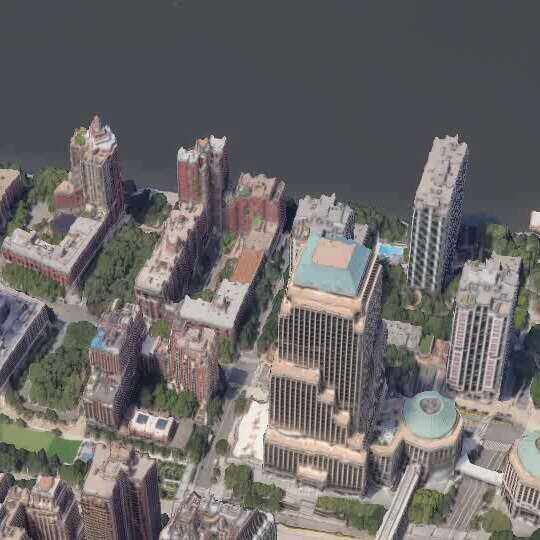} \\
    \vspace*{-10pt} \\

    \imagecell[0.16]{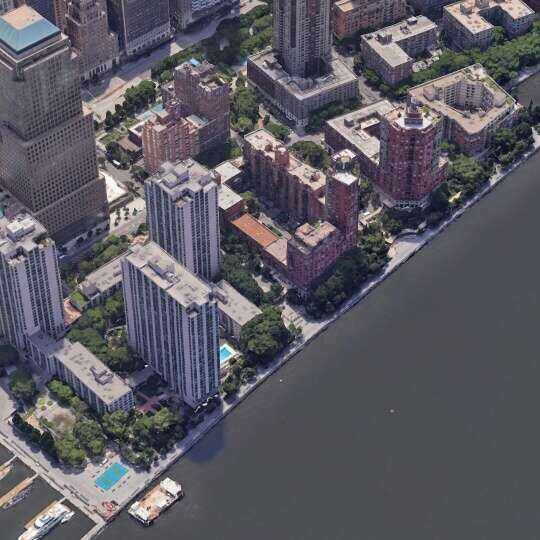} & 
    \imagecell[0.16]{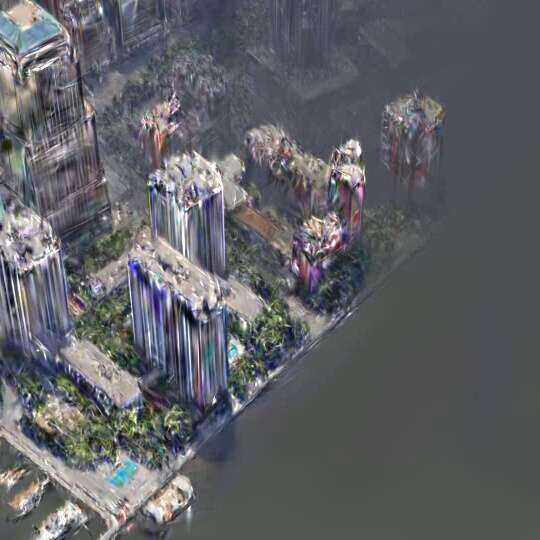} & 
    \imagecell[0.16]{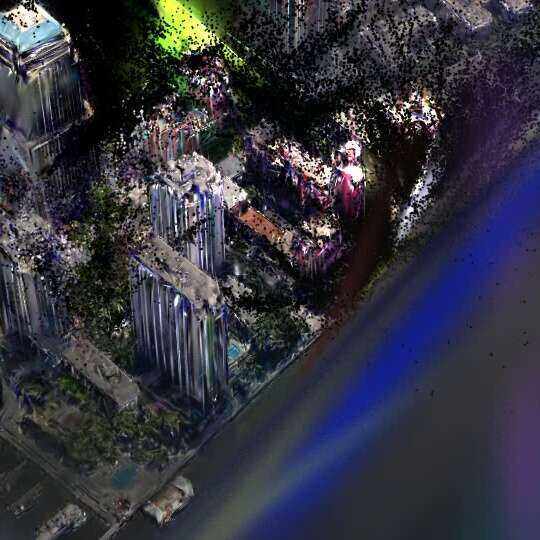} & 
    \imagecell[0.16]{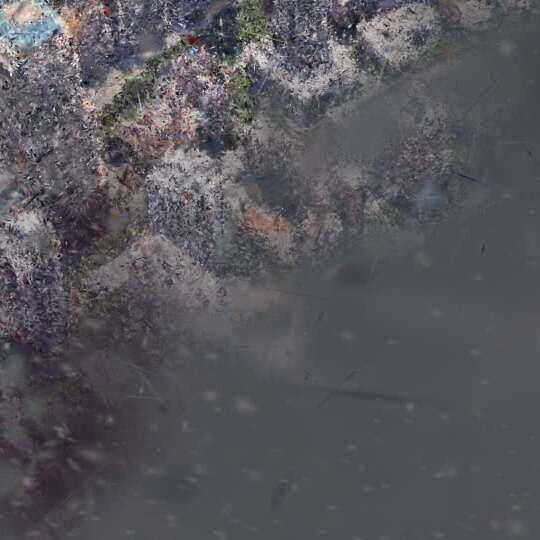} & 
    \imagecell[0.16]{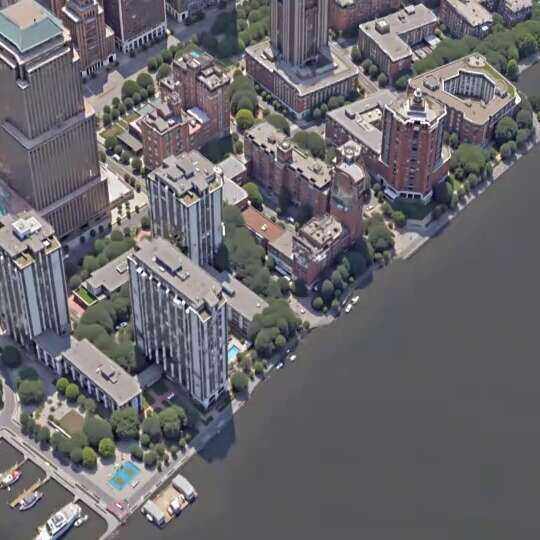} & 
    \imagecell[0.16]{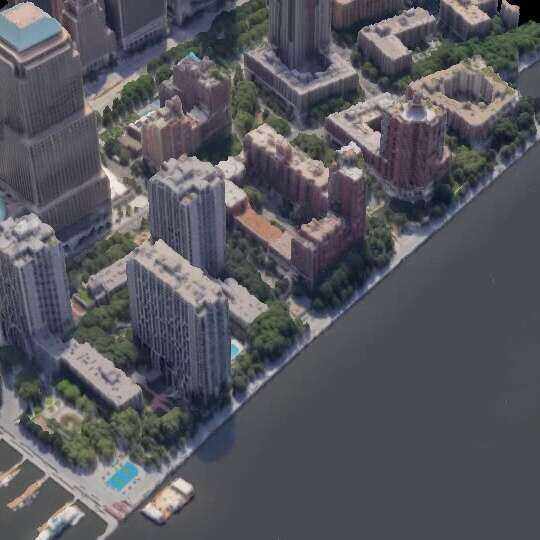} \\
    \vspace*{-10pt} \\

    \imagecell[0.16]{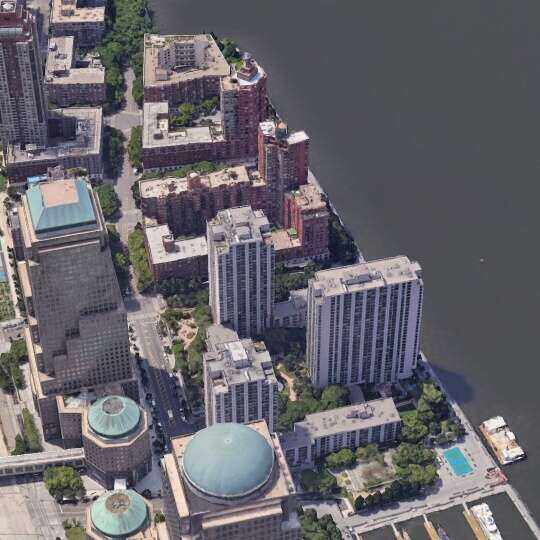} & 
    \imagecell[0.16]{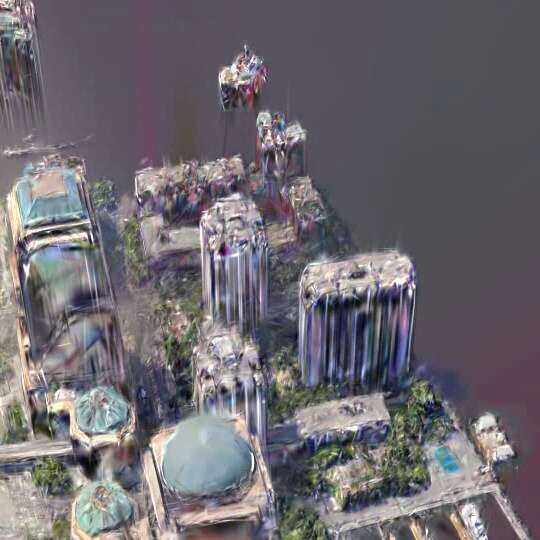} & 
    \imagecell[0.16]{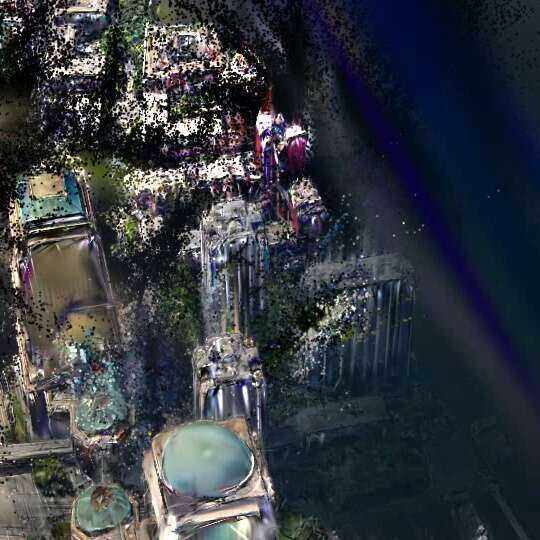} & 
    \imagecell[0.16]{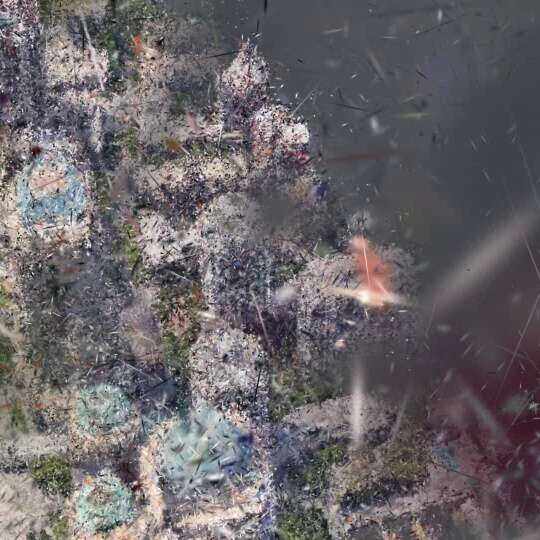} & 
    \imagecell[0.16]{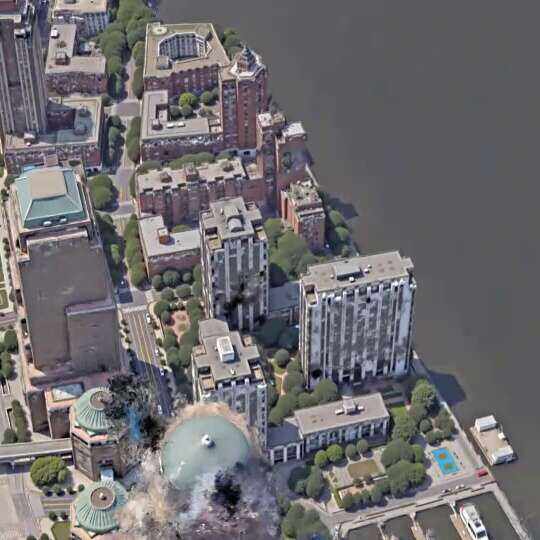} & 
    \imagecell[0.16]{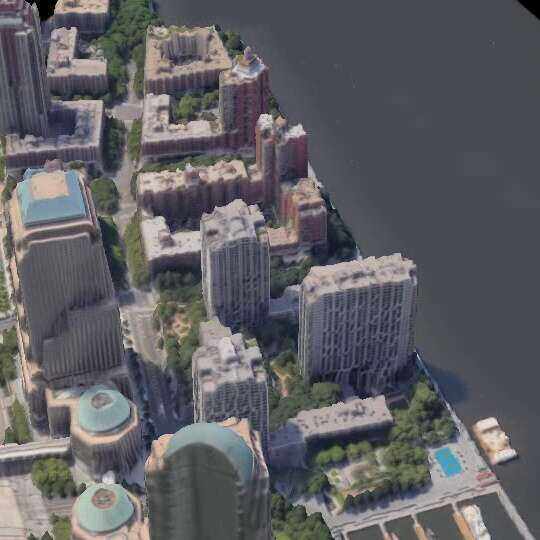} \\
    \vspace*{-10pt} \\
    
    \\
    \vspace*{-20pt}
    \\
    G.T. & 
    Mip-Splatting & 
    2DGS & 
    CityGS-X &
    Skyfall-GS & 
    Ours \\
    
    \end{tabular}
    \end{spacing}
	\caption{ 
    \textbf{Results of the NYC\_336 scene in the GoogleEarth dataset}. 
    Compared to baselines, our method successfully achieves high-quality city reconstruction from satellite imagery. 
    }
    \label{fig:supp:qual-nyc-336}
    \vspace*{-0.3cm}
\end{figure*}

\cref{fig:supp:qual-mc-far,fig:supp:qual-jax-004,fig:supp:qual-jax-068,fig:supp:qual-jax-214,fig:supp:qual-jax-260,fig:supp:qual-nyc-010,fig:supp:qual-nyc-219,fig:supp:qual-nyc-336} provide per-scene qualitative comparisons between our method, competing baselines, and the ground-truth test views.
Our reconstructions preserve fine-scale structural and texture details, such as building facades and roof patterns, which are often blurred, distorted, or missing in alternative methods.

\subsection{Qualitative Analysis at Varying Camera Altitudes}

To further investigate how our reconstructed city models behave under varying observation altitudes, we conduct a controlled rendering study with fixed camera intrinsics and identical view geometry. 
Specifically, we fix the camera's horizontal position, look-at point, and intrinsics ($f_x, f_y=1500$), while systematically varying the camera height at $[50, 200, 400, 600, 800]$ meters. 

\begin{figure*}[p]
	\centering
    \begin{spacing}{1} 
    \setlength\tabcolsep{1pt}
    \begin{tabular}{cccc}

    {\rotatebox[origin=c]{90}{\textbf{Skyfall-GS}} \hspace{1pt}} &
    \imagecell[0.27]{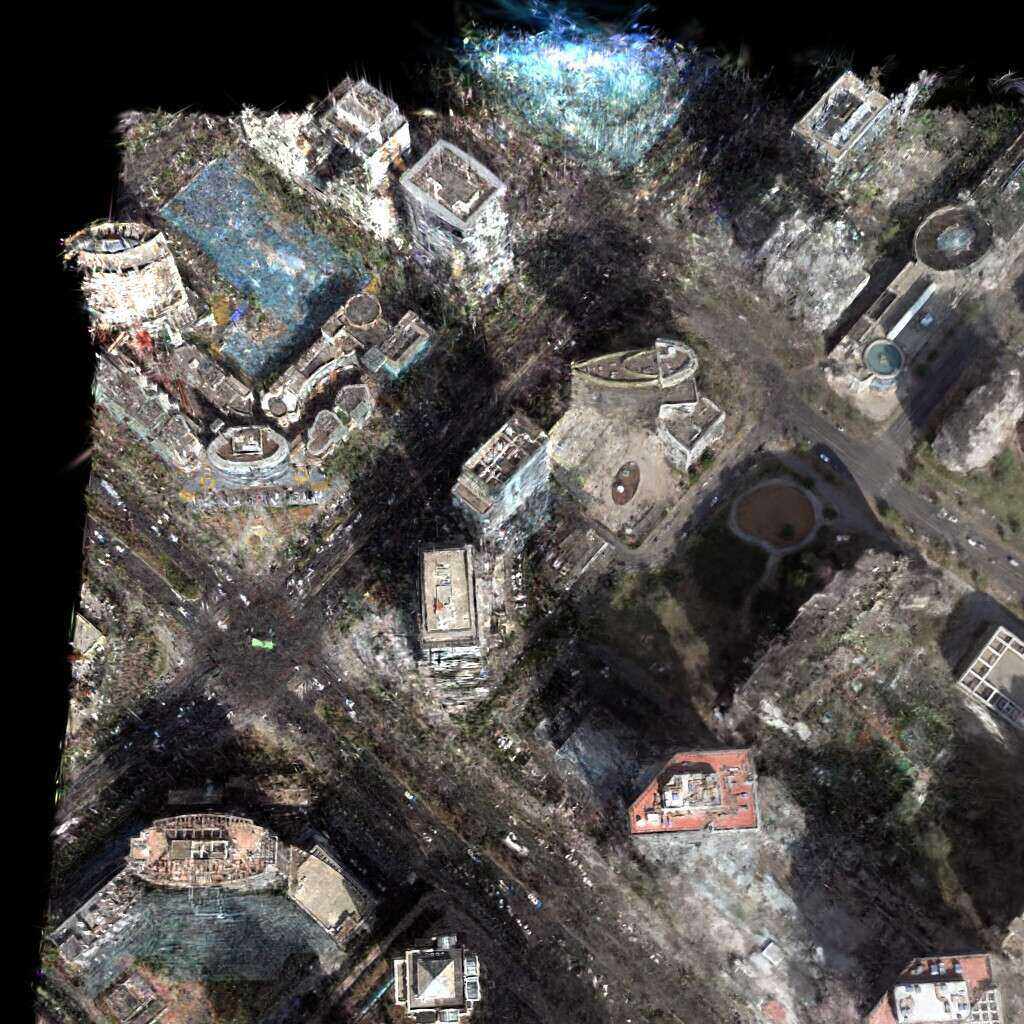} & 
    \imagecell[0.27]{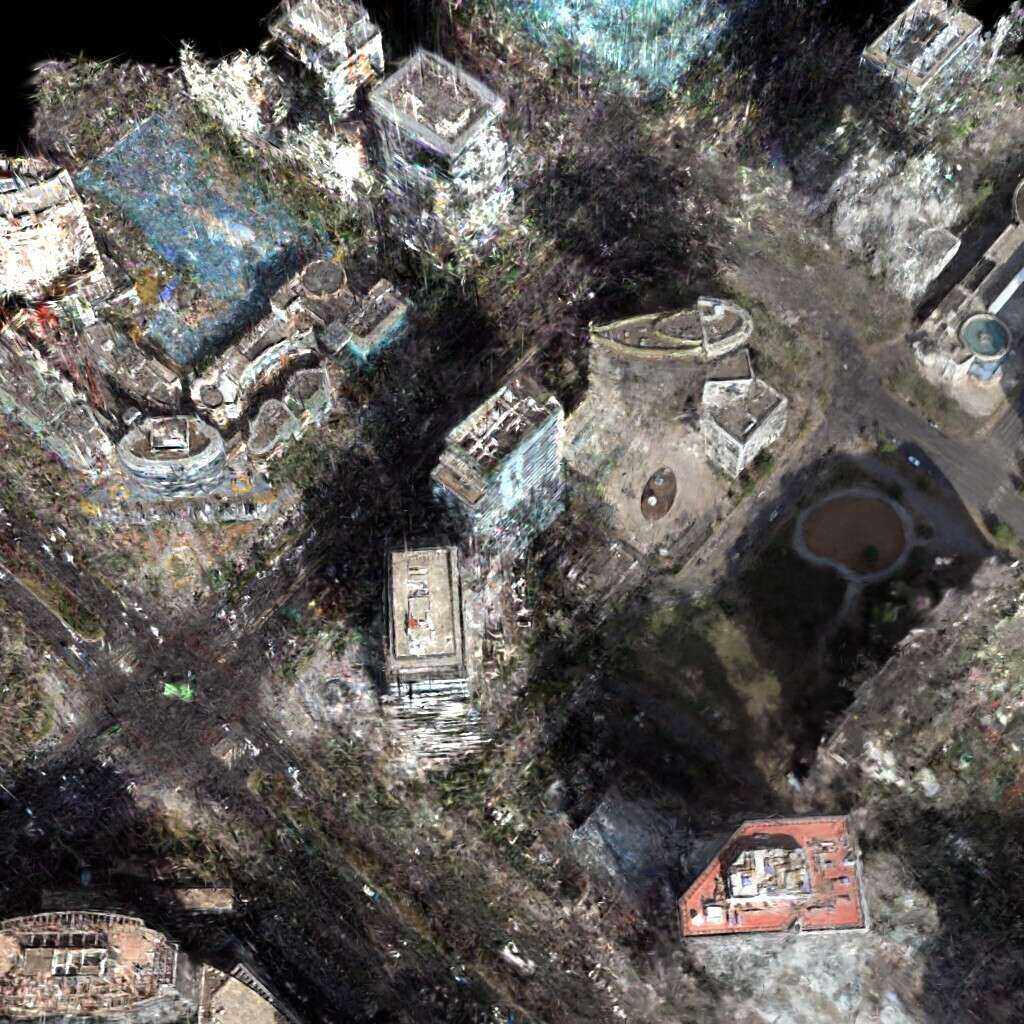} &
    \imagecell[0.27]{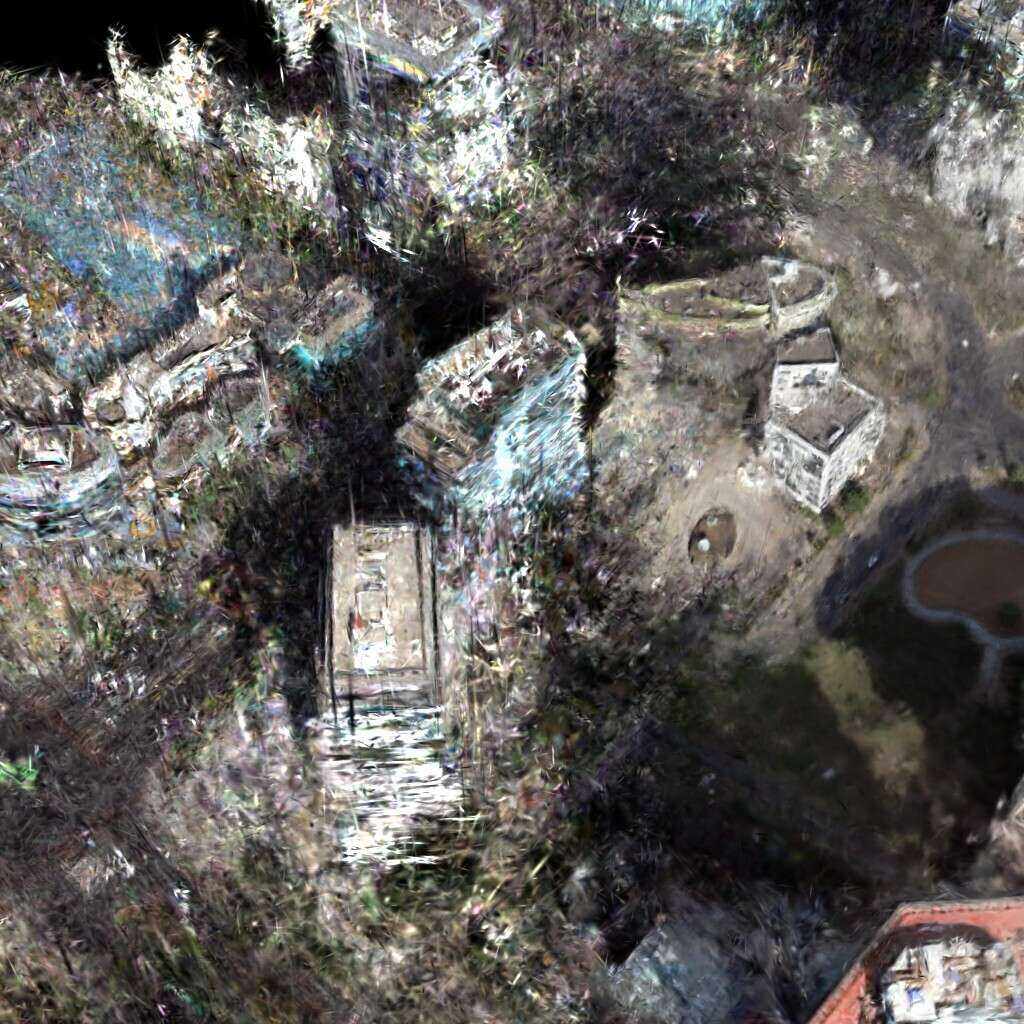} \\ 
    \vspace*{-10pt} \\

    {\rotatebox[origin=c]{90}{\textbf{Ours}} \hspace{1pt}} &
    \imagecell[0.27]{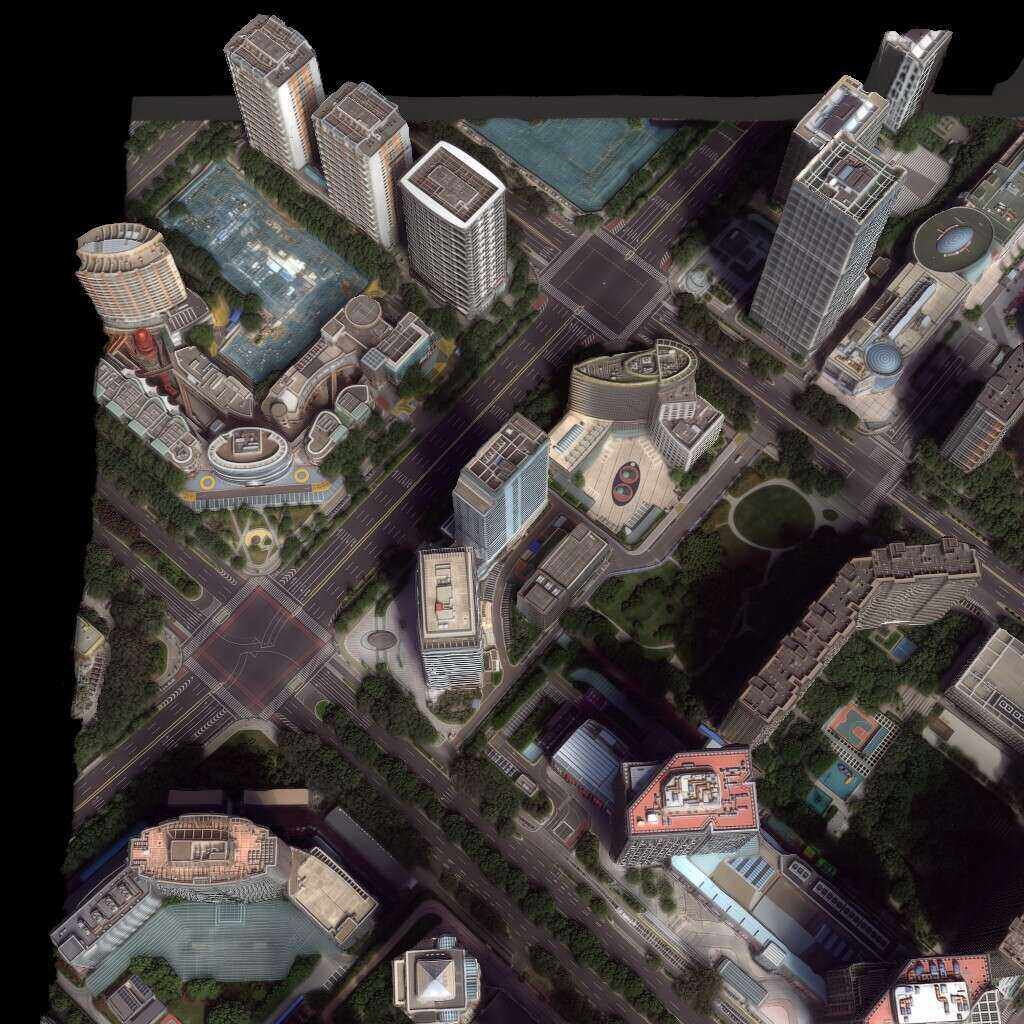} & 
    \imagecell[0.27]{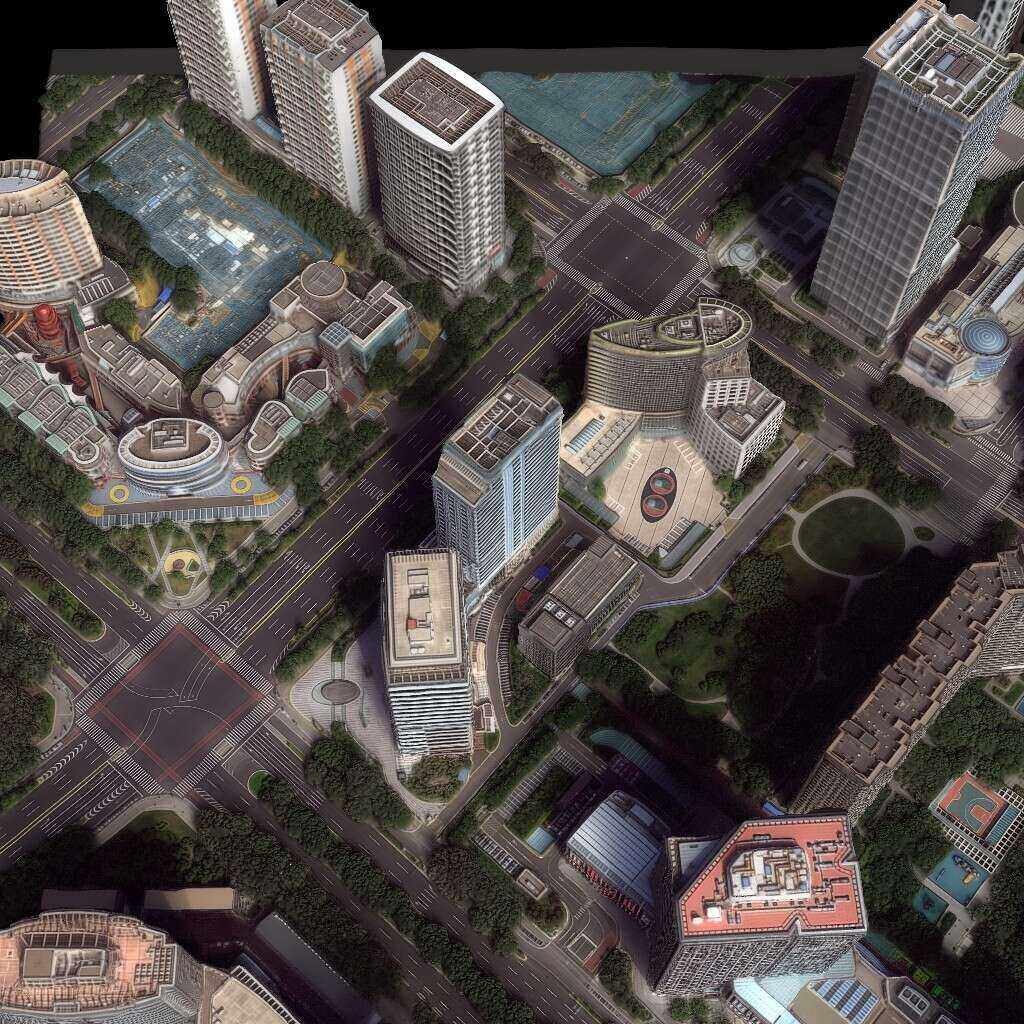} & 
    \imagecell[0.27]{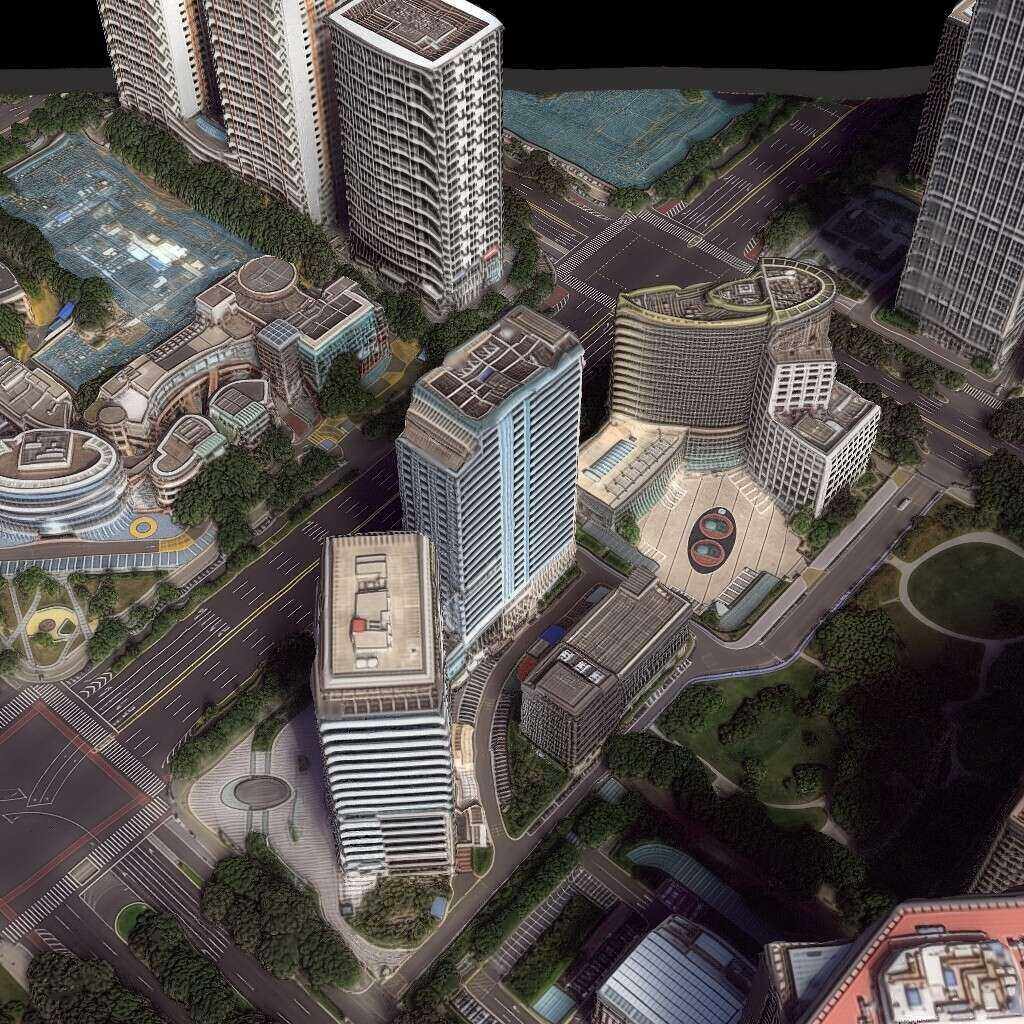} \\

    \\
    \vspace*{-20pt}
    \\
    &
    800m & 
    600m & 
    400m \\

    {\rotatebox[origin=c]{90}{\textbf{Skyfall-GS}} \hspace{1pt}} &
    \imagecell[0.27]{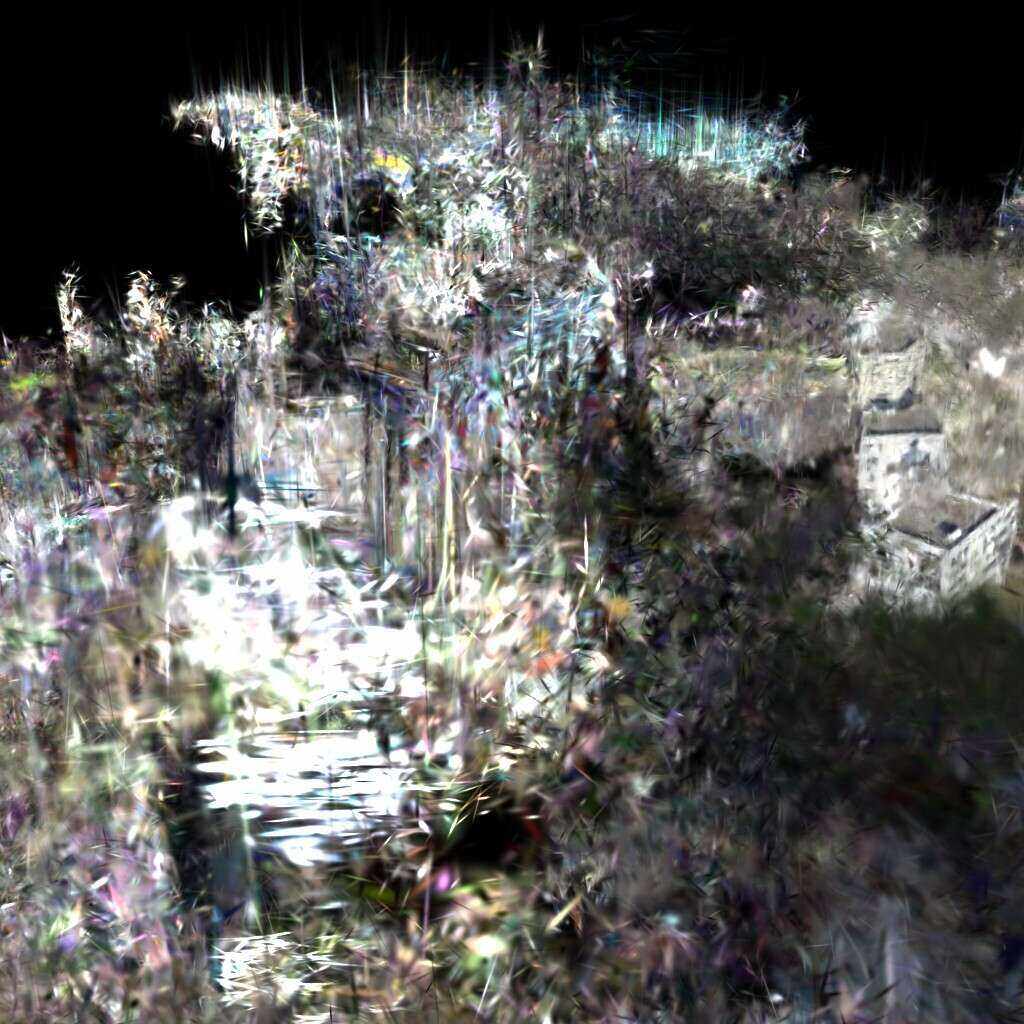} & 
    \imagecell[0.27]{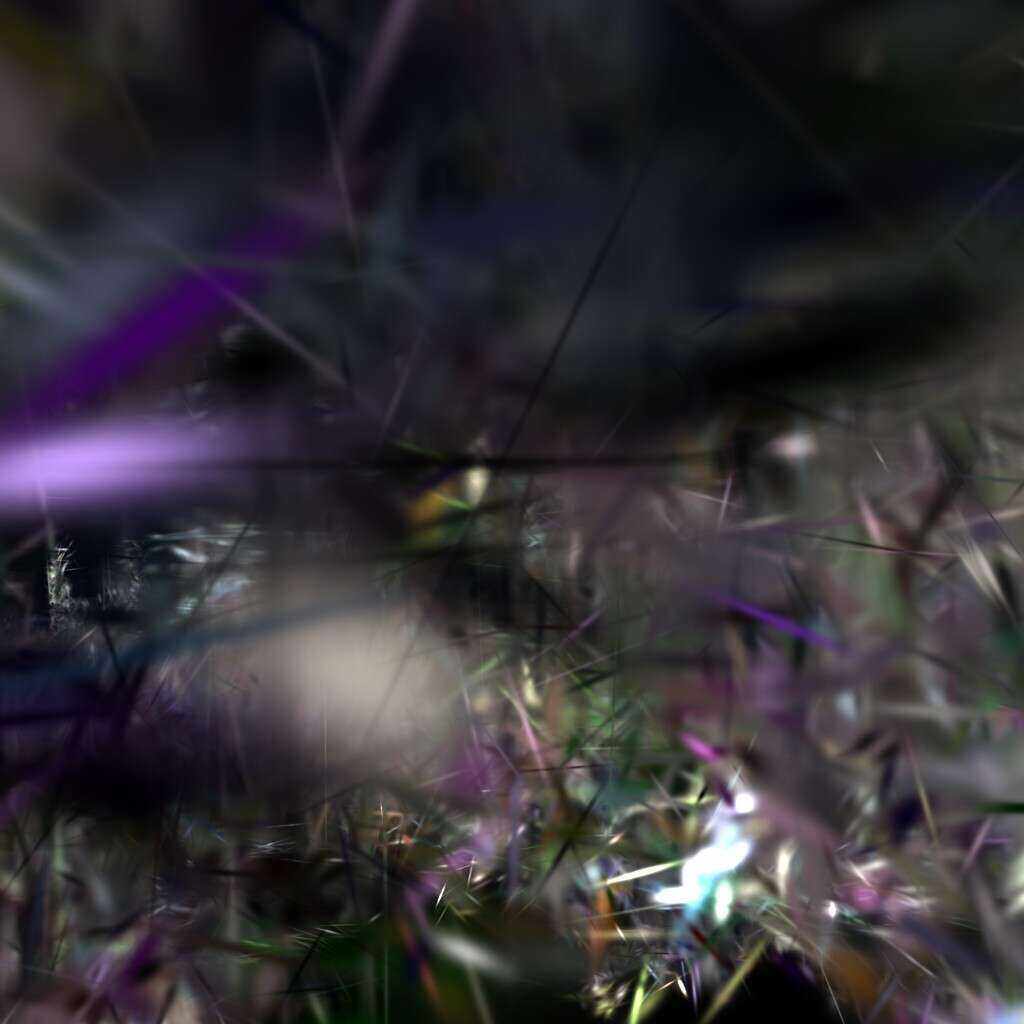} & 
    \imagecell[0.27]{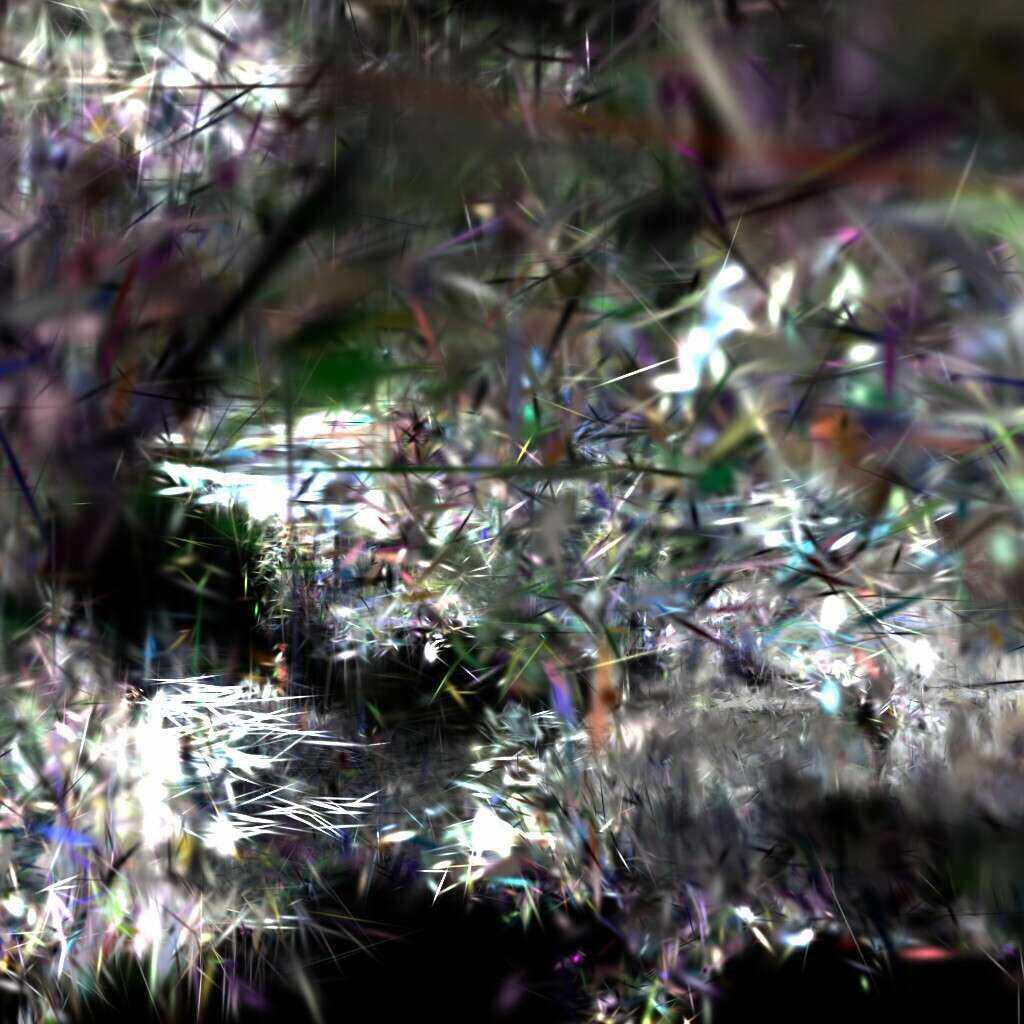} \\
    \vspace*{-10pt} \\

    {\rotatebox[origin=c]{90}{\textbf{Ours}} \hspace{1pt}} &
    \imagecell[0.27]{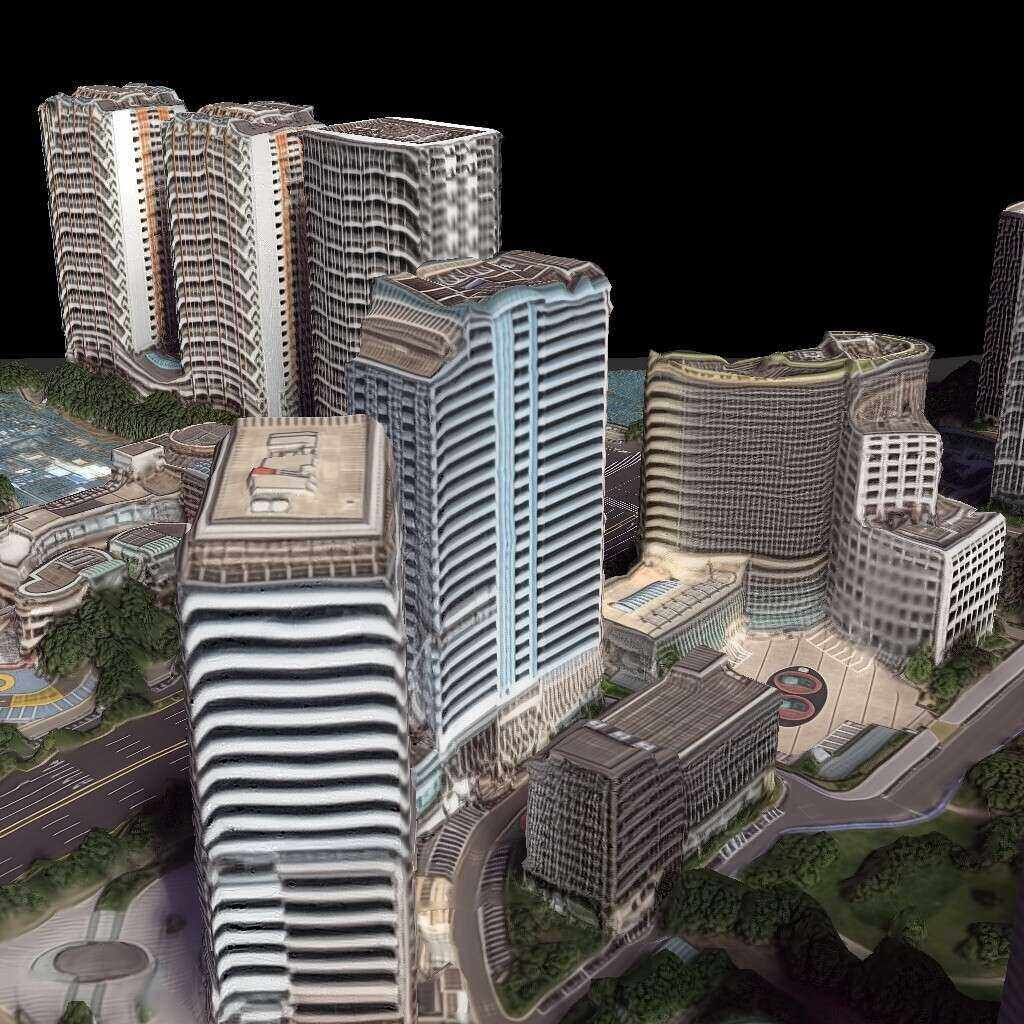} & 
    \imagecell[0.27]{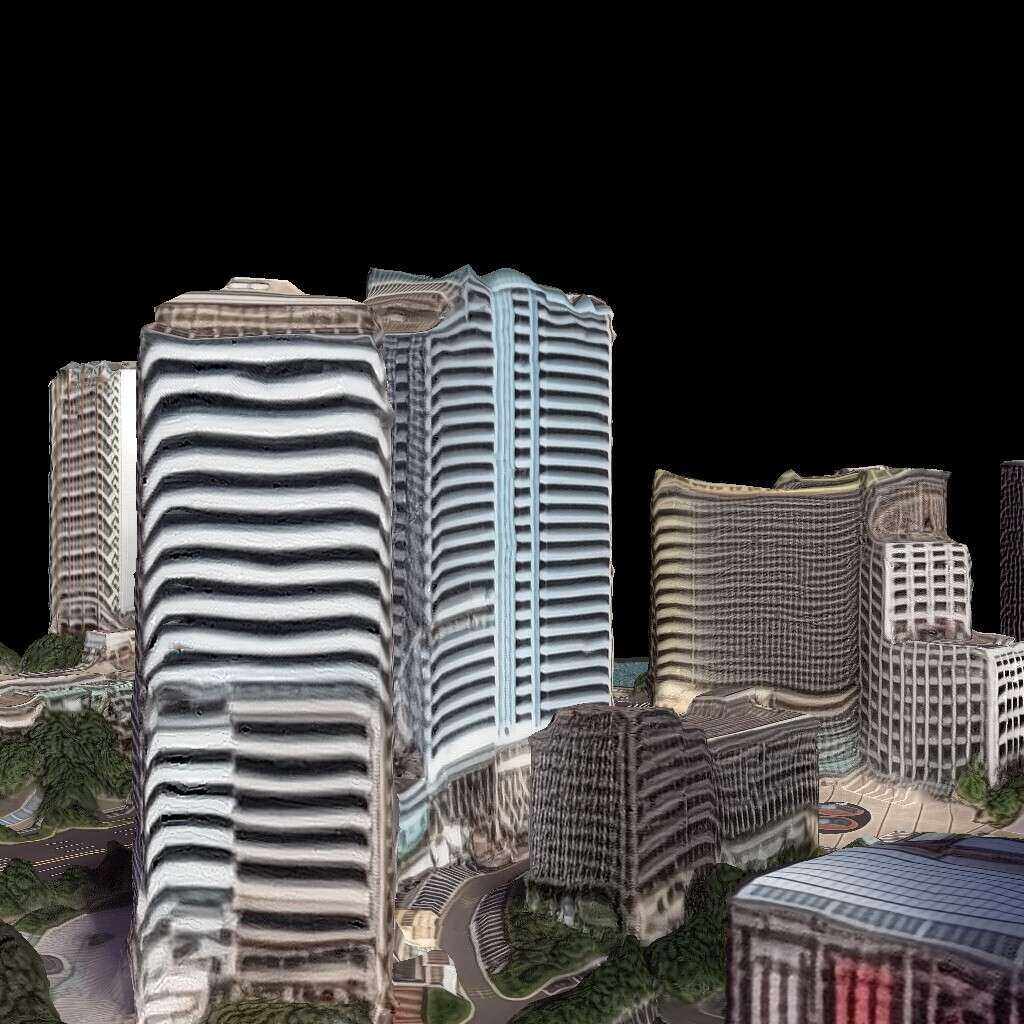} & 
    \imagecell[0.27]{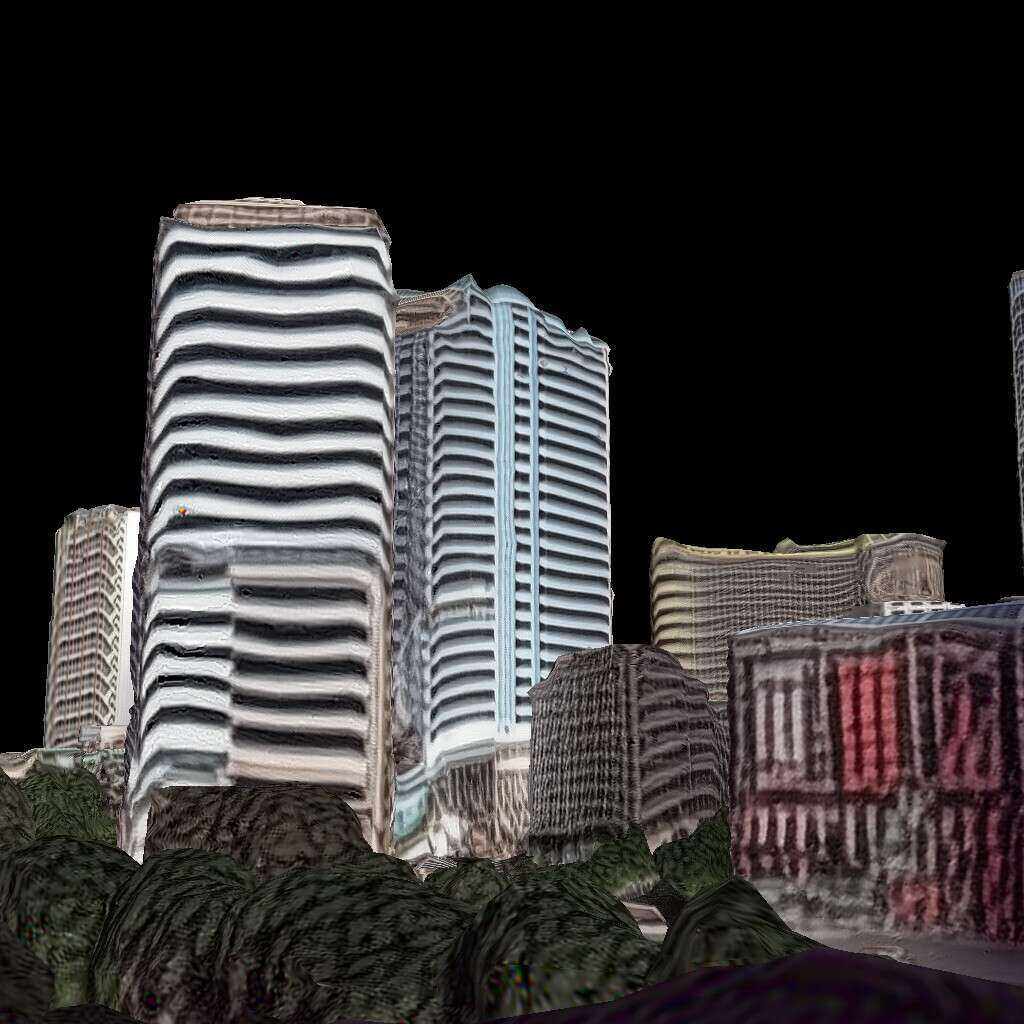} \\
    \vspace*{-10pt} \\
    
    \\
    \vspace*{-20pt}
    \\
    &
    200m &
    100m & 
    50m \\
    
    \end{tabular}
    \end{spacing}
	\caption{ 
    \textbf{Qualitative comparison between our method and Skyfall-GS across different camera altitudes}. 
    While our reconstructions may exhibit minor seam artifacts, Skyfall-GS even suffers from substantial quality degradation at lower altitudes.
    }
    \label{fig:supp:qual-alt}
    \vspace*{-0.3cm}
\end{figure*}

\cref{fig:supp:qual-alt} presents a qualitative comparison between our method and Skyfall-GS~\cite{lee2025skyfall} in real-world scenes. Our model exhibits stronger robustness, consistently delivering superior results across different observation altitudes. As the camera height decreases, the performance gap becomes increasingly evident: our method preserves markedly sharper facade textures and more refined geometric details. While our reconstructions may exhibit minor seams due to the joint optimization over multiple satellite views, Skyfall-GS suffers from pronounced quality degradation at lower altitudes, including widespread blurring and noticeable ``floating object'' artifacts.

\subsection{Close-Up Views}

The extreme viewpoint gap between orbital inputs and low-altitude target views poses a particular challenge for satellite-based reconstruction.
To further evaluate our method under such challenging conditions, we provide additional close-up comparisons that focus on fine-scale geometric and texture details.

\begin{figure*}[p]
	\centering
    \begin{spacing}{1} 
    \setlength\tabcolsep{1pt}
    \begin{tabular}{ccc}

    \imagecell[0.29]{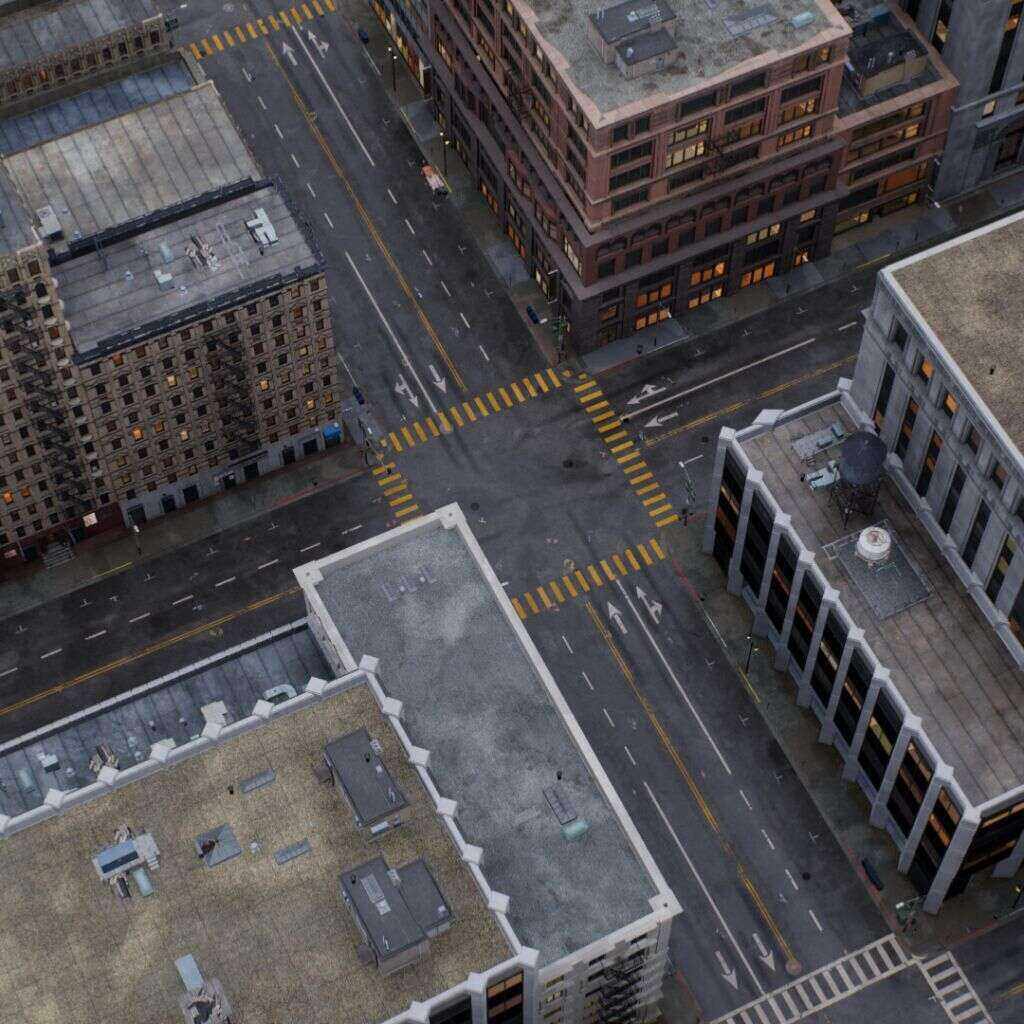} & 
    \imagecell[0.29]{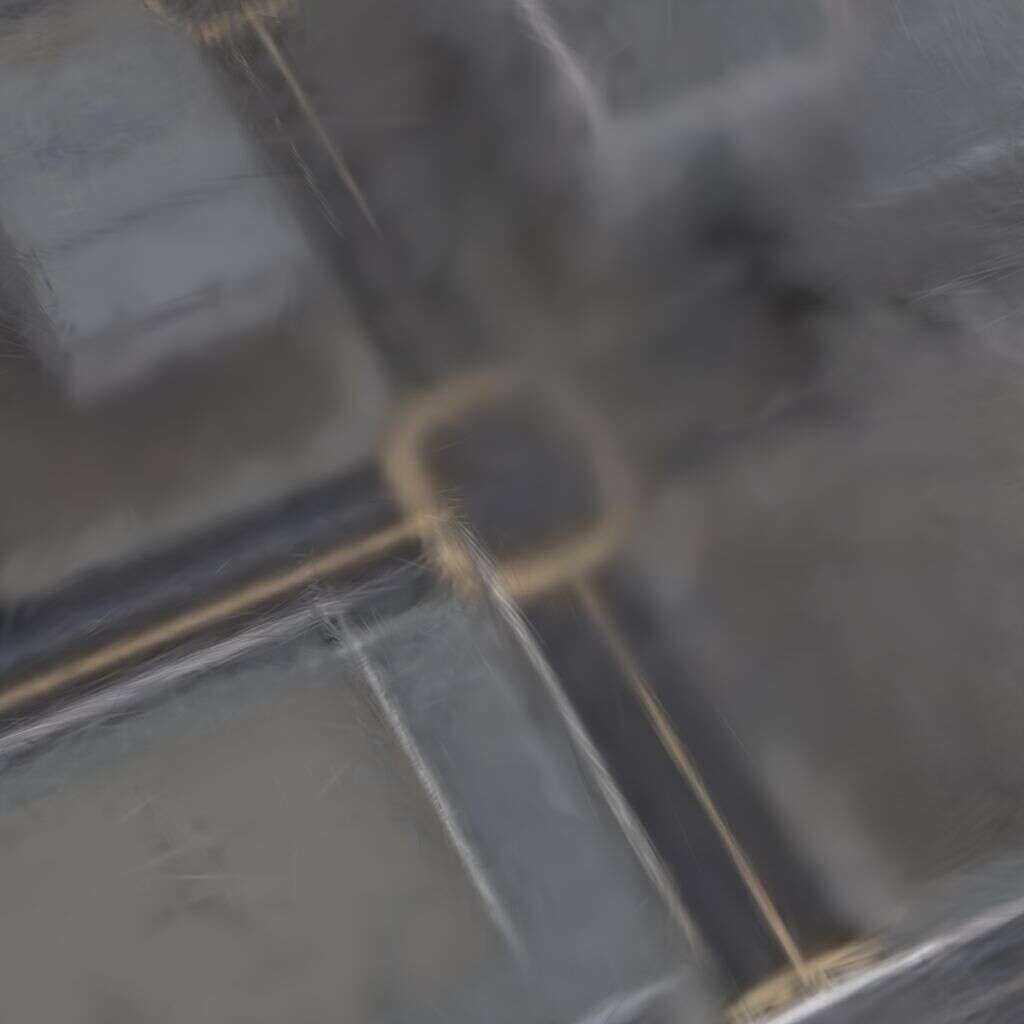} & 
    \imagecell[0.29]{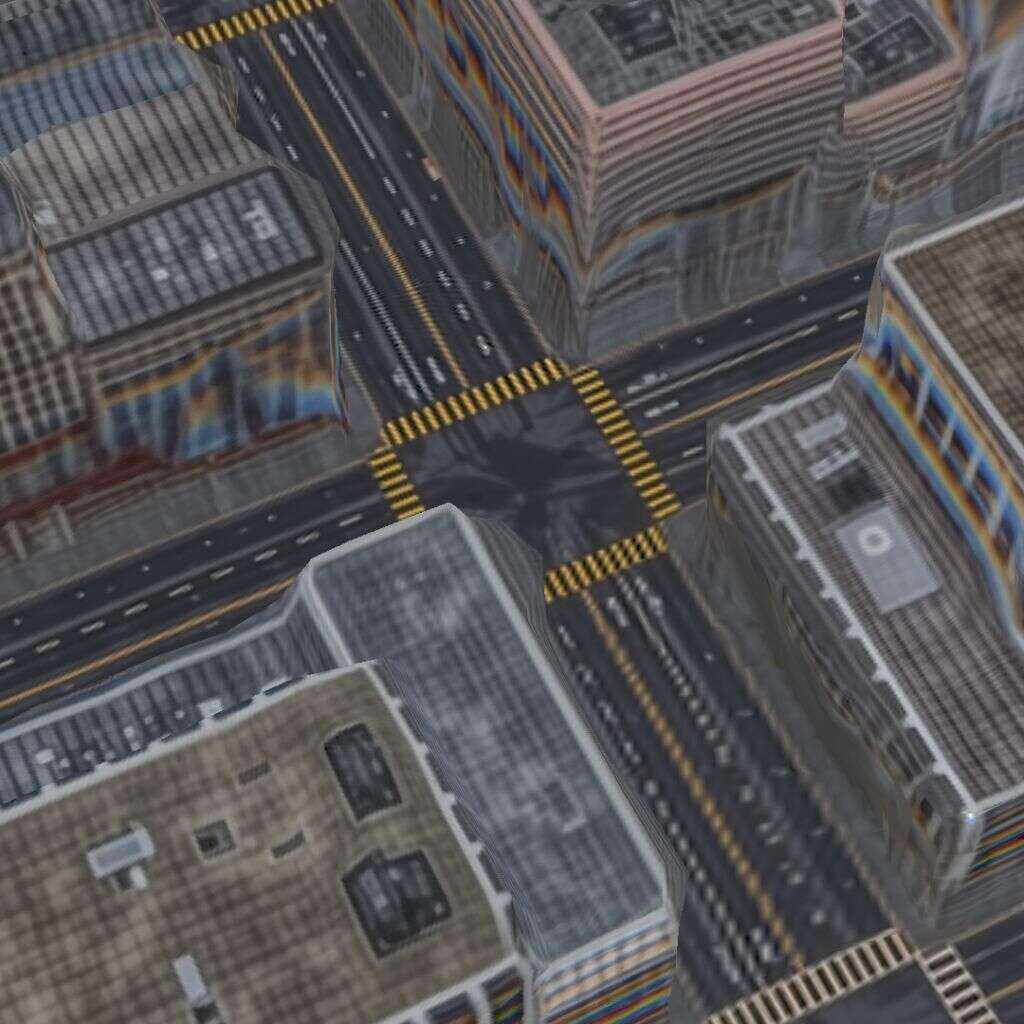} \\
    \vspace*{-10pt} \\

    \imagecell[0.29]{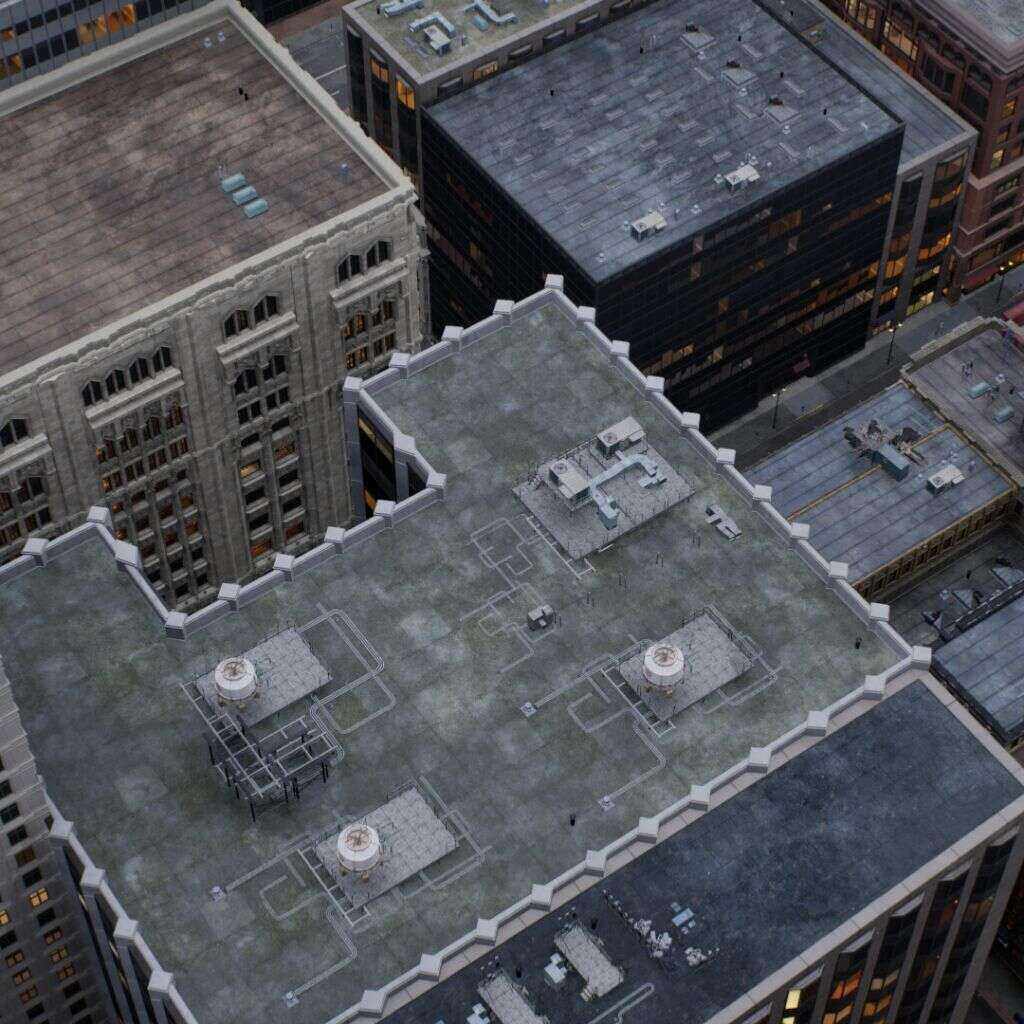} & 
    \imagecell[0.29]{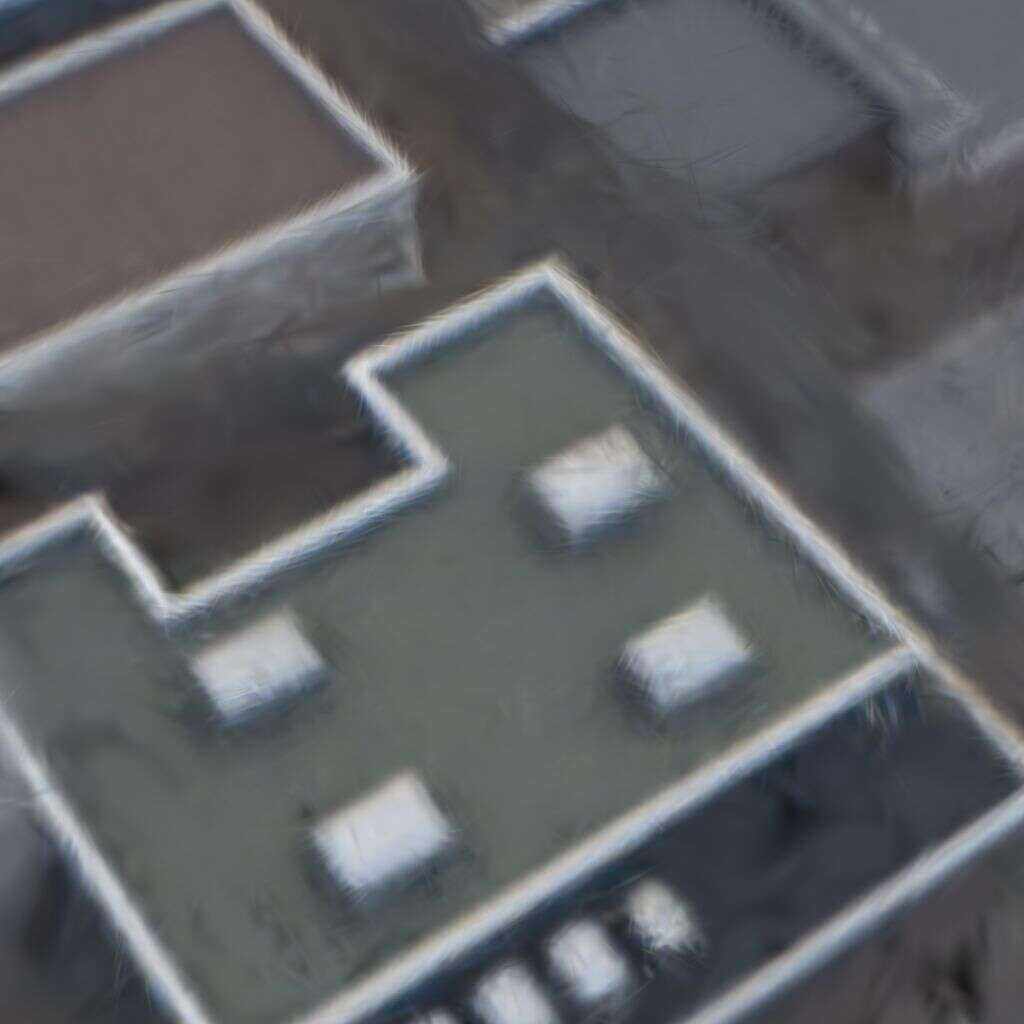} & 
    \imagecell[0.29]{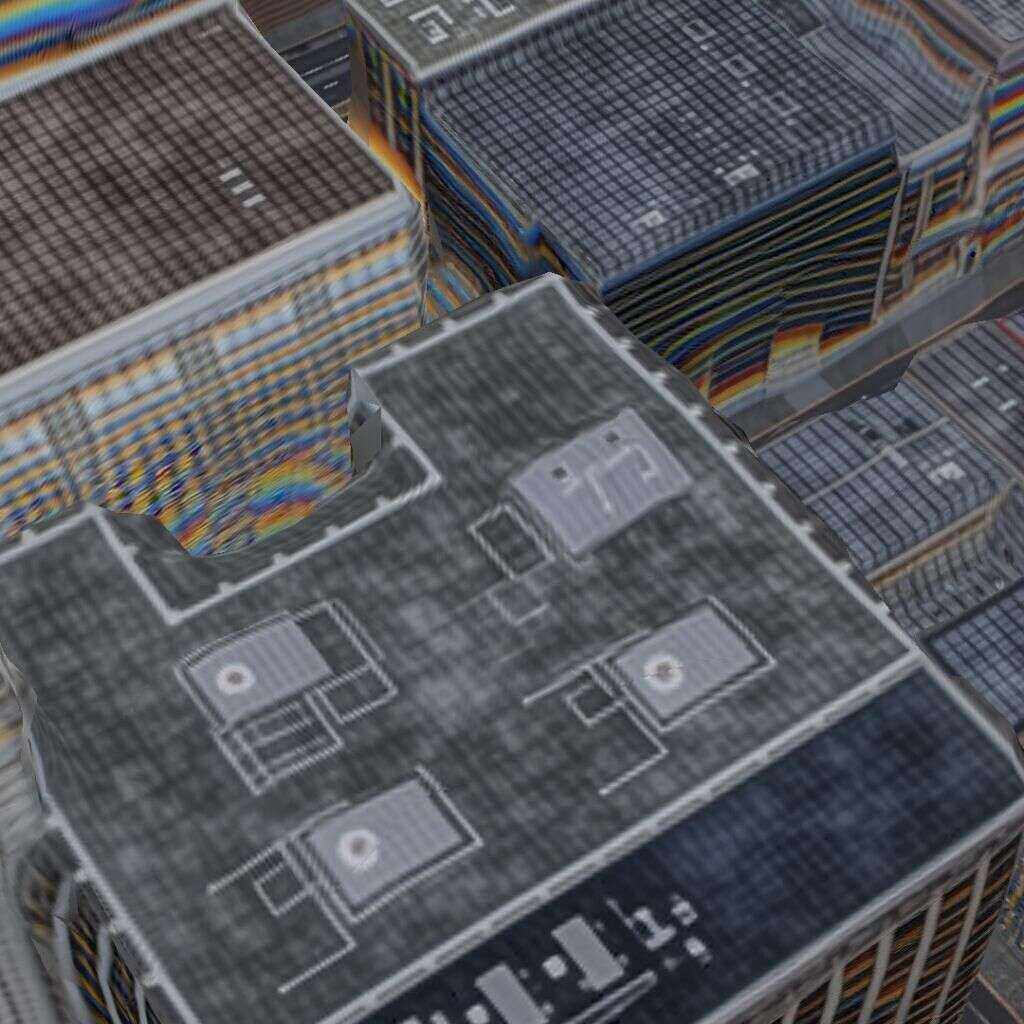} \\
    \vspace*{-10pt} \\

    \imagecell[0.29]{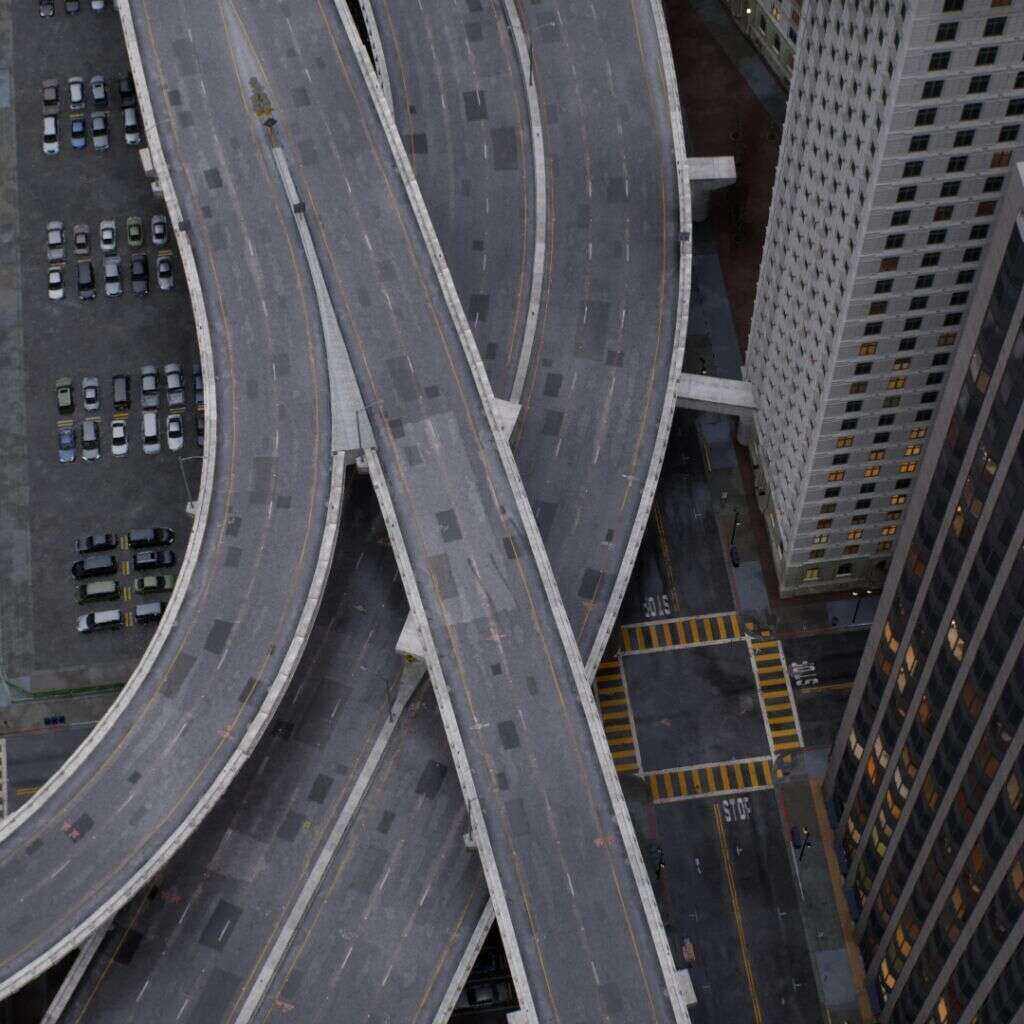} & 
    \imagecell[0.29]{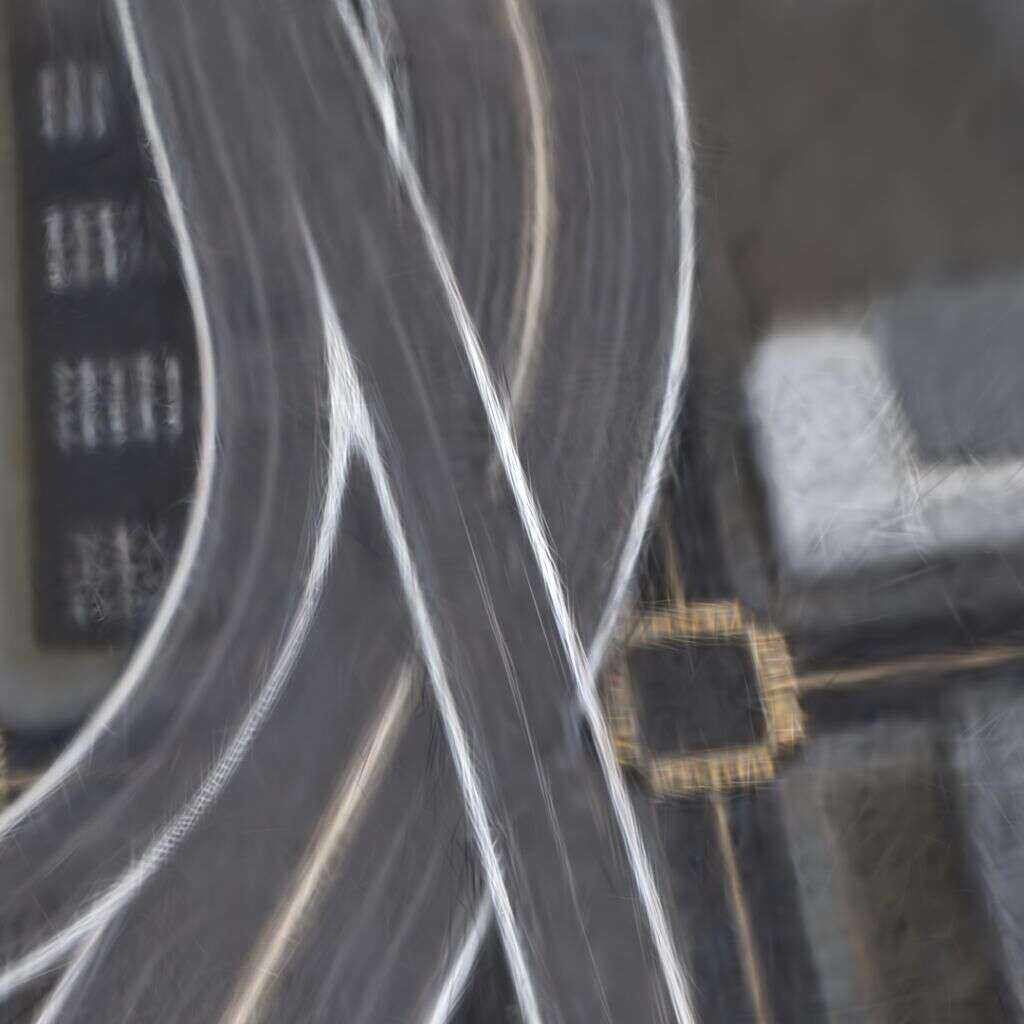} & 
    \imagecell[0.29]{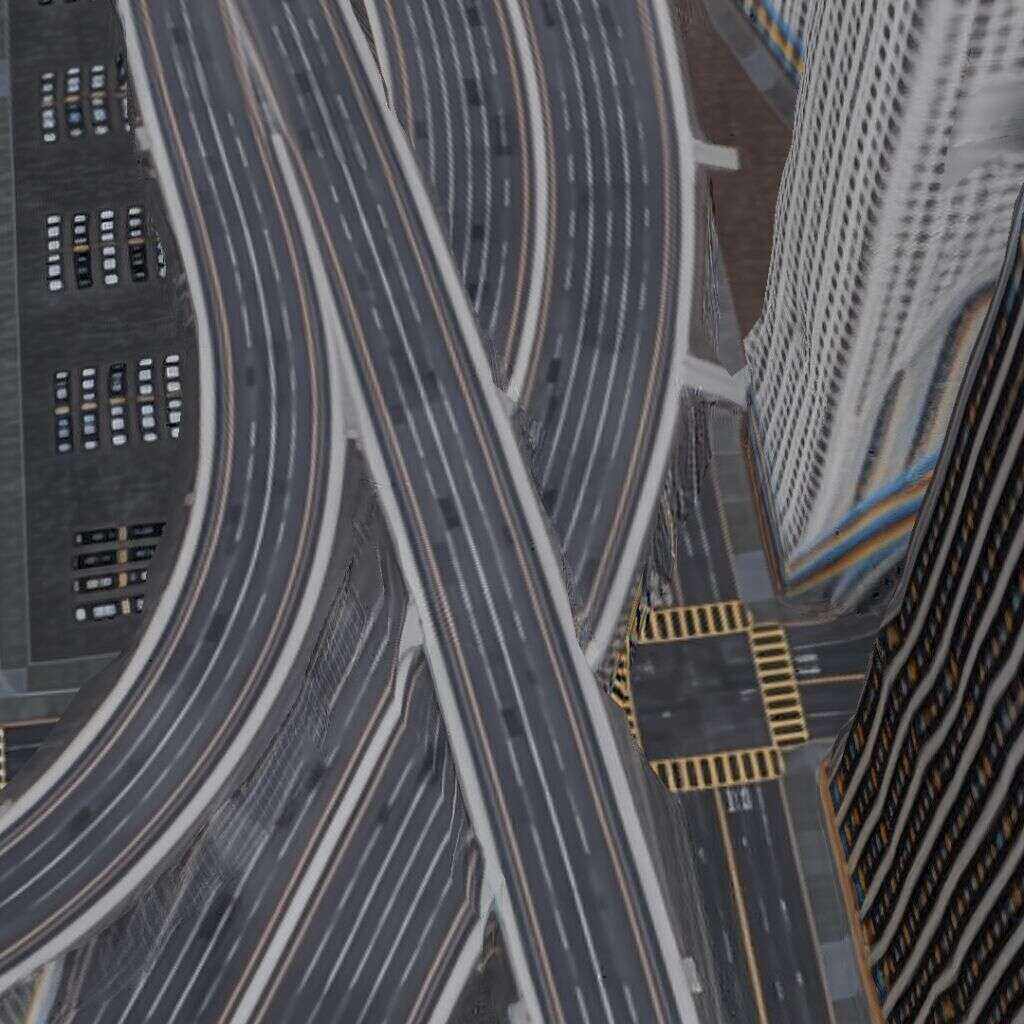} \\
    \vspace*{-10pt} \\

    \imagecell[0.29]{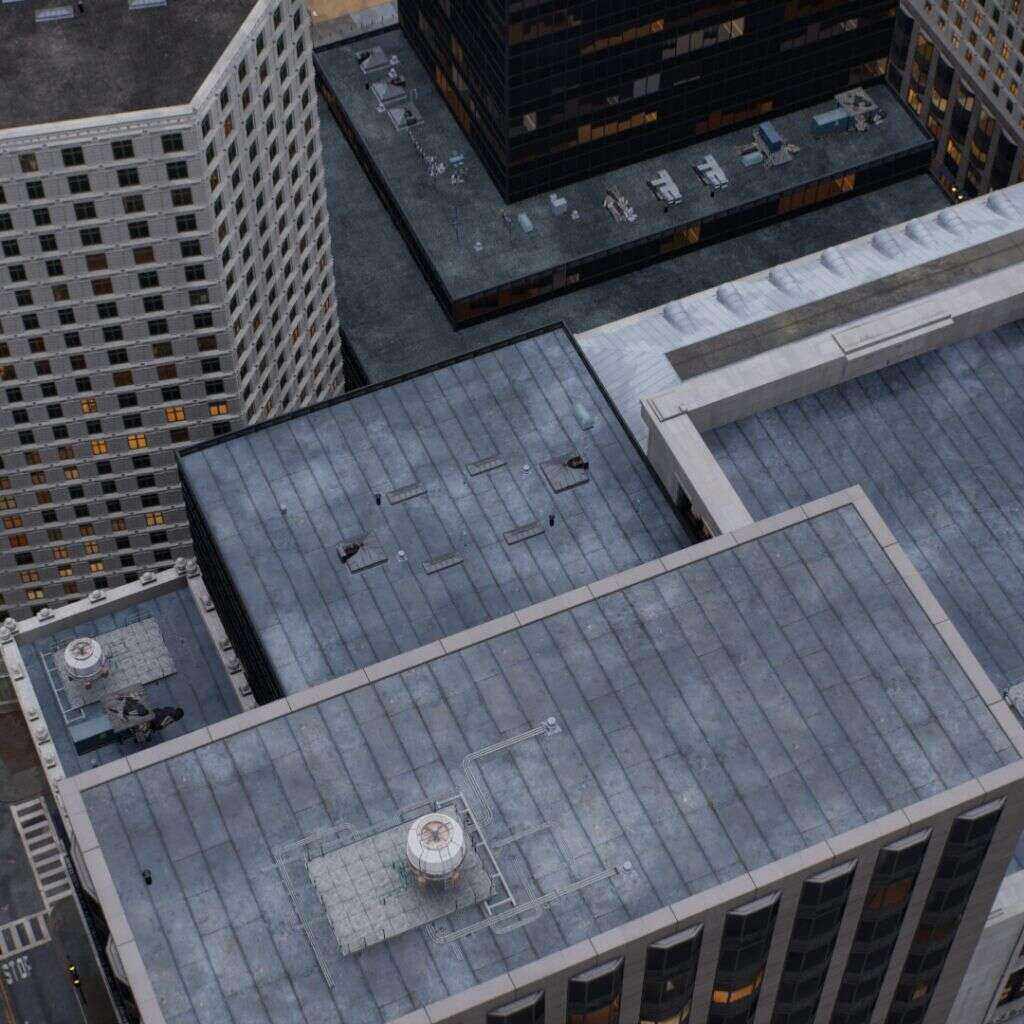} & 
    \imagecell[0.29]{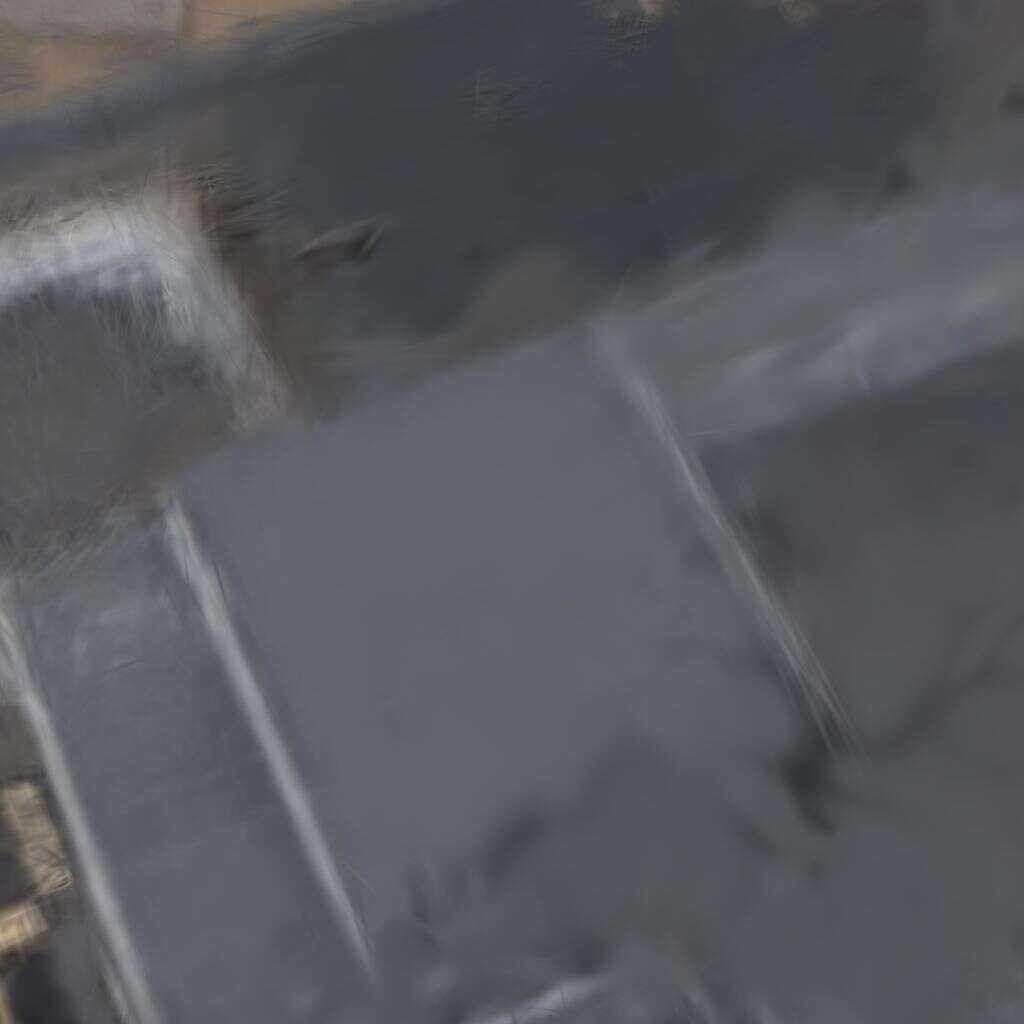} & 
    \imagecell[0.29]{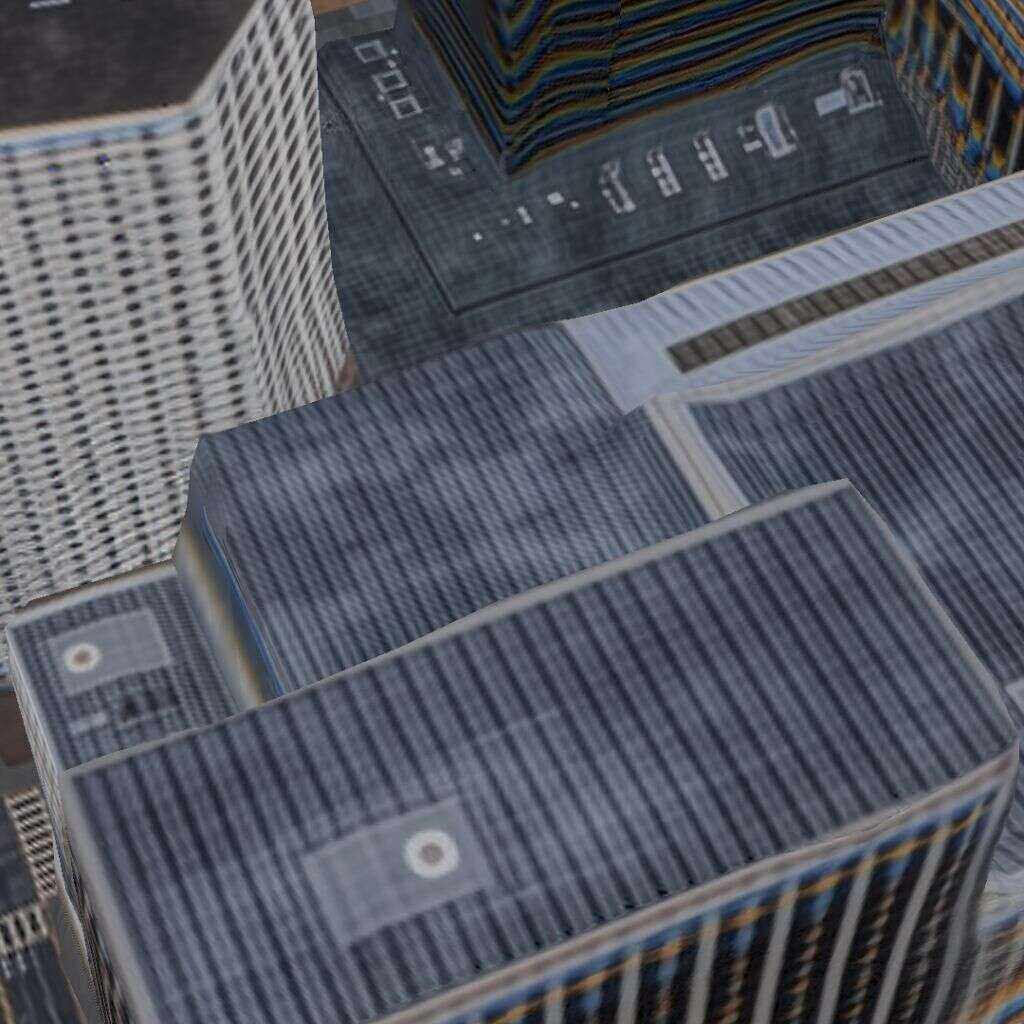} \\
    \vspace*{-10pt} \\
    
    \\
    \vspace*{-20pt}
    \\
    G.T. & 
    Skyfall-GS & 
    Ours \\
    
    \end{tabular}
    \end{spacing}
	\caption{ 
    \textbf{Close-Up views of reconstruction results of the MatrixCity-Satellite scene}. 
    Our method retains small-scale details such as window grids, facade lines, and roof textures significantly better than Skyfall-GS \cite{lee2025skyfall}, evidencing the benefit of our deterministic diffusion-based texture refinement.
    }
    \label{fig:supp:qual-mc-close}
    \vspace*{-0.3cm}
\end{figure*}

\cref{fig:supp:qual-mc-close} shows close-up novel views on the MatrixCity-Satellite benchmark, comparing our results against Skyfall-GS~\cite{lee2025skyfall} and the ground truth.
Our method produces sharper facades and more detailed roof textures, with significantly fewer ghosting and aliasing artifacts.

\begin{figure*}[p]
	\centering
    \begin{spacing}{1} 
    \setlength\tabcolsep{1pt}
    \begin{tabular}{cccc}

    \imagecell[0.24]{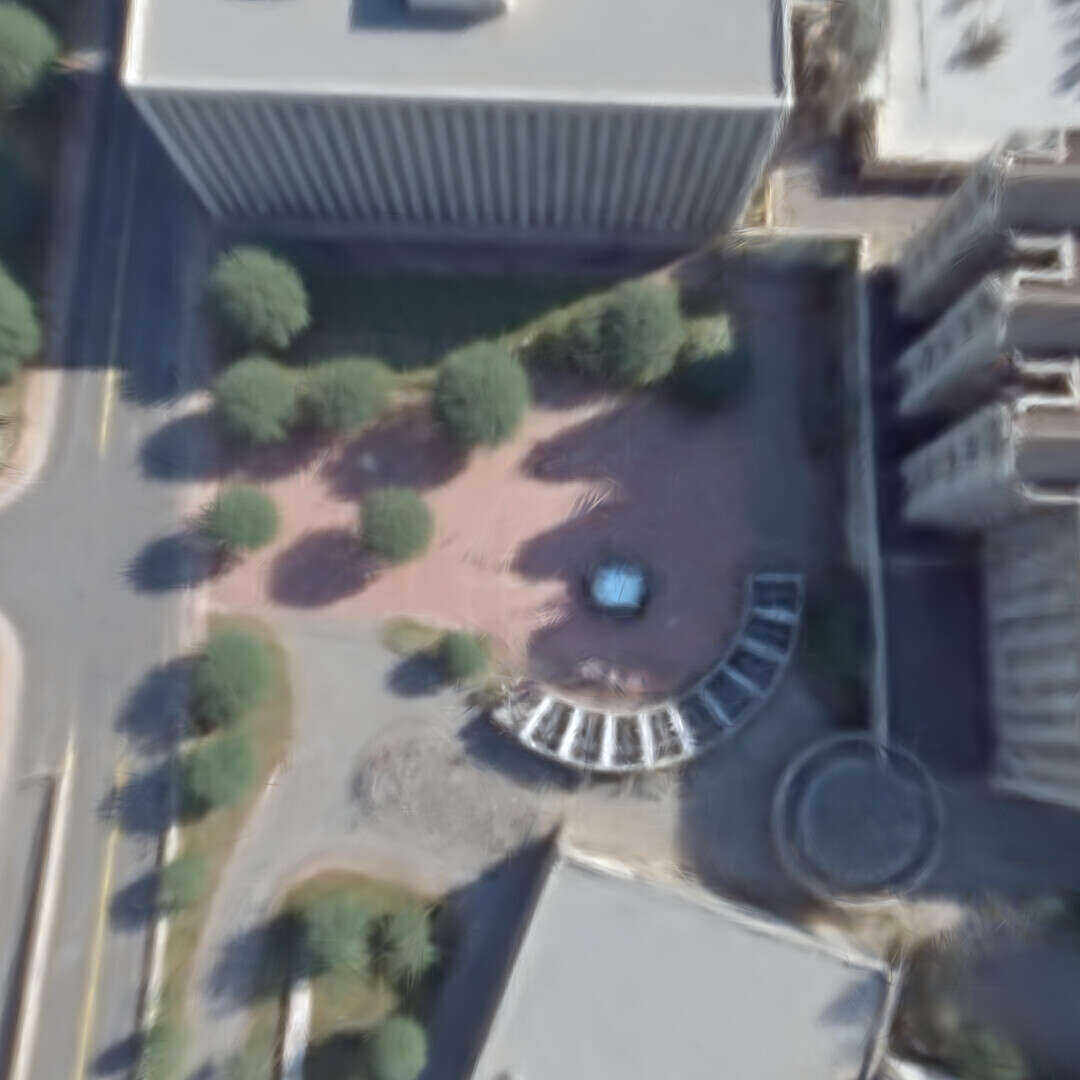} & 
    \imagecell[0.24]{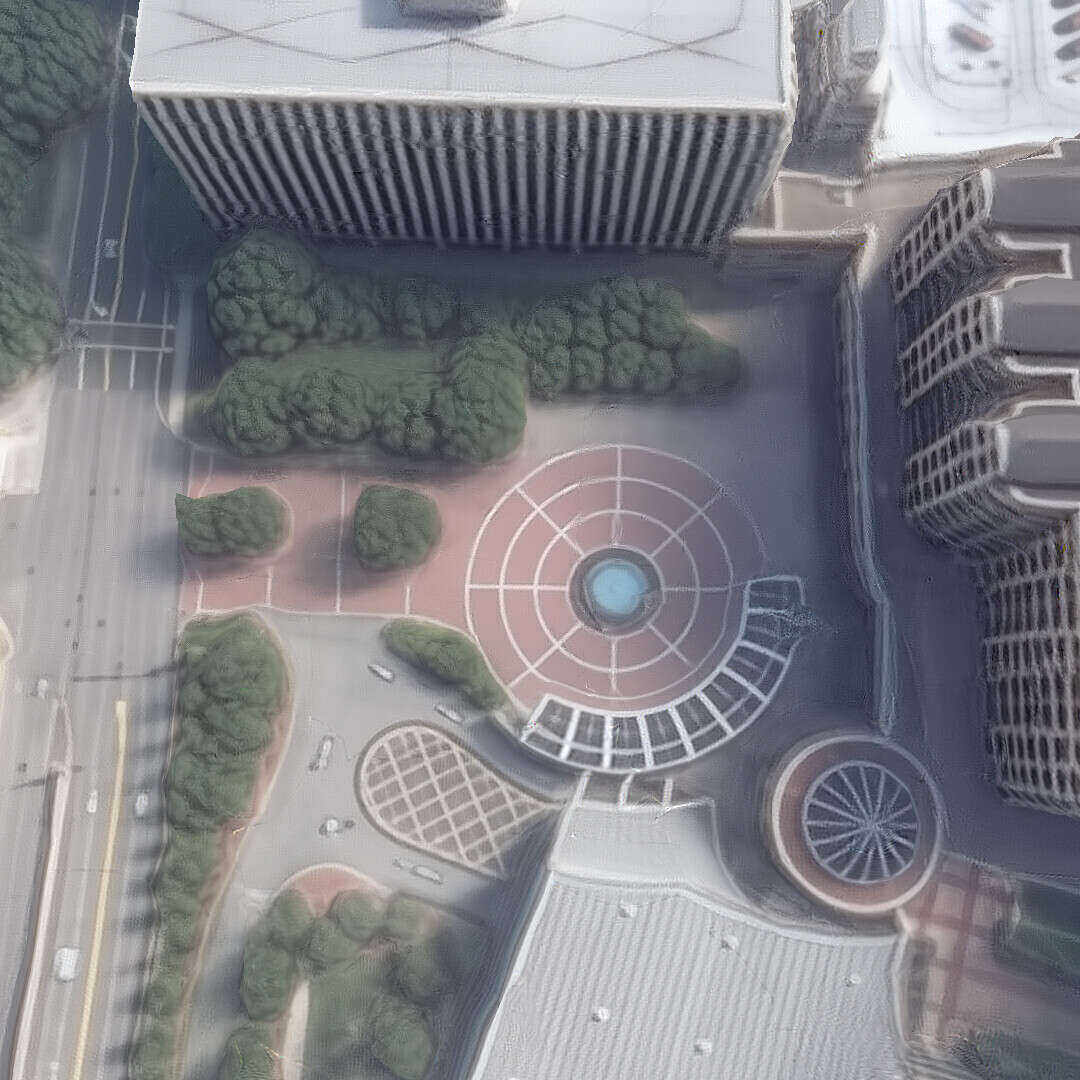} &
    \imagecell[0.24]{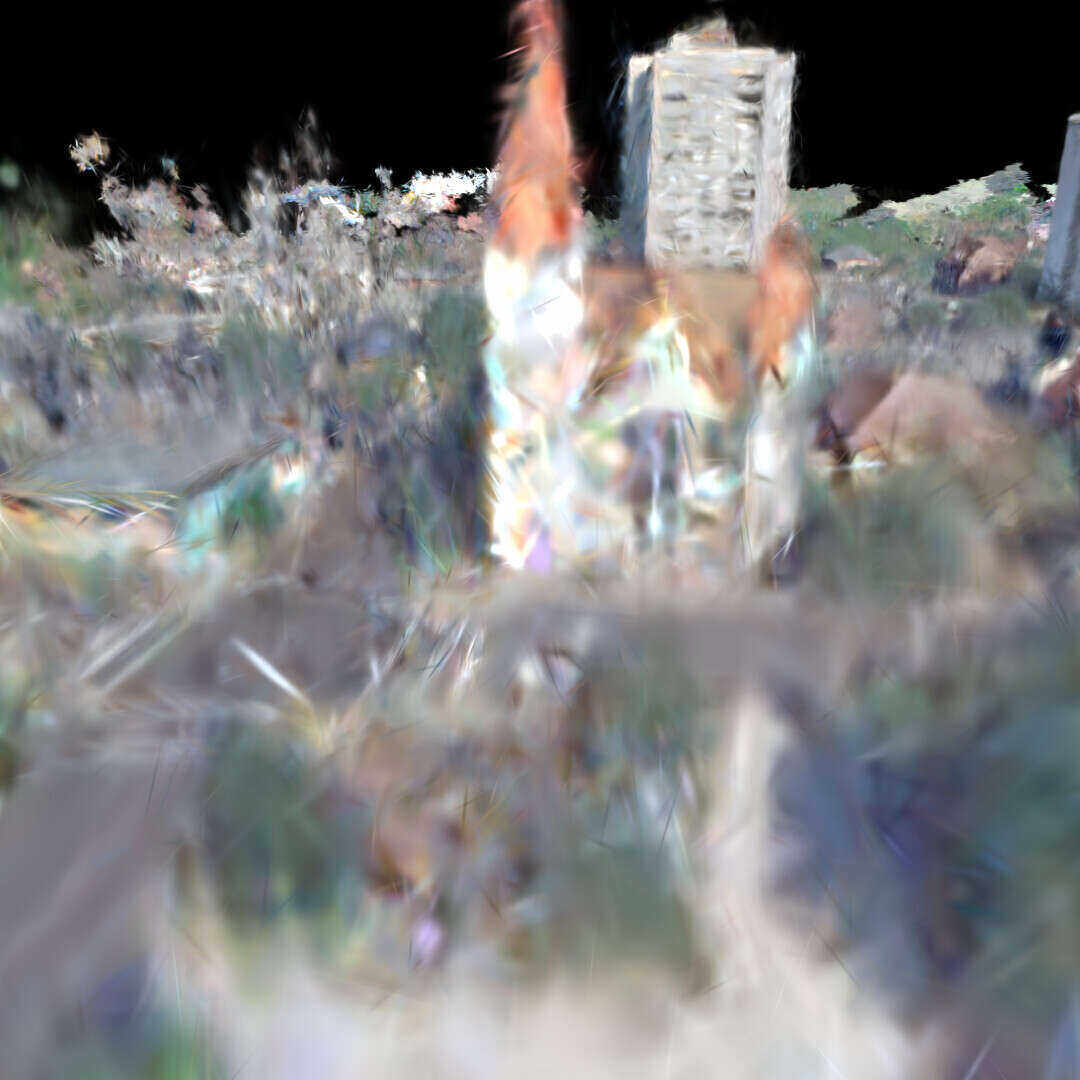} & 
    \imagecell[0.24]{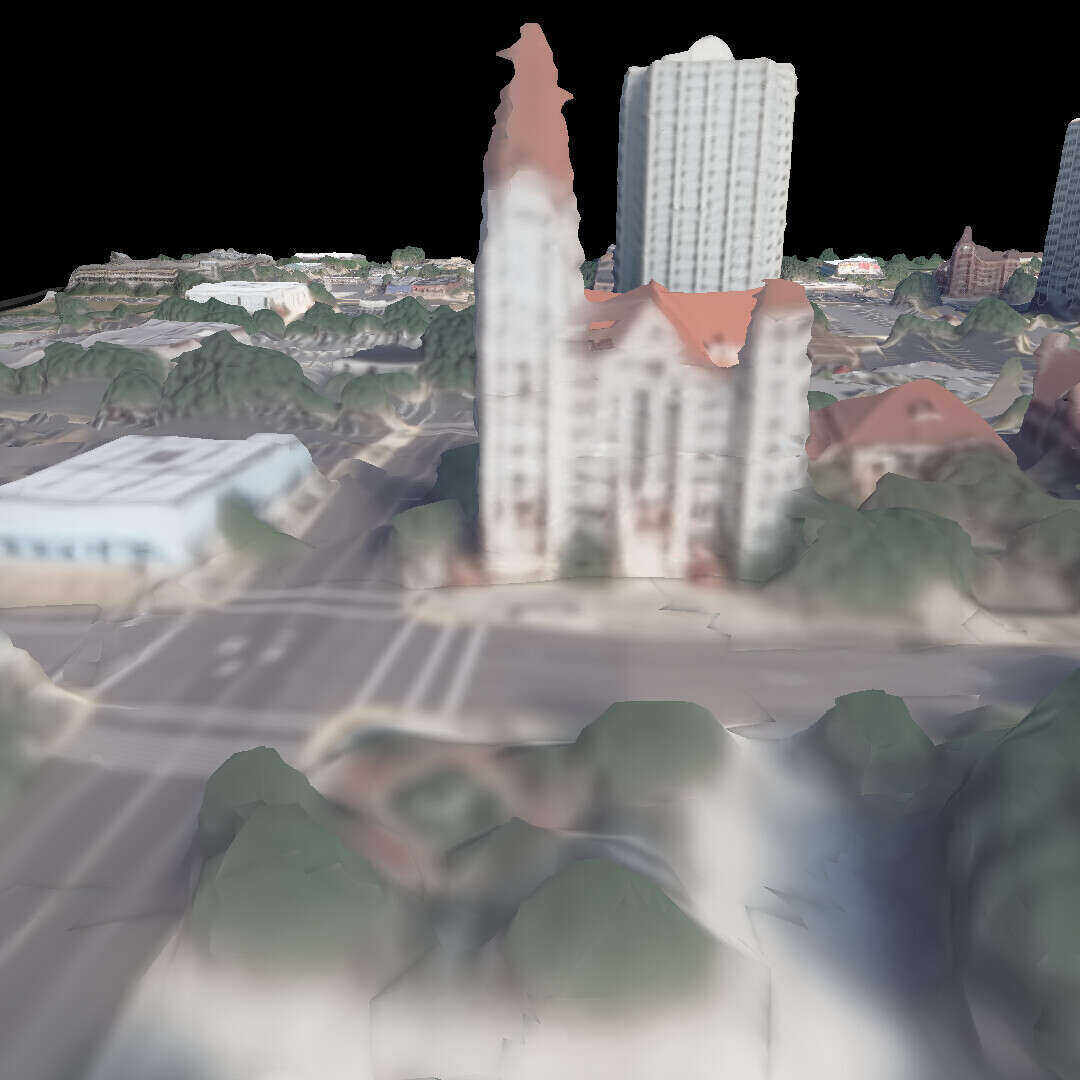} \\
    \vspace*{-10pt} \\
    
    \imagecell[0.24]{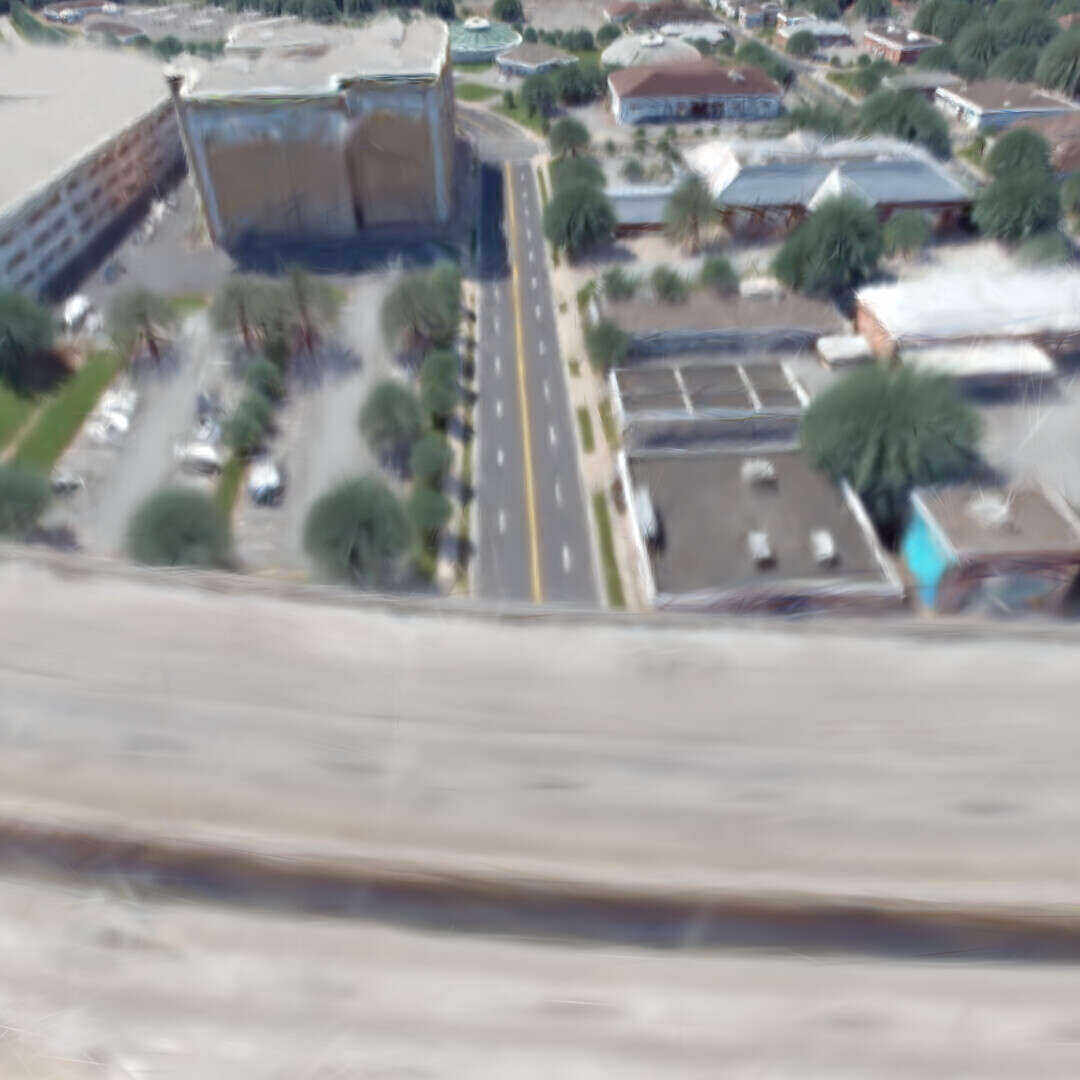} & 
    \imagecell[0.24]{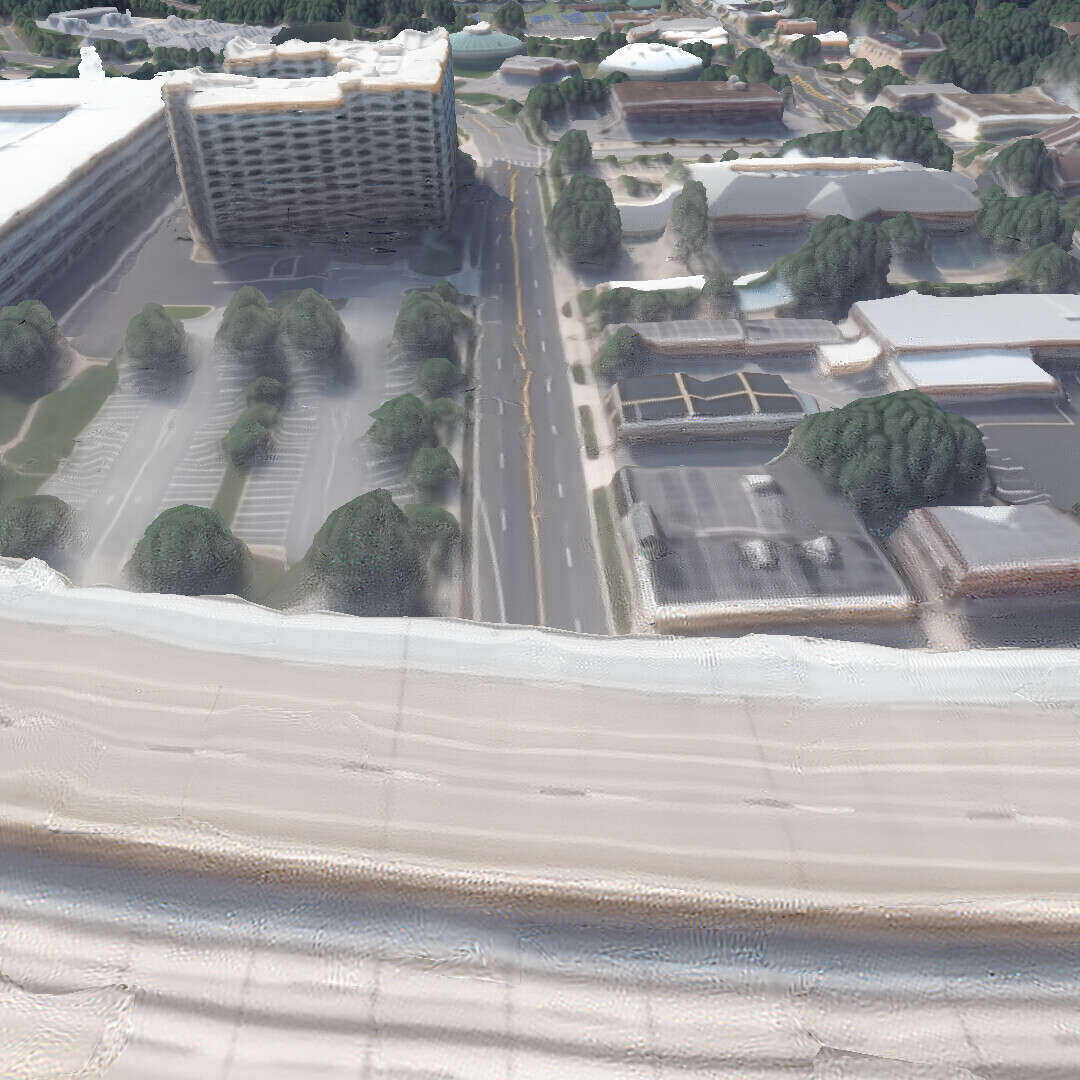} &
    \imagecell[0.24]{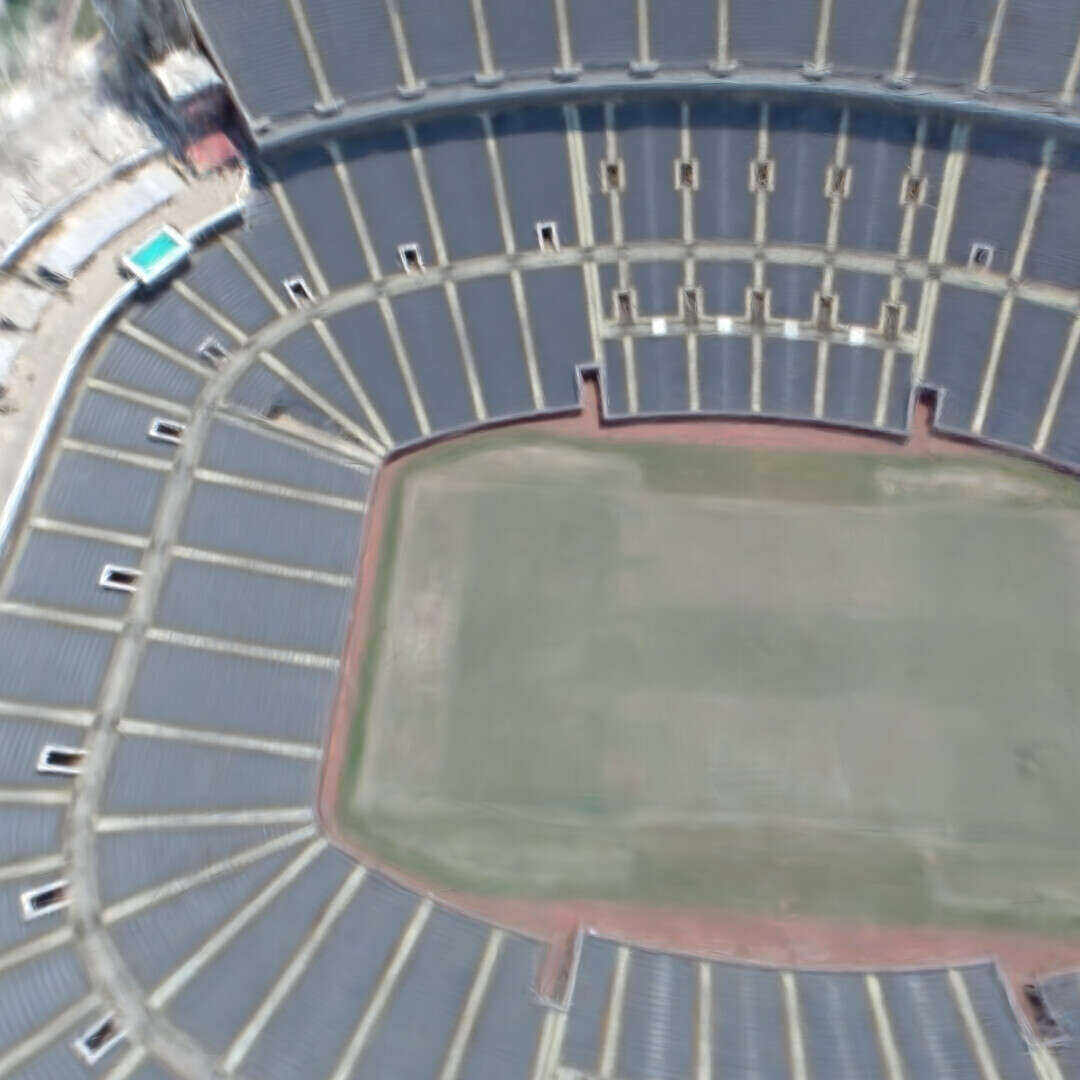} & 
    \imagecell[0.24]{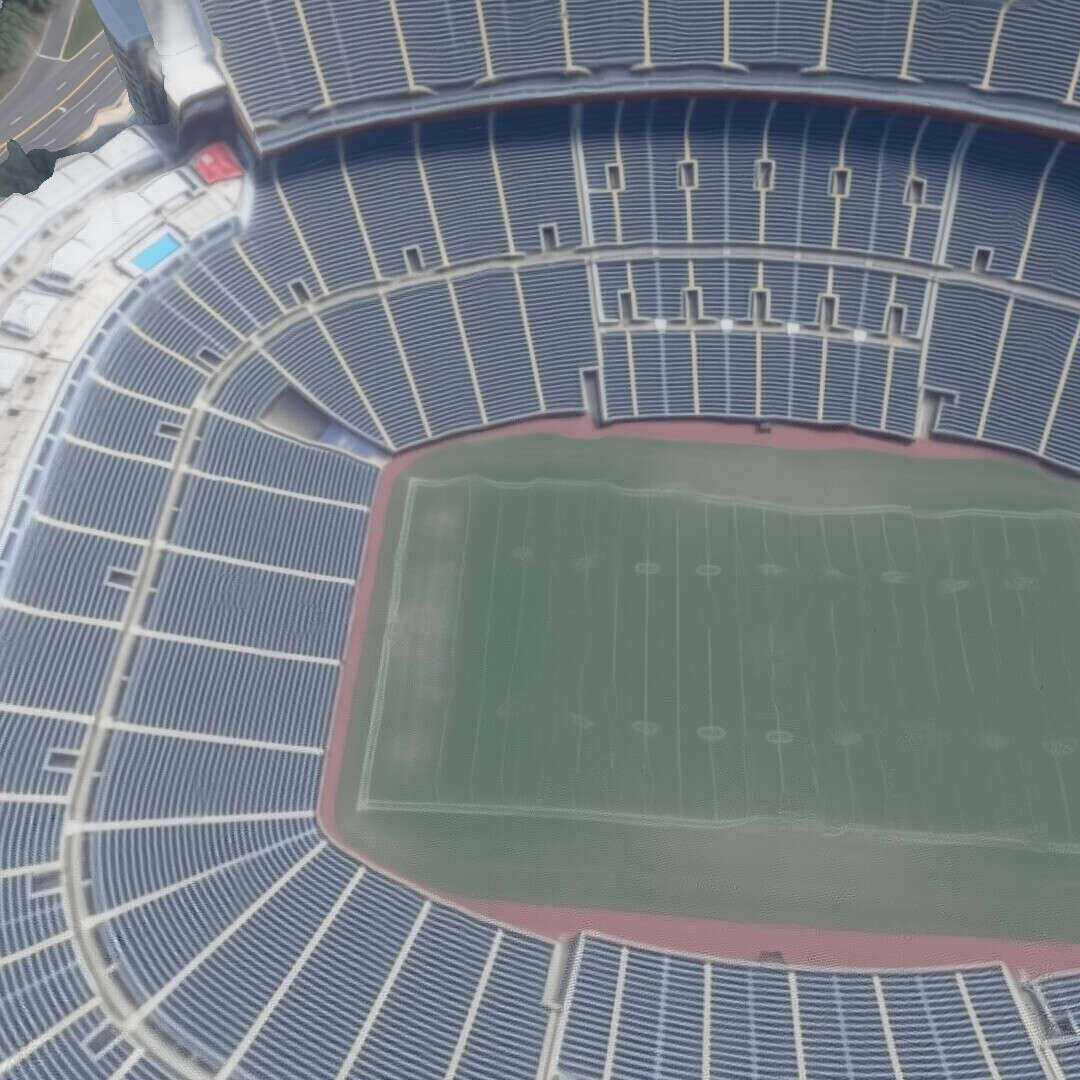} \\
    \vspace*{-10pt} \\
    
    \imagecell[0.24]{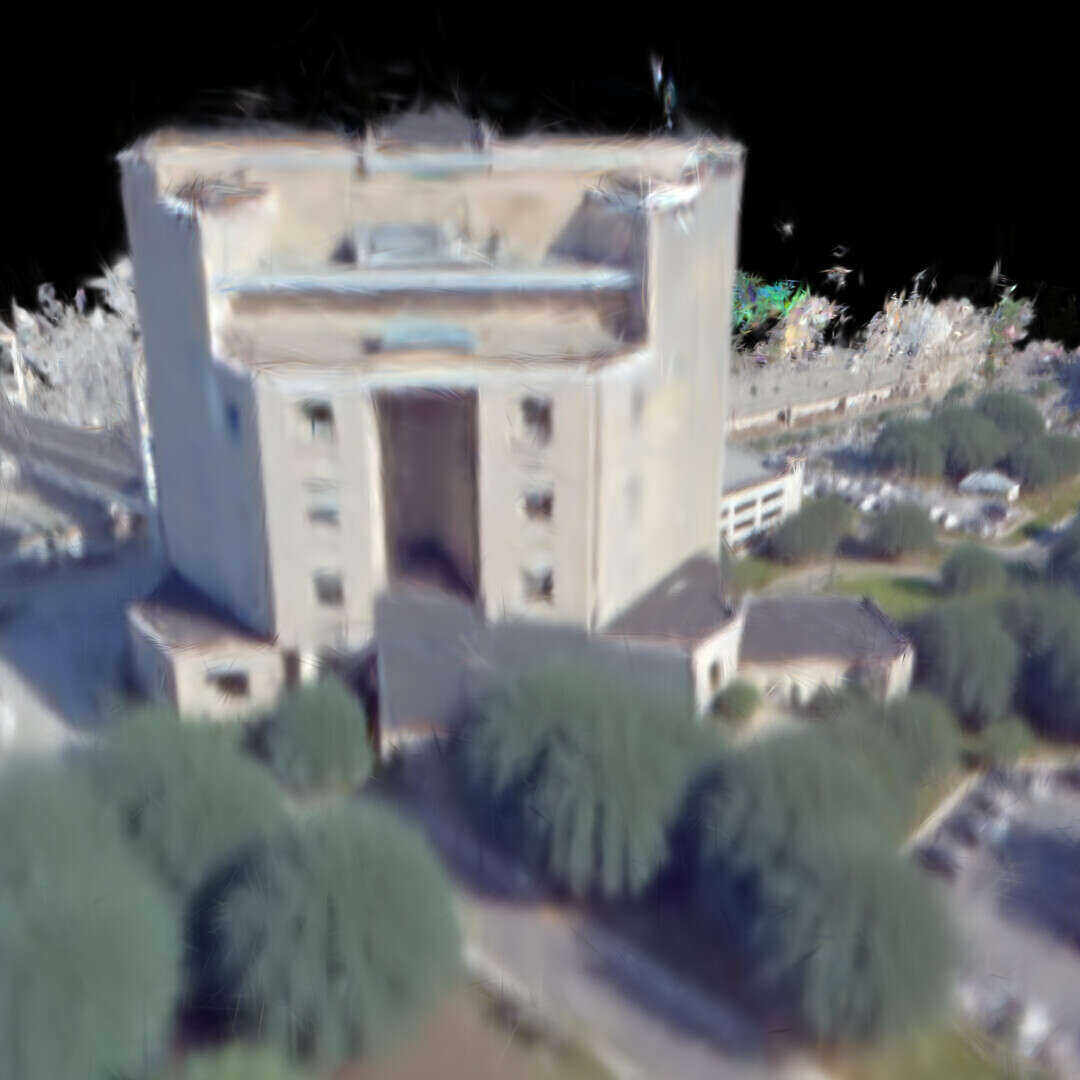} & 
    \imagecell[0.24]{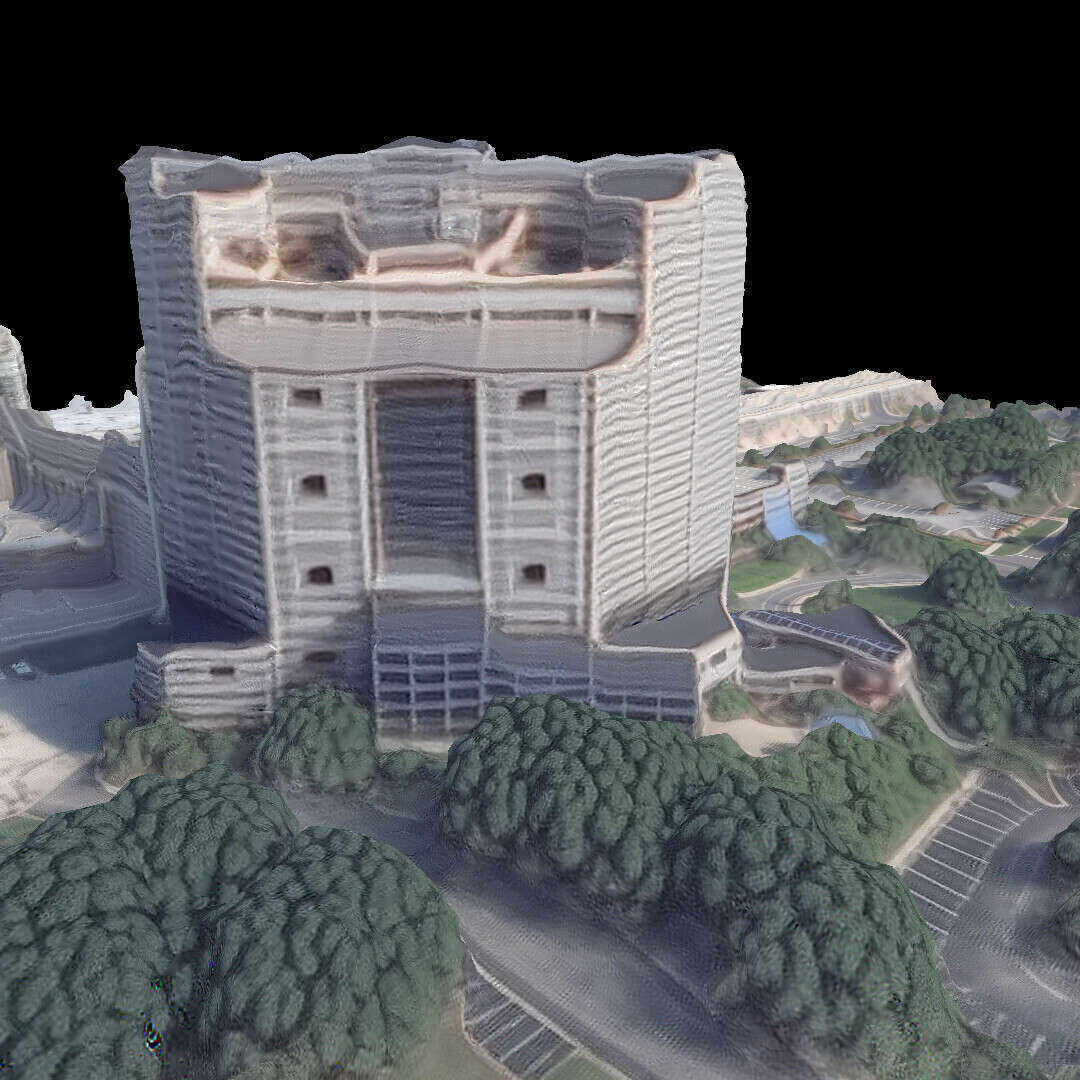} &
    \imagecell[0.24]{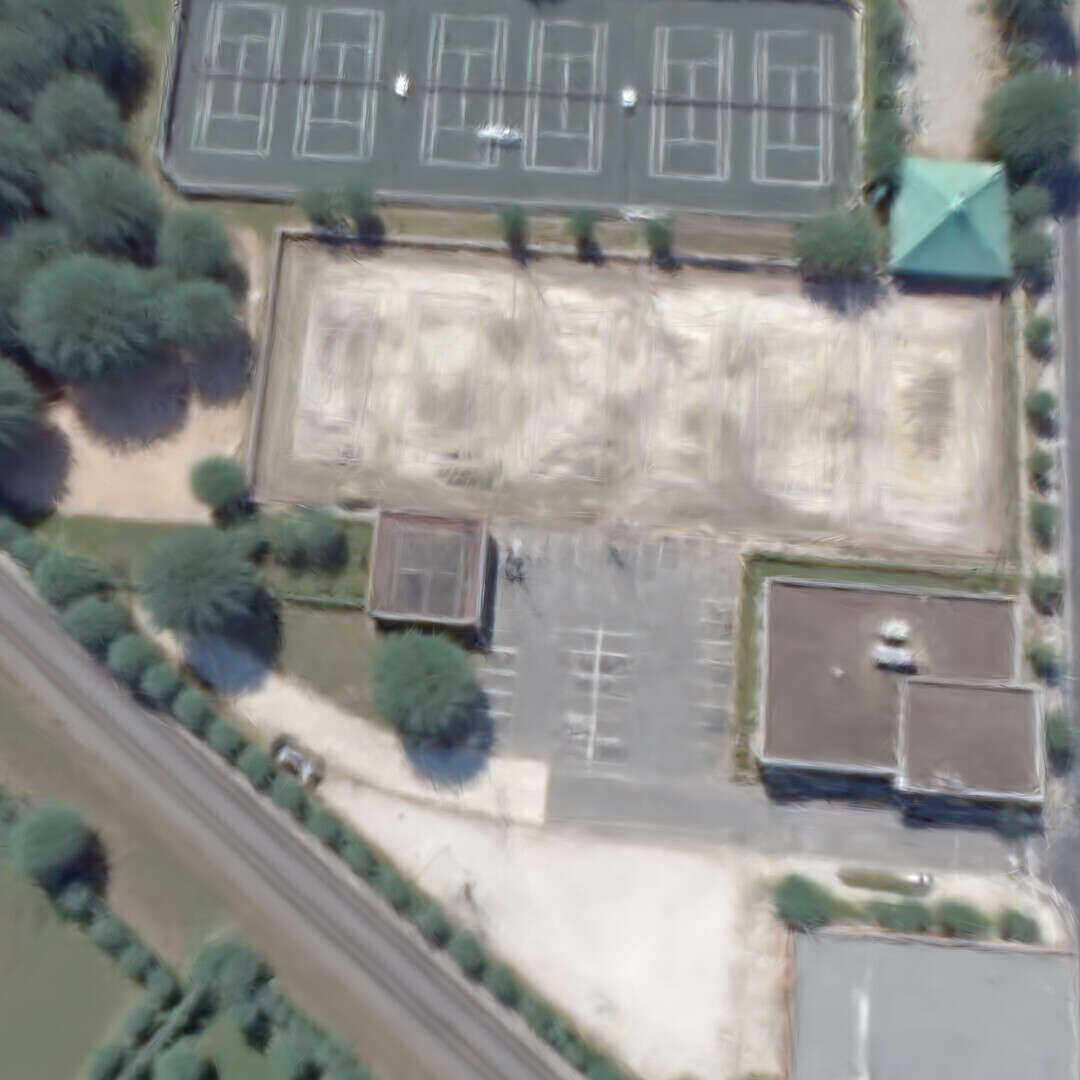} & 
    \imagecell[0.24]{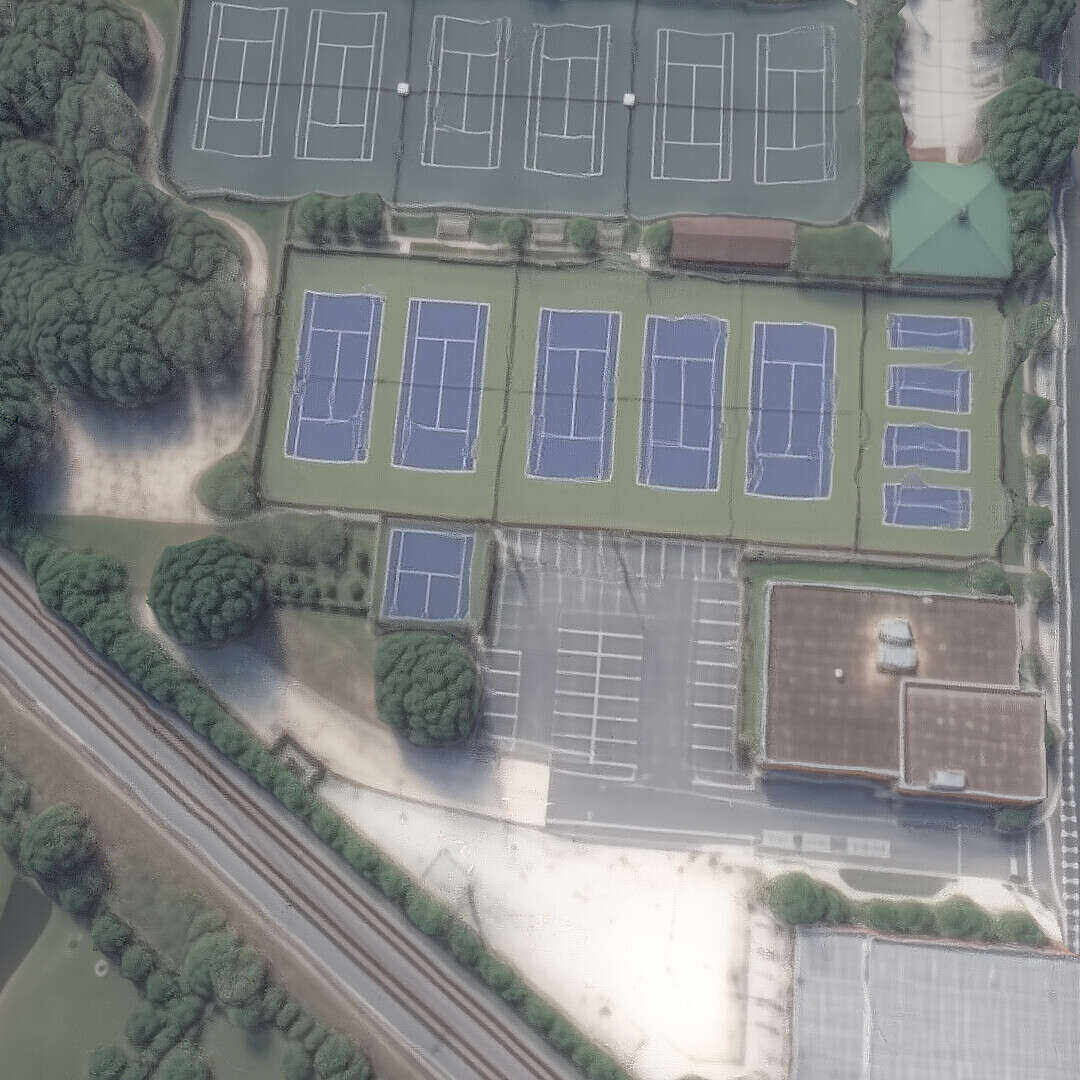} \\
    \vspace*{-10pt} \\
    
    \imagecell[0.24]{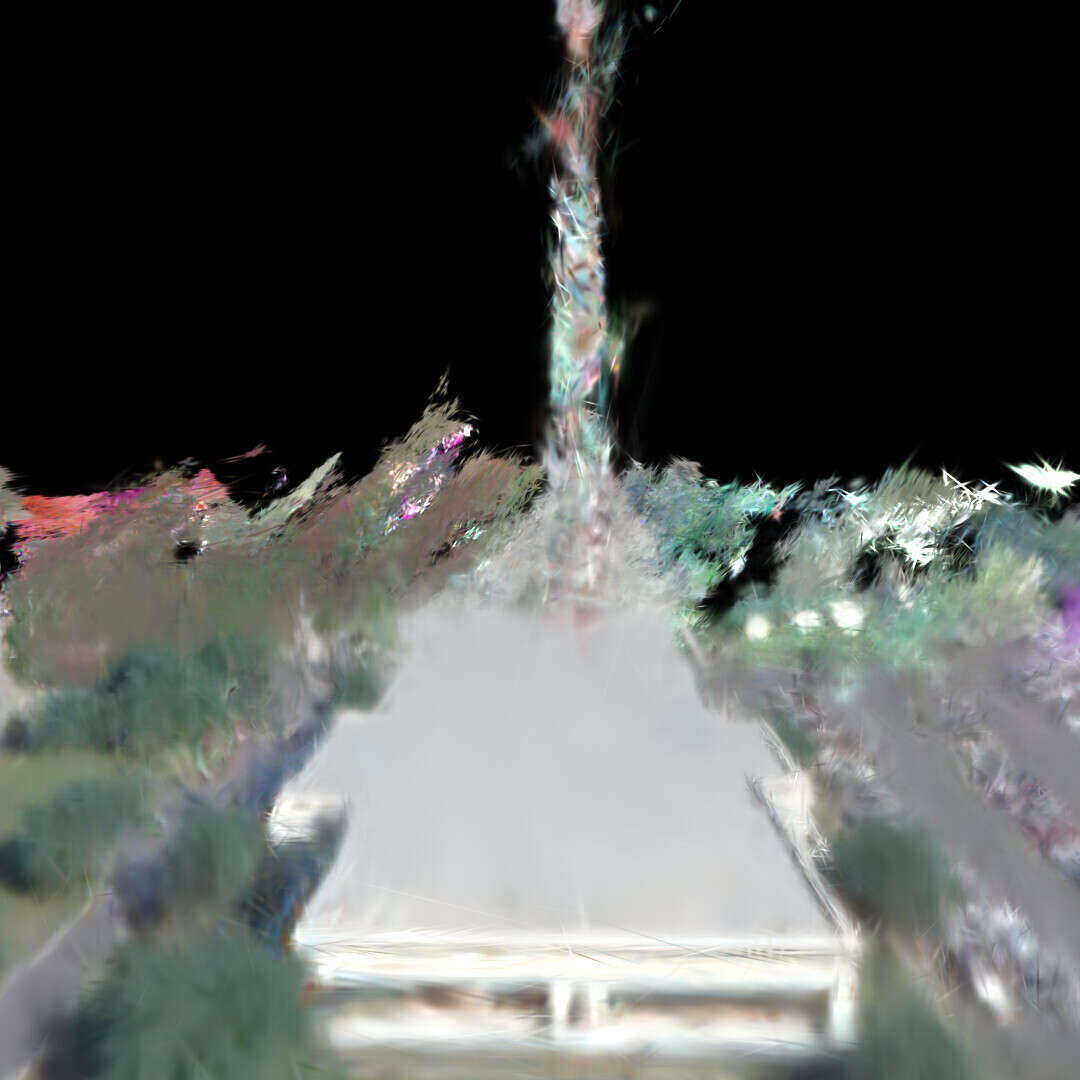} & 
    \imagecell[0.24]{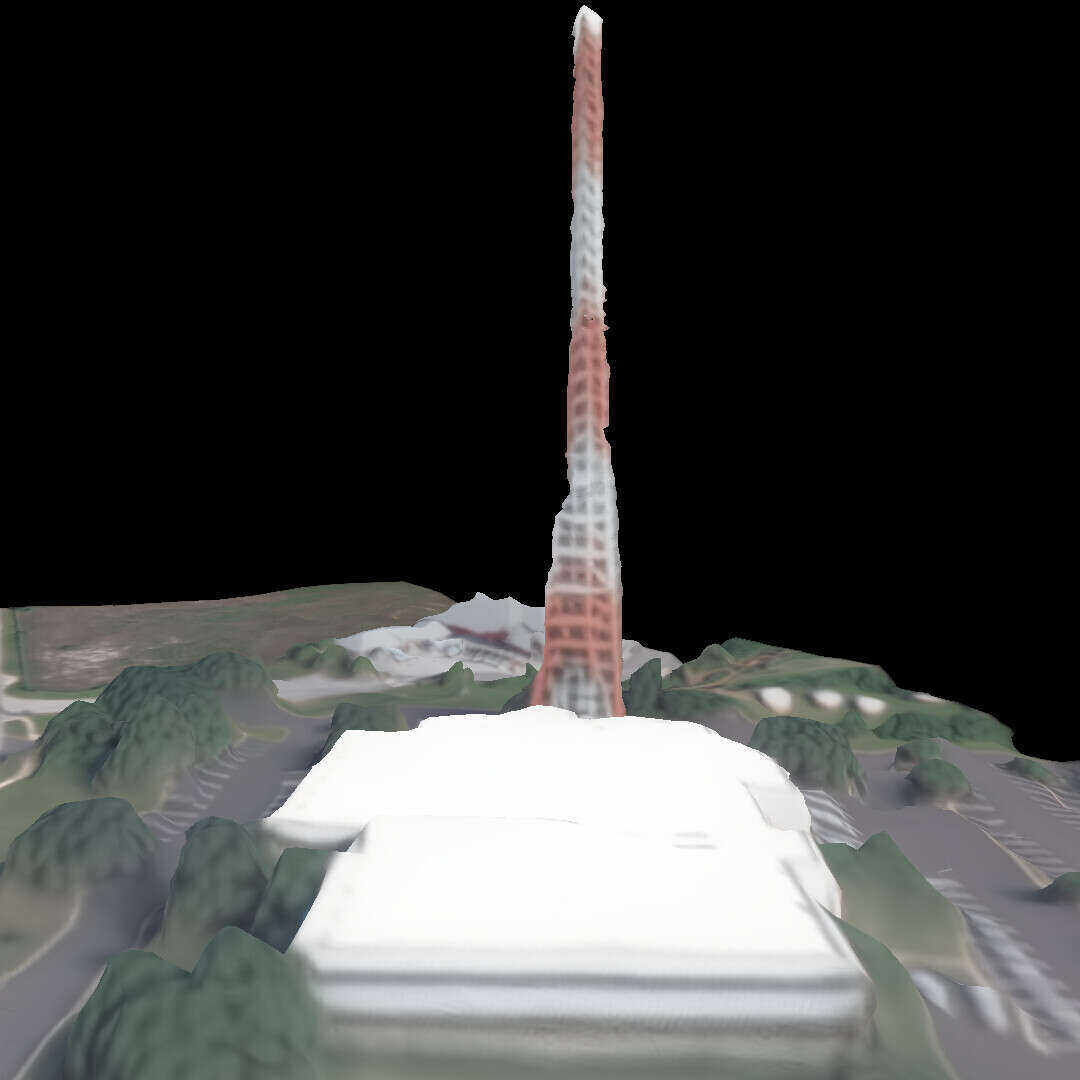} &
    \imagecell[0.24]{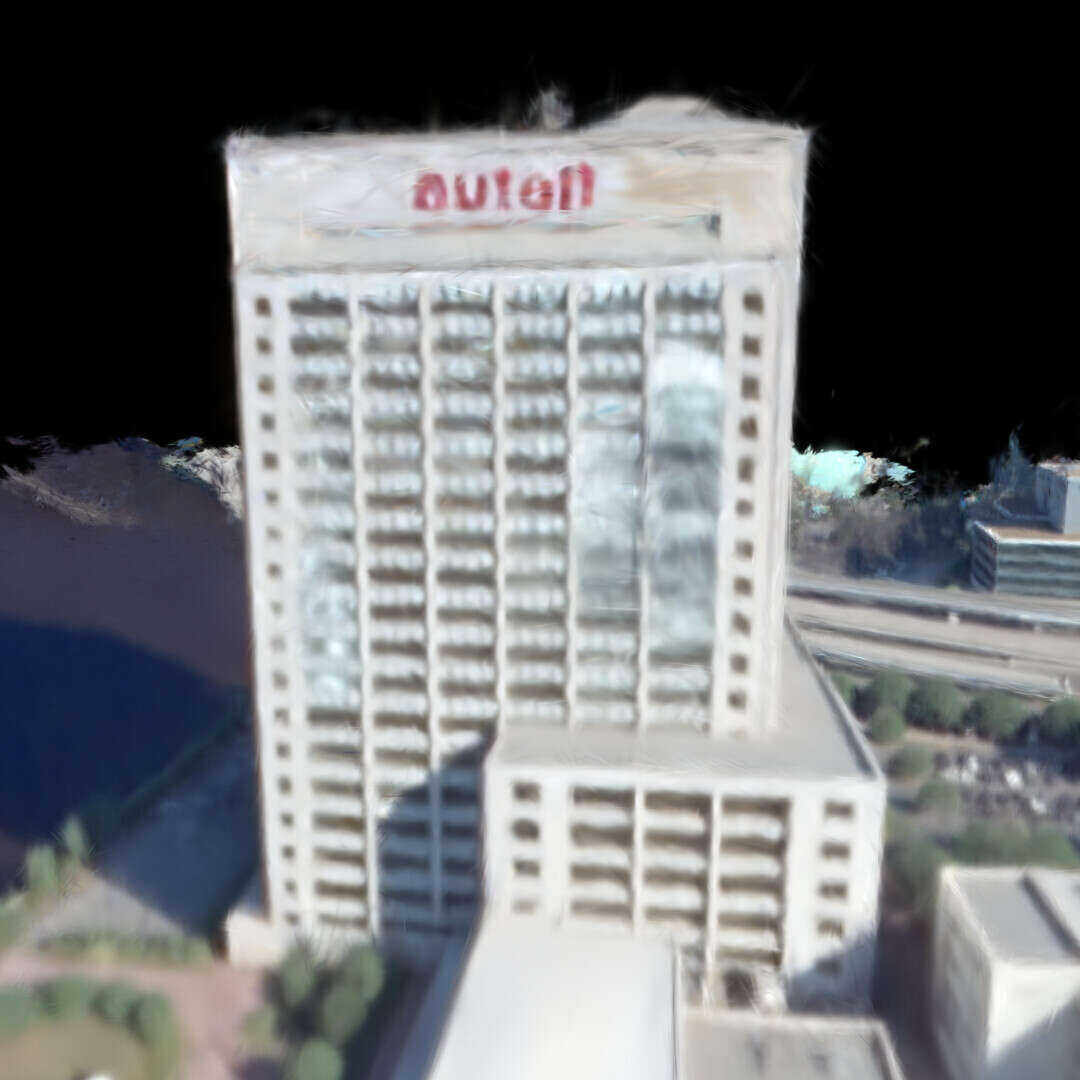} & 
    \imagecell[0.24]{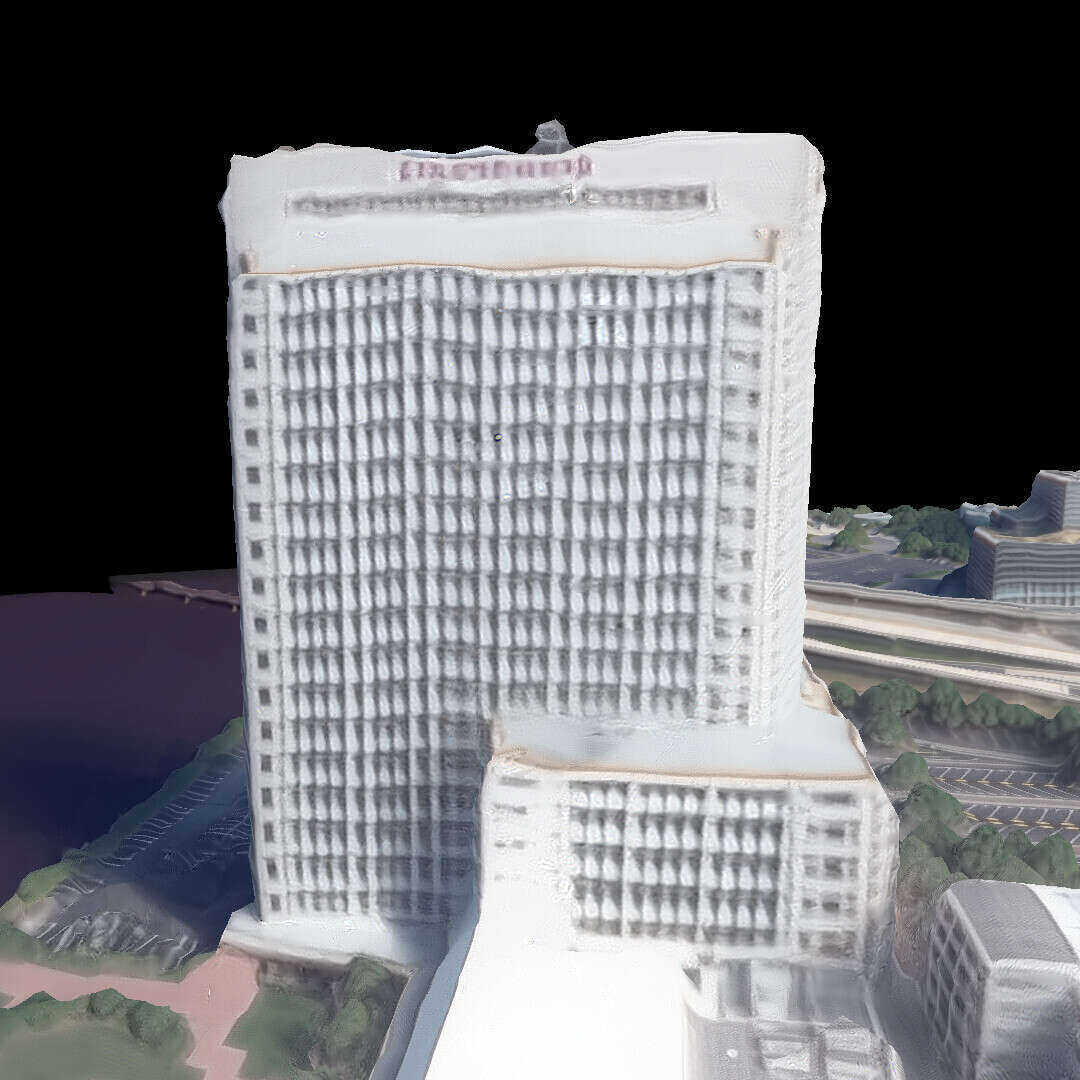} \\
    \vspace*{-10pt} \\
    
    \\
    \vspace*{-20pt}
    \\
    Skyfall-GS & 
    Ours &
    Skyfall-GS & 
    Ours \\
    
    \end{tabular}
    \end{spacing}
	\caption{ 
    \textbf{Close-Up views of reconstruction results of the DFC 2019 datasets}. 
    Our reconstructions preserve fine facade structures and building edges under low-altitude viewpoints, whereas Skyfall-GS \cite{lee2025skyfall} often produces blurred or smeared textures, confirming our improved near-view fidelity on real satellite data. 
    }
    \label{fig:supp:qual-jax-close}
    \vspace*{-0.3cm}
\end{figure*}

\begin{figure*}[p]
	\centering
    \begin{spacing}{1} 
    \setlength\tabcolsep{1pt}
    \begin{tabular}{cccc}
    
    \imagecell[0.24]{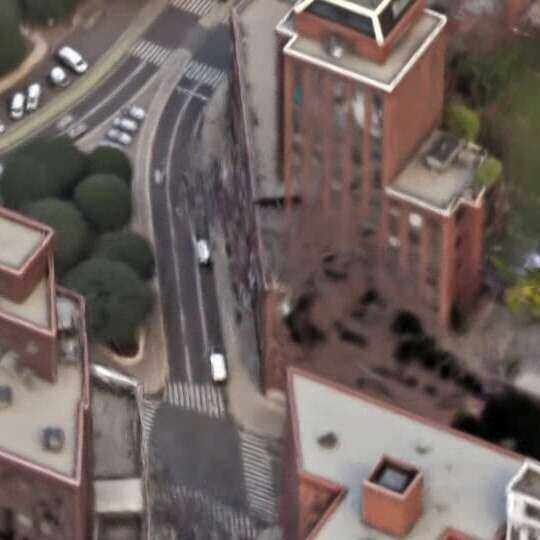} & 
    \imagecell[0.24]{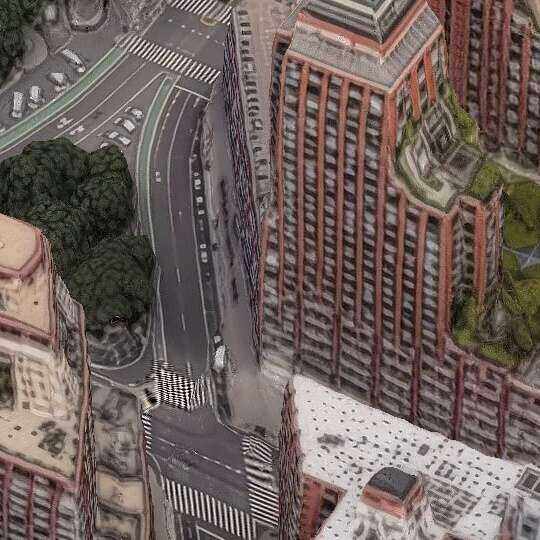} &
    \imagecell[0.24]{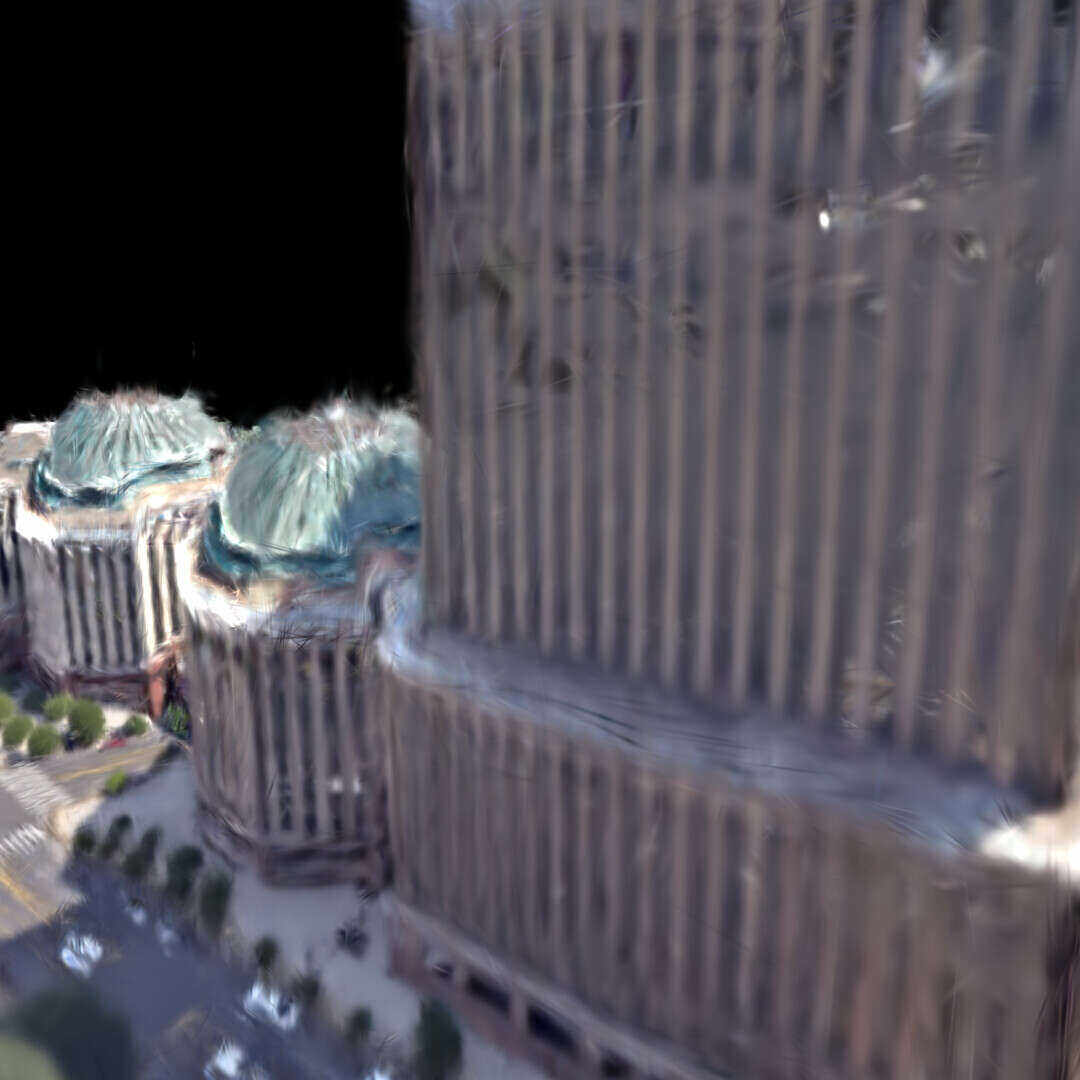} & 
    \imagecell[0.24]{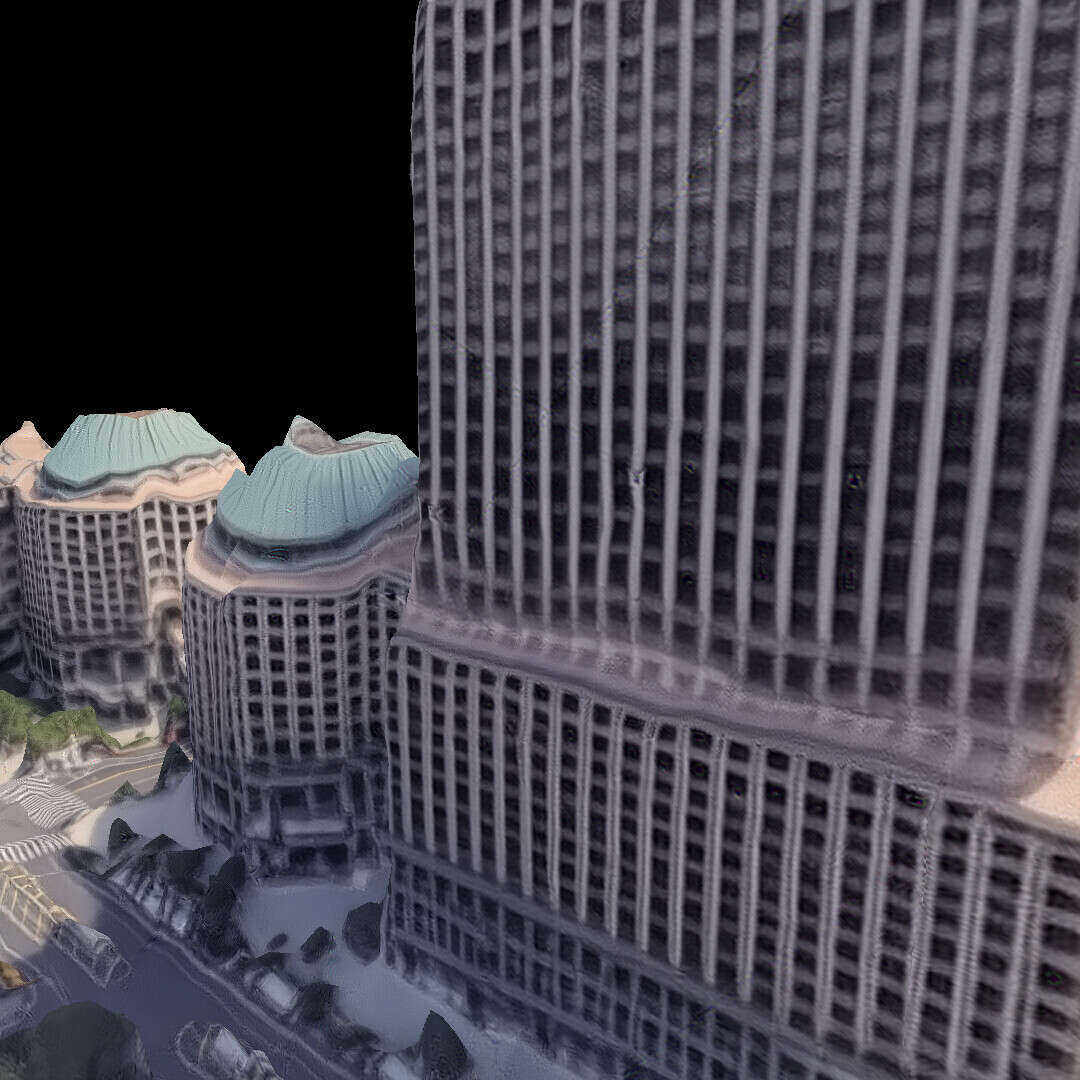} \\
    \vspace*{-10pt} \\

    \imagecell[0.24]{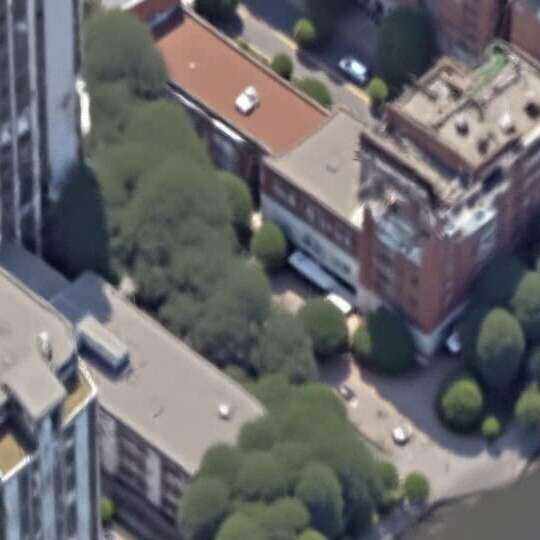} & 
    \imagecell[0.24]{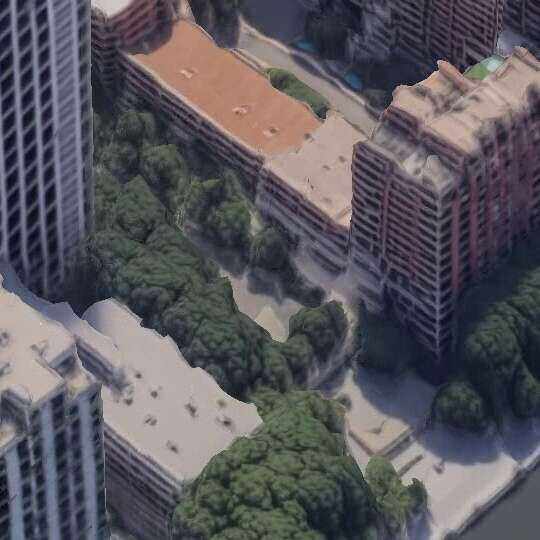} &
    \imagecell[0.24]{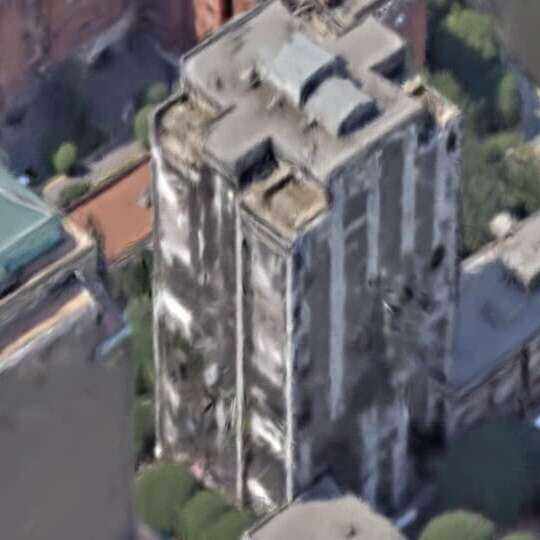} & 
    \imagecell[0.24]{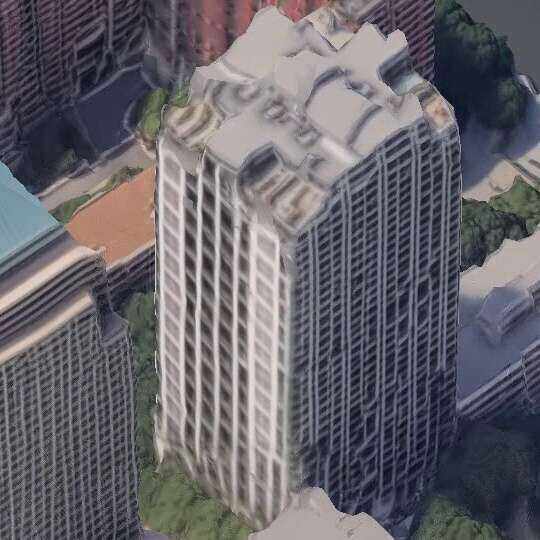} \\
    \vspace*{-10pt} \\

    \imagecell[0.24]{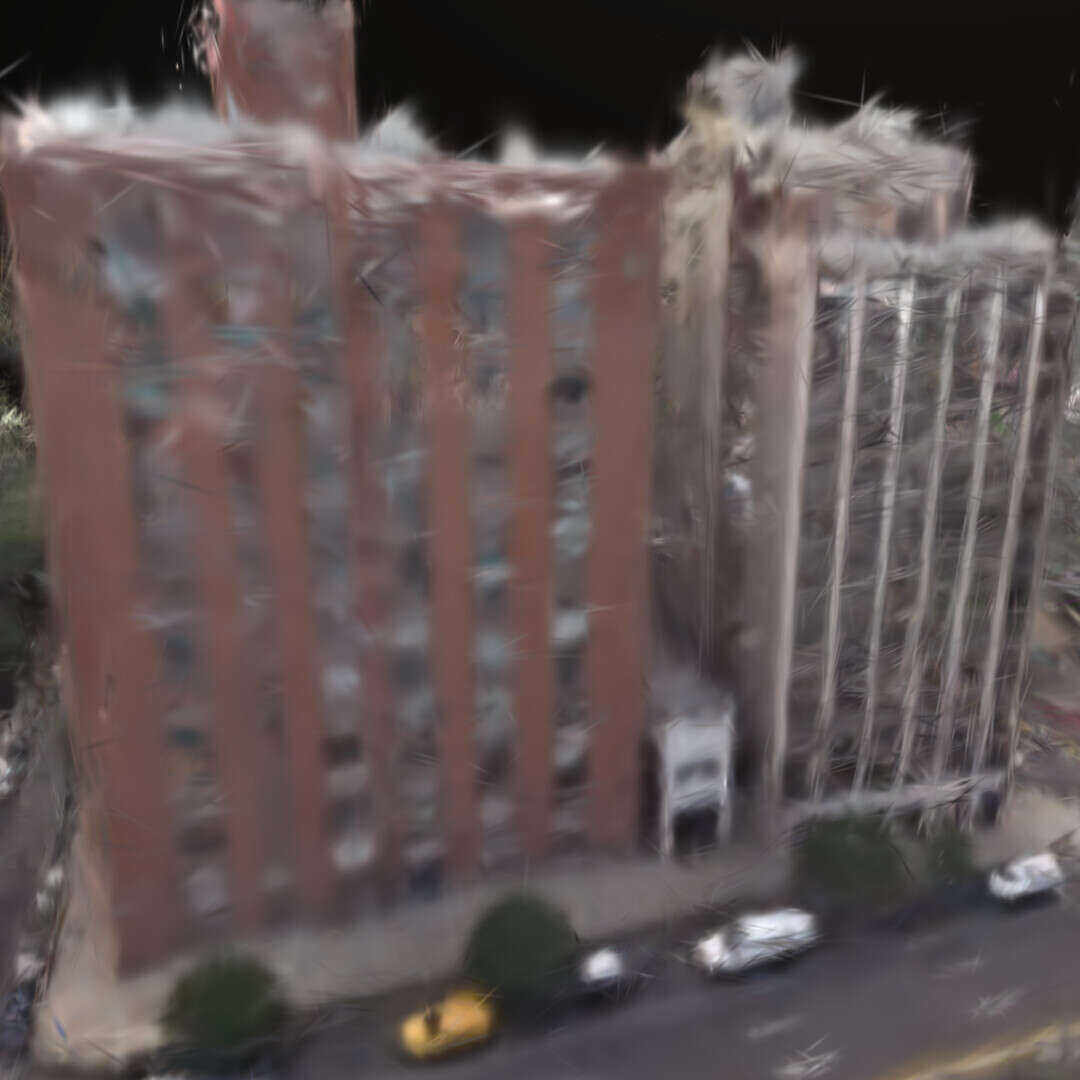} & 
    \imagecell[0.24]{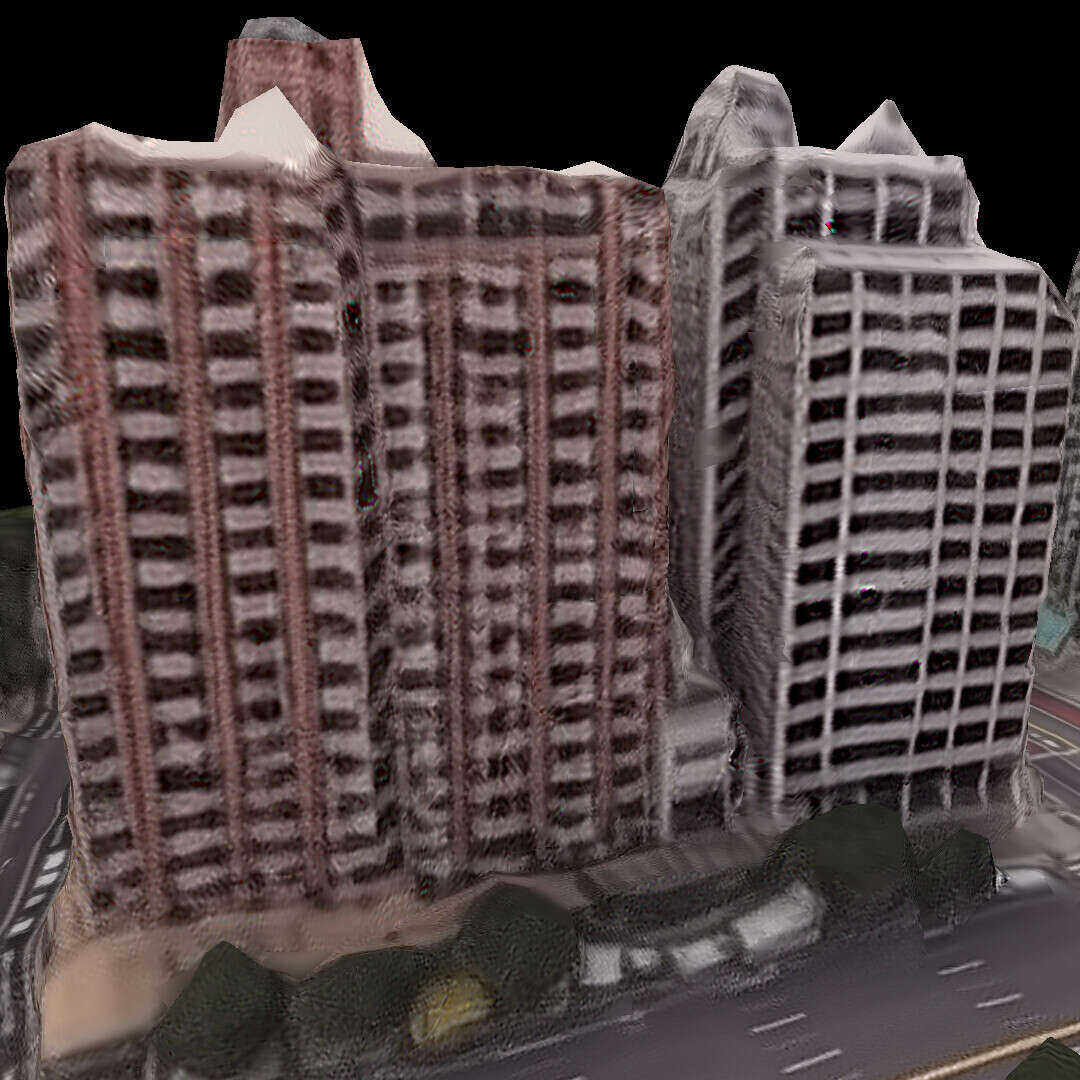} &
    \imagecell[0.24]{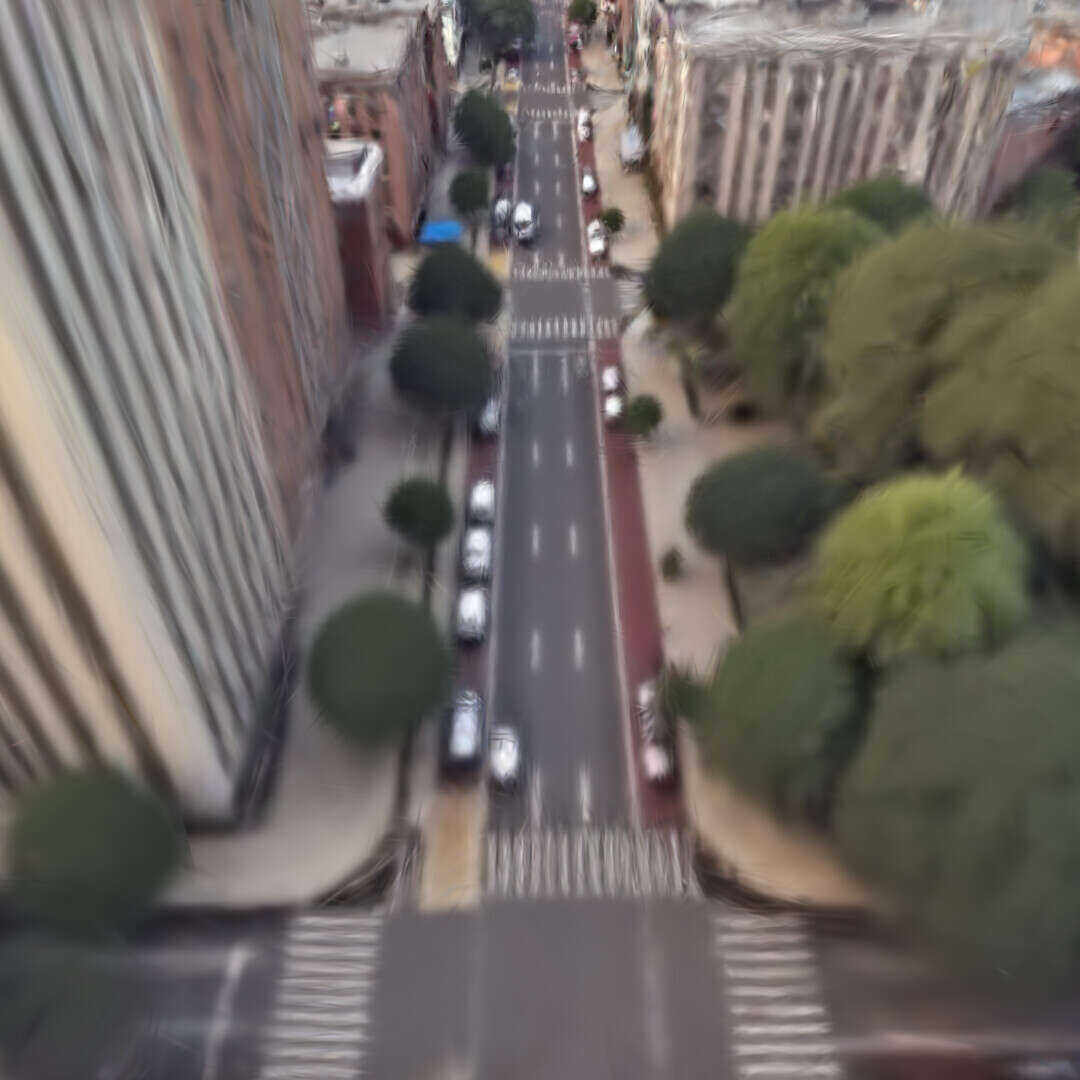} & 
    \imagecell[0.24]{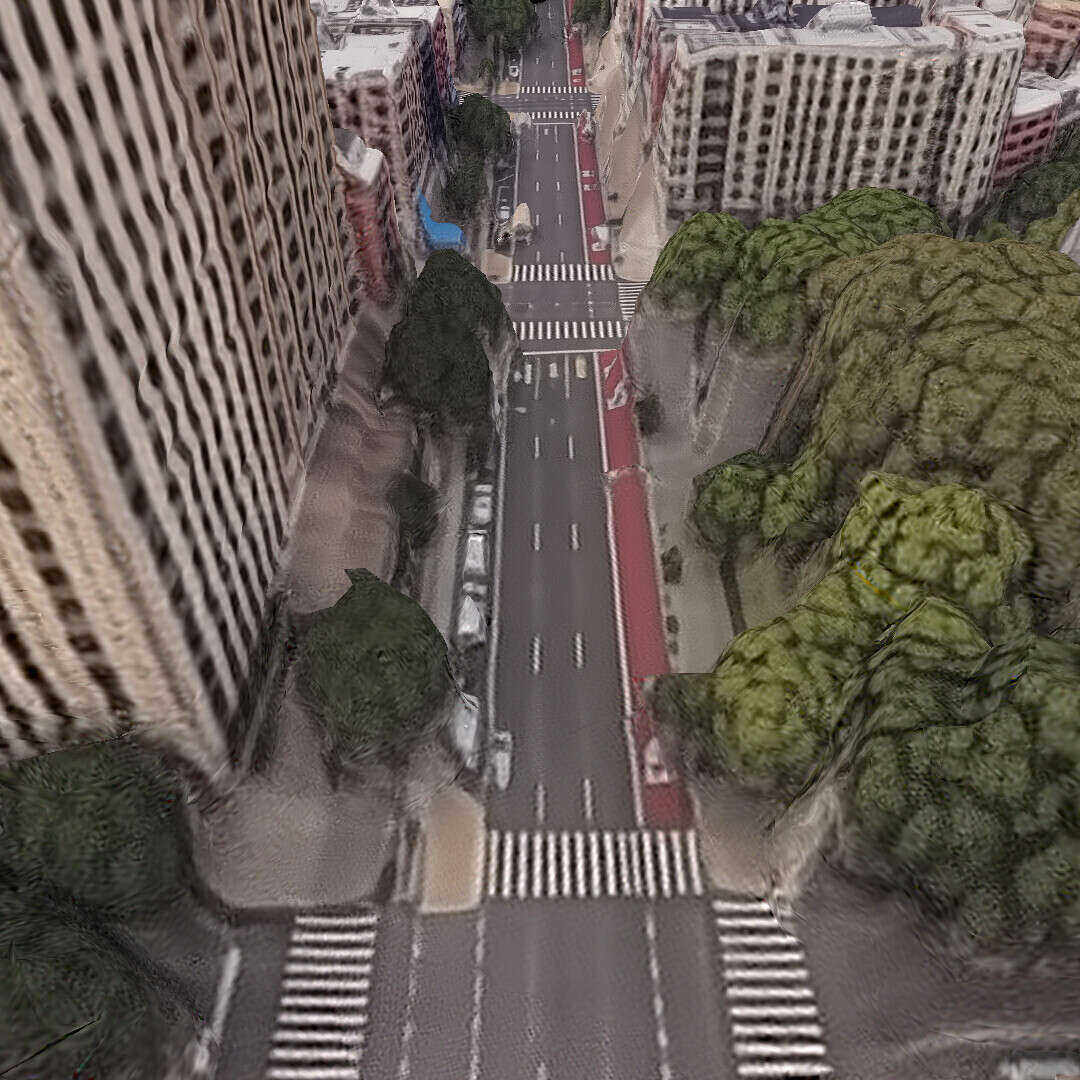} \\
    \vspace*{-10pt} \\

    \imagecell[0.24]{figures/supp/close/NYC_336/skyfall-gs/1.jpg} & 
    \imagecell[0.24]{figures/supp/close/NYC_336/ours-stage2/1.jpg} &
    \imagecell[0.24]{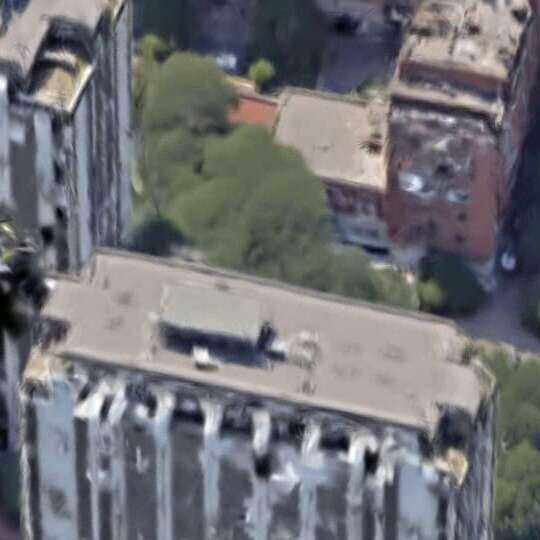} & 
    \imagecell[0.24]{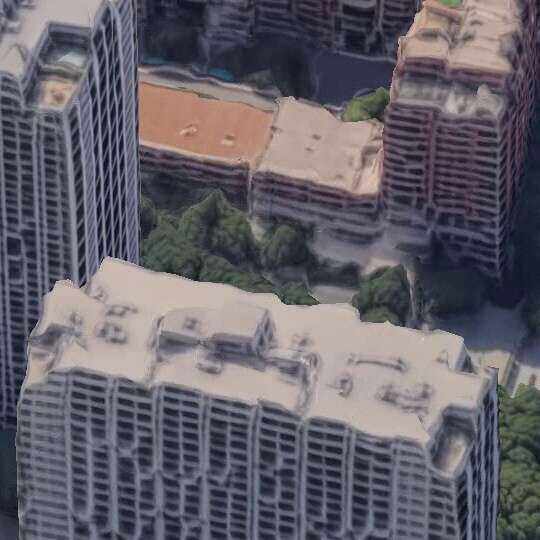} \\
    
    \vspace*{-10pt} \\
    
    \\
    \vspace*{-20pt}
    \\
    Skyfall-GS & 
    Ours &
    Skyfall-GS & 
    Ours \\
    
    \end{tabular}
    \end{spacing}
	\caption{ 
    \textbf{Close-Up views of reconstruction results of the Google Earth datasets}. 
    Our reconstructions preserve fine facade structures and building edges under low-altitude viewpoints, whereas Skyfall-GS \cite{lee2025skyfall} produces smeared textures and even broken parts, confirming our improved near-view fidelity. 
    }
    \label{fig:supp:qual-nyc-close}
    \vspace*{-0.3cm}
\end{figure*}

Similarly, \cref{fig:supp:qual-jax-close,fig:supp:qual-nyc-close} present close-up comparisons on DFC 2019 and GoogleEarth, respectively.
Across all real-world scenes, our reconstructions retain fine-grained details such as window patterns and vertical edges better than Skyfall-GS, validating the effectiveness of our two-stage geometry and appearance modeling for close-up observations.

\subsection{Additional Real Urban Scenes}

Beyond the benchmark datasets used in the main paper, we further validate the generalization ability of our method on two additional real-world urban scenes.
These scenes are of the same type and scale as the \textit{Urban Scene} in the main paper, but are completely disjoint in terms of geographic location and appearance.

\begin{figure*}[p]
	\centering
    \includegraphics[width=\linewidth]{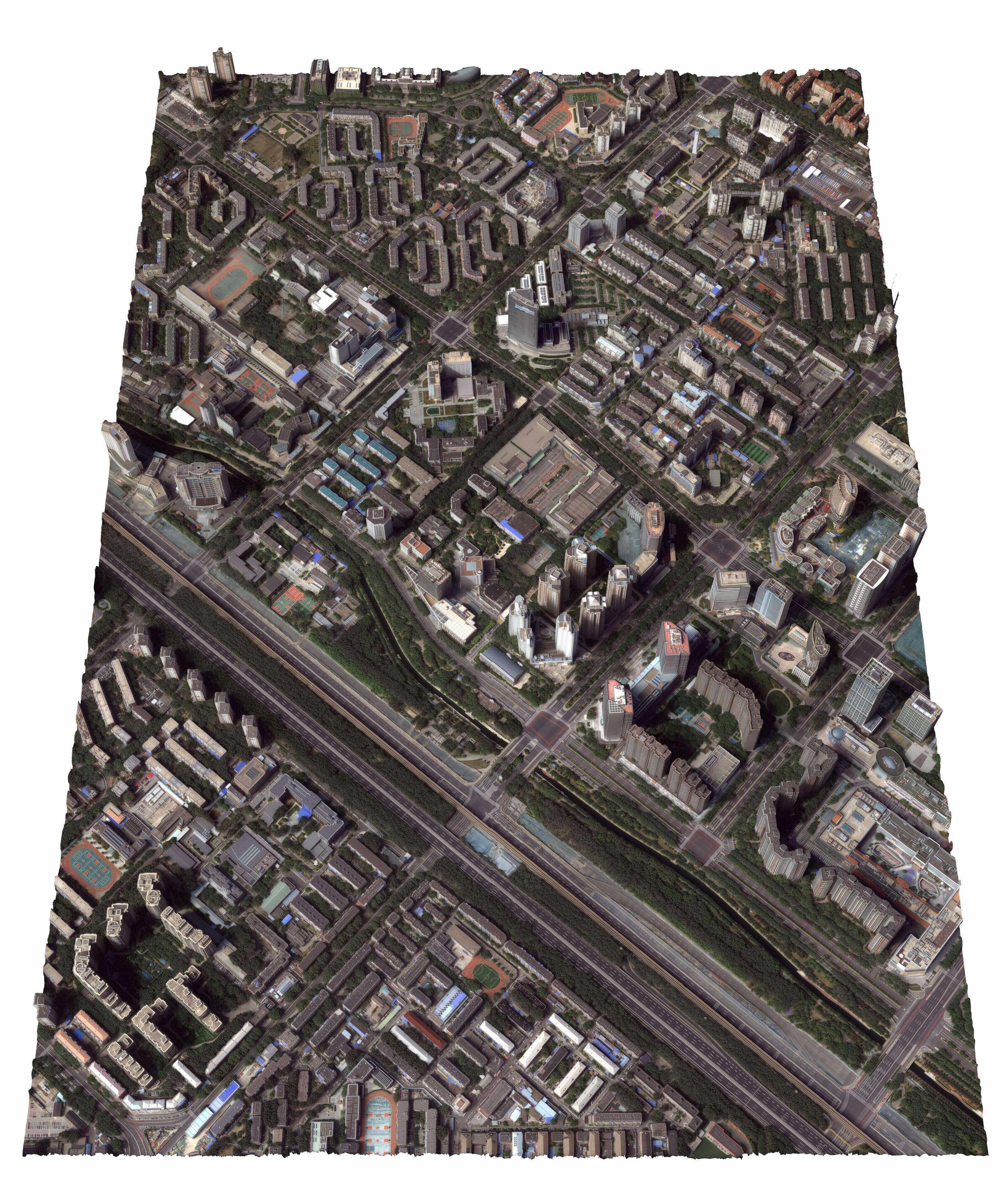}
	\caption{ 
    \textbf{Bird’s-Eye view of a Real Urban Scene Reconstruction}. 
    Our pipeline generalizes to new urban areas without dataset-specific tuning while preserving overall layout, building geometry, and texture plausibility. 
    }
    \label{fig:supp:large-jiuxianqiao}
    \vspace*{-0.3cm}
\end{figure*}

\begin{figure*}[p]
	\centering
    \includegraphics[width=\linewidth]{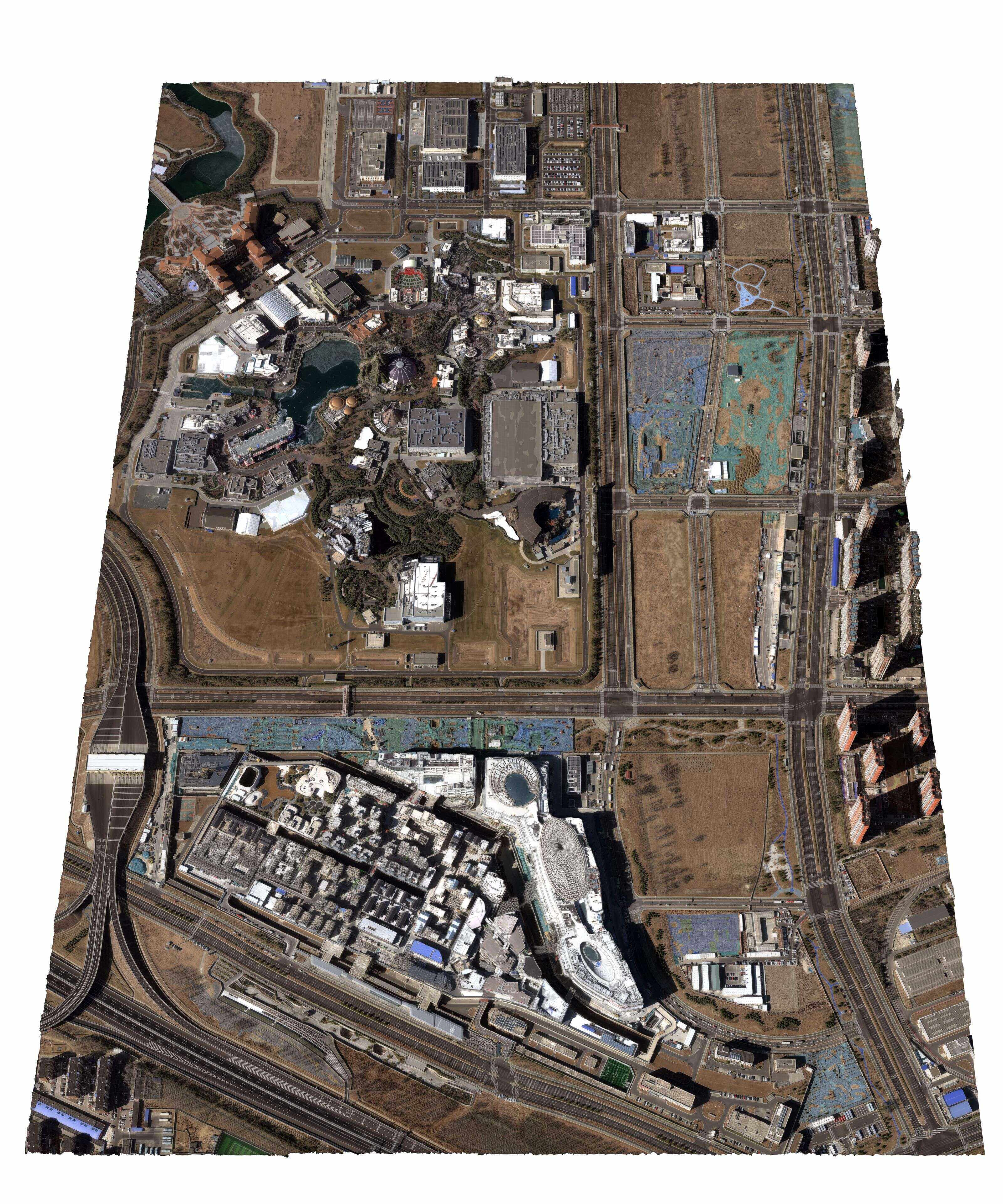}
	\caption{ 
    \textbf{Bird’s-Eye view of a Real Urban Scene Reconstruction}. 
    Our pipeline generalizes to new urban areas without dataset-specific tuning while preserving overall layout, building geometry, and texture plausibility. 
    }
    \label{fig:supp:large-tongzhou}
    \vspace*{-0.3cm}
\end{figure*}

\cref{fig:supp:large-jiuxianqiao,fig:supp:large-tongzhou} show bird’s-eye overviews of two scenes reconstructed by our method.
The results demonstrate that, without dataset-specific tuning, our approach can robustly handle diverse urban layouts, ranging from dense high-rise clusters to wide road networks and large open areas, while maintaining globally coherent geometry and visually plausible textures across each entire scene.

\subsection{Ablation Study}
\label{sub-sec:ablation-matrix}

We provide additional ablation results to complement the analysis in the main paper.

\begin{figure*}[p]
	\centering
    \begin{spacing}{1} 
    \setlength\tabcolsep{1pt}
    \begin{tabular}{cccccc}
    
    \imagecell[0.16]{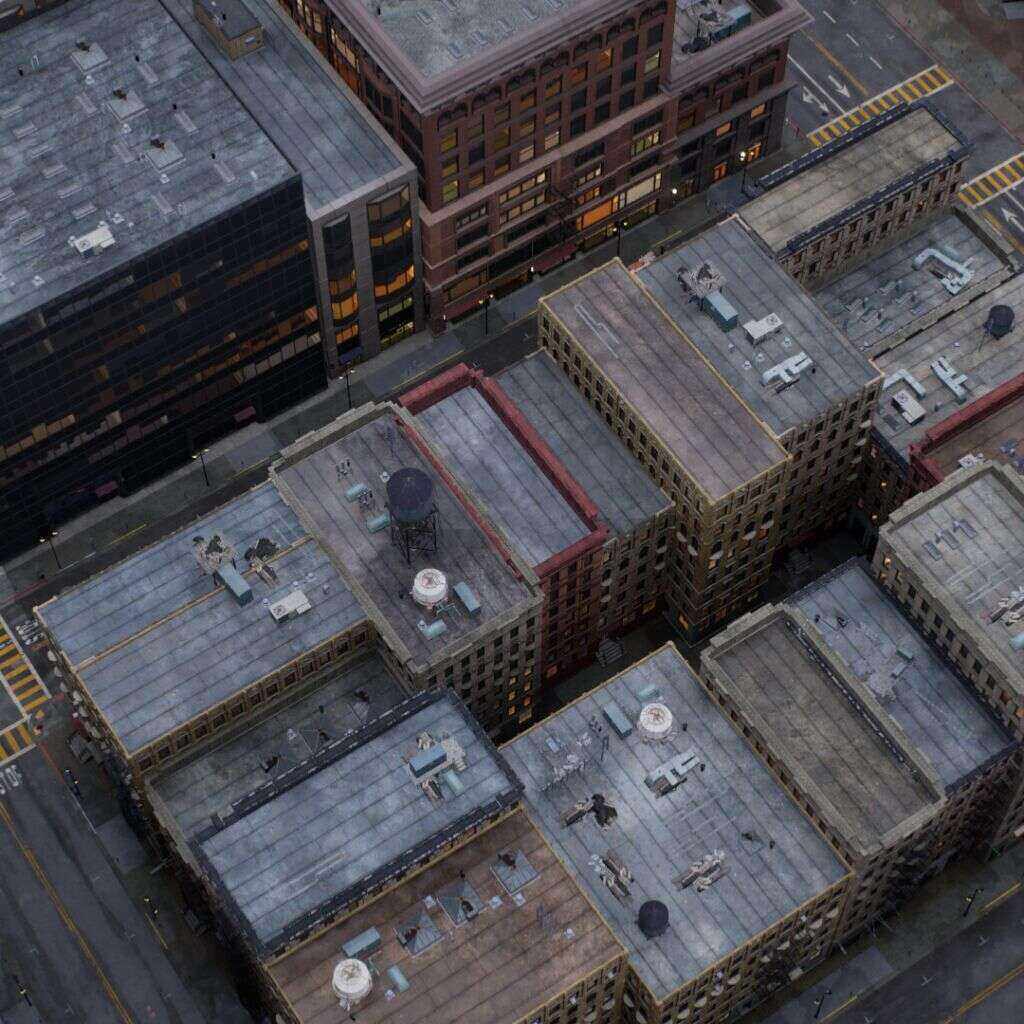} & 
    \imagecell[0.16]{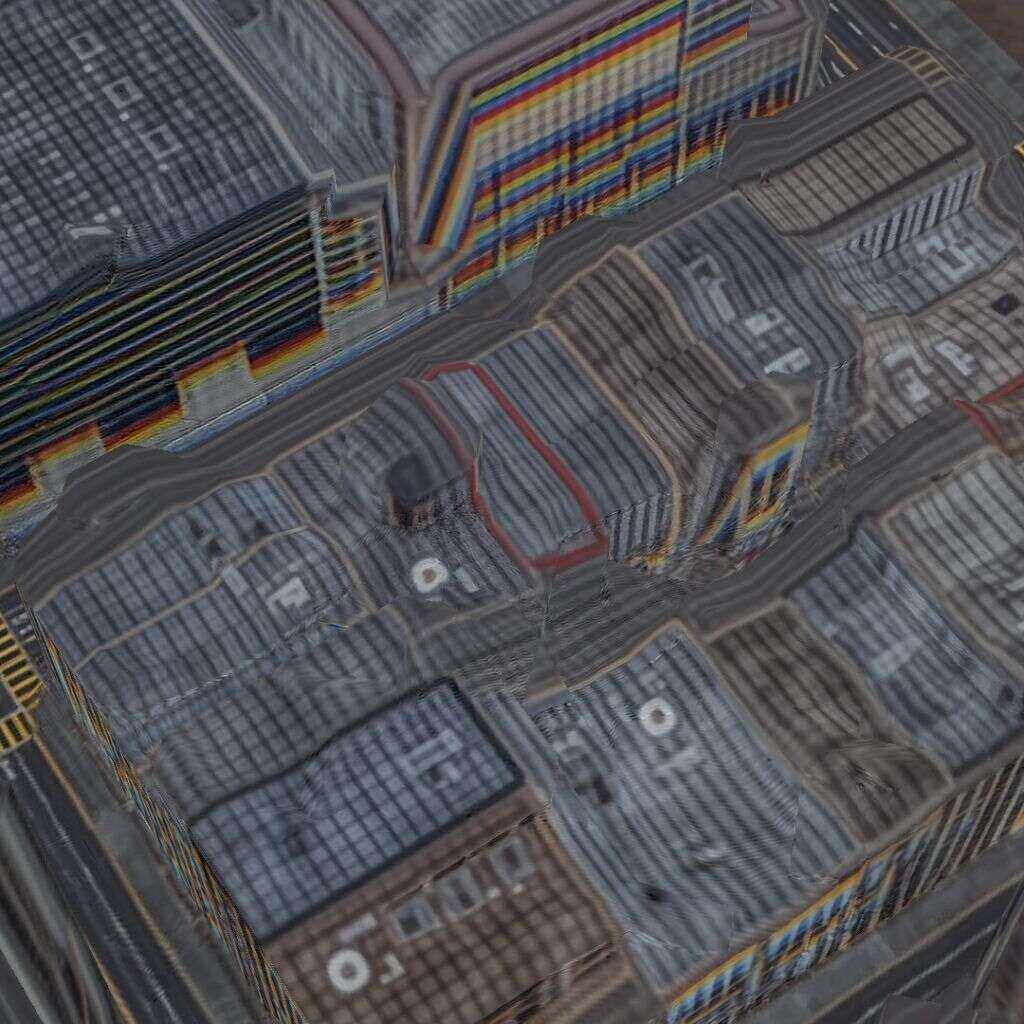} & 
    \imagecell[0.16]{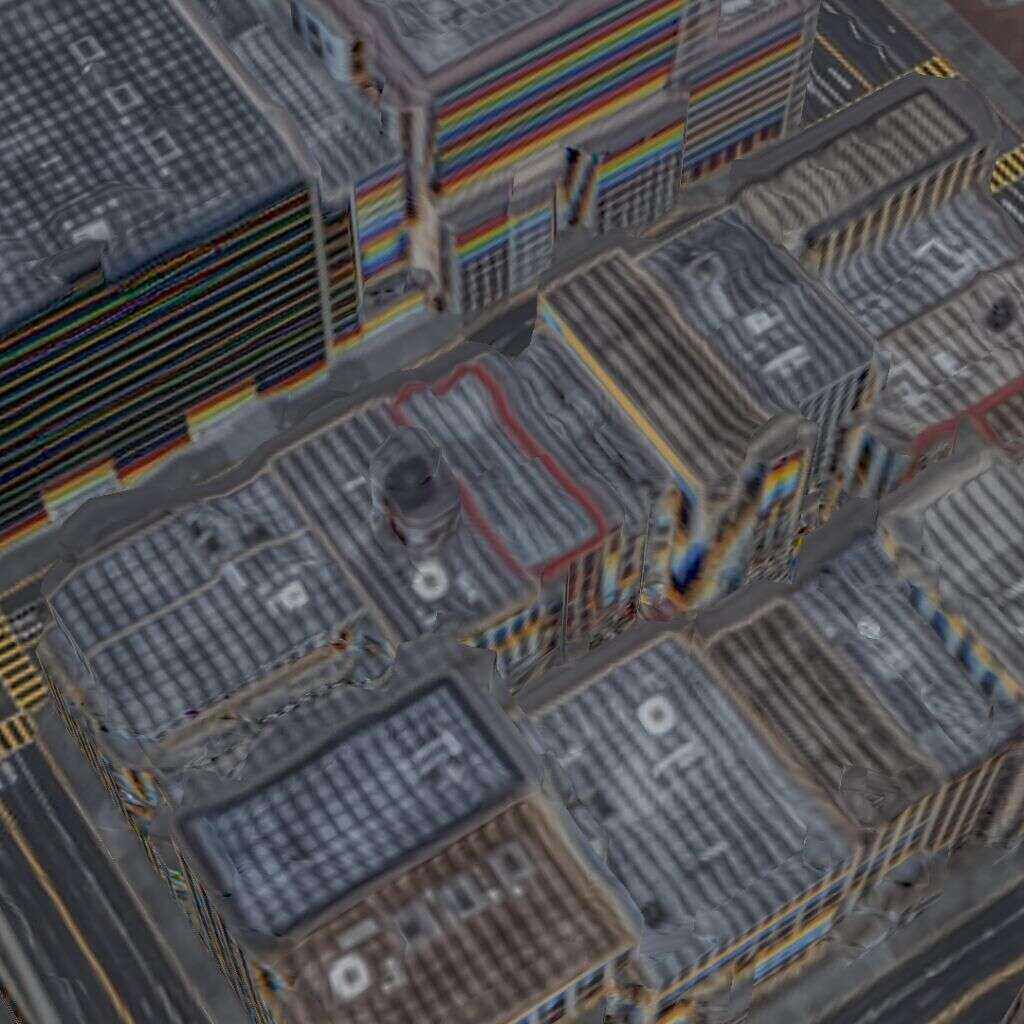} & 
    \imagecell[0.16]{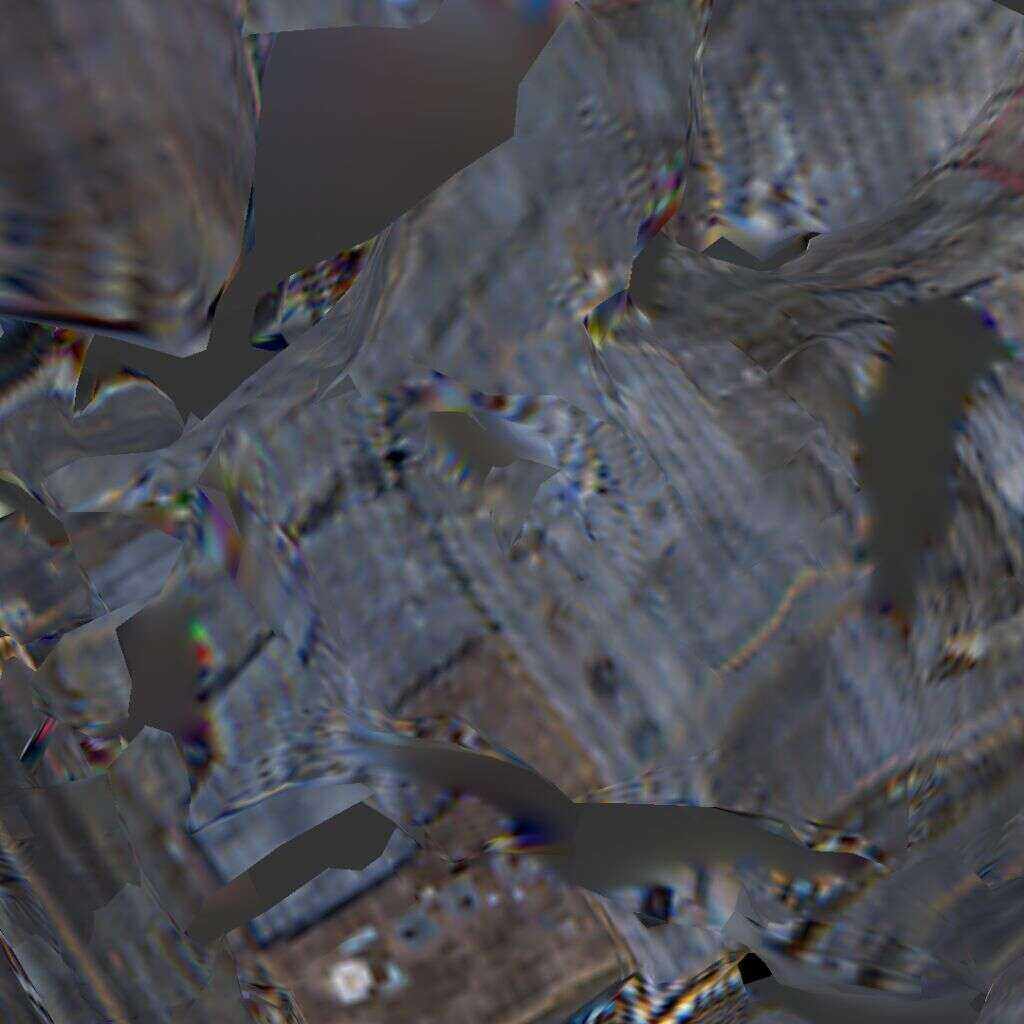} & 
    \imagecell[0.16]{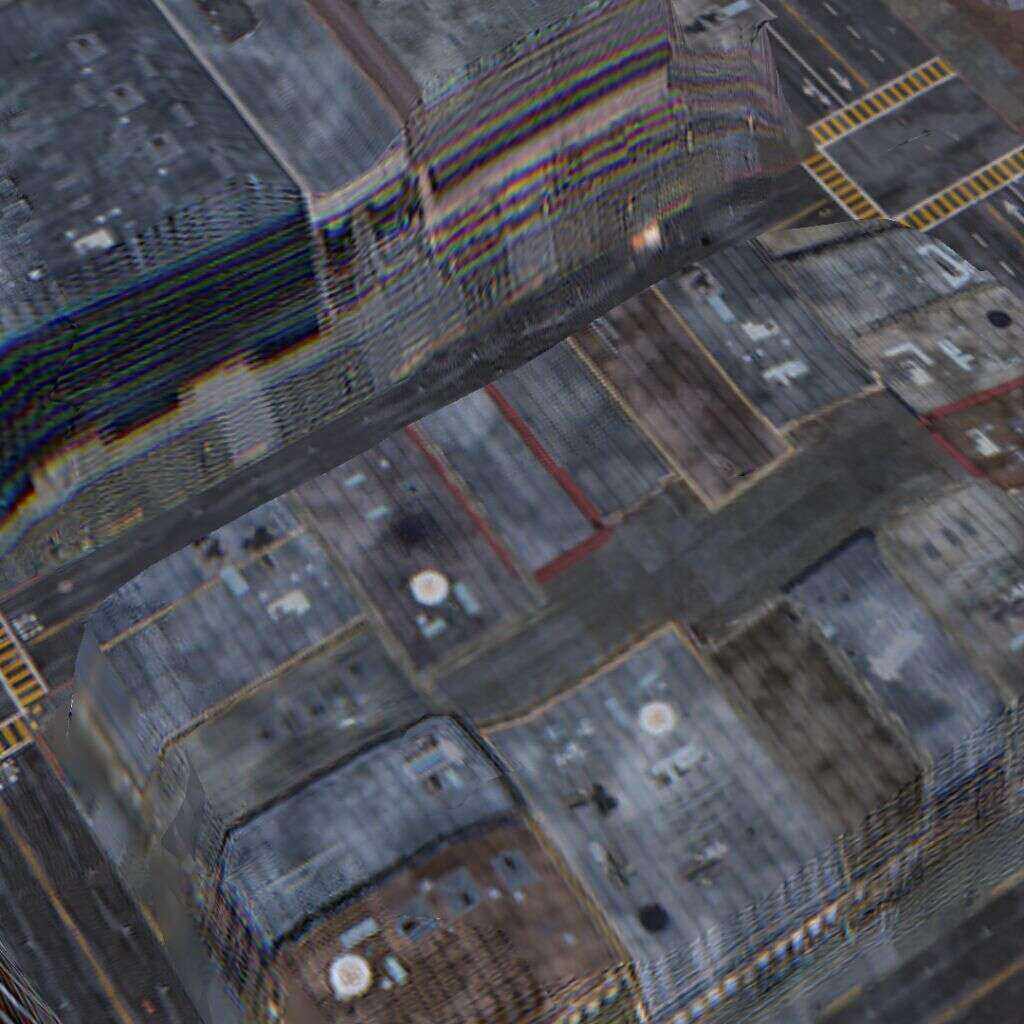} & 
    \imagecell[0.16]{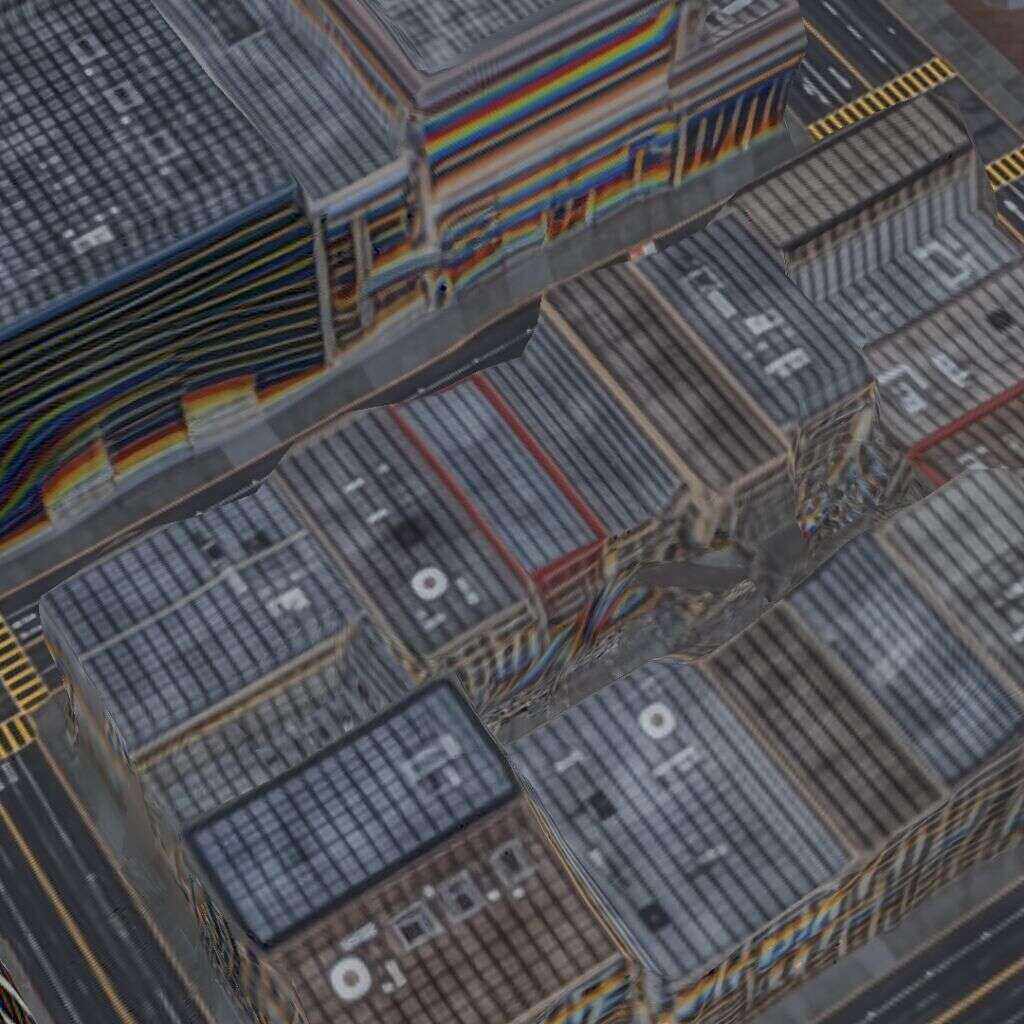} \\
    \vspace*{-10pt} \\
    
    \imagecell[0.16]{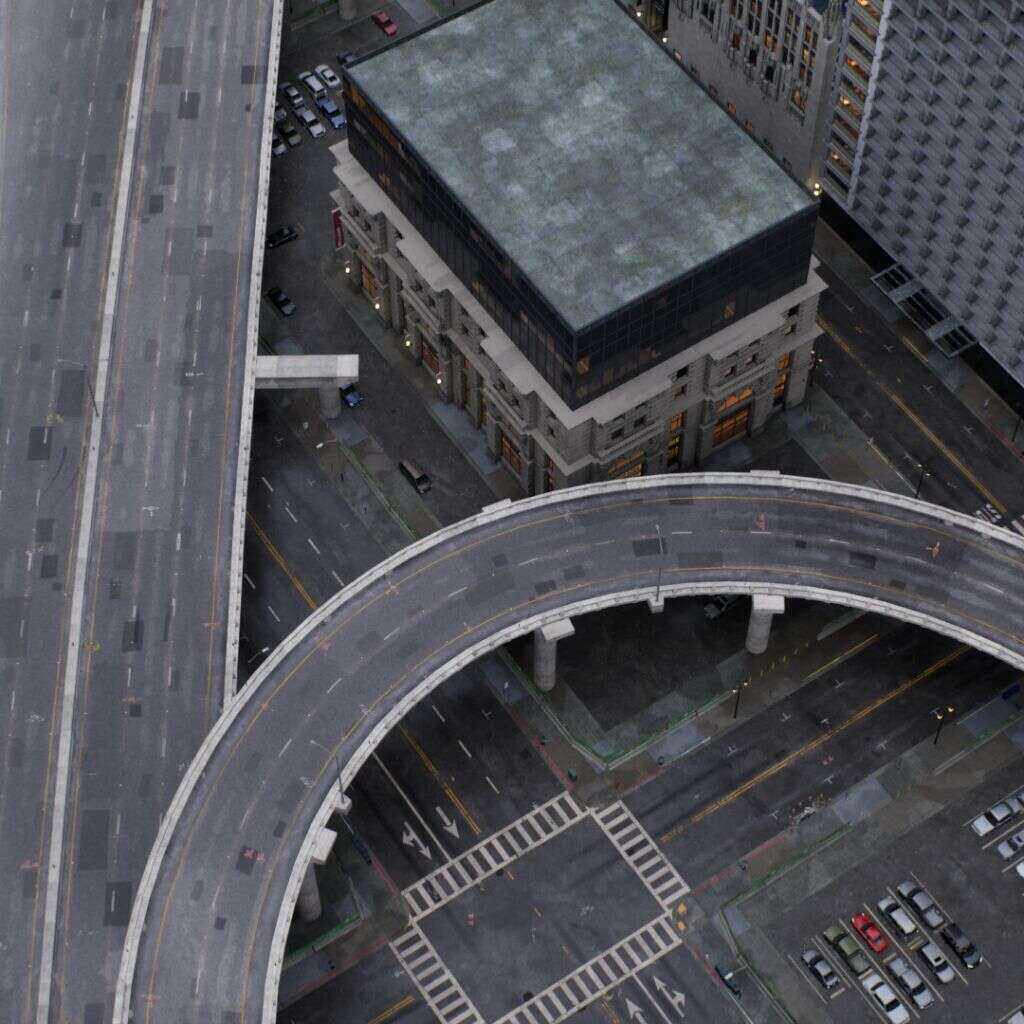} & 
    \imagecell[0.16]{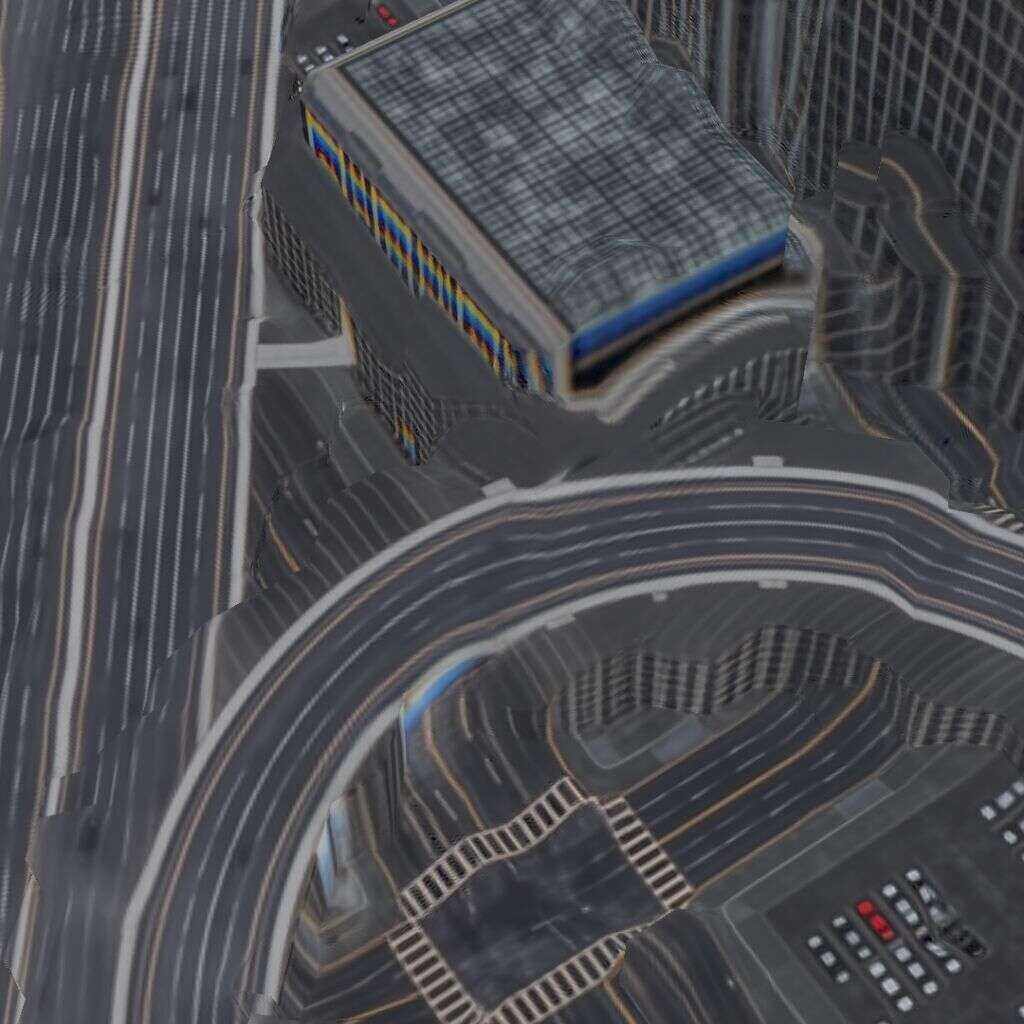} & 
    \imagecell[0.16]{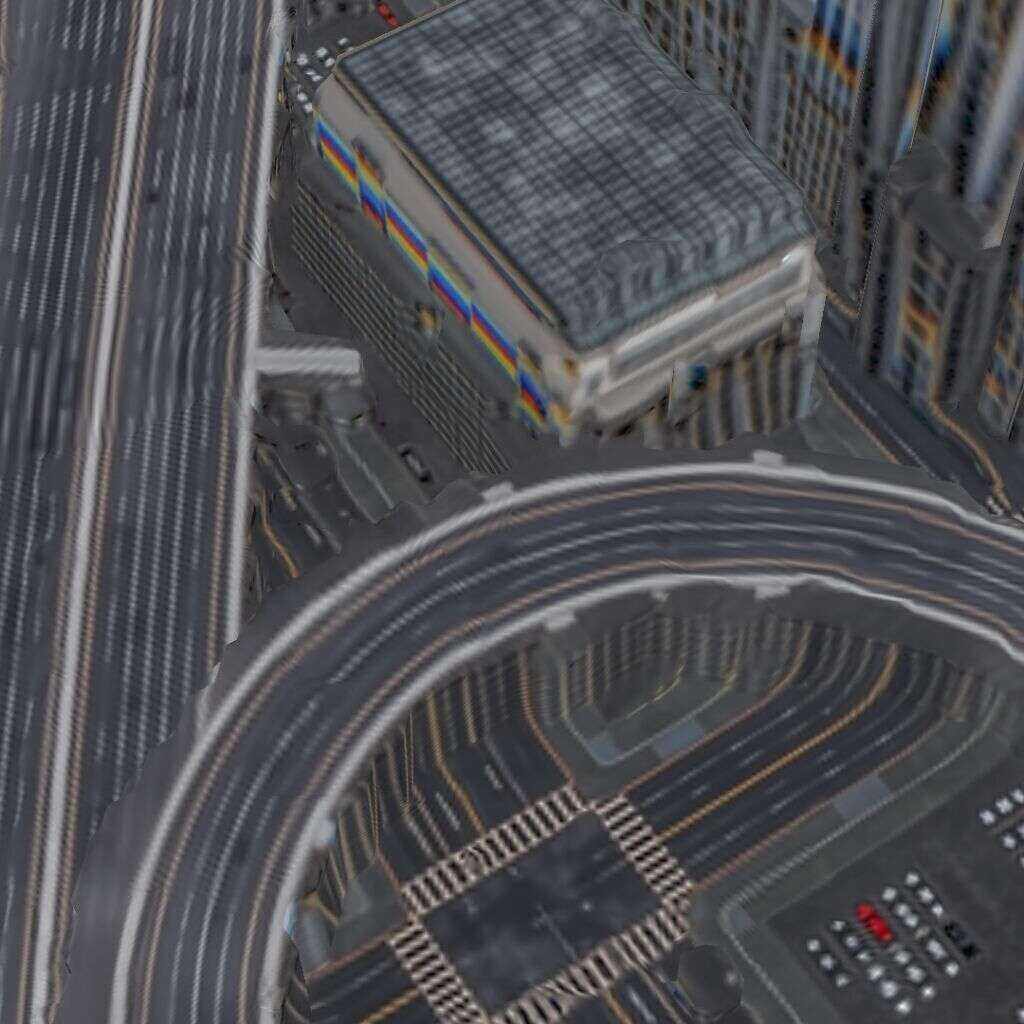} & 
    \imagecell[0.16]{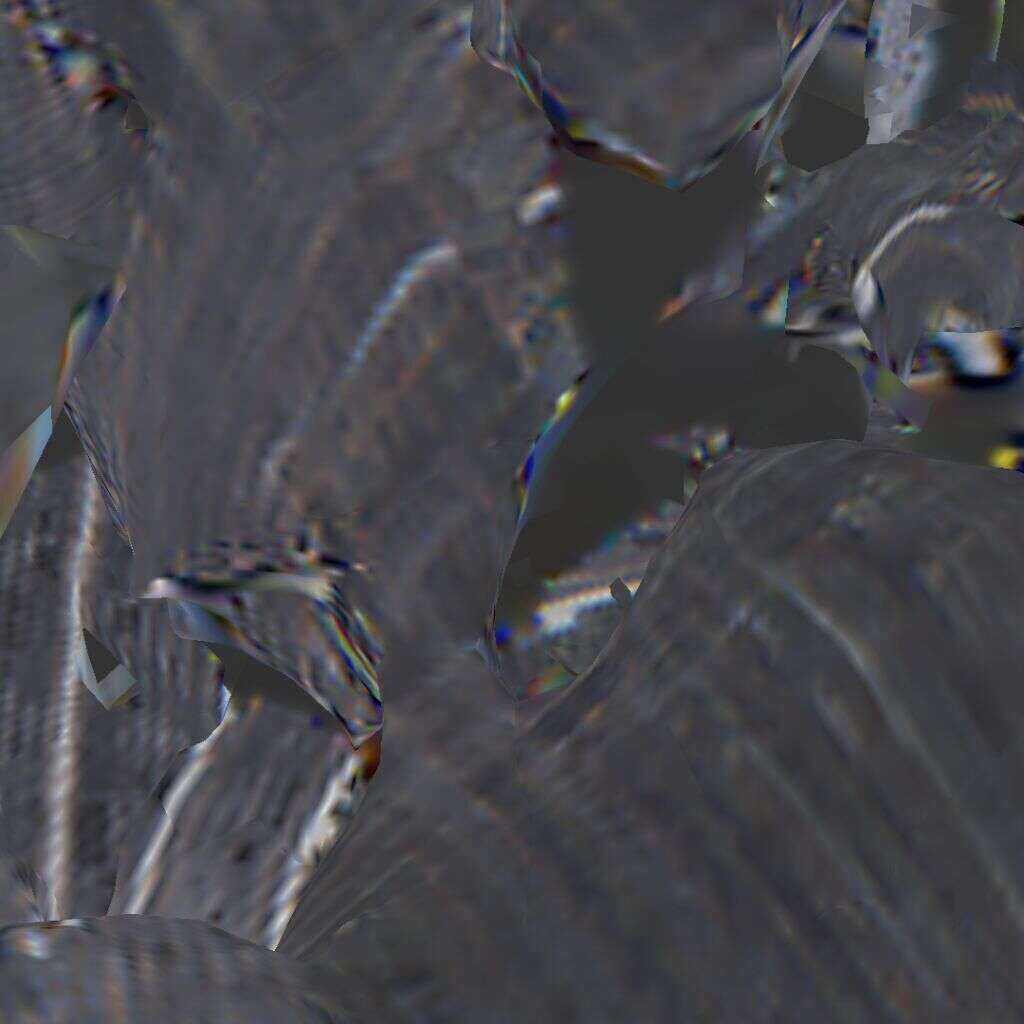} & 
    \imagecell[0.16]{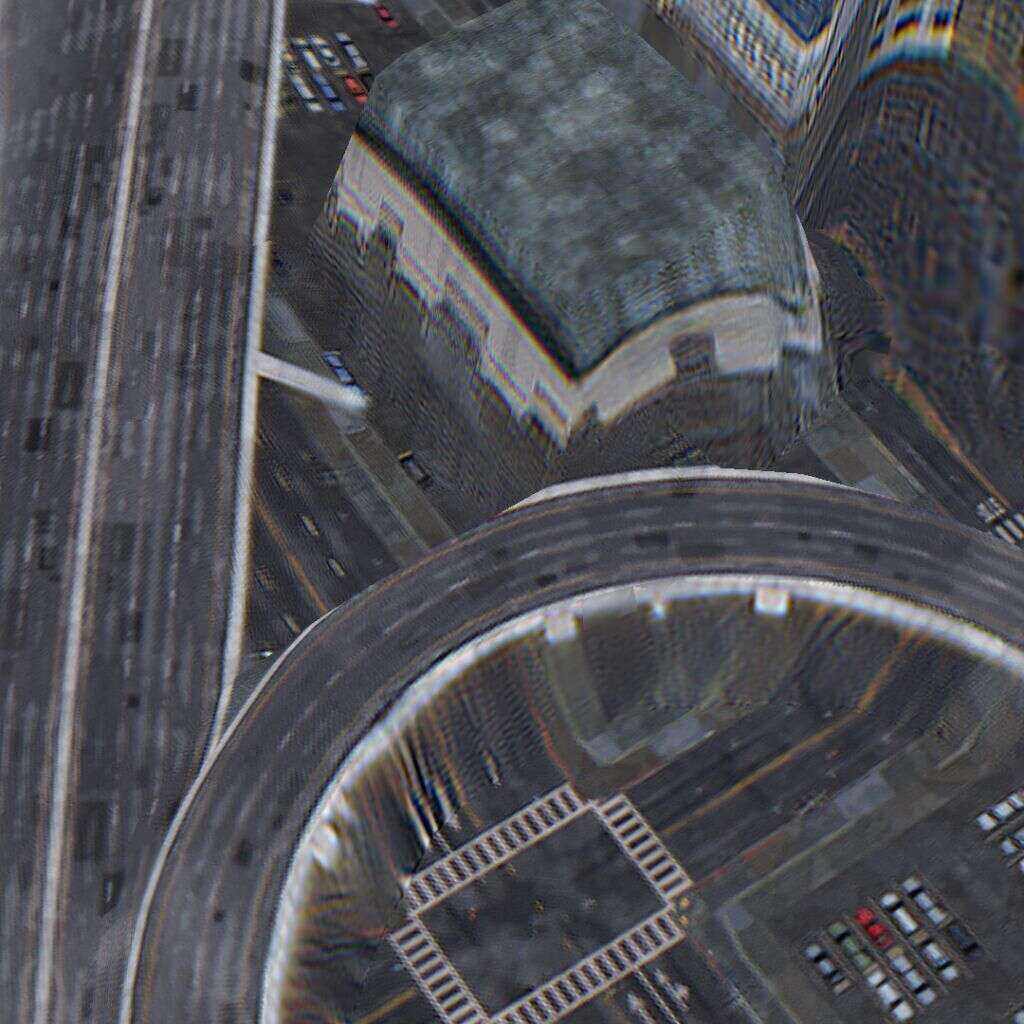} & 
    \imagecell[0.16]{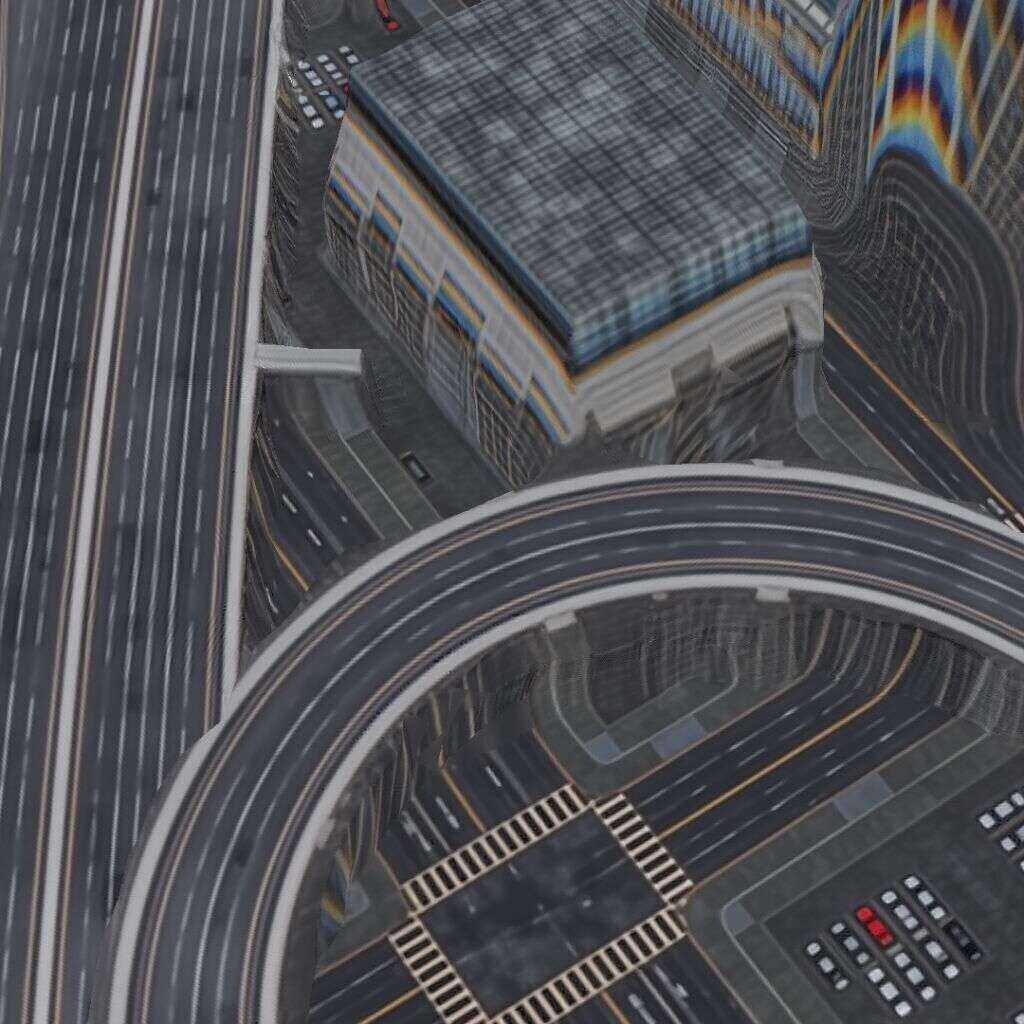} \\
    \vspace*{-10pt} \\
    
    \imagecell[0.16]{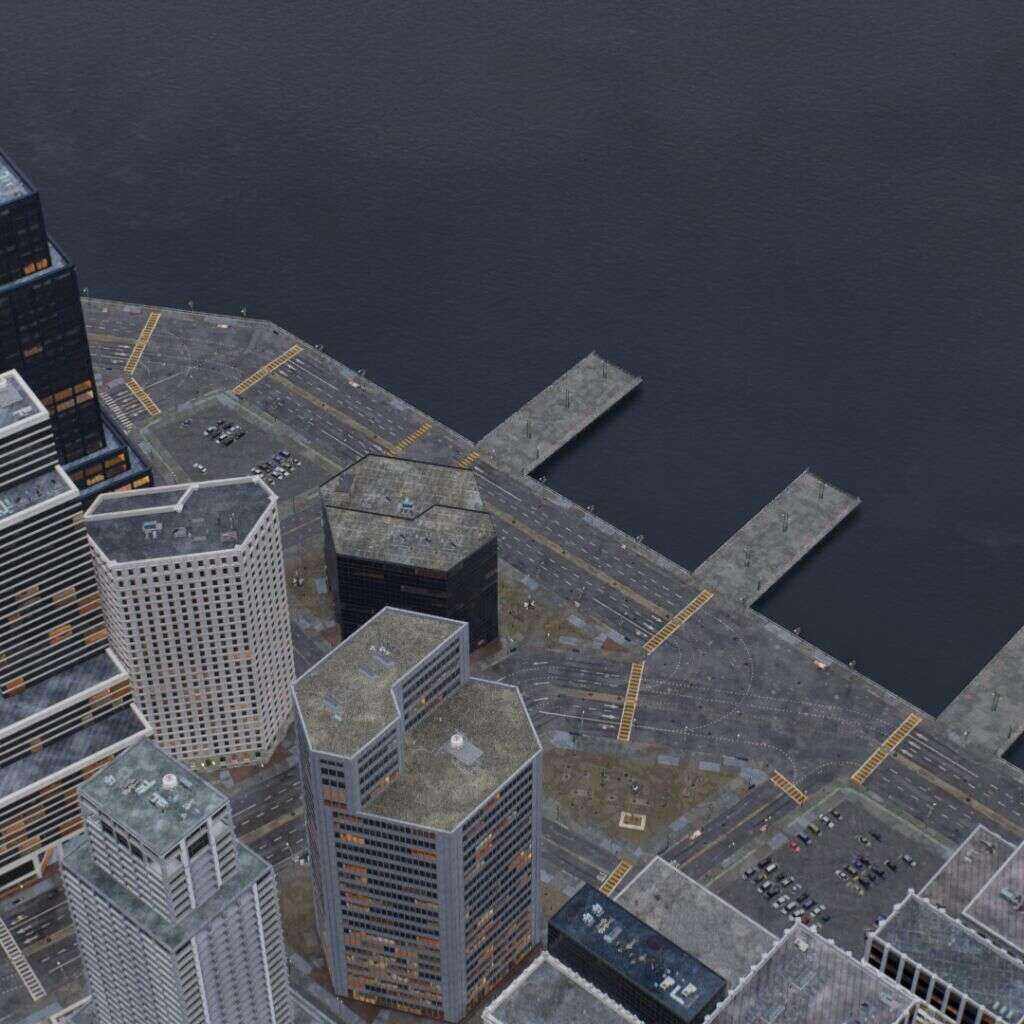} & 
    \imagecell[0.16]{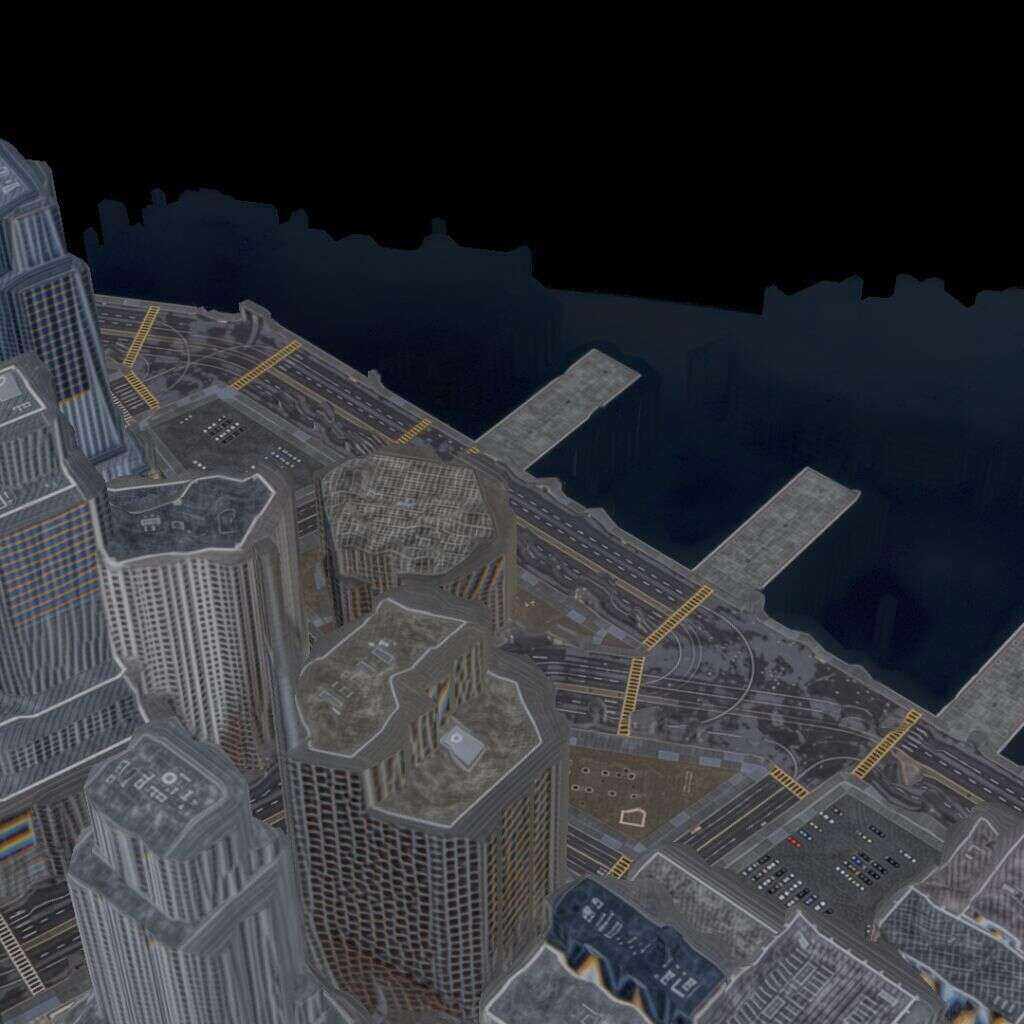} & 
    \imagecell[0.16]{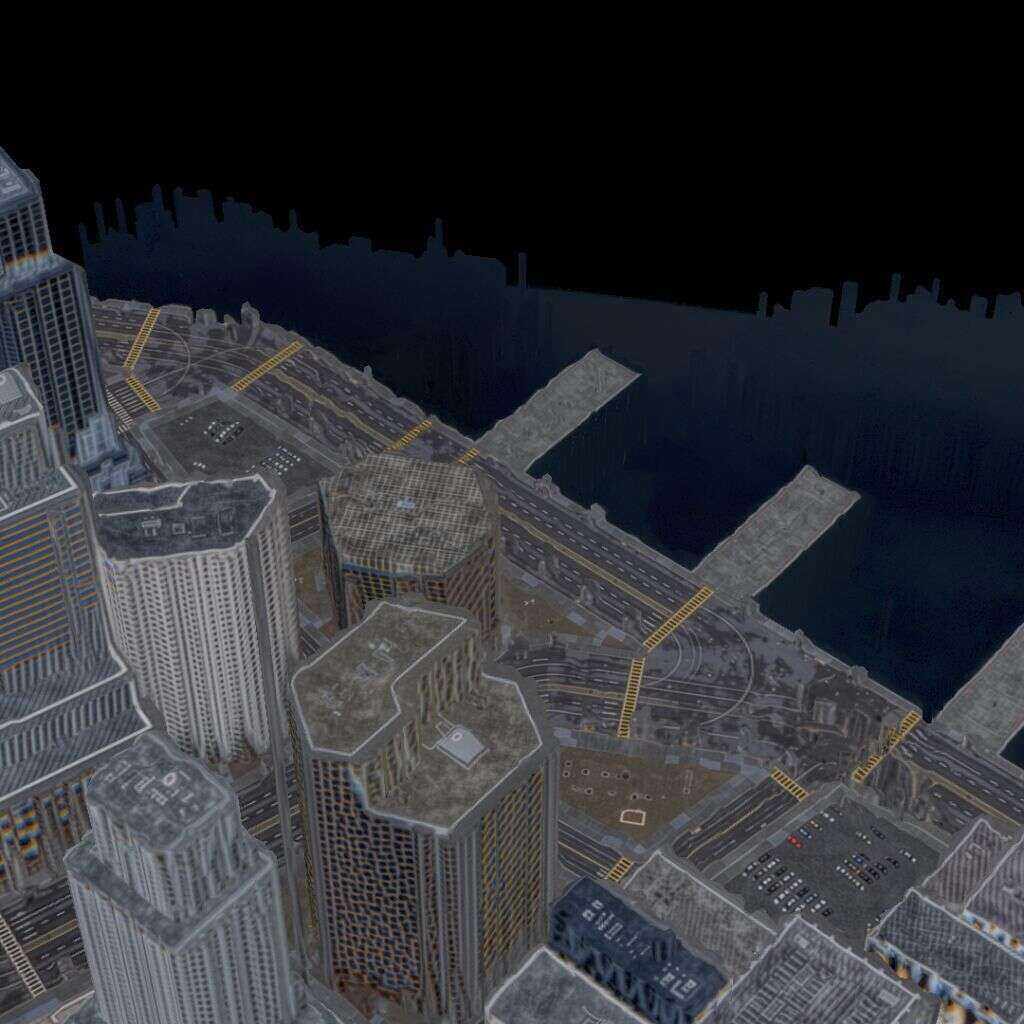} & 
    \imagecell[0.16]{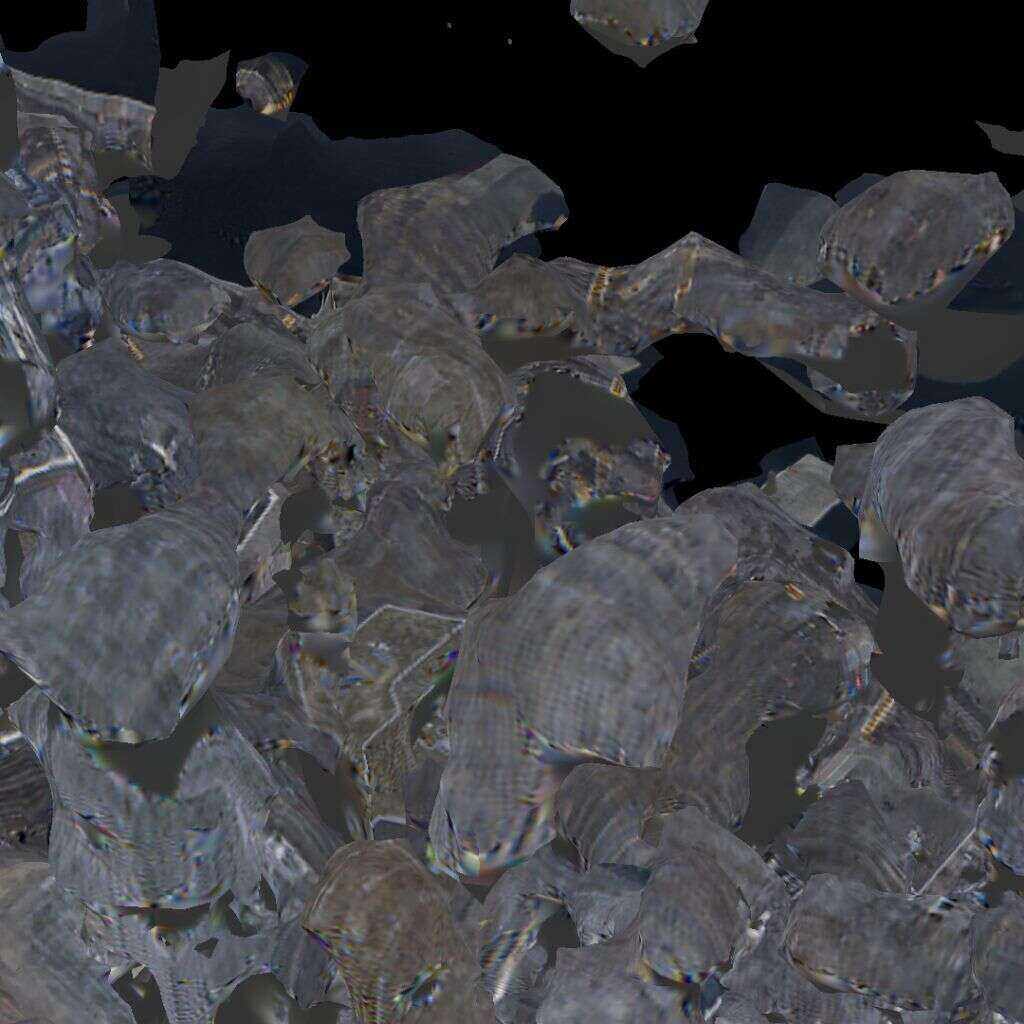} & 
    \imagecell[0.16]{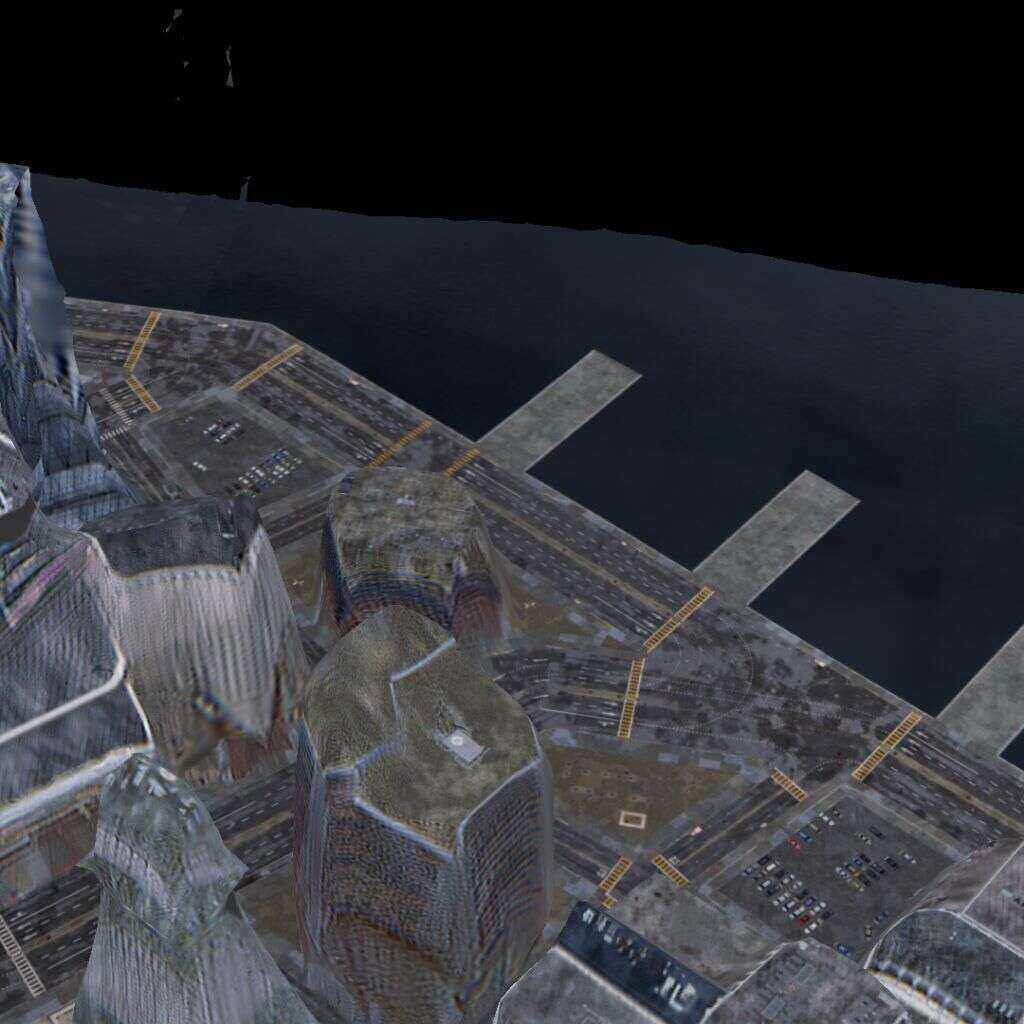} & 
    \imagecell[0.16]{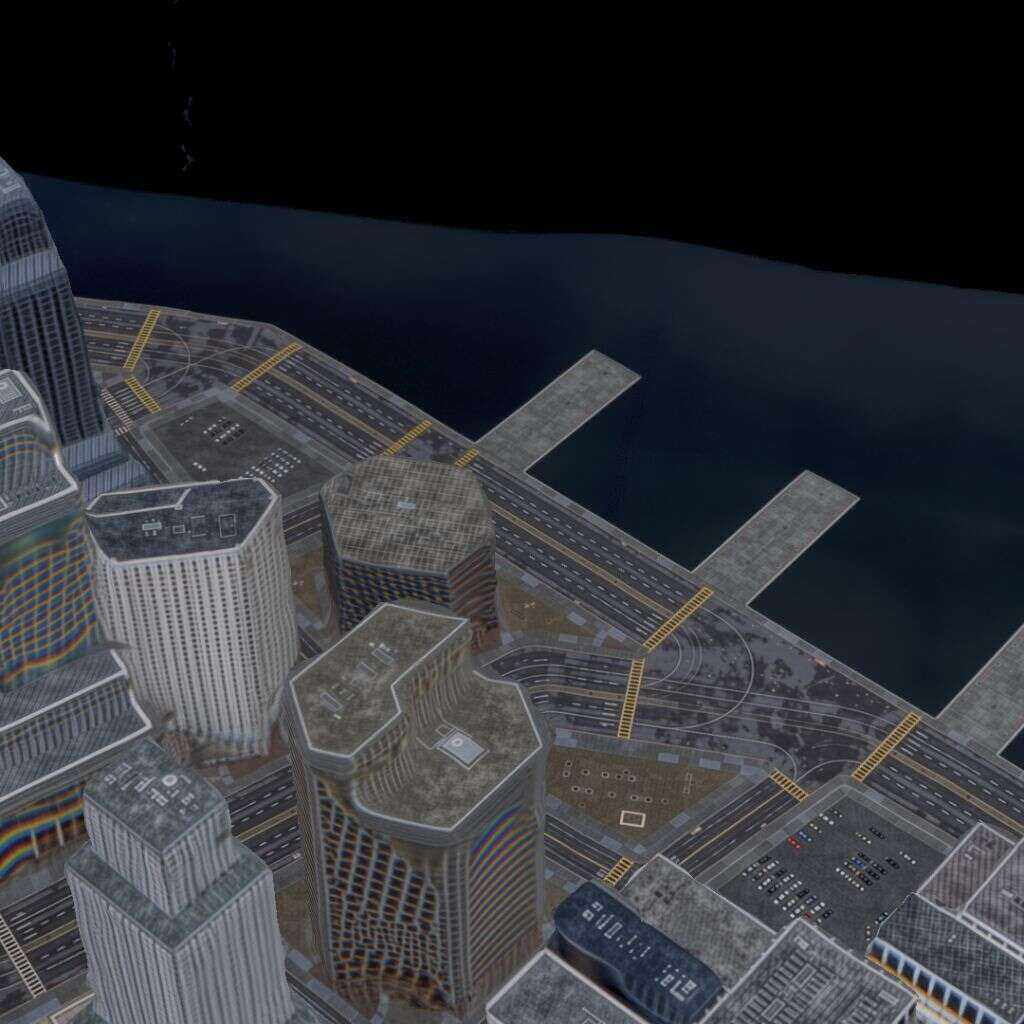} \\
    \vspace*{-10pt} \\
    
    \imagecell[0.16]{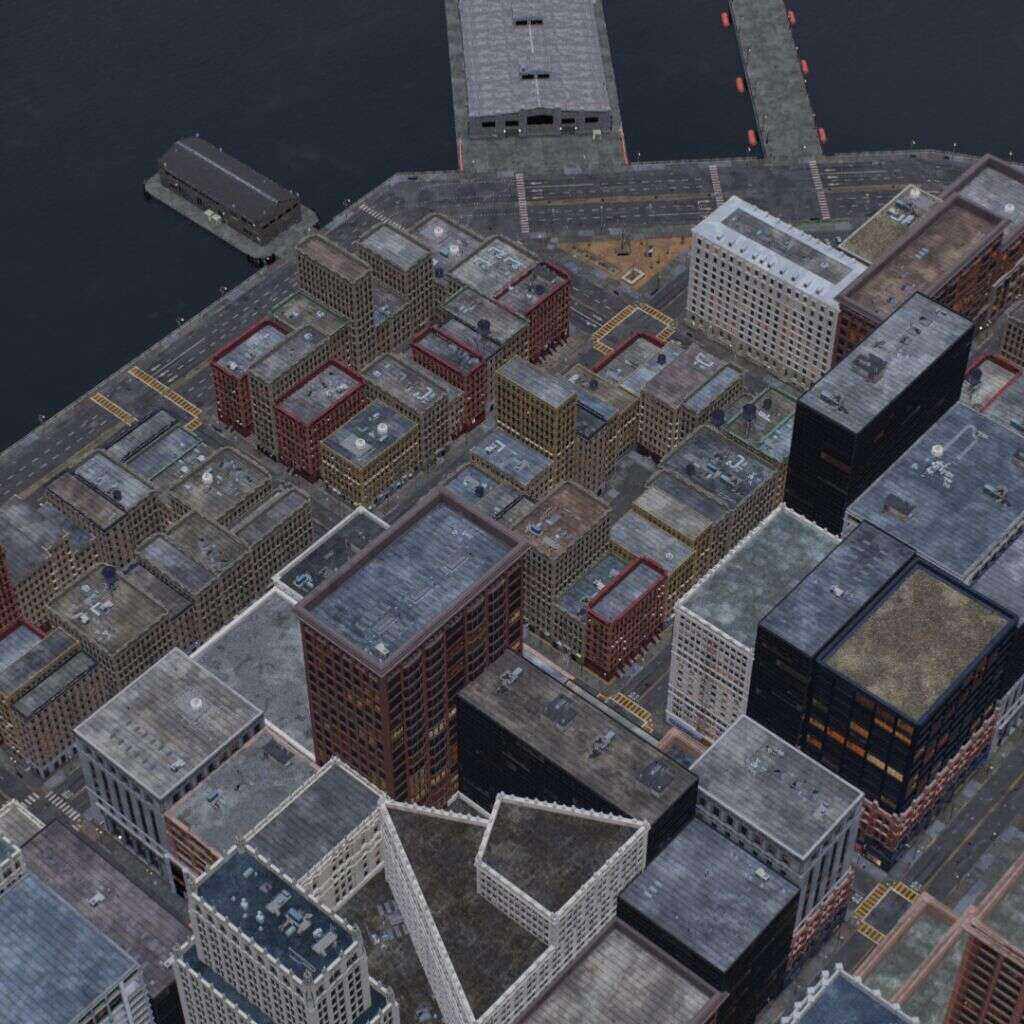} & 
    \imagecell[0.16]{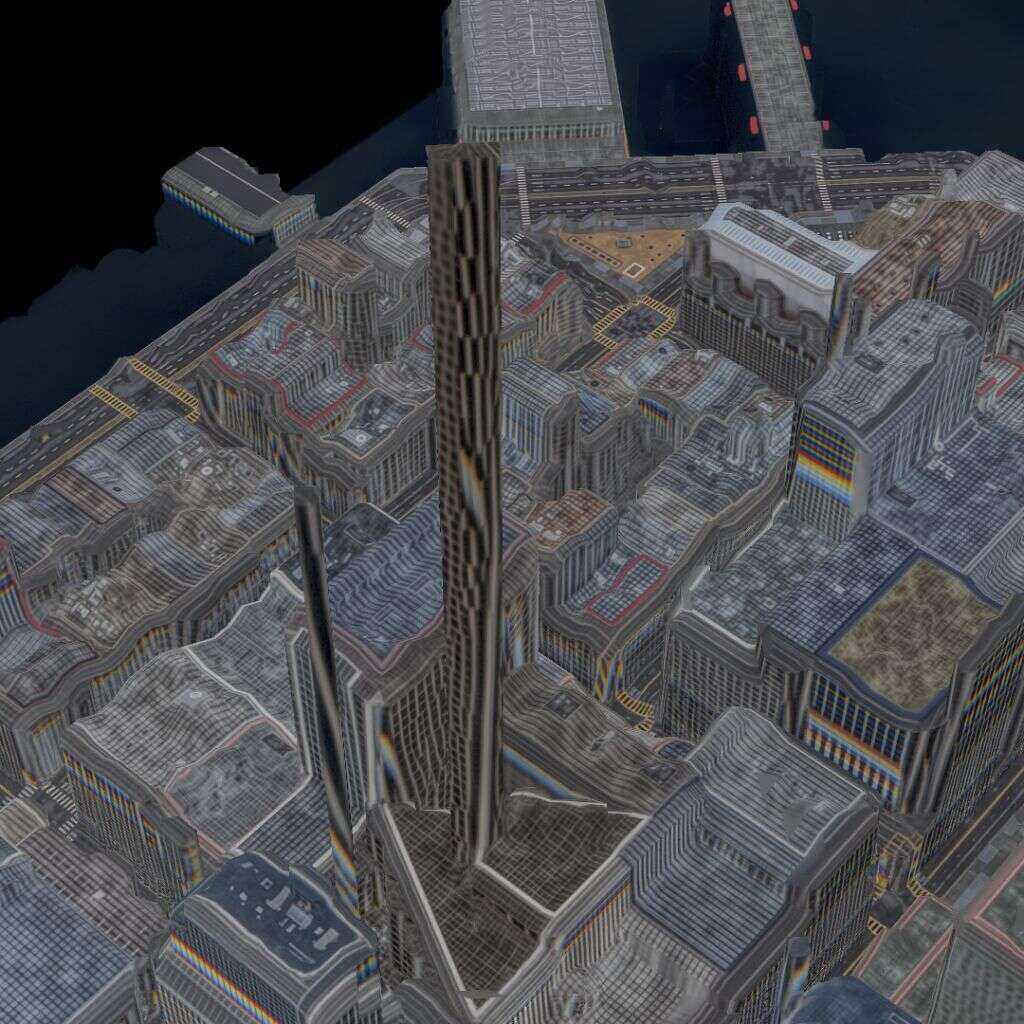} & 
    \imagecell[0.16]{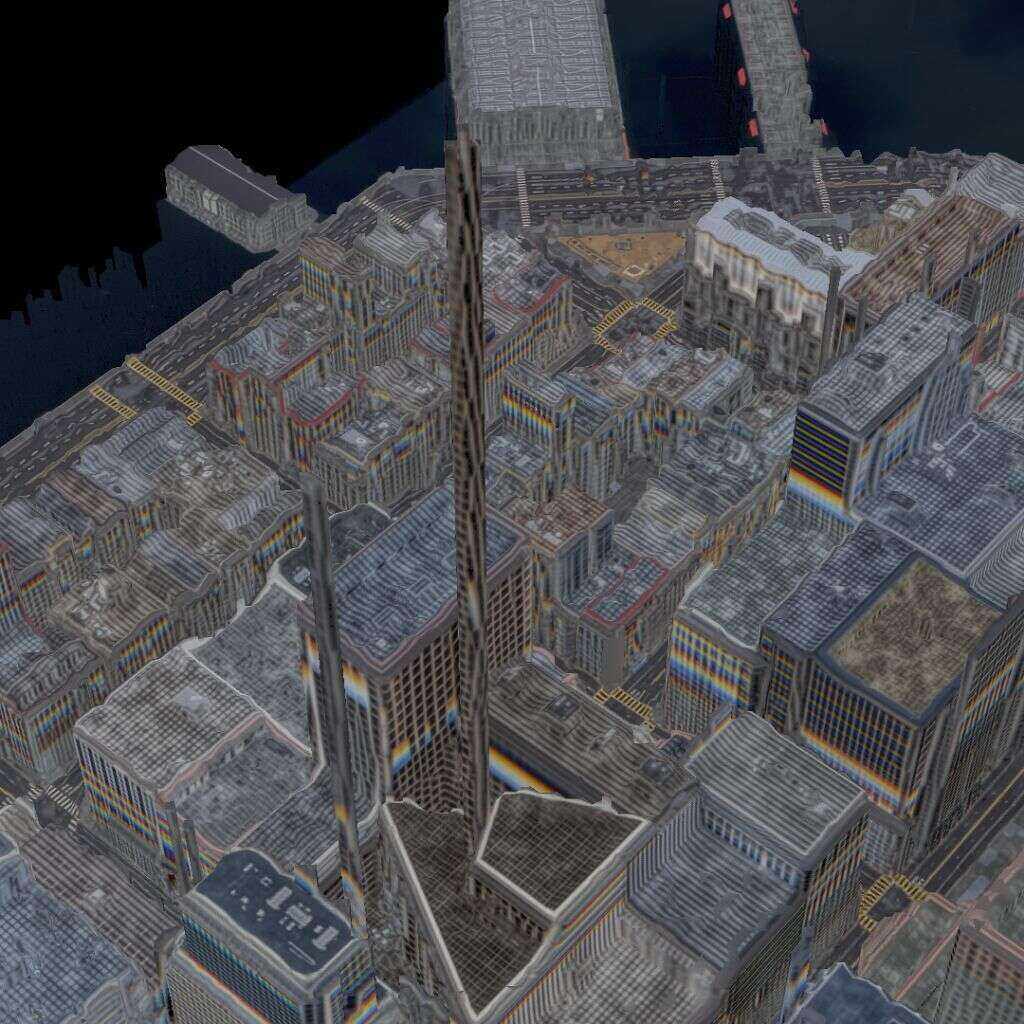} & 
    \imagecell[0.16]{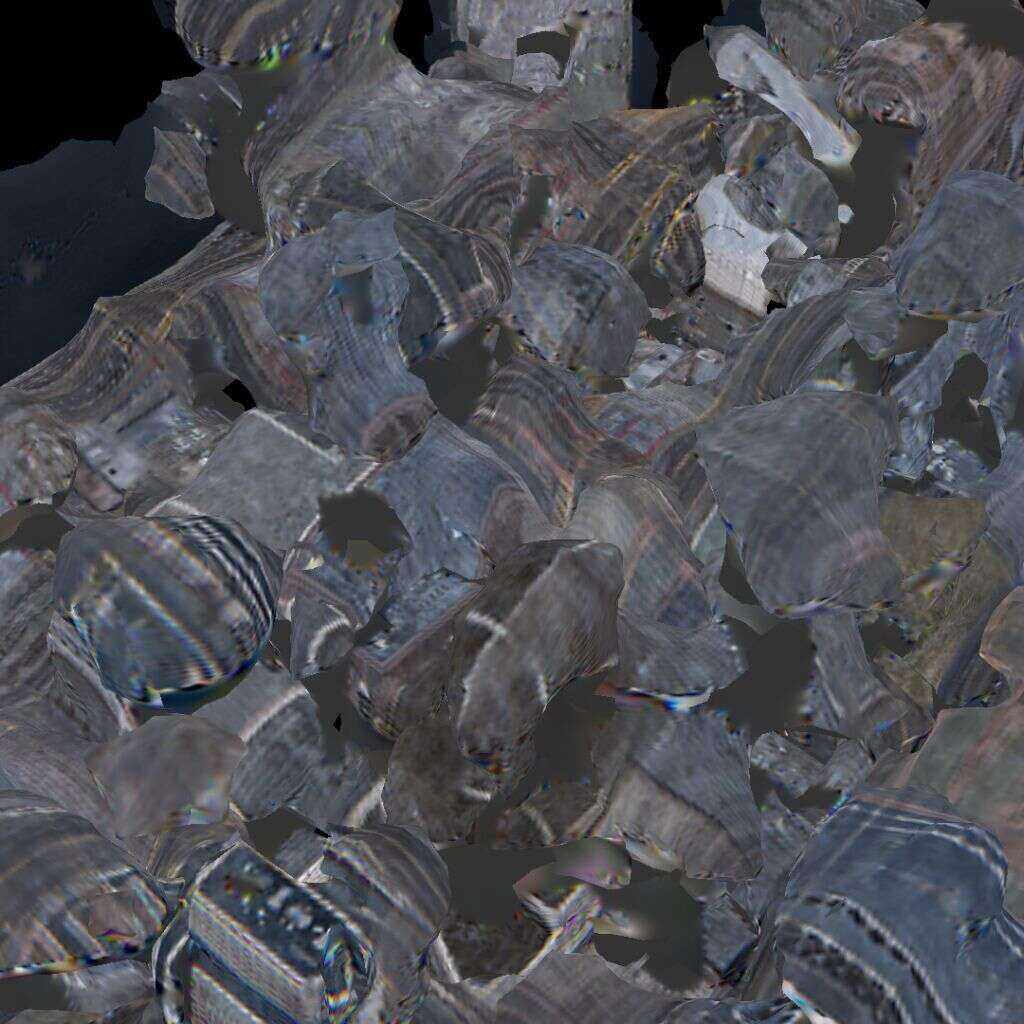} & 
    \imagecell[0.16]{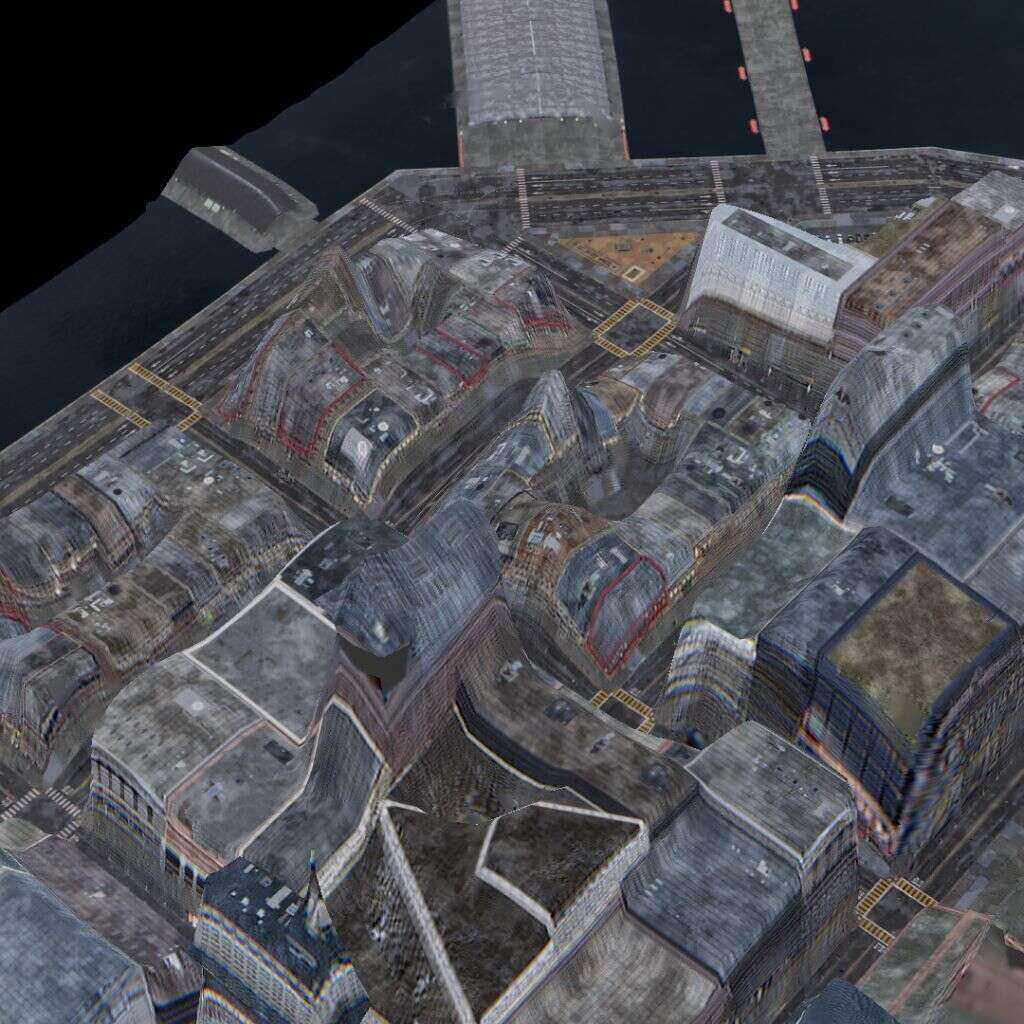} & 
    \imagecell[0.16]{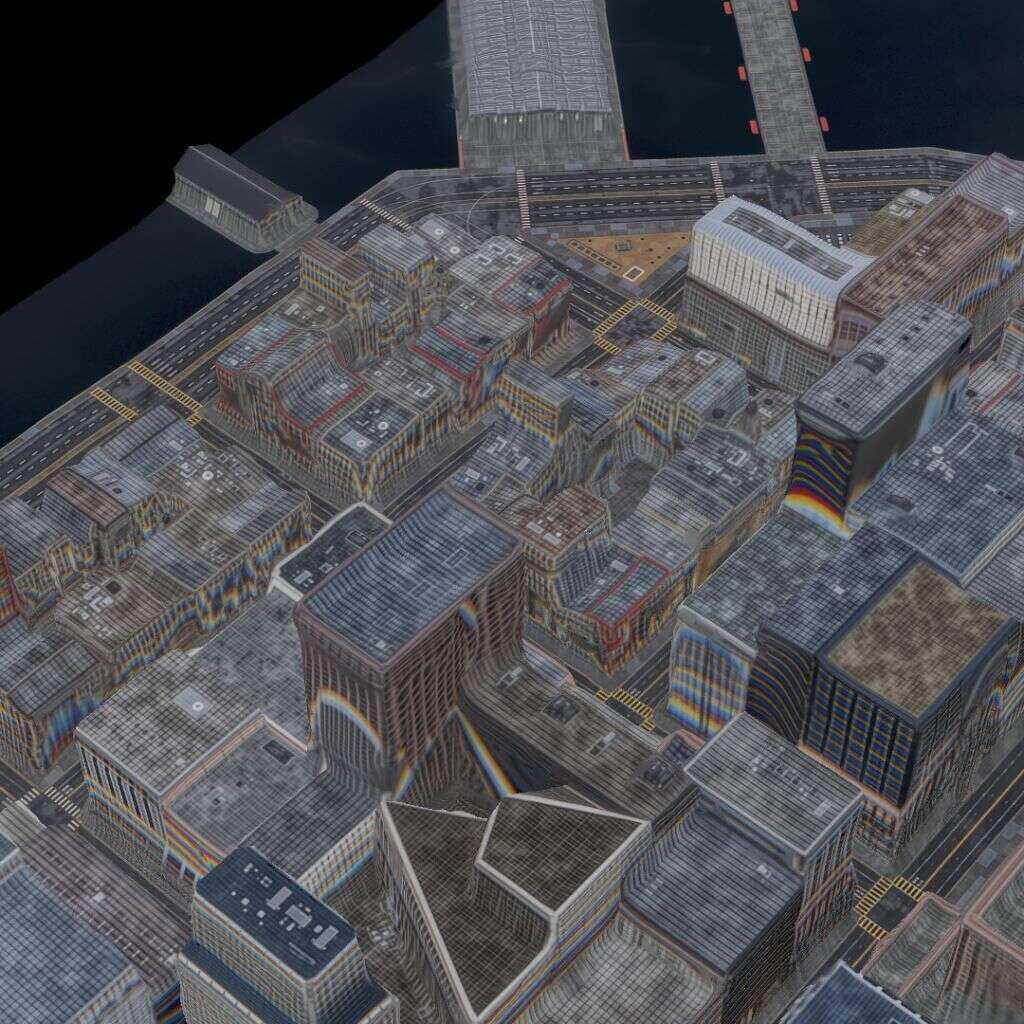} \\
    \vspace*{-10pt} \\
    
    \imagecell[0.16]{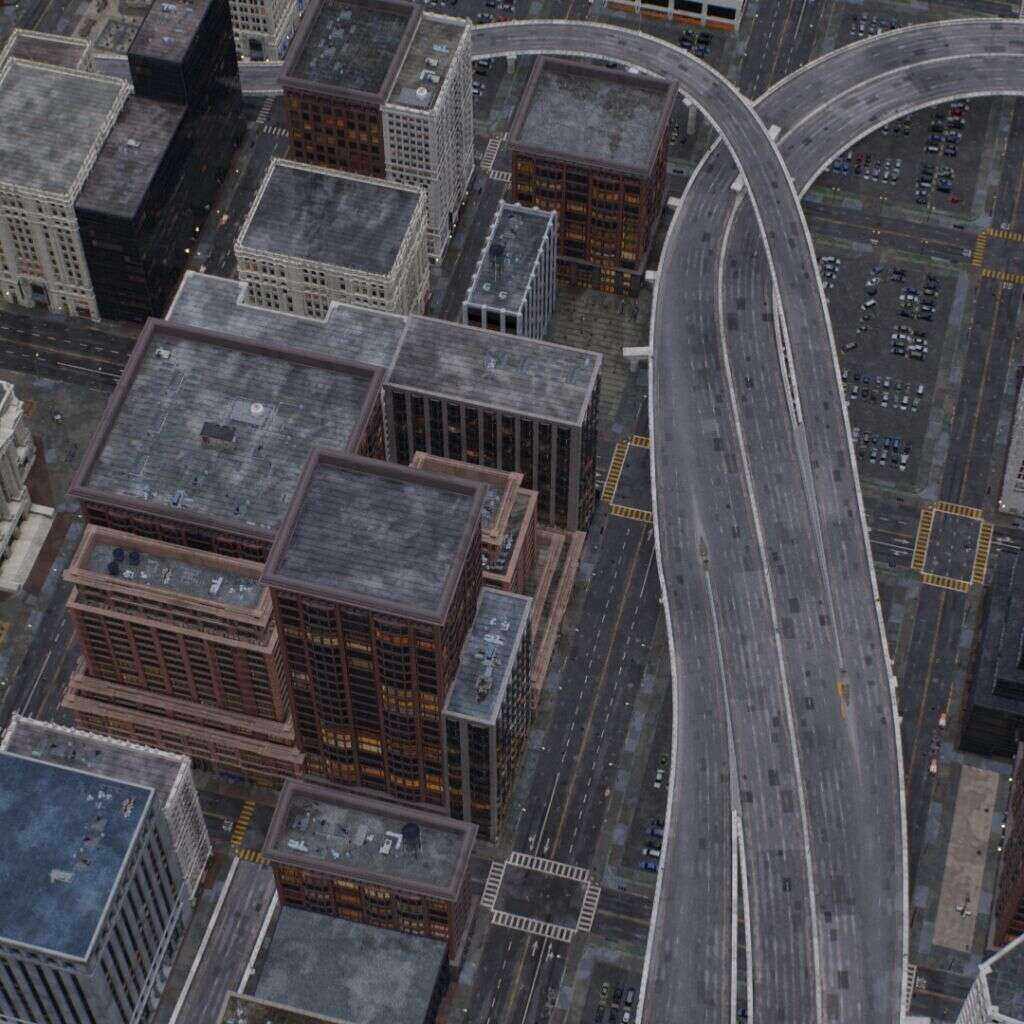} & 
    \imagecell[0.16]{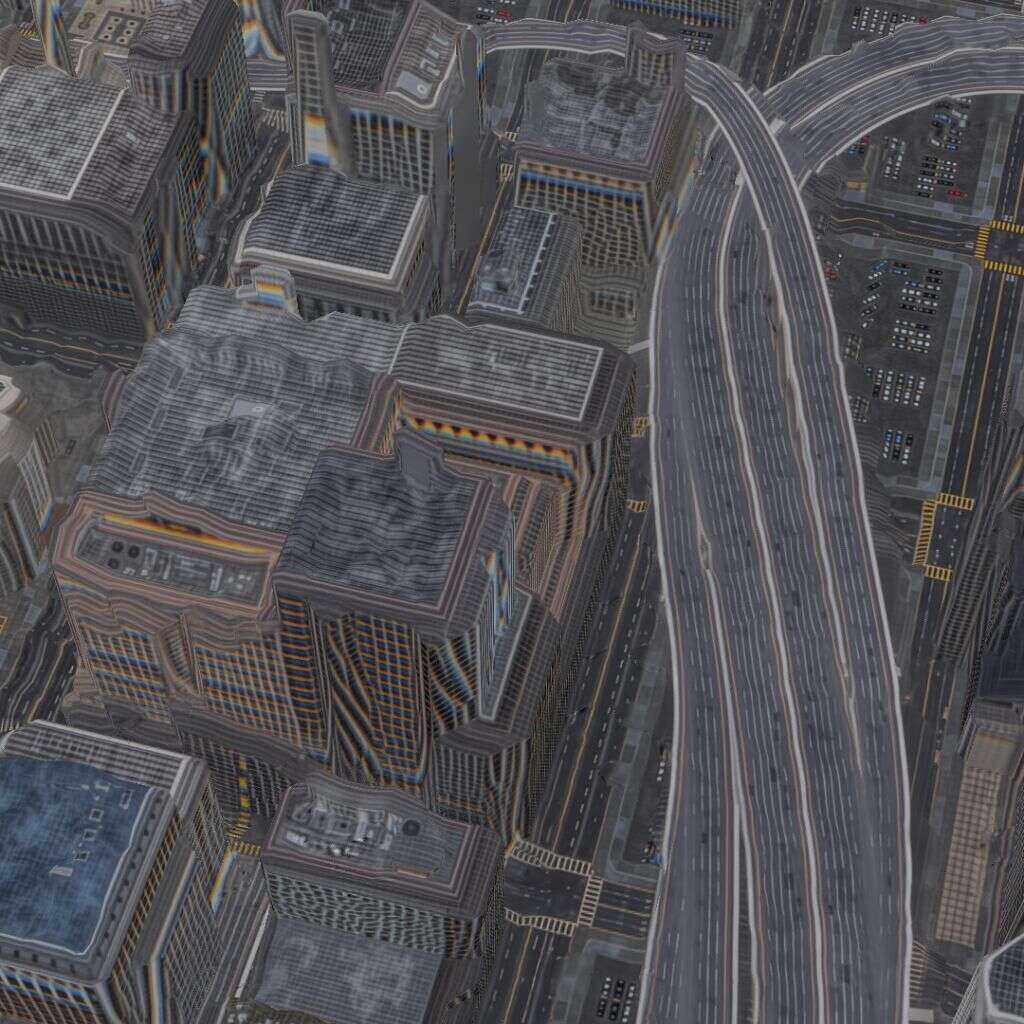} & 
    \imagecell[0.16]{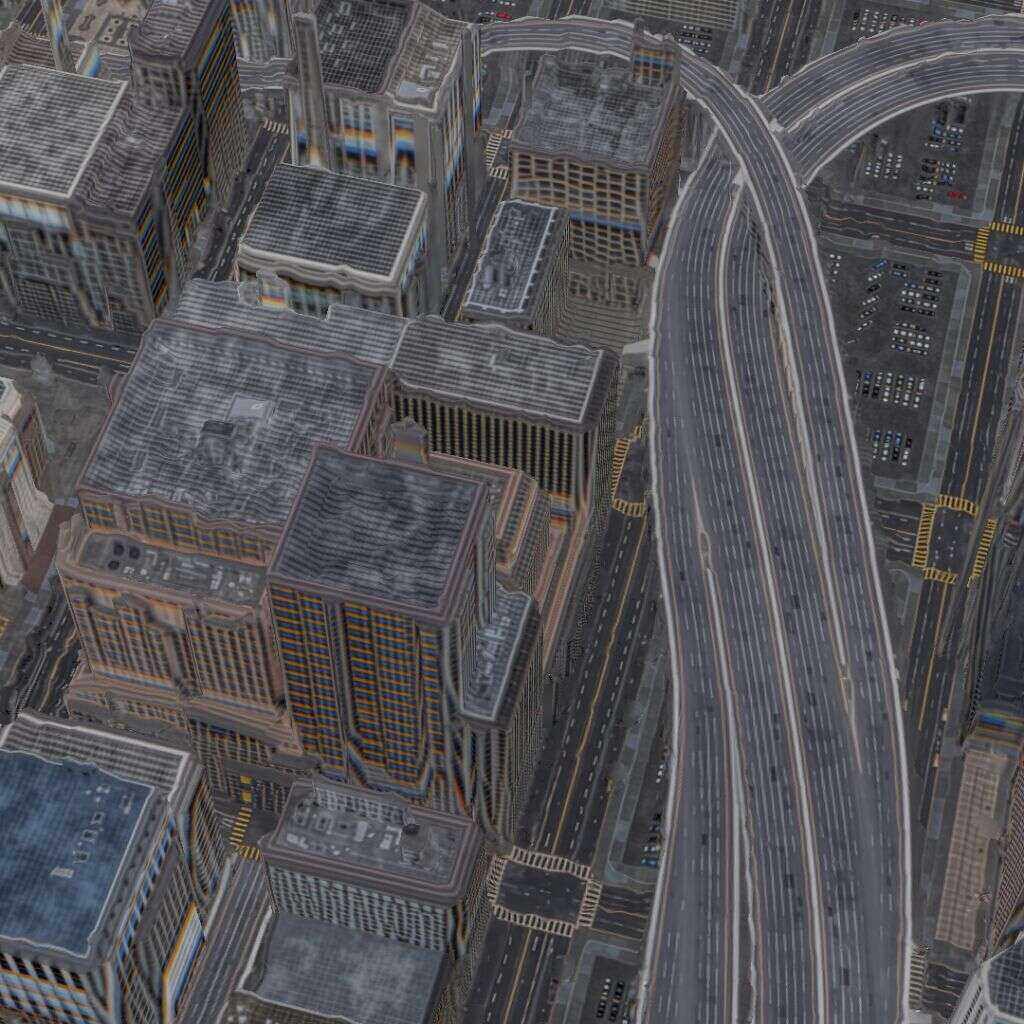} & 
    \imagecell[0.16]{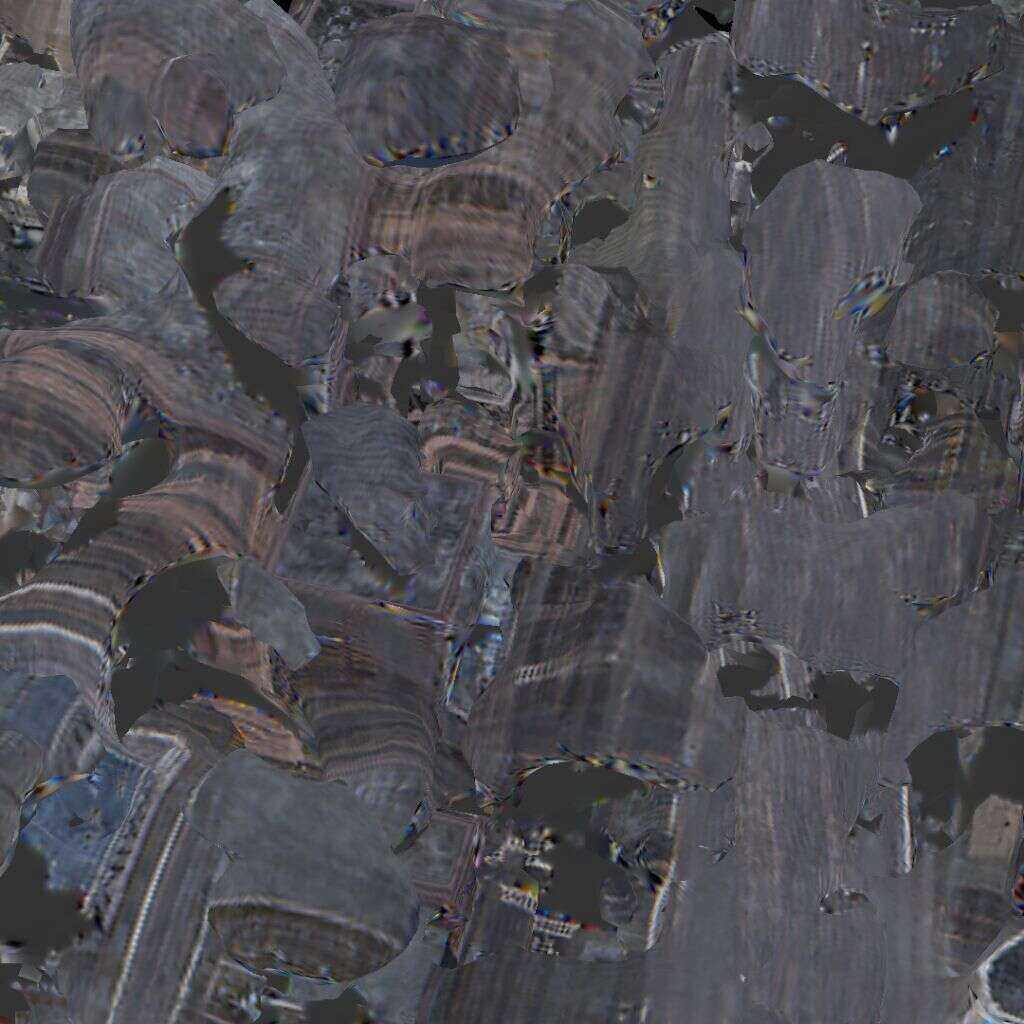} & 
    \imagecell[0.16]{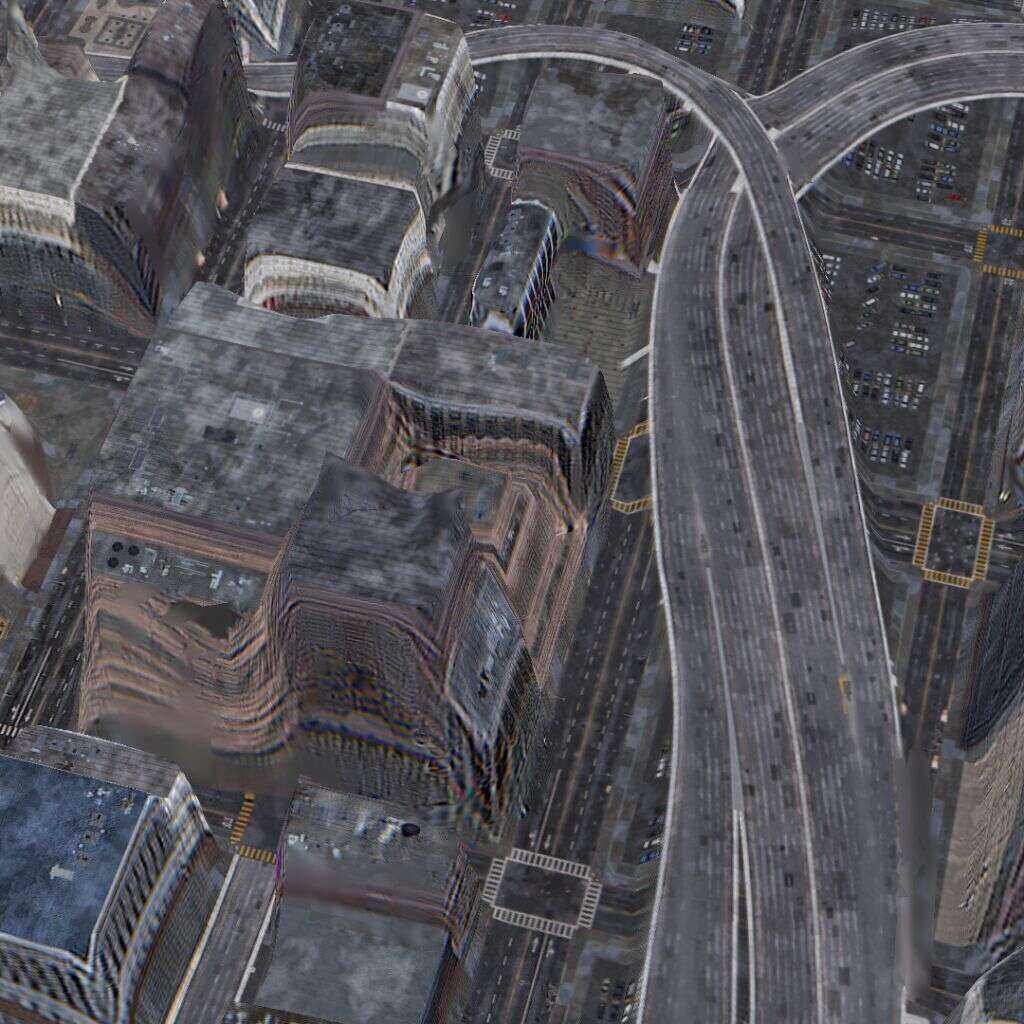} & 
    \imagecell[0.16]{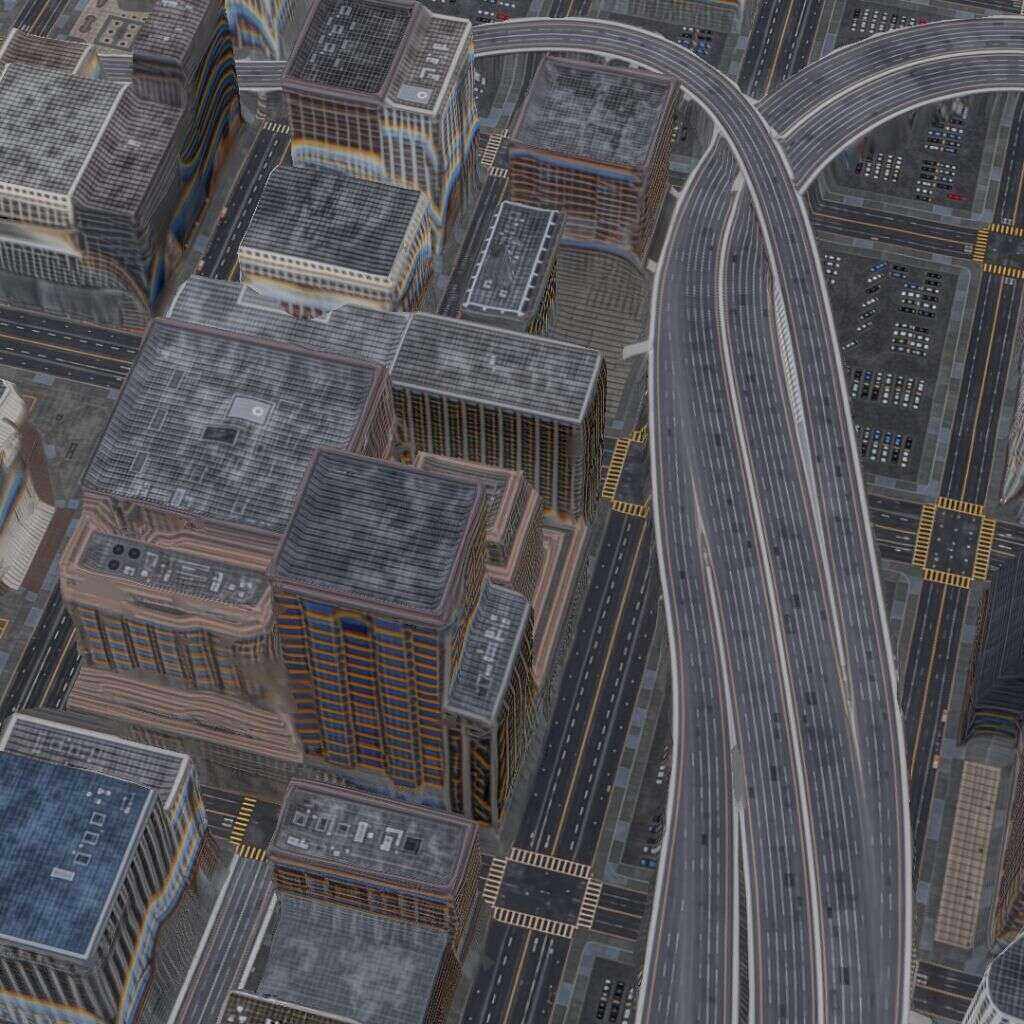} \\
    \vspace*{-10pt} \\
    
    \imagecell[0.16]{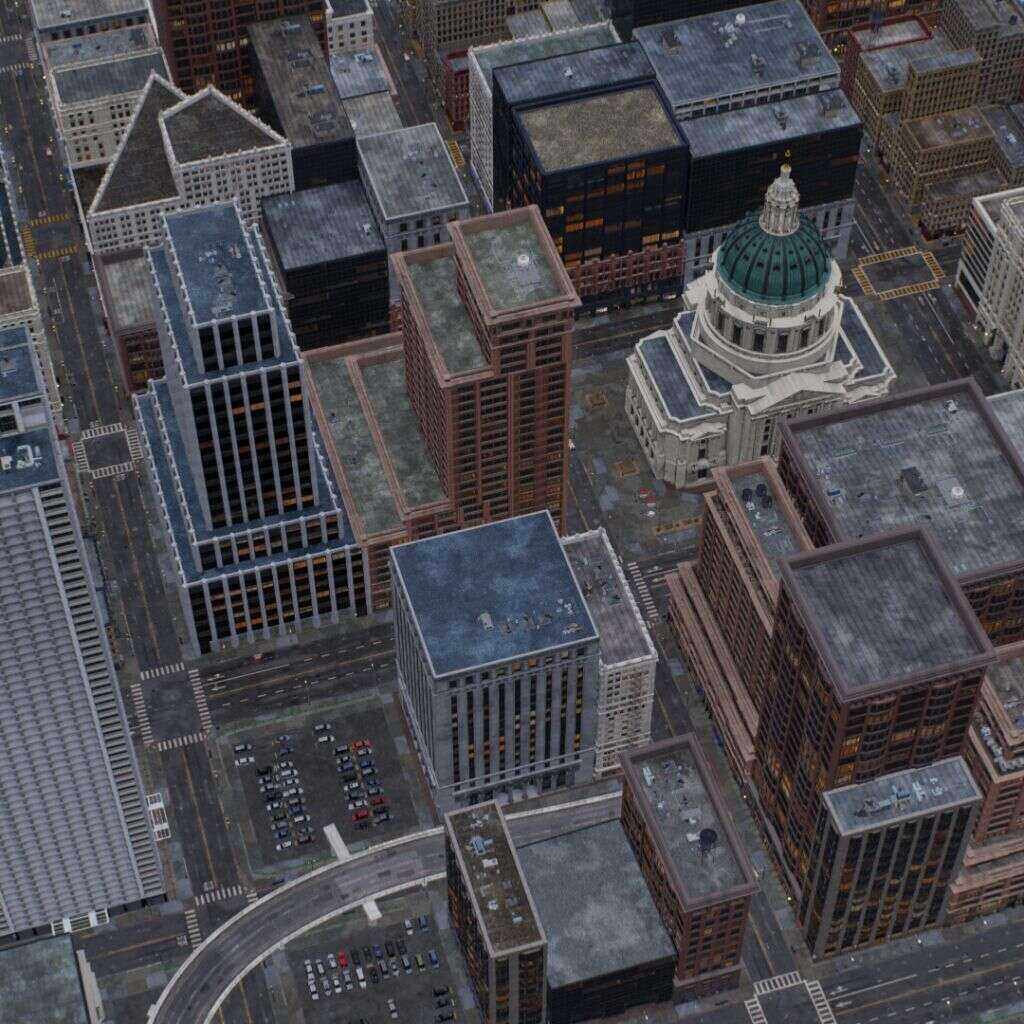} & 
    \imagecell[0.16]{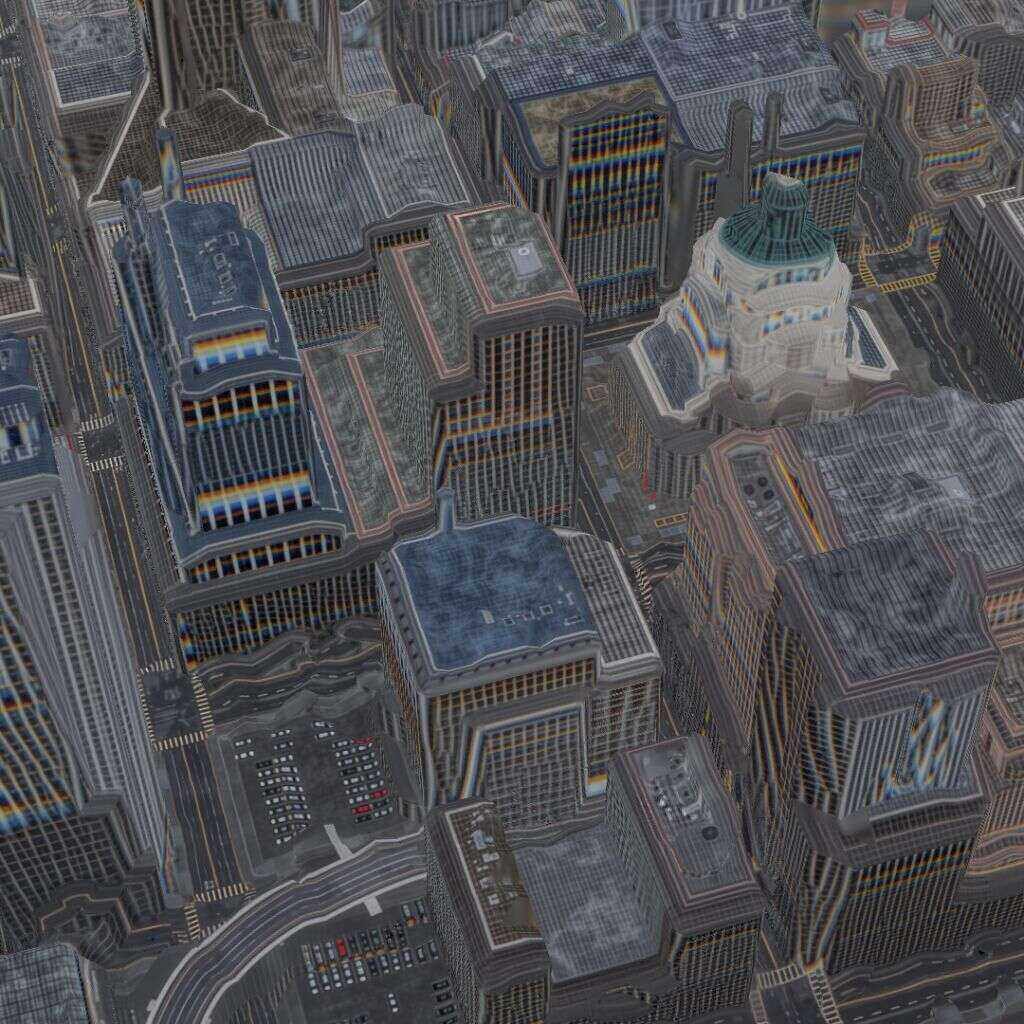} & 
    \imagecell[0.16]{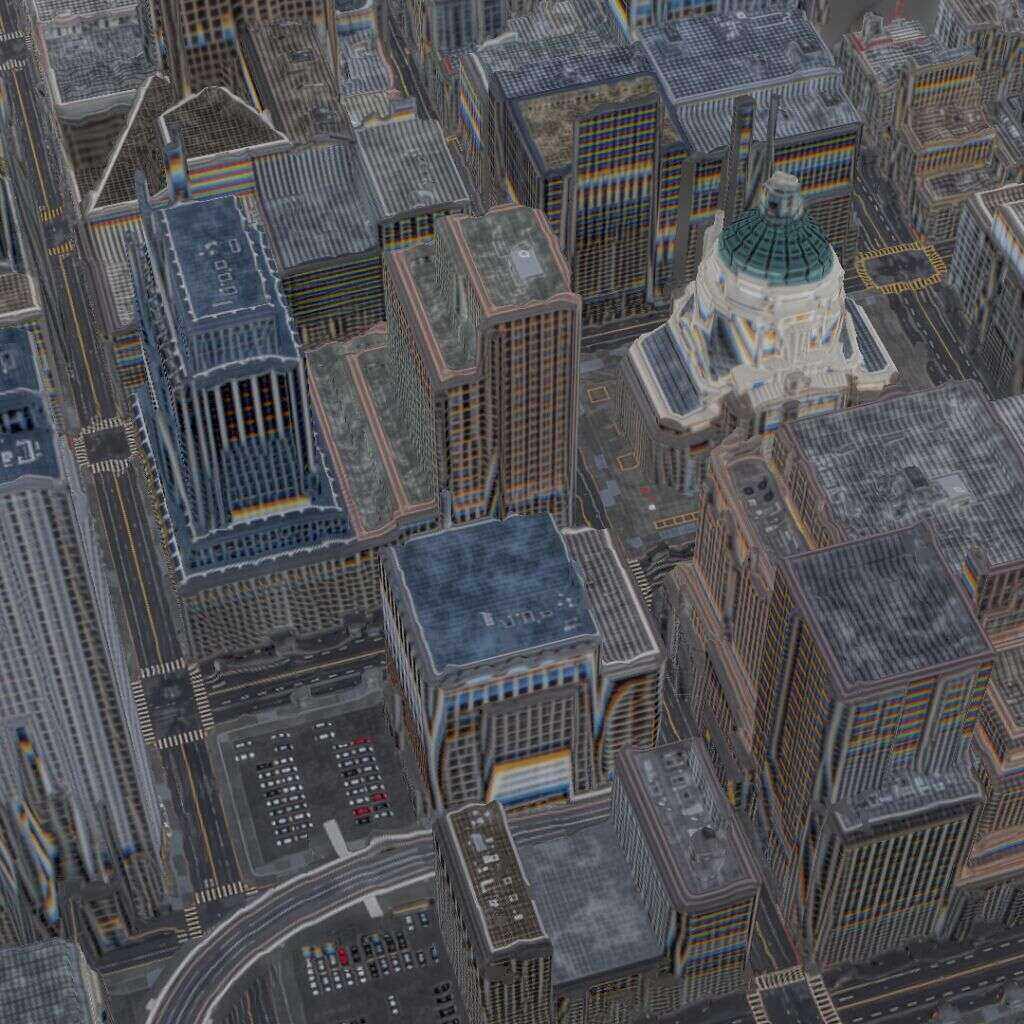} & 
    \imagecell[0.16]{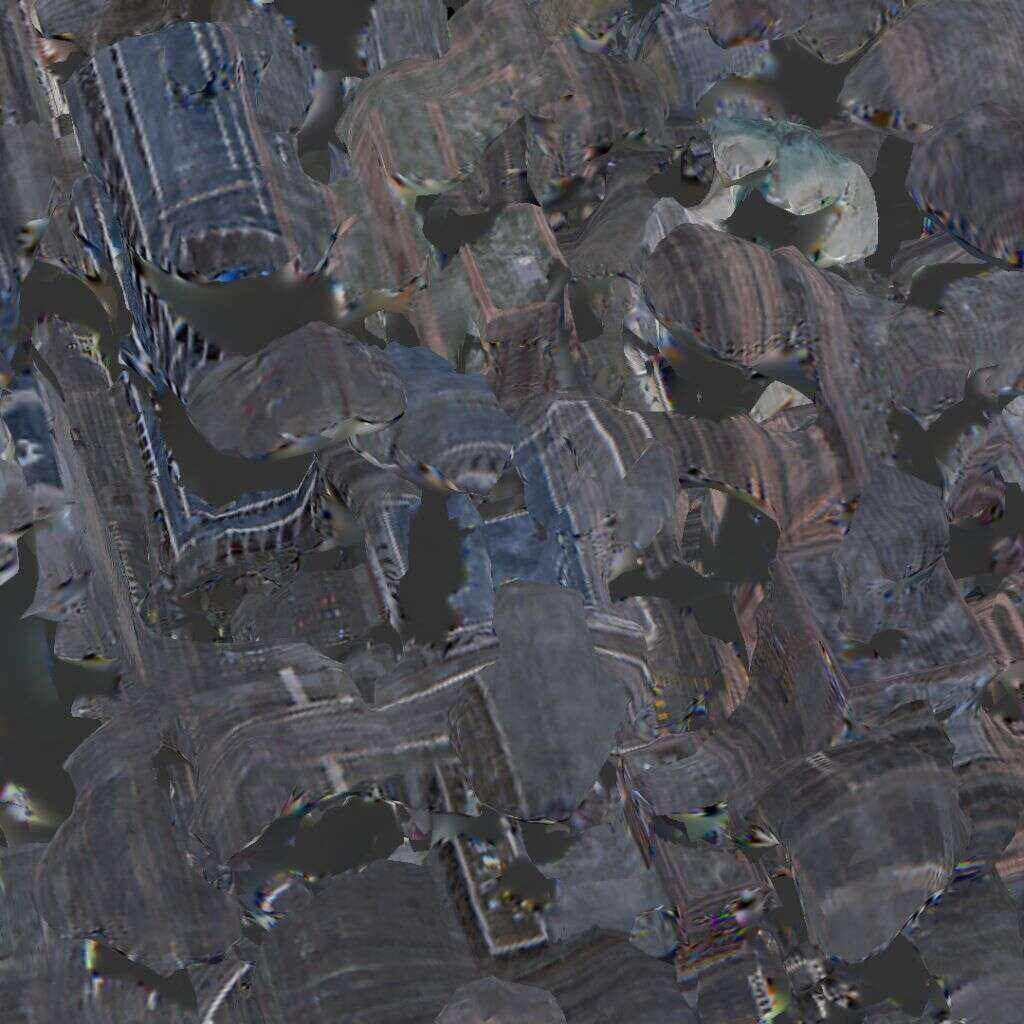} & 
    \imagecell[0.16]{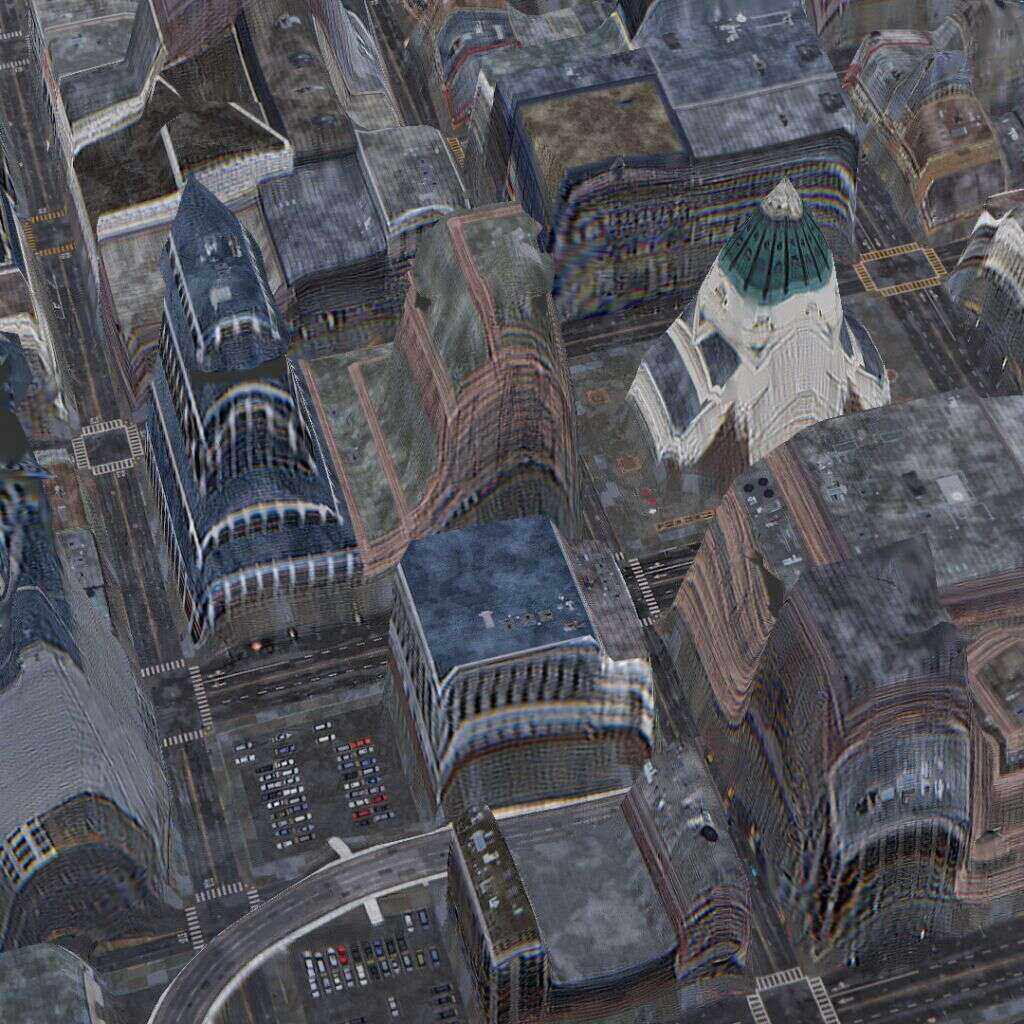} & 
    \imagecell[0.16]{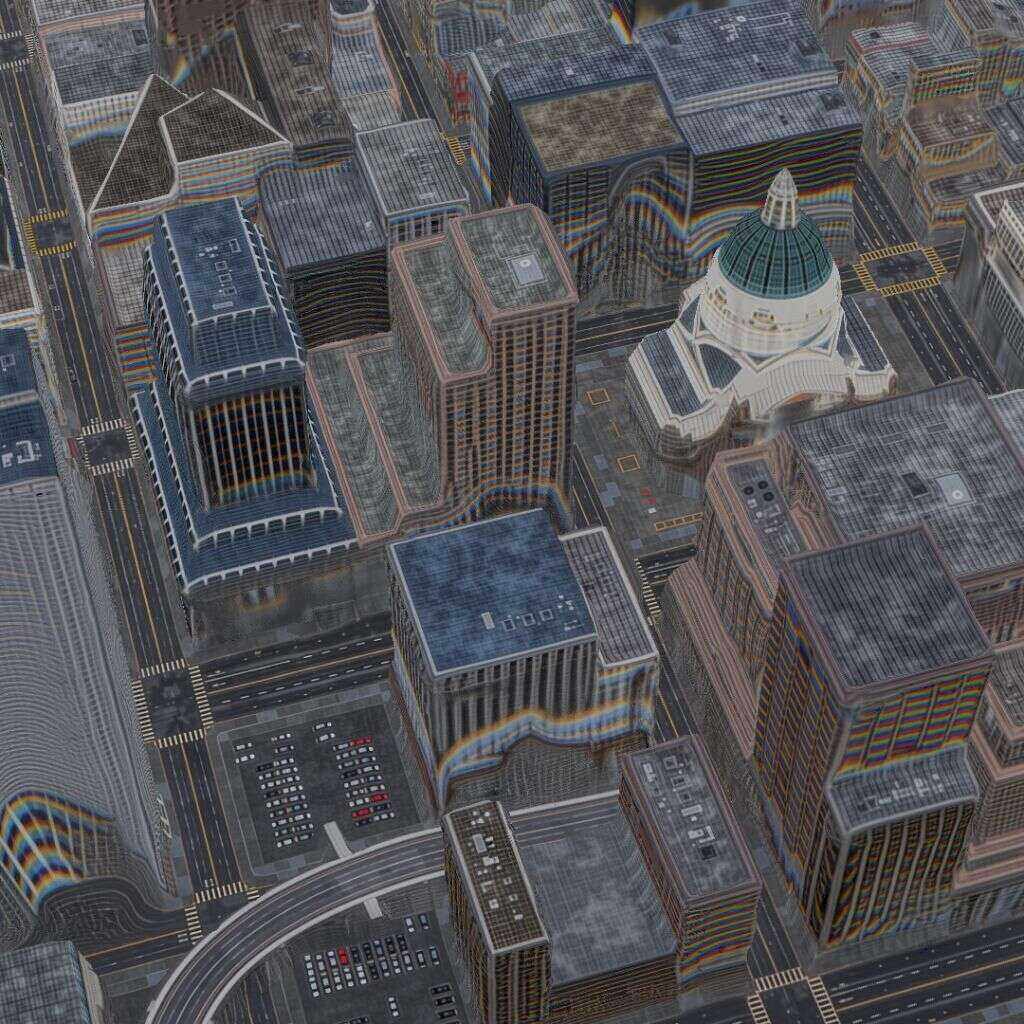} \\
    \vspace*{-10pt} \\

    \\
    \vspace*{-20pt}
    \\
    G.T. & 
    MC 128 & 
    MC 256 & 
    3D SDF &
    CD & 
    Ours \\
    
    \end{tabular}
    \end{spacing}
	\caption{ 
    \textbf{Visualization of ablation results on geometry}. 
    We compare our full geometry pipeline against variants using low-/high-resolution Marching Cubes, a generic 3D SDF with FlexiCubes, and a Chamfer-distance-only supervision.
    Only our Z-Monotonic SDF with height-map + regularized training recovers clean roofs, vertical facades, and watertight, artifact-free structures.
    }
    \label{fig:supp:ablation-geo}
    \vspace*{-0.3cm}
\end{figure*}

\cref{fig:supp:ablation-geo} compares different geometric representations and meshing strategies on the MatrixCity-Satellite scene.
From left to right, we visualize the ground truth, a naive 2.5D mesh obtained via Marching Cubes at low and high resolutions (MC 128/256), a full 3D SDF optimized with FlexiCubes, a variant trained with a Chamfer distance-based loss only, and our full Z-Monotonic SDF model.
Higher-resolution Marching Cubes still suffer from stair-step artifacts, and full 3D SDFs completely fail due to topological inconsistencies.
Our Z-Monotonic SDF yields cleaner roofs and vertically extruded facades, leading to the most faithful reconstruction.

\begin{figure*}[p]
	\centering
    \begin{spacing}{1} 
    \setlength\tabcolsep{1pt}
    \begin{tabular}{cccc}

    \imagecell[0.24]{figures/supp/ablation/gt/5.jpg} & 
    \imagecell[0.24]{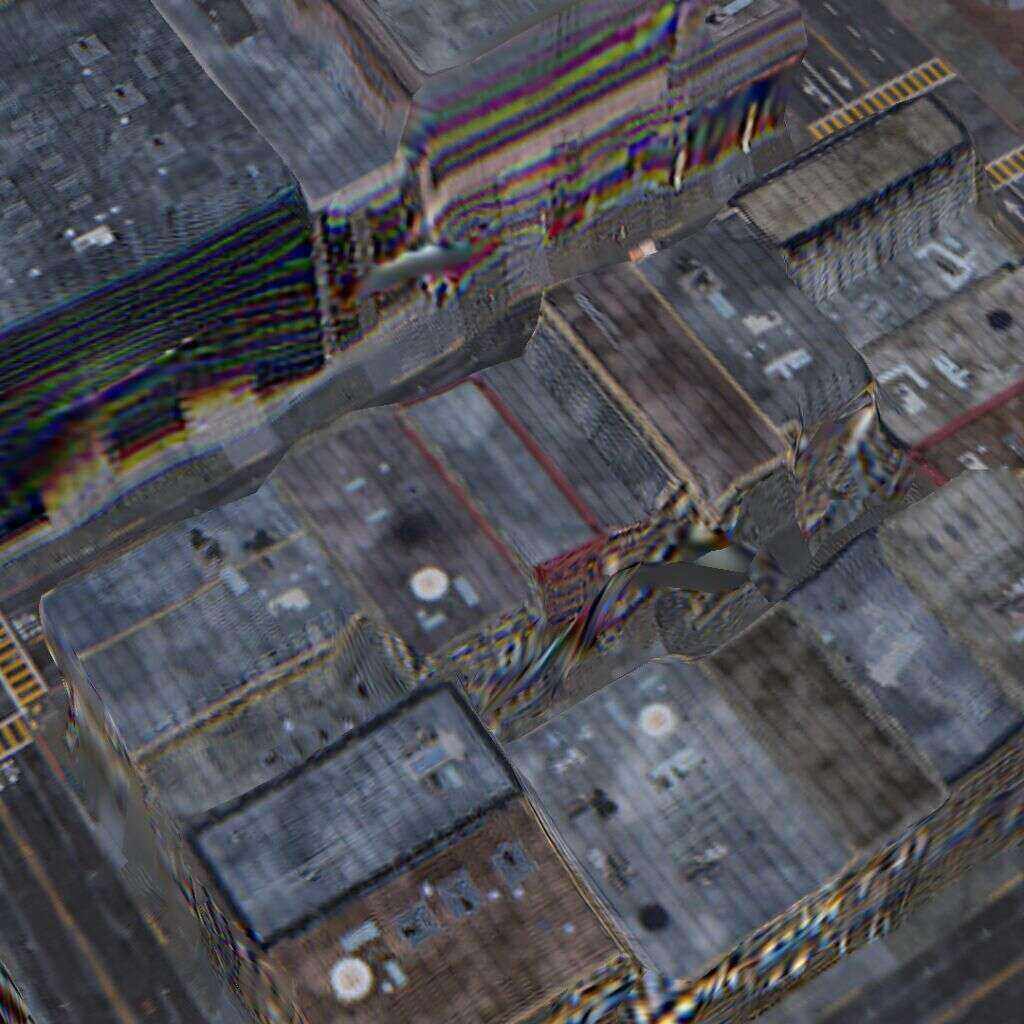} & 
    \imagecell[0.24]{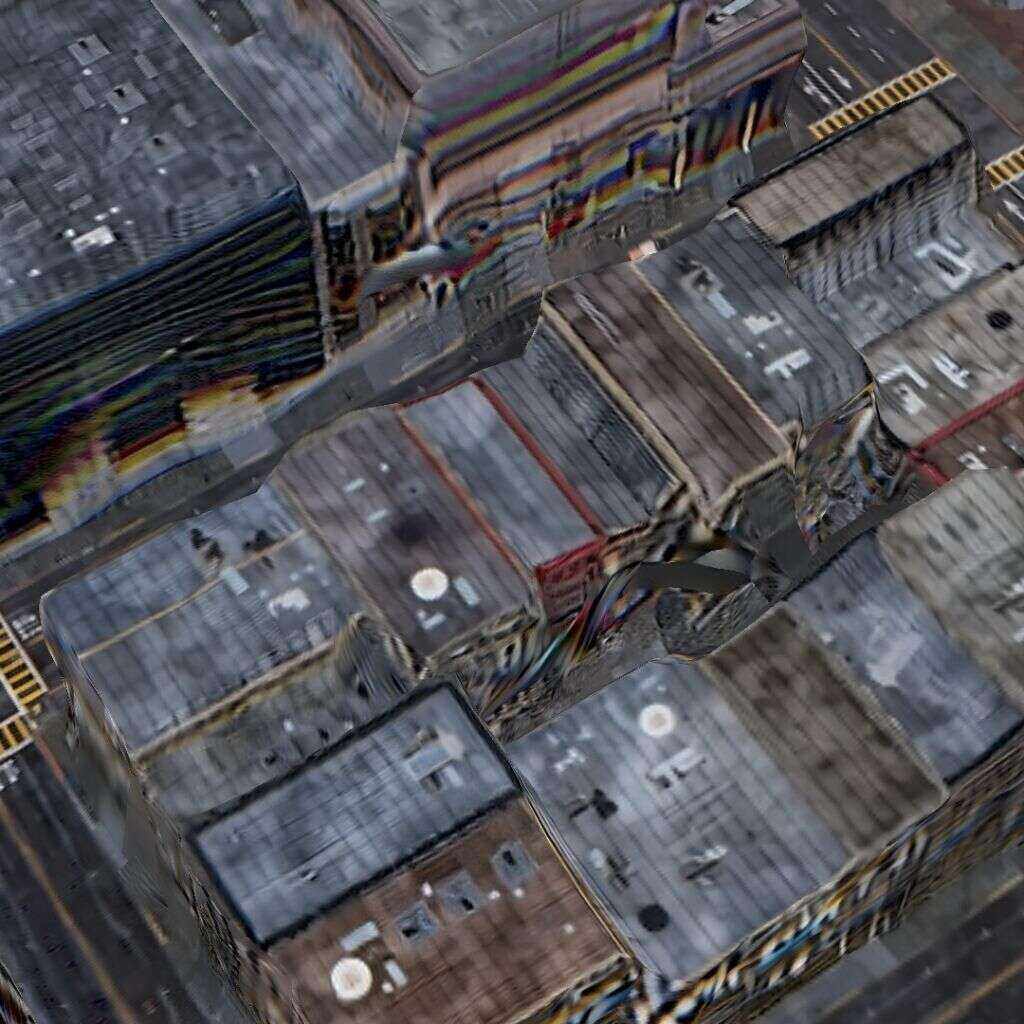} & 
    \imagecell[0.24]{figures/supp/ablation/ours-stage2/5.jpg} \\
    \vspace*{-10pt} \\

    \imagecell[0.24]{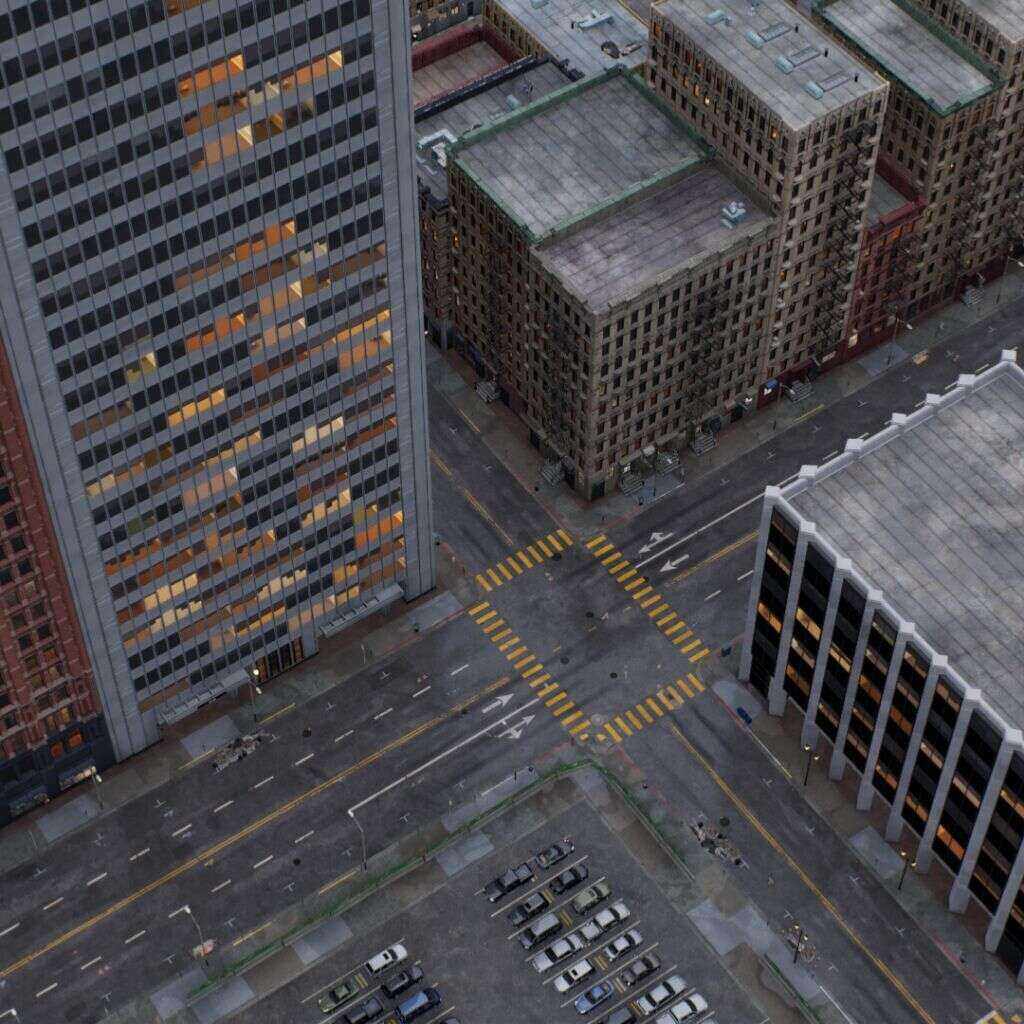} & 
    \imagecell[0.24]{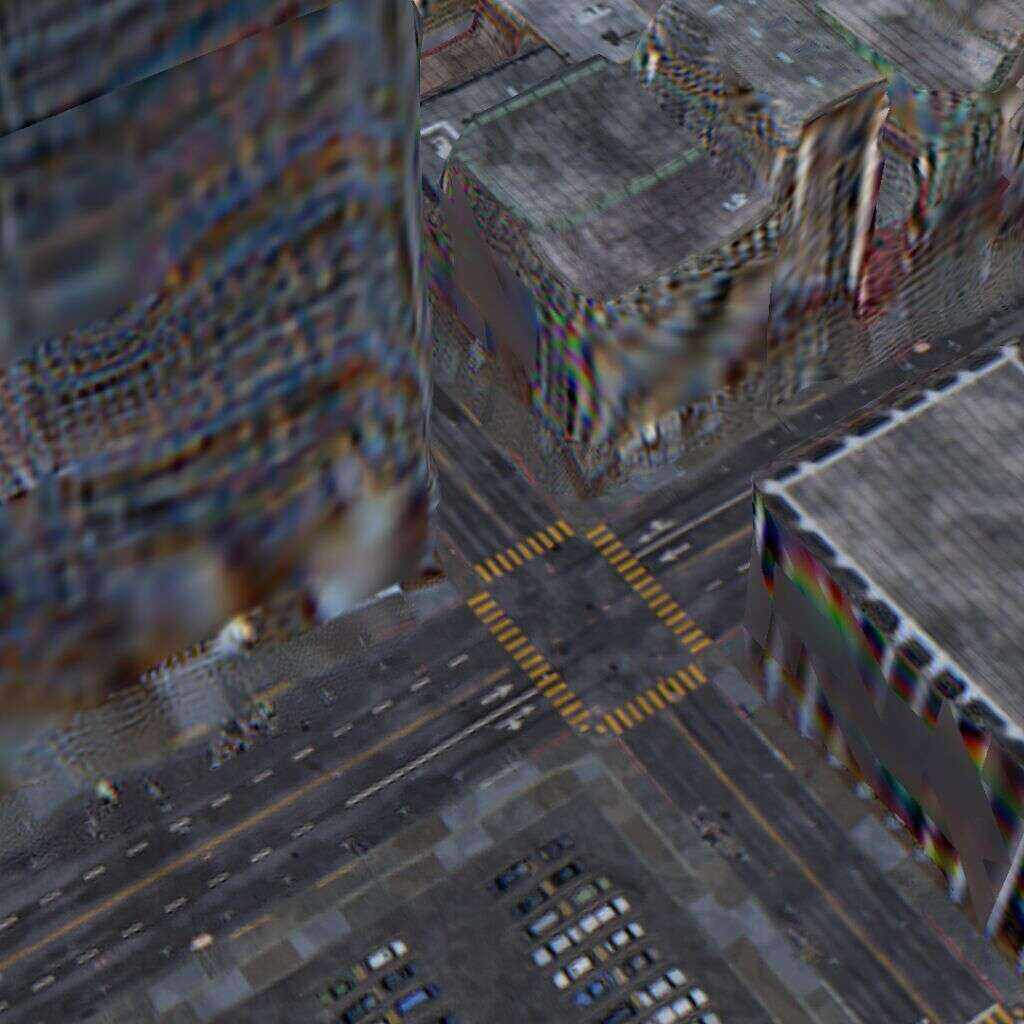} & 
    \imagecell[0.24]{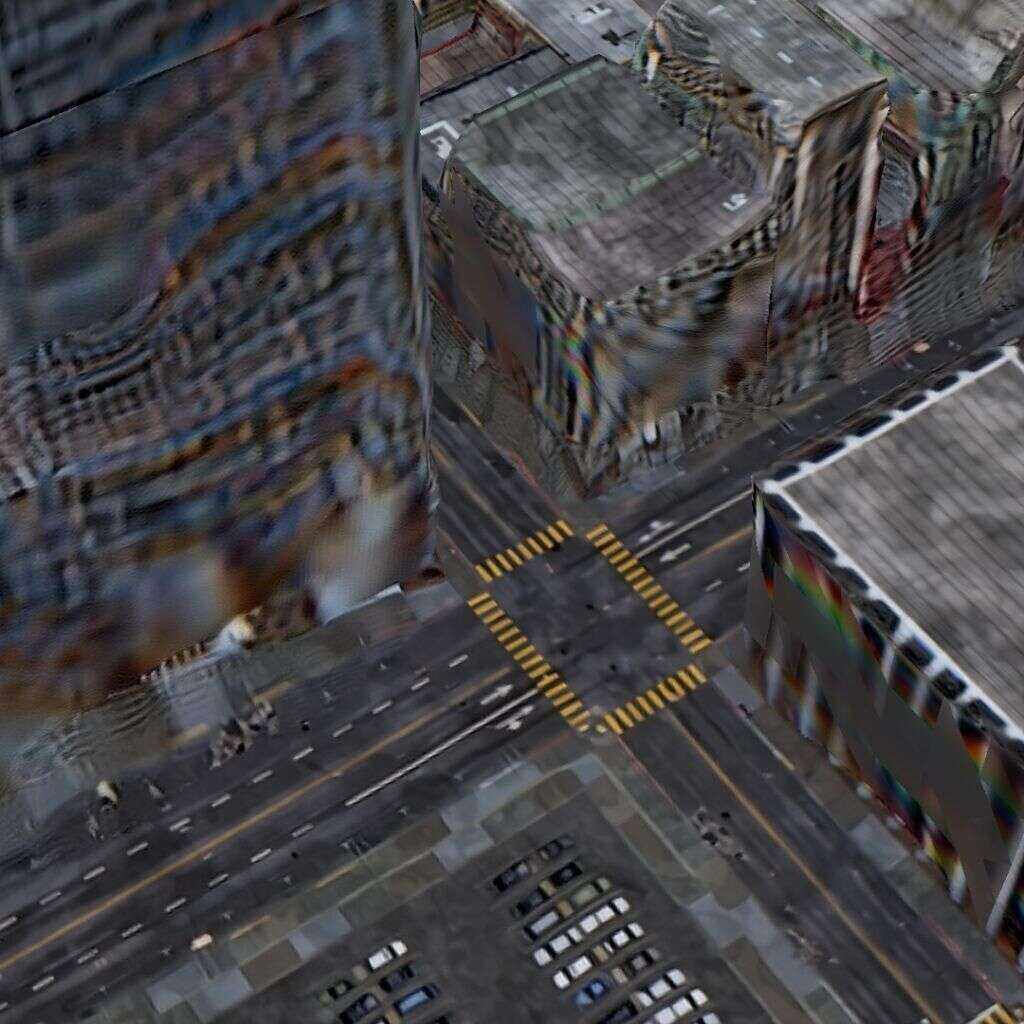} & 
    \imagecell[0.24]{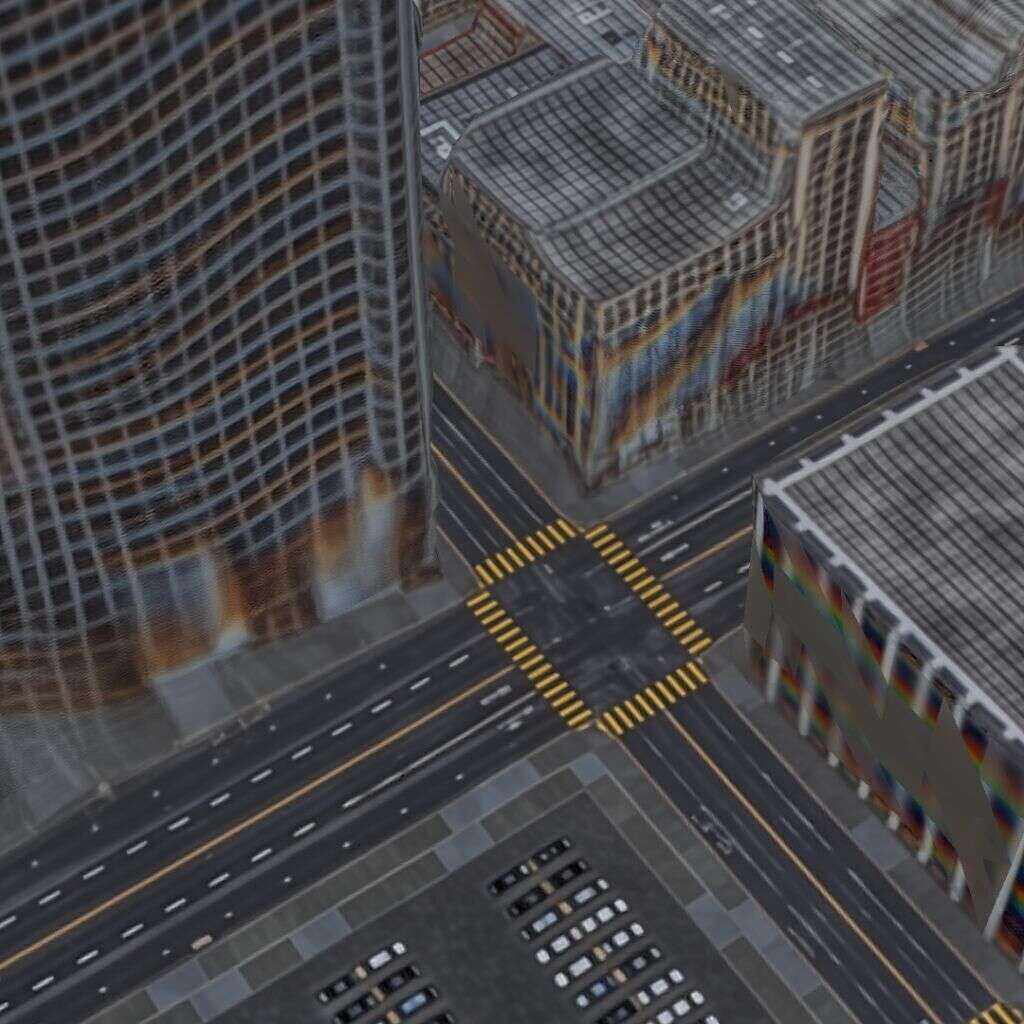} \\
    \vspace*{-10pt} \\

    \imagecell[0.24]{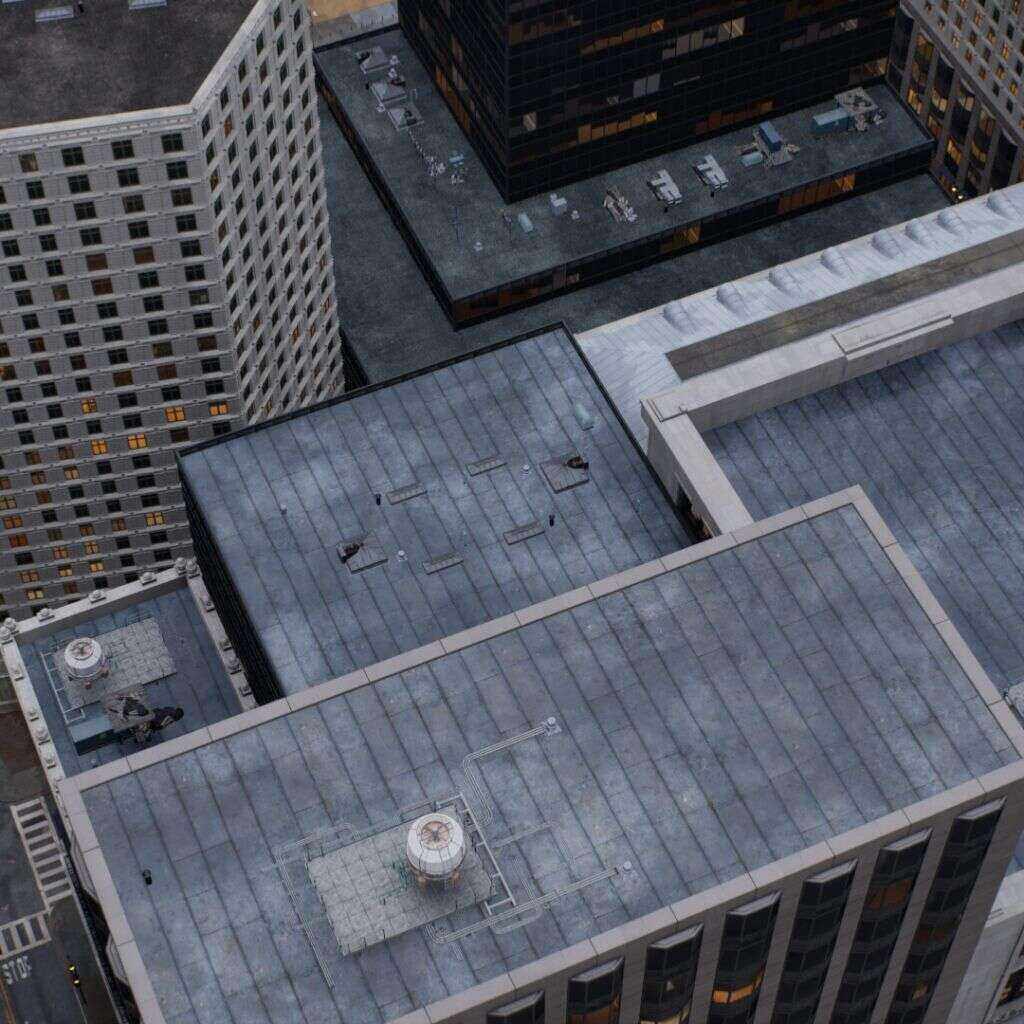} & 
    \imagecell[0.24]{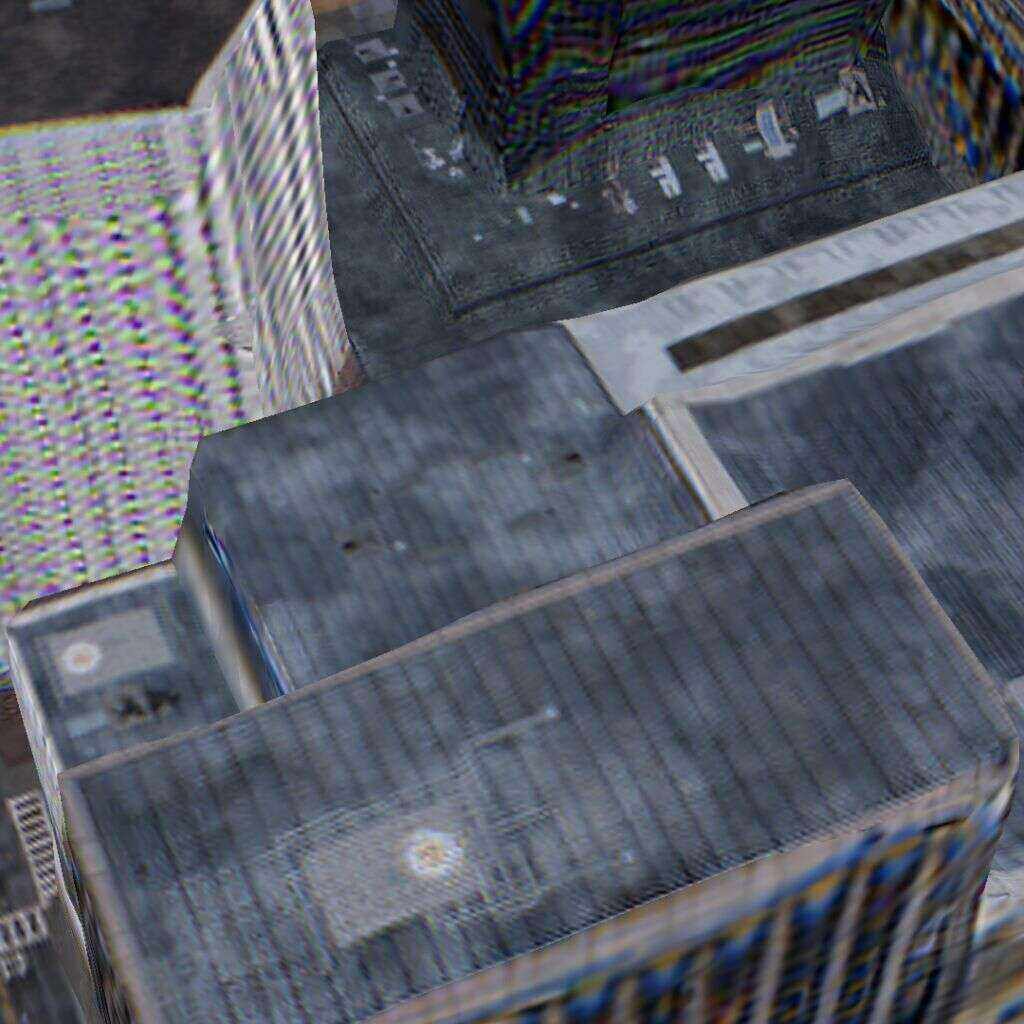} & 
    \imagecell[0.24]{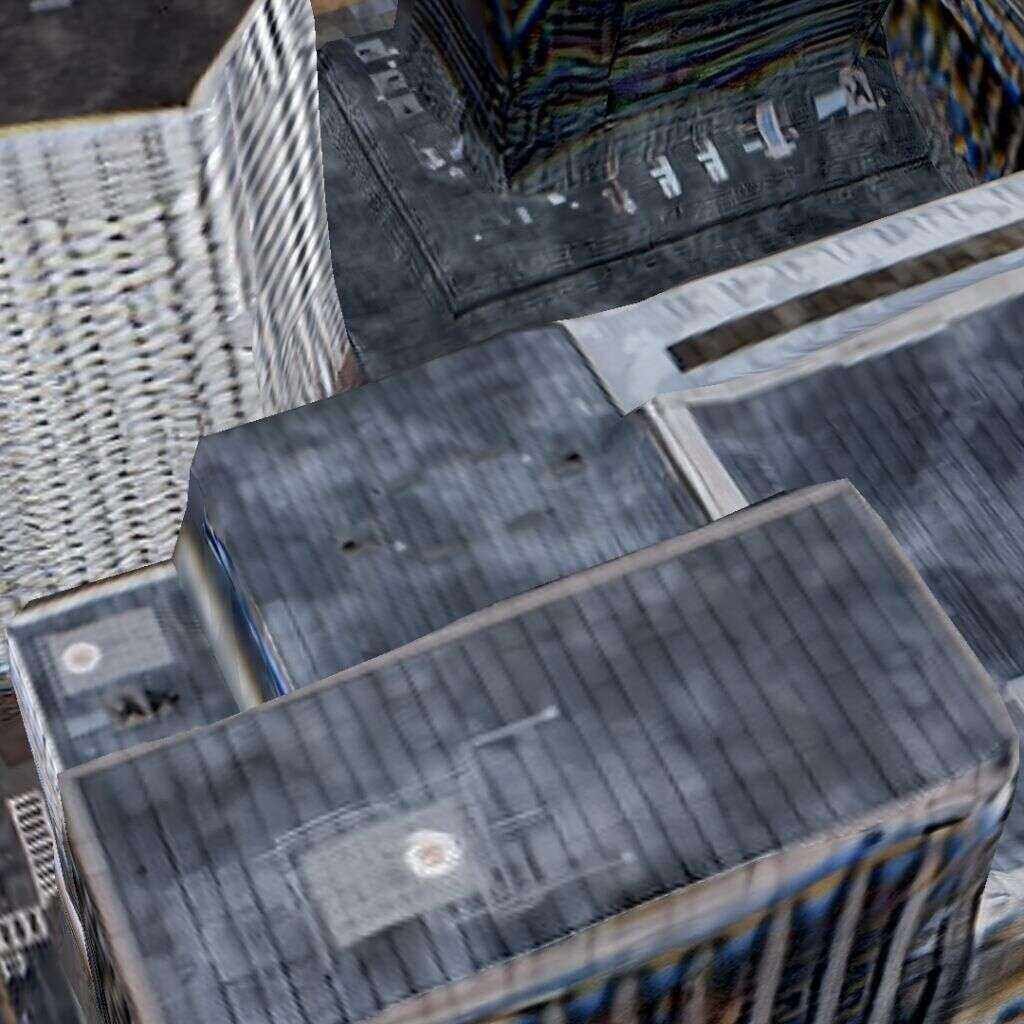} & 
    \imagecell[0.24]{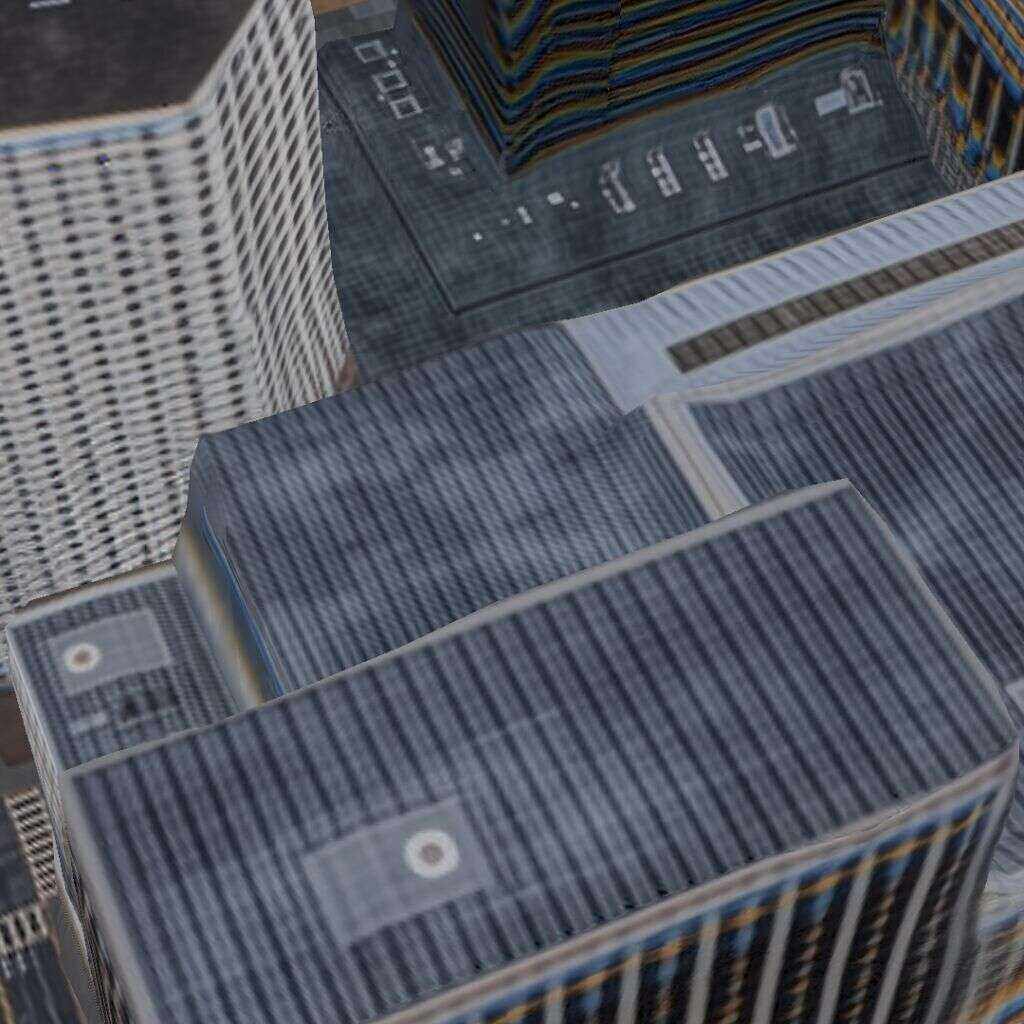} \\
    \vspace*{-10pt} \\
    
    \imagecell[0.24]{figures/supp/ablation/gt/29.jpg} & 
    \imagecell[0.24]{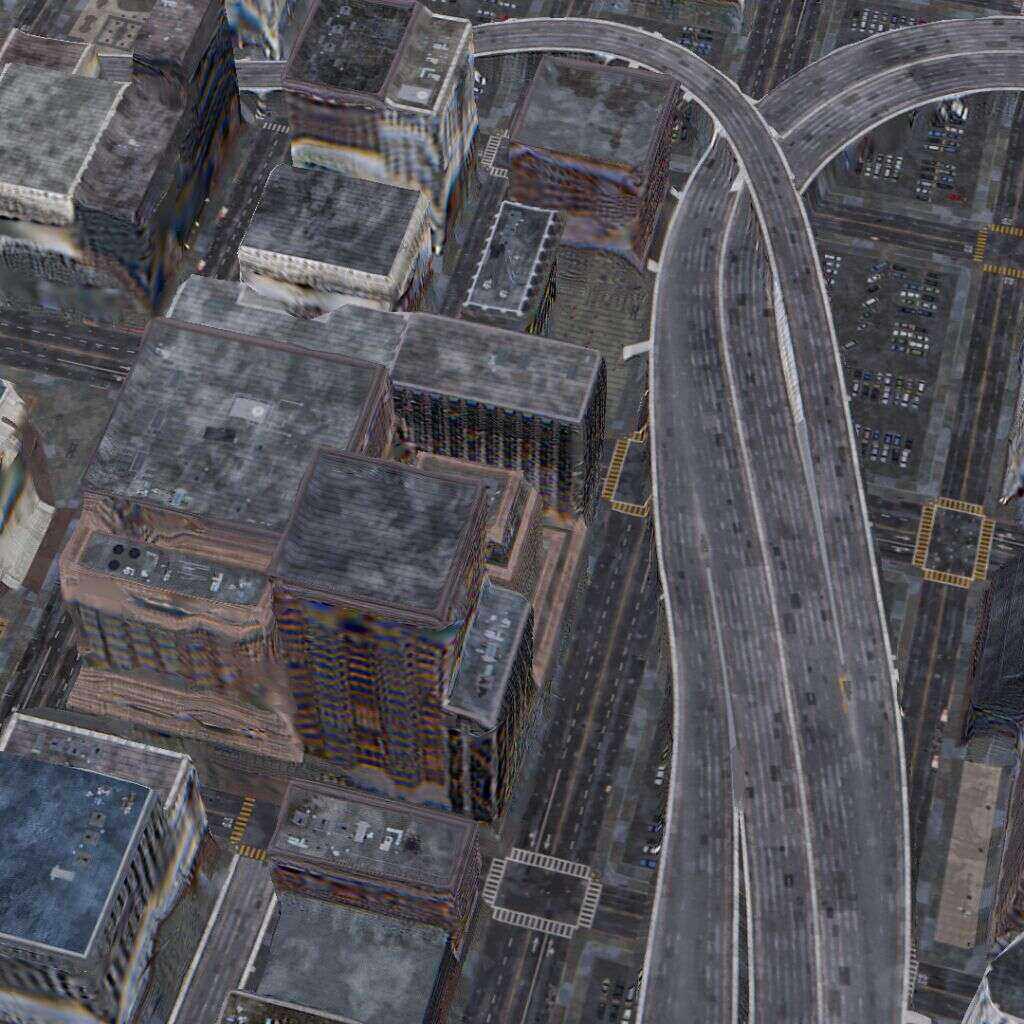} & 
    \imagecell[0.24]{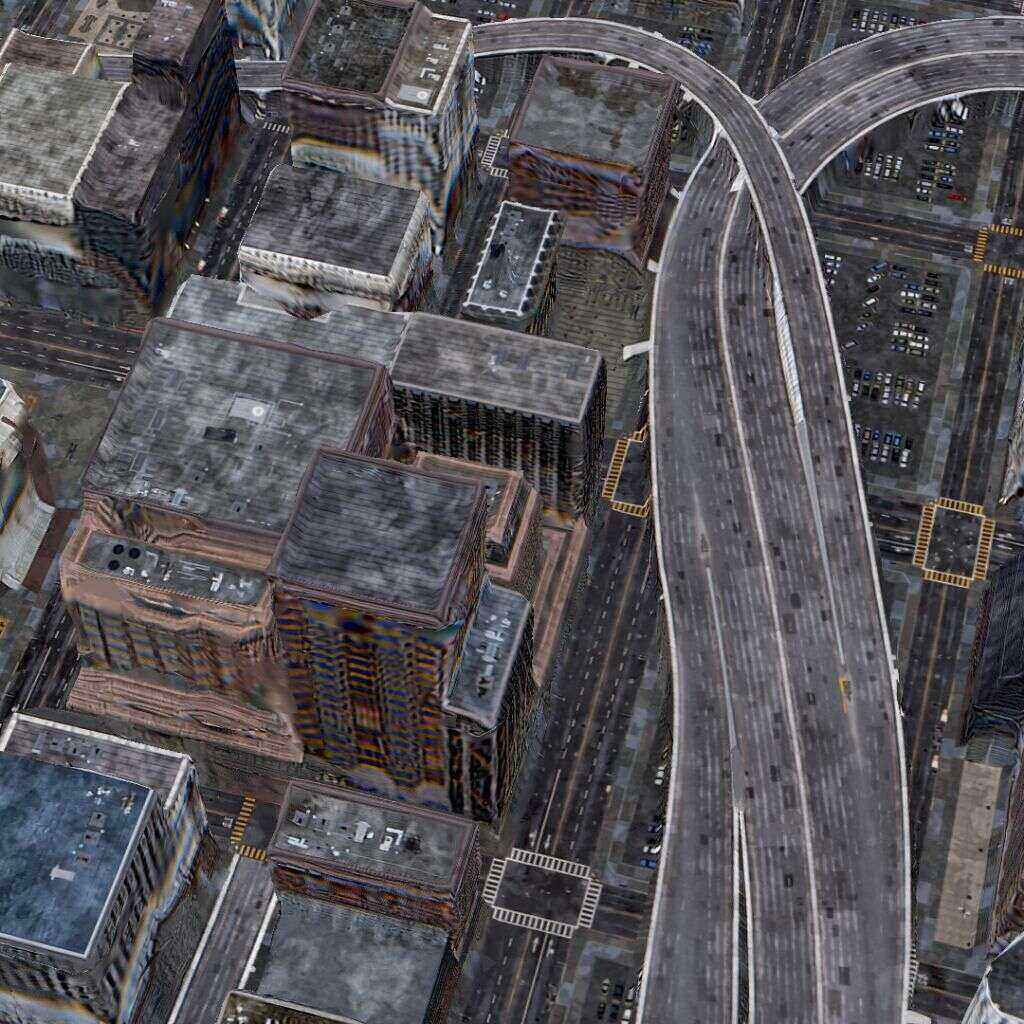} & 
    \imagecell[0.24]{figures/supp/ablation/ours-stage2/29.jpg} \\
    \vspace*{-10pt} \\
    
    \\
    \vspace*{-20pt}
    \\
    G.T. & 
    w/o Image Restoration & 
    w/ Flux-Kontext & 
    Ours \\
    
    \end{tabular}
    \end{spacing}
	\caption{ 
    \textbf{Visualization of ablation results on appearance modeling}. 
    Disabling image restoration or directly plugging in a generic FLUX-Kontext \cite{flux2024} model produces blurry and inconsistent textures, whereas our fine-tuned, deterministic restorer yields both sharp and view-consistent appearances that best match the ground truth.
    }
    \label{fig:supp:ablation-app}
    \vspace*{-0.3cm}
\end{figure*}

\cref{fig:supp:ablation-app} examines the impact of different appearance modeling choices.
We show the ground truth, a variant excluding our image restoration network (``w/o Image Restoration''), a variant that directly uses the off-the-shelf FLUX-Kontext model for enhancement, and our full appearance pipeline.
Without explicit restoration, textures remain blurry and exhibit baked-in projection artifacts.
Using FLUX-Kontext improves sharpness but introduces view-inconsistent hallucinations.
In contrast, our fine-tuned, deterministic restorer produces sharp and globally consistent textures across views, which is critical for stable texture optimization.

\subsection{Applications}
\label{sub-sec:aerial}

\begin{figure*}[p]
	\centering
    \includegraphics[width=0.8\linewidth]{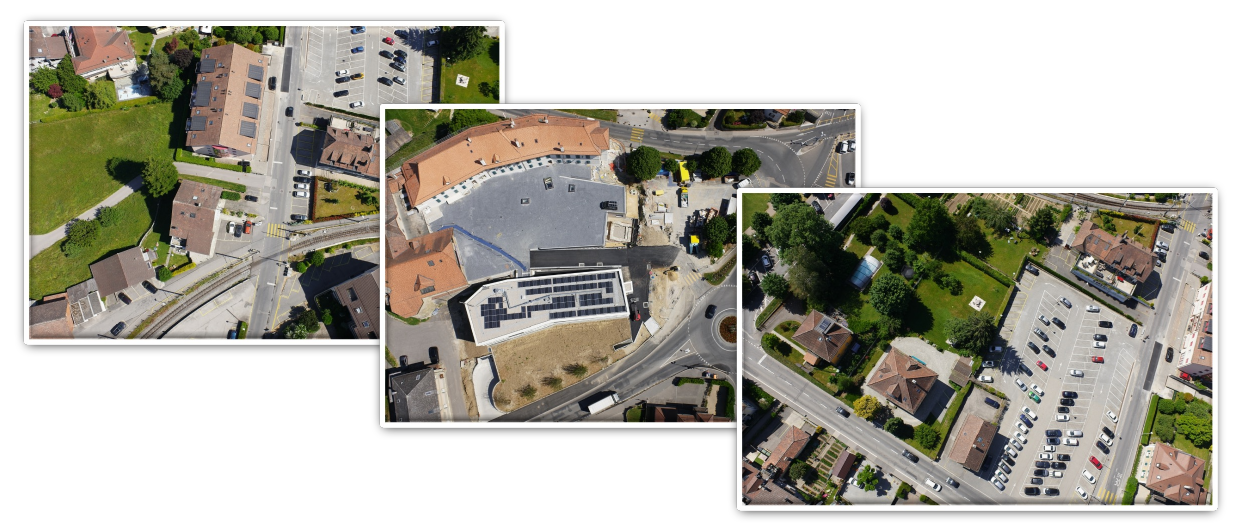}
    \\
    \textbf{Input}
    \\
    \includegraphics[width=\linewidth]{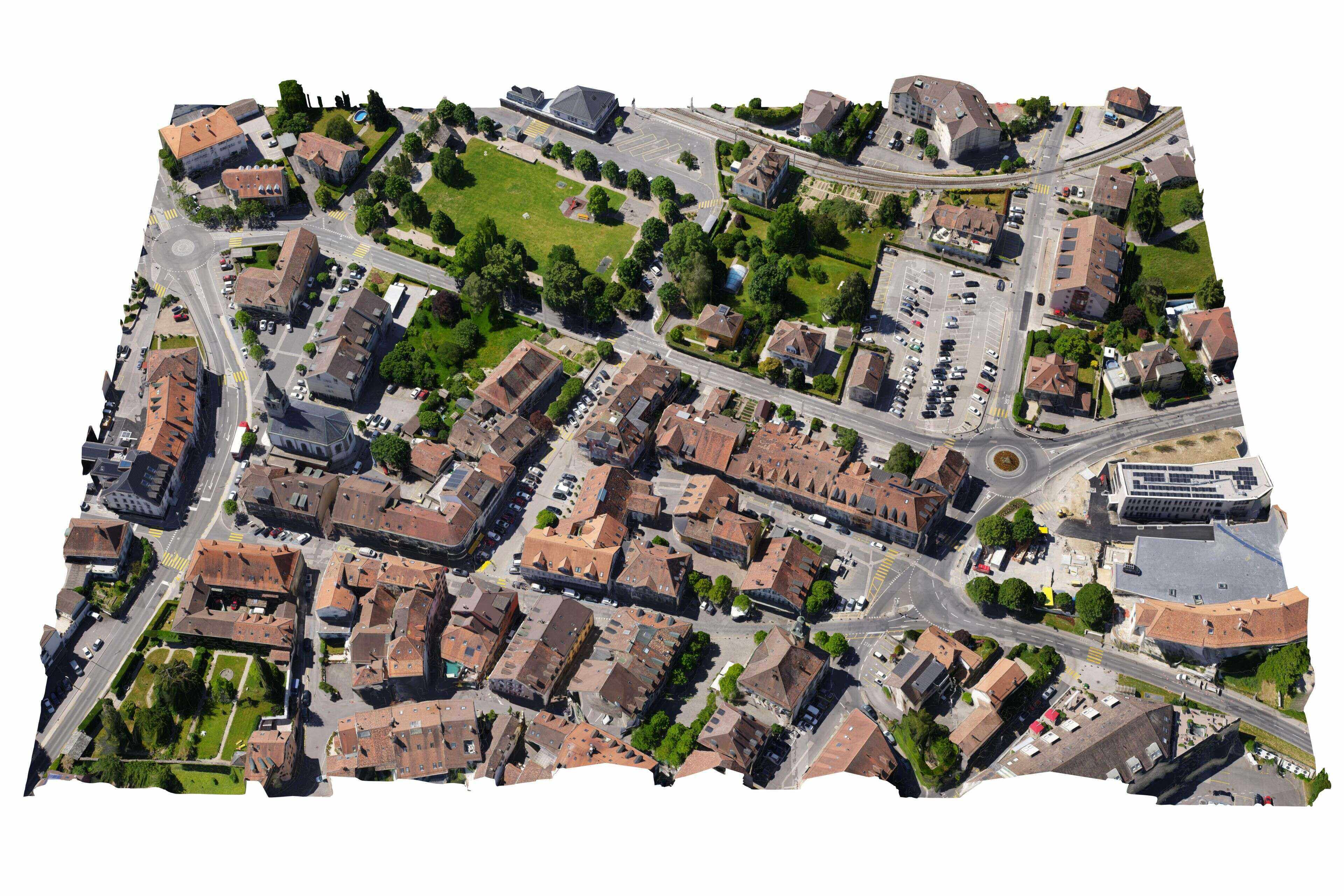}
    \\
    \textbf{Result}
	\caption{ 
    \textbf{Overview of aerial-view reconstruction results of our method.}
    We demonstrate that the same pipeline can be applied to an aerial-view reconstruction task: given a set of oblique aerial images (top), our method reconstructs a geometrically accurate and visually realistic scene (bottom), illustrating its applicability beyond satellite imagery to general photogrammetric settings.
    }
    \label{fig:supp:aerial}
    \vspace*{-0.3cm}
\end{figure*}

We illustrate an additional application of our framework beyond satellite-to-ground reconstruction.
\cref{fig:supp:aerial} shows an example where our method is applied to reconstruct an aerial-view scene from a set of oblique aerial images \cite{Pix4DmaticDataset}.
The top row visualizes a subset of the input views, while the bottom row shows renderings from novel viewpoints using our reconstructed mesh and textures.
The results demonstrate that our 2.5D geometry prior and generative appearance refinement are also applicable to aerial photogrammetry, yielding high-quality assets.

\subsection{Comparison with Remote Sensing Methods}
\label{sub-sec:supp:rs-comparison}

In the remote sensing community, several methods have also investigated 3D reconstruction from satellite imagery~\cite{derksen2021shadow, mari2022sat, mari2023multi, zhang2023sparsesat, zhang2024satensorf, aira2025gaussian, bai2025satgs}. 
These methods are primarily designed for accurate digital surface model (DSM) estimation and elevation recovery over large areas, and thus mainly focus on metric height accuracy. 
In contrast, our work targets high-fidelity reconstruction of the \emph{full 3D scene appearance}, including building facades and complex urban details, for photorealistic novel view synthesis from ground and near-ground viewpoints. 
This difference in objective naturally leads to complementary strengths: DSM-oriented methods excel at producing geographically accurate surface models, whereas our method emphasizes visually coherent, view-consistent 3D assets suitable for immersive rendering and city-scale visual applications.

Following Skyfall-GS~\cite{lee2025skyfall}, we compare our method against two representative methods Sat-NeRF~\cite{zhang2024satensorf} and EOGS~\cite{aira2025gaussian} on the DFC 2019 dataset in terms of novel view synthesis quality. 
As shown in \cref{table:exp:rs}, our method achieves superior performance across all metrics, indicating that our reconstruction better preserves the visual fidelity of the 3D scene from challenging novel viewpoints.

\begin{table}[t]
\centering
\renewcommand{\arraystretch}{1.1}
\setlength{\tabcolsep}{8pt}
\begin{tabular}{lccc}
\toprule
\textbf{Method} & \textbf{PSNR} $\uparrow$ & \textbf{SSIM} $\uparrow$ & \textbf{LPIPS} $\downarrow$ \\
\midrule
Sat-NeRF~\cite{zhang2024satensorf} & 10.220 & 0.278 & 0.816 \\
EOGS~\cite{aira2025gaussian}      & 7.338 & 0.181 & 0.931 \\
Ours                              & \textbf{13.059} & \textbf{0.358} & \textbf{0.556} \\
\bottomrule
\end{tabular}
\caption{\textbf{Quantitative comparison with remote sensing methods on DFC 2019.} 
Our method achieves higher PSNR and SSIM and lower LPIPS for novel view synthesis compared to Sat-NeRF and EOGS, highlighting its focus on reconstructing visually faithful 3D scene appearance.}
\label{table:exp:rs}
\vspace{-3mm}
\end{table}

\begin{figure}[t]
    \centering
    \begin{spacing}{1}
    \setlength\tabcolsep{1pt}
    \begin{tabular}{cccc}
        \imagecell[0.24]{figures/supp/jax/JAX_068/gt/3.jpg} & 
        \imagecell[0.24]{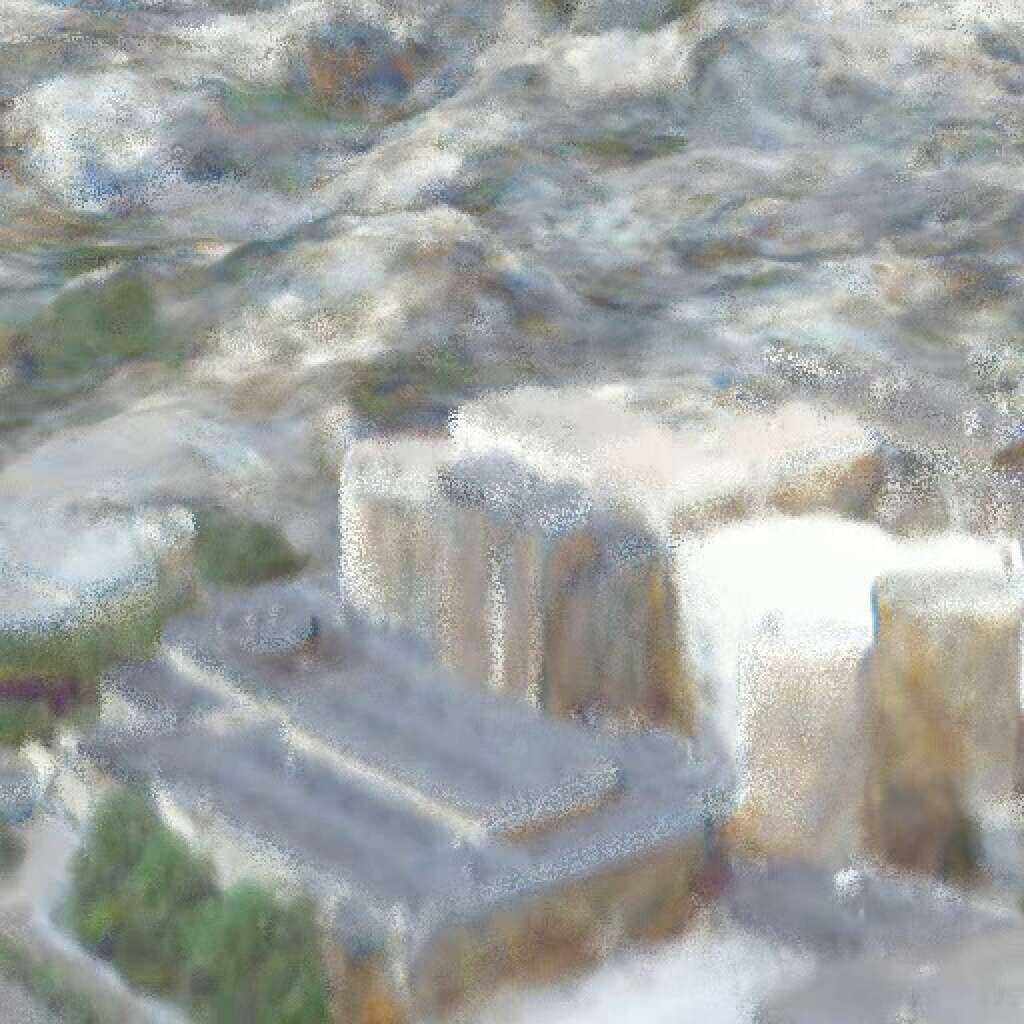} & 
        \imagecell[0.24]{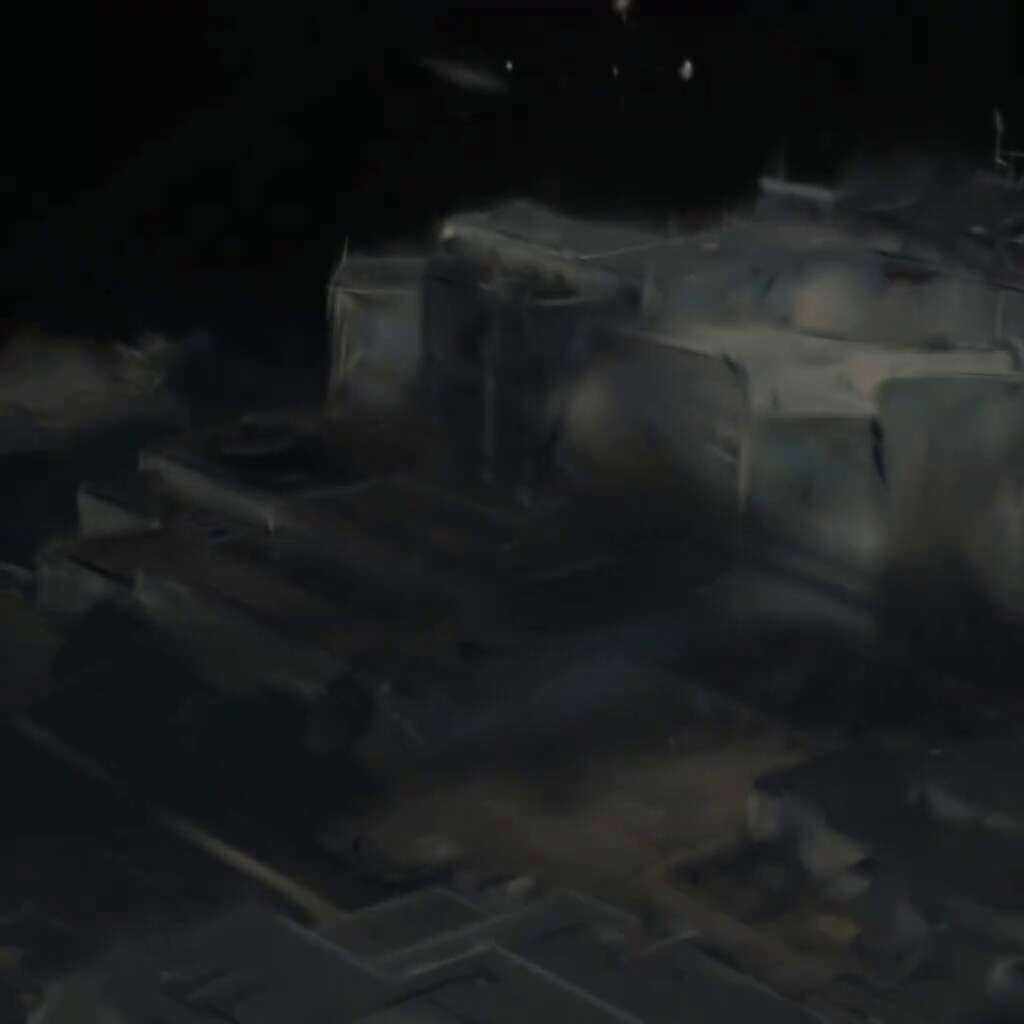} & 
        \imagecell[0.24]{figures/supp/jax/JAX_068/ours-stage2/3.jpg} \\ 
        \vspace{-9pt} \\
        \imagecell[0.24]{figures/supp/jax/JAX_214/gt/3.jpg} & 
        \imagecell[0.24]{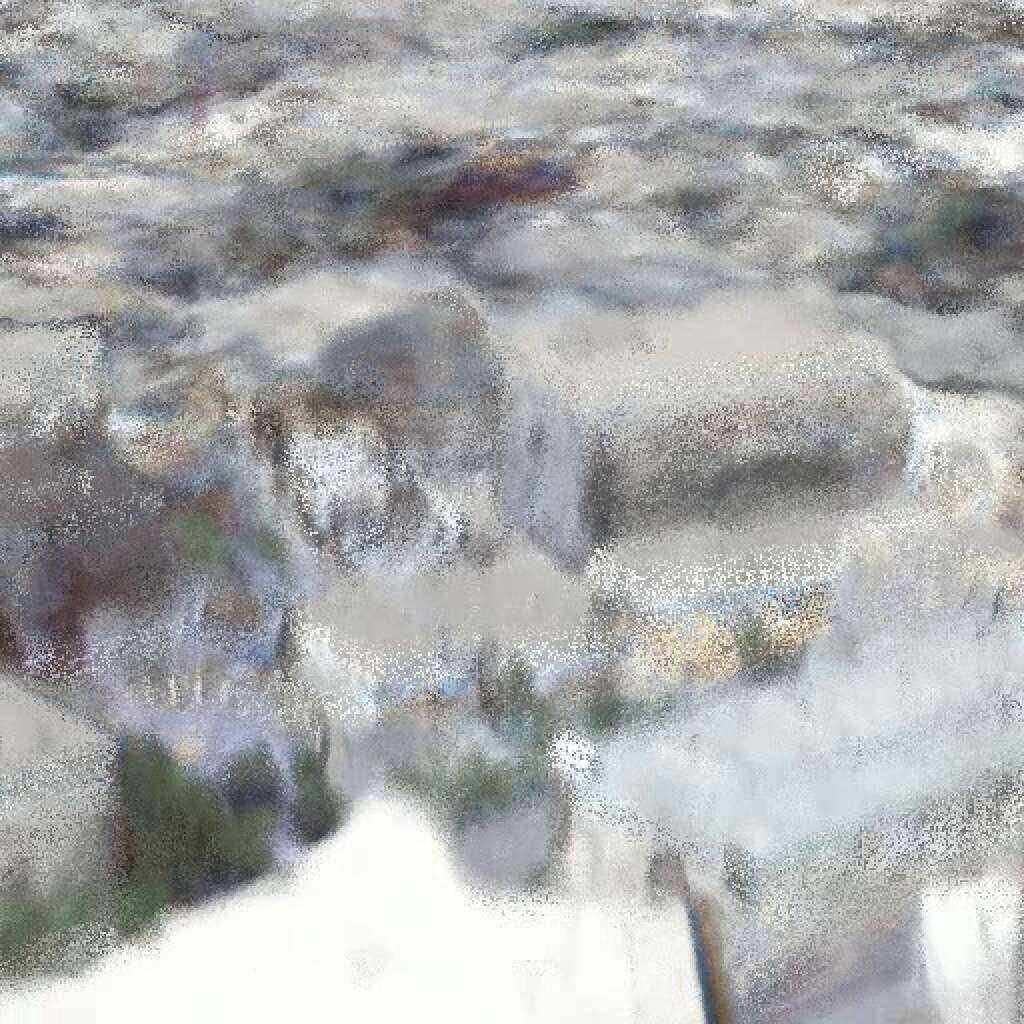} & 
        \imagecell[0.24]{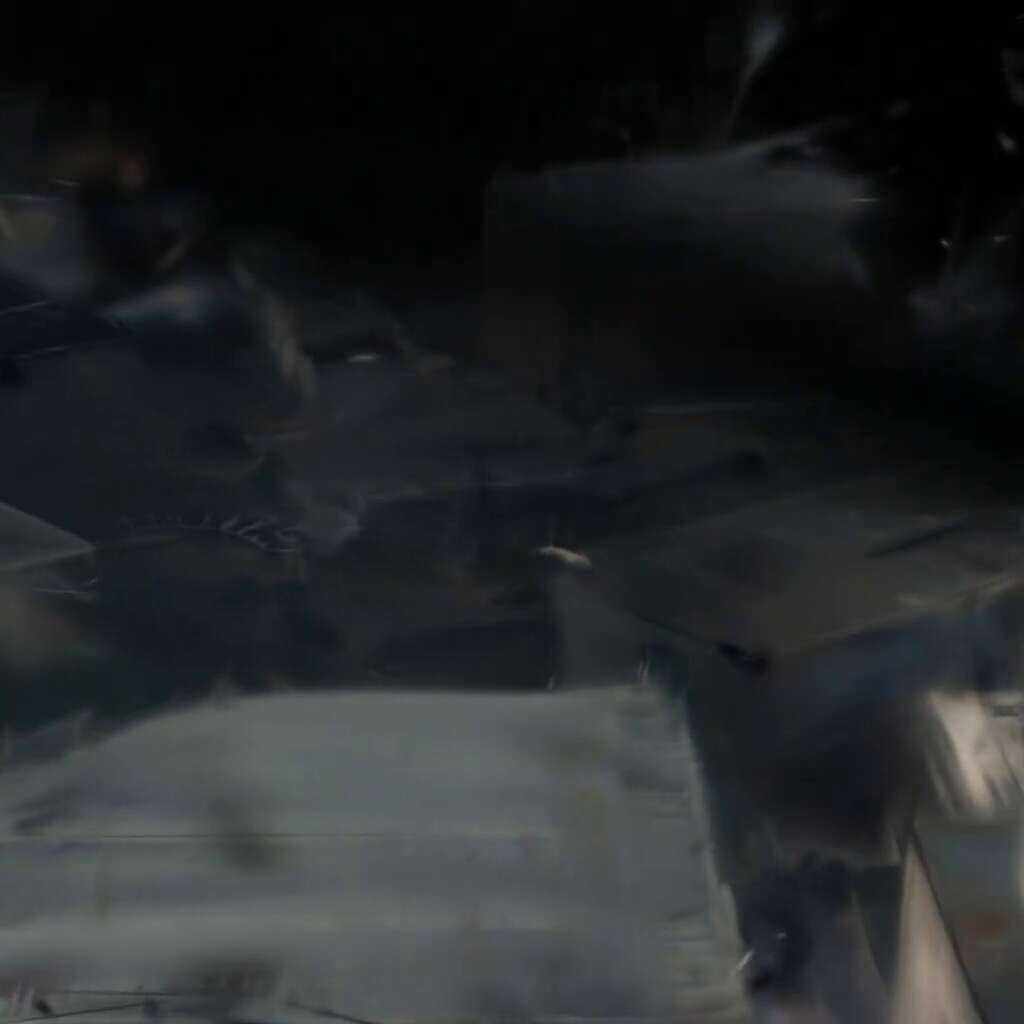} & 
        \imagecell[0.24]{figures/supp/jax/JAX_214/ours-stage2/3.jpg} \\
        \vspace{-6pt} \\
        {G.T.} & {Sat-NeRF} & {EOGS} & {Ours} \\
    \end{tabular}
    \end{spacing}
    \caption{\textbf{Qualitative comparison with remote sensing methods on DFC 2019.} 
    Compared to Sat-NeRF and EOGS, our method produces sharper facades and more realistic textures from oblique and near-ground viewpoints, reflecting our focus on reconstructing the visual appearance of the full 3D scene.}
    \label{fig:qualitative-rs}
    \vspace{-3mm}
\end{figure}

We further provide qualitative comparisons in \cref{fig:qualitative-rs}. 
As illustrated, Sat-NeRF and EoGS are able to recover plausible elevation and coarse structure, but their rendered views tend to exhibit oversmoothed textures and limited facade details when viewed from oblique or near-ground viewpoints. 
In contrast, our method produces sharper, more realistic facades and richer high-frequency details. 
These results underscore the complementary nature of our approach with respect to DSM-oriented remote sensing methods.

\begin{figure}
    \centering
    \setlength\tabcolsep{2pt}
    \begin{tabular}{cccc}
        \imagecell[0.24]{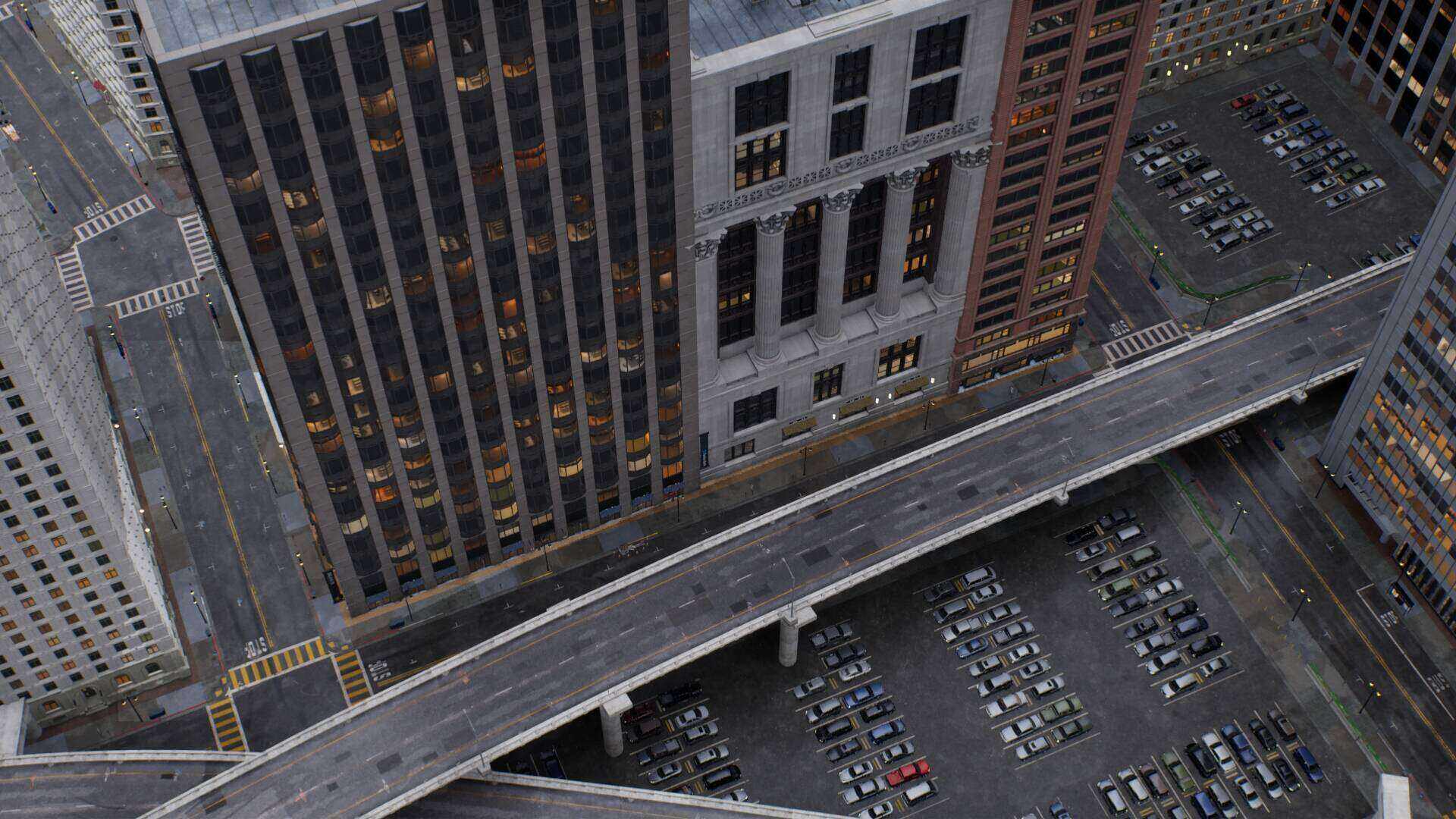} &
        \imagecell[0.24]{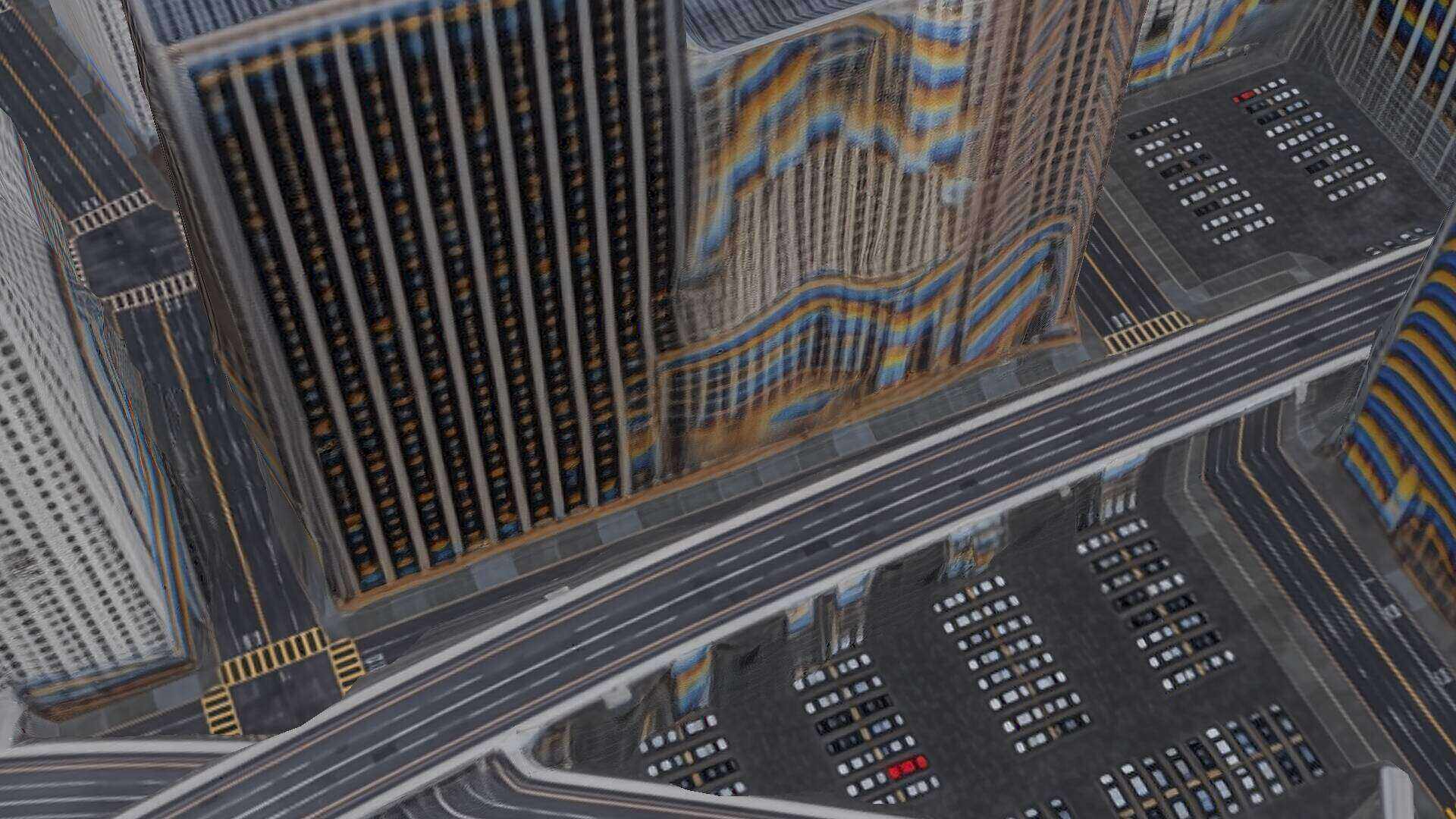} &
        \imagecell[0.24]{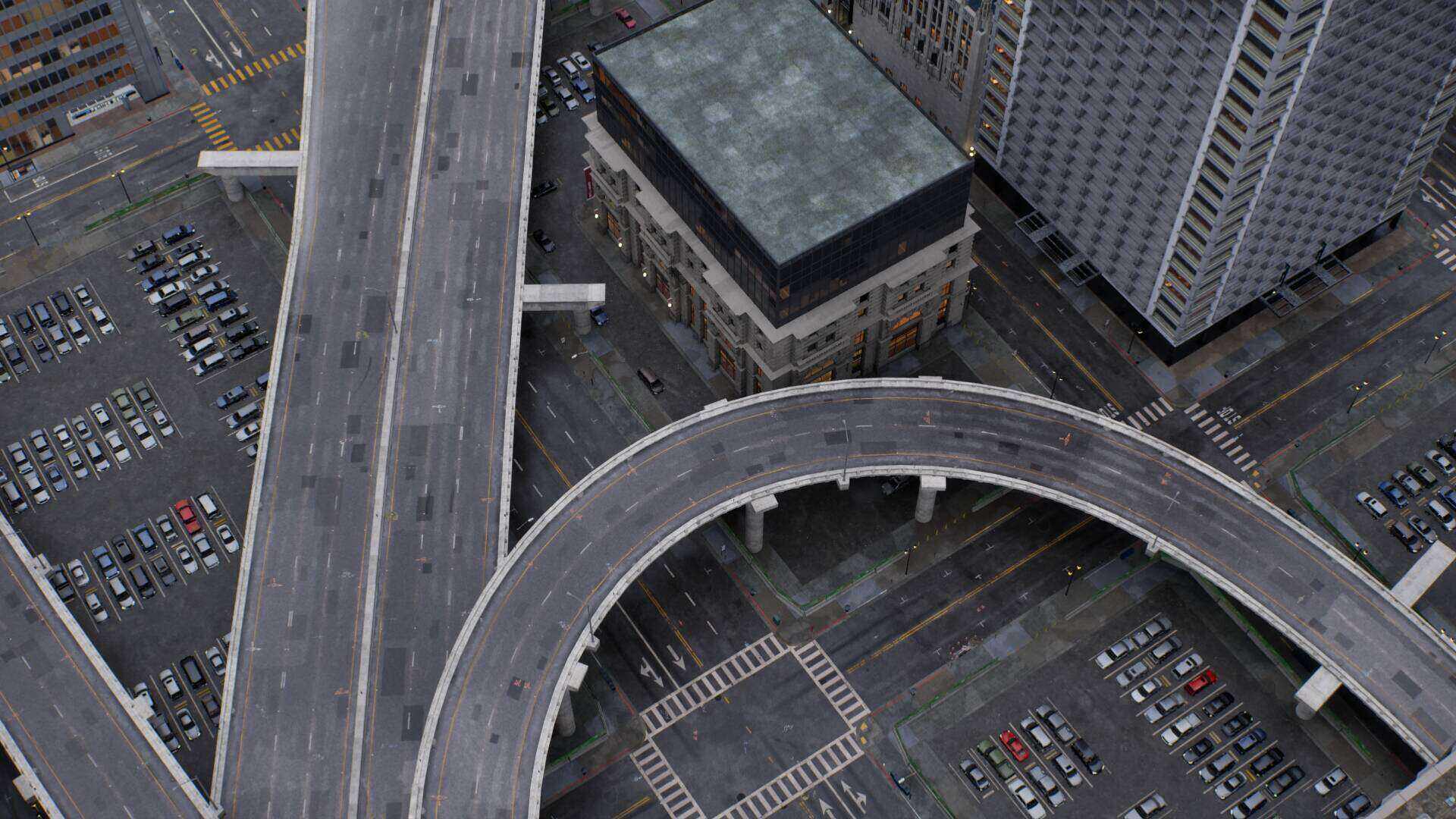} &
        \imagecell[0.24]{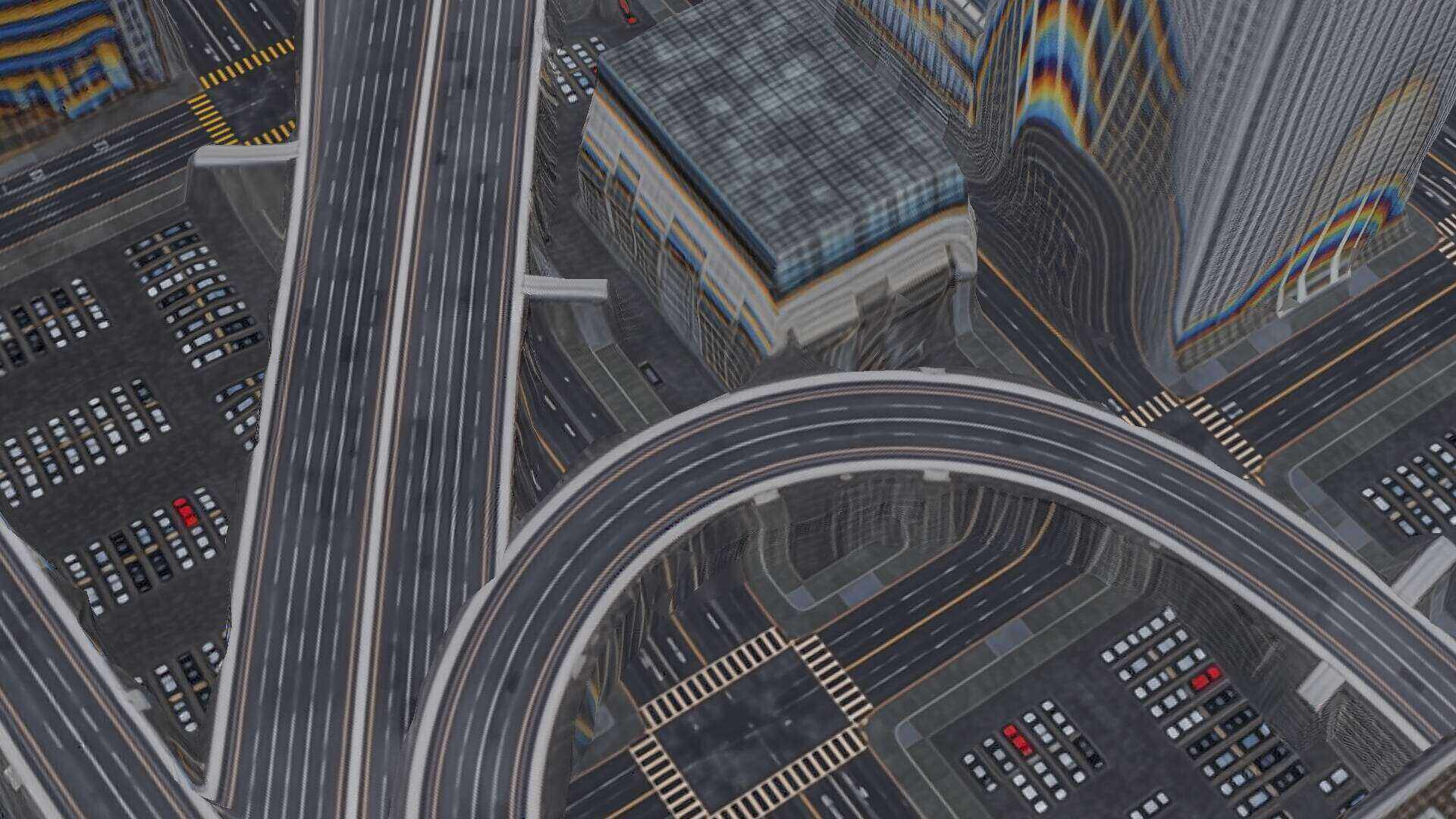} \\
        \vspace{-6pt} \\
        {G.T.} & {Ours} & {G.T.} & {Ours}
    \end{tabular}
    \caption{\textbf{Failure case on non-monotonic structures.}
    Non-monotonic scenes from the \textit{MatrixCity-Satellite} dataset where our 2.5D prior fails to capture the multi-layer geometry. We believe that future work could further alleviate these issues.}
    \label{fig:failure-bridge}
\end{figure}

\subsection{Failure Cases}
\label{sub-sec:supp:failure-cases}

As discussed in the Limitations section, our reliance on a 2.5D Z-monotonic geometric prior prevents our method from faithfully modeling non-monotonic structures such as bridges, overpasses, and other multi-layer configurations.
In these cases, the Z-monotonic assumption forces the geometry to collapse along the vertical axis, leading to missing underpasses.
\cref{fig:failure-bridge} shows two representative examples from the \textit{MatrixCity-Satellite} dataset, where the ground-truth geometry exhibits multiple vertical layers, while our reconstruction either merges them into a single surface or produces incomplete structures.
We believe that future work could further alleviate these issues, for example by augmenting the 2.5D scaffold with local full-3D representations or topology-aware modules that handle multi-layer and non-monotonic structures, thereby making the framework more flexible in such challenging cases.

\subsection{Additional Comparison with Mesh-Recovery Baselines}
\label{sub-sec:supp:mesh-baselines}

\begin{figure}[htbp]
    \centering
    \begin{spacing}{1.0}
    \setlength\tabcolsep{3pt}
    \begin{tabular}{ccccccc}
    \toprule
    \textbf{Method} & NKSR \tiny{({\color{cyan}PC})} & {\small{LightweightMR}} \tiny{$\binom{\text{{\color{cyan}PC}}}{\text{{\color{teal}SDF}}}$} & NDC \tiny{({\color{magenta}Vox})} & ODC \tiny{({\color{magenta}Vox})} & Ours   \\
    \midrule
    \textbf{CD $\downarrow$} & 0.0822 & 0.1461 & 0.0499 & 0.0633 & \textbf{0.0357}  \\
    \bottomrule
    \vspace{10pt}
    \end{tabular}
    \setlength\tabcolsep{1pt}
    \begin{tabular}{ccccc}
    \imagecell[0.19]{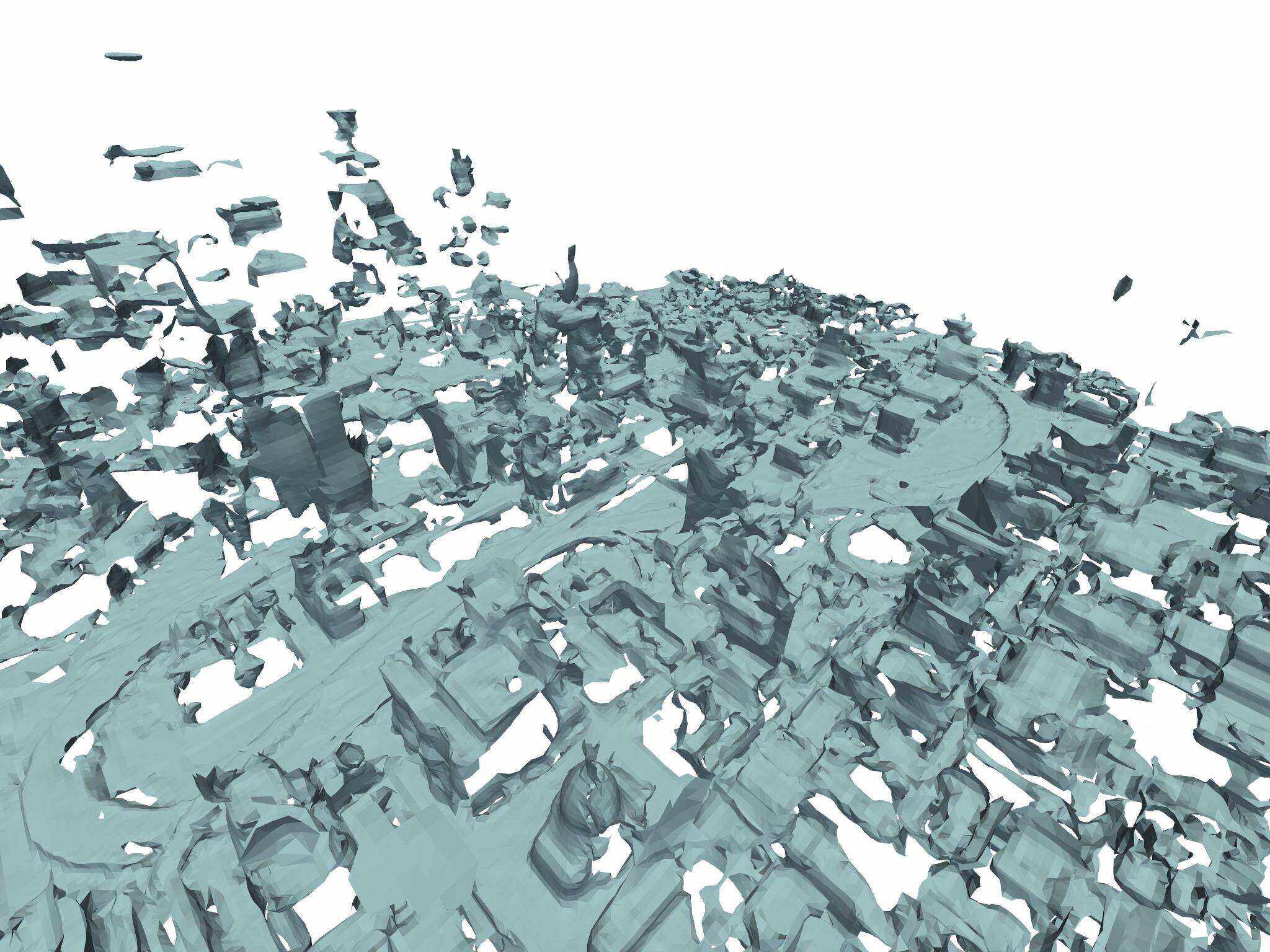} & 
    \imagecell[0.19]{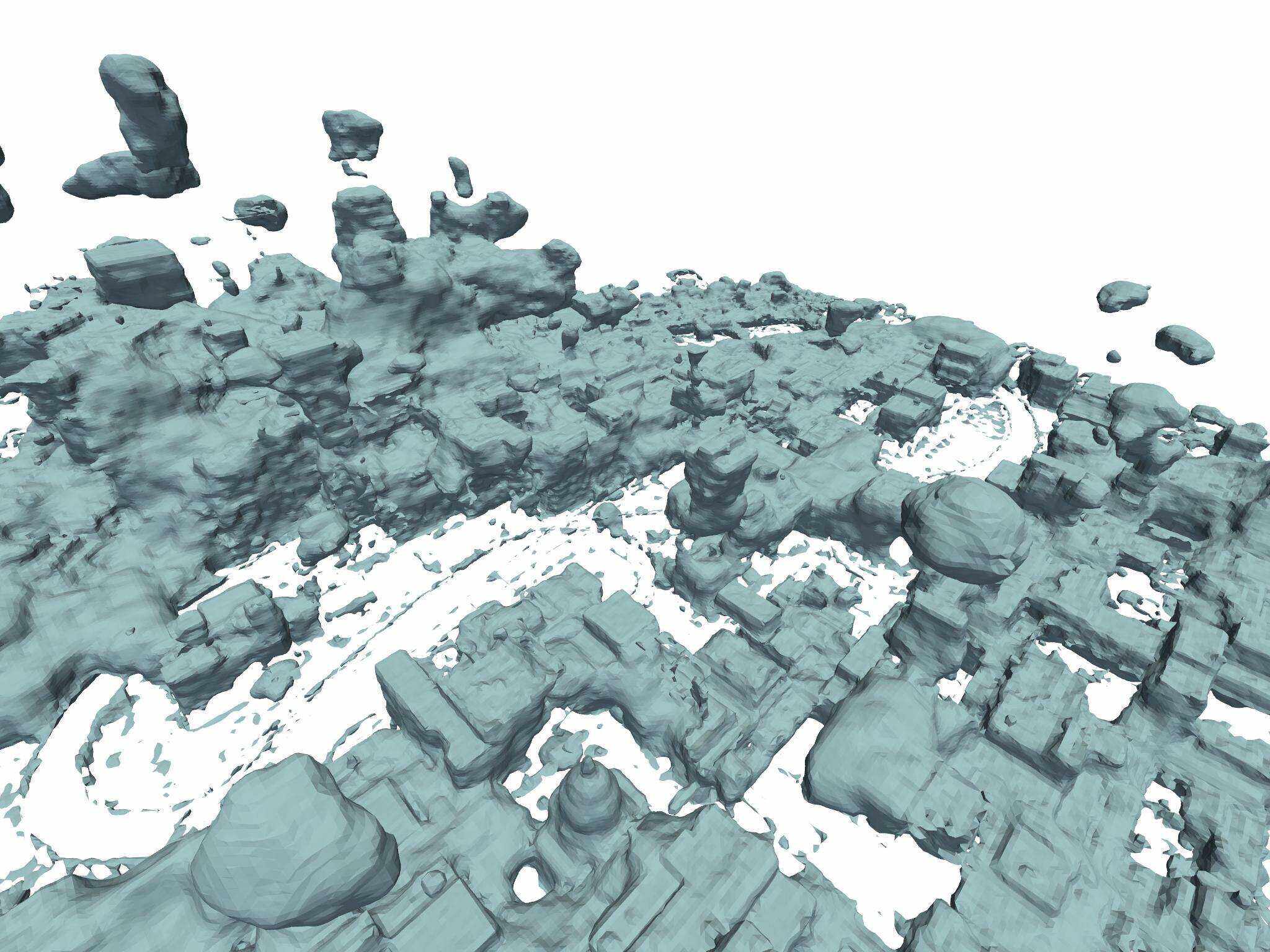} & 
    \imagecell[0.19]{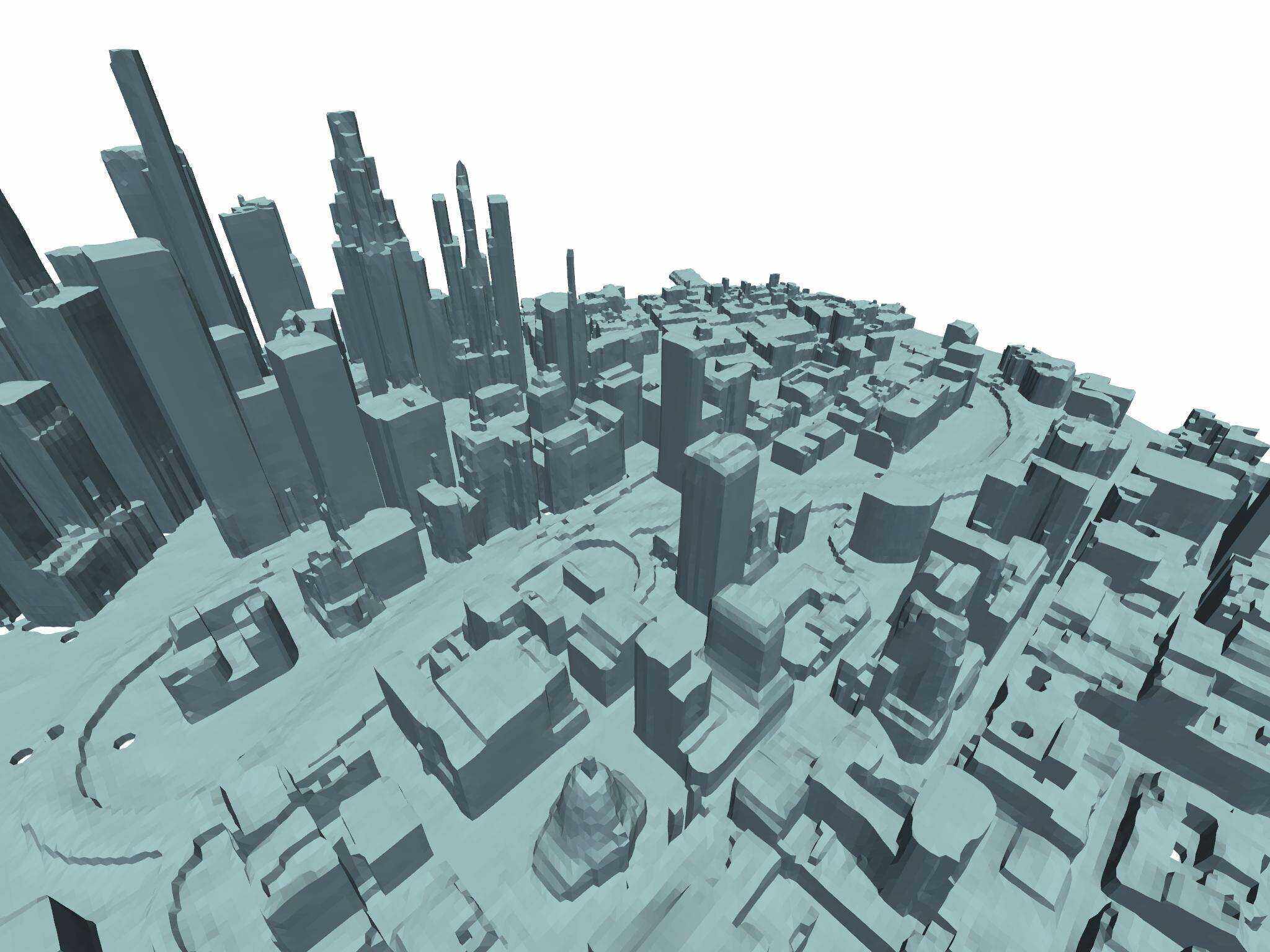} & 
    \imagecell[0.19]{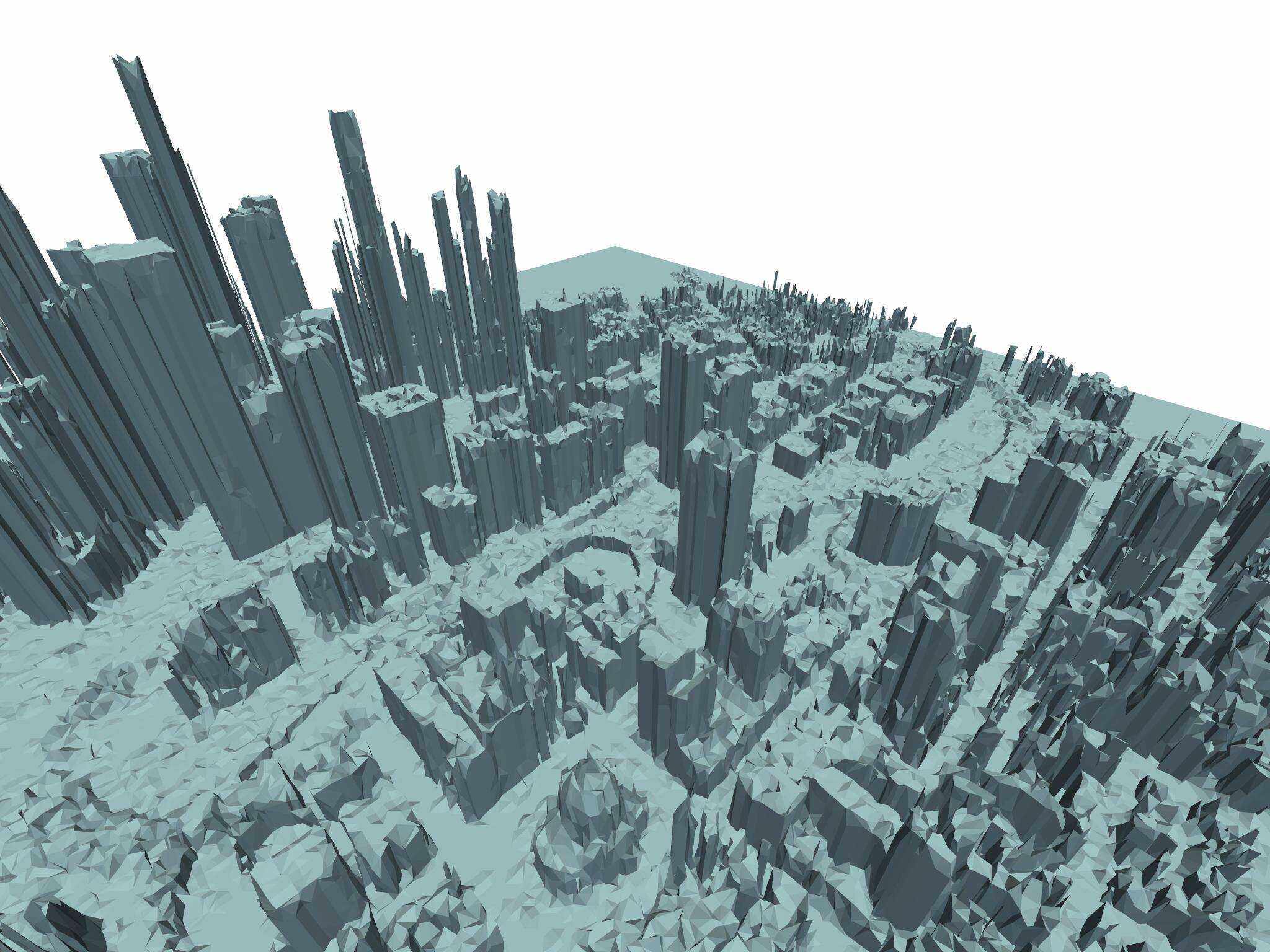} & 
    \imagecell[0.19]{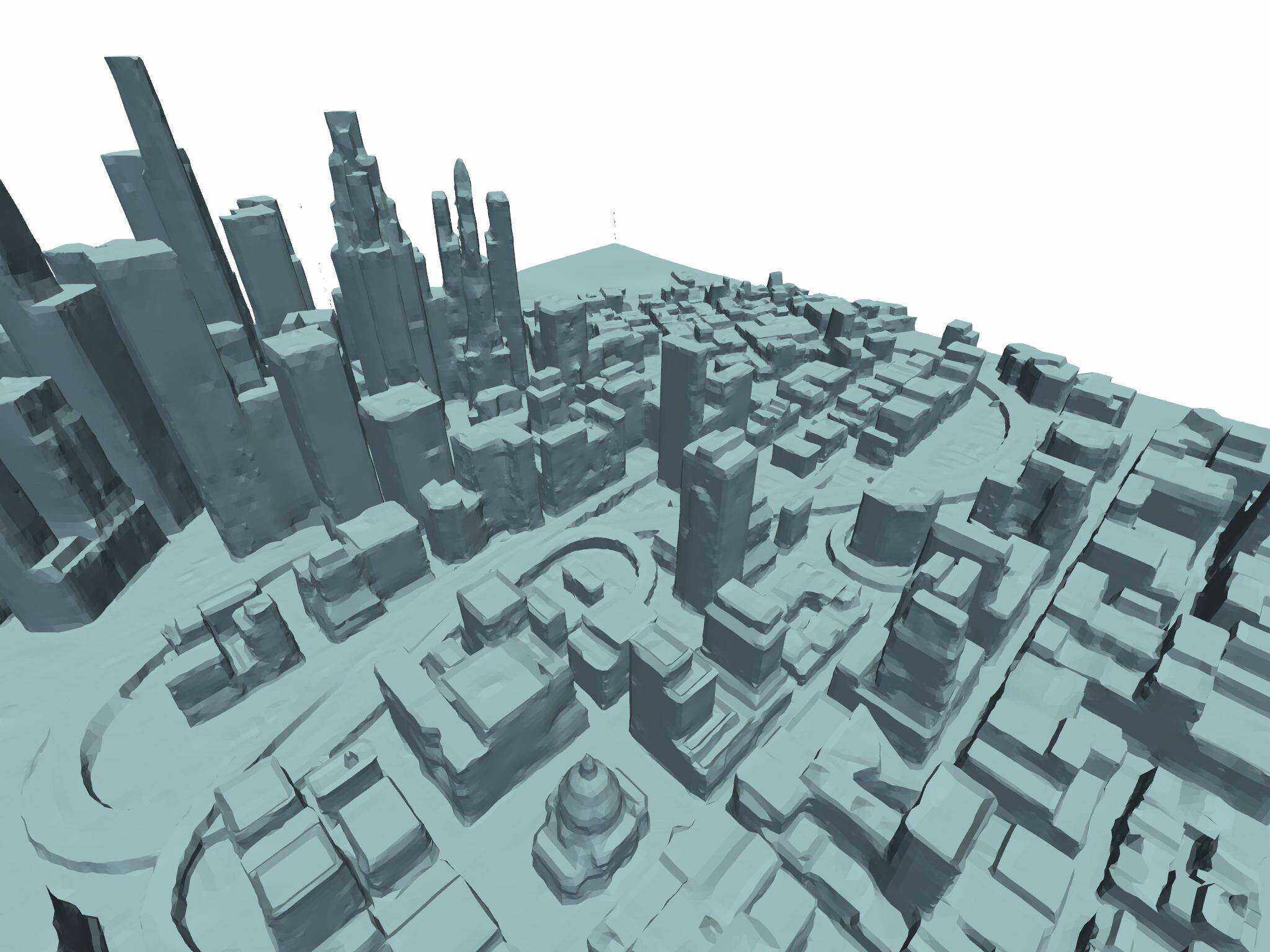} \\
    NKSR \tiny{({\color{cyan}PC})} & {\small{L. MR}} \tiny{$\binom{\text{{\color{cyan}PC}}}{\text{{\color{teal}SDF}}}$} & NDC \tiny{({\color{magenta}Vox})} & ODC \tiny{({\color{magenta}Vox})} & Ours
    \end{tabular}
    \end{spacing}
    \caption{Mesh-recovery baselines on \textit{MatrixCity-Satellite} dataset with same input. 
    Due to extremely limited wall parallax, MVS points are sparse or empty on walls, making {\color{cyan}(PC) point} or {\color{teal}(SDF) SDF}-based methods incomplete.
    {\color{magenta}(Vox) voxel}-based methods face a resolution-sparsity trade-off, often producing holes or artifacts.}
    \label{fig:supp:mesh-baselines}
\end{figure}

To better position our geometry module with respect to mesh-recovery approaches, we further compare against representative mesh-recovery baselines, including NKSR~\cite{huang2023neural}, LightweightMR~\cite{zhang2025high}, NDC~\cite{chen2022neural}, and ODC~\cite{hwang2024occupancy}.
\cref{fig:supp:mesh-baselines} shows that generic mesh-recovery methods remain challenged in this setting because the input MVS observations are dense on roofs and ground but sparse or empty on building facades under extreme off-nadir views. 
As a result, point-based and SDF-based approaches often produce incomplete wall geometry, while voxel-based methods face a resolution -- sparsity trade-off and may introduce holes or discretization artifacts. 
In contrast, our Z-Monotonic SDF imposes the structural prior directly at the representation level, yielding cleaner watertight meshes with sharper vertical facades.

\subsection{Ground-View Locations}
\label{sub-sec:supp:viewpoints}

\begin{figure}[htbp]
    \centering
    \begin{spacing}{1.0}
    \setlength\tabcolsep{1pt}
    \begin{tabular}{cccc}
    \CroppedImageWithStarMC{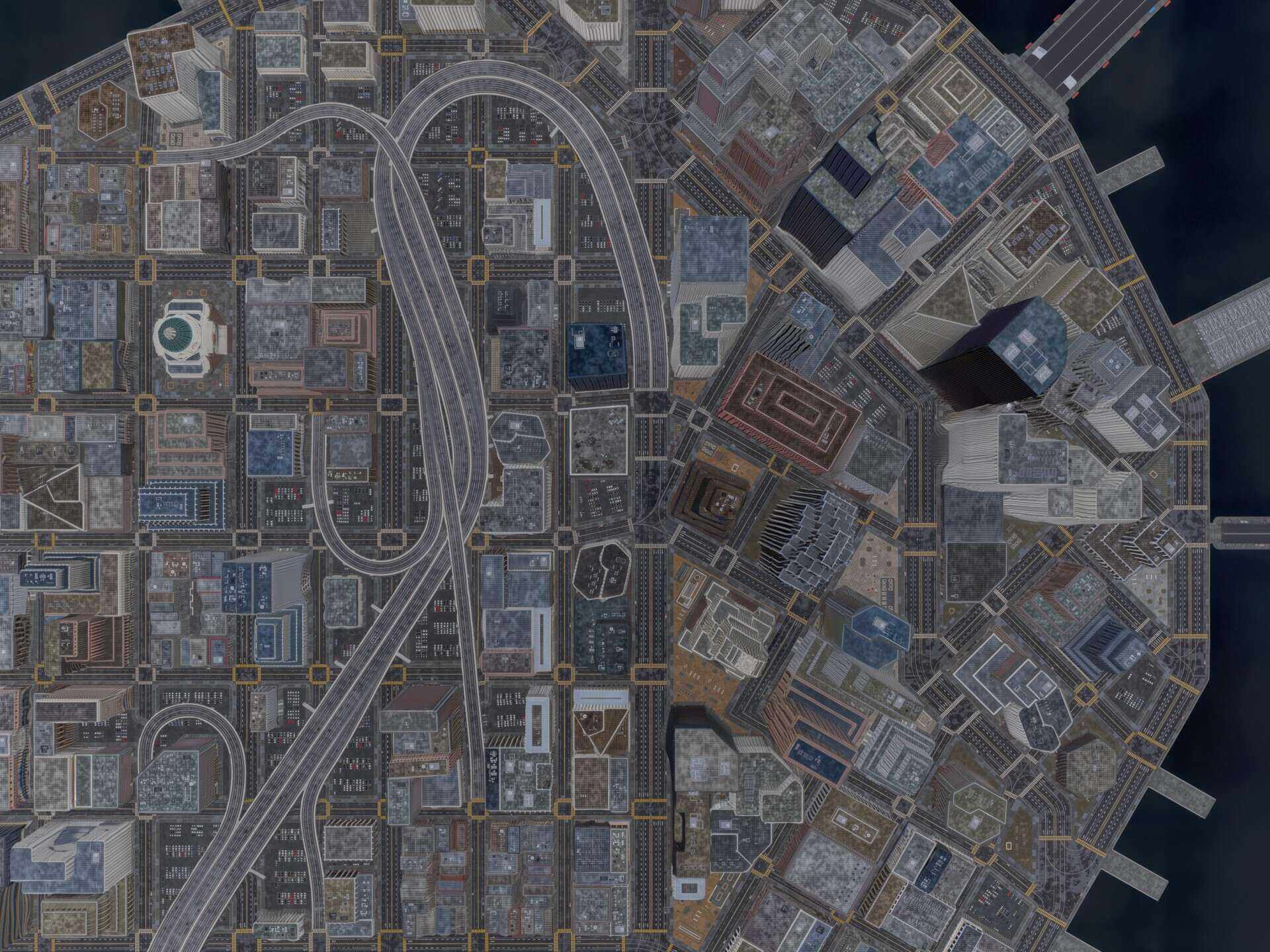}{0.23\linewidth}{0.155}{0.05} &
    \CroppedImageWithStarJAX{figures/qualitative/input/JAX_068.jpg}{0.23\linewidth}{0.03}{0.42} &
    \CroppedImageWithStarNYC{figures/qualitative/input/NYC_336.jpg}{0.23\linewidth}{0.52}{0.27} &
    \CroppedImageWithStarWJ{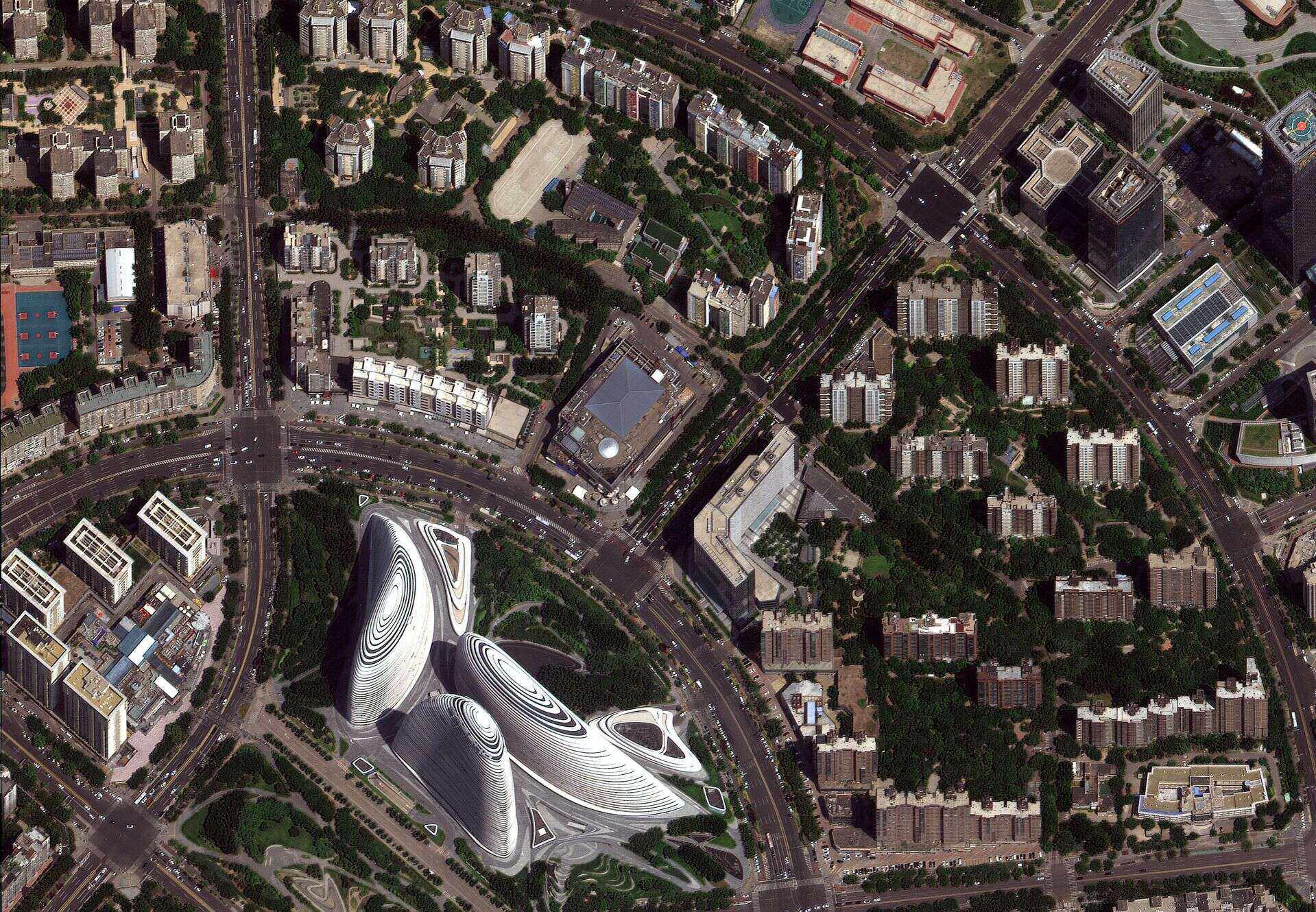}{0.23\linewidth}{0.6}{0.05}
    \end{tabular}
    \end{spacing}
    \caption{$\color{red}{\star}$ denotes viewpoints of  Fig.~6 results in the main paper. 
    }
    \label{fig:supp:ill-posed-problem}
\end{figure}

To clarify the camera placement of the qualitative results in \cref{fig:qualitative}, we mark the corresponding ground-view locations in \cref{fig:supp:ill-posed-problem}. The starred markers indicate the viewpoints used for rendering the close-range results.

\subsection{Real-Scene Geometry Evaluation}
\label{sub-sec:supp:real-geo}

For real-scene geometry evaluation, dense ground-truth geometry is unavailable for the full Urban Scene. 
We therefore further collect LiDAR data for a partially overlapping region and report a one-sided Chamfer distance from LiDAR to mesh after alignment on the overlap region. 
As shown in \cref{tab:supp:eval-geo-real}, our method achieves the lowest error among all baselines.

\begin{table}[htbp]
\centering
\setlength\tabcolsep{3pt}
\begin{tabular}{cccccc}
\toprule
\textbf{Method} & 2DGS & Mip-Splatting & CityGS-X & Skyfall-GS & Ours   \\
\midrule
\textbf{CD $\downarrow$} & 33.064 & 5.785 & FAIL & 30.345 & \textbf{3.825}  \\
\bottomrule
\end{tabular}
\caption{
    One-sided Chamfer Distance (LiDAR $\rightarrow$ mesh) on \emph{Urban Scene} after alignment on the overlap region. The lower the better.}
\label{tab:supp:eval-geo-real}
\end{table}

\subsection{UV Atlas Visualization}
\label{sub-sec:supp:uv}

We further provide a visualization of the UV atlas used in our appearance stage, as well as an additional ablation on the number of input satellite views. 
For UV parameterization, we use the UV atlasing tool in Open3D on the final merged mesh and keep the atlas fixed during texture optimization. 
\cref{fig:supp:uv-sparsity} (a) visualizes a representative UV map on the Urban Scene.

\subsection{Input Sparsity Ablation}
\label{sub-sec:supp:sparsity}

\begin{figure*}[htbp]
    \centering
    \begin{spacing}{1.0}
    \setlength\tabcolsep{1pt}
    \begin{tabular}{cccc}
    \imagecell[0.19]{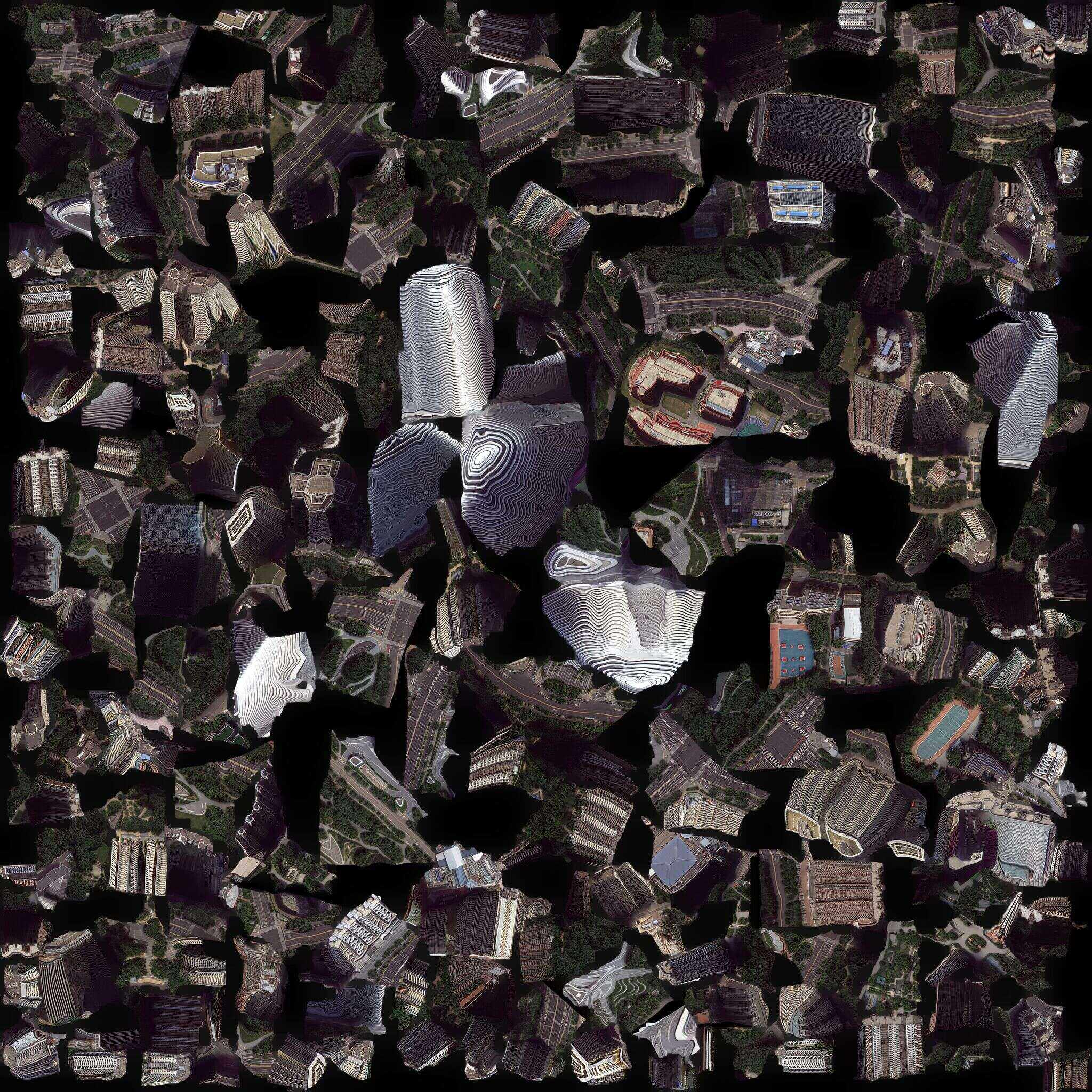} & 
    \imagecell[0.25]{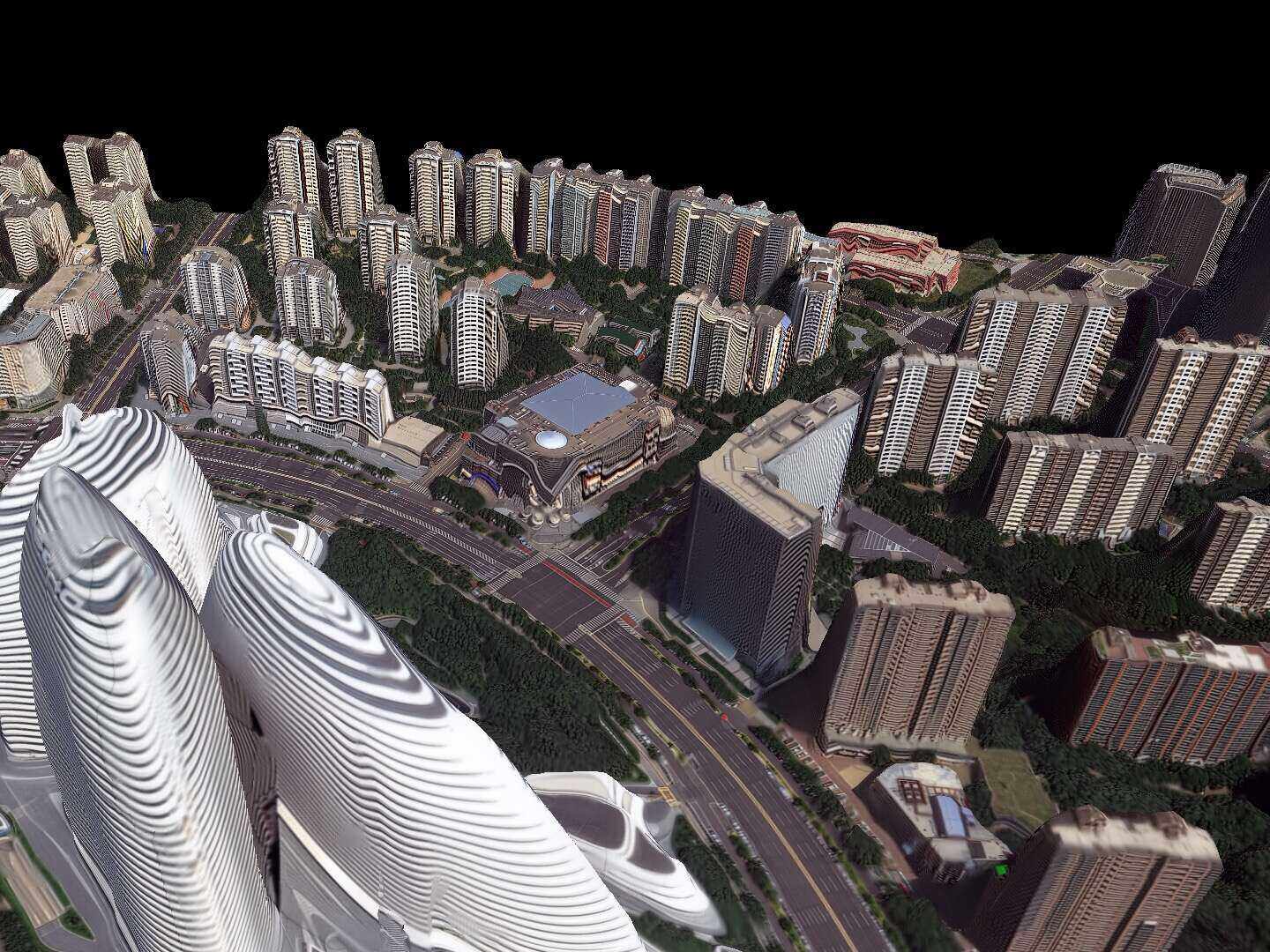} & 
    \imagecell[0.25]{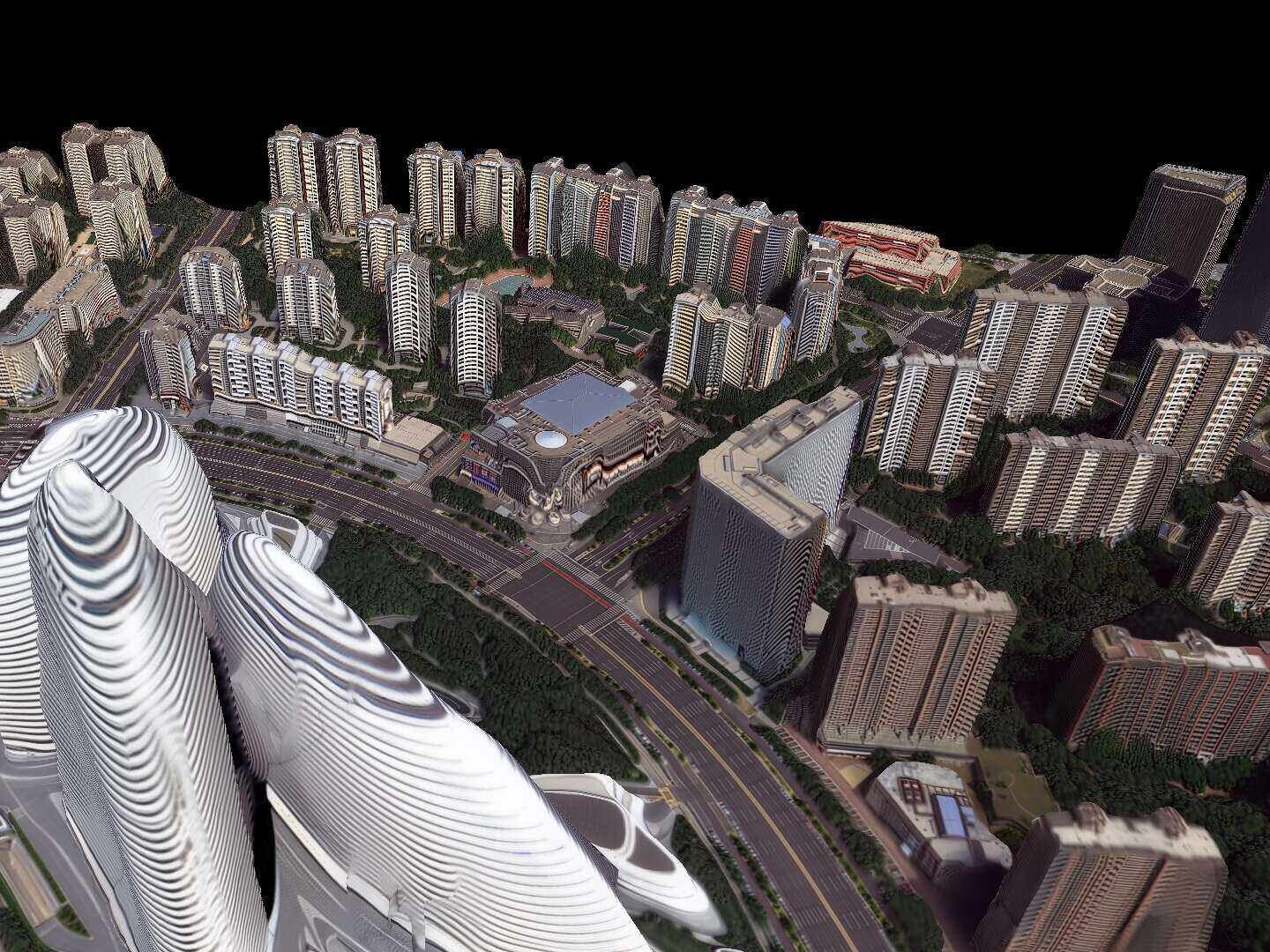} & 
    \imagecell[0.25]{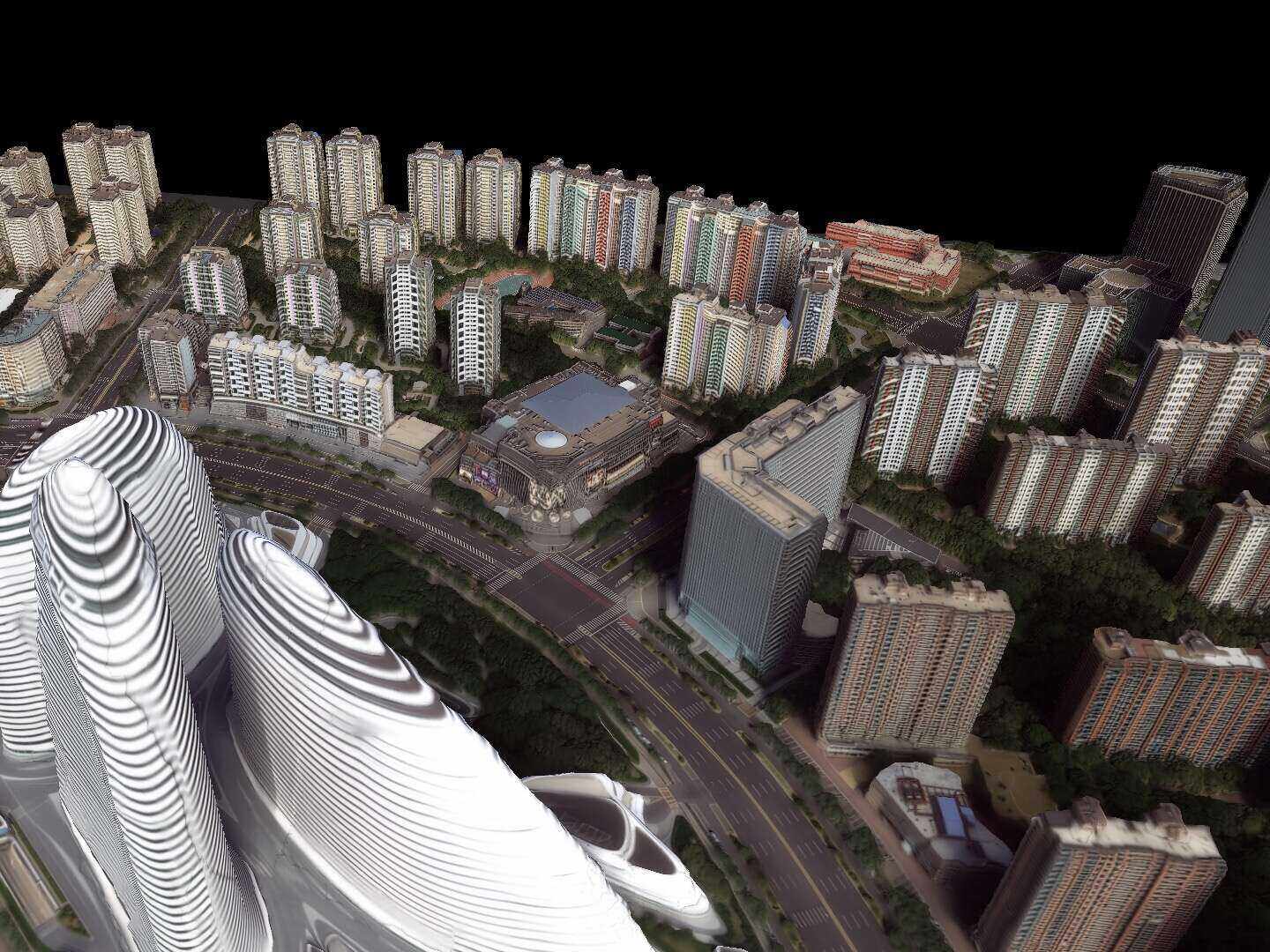} \\
    (a) {\small{UV Map}} & (b) 3 Views & (c) 5 Views & (d) 11 Views
    \end{tabular}
    \end{spacing}
    \caption{(a) A UV map of results on the Urban Scene. (b) -- (d) Qualitative ablation on input sparsity using 3, 5, and 11 input satellite images, respectively. The quality of our results remains comparable with fewer views.}
    \label{fig:supp:uv-sparsity}
\end{figure*}

To evaluate robustness to input sparsity, we reconstruct the same real Urban Scene using 3, 5, and 11 input satellite images sampled from the same capture set. 
As shown in \cref{fig:supp:uv-sparsity} (b) -- (d), our method maintains comparable geometry and visual quality even with fewer views, indicating that the geometric prior effectively stabilizes optimization in sparse-view regimes.

\subsection{Additional Clarifications}
\label{sec:supp:clarifications}

\paragraph{Why the problem is ill-posed.}
Under extreme off-nadir satellite imaging, vertical facades exhibit very limited parallax and are heavily foreshortened. 
Consequently, different 3D wall geometries may induce nearly indistinguishable observations in the input views. 
The inverse mapping from sparse satellite images to full 3D city geometry is therefore non-unique, especially on facades where MVS points are sparse or absent. 
Our Z-Monotonic prior reduces this ambiguity by restricting the solution space to 2.5D urban surfaces.

\paragraph{Training strategy of the restoration network.}
The restoration network $D$ is trained once on a large paired dataset of urban renderings and is then reused across all benchmarks without per-dataset retraining or test-scene finetuning.

\paragraph{Optimization details of iterative texture refinement.}
At each refinement iteration, we sample pseudo views from a regular grid of simulated UAV poses over the scene bounding box. 
These poses are synthetically defined rather than obtained from ground-truth UAV trajectories.
The rendered views are passed through $D$ to obtain restored targets, which are then used to supervise the texture optimization. 
We randomly shuffle the sampled views within each iteration and empirically observe negligible sensitivity to the optimization order.

\paragraph{Snow simulation.}
The snowy results in \cref{fig:app-sim} are produced using Blender on top of our reconstructed mesh. 
This application demonstrates that the mesh output of our method can serve as an application-ready asset for downstream urban simulation.

\end{document}